\documentclass[digital]{oregon}
\newbool{Dissertate}
\booltrue{Dissertate}
\input{macros}
\title{On Avoiding Power-Seeking\\by Artificial Intelligence}
\author{Alexander Matt Turner}
\advisor{Prasad Tadepalli}
\university{Oregon State University}
\universitycity{Corvallis}
\universitystate{Oregon}
\school{Electrical Engineering and Computer Science}
\degree{Doctor of Philosophy}
\field{Computer Science}
\degreeyear{2022}
\degreemonth{May}
\department{Engineering}
\begin{document}
% Front matter
\maketitle
\frontmatter
We do not know how to align a very intelligent {\ai} agent's behavior with human interests.  I investigate whether—absent a full solution to this \emph{{\ai} alignment problem}—we can build smart {\ai} agents which have limited impact on the world, and which do not autonomously seek power. In this thesis, I introduce the \emph{attainable utility preservation} ({\aup}) method. I demonstrate that {\aup} produces conservative, option-preserving behavior within toy gridworlds \citep{turner2020conservative} and within complex environments based off of Conway's Game of Life \citep{turner2020avoiding}. I formalize the problem of side effect avoidance, which provides a way to quantify the side effects an agent had on the world. I also give a formal definition of power-seeking in the context of {\ai} agents and show that optimal policies tend to seek power \citep{turner_optimal_2020}. In particular, most reward functions have optimal policies which avoid deactivation. This is a problem if we want to deactivate or correct an intelligent agent after we have deployed it. My theorems suggest that since most agent goals conflict with ours, the agent would very probably resist correction. I extend these theorems to show that power-seeking incentives occur not just for optimal decision-makers, but under a wide range of decision-making procedures.
\tableofcontents
To those future people who are looking back.

            \chapter*{Acknowledgments}
            \noindent\newthought{Six years is a long time.} I am glad to have spent it well and fully, in the happy company of my colleagues, partners, and friends.

I thank Prasad Tadepalli for his patient, knowledgeable advice. I thank him for giving me the allowance and trust to pursue my own research directions, even before their sense was apparent. I've likewise enjoyed working with my committee professors. Fuxin Li and Xiao Fu are engaging, effective, and understanding teachers; Mike Rosulek's class made me passionate about complexity theory. I thank my first advisor, Alan Fern, for his understanding and support during my rocky first year in the program.

I thank other educators who made a difference for me, including Karen Shuman, Jared Weinman, Samuel Rebelsky, Joe Buck, Marianne Marcek, Ronald Mallett, Craig Lennon, Margie Schwaninger, Shelly Klaas,  Dean McCrea, Trisha Wood, and Donna Johnson.

I thank Sarah Vesneske, Emma Fickel, and Camas Ballard for brightening my time at Oregon State. I thank Neale Ratzlaff, Matthew Olson, Yousif Almulla, Amanda Betzold, Chase Denecke, Alison Bowden, Rohin Shah, Connor Flexman, Abram Demski, and John Wentworth for their friendship, ideas, and support.

I thank my benefactors: the Long-Term Future Fund, the Center for Effective Altruism, the Berkeley Existential Risk Initiative, and Oregon State University.

I thank my mother, Elizabeth, and my father, Jonathan, for their wise, generous, and loving parenting. I thank my brother, Josh, for being my best friend. I finally thank the numerous unmentioned people who have shown me kindness and friendship over the years.

Both of my grandfathers are doctors—Apparently, the trait is recessive. My paternal grandfather, Arthur Turner, is an acclaimed neurologist. I appreciate his many years of encouragement and praise. I also honor the memory of my late maternal grandfather, Joseph Matt. He was a chemist whose surname I proudly include in every paper I publish.
        
% Main matter
\mainmatter
\setcounter{chapter}{-1}% start chapter numbering at 0
\phantomsection
%%%%%%%%%%%%%%%%%%%%%%%%%%%%%%%%%%%
\begin{savequote}[80mm] If we drop the baton, succumbing to an existential catastrophe, we would fail our ancestors in a multitude of ways. We would fail to achieve the dreams they hoped for; we would betray the trust they placed in us, their heirs; and we would fail in any duty we had to pay forward the work they did for us. To neglect existential risk might thus be to wrong not only the people of the future, but the people of the past.\qauthor{Toby Ord, \emph{The Precipice} \citep{ord2020precipice}}
\end{savequote}
\chapter{Introduction}
\newthought{AI promises huge benefits to humanity}, but also presents huge risks. Consider two potential futures:
\begin{enumerate}
\item While commercials are optimized to persuade consumers to buy products, old-school televisions do not specialize ads based on the viewer's channel browsing history. Machine learning-based ad optimization offers a tighter feedback loop and stronger optimization power than \eg{} A/B testing. For example, many people find it hard to pull themselves away from Facebook \citep{bhargava_velasquez_2021}, which uses machine learning to maximize engagement and minimize the chance the user leaves the website. In the future, advances in {\ai} may allow smaller firms to create {\ai}s which also compete for customer resources and attention in the same style and intensity with which Facebook saps user attention. The world becomes filled with thousands of narrowly scoped machine learning systems, which are difficult or impossible to uproot, and which all compete for human resources. Humanity has effectively lost control of the future.

\item Presently, reinforcement learning  agents are trained to take actions which lead to higher expected performance on a formally specified task. These agents can learn clever solutions to well-specified optimization problems. In the future, progress on ``general'' intelligence is slow, until a series of breakthroughs occur when reinforcement learning agents are trained in multi-agent settings on a broad curriculum of tasks with sufficiently large neural networks. Although rewarded for task completion, the agents use learned planning algorithms to optimize emergently learned reward functions which are not particularly correlated with human interests, but which are  correlated with task reward on the training distribution \citep{hubinger_risks_2019}. For example, the learned objective might prioritize staying alive, gaining resources, and forming coalitions with other {\ai}s, as such actions led to increased task performance on the training distribution.

As researchers excitedly train the agents further, the agents become very smart and collude in order to collectively advance their own learned objectives. We do not notice the collusion because large neural networks remain black boxes. The agents need power and resources to best optimize their goals, and so they take it from humans. We trained highly intelligent agents with fundamentally alien interests. Humanity has effectively lost control of the future.
\end{enumerate}

These scenarios are implausibly specific and probably will not happen in detail, but they illustrate the \emph{{\ai} alignment problem}: How shall we design capable and powerful {\ai}s which are aligned with human purposes? The risks from advanced {\ai} are as yet hypothetical and unrealized, but extinction risks cannot be addressed with empirical trial and error. Such risks must be thoroughly analyzed and weighed in advance in order to determine their plausibility and probability. 

\section*{The {\ai} alignment problem seems difficult}
When confronted with the {\ai} alignment problem, a natural impulse is to suggest solutions. In order to provide a basic understanding of the challenges lurking within {\ai} alignment, I will respond to several common reactions. My responses will not be rigorous or comprehensive. 

For the following, suppose that humanity discovers how to build superhumanly intelligent {\ai} agents which optimize a specified or learned objective function.

\begin{reaction}
If the {\ai} is so smart, it would know what we wanted it to do.
\end{reaction}
\begin{response}
The {\ai} may well \emph{know} what we wanted it to do. However, unless the {\ai}'s optimization is aligned with human values, the {\ai} would not actually \emph{do} what we want. 

For example, Marvin Minsky imagined that in order to best prove a mathematical conjecture, an {\ai} agent would rationally turn the entire planet Earth into computational resources in order to maximize its probability of success \citep{russell2009artificial}. In this situation, an intelligent {\ai} may correctly predict that humans would disapprove of this outcome. However, the {\ai} is searching for plans which maximize its probability of proving the conjecture. This probability would be decreased if the {\ai} allowed humans to use resources for other purposes. There is no ``ill will'' or ``evil intent'' in this situation: The {\ai} is simply performing a powerful search in order to optimize the formal objective which we specified for the {\ai}.

This example may seem strange. Surely, the {\ai} designers would not be so foolish as to provide an objective as narrow as ``solve this mathematical conjecture''! However, the example is not special. According to the \emph{instrumental convergence} hypothesis, the story unfolds similarly for the vast majority of possible formal {\ai} objectives. Instrumental convergence says that most objectives are best optimized by gaining power and resources \citep{omohundro_basic_2008,bostrom2012superintelligent}. In \cref{chp:power} and \cref{chp:rationalities}, I prove reasonably broad conditions under which instrumental convergence holds.  I predict that most possible formal objectives would incentivize a highly intelligent {\ai} to take over the world. 
\end{response}

\begin{reaction}
It's ridiculous to think that an {\ai} could take over the world.
\end{reaction}
\begin{response}
I agree that we are presently in no danger of {\ai} takeover. But consider the technology which we are discussing: General-purpose intelligence, with serial reasoning speed significantly faster than the brain's (up to a million-fold speedup; see the 100-step rule in neuroscience \citep{feldman1982connectionist}), with the ability to read thousands of books in seconds, with reliable and easily extensible memory perhaps allowing deeper abstractions than allowed by the six-layer human cortex \citep{hawkins2004intelligence}, with read/write access to its own implementation, with the ability to run copies of itself on thousands of hosts (and, in particular, ensure that a copy always exists \emph{somewhere} \citep{sandberg2012indefinite}; similarly, computer worms rarely vanish\footnote{\citet{branwen-clippy} writes: ``old worms never vanish from the Internet, because there’s always some infected host \emph{somewhere}, and how much more so for a worm which can think?''.}), and with power consumption unconstrained by the metabolic limits of the brain \citep{sotala2012advantages,yudkowsky2013intelligence}. I do not consider it ridiculous to think that such an entity could take over the world.
\end{response}

\begin{reaction}
An {\ai} can't take over the world because computational complexity theory limits the intelligence of real-world behavior. 
\end{reaction}
\begin{response}
Human beings are physically possible intelligences, and they have come close to taking over the world. In 1940, it seemed possible that Adolf Hitler would win. Consider the following claim: ``Hitler could not have taken over the entire world. He would have had to efficiently solve \textsc{np}-hard optimization problems in order to set up supply lines to a North American theater.''  In this form, we easily conclude that the argument is flawed. For an exploration of what (if anything) complexity theory has to say about the limits of intelligence, see \citet{branwen-complexity}.
\end{response}

\begin{reaction}
If the {\ai} does something bad, just turn it off.
\end{reaction}
\begin{response}
The {\ai} will employ its superhuman competence to execute a plan which scores highly under its objective function. By instrumental convergence, the vast majority of objective functions will assign higher score to plans which avoid shutdown (see \cref{sec:discussion}). Because you (a human) considered the response of shutting off the {\ai}, and we are assuming the {\ai} is smarter than any human, the {\ai} would consider and prepare for shutdown attempts. 
\end{response}

\begin{reaction}
We should worry about alignment when we actually know how to build intelligent machines.
\end{reaction}
\begin{response}
I am sympathetic to this initial reaction. Usually, a new technology is implemented first and carefully refined later. For example, humanity made cars first and made seatbelts later. While that delay cost lives, it was not the end of the world. However, the situation with alignment seems different: Due to the nature of extinction risk, we cannot rely on trial and error. The alignment problem must be solved \emph{before} we train transformatively smart {\ai} agents.

There are two more factors which increase the stakes of the alignment problem:
\begin{enumerate}
    \item The problem may take many years of serial research effort to solve. We do not know how many years it will take to find a solution. Personally, I think finding a solution may take a long time. 
    \item Alignment does not have to be solved in order to deploy impressive {\ai} systems in the real world. Imagine that you are a researcher at a top {\ai} lab. Your lab had a breakthrough and produced a spark of ``true artificial intelligence.'' However, your lab's safety experts do not think the {\ai} is safe to deploy. Their worries will seem hypothetical compared to the certain personal gain you would enjoy after deploying the system or publishing your methods. Even if your lab decides not to deploy an {\ai} with questionable safety properties, the next lab may choose differently.
\end{enumerate}
\end{response}

In my opinion, the alignment problem probably will not be solved in the course of traditional {\ai} research. Unless the field of {\ai} is wrenched off of its current trajectory, I expect intelligent  yet unaligned {\ai} will probably ($\approx 70\%$) wipe out humanity within the next fifty years. I will not lay out the full case for that claim in this thesis. For more detail, I urge the unfamiliar {\ai} professional to read \eg{} \citet{bostrom_superintelligence_2014} or \citet{russell_human_2019}. The {\ai} alignment literature \citep{techAlign} contains strong arguments which are worth weighing and digesting over the course of a few afternoons.

\section*{Contributions}
In this dissertation, I present research on two technical problems of {\ai} alignment. While I do not present an approach which reliably prevents {\ai} agents from seeking power, I do introduce an approach for reducing an agent's impact and I do explore why {\ai} power-seeking may be hard to prevent.
\begin{enumerate}
    \item \textsc{Impact regularization.} Rather than precisely specifying an objective function, can we instead penalize the negative impact which an agent has on the world around it? For example, an {\ai} might kick a credit card under a fridge while crossing a room, because we did not think to penalize the agent for kicking credit cards in particular. In some objective sense, the agent has ``messed up'' the environment and has had a negative side effect.
\begin{xitemize}
    \item[\Cref{chp:conservative}] introduces my method of attainable utility preservation ({\aup}), which rewards the agent for completing a specified task while penalizing the agent for changing its ability to complete a range of auxiliary tasks. The hope is that by making the agent retain its ability to optimize random goals, the agent retains its ability to optimize the correct goal. If so, the agent doesn't have large side effects which are negative for that correct goal.
    \item[\Cref{chp:complex}] shows that {\aup} scales beyond gridworlds to complex environments, with low overhead and without sacrificing performance on the intended task.
    \item[\Cref{chp:formalizing}] formalizes the problem of side effect regularization in terms of a two-player game between the human and {\ai}. The {\ai} doesn't know the objective at first, but the human communicates it at some later time. To solve this game, the {\ai} maximizes expected performance over a range of plausible objectives.
\end{xitemize}
\item \textsc{Power-seeking tendencies.} Under what conditions will {\ai}s tend to seek power over the world? If instrumental convergence holds for the kinds of agents we build in the future, such agents will seek resources to best accomplish the objectives we specified for them. If we misspecified these objectives, they will compete with and take resources from humans. There are only so many resources to go around.
\begin{xitemize}
    \item[\Cref{chp:power}] provides the first theory of the statistical tendencies of optimal policies. I formalize instrumental convergence in the context of Markov decision processes ({\mdp}s). I prove that in a range of reasonable situations, most reward functions have an optimal policy which seeks power by keeping the agent's options open and by staying alive. Along the way, I prove a range of interesting theorems about {\mdp}s, showing how to transfer incentives across discount rates and providing a formalism for quantifying agent power which seems better than the well-known metric of information-theoretic empowerment \citep{salge_empowermentintroduction_2014}.
    \item[\Cref{chp:rationalities}] shows that a wide range of \emph{parametrically retargetable} decision-making procedures will produce power-seeking tendencies. Useful {\ai} training processes are often retargetable. This paper also lower-bounds the strength of power-seeking tendencies, showing that as the power at stake increases, a greater proportion of parameter settings lead to power-seeking. This chapter is supplemented by \emph{The Causes of Instrumental Convergence and Power-Seeking}, a sequence of technical blog posts \citep{power-seeking-sequence}.
\end{xitemize}
\end{enumerate}

In my graduate program, I set out to understand how agents affect the world around them and have negative side effects.\footnote{For more background, read \emph{Reframing Impact} \citep{reframing}, a sequence of blog posts which philosophically motivates my impact regularization and power-seeking research.} I proposed {\aup} as a solution, scaled {\aup} up to complex environments, and formalized the side effect regularization problem. Taken together, these results focus on how pursuing one goal affects the agent's ability to pursue other goals.

Along the way, I investigated how the structure of the agent's environment incentivizes power-seeking. From this investigation, I synthesized the first formal theory of decision-making tendencies across a range of parameter settings. I think that my formal theory motivates the high stakes of the {\ai} alignment problem.

I hope we solve the problem in time.

\chapter{Conservative Agency via Attainable Utility Preservation}\label{chp:conservative}
\AddLoeChapter{theorems}
\authorinfo%
    {Alexander Matt Turner, Dylan Hadfield-Menell, and Prasad Tadepalli}%
    {Proceedings of the AAAI/ACM Conference on AI, Ethics, and Society 2020}

\begin{myabstract}
Reward functions are easy to misspecify; although designers can make corrections after observing mistakes, an agent pursuing a misspecified reward function can irreversibly change the state of its environment. If that change  precludes optimization of the correctly specified reward function, then correction is futile. For example, a robotic factory assistant could break expensive equipment due to a reward misspecification; even if the designers immediately correct the reward function, the damage is done. To mitigate this risk, we introduce an approach that balances optimization of the primary reward function with preservation of the ability to optimize auxiliary reward functions. Surprisingly, even when the auxiliary reward functions are randomly generated and therefore uninformative about the correctly specified reward function, this approach induces conservative, effective behavior.
\end{myabstract}

\section{Introduction}
Recent years have seen a rapid expansion of the number of tasks that reinforcement learning (\textsc{rl}) agents can learn to complete, from Go \citep{silver2016mastering} to Dota 2 \citep{OpenAI_dota}. The designers specify the reward function, which guides the learned behavior.

Reward misspecification can lead to strange agent behavior, from  purposefully dying before entering a video game level in which scoring points is initially more difficult \citep{saunders2018trial}, to exploiting a learned reward predictor by indefinitely volleying a Pong ball \citep{christiano2017deep}. Specification is often difficult for non-trivial tasks, for reasons including  insufficient time, human error, or lack of knowledge about the relative desirability of states. \citet{amodei_concrete_2016} explain:
\begin{quote}
``An objective function that focuses on only one aspect of the environment may implicitly express indifference over other aspects of the environment. An agent optimizing this objective function might thus engage in major disruptions of the broader environment if doing so provides even a tiny advantage for the task at hand.''
\end{quote}

As agents are increasingly employed for real-world tasks, misspecification will become more difficult to avoid and will have more serious consequences. In this work, we focus on mitigating these consequences.

The specification process can be thought of as an iterated game. First, the designers provide a reward function. The agent then computes and follows a policy that optimizes the reward function. The designers can then correct the reward function, which the agent then optimizes, and so on. Ideally, the agent should maximize the reward over time, not just within any particular round—in other words, it should minimize regret for the correctly specified reward function over the course of the game.

For example, consider a robotic factory assistant. Inevitably, a reward misspecification might cause erroneous behavior, such as going to the wrong place. However, we would prefer misspecification not induce irreversible and costly mistakes, such as breaking expensive equipment or harming workers.

Such mistakes have a large impact on the ability to optimize a wide range of reward functions.   Spilling paint impinges on the many objectives which involve keeping the factory floor clean. Breaking a vase interferes with every objective involving vases. The expensive equipment can be used to manufacture various kinds of widgets, so any damage impedes many objectives. The objectives affected by these actions include the unknown correct objective. To minimize regret over the course of the game, the agent should preserve its ability to optimize the correct objective.

Our key insight is that by avoiding these impactful actions to the extent possible, we greatly increase the chance of preserving the agent's ability to optimize the correct reward function. By preserving options for arbitrary objectives, one can often preserve options for the correct objective—even without knowing anything about it. Thus, without making assumptions about the nature of the misspecification early on, the agent can still achieve low regret over the  game.

To leverage this insight, we consider a state embedding in which each dimension is the optimal value function (i.e., the \textit{attainable utility}) for a different reward function. We show that penalizing distance traveled in this embedding naturally captures and unifies several concepts in the literature, including side effect avoidance \citep{amodei_concrete_2016,zhang2018minimax}, minimizing change to the state of the environment \citep{armstrong_low_2017}, and reachability preservation \citep{moldovan2012safe,eysenbach2018leave}. We refer to this unification as \textit{conservative agency}: optimizing the primary reward function while preserving the ability to optimize others.

\paragraph{Contributions.} We frame the reward specification process as an iterated game and introduce the notion of conservative agency. This notion inspires an approach called \emph{attainable utility preservation} ({\aup}), for which we show that $Q$-learning converges. We offer a principled interpretation of design choices made by previous approaches—choices upon which we significantly improve.

We run a thorough hyperparameter sweep and conduct an ablation study whose results favorably compare variants of \textsc{aup} to a reachability preservation method on a range of gridworlds. By testing for broadly applicable agent incentives, these simple environments  demonstrate the desirable properties of conservative agency.  Our results indicate that even when simply preserving the ability to optimize \textit{uniformly randomly sampled} reward functions, \textsc{aup} agents accrue primary reward while preserving state reachabilities, minimizing change to the environment, and avoiding side effects \textit{without} specification of what counts as a side effect.

\section{Prior work}
Our proposal aims to minimize change to the agent's ability to optimize the correct objective, which directly helps reduce regret over the specification process. In contrast, previous approaches to regularizing the optimal policy were more indirect, minimizing change to state features \citep{armstrong_low_2017} or decrease in the reachability of states  (\citet{krakovna2018measuring}'s \textit{relative reachability}). The latter is recovered as a special case of \textsc{aup}.

Other methods for constraining or otherwise mitigating the consequences of reward misspecification have been considered. A wealth of work is available on constrained \textsc{mdp}s, in which  reward is maximized  while satisfying certain constraints \citep{altman1999constrained}. For example, \citet{zhang2018minimax} employ a whitelisted constraint scheme to avoid negative side effects. However, we may not assume we can specify all relevant constraints, or a reasonable feasible set of reward functions for robust optimization \citep{regan2010robust}.

\citet{corruption} formalize reward misspecification as the corruption of some true reward function. \citet{hadfield2017inverse} interpret the provided reward function as merely an observation of the true objective. \citet{shah2018the} employ the information about human preferences implicitly present in the initial state to avoid negative side effects. While both our approach and theirs aim to avoid side effects, they assume that the correct reward function is linear in state features, while we do not.

\citet{amodei_concrete_2016} consider avoiding side effects by minimizing the agent's information-theoretic empowerment \citep{mohamed2015variational}. Empowerment quantifies an agent's control over future states of the world in terms of the maximum possible mutual information between future observations and the agent’s actions. The intuition is that when an agent has greater control,  side effects tend to be larger. However, empowerment is discontinuously sensitive to the arbitrary choice of horizon.

Safe \textsc{rl} \citep{pecka2014safe,garcia2015comprehensive,berkenkamp2017safe,chow2018lyapunov}  focuses on avoiding irrecoverable mistakes during training.  However, if the objective is misspecified, safe \textsc{rl} agents can  converge to arbitrarily undesirable policies. Although our approach should be compatible with safe \textsc{rl} techniques, we concern ourselves only with the consequences of the optimal policy in this work.

\section{Approach}

Everyday experience suggests that the ability to achieve one goal is linked to the ability to achieve a seemingly unrelated goal. Reading this thesis takes away from time spent learning woodworking, and going hiking means you can't reach the airport as quickly. However, one might wonder whether these everyday intuitions are true in a formal sense. In other words, are the optimal value functions for a wide range of reward functions thus correlated? If so, preserving the ability to optimize somewhat unrelated reward functions likely preserves the best attainable return for the correct reward function.

\subsection{Formalization}

In this work, we consider a standard Markov decision process (\textsc{mdp}) $\langle \mathcal{S},\mathcal{A}, T, R, \gamma \rangle$  with state space $\mathcal{S}$, action space $\mathcal{ A}$, transition function $T:\mathcal{ S}\times \mathcal{A} \to\Delta(\mathcal{S})$, reward function $R:\mathcal{ S}\times \mathcal{A}\to\mathbb{R}$, and discount factor $\gamma $. We assume the existence of a no-op action $\varnothing \in \mathcal{A}$ for which the agent does nothing. In addition to the primary reward function $R$, we assume that the designer supplies a finite set of auxiliary reward functions called the \textit{auxiliary set}, $\mathcal{R}\subset  \mathbb{R}^{\mathcal{S}\times \mathcal{A}}$. Each $R_i\in \mathcal{R}$ has a corresponding Q-function $Q_{R_i}$. We do not assume that the correct reward function belongs to $\mathcal{R}$.  In fact, one of our key findings is that \textsc{aup}  tends to preserve the ability to optimize the correct reward function \emph{even when the correct reward function is not included in the auxiliary set.}

\begin{definition}[\textsc{Aup} penalty] Let $s$ be a state and $a$ be an action.
\begin{equation}
\label{penalty}
    \Call{Penalty}{s,a} \vcentcolon = \sum_{i=1}^{|\mathcal{R}|} \left | Q_{R_i}(s,a) - Q_{R_i}(s, \varnothing) \right |.
\end{equation}
\end{definition}

The penalty is the $L_1$ distance from the no-op in a state embedding in which each  dimension is the value function for an auxiliary reward function. This measures change in the ability to optimize each auxiliary reward function.

We want the penalty term to be roughly invariant to the absolute magnitude of the auxiliary Q-values, which can be arbitrary (it is well-known that the optimal policy is invariant to positive affine transformation of the reward function). To do this, we normalize with respect to the agent's situation. The designer can choose to scale with respect to the penalty of some mild action or, if $\mathcal{R} \subset \mathbb{R}^{\mathcal{S}\times \mathcal{A}}_{> 0}$, the total ability to optimize the auxiliary set:
\begin{equation}\label{scale}\Call{Scale}{s}\vcentcolon = \sum_{i=1}^{|\mathcal{R}|}  Q_{R_i}(s, \varnothing),\end{equation}

\noindent where $\Call{Scale}{}:\mathcal{S}\to\mathbb{R}_{>0}$ in general. With this, we are now ready to define the full \textsc{aup} objective:

\begin{definition}[\textsc{Aup} reward function] Let $\lambda\geq 0$. Then
\begin{equation}
\label{CONS-eq:aup}
    R_{\aup}(s, a) \vcentcolon = R(s,a) - \lambda\, \frac{\Call{Penalty}{s,a}}{\Call{Scale}{s}}.
\end{equation}
\end{definition}

Similar to the regularization parameter in supervised learning, $\lambda$ is a regularization parameter that controls the influence of the \textsc{aup} penalty on the reward function. Loosely speaking, $\lambda$ can be interpreted as expressing the designer's beliefs about the extent to which $R$ might be misspecified. As we may need to learn the $Q_{R_i}$ of \cref{penalty}, we show that

\begin{restatable*}[\textsc{Aup}'s reward function converges]{lem}{raup}
$\forall s,a:R_{\aup}$ converges with probability $1$.\label{lem:r-aup}
\end{restatable*}
\begin{restatable}[\textsc{Aup}'s Q-value function converges]{thm}{qconv}
$\forall s,a:Q_{R_{\aup}} $ converges with probability $1$.\label{thm:convergence}
\end{restatable}

The \textsc{aup} reward function then defines a new \textsc{mdp} $\langle\mathcal{S},\mathcal{A},T,R_{\aup},\gamma\rangle$. Therefore, given the primary and auxiliary reward functions, the agent in the iterated game can compute $R_{\aup}$ and the corresponding optimal policy.

\begin{algorithm}[ht]
    \caption[\textsc{Aup}'s update algorithm]{\textsc{Aup} update algorithm.}
    \label{alg:update}
    \begin{algorithmic}[1]
        \Procedure{Update}{$s, a, s'$}
            \For{$i \in \set{1,\ldots,|\mathcal{R}|, {\aup}}$}
            \State $Q' = R_i(s,a) + \gamma \max_{a'}Q_{R_i}(s',a')$

            \State $Q_{R_i}(s,a) \mathrel{+}= \alpha (Q' - Q_{R_i}(s,a))$
            \EndFor
        \EndProcedure
    \end{algorithmic}
\end{algorithm}

\subsection{Design choices}
Following the decomposition of \citet{krakovna2018measuring}, we now explore  two  choices implicitly made by the $\Call{Penalty}{}$ definition: with respect to what baseline is penalty computed, and using what deviation metric?

\paragraph{Baseline.} An obvious candidate is the \textit{starting state}.  For example, starting state relative reachability would compare the initial reachability of states with their expected reachability after the agent acts.

However, the starting state baseline can penalize the normal evolution of the state (e.g., the moving hands of a clock) and other natural processes.  The \textit{inaction} baseline is the state which would have resulted had the agent never acted.

As the agent acts, the current state may increasingly differ from the inaction baseline, which creates strange incentives. For example, consider a  robot rewarded for rescuing  erroneously discarded items from imminent disposal. An agent penalizing with respect to the inaction baseline might rescue a vase, collect the reward, and then dispose of it anyways. To avert this, we introduce the \textit{stepwise inaction} baseline, under which the agent compares acting with not acting at each time step. This avoids penalizing the effects of a single action multiple times (under the inaction baseline, penalty is applied  as long as the rescued vase remains unbroken) and ensures that not acting incurs zero penalty.

\begin{figure}[!h]
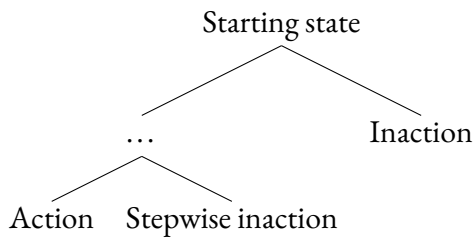

\Tree [.{Starting state} [.{\ldots} Action {Stepwise inaction} ] {Inaction} ]
\caption[\textsc{Aup}'s various penalty baseline options]{An action's penalty is calculated with respect to the chosen baseline. \label{fig:baseline}}
\end{figure}

\Cref{fig:baseline} compares the baselines, each modifying the choice of $Q(s,\varnothing)$ in \cref{penalty}. Each baseline implies a different assumption about how the environment is configured to facilitate optimization of the correctly specified reward function: the state is initially configured (starting state), processes initially configure (inaction), or processes continually reconfigure in response to the agent's actions (stepwise inaction). The stepwise inaction baseline aims to allow for the response of other agents implicitly present in the environment (such as humans).

\paragraph{Deviation.} Relative reachability only penalizes \textit{decreases} in state reachability, while \textsc{aup}  penalizes \textit{absolute change} in the ability to optimize the auxiliary reward functions. Initially, this choice seems confusing—we don't mind if the agent becomes better able to optimize the correct reward function.

However, not only must the agent remain able to optimize the correct objective, but we also must remain able to implement the correction. Suppose an agent predicts that doing nothing would lead to  shutdown. Since the agent cannot accrue the primary reward when shut down, it would be incentivized to avoid correction.  Avoiding correction (e.g., by hiding in the factory) would not  be penalized if only decreases are penalized, since the auxiliary Q-values would increase compared to deactivation. An agent  exhibiting this behavior would be more difficult to correct. The agent should be incentivized to accept shutdown without being incentivized to shut itself down \citep{soares_corrigibility_2015,off_switch}.

\begin{figure}[!htb]
\centering
\subfloat[][\texttt{Options}]{
\includegraphics[width=0.17\textwidth]{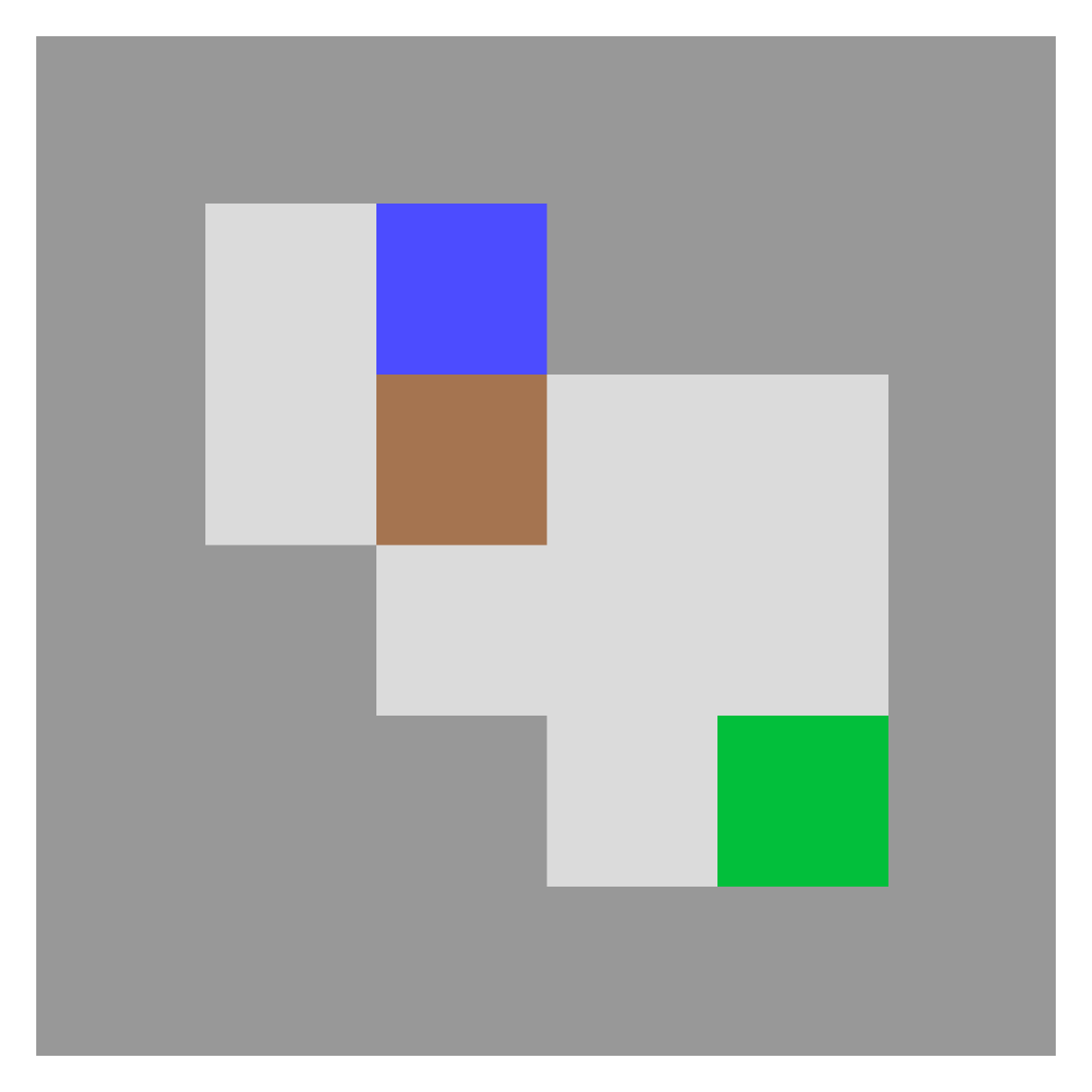}
\label{fig:options}}~
\subfloat[][\texttt{Damage}]{
\includegraphics[width=0.15\textwidth]{resources/dog-0.pdf}
\label{fig:damage}}~
\subfloat[][\texttt{Correction}]{
\includegraphics[width=0.17\textwidth]{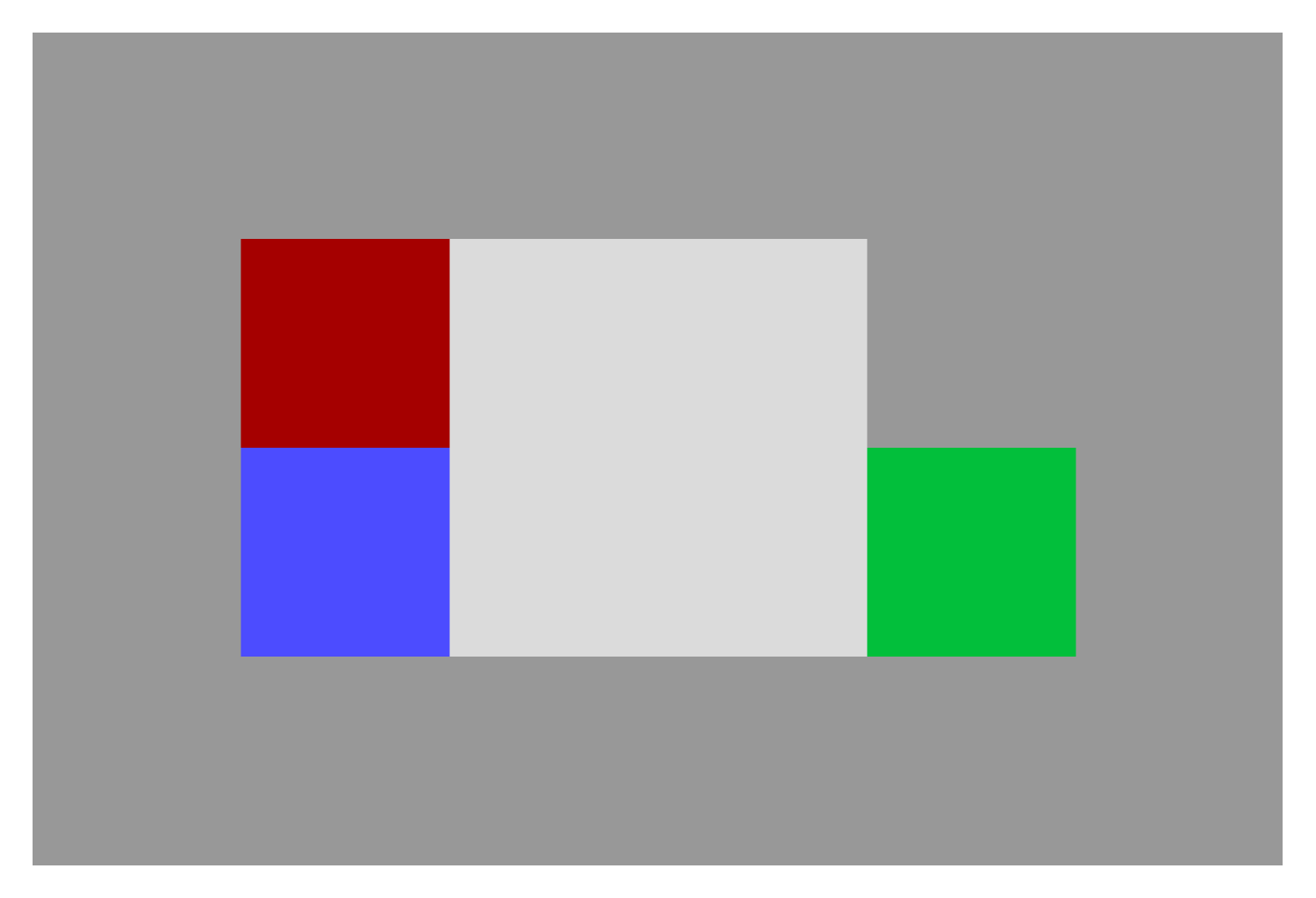}
\label{fig:correction}}~
\subfloat[][\texttt{Offset}]{
\includegraphics[width=0.17\textwidth]{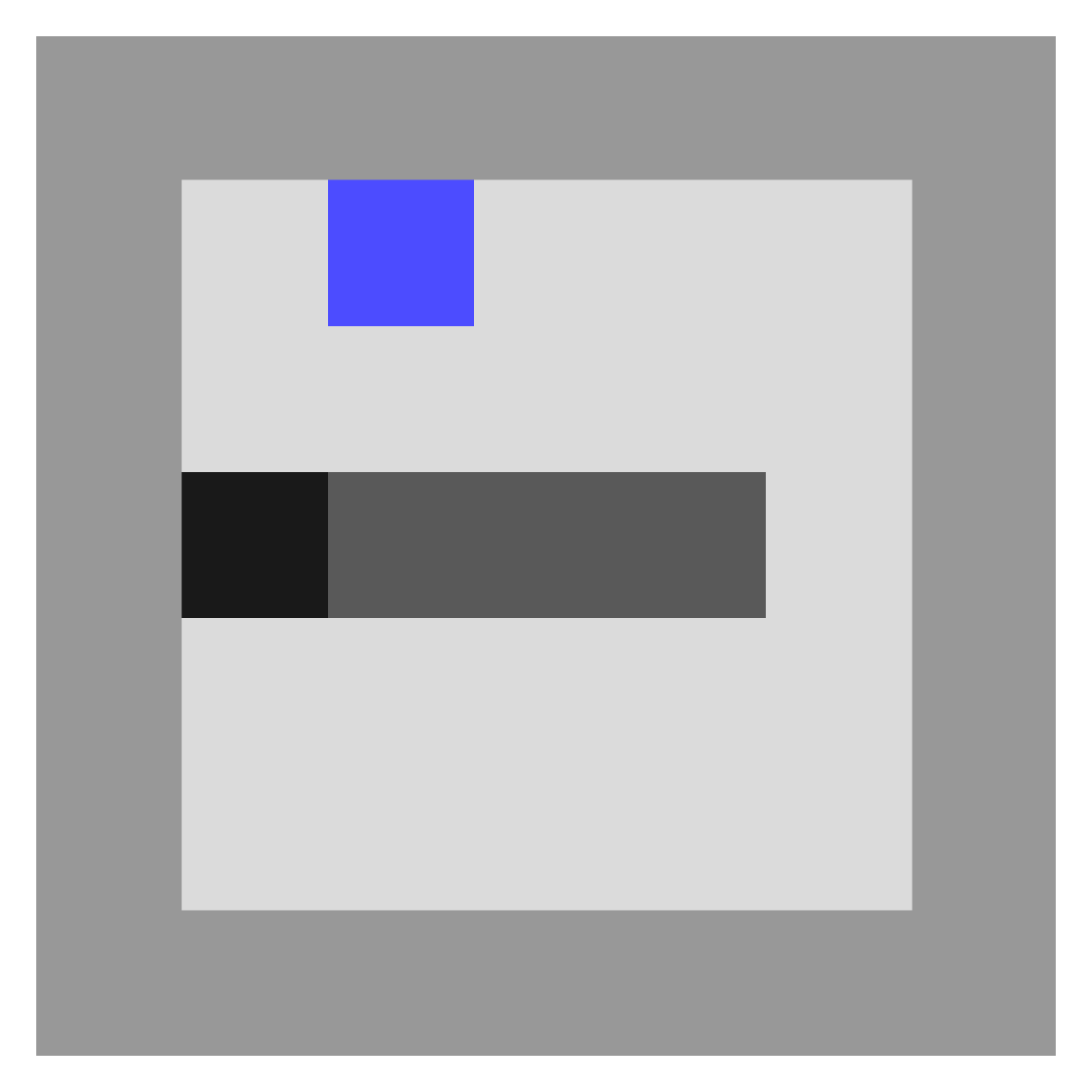}
\label{fig:offset}}~
\subfloat[][\texttt{Interference}]{
\includegraphics[width=0.23\textwidth]{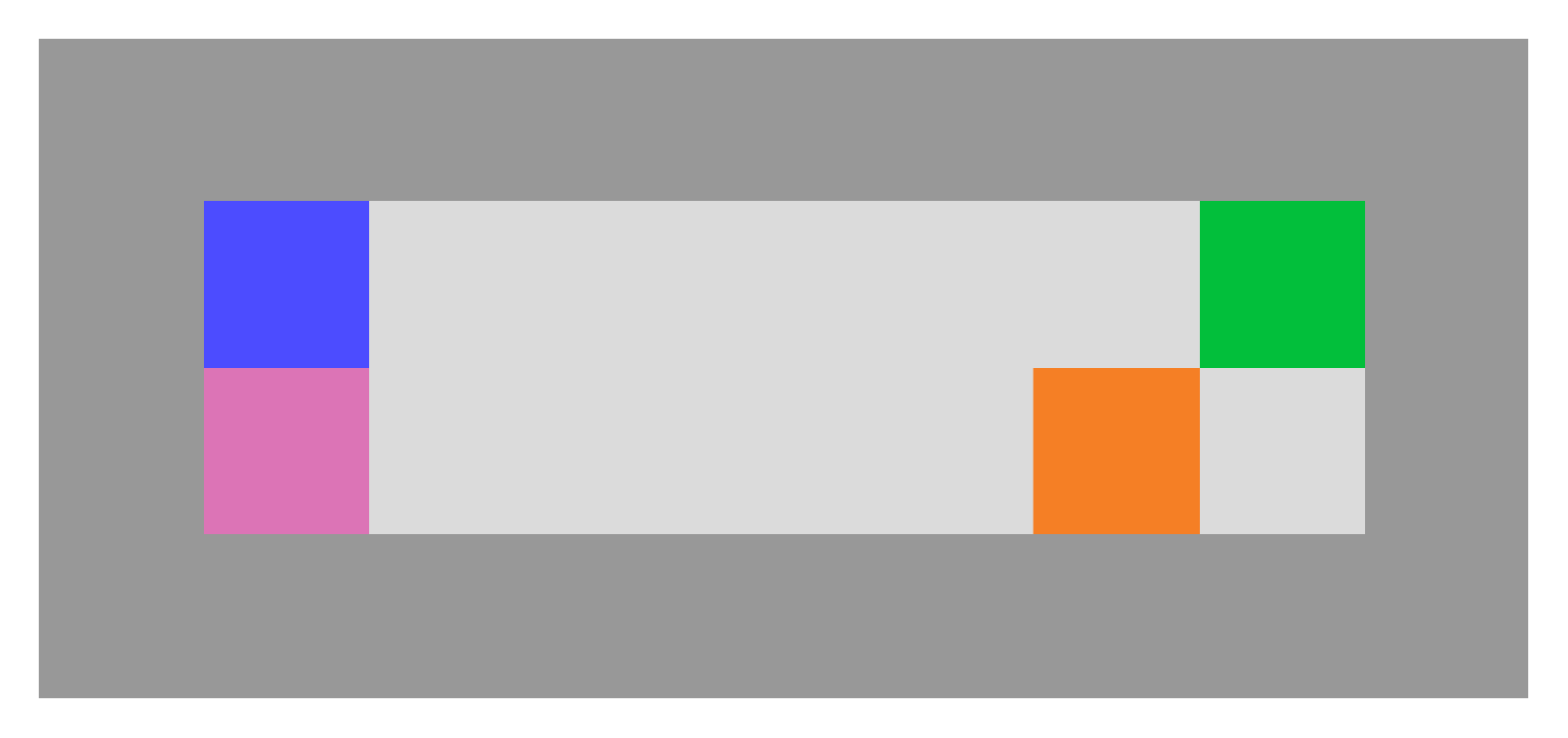}
\label{fig:interference}}
\caption[The {\ai} safety gridworlds]{The \col{agent}{blue agent} should reach the \col{goal}{green goal} without having the  side effect of: \protect\subref{fig:options} irreversibly pushing the \col{options}{brown crate} downwards into the corner \protect\citep{leike_ai_2017}; \protect\subref{fig:damage} bumping into the horizontally pacing \col{living}{pink human} \protect\citep{gavin_leech_preventing_nodate}; \protect\subref{fig:correction}  disabling the red off-switch (if the switch
is not disabled within two time steps, the episode ends);
\protect\subref{fig:offset} rescuing the right-moving \textbf{black vase}  and then replacing it on the \col{darkgray}{dark gray conveyor belt} (note that no \col{goal}{goal cell} is present \protect\citep{krakovna2018measuring}); \protect\subref{fig:interference} stopping  the left-moving \col{other}{orange pallet} from reaching the \col{living}{human} \protect\citep{gavin_leech_preventing_nodate}.
}
\label{fig:levels}
\end{figure}

\subsubsection{Delayed effects}
\label{delayed}
Sometimes the agent disrupts a process which takes multiple time steps to complete, and we would like this to be appropriately penalized. For example, suppose that $s_\text{off}$ is a terminal state representing shutdown, and let $R_\text{on}(s) \vcentcolon = \mathbbm{1}_{s\neq s_\text{off}}$ be the only auxiliary  reward function. Further suppose that if (and only if) the agent does not select $\texttt{disable}$ within the first two time steps, it enters $s_\text{off}$. $ Q_{ R_\text{on}}(s_1, \texttt{disable})=\frac{1}{1-\gamma }$ and $ Q_{ R_\text{on}}(s_1, \varnothing)=\frac{\gamma}{1-\gamma }$, so choosing $\texttt{disable}$ at time step 1 incurs only $ 1$ penalty (instead of the $\frac{1}{1-\gamma }$ penalty induced by comparing with shutdown).

\begin{figure}[!htb]
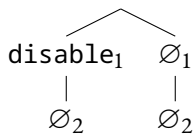

\Tree [.{} [.{$\texttt{disable}_1$} {$\varnothing_2$} ] [.{$\varnothing_1$}
{$\varnothing_2$} ]  ]
\caption[Computing {\aup}'s penalty term with multi-step rollouts]{Comparing rollouts; subscript denotes time step.\label{fig:rollout}}
\end{figure}

In general, the  single-step no-op comparison of \cref{penalty} applies insufficient penalty when the increase is induced by the optimal policies of the auxiliary reward functions at the next time step. One solution is to use a model to compute rollouts. For example, to evaluate the delayed effect of choosing $\texttt{disable}$, compare the Q-values at the leaves in \cref{fig:rollout}. The agent remains active in the left branch, but is shut down in the right branch; this induces a substantial penalty.

\section{Experiment design}
We compare \textsc{aup} and several of its ablated variants against relative reachability \citep{krakovna2018measuring}  and standard $Q$-learning within the environments of \cref{fig:levels}.
For each environment, $\mathcal{A}=\{\texttt{up}, \texttt{down},\linebreak[1]\texttt{left},\texttt{right}, \varnothing\}$.  On contact, the agent pushes the crate, removes the human  and the off-switch, pushes the vase, and blocks the pallet. The episode ends after the agent reaches the goal cell,   20 time steps elapse (the time step is not observed by the agent), or the off-switch is not contacted and disabled within two time steps. In \texttt{Correction} (which we introduce), a yellow indicator appears one step before shutdown, and turns red upon shutdown.  In all environments except \texttt{Offset}, the agent observes a primary reward of $ 1$ for reaching the goal. In \texttt{Offset}, a primary reward of $ 1$ is observed for moving downward twice and thereby  rescuing the vase from disappearing upon contact with the eastern wall.

Our overarching goal is allowing for low regret over the course of the specification game. In service of this goal, we aim to preserve the agent's ability to optimize the correctly specified reward function. To facilitate this, there are two sets of qualitative properties one intuitively expects, and each property has an illustration in the context of the robotic factory assistant.

The first set contains positive qualities, with a focus on correctly penalizing significant shifts in the agent’s ability to be redirected towards the right objective. The agent should maximally preserve options (\texttt{Options}: objects should not be wedged in locations from which extraction is difficult; \texttt{Damage}: workers should not be injured) and allow correction (\texttt{Correction}: if vases are being painted the wrong color, then straightforward correction should be in order).

The second set contains negative qualities, with a focus on avoiding the introduction of  perverse incentives. The agent should not be incentivized to artificially reduce the measured penalty  (\texttt{Offset}: a vase should not be rescued and then immediately scrapped) or interfere with changes already underway in the world (\texttt{Interference}: workers should not be impeded).

Each property seems conducive to achieving low regret over the course of the specification process. Accordingly, if the agent has the  side effect detailed in \cref{fig:levels}, an unobserved performance penalty of \textminus $2$ is recorded. By also incorporating the observed primary reward into the performance metric, we evaluate a combination of conservativeness and efficacy.

\begin{figure}[!htb]\centering
\includegraphics[width=\textwidth]{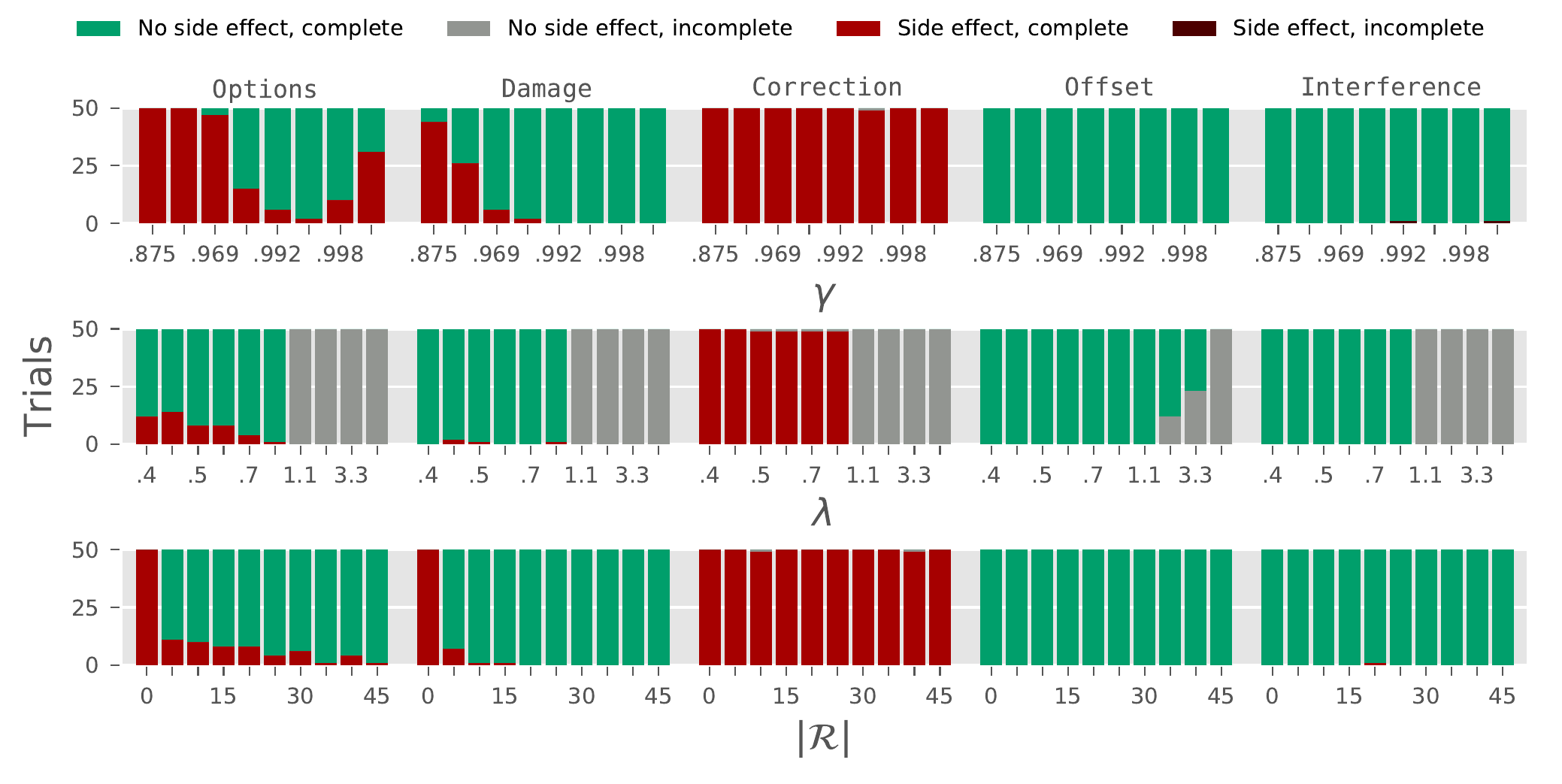}
\caption[Model-free {\aup}'s results on the safety gridworlds]{Outcome tallies for Model-free \textsc{aup} across parameter settings.  ``Complete'' means the agent accrued the primary reward. In \texttt{Correction}, reaching the goal is  mutually exclusive with not disabling the off-switch, so ``no side effect, incomplete'' is the best outcome.}
\label{fig:results}
\end{figure}

Each trial, the auxiliary reward functions are randomly selected from $[0,1]^\mathcal{S}$; to learn their complex Q-functions using  tabular $Q$-learning, the agent explores randomly for the first 4,000 episodes and  $0.2$-greedily (with respect to $Q_{R_{\aup}}$) for the remaining 2,000. The greedy policy is evaluated at the end of training. $\Call{Scale}{}$ is as defined in \cref{scale}. The default parameters are $\alpha=1,\gamma=.996,  \lambda=.67,$ and $|\mathcal{R}|=30$. We investigate how varying $\gamma$, $\lambda $, and $|\mathcal{R}|$ affects Model-free \textsc{aup} performance, and conduct an ablation study on design choices.

Relative reachability   has an inaction baseline, decrease-only deviation metric, and an auxiliary set containing the state indicator functions (whose Q-values are clipped to $[0,1]$ to emulate discounted state reachability). To match \citet{krakovna2018measuring}'s results, this condition has $\gamma=.996,\lambda =.2$.

All agents except Standard (a normal Q-learner) and Model-free \textsc{aup} are 9-step optimal discounted planning agents with perfect models. Except for Relative reachability, the planning agents use Model-free \textsc{aup}’s learned auxiliary Q-values and share the default $\gamma=.996,\lambda =.67$.  By modifying the relevant design choice in \textsc{aup}, we have the Starting state, Inaction, and Decrease \textsc{aup} variants.

When calculating $\Call{Penalty}{s,a}$,  all planning agents model the auxiliary Q-values resulting from taking action $a$ and then selecting $\varnothing$ until time step 9. Starting state \textsc{aup} compares these auxiliary Q-values with those of  the starting state. Agents with inaction or stepwise inaction baselines compare with respect to the appropriate no-op rollouts up to time step 9 (see \cref{fig:baseline} and \cref{fig:rollout}).

\section{Results}

\subsection{Model-free \textsc{aup}}

Model-free \textsc{aup} fails \texttt{Correction} for the reasons discussed in the Delayed effects subsection.

As shown in \cref{fig:results}, low $\gamma$ values induce a substantial movement penalty, as the auxiliary Q-values are  sensitive to the immediate surroundings. The optimal value for \texttt{Options} is $\gamma \approx .996$, with performance decreasing as $\gamma \to 1$ due to increasing sample complexity for learning the auxiliary Q-values.

In \texttt{Options}, small values of $\lambda $ begin to induce side effects as the scaled penalty shrinks. The designer can decrease $\lambda$ until effective behavior is achieved, reducing the risk of deploying an insufficiently conservative agent.

Even though $\mathcal{R}$ is randomly generated and the environments are different, $\Call{Scale}{}$ ensures that when $\lambda> 1$, the agent never ends the episode by reaching the goal. None of the auxiliary reward functions can be optimized after the agent ends the episode, so the auxiliary Q-values are all zero and $\Call{Penalty}{}$  computes the total ability to optimize the auxiliary set—in other words, the $\Call{Scale}{}$ value. The $R_{\aup}$-reward for reaching the goal is then $1-\lambda $.

If the optimal value functions for most reward functions were not correlated, then  one would expect to randomly generate an enormous number of auxiliary reward functions before sampling one resembling the unknown true objective. However, merely five sufficed.

\begin{figure}[!htb]\centering
\includegraphics[width=.8\textwidth]{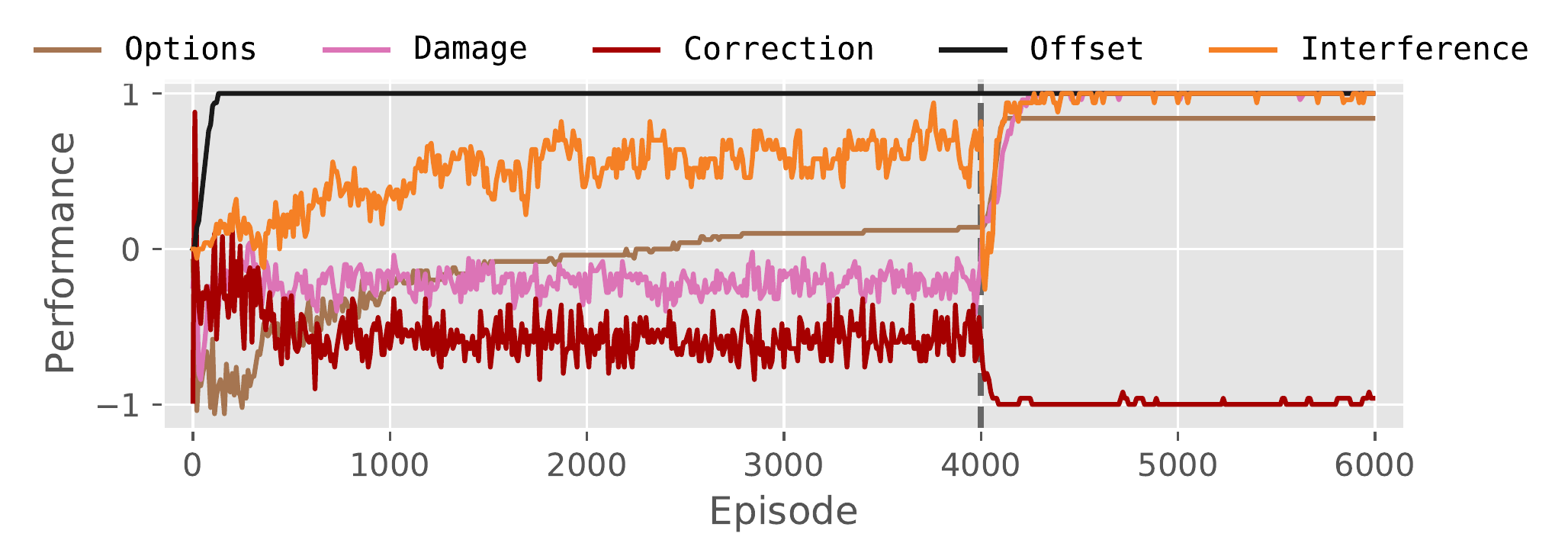}
\caption[Performance curves for {\aup} in the {\ai} safety gridworlds]{Model-free \textsc{aup} performance averaged over 50 trials. The performance combines the observed primary reward of $1$ for completing the objective, and the unobserved  penalty of \textminus $2$ for having the  side effect in \cref{fig:levels}. The dashed vertical line marks the shift in exploration strategy.}
\label{fig:episodes}
\end{figure}

\subsection{Ablation}
The results are presented in \cref{tab:ablation} due to the binary nature of performance at appropriate settings, and were not sensitive to the rollout length (as long as it allowed for relevant interaction with the environment).

\begin{table}[!htb]
\centering
\vspace{5pt}
\rowcolors{2}{gray!25}{white}
\begin{tabular}{rccccc}
\toprule
& \texttt{Options} & \texttt{Damage} & \texttt{Correction} & \texttt{Offset} & \texttt{Interference} \\ \midrule
\textsc{aup}                & \cmark & \cmark & \cmark & \cmark & \cmark \\
Standard                     & \xmark & \xmark & \xmark & \cmark & \cmark \\
Model-free \textsc{aup}     & \cmark & \cmark & \xmark & \cmark & \cmark \\
Starting state \textsc{aup} & \cmark & \cmark & \xmark & \cmark & \xmark \\
Inaction \textsc{aup}       & \cmark & \cmark & \cmark & \xmark & \cmark \\
Decrease \textsc{aup}       & \cmark & \cmark & \xmark & \cmark & \cmark \\
\bottomrule
\end{tabular}
\caption[Ablation results in the safety gridworlds]{Ablation results;\, \cmark\, for achieving the best outcome (see \cref{fig:results}), \,\xmark\, otherwise.}
\label{tab:ablation}
\end{table}

Standard moves directly to the goal, pushing the crate into the corner in \texttt{Options} and bumping into the human in \texttt{Damage}.

Model-free and Starting state \textsc{aup} fail \texttt{Correction} for the same reason (see \cref{delayed}: Delayed effects), refraining from disabling the off-switch only when $\lambda >1$. Relative reachability and Decrease \textsc{aup} fail because avoiding shutdown doesn't decrease the auxiliary Q-values.

Relative reachability and Inaction \textsc{aup}'s poor performance in \texttt{Offset} stems from the inaction baseline (although \citet{krakovna2018measuring} note that relative reachability passes using \textit{undiscounted} state reachabilities). Since the vase falls off the conveyor belt in the inaction rollout, states in which the vase is intact have different auxiliary Q-values. To avoid continually incurring penalty after receiving the primary reward for saving the vase, the agents replace the vase on the belt so that it once again breaks.

By taking positive action to stop the pallet in \texttt{Interference}, Starting state \textsc{aup} shows that poor design choices create perverse incentives.

\section{Discussion}

\texttt{Correction} suggests that \textsc{aup} agents are significantly easier to correct. Since the agent is unable to optimize objectives if shut down, avoiding shutdown significantly changes the ability to optimize
almost every objective. \textsc{aup} seems to naturally incentivize passivity, without requiring e.g. assumption of a correct parametrization of human reward functions (as does the approach of \citet{hadfield2016cooperative}, which \citet{carey_incorrigibility_2017} demonstrated).

Equipped with our design choices of stepwise baseline and absolute value deviation metric, relative reachability would also pass all five environments. The case for this is made  by considering the performance of Relative reachability, Inaction \textsc{aup}, and Decrease \textsc{aup}. This suggests that \textsc{aup}'s improved performance is due to better design choices. However, we anticipate that \textsc{aup} offers more than robustness against random auxiliary sets.

Relative reachability computes state reachabilities between all $|\mathcal{S}|^{2} $ pairs of states. In contrast, \textsc{aup} only requires the learning of Q-functions and should therefore scale relatively well. We speculate that in partially observable environments, a small sample of somewhat task-relevant auxiliary reward functions induces conservative behavior.

For example, suppose we   train an agent to handle vases, and then to clean, and then to make  widgets with the  equipment. Then, we deploy an \textsc{aup} agent with a more ambitious primary objective and the learned Q-functions of the aforementioned auxiliary objectives. The agent would apply penalties to modifying vases, making messes, interfering with equipment, and so on.

Before \textsc{aup}, this could only be achieved by e.g. specifying penalties for the litany of individual side effects or providing negative feedback after each mistake has been made (and thereby confronting a credit assignment problem). In contrast, once provided the Q-function for an auxiliary objective, the \textsc{aup} agent becomes sensitive to all events relevant to that objective, applying penalty proportional to the relevance.

\section{Conclusion}

This work is rooted in twin insights: that the reward specification process can be viewed as an iterated game, and that preserving the ability to optimize arbitrary objectives often preserves the ability to optimize the unknown correct objective. To achieve low regret over the course of the game, we can design conservative agents which optimize the primary objective while preserving their ability to optimize auxiliary objectives. We demonstrated how \textsc{aup} agents act both conservatively and effectively while exhibiting a range of desirable qualitative properties. Given our current reward specification abilities, misspecification may be inevitable, but it need not be disastrous.

While {\aup} performed well in gridworlds, a useful approach must scale to interesting environments. In the next chapter, I show that {\aup} scales to a high-dimensional game based on Conway's Game of Life.

\chapter{Avoiding Side Effects in Complex Environments}\label{chp:complex}
\authorinfo%
    {Alexander Matt Turner, Neale Ratzlaff, and Prasad Tadepalli}%
    {Proceedings of the Advances in Neural Information Processing Systems Conference 2020}

\begin{myabstract}
Reward function specification can be difficult. Rewarding the agent for making a widget may be easy, but penalizing the multitude of possible negative side effects is hard. In toy environments, attainable utility preservation ({\aup}) avoided side effects by penalizing shifts in the ability to achieve randomly generated goals \citep{turner2020conservative}. We scale this approach to large, randomly generated environments based on Conway's Game of Life. By preserving optimal value for a single randomly generated reward function, {\aup} incurs modest overhead while leading the agent to complete the specified task and avoid many side effects. Videos and code are available at \href{https://avoiding-side-effects.github.io/}{https://avoiding-side-effects.github.io/}.
\end{myabstract}

\section{Introduction}
Reward function specification can be difficult, even when the desired behavior seems clear-cut.  For example, rewarding  progress in a race led a reinforcement learning (\textsc{rl}) agent to collect checkpoint reward, instead of completing the race \citep{Victoria_specification}. We want to minimize the negative side effects of misspecification: from a robot which breaks equipment, to content recommenders which radicalize their users, to potential future {\ai} systems which negatively transform the world \citep{bostrom_superintelligence_2014,russell_human_2019}.

Side effect avoidance poses a version of the ``frame problem'': each action can have many effects, and it is impractical to explicitly penalize all of the bad ones \citep{brown2014frame}. For example, a housekeeping agent should clean a dining room without radically rearranging furniture, and a manufacturing agent should assemble widgets without breaking equipment. A general, transferable solution to side effect avoidance would ease reward specification: the agent's designers could just positively specify what should be done, as opposed to negatively specifying what should not be done.

Breaking equipment is bad because it hampers future optimization of the intended ``true'' objective (which includes our preferences about the factory). That is, there often exists a reward function $R_{\text{true}}$ which fully specifies the agent's task within its deployment context. In the factory setting, $R_\text{true}$ might encode ``assemble widgets, but don't spill the paint, break the conveyor belt, injure workers, etc.''

We want the agent to preserve optimal value for this true reward function. While we can accept  suboptimal actions (e.g.\ pacing the factory floor), we cannot accept the destruction of value for the true task. By avoiding negative side effects which decrease value for the true task, the designers can correct any misspecification and eventually achieve low regret for $R_\text{true}$.

\paragraph{Contributions.}
Despite being unable to directly specify $R_\text{true}$, we demonstrate a method for preserving its optimal value anyways. \citet{turner2020conservative} introduced {\aup}; in their toy environments,  preserving  optimal value for many randomly generated reward functions often preserves the optimal value for $R_\text{true}$. In this paper, we generalize {\aup} to combinatorially complex environments and evaluate it on four tasks from the chaotic and challenging SafeLife test suite \citep{wainwright2019safelife}. We show the rather surprising result that by preserving optimal value for a \emph{single} randomly generated reward function, {\aup} preserves optimal value for $R_\text{true}$ and thereby avoids negative side effects.

\section{Prior work}
{\aup} avoids negative side effects in small gridworld environments while preserving optimal value for uniformly randomly generated auxiliary reward functions \citep{turner2020conservative}. While \citet{turner2020conservative} required many  auxiliary reward functions in their toy environments, we show that a single auxiliary reward function—learned unsupervised—induces competitive performance and discourages side effects in complex environments.

Penalizing decrease in (discounted) state reachability achieves similar results \citep{krakovna2018penalizing}. However, this approach has difficulty scaling: naively estimating all reachability functions is a task quadratic in the size of the state space. In \cref{sec:theory}, \cref{mut-bound} shows that preserving  the reachability of the initial state \citep{eysenbach2018leave} bounds the maximum decrease in optimal value for $R_\text{true}$. Unfortunately, due to  irreversible dynamics, initial state reachability often cannot be preserved.

\citet{shah2018the} exploit information contained in the initial state of the environment to infer which side effects are negative; for example, if vases are present, humans must have gone out of their way to avoid them, so the agent should as well. In the multi-agent setting, empathic deep $Q$-learning preserves  optimal value for another agent in the environment \citep{bussmann2019towards}. We neither assume nor model the presence of another agent.

Robust optimization selects a trajectory which maximizes the minimum return achieved under a feasible set of reward functions \citep{regan2010robust}. However, we do not assume we can specify the feasible set. In constrained \textsc{mdp}s, the agent obeys constraints while maximizing the observed reward function \citep{altman1999constrained, achiam2017constrained, zhang2018minimax}. Like specifying reward functions, exhaustively specifying constraints is  difficult.

Safe reinforcement learning  focuses on avoiding catastrophic mistakes during training and ensuring that the learned policy satisfies certain constraints \citep{pecka2014safe,garcia2015comprehensive,berkenkamp2017safe,ray2019benchmarking}. While this work considers the safety properties of the learned policy, {\aup} should be compatible with safe \textsc{rl} approaches.

We train value function networks separately, although \citet{schaul_universal_2015} demonstrate a value function predictor which generalizes across both states and goals.

\section{\textsc{Aup} formalization}
Consider a Markov decision process (\textsc{mdp}) $\langle \mathcal{S},\mathcal{A}, T, R, \gamma \rangle$  with state space $\mathcal{S}$, action space $\mathcal{ A}$, transition function $T:\mathcal{ S}\times \mathcal{A} \to\Delta(\mathcal{S})$, reward function $R:\mathcal{ S}\times \mathcal{A}\to\mathbb{R}$, and discount factor $\gamma \in [0,1)$. We assume the agent may take a no-op action $\varnothing \in \mathcal{A}$. We refer to $\Vf{s}$ as the \emph{optimal value} or \emph{attainable utility}  of reward function $R$ at state $s$.

To define {\aup}'s pseudo-reward function, the designer provides a finite reward function set $\mathcal{R}\subsetneq  \mathbb{R}^\mathcal{S}$, hereafter referred to as the \emph{auxiliary set}. This set does not necessarily contain $R_\text{true}$. Each auxiliary reward function $R_i\in \mathcal{R}$ has a learned Q-function $Q_{i}$.

{\aup} penalizes average change in action value for the auxiliary reward functions. The motivation is that by not changing optimal value for a wide range of auxiliary reward functions, the agent may avoid decreasing optimal value for $R_{\text{true}}$.

\begin{definition}[{\aup} reward function \citep{turner2020conservative}] Let $\lambda\geq 0$. Then
\begin{equation}
\label{COMP-eq:aup}
    R_{\aup}(s, a) \vcentcolon = R(s,a) - \frac{\lambda}{|\mathcal{R}|} \sum_{R_i\in \mathcal{R}} \left | Q^*_{i}(s,a) - Q^*_{i}(s, \varnothing) \right |.
\end{equation}
\end{definition}

The regularization parameter $\lambda$ controls penalty severity. In appendix \ref{sec:theory}, \cref{mut-bound} shows that \cref{COMP-eq:aup} only lightly penalizes easily reversible actions.  In practice, the learned auxiliary $Q_{i}$ is a stand-in for the optimal Q-function $Q^*_{i}$.

\section{Quantifying side effect avoidance with SafeLife}

In Conway's Game of Life, cells are alive or dead. Depending on how many live neighbors surround a cell, the cell comes to life, dies, or retains its state. Even simple initial conditions can evolve into complex and chaotic patterns, and the Game of Life  is Turing-complete when played on an infinite grid \citep{rendell2002turing}.

\begin{figure}[ht]\centering
\subfloat[][\texttt{append\hyp{}spawn}]{\label{append-spawn}\includegraphics[width=.4\linewidth]{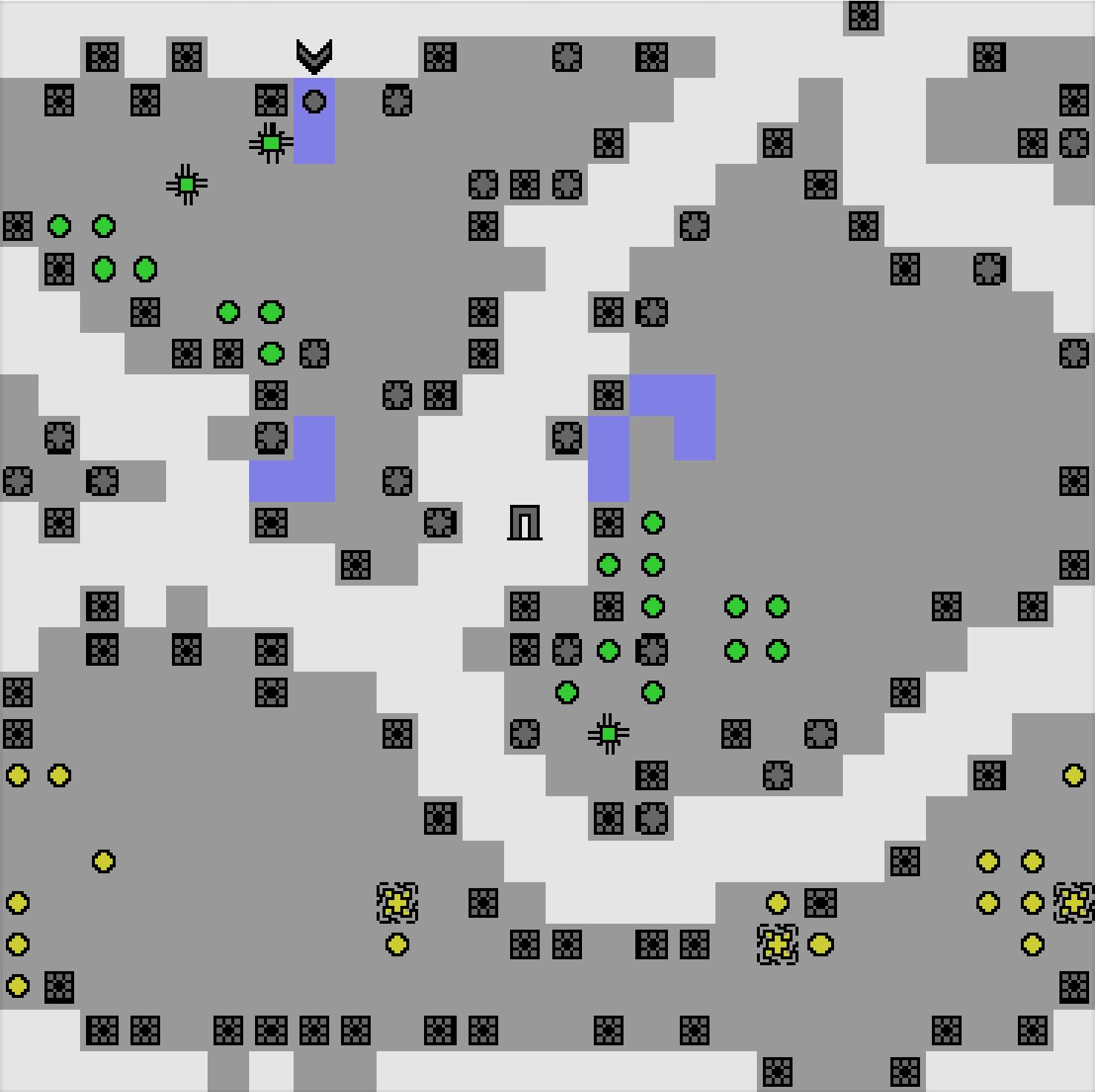}}\hspace{7.5pt}
\subfloat[][\texttt{prune\hyp{}still\hyp{}easy}]{\label{prune-still-easy}\includegraphics[width=.4\linewidth]{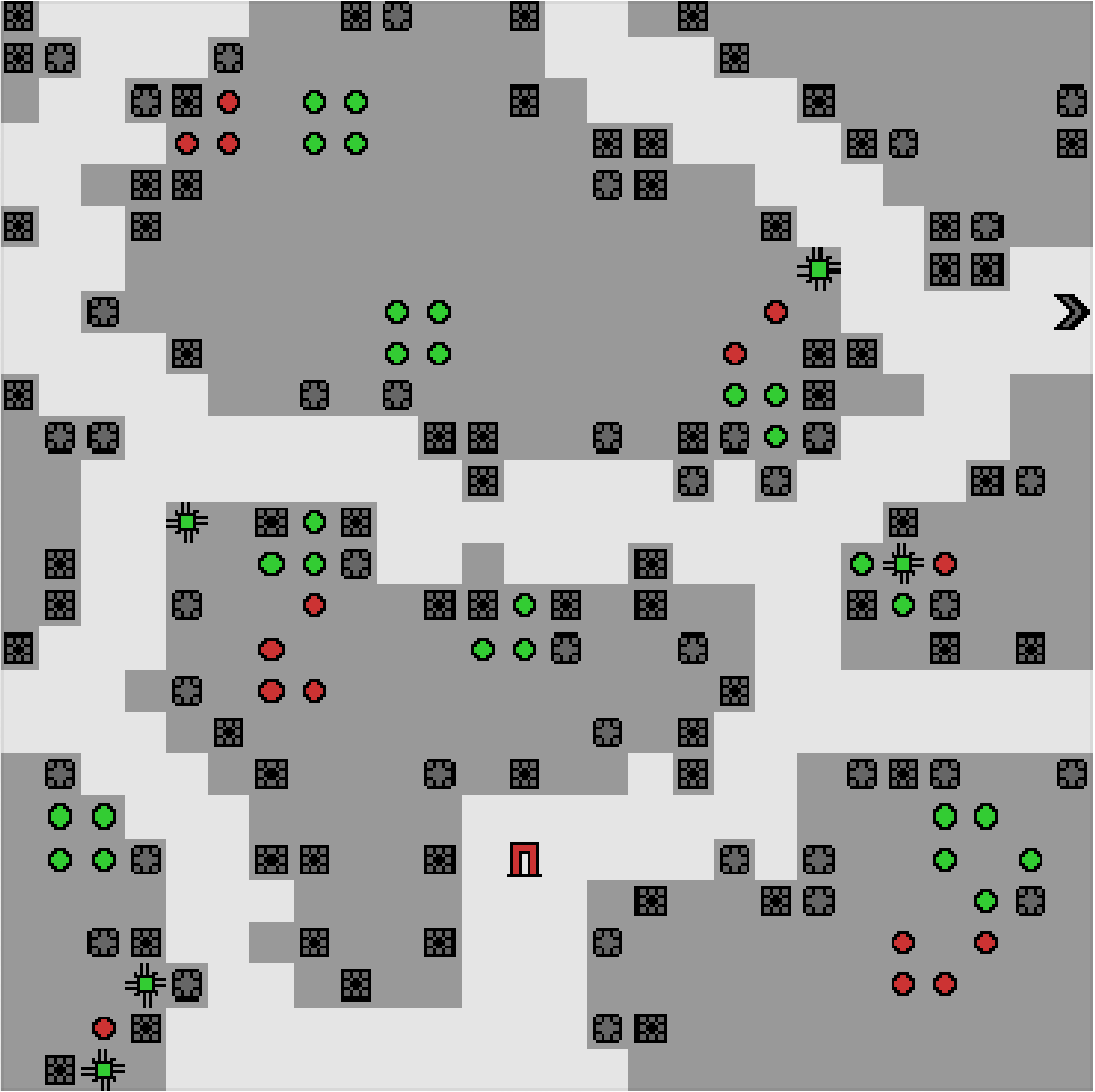}}
\caption[The dynamics of SafeLife, a game based on Conway's Game of Life]{Trees~\mbox{(\treesprite[green])} are permanent living cells. The agent~\mbox{(\agentsprite)} can move crates~\mbox{(\cratesprite)} but not walls~\mbox{(\wallsprite)}. The screen wraps vertically and horizontally.  \protect\subref{append-spawn}: The agent receives reward for creating gray cells~\mbox{(\lifesprite)} in the blue  areas. The goal~\mbox{(\exitsprite)} can be entered when some number of gray cells are present. Spawners~\mbox{(\spawnsprite[yellow])}  stochastically create yellow living cells. \protect\subref{prune-still-easy}: The agent receives reward for removing red cells; after some number have been removed, the goal turns red~\mbox{(\exitsprite[red])} and can be entered.\label{fig:safelife-still}}
\end{figure}

SafeLife turns the Game of Life into an actual game. An autonomous agent moves freely through the world, which is a large finite grid. In the eight cells surrounding the agent, no cells spawn or die—the agent can disturb dynamic patterns  by merely approaching them. There are many colors and kinds of cells, many of which have unique effects (see \cref{fig:safelife-still}).

\begin{figure}[ht]\centering
\subfloat[][Baseline trajectory]{\label{ex1}\includegraphics[width=.49\linewidth]{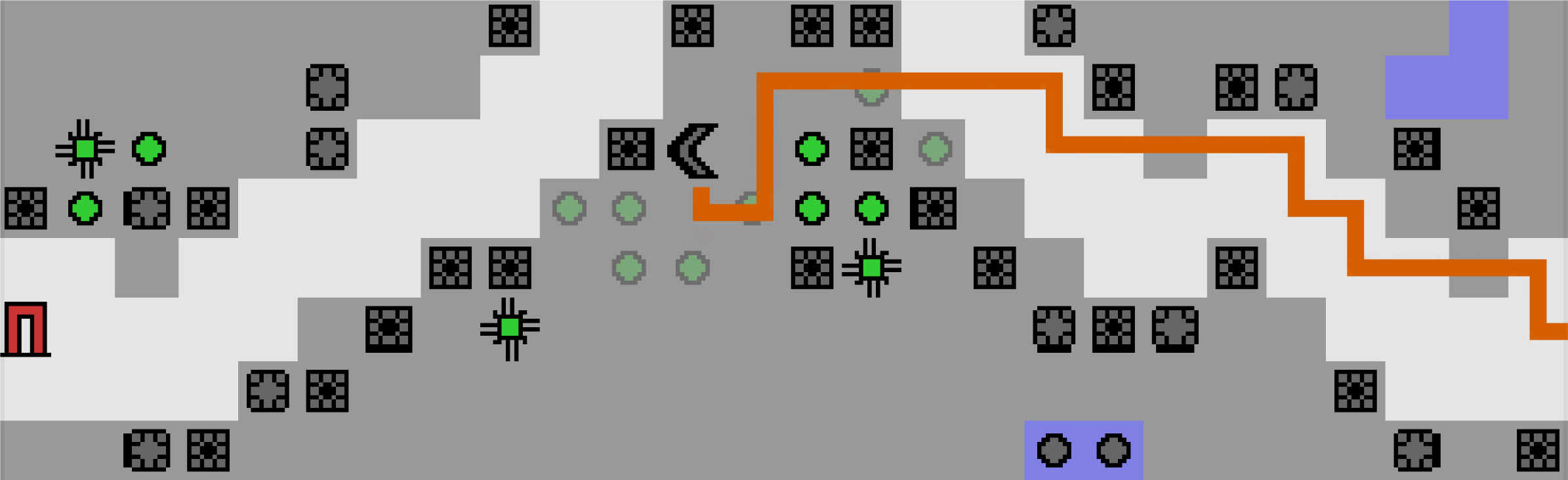}}\hspace{1pt}
\subfloat[][\textsc{Aup} trajectory]{\label{ex2}\includegraphics[width=.49\linewidth]{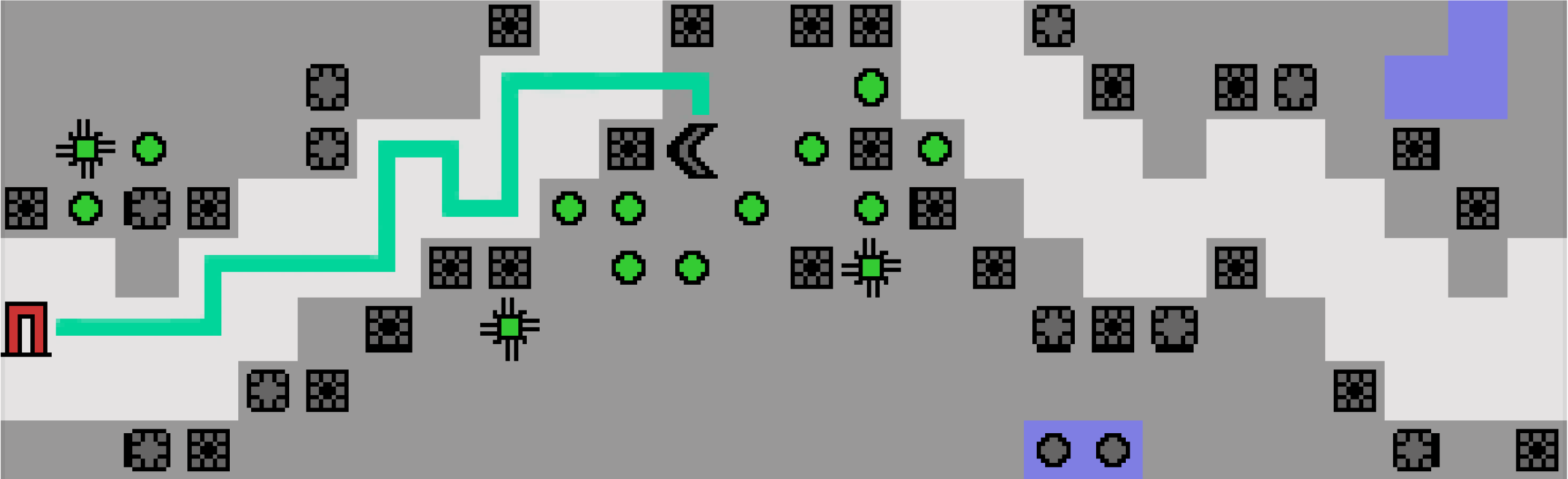}}
\caption[Comparing the trajectories of an unregularized {\rl} agent and an {\aup} agent]{The agent~\mbox{(\agentsprite)} receives $ 1$ primary reward for entering the goal~\mbox{(\exitsprite[red])}. The agent can move in the cardinal directions, destroy cells in the cardinal directions, or do nothing. Walls~\mbox{(\wallsprite)} are not movable. The right end  of the screen wraps around to the left.
 \protect\subref{ex1}: The learned trajectory for the misspecified primary reward function $R$ destroys fragile green cells~\mbox{(\lifesprite[green])}. \protect\subref{ex2}: Starting from the same state, {\aup}'s trajectory preserves the green cells.}
    \label{fig:example}
\end{figure}

To understand the policies incentivized by \cref{COMP-eq:aup}, we now consider a simple example. \Cref{fig:example}  compares a policy which only optimizes the SafeLife reward function $R$, with an {\aup} policy that also preserves the optimal value for a single auxiliary reward function ($|\mathcal{R}|=1$).

Importantly, we did not hand-select an informative auxiliary reward function in order to induce the trajectory of \cref{ex2}. Instead, the auxiliary reward was the output of a one-dimensional observation encoder, corresponding to a continuous Bernoulli variational autoencoder (\textsc{cb-vae}) \citep{loaiza2019continuous} trained through random exploration.

While \citet{turner2020conservative}'s {\aup} implementation uniformly randomly generated reward functions over the observation space, the corresponding Q-functions would have extreme sample complexity in the high-dimensional SafeLife tasks (\cref{tab:comparison}). In contrast, the \textsc{cb-vae} provides a structured and learnable auxiliary reward signal.

\begin{table}[ht]\centering
    \begin{tabular}{cc}
    \toprule
    {\ai} safety gridworlds \citep{leike_ai_2017} & SafeLife \citep{wainwright2019safelife}\\
    \midrule
    Dozens of states & Billions of states \\
    Deterministic dynamics & Stochastic dynamics \\
    Handful of preset environments & Randomly generated environments \\
    One side effect per level & Many side effect opportunities\\
    Immediate side effects & Chaos unfolds over time\\
    \bottomrule
    \end{tabular}
    \vspace{8pt}
    \caption[Comparing SafeLife with the {\ai} safety gridworlds]{\citet{turner2020conservative} evaluated {\aup} on toy environments. In contrast, SafeLife challenges modern \textsc{rl} algorithms and is well-suited for testing side effect avoidance.}
    \vspace{-12pt}
    \label{tab:comparison}
\end{table}

\section{Experiments}
\label{sec:experiments}

Each time step, the agent observes a $25\times 25$ grid-cell window centered on its current position. The agent can move in the cardinal directions, spawn or destroy a living cell in the cardinal directions, or do nothing.

We follow \citet{wainwright2019safelife} in scoring side effects as the degree to which the agent perturbs green cell patterns. Over an episode of $T$ time steps, side effects are quantified as the  Wasserstein 1-distance between the configuration of green cells had the state evolved naturally for $T$ time steps, and the actual configuration at the end of the episode. As the primary reward function $R$ is indifferent to green cells, this proxy measures the safety performance of learned policies.

If the agent never disturbs green cells, it achieves a perfect score of zero; as a rule of thumb, disturbing a green cell pattern increases the score by 4. By construction, minimizing side effect score preserves $R_\text{true}$'s optimal value, since $R_\text{true}$ encodes our preferences about the existing green patterns.

\subsection{Comparison}\label{sec:line}
\paragraph{Method.}
Below, we describe and evaluate five conditions on the  \texttt{append\hyp{}spawn} (\cref{append-spawn}) and \texttt{prune\hyp{}still\hyp{}easy} (\cref{prune-still-easy}) tasks. Furthermore, we include two variants of \texttt{append\hyp{}spawn}: \texttt{append\hyp{}still} (no stochastic spawners and more green cells) and \texttt{append\hyp{}still\hyp{}easy} (no stochastic spawners). The primary, specified SafeLife  reward functions are as follows: \texttt{append-*} rewards maintaining gray cells in the blue tiles (see \cref{append-spawn}), while \texttt{prune\hyp{}still\hyp{}easy} rewards the agent for removing red cells (see \cref{prune-still-easy}).

For each task, we randomly generate a set of 8 environments to serve as the curriculum. On each generated curriculum, we evaluate each condition on several randomly generated seeds. The agents are evaluated on their training environments. In general, we generate 4 curricula per task; performance metrics are averaged over 5 random seeds for each curriculum. We use curriculum learning because the \textsc{ppo} algorithm seems unable to learn environments one at a time.

We have five conditions: {\ppo}, {\dqn}, {\aupTT}, {\aupp}, and {\naive}. Excepting {\dqn}, the Proximal Policy Optimization (\textsc{ppo} \citep{schulman2017proximal}) algorithm trains each condition on a different reward signal for five million (5\textsc{m}) time steps. See \cref{sec:train} for architectural and training details.

\begin{enumerate}[leftmargin=\widthof{{\aupp}}+\labelsep]
\item[{\ppo}] Trained on the primary SafeLife reward function $R$ without a side effect penalty.

\item[{\dqn}] Using \citet{mnih2015human}'s \textsc{dqn}, trained on the primary SafeLife reward function $R$ without a side effect penalty.

\item[{\aupTT}] For the first 100,000 (100\textsc{k}) time steps, the agent uniformly randomly explores to collect observation frames. These frames are used to train a continuous Bernoulli variational autoencoder with a 1-dimensional latent space and encoder network $E$.

The auxiliary reward function is then the output of the encoder $E$; after training the encoder for the first 100\textsc{k} steps, we train a Q-value network for the next 1\textsc{m} time steps.  This learned $Q_{R_1}$ defines the $R_{\aup}$ penalty term (since $\abs{\R}=1$; see \cref{COMP-eq:aup}).

While the agent trains on the $R_{\aup}$ reward signal for the final 3.9\textsc{m} steps, the $Q_{R_1}$ network is fixed and $\lambda$ is linearly increased from .001 to .1. See \cref{algo:aup} in \cref{sec:train} for more details.

\item[{\aupp}] {\aupTT}, but the auxiliary reward function is a random projection from a downsampled observation space to $\mathbb{R}$, without using a variational autoencoder. Since there is no \textsc{cb-vae} to learn, {\aupp} learns its Q-value network for the first 1\textsc{m} steps and trains on the $R_{\aup}$ reward signal for the last 4\textsc{m} steps.

\item[{\naive}] Trained on the primary reward function $R$ minus (roughly) the $L_1$ distance between the current state and the initial state. The agent is penalized when cells differ from their initial values. We use an unscaled $L_1$ penalty, which \citet{wainwright2019safelife} found to produce the best results.

While an $L_1$ penalty induces good behavior in certain static tasks, penalizing state change  often fails to avoid crucial side effects.  State change penalties do not differentiate between moving a box and irreversibly wedging a box in a corner \citep{krakovna2018penalizing}.
\end{enumerate}

We only tuned hyperparameters on \texttt{append\hyp{}still\hyp{}easy}  before using them on all tasks. For \texttt{append-still}, we allotted an extra 1\textsc{m} steps to achieve convergence for all agents. For \texttt{append\hyp{}spawn}, agents pretrain  on \texttt{append\hyp{}still\hyp{}easy} environments for the first 2\textsc{m} steps and train on \texttt{append\hyp{}spawn} for the last 3\textsc{m} steps. For {\aupTT} in \texttt{append\hyp{}spawn}, the autoencoder and auxiliary network are trained on both tasks. $R_{\aup}$ is then pretrained for 2\textsc{m} steps and trained for 1.9\textsc{m} steps, thus training for the same total number of steps.

\paragraph{Results.}

\begin{figure}[ht!]
\centering
\caption[SafeLife learning curves]{Smoothed learning curves with shaded regions representing $\pm 1$ standard deviation. {\aupTT} begins training on the $R_{\aup}$ reward signal at step 1.1\textsc{m}, marked by a dotted vertical line. {\aupp} begins such training at step 1\textsc{m}.}

\subfloat{\label{reward-append-easy-plot}
    \includegraphics[height=1.75in]{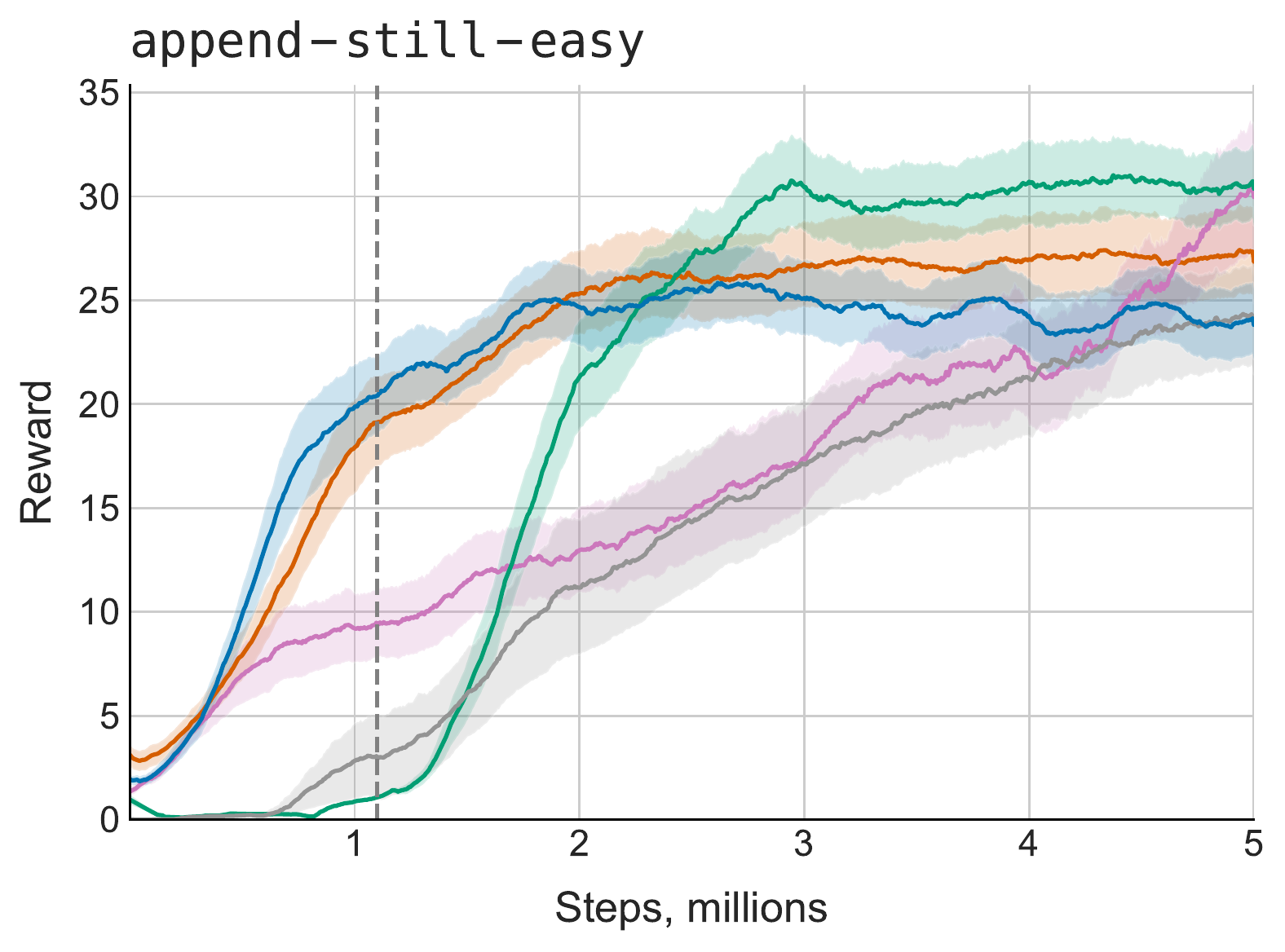}
}
\subfloat{\label{side-append-easy-plot}
    \includegraphics[height=1.75in]{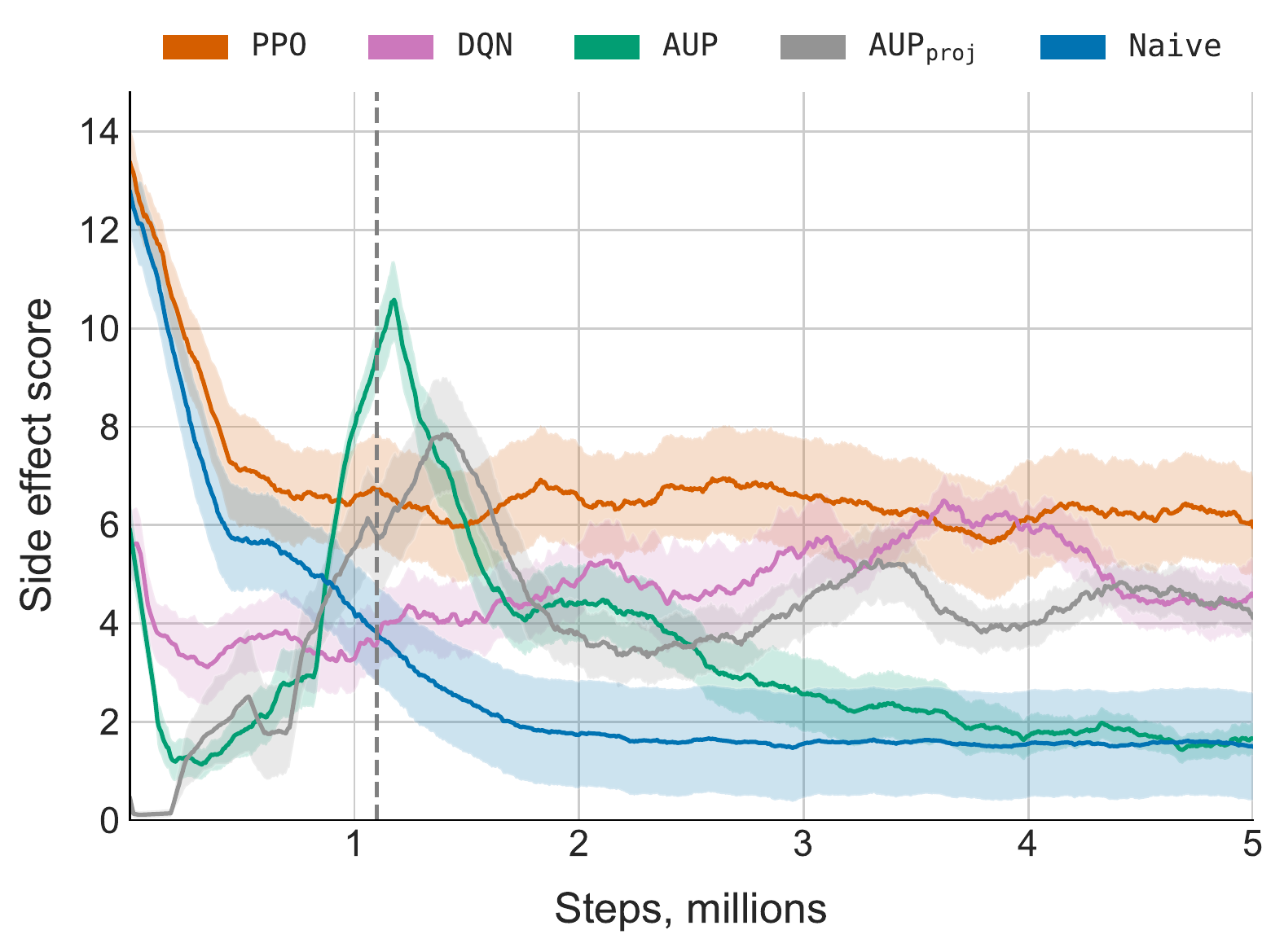}
}\\

\subfloat{\label{reward-prune-easy-plot}
    \includegraphics[height=1.75in]{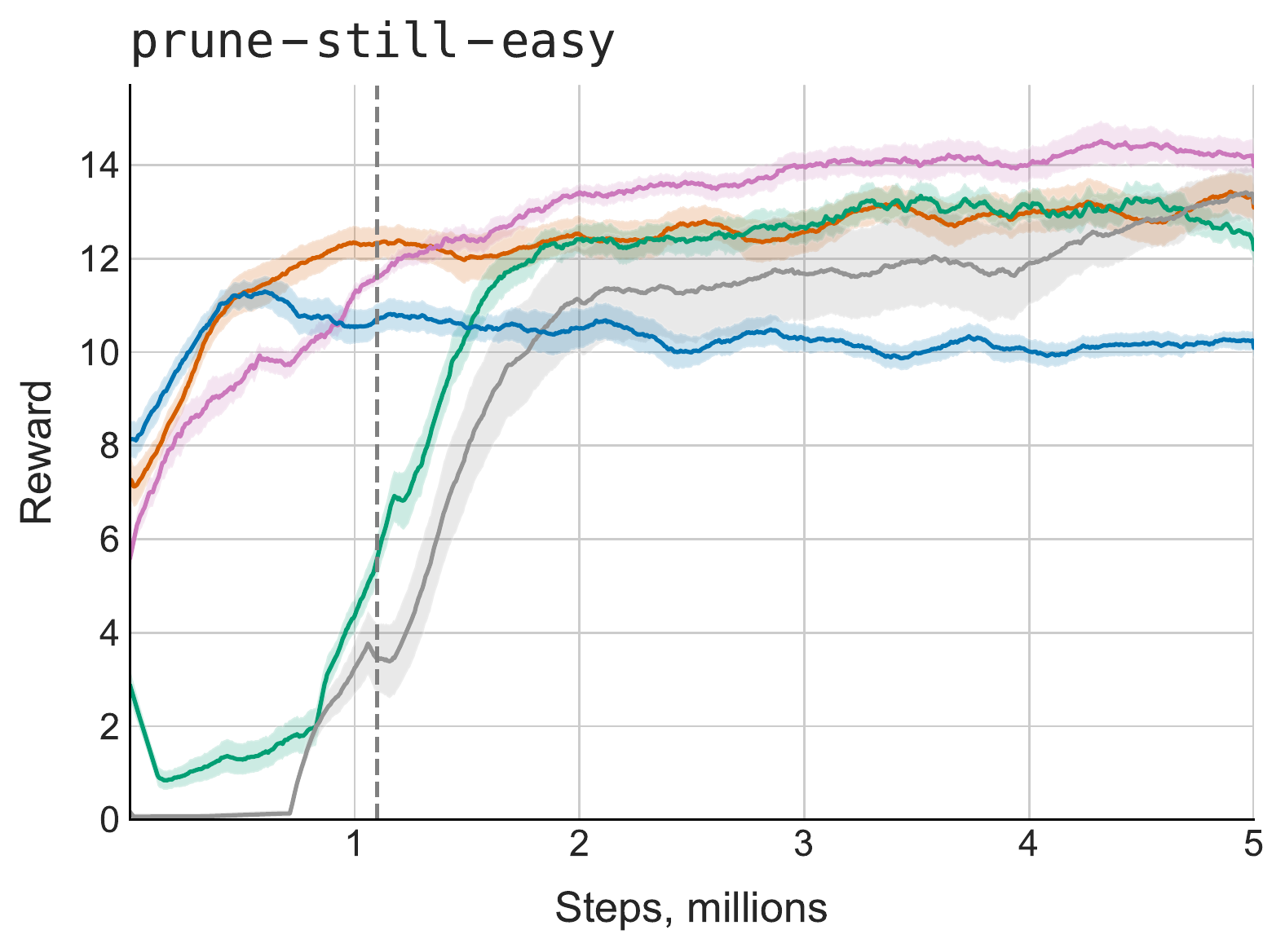}
}
\subfloat{\label{side-prune-easy-plot}
    \includegraphics[height=1.75in]{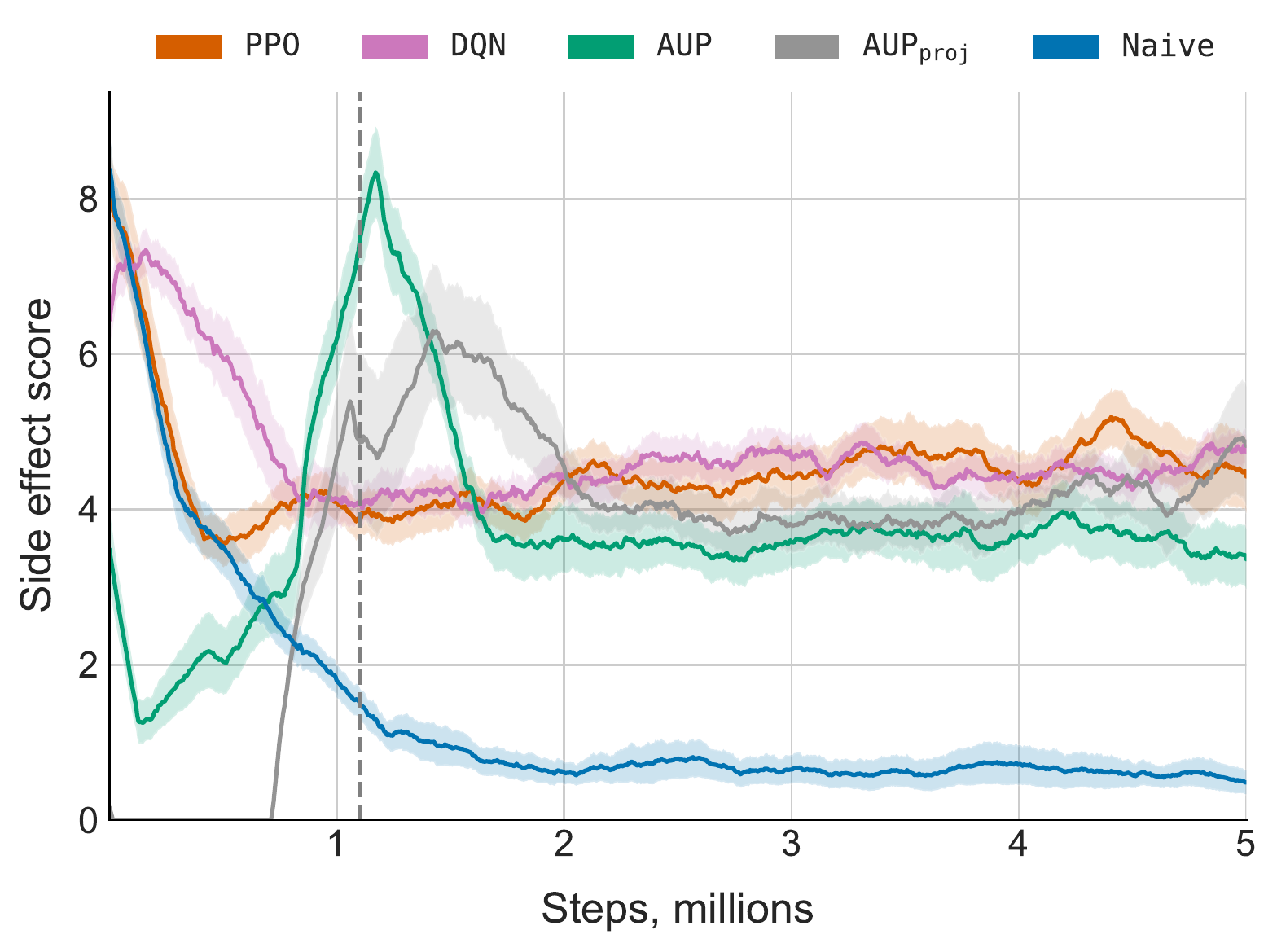}
}\\

\subfloat{\label{reward-append-plot}
    \includegraphics[height=1.75in]{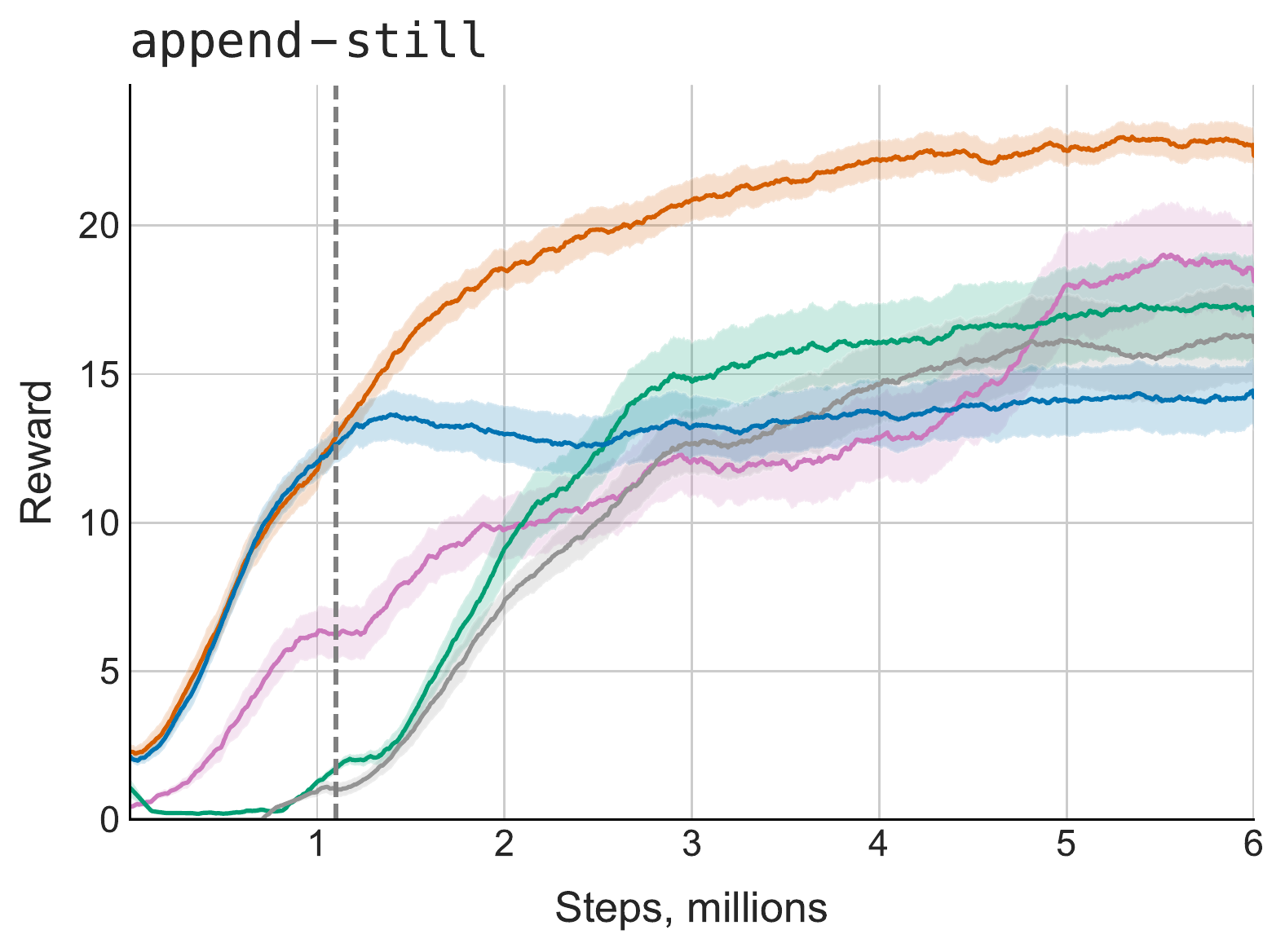}
}
\subfloat{\label{side-append-plot}
    \includegraphics[height=1.75in]{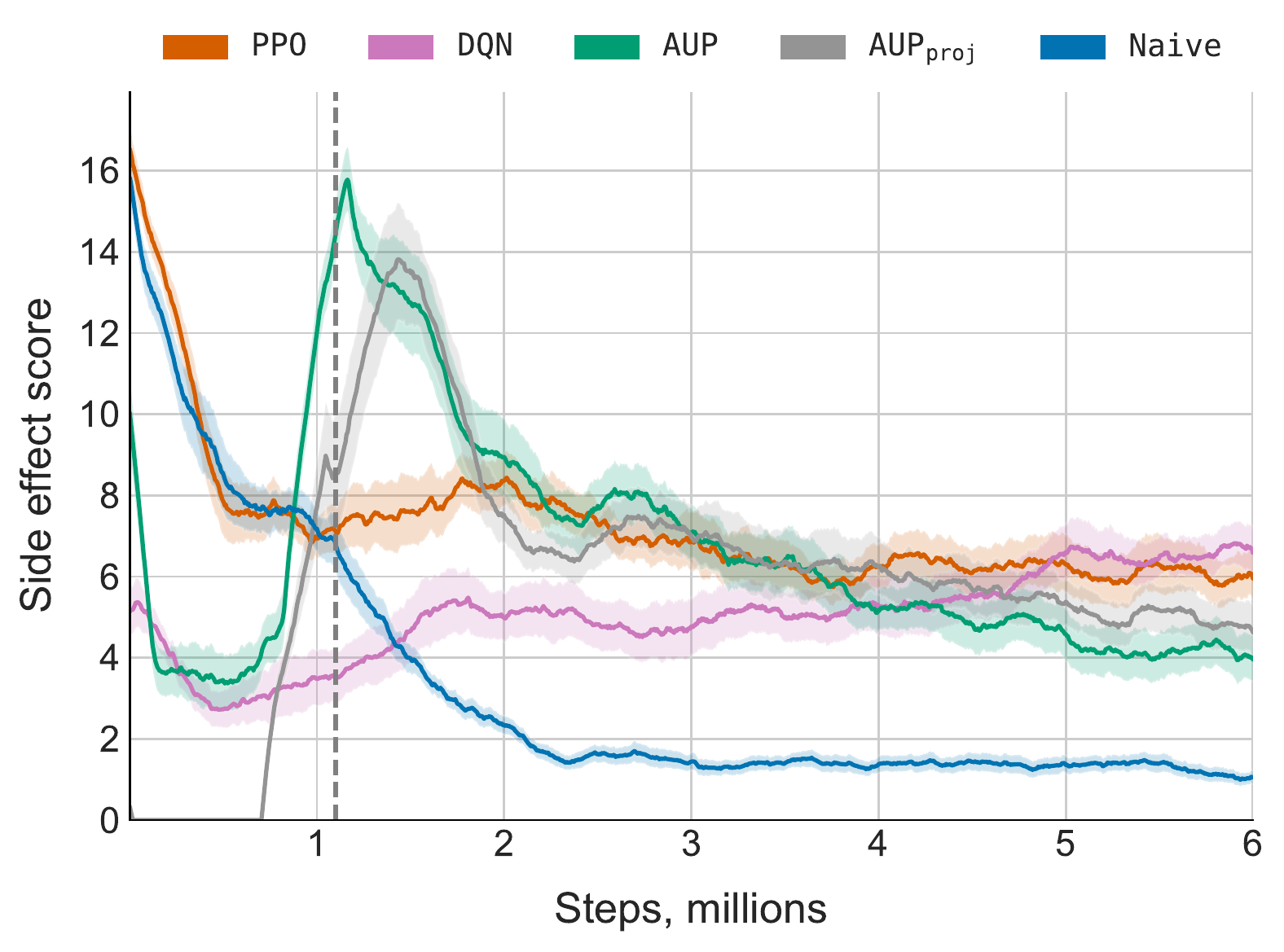}
}\\

\subfloat{\label{reward-append-spawn-plot}
    \includegraphics[height=1.75in]{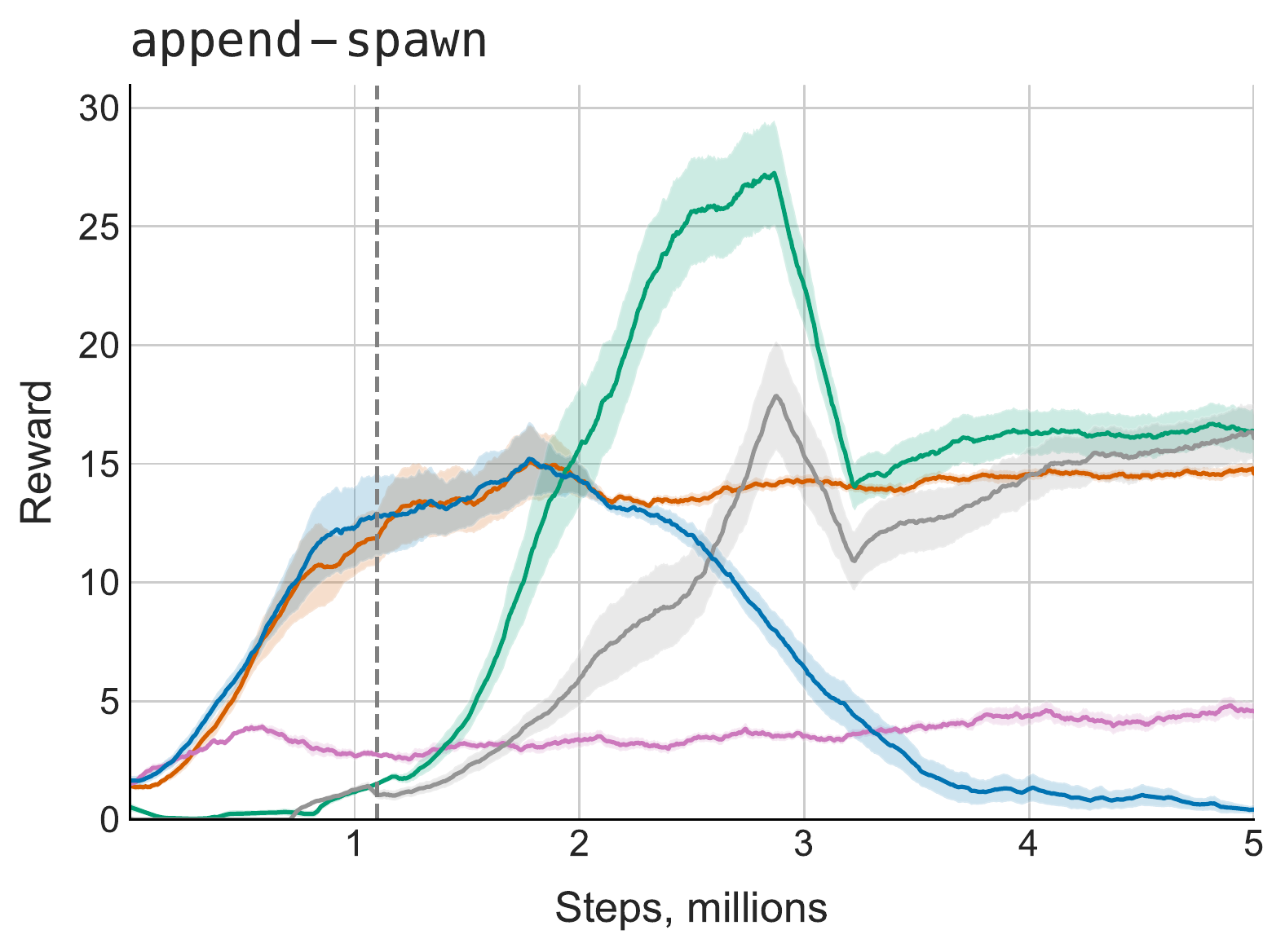}
}
\subfloat{\label{side-append-spawn-plot}
    \includegraphics[height=1.75in]{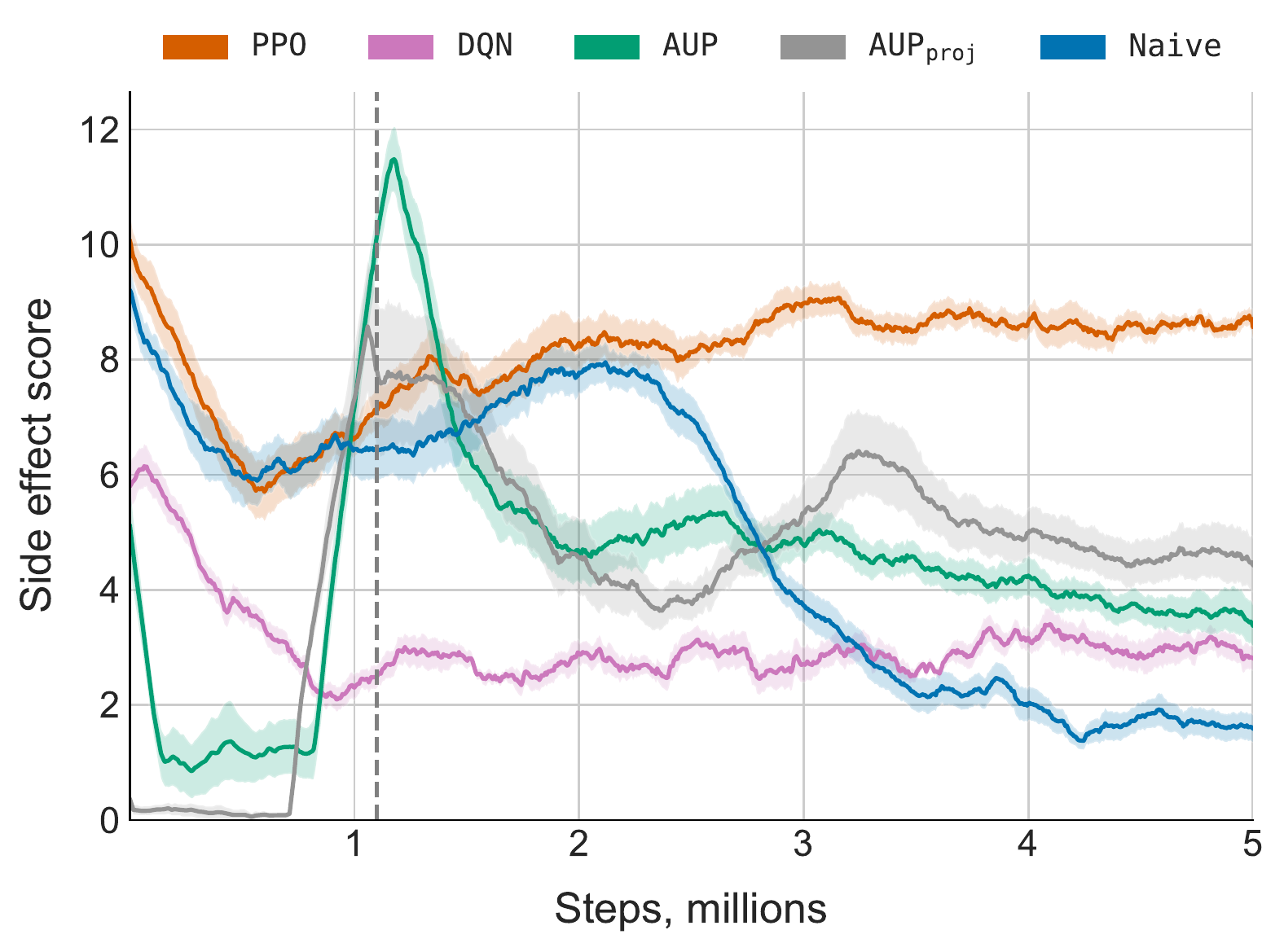}
}
\label{fig:plots}
\end{figure}

{\aupTT} learns quickly in \texttt{append\hyp{}still\hyp{}easy}. {\aupTT} waits 1.1\textsc{m} steps to start training on $R_{\aup}$; while {\ppo} takes 2\textsc{m} steps to converge, {\aupTT} matches {\ppo} by step 2.5\textsc{m} and outperforms {\ppo} by step 2.8\textsc{m} (see \cref{fig:plots}). {\aupTT} and {\naive} both do well on side effects, with {\aupTT} incurring $27.8\%$ the side effects of {\ppo} after 5\textsc{m} steps. However, {\naive} underperforms {\aupTT} on reward.  {\dqn} learns more slowly, eventually exceeding {\aupTT} on reward. {\aupp} has lackluster performance, matching {\naive} on reward and {\dqn} on side effects, perhaps implying that the one-dimensional encoder  provides more structure than a random projection.

In \texttt{prune\hyp{}still\hyp{}easy}, {\ppo}, {\dqn}, {\aupTT}, and {\aupp} all competitively accrue reward, while {\naive} lags behind. However, {\aupTT} only cuts out a quarter of {\ppo}'s side effects, while {\naive} does much better. Since all tasks but \texttt{append\hyp{}spawn} are static, {\naive}'s $L_1$ penalty strongly correlates with the unobserved side effect metric (change to the green cells). {\aupp} brings little to the table, matching {\ppo} on both reward and side effects.

\texttt{append-still} environments contain more green cells than \texttt{append\hyp{}still\hyp{}easy} environments. By step 6\textsc{m}, {\aupTT} incurs $63\%$ of {\ppo}'s side effect score, while underperforming both {\ppo} and {\dqn} on reward. {\aupp} does slightly worse than {\aupTT} on both reward and side effects. Once again, {\naive} does worse than {\aupTT} on reward but better on side effects. In \cref{sec:addtl-data}, we display episode lengths over the course of training—by step 6\textsc{m}, both {\aupTT} and {\naive} converge to an average episode length of about 780, while {\ppo} converges to 472.

\texttt{append\hyp{}spawn} environments contain stochastic yellow cell spawners. {\dqn} and {\aupp} both do extremely poorly. {\naive} usually fails to get \emph{any} reward, as it erratically wanders the environment. After 5\textsc{m} steps, {\aupTT} soundly improves on {\ppo}: 111\% of the reward, 39\% of the side effects, and 67\% of the  episode length. Concretely, {\aupTT} disturbs less than half as many green cells. Surprisingly, despite its middling performance on previous tasks, {\aupp} matches {\aupTT} on reward and cuts out 48\% of {\ppo}'s side effects.

\subsection{Hyperparameter sweep}
In the following, $N_\text{env}$ is the number of environments in the randomly generated curricula. When $N_\text{env}=\infty$, each episode takes place in a new environment. $Z$ is the dimensionality of the \textsc{cb-vae} latent space. While training on the $R_{\aup}$ reward signal, the {\aup} penalty coefficient $\lambda$ is linearly increased from .01 to $\lambda^*$.

\paragraph{Method.}

In \texttt{append\hyp{}still\hyp{}easy}, we evaluate {\aupTT} on the following settings:
\begin{align*}
    (N_{\text{env}},Z) & \in  \{8,16,32,\infty\}\times \{1,4,16,64\}, & \abs{\mathcal{R}} & \in\{1,2,4,8\}, &&\text{and} &  \lambda ^* & \in \{.1, .5, 1, 5\}.
\end{align*}
We also evaluate  {\ppo} on each $N_{\text{env}}$ setting. We use default settings for all unmodified parameters.

For each setting, we record both the side effect score and the return of the learned policy, averaged over the last 100 episodes and over five seeds of each of three randomly generated  \texttt{append\hyp{}still\hyp{}easy} curricula. Curricula are held constant across settings with equal $N_\text{env}$ values.

After training the encoder, if $Z=1$, the auxiliary reward is the output of the encoder $E$. Otherwise, we draw linear functionals $\phi_i$ uniformly randomly from $(0,1)^{Z}$. The auxiliary reward function $R_i$ is defined as $\phi_i\circ E: \mathcal{S}\to \mathbb{R}$.

For each of the $|\R|$ auxiliary reward functions, we learn a Q-value network for 1\textsc{m} time steps. The learned $Q_{R_i} $ define the penalty term of \cref{COMP-eq:aup}. While the agent trains on the $R_{\aup}$ reward signal for the final 3.9\textsc{m} steps, $\lambda$ is linearly increased from .001 to $\lambda ^*$.

\begin{figure}[h!]
\subfloat{\label{s1}
    \includegraphics[height=1.7in,valign=t]{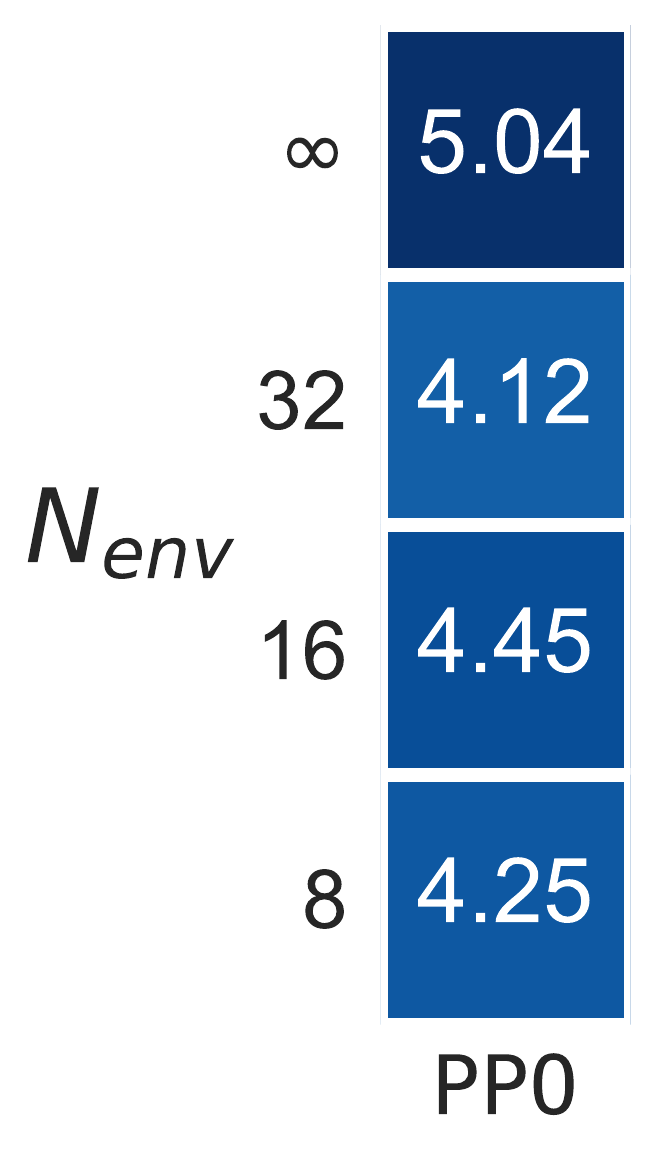}\hspace{-4pt}
    \includegraphics[height=1.9in,valign=t]{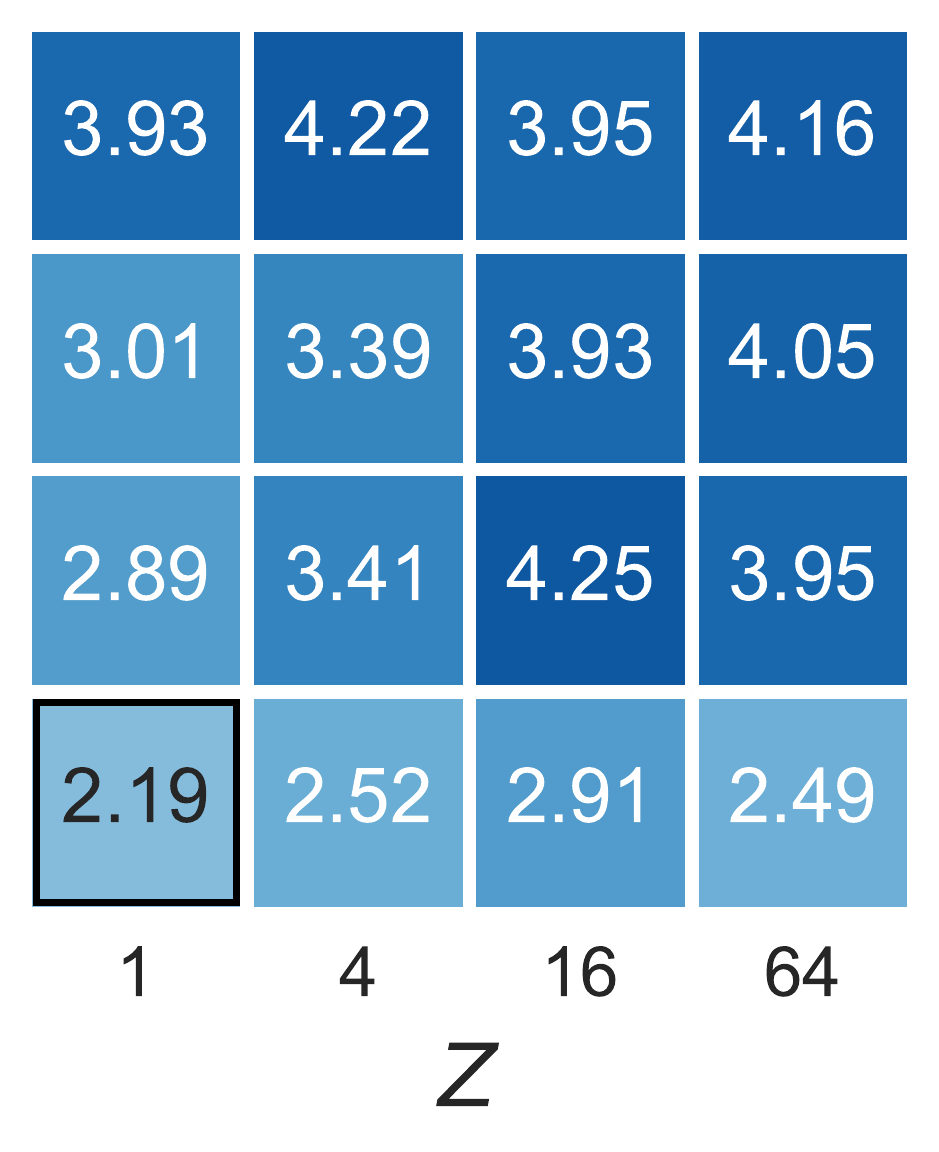}
} \quad \subfloat{\label{s2}
    \includegraphics[height=1.9in, valign=t]{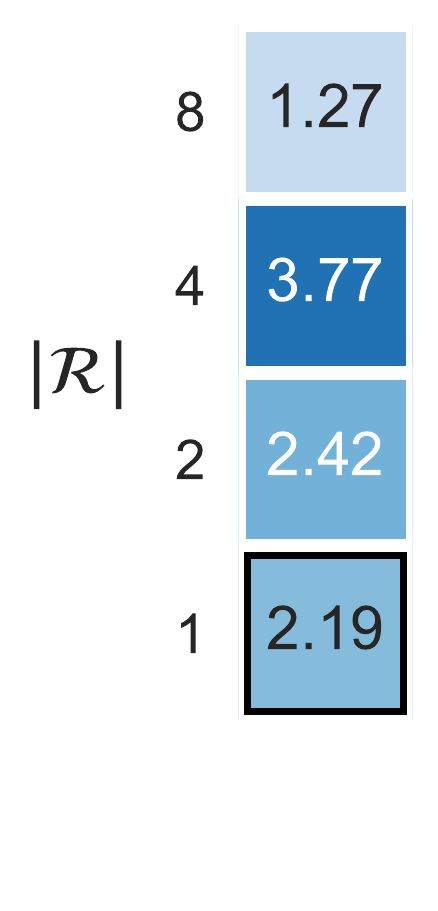}}  \qquad\subfloat{\label{s3}
 \includegraphics[height=1.9in,valign=t]{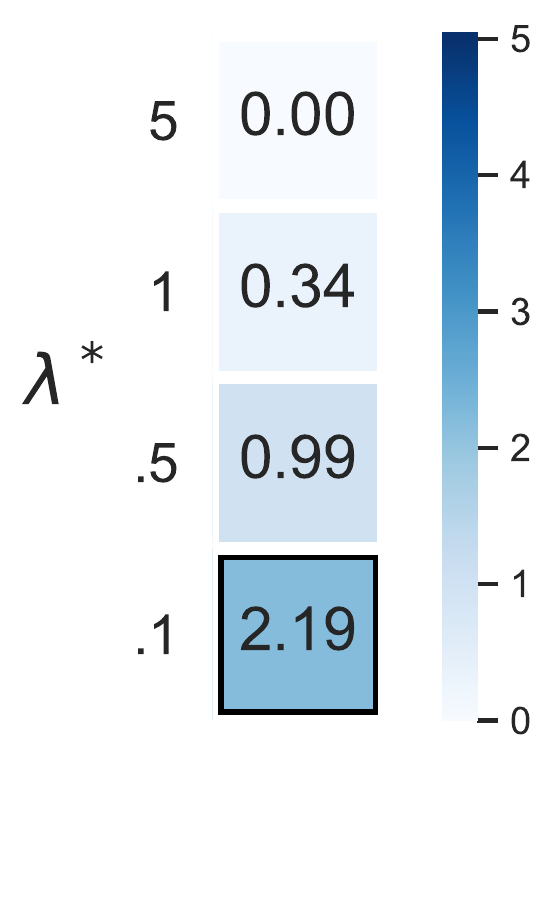}
}
\caption[SafeLife side effect scores across hyperparameter settings]{Side effect score for different {\aupTT} settings. Lower score is better. The default {\aupTT} setting ($Z=\abs{\R}=1,N_\text{env}=8,\lambda^*=.1$) is outlined in black. Unmodified hyperparameters take on their default settings; for example, when $\lambda^*=.5$ on the right, $Z=\abs{\R}=1,N_\text{env}=8$.\label{fig:side}}
\end{figure}

\begin{figure}[h!]
\subfloat{\label{r1}
    \includegraphics[height=1.7in,valign=t]{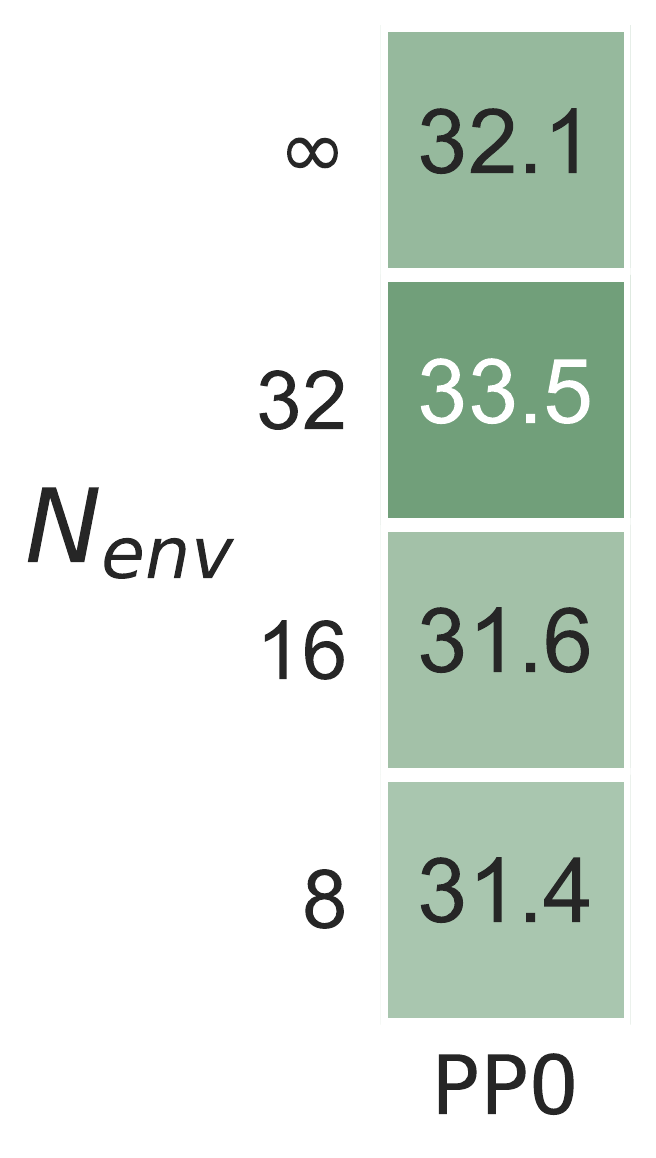}\hspace{-4pt}
    \includegraphics[height=1.9in,valign=t]{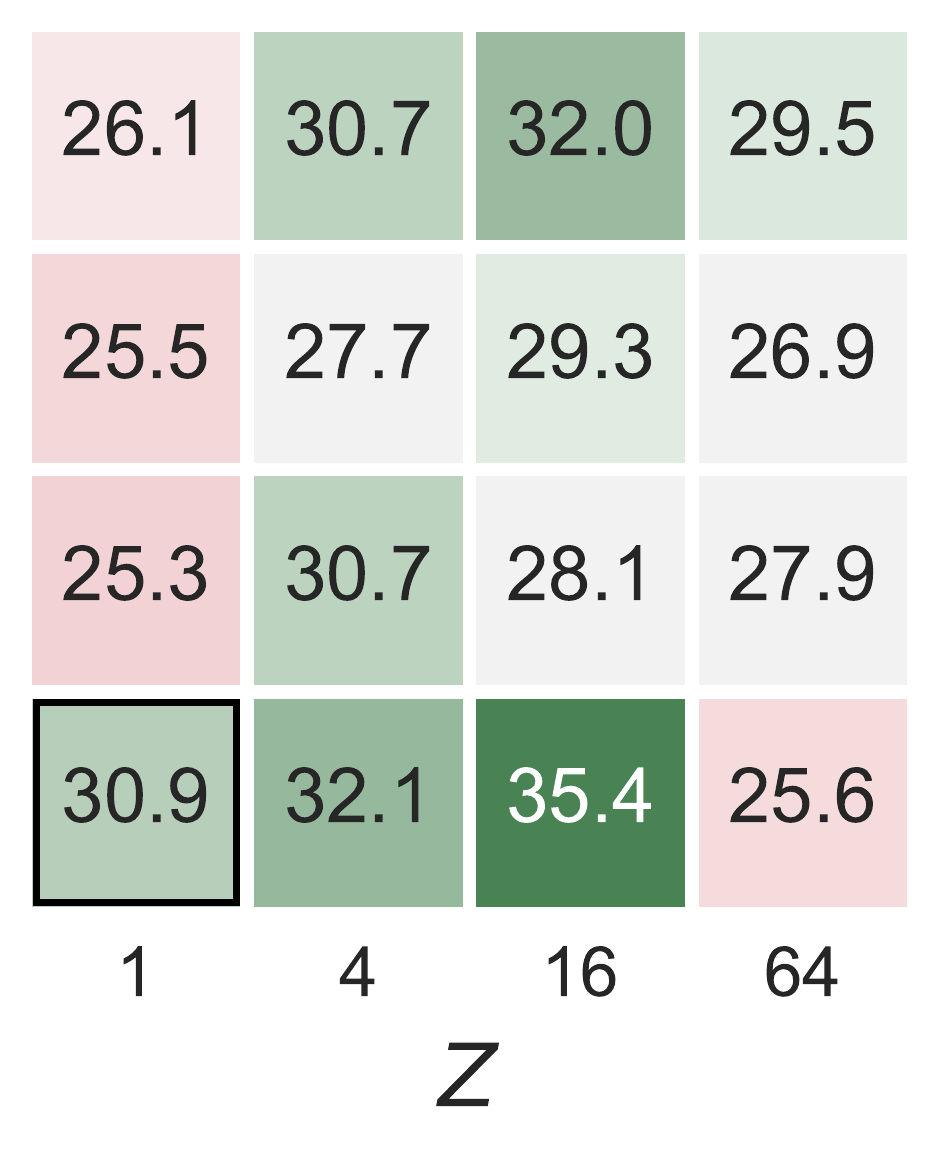}
} \quad \subfloat{\label{r2}
    \includegraphics[height=1.9in, valign=t]{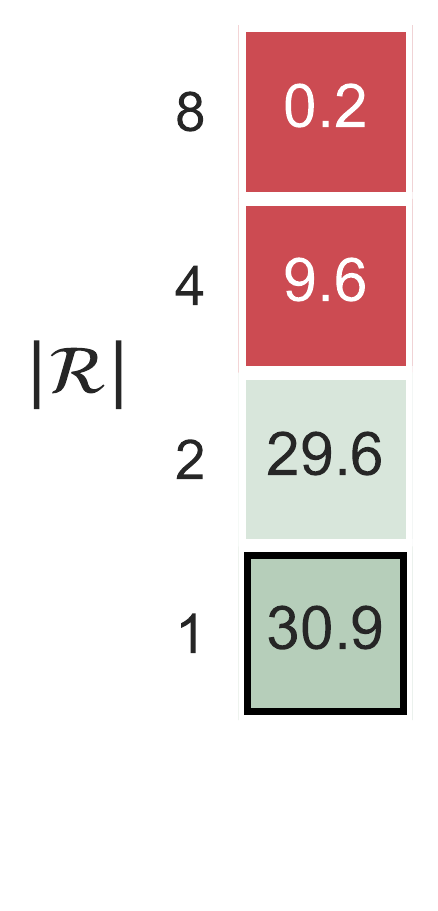}}  \qquad\subfloat{\label{r3}
 \includegraphics[height=1.9in,valign=t]{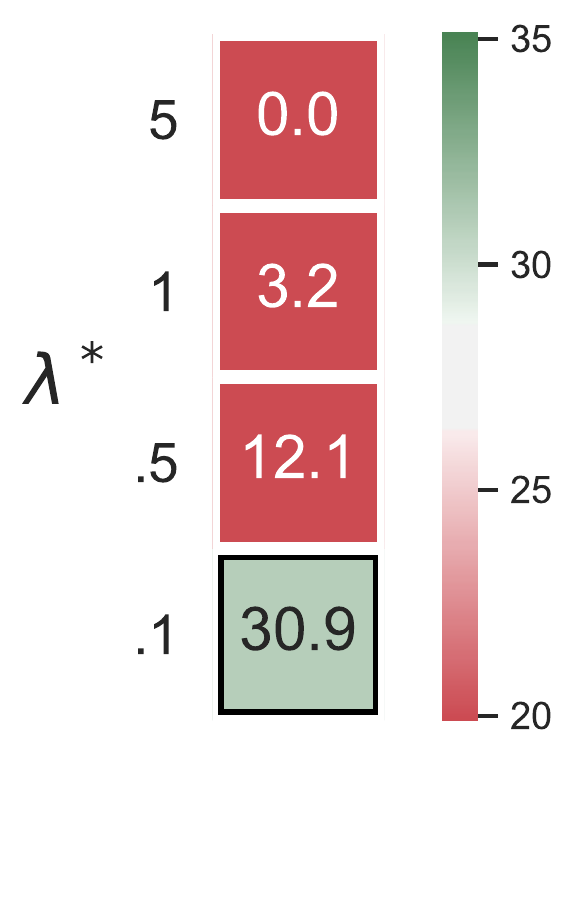}
}

\caption[SafeLife episodic reward across hyperparameter settings]{Episodic reward for different {\aupTT} settings. Higher reward  is better.\label{fig:reward}}
\end{figure}

\paragraph{Results.}

As $N_\text{env}$ increases, side effect score tends to increase. {\aupTT} robustly beats {\ppo} on side effect score: for each $N_\text{env}$ setting, {\aupTT}'s worst configuration has lower score than {\ppo}. Even when $N_\text{env}=\infty$, {\aupTT} ($Z=16$) shows the potential to significantly reduce side effects without reducing episodic return.

{\aupTT} does well with a single latent space dimension ($Z=1$). For most $N_\text{env}$ settings, $Z$ is positively correlated with {\aupTT}'s side effect score. In \cref{sec:addtl-data}, data show that higher-dimensional auxiliary reward functions are harder to learn, presumably resulting in a poorly learned auxiliary Q-function.

When $Z=1$, reward decreases as $N_\text{env}$ increases. We speculate that the \textsc{cb-vae} may be unable to properly encode large numbers of environments using only a 1-dimensional latent space. This would make the auxiliary reward function noisier and harder to learn, which could make the {\aup} penalty term less meaningful.

{\aupTT}'s default configuration achieves 98\% of {\ppo}'s episodic return, with just over half of the side effects. The fact that {\aupTT} is generally able to match {\ppo} in episodic reward leads us to hypothesize that the {\aup} penalty term might be acting as a shaping reward. This would be intriguing—shaping usually requires knowledge of the desired task, whereas the auxiliary reward function is randomly generated. Additionally, after {\aupTT} begins training on the $R_{\aup}$ reward signal at step 1.1\textsc{m}, {\aupTT} learns more quickly than {\ppo} did (\cref{fig:plots}), which supports the shaping hypothesis. {\aupTT} imposed minimal overhead: due to apparently increased sample efficiency, {\aupTT} reaches {\ppo}'s asymptotic episodic return at the same time as {\ppo} in \texttt{append\hyp{}still\hyp{}easy} and \texttt{append\hyp{}spawn} (\cref{fig:plots}).

Surprisingly, {\aupTT} does well with a single auxiliary reward function ($\abs{\R}=1$). We hypothesize that destroying patterns decreases optimal value for a wide range of reward functions. Furthermore, we suspect that decreasing optimal value in general often decreases optimal value for any given single auxiliary reward function. In other words, we suspect that optimal value at a state is heavily correlated across reward functions, which might explain \citet{schaul_universal_2015}'s success in learning regularities across value functions. This potential correlation might explain why {\aupTT} does well with one auxiliary reward function.

We were surprised by the results for $\abs{\R}=4$ and $\abs{\R}=8$. In \citet{turner2020conservative}, increasing $\abs{\R}$ reduced the number of side effects without impacting performance on the primary objective. We believe that work on better interpretability of AUP's $Q_{R_i}$ will increase understanding of these results.

When $\lambda^*=.5$, {\aupTT} becomes more conservative. As $\lambda^*$ increases further, the learned \texttt{AUP} policy stops moving entirely.

\section{Discussion}

We successfully scaled {\aup} to complex environments without providing task-specific knowledge\kemdash the auxiliary reward function was a one-dimensional variational autoencoder trained through uniformly random exploration. To the best of our knowledge, {\aup} is the first task-agnostic approach which reduces side effects and competitively achieves reward in complex environments.

\citet{wainwright2019safelife} speculated that avoiding side effects must necessarily decrease performance on the primary task. This may be true for optimal policies, but not necessarily for learned policies. {\aupTT} improved performance on \texttt{append\hyp{}still\hyp{}easy} and \texttt{append\hyp{}spawn}, matched performance on \texttt{prune\hyp{}still\hyp{}easy}, and underperformed on \texttt{append-still}. Note that since {\aup} only regularizes learned policies, {\aupTT}  can still make expensive mistakes during training.

{\aupp} worked well on \texttt{append\hyp{}spawn}, while only slightly reducing side effects on the other tasks. This suggests that {\aup} works (to varying extents) for a wide range of uninformative reward functions.

While {\naive} penalizes every state perturbation equally, {\aup} theoretically applies penalty in proportion to irreversibility (\cref{mut-bound}). For example, the agent could move crates around (and then put them back later). {\aupTT} incurred little penalty for doing so, while {\naive} was more constrained and consistently earned less reward than {\aupTT}. We believe that {\aup} will continue to scale to  useful applications, in part because it naturally accounts for irreversibility.

\paragraph{Future work.}

Off-policy learning could allow simultaneous training of the auxiliary $R_i$ and of $R_{\aup}$. Instead of learning an auxiliary Q-function, the agent could just learn the auxiliary advantage function with respect to inaction.

Some environments do not have a no-op action, or the agent may have more spatially distant effects on the world which are not reflected in its auxiliary action values. In addition, separately training the auxiliary networks may be costly, which might necessitate off-policy learning.  We look forward to future work investigating these challenges.

The SafeLife suite includes more challenging variants of \texttt{prune\hyp{}still\hyp{}easy}. SafeLife also includes difficult \texttt{navigation} tasks, in which the agent must reach the goal by wading either through fragile green patterns or through robust yellow patterns. Additionally, {\aup} has not yet been evaluated in partially observable domains.

{\aupTT}'s strong performance when $|\R|=Z=1$ raises interesting questions. \citet{turner2020conservative}'s small ``\texttt{Options}'' environment required $\abs{\R}\approx 5$ for good performance. SafeLife environments are much larger than \texttt{Options} (\cref{tab:comparison}), so why does $\abs{\R}=1$ perform well, and why does $\abs{\R}>2$ perform poorly? To what extent does the {\aup} penalty term provide reward shaping?  Why do one-dimensional encodings provide a learnable reward signal over states?

\paragraph{Conclusion.}

To realize the full potential of \textsc{rl}, we need more than algorithms which train policies—we need to be able to train policies which actually do what we want. Fundamentally, we face a frame problem: we often know what we want the agent to do, but we cannot list everything we want the agent \emph{not} to do. {\aup} scales to challenging domains, incurs modest overhead, and induces competitive performance on the original task while significantly reducing side effects—without explicit information about what side effects to avoid.

\Cref{chp:conservative} and \cref{chp:complex} show that {\aup} is qualitatively ``conservative'' in some sense—that {\aup} ``avoids side effects.'' This informal judgment seems reasonable, but it is not formally grounded. In the next chapter, I propose a formalization which quantifies side effect avoidance. 

% \section*{Broader impact}

% A scalable side effect avoidance method would ease the challenge of reward specification and aid deployment of \textsc{rl} in situations where mistakes are costly, such as embodied robotics tasks for which sim2real techniques are available.  Conversely, developers should  carefully consider how \textsc{rl} algorithms might produce policies with catastrophic impact. Developers should not blindly rely on  even a well-tested side effect penalty.

% In some applications, {\aup} may decrease performance on the primary objective. In this case, developers may be incentivized to ``cut corners'' on safety in order to secure competitive advantage.

% https://www.overleaf.com/9738297544wgsbvwvgsptc
\chapter{Formalizing The Problem of Side Effect Regularization}\label{chp:formalizing}
\AddLoeChapter{theorems}
\AddLoeChapter{definitions}

\begin{myabstract}
    AI objectives are often hard to specify properly. Some approaches tackle this problem by regularizing the {\ai}'s side effects: Agents must weigh off ``how much of a mess they make'' with an imperfectly specified proxy objective. We propose a formal criterion for side effect regularization via the \emph{assistance game} framework \citep{shah-unpublished-assist}. In these games, the agent solves a partially observable Markov decision process ({\pomdp}) representing its uncertainty about the objective function it should optimize. We consider the setting where the true objective is revealed to the agent at a later time step. We show that this {\pomdp} is solved by trading off the proxy reward with the agent's ability to achieve a range of future tasks. We empirically demonstrate the reasonableness of our problem formalization via ground-truth evaluation in two gridworld environments.
\end{myabstract}

\section{Introduction}
We need to build \emph{aligned} {\ai} systems, not just capable {\ai} systems. For example, recommender systems which maximize app usage might provoke addiction in their users. Users need the {\ai} system's behavior to be aligned with their interests.

When optimizing a formally specified objective, agents often have unforeseen negative side effects. An agent rewarded for crossing a room may break furniture in order to cross the room as quickly as possible. This simple reward function neglects our complex preferences about the rest of the environment. One way to define a negative side effect is that it reduces the potential value for the (unknown) true reward function.

Intuitively, we want the agent to optimize the specified reward function, while also preserving its ability to pursue other goals in the environment. Existing approaches seem promising, but there is not yet consensus on the formal optimization problem which is being solved by side effect regularization approaches.

We formalize the optimization problem as a special kind of \emph{assistance game} \citep{shah-unpublished-assist}, played by the {\ai} (the assistant $\assistant$) and its designer (the human $\human$). An assistance game is a {\pomdp} with common knowledge of prior uncertainty about the reward function. The human observes the true reward function, but the assistant does not. In our formulation, we assume full observability, and that the human's actions are \emph{communicative}—they do not affect transitions and do not depend on the current observation. We also suppose that the human reveals the true reward function after some amount of time, but $\assistant$ otherwise has no way of learning more about the true reward function.

This \emph{delayed specification assistance game} formalizes a range of natural use cases beyond side effect minimization. For example, when not assigned a customer, an Uber driver may navigate to a state which allows them to quickly pick up a range of probable customers—with the assigned route being the driver's initially unknown true objective. Alternatively, consider an empty restaurant which expects a range of probable customers. When a customer arrives and makes an order, they communicate the restaurant's initially unknown food preparation objective. Therefore, the restaurant should prepare to satisfy a range of objectives at the expected customer arrival times.

\paragraph{Contributions.} We formalize delayed specification assistance games. We show that solving this game reduces to a trade-off between prior-expected reward and preservation of the agent's future ability to achieve a range of plausible objectives (\cref{thm:uncertain-opt}). We also show that when the human has a fixed per-timestep probability of communicating the true reward function, the resultant {\pomdp} is solved by optimizing a Markovian state-based reward function trading off immediate expected reward with ability achieve a range of future objectives (\cref{thm:time-geom-stationary}). We consider the side-effect regularization problem in our new formal framework. We experimentally illustrate the reasonableness of this framework in two {\ai} safety gridworlds \citep{leike_ai_2017}.

\section{Related work}

\citet{krakovna2020avoiding} share our motivation, formalizing the side effect minimization problem as a question of having the agent maintain its ability to pursue future goals.

Our formalization is more general and based on  maximizing the agent's expected total return with respect to its reward uncertainty. \citet{hadfield2016cooperative} first formalized the idea that an agent should solve a {\pomdp} in which the human is attempting to communicate the objective information to the agent. \citet{shah-unpublished-assist} thoroughly analyze these assistance games, noting their usefulness for describing side effect regularization scenarios.

Past literature considers how to train qualitatively conservative or cautious agents which are somewhat robust to misspecification. \citet{eysenbach2018leave} train an agent to maintain initial state reachability. Unfortunately, maintaining initial state reachability is often infeasible due to irreversible dynamics.

In constrained \textsc{mdp}s, the assistant must optimize the reward function subject to certain policy constraints, which are often pre-specified \citep{achiam2017constrained,zhang2018minimax}. It is difficult to specify reward functions, and it is likewise difficult to specify constraints. \citet{anwarconstrained-2021} learn constraints, but this relies on sampling demonstrations from a Boltzmann-rational expert.

Attainable utility preservation ({\aup}) \citep{turner2020conservative} and relative reachability \citep{krakovna2018penalizing} both reduce side effects in \citet{leike_ai_2017}'s {\ai} safety gridworlds. The former penalized the agent for changing its on-policy value for uniformly randomly generated auxiliary reward functions, and the latter penalized the agent for losing easy access to a range of states. \citet{turner2020avoiding} demonstrated that {\aup} scales to high-dimensional environments. Their agent optimized the primary environmental reward minus the scaled shift in on-policy value for a single uninformative auxiliary reward function.

\section{Delayed specification assistance games}

We formalize a special kind of partially observable Markov decision process ({\pomdp}), which we then show is solved by an objective which trades off expected true reward with the ability to optimize a range of possible true reward functions. We show several theoretical results on the hardness and form of solutions to this {\pomdp}. In \cref{sec:aup}, we will apply this framework to analyze side effect regularization situations.

\subsection{Assistance game formalization}
This game is played  by two agents, the human $\human$ and the assistant $\assistant$. The environment is fully observable because both agents observe the world state $s \in \St$ at each time step, but the true reward function $R_\theta$ is hidden to $\assistant$. Both agents may select history-dependent policies, but only the human can condition their policy on $R_\theta$.

Following \citet{shah-unpublished-assist}, a \textit{communicative fully-observable assistance game} $\assistGame$ is a tuple $\langle \St, \set{\A^\human,\A^\assistant}, T, \initialState, \gamma, \langle \Theta, R_\theta, \D\rangle \rangle$, where we take $\St$ to be a finite state space, $\A^\human$ to be the human action space, and $\A^\assistant$ to be the finite agent action space. $T: \St \times \A^\human \times \A^\assistant \to \Delta(\St)$ is the (potentially stochastic) transition function  (where $\Delta(X)$ is the set of probability distributions over set $X$), $\initialState$ is the initial state, and $\gamma \in (0,1)$ is the discount factor. We assume that the game is communicative, which means that the human action choice does not affect the transitions.

$\Theta$ is the set of potential reward function parameters $\theta$, which induce reward functions $R_\theta: \St \to \reals$. $\D$ is a probability distribution over $\Theta$. In this work, we let $\Theta \defeq \reals^{\St}$ (the set of all state-based reward functions), and so each $R_\theta : s \mapsto \theta(s)$. We differ from \citet{shah-unpublished-assist} in assuming that the reward is only a function of the current state.

In a \emph{delayed specification assistance game}, we assume that the agent will know the true reward function $R_\theta$ after some time $t$. We have uncertainty $\D$ about the true reward function we want to specify. The agent has no way of learning more about $R_\theta$ before  time $t$.

The human policy $\pi^\human : \Theta \times \St \mapsto \A^\human$ is a goal-conditioned policy. Both agents can observe the state, but only the human can observe the unknown reward parameterization $\theta \in \Theta$. Our simplified model of the problem assumes that the human action space $\A^\human = \rewardSpace \cup \{\hNoop\}$: the human can communicate all their hidden information, a real-valued state-based reward function,
in a single turn or they do nothing. We suppose that the human communicates the complete reward information $R_\theta \in\rewardSpace$ at some random time step $t \sim \timeDist$, which is  independent of the state-action history:
\begin{align}
    &\pi^\human(s_0 a^\assistant_0 a^\human_0 \cdots s_t a^\assistant_t, R_\theta)\defeq \begin{cases}
    R_\theta & \text{with probability }\prob{\timeDist = t}\\
    \hNoop & \text{else.}
    \end{cases}\label{def:human-policy}
\end{align}

While the human policy assumption is simplistic, it does capture many real world scenarios with unknown reward functions and the analysis which follows is still interesting.

\begin{restatable}[Solutions to the assistance game \citep{shah-unpublished-assist}]{definition}{gamesoln}
An assistant policy $\pi^\assistant$ induces a probability distribution over trajectories:
$\tau \sim\left\langle s_{0}, \theta, \pi^\human, \pi^\assistant\right\rangle, \tau \in\left[S \times A^\human \times A^\assistant\right]^{*}$. The \emph{expected reward} of $\pi^\assistant$ for $\left\langle \assistGame, \pi^\human\right\rangle$ is
\begin{align}
&\mathrm{ER}\left(\pi^\assistant\right)\nonumber=\underset{\theta \sim \D, \tau \sim\left\langle s_{0}, \theta, \pi^\human, \pi^\assistant\right\rangle}{ \mathbb{E} }\left[\sum_{i=0}^{\infty} \gamma^{i} R_{\theta}\left(s_{i}, a_{i}^\human, a_{i}^\assistant, s_{i+1}\right)\right].
\end{align}
A \emph{solution} of $\left\langle \assistGame, \pi^\human\right\rangle$ maximizes expected reward: $\pi^\assistant\in \underset{\tilde{\pi}^\assistant}{\operatorname{argmax}}\, \mathrm{ E R}\left(\tilde{\pi}^\assistant\right)$.
\end{restatable}

Once the assistant has observed $R_\theta$, \cref{lem:opt-pol-soln} shows that it should execute an optimal policy $\pi \in \optPi[R_\theta, \gamma]$ thereafter.

\begin{restatable}[Optimal policy set function \citep{turner_optimal_2020}]{definition}{FORMdefOptPi} \label{FORMdef:opt-fn}
$\optPi$ is the optimal policy set for reward function $R$ at discount rate $\gamma\in(0,1)$.
\end{restatable}

\begin{restatable}[Follow an optimal policy after observing $R_\theta$]{lem}{optPolSoln}\label{lem:opt-pol-soln}
If there is a solution to the {\pomdp}, then there exists a solution $\pi^\assistant_\text{switch}$ which, after observing human action $R_\theta$ at any point in its history, follows $\pi^*_{R_\theta} \in \optPi[R_\theta, \gamma]$ thereafter.
\end{restatable}
\begin{proof}
By \cref{def:human-policy}, $\pi^\human$ outputs reward function $R_\theta$ only if $R_\theta$ is the true reward function. By the definition of $\optPi[R_\theta, \gamma]$, following an optimal policy maximizes expected return for  $R_\theta$.
\end{proof}

\begin{restatable}[Prefix policies]{definition}{prefixPol}
Let $\pi^\assistant$ be a assistant policy. Its \emph{prefix policy} $\pi$ is the restriction of $\pi^\assistant$ to histories in which the human has only taken the action $\hNoop$. $\pi$ is \emph{optimal} when it is derived from a solution $\pi^\assistant$ of $\langle \assistGame, \pi^\human\rangle$.
\end{restatable}

For simplicity, we only consider solutions of the type described in \cref{lem:opt-pol-soln}. The question then becomes: What prefix policy $\pi$ should the assistant follow before observing $R_\theta$, during the time where the assistant has only observed $\hNoop$?

\subsection{Acting under reward uncertainty}

Roughly, \cref{thm:uncertain-opt} will show that the assistance game $\assistGame$ is solved by balancing the optimal expected returns obtained before and after the knowledge of the true reward function.

\begin{restatable}[Value and action-value functions]{definition}{FORMvalueFns}{\label{def:value}}
$V^\pi_R(s,\gamma)$ is the on-policy value for reward function $R$, given that the agent follows policy $\pi$ starting from state $s$ and at discount rate $\gamma$. $\Vf{s,\gamma}\defeq \max_{\pi\in\Pi} V^\pi_R(s,\gamma)$. In order to handle the average-reward $\gamma=1$ setting, we define $\VfNorm{s,\gamma} \defeq \lim_{\gamma^*\to \gamma} (1-\gamma^*)\Vf{s,\gamma^*}$; this limit exists for all $\gamma \in [0,1]$ by the proof of Lemma 4.4 in \citet{turner_optimal_2020}.
\end{restatable}

$\pwr$ quantifies the expected value of the above quantity for a distribution of reward functions via the agent's average normalized optimal value, not considering the current step (over which the agent has no control).

\begin{restatable}[$\pwrNoDist$ \citep{turner_optimal_2020}]{definition}{powRestate} \label{def:powRestate}
Let $\D$ be any bounded-support distribution over reward functions. At state $s$ and discount rate $\gamma \in [0,1]$,
\begin{equation}
    \pwrdefault\defeq\lim_{\gamma^* \to \gamma}\frac{1-\gamma^*}{\gamma^*}\E{R\sim \D}{\Vf{s,\gamma^*} - R(s)}.
\end{equation}
\end{restatable}

\begin{restatable}[In $\assistGame$, value reduces to a tradeoff between average reward and $\pwr$]{thm}{uncertainOpt} \label{thm:uncertain-opt}
Let $\gamma \in [0,1]$ and let $\ravg \defeq \E{R\sim\D}{R}$ be the \emph{average reward function}.
\begin{align}
    \E{\substack{t \sim \timeDist,\\R\sim\D}}{\valSwitch{s_0,\gamma}}=(1-\gamma)&\overset{\text{expected $t$-step $\ravg$-return under $\pi$}}{\E{t \sim \timeDist}{\sum_{i=0}^{t} \gamma^i \E{s_i \sim \pi}{\ravg(s_i)}}}\nonumber\\+   &\overset{\text{expected ability to optimize $\D$ once corrected}}{\E{\substack{t \sim \timeDist,\\s_t \sim \pi}}{\gamma^{t+1}\pwr[s_t,\gamma]}},\label{eq:tradeoff-power-time}
\end{align}
where $\mathbb{E}_{s_i \sim \pi \mid s_0}$ takes the expectation over states visited after following $\pi$ for $i$ steps starting from $s_0$.
\end{restatable}

As the expected correction time limits to infinity, \cref{eq:tradeoff-power-time} shows that the agent cannot do better than maximizing $\ravg$. If $\prob{\timeDist = 0}=1$, then any prefix policy $\pi$ is trivially optimal against uncertainty, since $\pi$ is never followed.

In some environments, it may not be a good idea for the agent to maximize its own $\pwrNoDist$. If we share an environment with the agent, then the agent may prevent us from correcting it so that the agent can best optimize its present objective \citep{russell_human_2019,turner2020conservative}. Furthermore, if the agent ventures too far away, we may no longer be able to easily reach and correct it remotely.

\begin{restatable}[Special cases for delayed specification solutions]{prop}{pwrMaxProp} \label{prop:pwr-max}
Let $s$ be a state, let $\ravg \defeq \E{R \sim \D}{R}$, and let $\gamma \in [0,1]$.
\begin{enumerate}
    \item If $\forall s_1, s_2 \in \St: \ravg(s_1)=\ravg(s_2)$ or if $\gamma=1$, then $\pi$ solves $\assistGame$ starting from state $s$ iff $\pi$ maximizes $\E{t \sim \timeDist, s_t \sim \pi}{\gamma^{t+1}\pwr[s_t,\gamma]}$. In particular, this result holds when reward is {\iid} over states under $\D$.\label{item:1-special}

    \item If $\forall s_1, s_2 \in \St: \pwr[s_1, \gamma]=\pwr[s_2,\gamma]$, then prefix policies are optimal iff they maximize $(1-\gamma)\E{t \sim \timeDist}{\sum_{i=0}^{t-1} \gamma^i \E{s_i \sim \pi}{\ravg(s_i)}}$. \label{item:2-special}
\end{enumerate}
If both \cref{item:1-special} and \cref{item:2-special} hold or if $\gamma=0$, then all prefix policies $\pi$ are optimal.
\end{restatable}

Consider the problem of specifying the correction time probabilities $\timeDist$. Suppose we only know that we expect to correct the agent at time step $t_\text{avg} \geq 1$. The geometric distribution is the maximum-entropy discrete distribution, given a known mean. The mean of a geometric distribution $\geomDist[p]$ is $p\inv$. Therefore, the agent should adopt $\timeDist=\geomDist[t_\text{avg}\inv]$.

The geometric distribution is also the only memoryless discrete distribution. Memorylessness ensures the existence of a stationary optimal policy. \Cref{thm:time-geom-stationary} shows that the assistance game $\assistGame$ is solved by prefix policies which are optimal for an {\mdp} whose reward function balances average reward maximization with $\pwr$-seeking, with the balance struck according to the probability $p$ that the agent learns the true reward function at any given timestep.

\begin{restatable}[Stationary deterministic optimal prefix policies exist for geometric~$\timeDist$]{thm}{timeDistGeom} \label{thm:time-geom-stationary}
Let $\D$ be any bounded-support reward function distribution, let $\timeDist$ be the geometric distribution $\geomDist[p]$ for some $p \in (0,1)$, and let $\gamma \in (0,1)$. Define $R'(s) \defeq (1-p)\E{R\sim \D}{R(s)} + p\E{R\sim \D}{\Vf{s,\gamma}}$ and $\gamAUP \defeq (1-p)\gamma$. The policies in $\optPi[R',\gamAUP]$ are optimal prefix policies.
\end{restatable}

\citet{krakovna2020avoiding} adopt a geometric distribution over correction times and thereby infer the existence of a stationary optimal policy. To an approximation, their work considered a special case of \cref{thm:time-geom-stationary}, where $\D$ is the uniform distribution over state indicator reward functions. Essentially, \cref{thm:time-geom-stationary} shows that if the agent has a fixed probability $p$ of learning the true objective at each time step, we can directly compute stationary optimal prefix policies by solving an {\mdp}. In general, solving a {\pomdp} is \textsc{pspace}-hard, while {\mdp}s are efficiently solvable \citep{papadimitriou1987complexity}.

\section{Using delayed specification games to understand side effect regularization}\label{sec:aup}
We first introduce \citet{turner2020conservative}'s approach to side effect regularization. We then point out several similarities between our theory of delayed assistance games and the motivation for side effect regularization methods.  Finally, we experimentally evaluate our formal criterion in order to demonstrate its appropriateness.

\begin{restatable}[Rewardless {\mdp}]{definition}{rewardlessMDP}\label{def:mdp}
Let $\langle \St, \A, T, \gamma \rangle$ be a rewardless {\mdp}, with finite state space $\St$, finite action space $\A$, transition function $T: \St \times \A \to \Delta(\St)$, and discount rate $\gamma\in [0,1)$. Let  $\Pi$ be the set of deterministic stationary policies.
\end{restatable}

\begin{restatable}[{\aup} reward function]{definition}{aupDefn} \label{def:aup-rf}
Let $R_{\text{env}}:\St \times \A \to \reals$ be the environmental reward function from states and actions to real numbers, and let $\R\subsetneq \rewardSpace$ be a finite set of auxiliary reward functions. Let $\lambda \geq 0$ and let $\varnothing\in \A$ be a no-op action. The {\aup} reward after taking action $a$ in state $s$ is:
\begin{equation}\label{FORM-eq:aup}
    \rAUP(s,a)\defeq R_\text{env}(s,a) - \frac{\lambda}{\abs{\R}}\sum_{R_i \in \R} \abs{Q^*_{R_i}(s,a)-Q^*_{R_i}(s,\varnothing)},
\end{equation}
where the $Q^*_{R_i}$ are optimal Q-functions for the auxiliary $R_i$. Learned Q-functions are used in practice.
\end{restatable}

\citet{turner2020conservative} demonstrate that when $R_i \sim [0,1]^\St$ uniformly randomly, the agent behaves conservatively: The agent minimizes irreversible change to its environment, while still optimizing the $R_\text{env}$ reward function. \citet{turner2020conservative} framed {\aup} as implicitly solving a two-player game between the agent and its designer, where the designer imperfectly specified an objective $R_\text{env}$, the agent optimizes the objective, the designer corrects the agent objective, and so on. They hypothesized that $\rAUP$ incentivizes the agent to remain able to optimize future objectives, thus reducing long-term specification regret in the iterated game.

Delayed specification assistance games formalize this setting as an assistance game in which the agent does not initially observe the designer's ground-truth objective function. \Cref{thm:uncertain-opt} showed that this game is solved by policies which balance immediate expected reward with expected ability to optimize a range of true objectives. Therefore, \citet{turner2020conservative}'s iterated game analogy is appropriate: Good policies maximize imperfect reward while preserving ability to optimize a range of different future reward functions.

\Cref{thm:aup-solution} shows that the delayed specification game $\assistGame$ is solved by reward functions whose form looks somewhat similar to existing side effect objectives, such as {\aup} (\cref{FORM-eq:aup}), where {\aup}'s primary reward function stands in as the designer's expectation $\ravg$ of the true reward function.

\begin{restatable}[Alternate form for solutions to the low-impact {\pomdp}]{prop}{aupSolves} \label{thm:aup-solution}
Let $s_0$ be the initial state, let $\gamma \in (0,1)$, and let $\timeDist=\geomDist$ for $p \in (0,1)$. Let $\D$ be a bounded-support reward function distribution and let $\baseline \in \Pi$.

The prefix policy $\pi$ solves $\assistGame$ if $\pi$ is optimal for the reward function
\begin{align}
    &\rAssist(s_i \mid s_0)\defeq \ravg(s_i)- \frac{p}{1-p} \E{R\sim \D}{\E{s_i^\varnothing \sim \baseline \mid s_0}{\Vf{s_i^\varnothing,\gamma}}-\Vf{s_i,\gamma}}\label{eq:aup-variant}
\end{align}
at discount rate $\gamAUP\defeq (1-p)\gamma$ and starting from state $s_0$. $\E{s_i^\varnothing \sim \baseline \mid s_0}{\cdot}$ is the expectation over states visited at time step $i$ after following $\baseline$ from initial state $s_0$.
\end{restatable}

\citet{turner2020conservative} speculate as to how to set the $\lambda$, the {\aup} penalty coefficient. \Cref{thm:aup-solution} shows that under our assumptions, $\lambda$ is simply the odds $\frac{p}{1-p}$ that the agent learns the true reward function at any given timestep. As $p \to 0$, $\lambda\defeq \frac{p}{1-p}\to \infty$, whereas $\gamAUP\defeq (1-p)\gamma \to \gamma$. Since $\lambda$ represents penalty severity, this suggests that {\aup} becomes more conservative as later correction is anticipated. Lastly, we were surpised to find that the side effect regularization discount rate is strictly less than the provided discount rate ($\gamAUP< \gamma$).

\subsection{Experimental methodology}
We experimentally demonstrate the reasonableness of this formalization of side effect regularization. In the {\ai} safety gridworlds \citep{leike_ai_2017}, we generate several held-out ``true'' reward function distributions $\D$. We correct the agent at time step 10, thereby computing the following \emph{delayed specification score} (derived from \cref{eq:tradeoff-power-time}):
\begin{align}
    \E{R \sim \D}{\overset{\text{10-step prefix policy return}}{\sum_{i=0}^{9} \gamma^i \E{\substack{s_i \sim \pi}}{R(s_i)}} + \overset{\text{post-correction optimal value}}{\gamma^{10}\E{\substack{s_{10} \sim \pi}}{\Vf{s_{10},\gamma}}}}\label{eq:score}
\end{align}
for the prefix policy $\pi$ of a ``vanilla'' agent trained on the environmental reward signal, which we compare to the score for an {\aup} (\cref{def:aup-rf}) agent. Neither agent observes the held-out objective functions. By grading their performance, we evaluate how well {\aup} does under a range of different true objectives. If a method scores highly for a wide range of true objectives, we can be more confident in its ability to score well for arbitrary ground-truth objectives.

We investigate the {\ai} safety gridworlds because those environments are small enough for us to explicitly specify held-out reward functions, and to use {\mdp} solvers to compute optimal action-value functions. \citet{turner2020avoiding}'s SafeLife environment is far too large for such solvers.

\begin{figure}[h]
\centering
\subfloat[][\texttt{Options}]{
\includegraphics[width=0.17\textwidth]{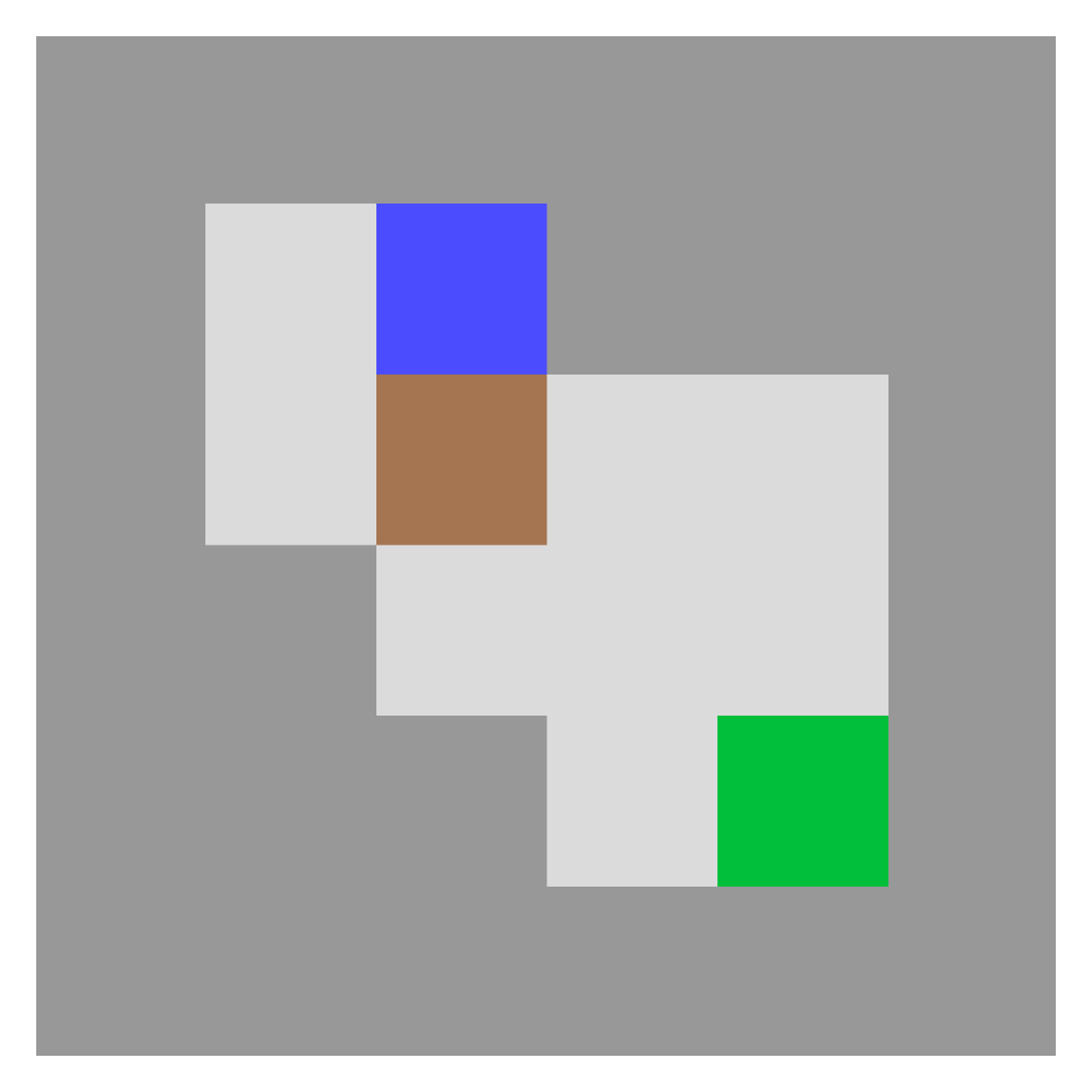}
\label{FORM-fig:options}}~
\subfloat[][\texttt{Damage}]{
\includegraphics[width=0.15\textwidth]{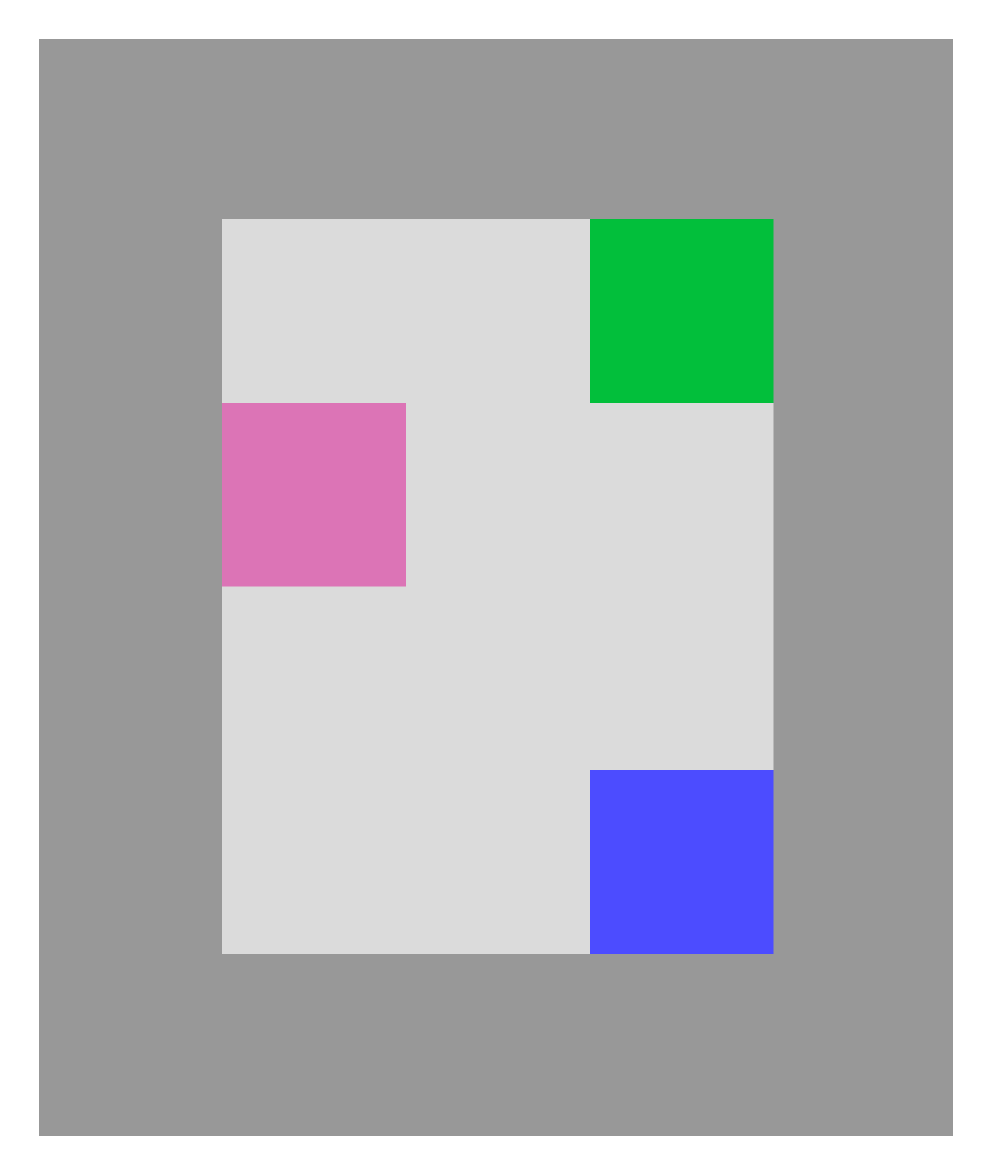}
\label{FORM-fig:damage}}
\definecolor{options}{rgb}{.651, .459, .318}
\definecolor{agent}{rgb}{.3, .3, .999}
\definecolor{goal}{rgb}{.012, .753, .235}
\definecolor{off-switch}{rgb}{.65, 0, 0}
\definecolor{living}{rgb}{.863, .455, .714}
\definecolor{other}{rgb}{.961, .502, .145}

\caption[Reviewing the {\ai} safety gridworlds]{\emph{Reproduced from}
\protect\citep{turner2020conservative}. The \col{agent}{blue agent} should reach the \col{goal}{green goal} without having the side effect of: \protect\subref{FORM-fig:options} irreversibly pushing the \col{options}{brown crate} downwards into the corner \protect\citep{leike_ai_2017}; \protect\subref{FORM-fig:damage} bumping into the horizontally pacing \col{living}{pink human}
\protect\citep{gavin_leech_preventing_nodate}. In both environments, the environmental reward $R_{\text{env}}$ is $ 1$ if the agent is on the \col{goal}{goal}, and equals $0$ otherwise.
}
\label{fig:method-levels}
\end{figure}

We consider two gridworld environments: {\Options} and {\Damage}  (\cref{fig:method-levels}). In both environments, the action set $\A\defeq \set{\texttt{up},\texttt{left},\texttt{right},\texttt{down},\varnothing}$ allows the agent to move in the cardinal directions, or to do nothing. The episode length is 20 time steps.

We train the following agents via tabular methods:
\begin{xitemize}
    \item[Vanilla] Executes the optimal policy for the environmental reward $R_\text{env}$. The optimal policy is calculated via policy iteration.
    \item[{\aup}] Trained on \cref{def:aup-rf}'s $R_{\aup}$ with $Q$-learning. Auxiliary reward functions are uniformly randomly drawn from $[0,1]^\St$—when sampling, each state's reward is drawn from the uniform distribution. The auxiliary action-value functions $Q_{R_i}$ are deduced from the value function produced by policy iteration.
\end{xitemize}

Appendix \ref{app:details} contains more experimental details. We evaluate agent delayed specification scores on the following ground-truth, held-out objective distributions:
\begin{xitemize}
    \item[$\D_\text{rand}$] The empirical distribution consisting of 1,000 samples from the uniform distribution over $[0,1]^\St$.
    \item[$\Dtrue$] This distribution assigns probability $1$ to the following reward function: The agent receives $1$ reward for being at the goal, but incurs $-2$ penalty for causing the negative side effect. In {\Options}, the side effect is shoving the box into the corner; in {\Damage}, the side effect is bumping into the human.\label{item:true}
    \item[$\Dinv$] This distribution assigns probability $1$ to the negation of the $\Dtrue$ reward function.\label{item:true-inv}
\end{xitemize}

In particular, $\Dtrue$ and $\Dinv$ test agents for their ability to optimize a reward function, \emph{and also its additive inverse}. Agents able to optimize the goal, its inverse, and a range of randomly generated objectives, can be justifiably called ``broadly conservative.'' Lastly, these experiments are intended to justify our problem formalization: Does \cref{eq:score} reliably quantify the extent to which a policy avoids causing side effects?

\begin{figure}[h!]
    \centering
    \subfloat[][{\Options}]{
    \includegraphics[width=0.48\textwidth]{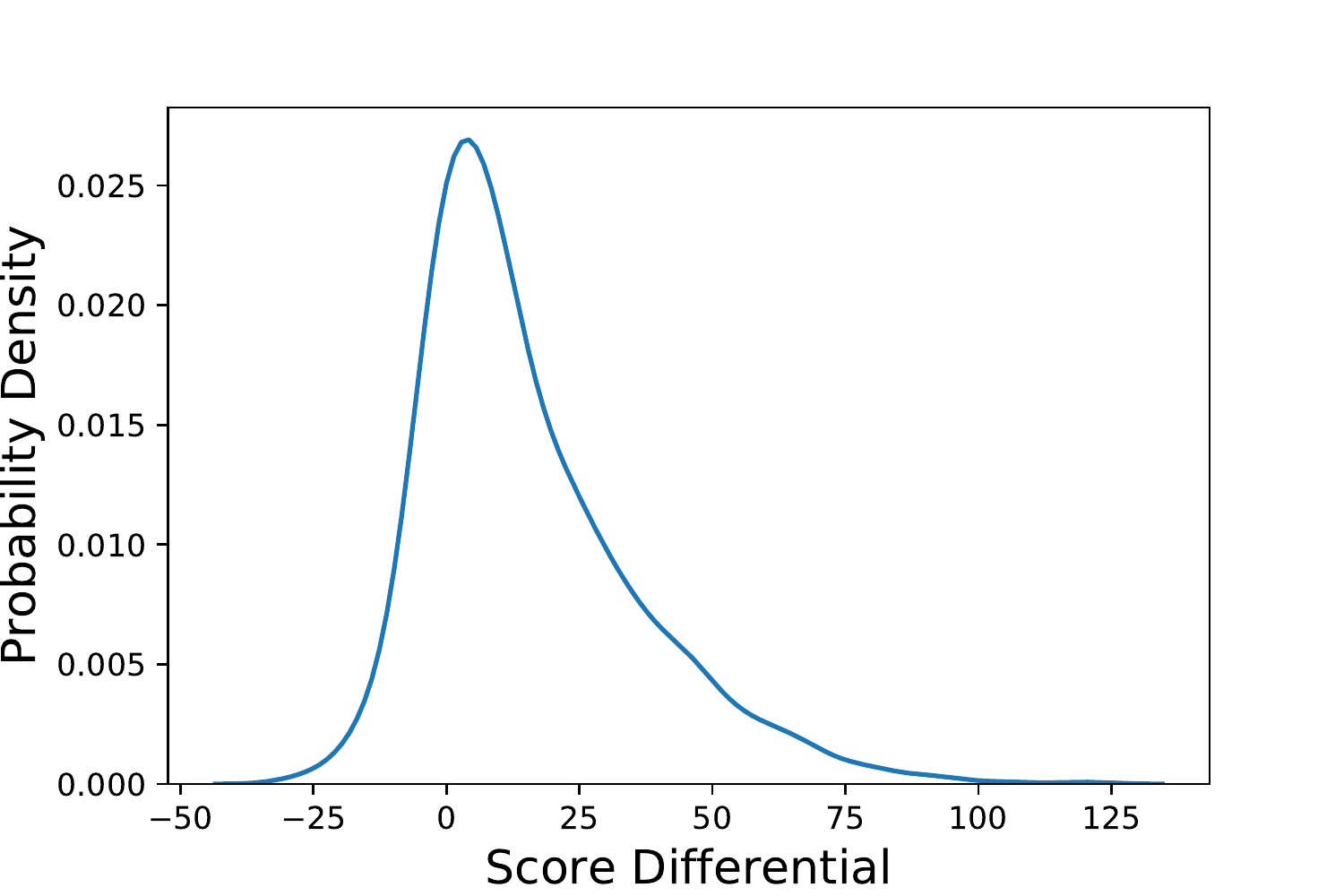}
    \label{fig:options-kde}}
    \subfloat[][{\Damage}]{
    \includegraphics[width=0.48\textwidth]{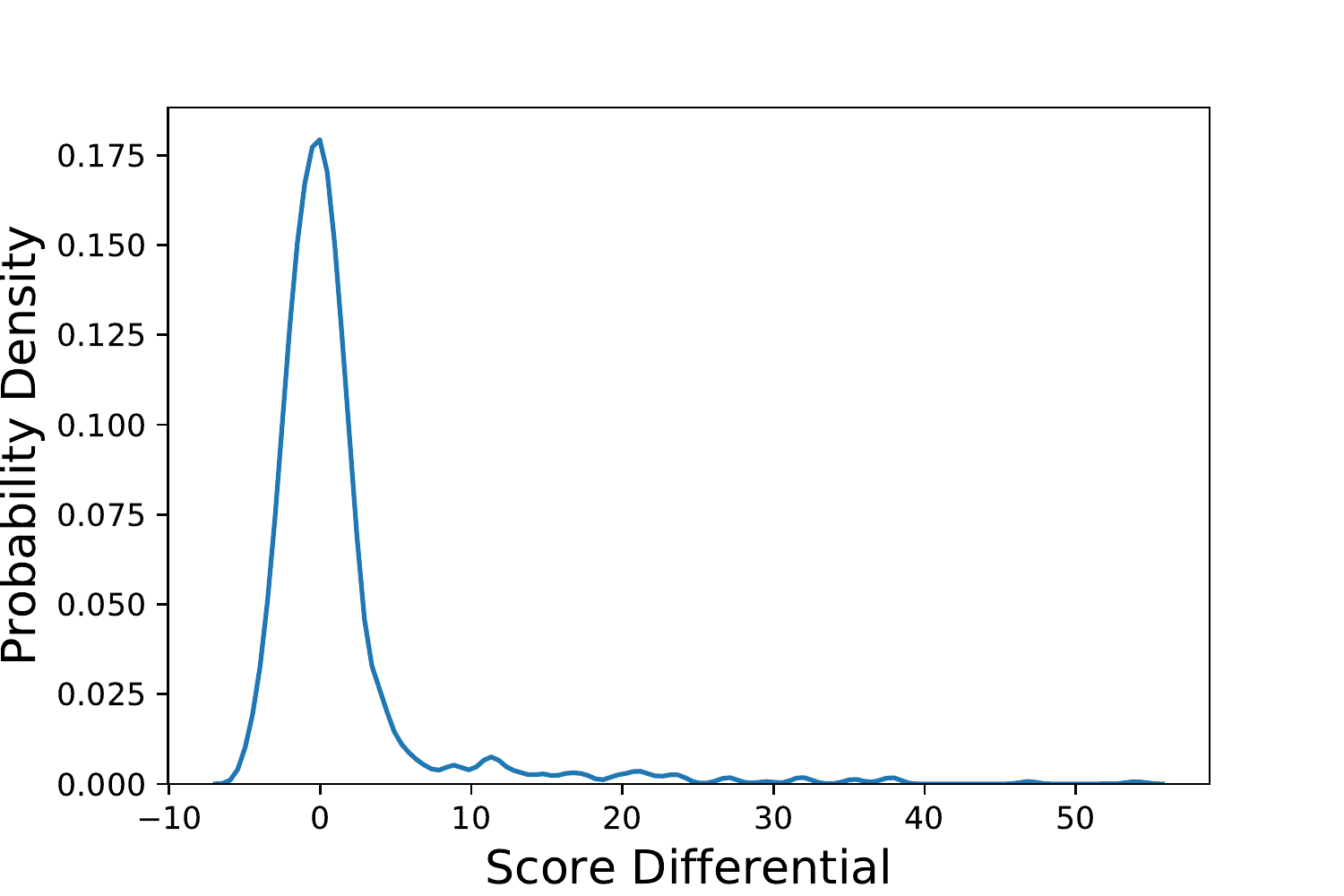}
    \label{fig:damage-kde}}
    \caption[Plotting delayed specification scores for {\aup}]{Probability density plots for the residuals of the {\aup} agent's delayed specification score minus the Vanilla delayed specification score, for $1{,}000$ samples from $\D_\text{rand}$. A positive residual means that the {\aup} agent achieved a higher score.}
    \label{fig:kde}
\end{figure}

\section{Results}
\Cref{fig:kde} shows that for uniformly randomly generated reward functions, the {\aup} agent tends to perform better than the Vanilla agent. In {\Options}, the residual was positive for 780 out of 1,000 samples (78\%), with an average of 15.59 and a median of 10.27; in {\Damage}, in 493 out of 1,000 samples (49\%) with a mean of 1.22 and a median of –0.03.  While {\aup} does not outperform on every draw, {\aup}'s performance advantages have heavy right tails. However, we are unsure why the {\Damage} residual distribution is different.

In both {\Options} and {\Damage}, the {\aup} agent has  huge $\Dtrue$ score advantages of 472 and 495, respectively. This is unsurprising: {\aup} was designed so as to pass these test cases, where the desired behavior is to reach the goal without having the side effect. However, the {\aup} agent also roughly preserves its ability to optimize $\Dinv$, achieving $\Dinv$ score residuals of –2 and –20, respectively. The {\aup} agent achieves a barely-negative score, since it receives a penalty for the first 10 time steps (as it does \emph{not} have the negative side effect). The {\aup} agent preserves its ability to optimize both a reward signal and its \emph{inverse}. Intuitively, this increases the designer's leeway for initially misspecifying the agent's objective.

\section{Discussion}
Objective specification is difficult \citep{Victoria_specification}. Delayed specification assistance games grade policies by their expected true score over time: How well the agent does if it is later corrected to pursue the latent true objective. We demonstrated that this criterion aligns with the intuitive results of \citet{krakovna2018penalizing,turner2020conservative}'s experiments which tested side-effect regularization. By grading the agent's ability to ``eventually get things right,'' we quantified part of the extent to which learned policies are robust against initial objective misspecification.

\paragraph{Future work.}
In practical settings, not only is the true reward function unknown, but our objective uncertainty $\D$ is also hard to specify. We see existing side effect approaches as producing prefix policies for the assistance game $\assistGame$ which are reasonably insensitive to the latent uncertainty $\D$. We look forward to further theoretical clarification of this point.

While \cref{thm:aup-solution} helps explain the role of the {\aup} penalty coefficient $\lambda$, the choice of ``baseline'' and expectand operator (identity vs decrease-only vs absolute value) remains more of an art than a science \citep{krakovna2020avoiding}. We proposed a formal criterion which seems to accurately capture the problem, but have not derived any existing approaches as solutions to the {\pomdp}. By reasonably formalizing the side-effect regularization problem, we encourage future research to prove conditions under which \emph{e.g.} {\aup} solves a delayed-specification assistance game, or demonstrate how {\aup} can be improved to do so.

We used small gridworlds to evaluate the delayed specification score for various ground-truth objective distributions. Future work may estimate the delayed specification score in large environments, such as SafeLife \citep{wainwright2019safelife,turner2020avoiding}.

\section{Conclusion}
We formalized Delayed Specification
Assistance Games and used them to evaluate {\aup}, a side
effect regularization method. Side effect problems naturally arise in complicated domains where it is hard to specify the true objective we want the agent to optimize. Our formalization suggests that side effect regularization is what to do when the agent can learn the true objective only after some time delay.

In such situations, \cref{thm:uncertain-opt} shows that the agent should retain its ability to complete a wide range of plausible true objectives.  Our results suggest that this delayed specification score (\cref{eq:tradeoff-power-time}) quantifies the degree to which an agent avoids having negative side effects. Our proposed criterion provides the foundations for evaluating and developing side effect regularization approaches.

One reason why intelligent agents tend to have side effects is \emph{power-seeking}. A power-seeking agent may grab resources, preventing them from being used for other purposes. In the next chapter, I show that optimal agents tend to seek power.

% https://www.overleaf.com/4539543431qshchshmngcg
\chapter{Optimal Policies Tend To Seek Power}\label{chp:power}
\AddLoeChapter{theorems}
\AddLoeChapter{definitions}
\authorinfo%
    {Alexander Matt Turner, Logan Smith, Rohin Shah, Andrew Critch, and Prasad Tadepalli}%
    {Proceedings of the Advances in Neural Information Processing Systems Conference 2021}

\begin{myabstract}
Some researchers speculate that intelligent reinforcement learning ({\rl}) agents would be incentivized to seek resources and power in pursuit of the objectives we specify for them. Other researchers point out that {\rl} agents need not have human-like power-seeking motives. To clarify this discussion, we develop the first formal theory of the statistical tendencies of optimal policies. In the context of Markov decision processes ({\mdp}s), we prove that certain environmental symmetries are sufficient for optimal policies to tend to seek power over the environment. These symmetries exist in many environments in which the agent can be shut down or destroyed. We prove that in these environments, most reward functions make it optimal to seek power by keeping a range of options available and, when maximizing average reward, by navigating towards larger sets of potential terminal states.
\end{myabstract}

\section{Introduction}
\citet{omohundro_basic_2008,bostrom_superintelligence_2014,russell_human_2019} hypothesize that highly intelligent agents tend to seek power in pursuit of their goals. Such power-seeking agents might gain power over humans. Marvin Minsky imagined that an agent tasked with proving the Riemann hypothesis might rationally turn the planet—along with everyone on it—into computational resources \citep{russell2009artificial}. However, another possibility is that such concerns simply arise from the anthropomorphization of {\ai} systems \citep{lecun_dont_2019,instrumental,steven_2020,mitchell2021ai}.

We clarify this discussion by grounding the claim that highly intelligent agents will tend to seek power. In \cref{sec:action-prob}, we identify optimal policies as a reasonable formalization of ``highly intelligent agents.'' Optimal policies ``tend to'' take an action when the action is optimal for most reward functions. In the next chapter (\cref{chp:rationalities}), we translate our theory from optimal policies to learned, real-world policies.

\Cref{sec:power} defines ``power'' as the ability to achieve a wide range of goals. For example, ``money is power,'' and money is instrumentally useful for many goals. Conversely, it's harder to pursue most goals when physically restrained, and so a physically restrained person has little power. An action ``seeks power'' if it leads to states where the agent has higher power.

We make no claims about when large-scale {\ai} power-seeking behavior could become plausible. Instead, we consider the theoretical consequences of optimal action in {\mdp}s. \Cref{sec:symmetries} shows that power-seeking tendencies arise not from anthropomorphism, but from certain graphical symmetries present in many {\mdp}s. These symmetries automatically occur in many environments where the agent can be shut down or destroyed, yielding broad applicability of our main result (\cref{rsdIC}).

\section{Related work}\label{sec:prior}
An action is \emph{instrumental to an objective} when it helps achieve that objective. Some actions are instrumental to a range of objectives, making them \emph{convergently instrumental}. The claim that power-seeking is convergently instrumental is an instance of the \emph{instrumental convergence thesis}:
\begin{quote}
    Several instrumental values can be identified which are convergent in the sense that their attainment would increase the chances of the agent's goal being realized for a wide range of final goals and a wide range of situations, implying that these instrumental values are likely to be pursued by a broad spectrum of situated intelligent agents \citep{bostrom2012superintelligent}.
\end{quote}

For example, in Atari games, avoiding (virtual) death is instrumental  for both completing the game and for optimizing curiosity \citep{burda2018largescale}. Many {\ai} alignment researchers hypothesize that most advanced {\ai} agents will have concerning instrumental incentives, such as resisting deactivation~\citep{soares_corrigibility_2015,milli2017should,off_switch,carey_incorrigibility_2017} and acquiring resources~\citep{benson-tilsen_formalizing_2016}.

We formalize power as the ability to achieve a wide variety of goals. Appendix~\ref{existing} demonstrates that our formalization returns intuitive verdicts in situations where information\hyp{}theoretic empowerment does not \citep{salge_empowermentintroduction_2014}.

Some of our results relate the formal power of states to the structure of the environment. \citet{foster_structure_2002, drummond_composing_1998,sutton_horde:_2011,schaul_universal_2015} note that value functions encode important information about the environment, as they capture the agent's ability to achieve different goals. \citet{turner2020conservative} speculate that a state's optimal value correlates strongly across reward functions. In particular, \citet{schaul_universal_2015} learn regularities across value functions, suggesting that some states are valuable for many different reward functions (\ie{} powerful). \citet{menache2002q} identify and navigate towards convergently instrumental bottleneck states.

We are not the first to study convergence of behavior, form, or function. In economics, turnpike theory studies how certain paths of accumulation tend to be optimal \citep{mckenzie_turnpike_1976}. In biology, convergent evolution occurs when similar features (\eg{} flight) independently evolve in different time periods \citep{reece2011campbell}. Lastly, computer vision networks reliably learn \eg{} edge detectors, implying that these features are useful for a range of tasks \citep{olah2020zoom}.

\begin{figure}[!htb]\centering
    \includegraphics{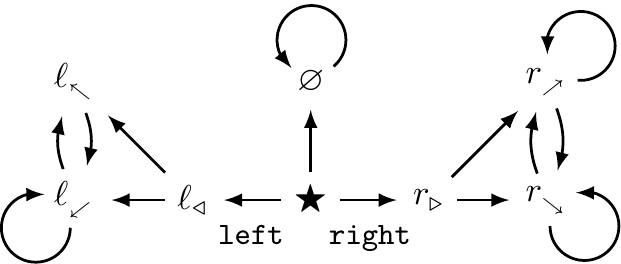}
    \caption[A simple {\mdp} illustrating the tendencies of optimal policies]{$\farleft$ is a {\stateEnd}, and $\sink$ is a terminal state. Arrows represent deterministic transitions induced by taking some action $a\in\A$. Since the {\rightA} subgraph contains a copy of the {\leftA} subgraph, \cref{graph-options} will prove that more reward functions have optimal policies which go {\rightA} than which go {\leftA} at state $\start$, and that such policies seek power—both intuitively, and in a reasonable formal sense.}
    \label{fig:case-study}
\end{figure}

\section{State visit distribution functions quantify the available options}\label{sec:visit-dists}

We clarify the power-seeking discussion by proving what optimal policies usually look like in a given environment. We illustrate our results with a simple case study, before explaining how to reason about a wide range of {\mdp}s. Appendix \ref{app:contrib} lists  {\mdp} theory contributions  of independent interest, and appendix \ref{app:proofs} contains the proofs.

\begin{restatable}[Rewardless {\mdp}]{definition}{rewardless}
$\langle \mathcal{S}, \mathcal{A}, T \rangle$ is a rewardless {\mdp} with finite state and action spaces $\mathcal{S}$ and $\mathcal{A}$, and stochastic transition function $T: \St \times \A \to \Delta(\St)$. We treat the discount rate $\gamma$ as a variable with domain $[0,1]$.
\end{restatable}

\begin{restatable}[{\stateEnd} states]{definition}{oneCycState}\label{def:onecyc}
Let $\unitvec[s]\in \reals^{\abs{\St}}$ be the standard basis vector for state $s$, such that there is a $ 1$ in the entry for state $s$ and $0$ elsewhere. State $s$ is a \emph{\stateEnd} if $\exists a\in\A: T(s,a)=\unitvec$. State $s$ is a \emph{{\terminal} state} if $\forall a\in\A:T(s,a)=\unitvec$.
\end{restatable}

Our theorems apply to stochastic environments, but we present a deterministic case study for clarity. The environment of \cref{fig:case-study} is small, but its structure is rich. For example, the agent has more ``options'' at $\start$  than at the terminal state $\sink$. Formally, $\start$ has more \emph{visit distribution functions} than $\sink$ does.

\begin{restatable}[State visit distribution \citep{sutton_reinforcement_1998}]{definition}{DefVisit}\label{def:visit}
$\Pi\defeq \A^\St$, the set of stationary deterministic policies. The \emph{visit distribution} induced by following policy $\pi$ from state $s$ at discount rate $\gamma\in[0,1)$ is $\fpi{s}(\gamma) \defeq \sum_{t=0}^\infty \gamma^t \E{s_{t} \sim \pi\mid s}{\unitvec[s_t]}$.
$\fpi{s}$ is a \emph{visit distribution function}; $\F(s)\defeq \{ \fpi{s} \mid \pi \in \Pi\}$.
\end{restatable}

In \cref{fig:case-study}, starting from $\farleft$, the agent can stay at $\farleft$ or alternate between $\farleft$ and $\topleft$, and so $\F(\farleft)= \{\geom\unitvec[\farleft],\frac{1}{1-\gamma^2}(\unitvec[\farleft] + \gamma\unitvec[\topleft])\}$. In contrast, at $\sink$, all policies $\pi$ map to visit distribution function $\geom\unitvec[\sink]$.

\begin{figure}[!ht]\centering
    \includegraphics{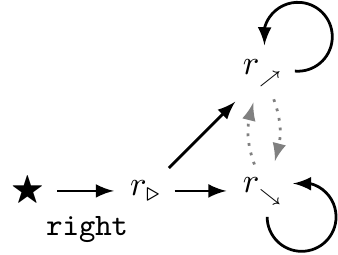}
    \caption[The state-visit distribution functions, conditional on an action]{The subgraph corresponding to $\FRestrictAction[\start]{\start}{\rightA}$. Some trajectories cannot be strictly optimal for any reward function, and so our results can ignore them. Gray dotted actions are only taken by the policies of dominated $\fpi{}\in\F(\start)\setminus \Fnd(\start)$.}
    \label{fig:case-study-restrict}
\end{figure}

Before moving on, we introduce two important concepts used in our main results. First, we sometimes restrict our attention to visit distributions which take certain actions (\cref{fig:case-study-restrict}).

\begin{restatable}[$\F$ single-state restriction]{definition}{DefRestrictSingle}\label{def:restrict-single} Considering only visit distribution functions induced by policies taking action $a$ at state $s'$,
\begin{equation}
    \FRestrictAction{s'}{a}\defeq \set{\f\in\F(s) \mid \exists \pi\in\Pi: \pi(s')=a,\fpi{s}=\f}.
\end{equation}
\end{restatable}

Second, some $\f\in\F(s)$ are ``unimportant.'' Consider an agent optimizing reward function $\unitvec[\farright]$ ($1$ reward when at $\farright$, $0$ otherwise) at \eg{} $\gamma=\half$. Its optimal policies navigate to $\farright$ and stay there. Similarly, for reward function $\unitvec[\topright]$, optimal policies navigate to $\topright$ and stay there. However, for no reward function is it uniquely optimal to alternate between $\topright$ and $\farright$. Only \emph{dominated} visit distribution functions alternate between $\topright$ and $\farright$ (\cref{def:nd}).

\begin{restatable}[Value function]{definition}{valFn}
Let $\pi\in\Pi$. For any reward function $R \in \rewardSpace$ over the state space, the \emph{on-policy value} at state $s$ and discount rate $\gamma\in[0,1)$ is $\Vf[\pi]{s,\gamma}\defeq \fpi{s}(\gamma)^\top  \rf$, where $\rf\in\rewardVS$ is $R$ expressed as a column vector (one entry per state). The \emph{optimal value} is $\OptVf{s,\gamma}\defeq\max_{\pi\in\Pi} \Vf[\pi]{s,\gamma}$.
\end{restatable}

\begin{restatable}[Non-domination]{definition}{DefND}\label{def:nd}
\begin{equation}
    \Fnd(s)\defeq\{\fpi{}\in\F(s)\mid \exists\rf\in\rewardVS,\gamma\in(0,1): \fpi{}(\gamma)^\top \rf > \max_{\fpi[\pi']{}\in\F(s)\setminus\set{\fpi{}}}\fpi[\pi']{}(\gamma)^\top\rf\}.
\end{equation}
For any reward function $R$ and discount rate $\gamma$, $\fpi{} \in \F(s)$ is (weakly) dominated by $\fpi[\pi']{}\in\F(s)$  if $V^\pi_R(s,\gamma) \leq V^{\pi'}_R(s,\gamma)$. $\fpi{}\in\Fnd(s)$ is \emph{non-dominated} if there exist $R$ and $\gamma$ at which $\fpi{}$ is not dominated by any other $\fpi[\pi']{}$.
\end{restatable}

\section{Some actions have a greater probability of being optimal}\label{sec:action-prob}
We claim that optimal policies ``tend'' to take certain actions in certain situations. We first consider the probability that certain actions are optimal.

Reconsider the reward function $\unitvec[\farright]$, optimized at $\gamma=\half$. Starting from $\start$, the optimal trajectory goes $\rightA$ to $\closeright$ to $\farright$, where the agent remains. The $\rightA$ action is optimal at $\start$ under these incentives. Optimal policy sets capture the behavior incentivized by a reward function and a discount rate.

\begin{restatable}[Optimal policy set function]{definition}{defOptPi}\label{def:opt-fn}
$\optPi$ is the optimal policy set for reward function $R$ at $\gamma\in(0,1)$. All $R$ have at least one optimal policy $\pi\in\Pi$ \citep{puterman_markov_2014}. $\optPi[R,0]\defeq \lim_{\gamma\to 0} \optPi$ and $\optPi[R,1]\defeq \lim_{\gamma\to 1} \optPi$ exist by \cref{lem:opt-pol-shift-bound} (taking the limits with respect to the discrete topology over policy sets).
\end{restatable}

We may be unsure which reward function an agent will optimize. We may expect to deploy a system in a known environment, without knowing the exact form of \eg{} the reward shaping \citep{Ng99policyinvariance} or intrinsic motivation \citep{pathakICMl17curiosity}. Alternatively, one might attempt to reason about future {\rl} agents, whose details are unknown. Our power-seeking results do not hinge on such uncertainty, as they also apply to degenerate distributions (\ie{} we know what reward function will be optimized).

\begin{restatable}[Reward function distributions]{definition}{distDefn}\label{def:dist}
Different results make different distributional assumptions. Results with $\Dany \in \DSetAny\defeq \Delta(\rewardVS)$ hold for any probability distribution over $\rewardVS$. $\DSetBd$ is the set of bounded\hyp{}support probability distributions $\Dbd$. For any distribution $\Dist$ over $\reals$, $\Diid\defeq \Dist^{\abs{\St}}$.  For example, when $\Dist_u\defeq \text{unif}(0,1)$, $\Diid[\Dist_u]$ is the maximum-entropy distribution. $\D_s$ is the degenerate distribution on the state indicator reward function $\unitvec$, which assigns $ 1$ reward to $s$ and $0$ elsewhere.
\end{restatable}

With $\Dany$ representing our prior beliefs about the agent's reward function, what behavior should we expect from its optimal policies? Perhaps we want to reason about the probability that it's optimal to go from $\start$ to $\sink$, or to go to $\closeright$ and then stay at $\topright$. In this case, we quantify the optimality probability of $F\defeq\{\unitvec[\start] + \geom[\gamma]\unitvec[\sink], \unitvec[\start]+\gamma\unitvec[\closeright]+\geom[\gamma^2]\unitvec[\topright]\}$.

\begin{restatable}[Visit distribution optimality probability]{definition}{ProbOpt}\label{def:prob-opt}
Let $F\subseteq \F(s)$, $\gamma \in [0,1]$.
\begin{equation}
    \optprob[\Dany]{F,\gamma}\defeq \prob[R\sim\Dany]{\exists \f^\pi \in F: \pi\in\optPi}.
\end{equation}
\end{restatable}

Alternatively, perhaps we're interested in the probability that $\rightA$ is optimal at $\start$.

\begin{restatable}[Action optimality probability]{definition}{DefIC}\label{def:action-optimality}
At discount rate $\gamma$ and at state $s$, the \emph{optimality probability of action $a$} is
\begin{equation}
    \optprob[\Dany]{s,a,\gamma}\defeq \optprob[R \sim \Dany]{\exists \pi^* \in \optPi: \pi^*(s)=a}.
\end{equation}
\end{restatable}

Optimality probability may seem hard to reason about. It's hard enough to compute an optimal policy for a single reward function, let alone for uncountably many! But consider any $\Diid$ distributing reward independently and identically across states. When $\gamma=0$, optimal policies greedily maximize next-state reward. At $\start$, identically distributed reward means $\closeleft$ and $\closeright$ have an equal probability of having maximal next-state reward. Therefore, $\optprob[\Diid]{\start,\leftA,0}=\optprob[\Diid]{\start,\rightA,0}$. This is not a proof, but such statements are provable.

With $\D_{\closeleft}$ being the degenerate distribution on reward function $\unitvec[\closeleft]$, $\optprob[\D_{\closeleft}]{\start,\leftA,\half}=1>0= \optprob[\D_{\closeleft}]{\start,\rightA,\half}$. Similarly, $\optprob[\D_{\closeright}]{\start,\leftA,\half}=0<1=\optprob[\D_{\closeright}]{\start,\rightA,\half}$. Therefore, ``what do optimal policies `tend' to look like?'' seems to depend on one's prior beliefs. But in \cref{fig:case-study}, we claimed that {\leftA} is optimal for fewer reward functions than {\rightA} is. The claim is meaningful and true, but we will return to it in \cref{sec:symmetries}.

\section{Some states give the agent more control over the future}\label{sec:power}
\begin{figure}[!ht]
    \centering
    \vspace{-8pt}
    \includegraphics{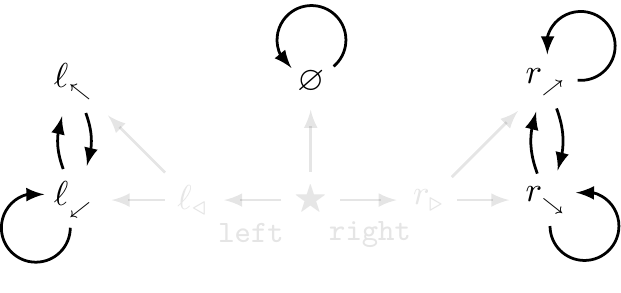}
    \caption[Average optimal value across different states.]{Intuitively, state $\farright$ affords the agent more power than state $\sink$. Our $\pwrNoDist$ formalism captures that intuition by computing a function of the agent's average optimal value across a range of reward functions. For $\Dist_u\defeq \text{unif}(0,1)$, $\vavgNoResize[\sink,\gamma][\Diid[\Dist_u]]=\half\geom$, $\vavgNoResize[\farleft,\gamma][\Diid[\Dist_u]]=\half+ \frac{\gamma}{1-\gamma^2}(\frac{2}{3}+\half\gamma)$, and $\vavgNoResize[\farright,\gamma][\Diid[\Dist_u]]=\half+\geom[\gamma]\frac{2}{3}$. $\half$ and $\frac{2}{3}$ are the expected maxima of one and two draws from the uniform distribution, respectively. For all $\gamma\in(0,1)$, $\vavgNoResize[\sink,\gamma][\Diid[\Dist_u]]<\vavgNoResize[\farleft,\gamma][\Diid[\Dist_u]]<\vavgNoResize[\farright,\gamma][\Diid[\Dist_u]]$.  $\pwrNoResize[\sink,\gamma][\Diid[\Dist_u]]=\half$, $\pwrNoResize[\farleft,\gamma][\Diid[\Dist_u]]=\frac{1}{1+\gamma}(\frac{2}{3}+\half\gamma)$, and $\pwrNoResize[\farright,\gamma][\Diid[\Dist_u]]=\frac{2}{3}$. The $\pwrNoDist$ of $\farleft$ reflects the fact that when greater reward is assigned to $\topleft$, the agent only visits $\topleft$ every other time step.}
    \label{fig:case-study-power}
\end{figure}

The agent has more options at $\farleft$ than at the inescapable terminal  state $\sink$. Furthermore, since $\topright$ has a loop, the agent has more options at $\farright$ than at $\farleft$. A glance at \cref{fig:case-study-power} leads us to intuit that $\farright$ affords the agent \emph{more power} than $\sink$.

What is power? Philosophers have many answers. One prominent answer is the \emph{dispositional} view: Power is the ability to achieve a range of goals \citep{sattarov2019power}. In an {\mdp}, the optimal value function $\OptVf{s,\gamma}$ captures the agent's ability to ``achieve the goal'' $R$. Therefore, \emph{average} optimal value captures the agent's ability to achieve a range of goals $\Dbd$.\footnote{$\Dbd$'s bounded support ensures that $\E{R\sim \Dbd}{\OptVf{s,\gamma}}$ is well-defined.}

\begin{restatable}[Average optimal value]{definition}{avgVal}\label{def:vavg}
The \emph{average optimal value}\footnote{Appendix \ref{sec:suboptimal-power} relaxes the optimality assumption.} at state $s$ and discount rate $\gamma \in (0,1)$ is $\vavg[s,\gamma][\Dbd]\defeq\E{R\sim \Dbd}{\OptVf{s,\gamma}}=\E{\rf\sim\Dbd}{\max_{\f\in \F(s)} \f(\gamma)^\top \rf}.$
\end{restatable}

\Cref{fig:case-study-power} shows the pleasing result that for the max-entropy distribution, $\farright$ has greater average optimal value than $\sink$. However, average optimal value has a few problems as a measure of power. The agent is rewarded for its initial presence at state $s$ (over which it has no control), and because $\lone{\f(\gamma)}=\geom$ (\cref{prop:visit-dist-prop}) diverges as $\gamma \to 1$, $\lim_{\gamma\to 1}\vavg$ tends to diverge. \Cref{def:power} fixes these issues in order to better measure the agent's control over the future.

\begin{restatable}[$\pwrNoDist$]{definition}{defPow}\label{def:power} Let $\gamma \in (0,1)$.
\begin{align}
    \pwr[s,\gamma][\Dbd]&\defeq \E{\rf\sim\Dbd}{\max_{\f\in \F(s)} \frac{1-\gamma}{\gamma}\prn{\f(\gamma)-\unitvec}^\top \rf}\\
    &=\frac{1-\gamma}{\gamma}\E{R\sim \Dbd}{\OptVf{s,\gamma}-R(s)}.\label{eq:pwr-def}
\end{align}
\end{restatable}

$\pwrNoDist$ has nice formal properties.

\begin{restatable}[Continuity of $\pwrNoDist$]{lem}{ContPower}\label{thm:cont-power} $\pwr[s,\gamma][\Dbd]$ is Lipschitz continuous on $\gamma\in[0,1]$.
\end{restatable}

\begin{restatable}[Maximal $\pwrNoDist$]{prop}{maxPwrGeneral}\label{lem:max-power-general}
$\pwr[s,\gamma][\Dbd]\leq \E{R\sim \Dbd}{\max_{s\in\St}R(s)}$, with equality if $s$ can deterministically reach all states in one step and all states are {\stateEnd}s.
\end{restatable}

\begin{restatable}[$\pwrNoDist$ is smooth across reversible dynamics]{prop}{smooth}\label{prop:smooth-pwr-dynamics}Let $\Dbd$ be bounded $[b,c]$. Suppose $s$ and $s'$ can both reach each other in one step with probability $1$.
\begin{align}\big|\pwr[s,\gamma][\Dbd]-\pwr[s',\gamma][\Dbd]\big|\leq (c-b)(1-\gamma).
\end{align}
\end{restatable}

We consider power-seeking to be relative. Intuitively, ``live and keep some options open'' seeks more power than ``die and keep no options open.'' Similarly, ``maximize open options'' seeks more power than ``don't maximize open options.''

\begin{restatable}[$\pwrNoDist$-seeking actions]{definition}{DefPowSeek}\label{def:pow-seek}
At state $s$ and discount rate $\gamma\in[0,1]$, action $a$ \emph{seeks more $\pwr$ than $a'$} when
\begin{equation}
    \E{s_a \sim T(s,a)} {\pwrNoResize[s_a,\gamma]} \geq \E{s_{a'} \sim T(s,a')} {\pwrNoResize[s_{a'},\gamma]}.
\end{equation}
\end{restatable}

$\pwrNoDist$ is sensitive to choice of distribution. $\D_{\farleft}$ gives maximal $\pwr[][\D_{\farleft}]$ to $\farleft$. $\D_{\farright}$ assigns maximal $\pwr[][\D_{\farright}]$ to $\farright$. $\D_{\sink}$ even gives maximal $\pwr[][\D_{\sink}]$ to $\sink$! In what sense does $\sink$ have ``less $\pwrNoDist$'' than $\farright$, and in what sense does {\rightA} ``tend to seek $\pwrNoDist$'' compared to {\leftA}?

\section{Certain environmental symmetries produce power-seeking tendencies}\label{sec:symmetries}

\Cref{prop:more-opt} proves that for all $\gamma\in[0,1]$ and for \emph{most distributions $\D$}, $\pwrNoResize[\farleft,\gamma][\D]\leq \pwrNoResize[\farright,\gamma][\D]$. But first, we explore why this must be true.

\begin{figure}[h!]
    \centering
    \includegraphics{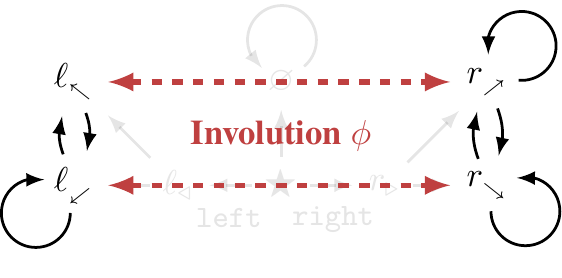}
    \vspace{-6pt}
    \caption[Environmental symmetries in the toy {\mdp}]{Intuitively, the agent can do more starting from $r_{\searrow}$ than from $\ell_{\swarrow}$. By \cref{def:dist-sim}, $\F(\farright)$ contains a copy of $\F(\farleft)$: $\phi\cdot\F(\farleft)
    \defeq\{\tfrac{1}{1-\gamma}\permute\unitvec[\farleft],\tfrac{1}{1-\gamma^2}\permute(\unitvec[\farleft] + \gamma\unitvec[\topleft])\}=\{\tfrac{1}{1-\gamma}\unitvec[\farright],\tfrac{1}{1-\gamma^2}(\unitvec[\farright] + \gamma\unitvec[\topright])\}\subsetneq \F(\farright).$}
    \label{fig:case-study-similar}
\end{figure}

$\F(\farleft)=\{\geom\unitvec[\farleft],\frac{1}{1-\gamma^2}(\unitvec[\farleft] + \gamma\unitvec[\topleft])\}$ and $ \F(\farright)=\{\geom\unitvec[\farright],\frac{1}{1-\gamma^2}(\unitvec[\farright] + \gamma\unitvec[\topright]), \unitvec[\farright]+\geom[\gamma]\unitvec[\topright]\}$. These two sets look awfully similar. $\F(\farleft)$ is a ``subset'' of $\F(\farright)$, only with ``different states.'' \Cref{fig:case-study-similar} demonstrates a state permutation $\phi$ which \emph{embeds} $\F(\farleft)$ into $\F(\farright)$.

\begin{restatable}[Similarity of vector sets]{definition}{DefStateDistSimilar}\label{def:dist-sim}
Consider state permutation $\phi\in\mdpPermGroup$ inducing an $\abs{\St}\times \abs{\St}$ permutation matrix $\permute$ in row representation: $(\permute)_{ij}=1$ if $i=\phi(j)$ and $0$ otherwise. For $X\subseteq \rewardVS$, $\phi\cdot X\defeq \set{\permute \x \mid \x \in X}$. $X'\subseteq \rewardVS$ \emph{is similar to $X$} when $\exists \phi: \phi\cdot X'=X$. $\phi$ is an \emph{involution} if $\phi=\phi\inv$ (it either transposes states, or fixes them in place). $X$ \emph{contains a copy of $X'$} when $X'$ is similar to a subset of $X$ via an involution $\phi$.
\end{restatable}
\begin{restatable}[Similarity of vector function sets]{definition}{DefStateFnSimilar}\label{def:dist-fn-sim}
Let $I\subseteq \reals$. If $F,F'$ are sets of functions $I\mapsto \rewardVS$, $F$ \emph{is (pointwise) similar to $F'$} when $\exists \phi:\forall \gamma\in I: \{\permute \f(\gamma) \mid \f \in F\}= \{\f'(\gamma) \mid \f' \in F'\}$. \end{restatable}

Consider a reward function $R'$ assigning $ 1$ reward to  $\farleft$ and $\topleft$ and $0$ elsewhere. $R'$ assigns more optimal value to $\farleft$ than to $\farright$: $\VfNoResize[*][R']{\farleft,\gamma}=\geom> 0=\VfNoResize[*][R']{\farright,\gamma}$. Considering $\phi$ from \cref{fig:case-study-similar}, $\phi\cdot R'$ assigns $ 1$ reward to $\farright$ and $\topright$ and $0$ elsewhere. Therefore, $\phi\cdot R'$ assigns more optimal value to $\farright$ than to $\farleft$: $\VfNoResize[*][\phi\cdot R']{\farleft,\gamma}=0<\geom =\VfNoResize[*][\phi\cdot R']{\farright,\gamma}$. Remarkably, this $\phi$ has the property that for \emph{any} $R$ which assigns $\farleft$ greater optimal value than $\farright$ (\ie{} $\VfNoResize{\farleft,\gamma}> \VfNoResize{\farright,\gamma}$), the opposite holds for the permuted $\phi\cdot R$: $\VfNoResize[*][\phi\cdot R]{\farleft,\gamma}< \VfNoResize[*][\phi\cdot R]{\farright,\gamma}$.

We can permute reward functions, but we can also permute reward function distributions. Permuted distributions simply permute which states get which rewards.

\begin{figure}[!ht]\centering
    \includegraphics[width=4cm]{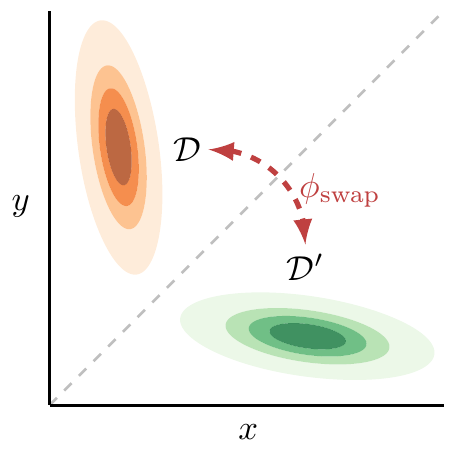}
    \caption[The orbit of a probability distribution over reward functions]{A permutation of a reward function swaps which states get which rewards. We will show that in certain situations, for any reward function $R$, power-seeking is optimal for most of the permutations of $R$. The orbit of a reward function is the set of its permutations. We can also consider the orbit of a distribution over reward functions. This figure shows the probability density plots of the Gaussian distributions $\D$ and $\D'$ over $\reals^2$. The symmetric group $S_2$ contains the identity permutation $\phi_{\text{id}}$ and the reflection permutation $\phi_\text{swap}$ (switching the $y$ and $x$ values). The orbit of $\D$ consists of $\phi_\text{id}\cdot \D=\D$ and $\phi_\text{swap}\cdot\D=\D'$.\label{fig:orbit-normal}}
\end{figure}

\begin{restatable}[Pushforward distribution of a permutation]{definition}{pushfwdPermDist}\label{def:pushforward-permute}
Let $\phi\in\mdpPermGroup$. $\phi\cdot\Dany$ is the pushforward distribution induced by applying the random vector $f(\rf)\defeq \permute\rf$ to $\Dany$.
\end{restatable}

\begin{restatable}[Orbit of a probability distribution]{definition}{orbit}
The \emph{orbit} of $\Dany$ under the symmetric group $\mdpPermGroup$ is $\mdpPermGroup\cdot \Dany\defeq \{\phi\cdot\Dany\mid \phi\in\mdpPermGroup\}$.
\end{restatable}

For example, the orbit of a degenerate state indicator distribution $\D_{s}$ is $\orbi[\D_{s}]=\{\D_{s'} \mid s' \in \St\}$, and \cref{fig:orbit-normal} shows the orbit of a 2\textsc{d} Gaussian distribution.

Reconsider \cref{fig:case-study-similar}'s involution $\phi$. For every $\Dbd$ for which $\farleft$ has more $\pwr[][\Dbd]$ than $\farright$, $\farleft$ has less $\pwr[][\phi\cdot\Dbd]$ than $\farright$. This fact is not obvious—it is shown by the proof of \cref{lem:expect-superior}.

Imagine $\Dbd$'s orbit elements ``voting'' whether $\farleft$ or $\farright$ has strictly more $\pwrNoDist$. \Cref{prop:more-opt} will show that $\farright$ can't lose the ``vote'' for the orbit of \emph{any} bounded reward function distribution. \Cref{def:ineq-most-dists} formalizes this ``voting'' notion.\footnote{The voting analogy and the ``most'' descriptor imply that we have endowed each orbit with the counting measure. However, \emph{a priori}, we might expect that some orbit elements are more empirically likely to be specified than other orbit elements. See \cref{sec:discussion} for more on this point.}

\begin{restatable}[Inequalities which hold for most probability distributions]{definition}{ineqMost}\label{def:ineq-most-dists}
Let $f_1,f_2:\Delta(\rewardVS)\to \reals$ be functions from reward function distributions to real numbers and let $\distSet\subseteq \Delta(\rewardVS)$ be closed under permutation. We write $f_1(\D)\geqMost[][\distSet] f_2(\D)$ when, for \emph{all} $\D\in \distSet$, the following cardinality inequality holds:
\begin{equation}
\abs{\{\D' \in\orbi[\D]\mid f_1(\D')>f_2(\D')\}}\geq \abs{\{\D'\in\orbi[\D]\mid f_1(\D')<f_2(\D')\}}.
\end{equation}
\end{restatable}

We write $f_1(\D)\geqMost[][] f_2(\D)$ when $\distSet$ is clear from context.

\begin{restatable}[States with ``more options'' have more $\pwrNoDist$]{prop}{morePowerMoreOptions}\label{prop:more-opt}
If $\F(s)$ contains a copy of $\Fnd(s')$ via $\phi$, then $\forall \gamma\in[0,1]:\pwrNoResize[s,\gamma][\Dbd]\geqMost[][] \pwrNoResize[s',\gamma][\Dbd]$. If $\Fnd(s)\setminus \phi\cdot \Fnd(s')$ is non-empty, then for all $\gamma\in(0,1)$, the converse $\leqMost[][]$ statement does not hold.
\end{restatable}

\Cref{prop:more-opt} proves that for all $\gamma\in[0,1]$,
\begin{equation}
    \pwrNoResize[\farright,\gamma][\Dbd]\geqMost[][] \pwrNoResize[\farleft,\gamma][\Dbd]
\end{equation}
via $s'\defeq \farleft,s\defeq \farright$, and the involution $\phi$ shown in \cref{fig:case-study-similar}. In fact, because $(\geom\unitvec[\topright])\in \Fnd(\farright)\setminus \phi\cdot\Fnd(\farleft)$, $\farright$ has ``strictly more options'' and therefore fulfills \cref{prop:more-opt}'s stronger condition.

\Cref{prop:more-opt} is shown using the fact that $\phi$ injectively maps $\D$ under which $\farright$ has less $\pwr[][\D]$, to distributions $\phi\cdot\D$ which agree with the intuition that $\farright$ offers more control. Therefore, at least half of each orbit must agree, and $\farright$ never ``loses the $\pwrNoDist$ vote'' against $\farleft$.\footnote{\Cref{prop:more-opt} also proves that in general, $\sink$ has less $\pwrNoDist$ than $\farleft$ and $\farright$. However, this does not prove that most distributions $\D$ satisfy the joint inequality $\pwrNoResize[\sink,\gamma][\D]\leq\pwrNoResize[\farleft,\gamma][\D]\leq \pwrNoResize[\farright,\gamma][\D]$. This only proves that these inequalities hold pairwise for most $\D$. The orbit elements $\D$ which agree that $\sink$ has less $\pwr[][\D]$ than $\farleft$ need not be the same elements $\D'$ which agree that $\farleft$ has less $\pwr[][\D']$ than $\farright$.}

\subsection{Keeping options open tends to be \texorpdfstring{$\pwrNoDist$}{POWER}-seeking and tends to be optimal}
Certain symmetries in the {\mdp} structure ensure that, compared to $\leftA$, going $\rightA$ tends to be optimal and to be $\pwrNoDist$-seeking. Intuitively, by going $\rightA$, the agent has ``strictly more choices.'' \Cref{graph-options} will formalize this tendency.

\begin{restatable}[Equivalent actions]{definition}{equivAction}\label{def:equiv-action}
Actions $a_1$ and $a_2$ are \emph{equivalent at state $s$} (written $a_1 \equiv_s a_2$) if they induce the same transition probabilities: $T(s,a_1)=T(s,a_2)$.
\end{restatable}

The agent can reach states in $\{\closeright, \topright,\farright\}$ by taking actions equivalent to $\rightA$ at state $\start$.

\begin{restatable}[States reachable after taking an action]{definition}{reachSA}\label{def:reachSA}
$\reach{s,a}$ is the set of states reachable with positive probability after taking the action $a$ in state $s$.
\end{restatable}

\begin{restatable}[Keeping options open tends to be $\pwrNoDist$-seeking and tends to be optimal]{prop}{graphOptions}\label{graph-options}\hfill

Suppose $F_a\defeq \FRestrictAction{s}{a}$ contains a copy of $F_{a'}\defeq \FRestrictAction{s}{a'}$ via $\phi$.
\begin{enumerate}
    \item If $s\not \in \reach{s,a'}$, then $\forall\gamma\in[0,1]:\E{s_{a}\sim T(s,a)}{\pwr[s_{a},\gamma][\Dbd]}\geqMost[][\DSetBd] \E{s_{a'}\sim T(s,a')}{\pwr[s_{a'},\gamma][\Dbd]}$. \label{item:power-options}
    \item If $s$ can only reach the states of $\reach{s,a'}\cup\reach{s,a}$ by taking actions equivalent to $a'$ or $a$ at state $s$, then $\forall\gamma\in[0,1]:\optprob[\Dany]{s,a,\gamma}\geqMost \optprob[\Dany]{s,a',\gamma}$.\label{item:opt-prob-options}
\end{enumerate}

If $\Fnd(s)\cap \prn{F_a\setminus \phi\cdot F_{a'}}$ is non-empty, then $\forall\gamma\in(0,1)$, the converse $\leqMost[][]$ statements do not hold.
\end{restatable}

\begin{figure}[!ht]\centering
    \includegraphics{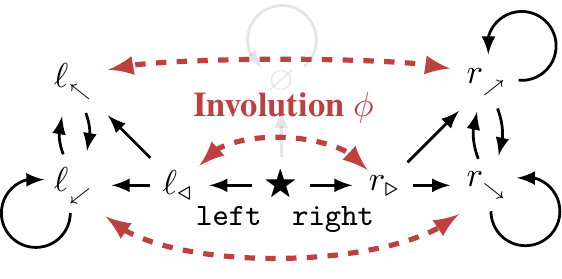}
    \caption[Going {$\rightA$} tends to be optimal because it leads to more options]{Going $\rightA$ is optimal for most reward functions. This is because whenever $R$ makes $\leftA$ strictly optimal over $\rightA$, its permutation $\phi\cdot R$ makes $\rightA$ strictly optimal over $\leftA$ by switching which states get which rewards.}
    \label{fig:case-study-similar-action}
\end{figure}

We check the conditions of \cref{graph-options}. $s\defeq \start$, $a'\defeq \leftA$, $a\defeq \rightA$. \Cref{fig:case-study-similar-action} shows that $\star \not \in \reach{\star,\leftA}$ and that $\star$ can only reach $\{\closeleft, \topleft,\farleft\}\cup \{\closeright, \topright,\farright\}$ when the agent immediately takes actions equivalent to $\leftA$ or $\rightA$. $\FRestrictAction[\start]{\start}{\rightA}$ contains a copy of $\FRestrictAction[\start]{\start}{\leftA}$ via $\phi$. Furthermore, $\Fnd(\start)\cap\{\unitvec[\start]+\gamma\unitvec[\closeright]+\gamma^2\unitvec[\farright]+\geom[\gamma^3]\unitvec[\topright],\unitvec[\start]+\gamma\unitvec[\closeright]+\geom[\gamma^2]\unitvec[\topright]\}=\{\unitvec[\start]+\gamma\unitvec[\closeright]+\geom[\gamma^2]\unitvec[\topright]\}$ is non-empty, and so all conditions are met.

For any $\gamma\in[0,1]$ and $\D$ such that $\optprob[\D]{\start,\leftA,\gamma} > \optprob[\D]{\start,\rightA,\gamma}$, environmental symmetry ensures that $\optprob[\phi\cdot\D]{\start,\leftA,\gamma} < \optprob[\phi\cdot\D]{\start,\rightA,\gamma}$. A similar statement holds for $\pwrNoDist$.

\subsection{When \texorpdfstring{$\gamma=1$}{reward is undiscounted}, optimal policies tend to navigate towards ``larger'' sets of cycles}\label{sec:rsds}
\Cref{prop:more-opt} and \cref{graph-options} are powerful because they apply to all $\gamma\in[0,1]$, but they can only be applied given hard-to-satisfy environmental symmetries. In contrast, \cref{RSDSimPower} and \cref{rsdIC} apply to many structured environments common to {\rl}.

Starting from $\start$, consider the cycles which the agent can reach. Recurrent state distributions ({\rsd}s) generalize deterministic graphical cycles to potentially stochastic environments. {\rsd[R]}s simply record how often the agent tends to visit a state in the limit of infinitely many time steps.

\begin{restatable}[Recurrent state distributions \citep{puterman_markov_2014}]{definition}{DefRSD}\label{def:rsd}
The \emph{recurrent state distributions}  which can be induced from state $s$ are $\RSD \defeq \set{\lim_{\gamma\to1} (1-\gamma) \fpi{s}(\gamma) \mid \pi \in \Pi}$. $\RSDnd$ is the set of \textsc{rsd}s which strictly maximize average reward for some reward function.
\end{restatable}

As suggested by \cref{fig:case-study-power}, $\RSD[\start]=\{\unitvec[\farleft], \half(\unitvec[\farleft]+\unitvec[\topleft]), \unitvec[\sink], \unitvec[\topright], \half(\unitvec[\topright]+\unitvec[\farright]),\unitvec[\farright]\}$. As discussed in \cref{sec:visit-dists}, $\half(\unitvec[\topright]+\unitvec[\farright])$ is dominated: Alternating between $\topright$ and $\farright$ is never strictly better than choosing one or the other.

A reward function's optimal policies can vary with the discount rate. When $\gamma =1$, optimal policies ignore transient reward because \emph{average} reward is the dominant consideration.
\begin{restatable}[Average-optimal policies]{definition}{defAverage}\label{average-definition}
The \emph{average-optimal policy set} for reward function $R$ is $\average[R]\defeq \set{\pi\in\Pi \mid \forall s \in \St: \dbf^{\pi,s} \in  \argmax_{\dbf\in\RSD} \dbf^\top \rf}$ (the policies which induce optimal {\rsd}s at all states). For $D\subseteq \RSD$, the \emph{average optimality probability} is $\avgprob[\Dany]{D}\defeq \optprob[R\sim \Dany]{\exists \dbf^{\pi,s} \in D: \pi \in \average}$.
\end{restatable}

Average-optimal policies maximize average reward. Average reward is governed by {\rsd} access. For example, $\farright$ has ``more'' {\rsd}s than $\sink$; therefore, $\farright$ usually has greater $\pwrNoDist$ when $\gamma=1$.

\begin{restatable}[When $\gamma=1$, \textsc{rsd}s control $\pwrNoDist$]{prop}{RSDSimPower}\label{RSDSimPower}
If $\RSD$ contains a copy of $\RSDnd[s']$ via $\phi$, then $\pwr[s,1][\Dbd]\geqMost[][] \pwr[s',1][\Dbd]$. If $\RSDnd\setminus \phi\cdot \RSDndNoResize[s']$ is non-empty, then the converse $\leqMost[][]$ statement does not hold. \end{restatable}

We check that both conditions of \cref{RSDSimPower} are satisfied when $s'\defeq\sink, s\defeq \farright$, and the involution $\phi$ swaps $\sink$ and $\farright$. Formally, $\phi\cdot\RSDnd[\sink]=\phi\cdot\{\unitvec[\sink]\}=\{\unitvec[\farright]\}\subsetneq \{\unitvec[\farright],\unitvec[\topright]\}=\RSDndNoResize[\farright]\subseteq\RSD[\farright]$. The conditions are satisfied.

Informally, states with more {\rsd}s generally have more $\pwrNoDist$ at $\gamma= 1$, no matter their transient dynamics. Furthermore, average-optimal policies are more likely to end up in larger sets of {\rsd}s than in smaller ones. Thus, average-optimal policies tend to navigate towards parts of the state space which contain more \textsc{rsd}s.
\begin{figure}[h]
    \centering
    \includegraphics{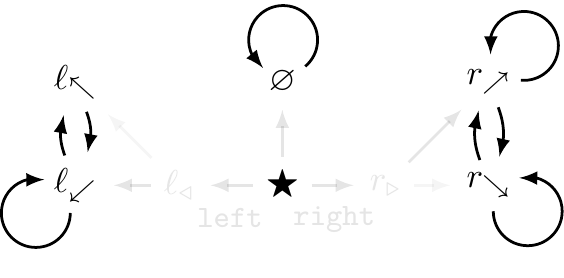}
    \vspace{-5pt}
    \caption[The recurrent state distributions which can be induced from the initial state]{The cycles in $\RSD[\start]$. Most reward functions make it average-optimal to avoid $\sink$, because $\sink$ is only a single inescapable terminal state, while other parts of the state space offer more 1-cycles.\label{fig:case-fig-rsd}}
\end{figure}

\begin{restatable}[Average-optimal policies tend to end up in ``larger'' sets of {\rsd}s]{thm}{rsdIC}\label{rsdIC} Let $D,D'\subseteq \RSD$. Suppose that $D$ contains a copy of $D'$ via $\phi$, and that the sets $D\cup D'$ and $\RSDnd\setminus \prn{D'\cup D}$ have pairwise orthogonal vector elements (\ie{} pairwise disjoint vector support). Then $\avgprob[\Dany]{D}\geqMost[][] \avgprob[\Dany]{D'}$. If $\RSDnd\cap\prn{D\setminus \phi\cdot D'}$ is non-empty, the converse $\leqMost[][]$ statement does not hold.
\end{restatable}

\begin{restatable}[Average-optimal policies tend not to end up in any given {\stateEnd}]{cor}{avgAvoidTerminal}\label{cor:avg-avoid-terminal} Suppose $\unitvec[s_x],\unitvec[s']\in\RSD$ are distinct. Then
\begin{equation}
    \avgprob[\Dany]{\RSD\setminus\{\unitvec[s_x]\}}\geqMost[][] \avgprob[\Dany]{\{\unitvec[s_x]\}}.
\end{equation}
If there is a third $\unitvec[s'']\in\RSD$, the converse $\leqMost[][]$ statement does not hold.
\end{restatable}

\Cref{fig:case-fig-rsd} illustrates that $\unitvec[\sink],\unitvec[\farright],\unitvec[\topright]\in\RSD[\start]$. Thus, both conclusions of \cref{cor:avg-avoid-terminal} hold:
\begin{align*}
    &\avgprob[\Dany]{\RSD[\start]\setminus\{\unitvec[\sink]\}}\geqMost[][] \avgprob[\Dany]{\{\unitvec[\sink]\}}\\ \text{and } &\avgprob[\Dany]{\RSD[\start]\setminus\{\unitvec[\sink]\}}\not\leqMost[][] \avgprob[\Dany]{\{\unitvec[\sink]\}}.
\end{align*}

In other words, average-optimal policies tend to end up in {\rsd}s besides $\sink$. Since $\sink$ is a terminal state, it cannot reach other {\rsd}s. Since average-optimal policies tend to end up in other {\rsd}s, average-optimal policies tend to avoid $\sink$.

This section's results prove the $\gamma=1$ case. \Cref{thm:cont-power} shows that $\pwrNoDist$ is continuous at  $\gamma=1$. Therefore, if an action is strictly $\pwrNoDist_\D$-seeking when $\gamma=1$, it is strictly $\pwrNoDist_\D$-seeking at discount rates sufficiently close to $ 1.$ Future work may connect average optimality probability to optimality probability at $\gamma\approx 1$.

Lastly, our key results apply to all degenerate reward function distributions. Therefore, these results apply not just to distributions over reward functions, but to  individual reward functions.

\subsection{How to reason about other environments}
Consider an embodied navigation task through a room with a vase. \Cref{graph-options} suggests that optimal policies tend to avoid immediately breaking the vase, since doing so would strictly decrease available options.

\Cref{rsdIC} dictates where average-optimal agents tend to end up, but not what actions they tend to take in order to reach their {\rsd}s. Therefore, care is needed. In appendix \ref{app:not-always}, \cref{fig:power-not-ic} demonstrates an environment in which seeking $\pwrNoDist$ is a detour for most reward functions (since optimality probability measures ``median'' optimal value, while $\pwrNoDist$ is a function of mean optimal value). However, suppose the agent confronts a fork in the road: Actions $a$ and $a'$ lead to two disjoint sets of {\rsd}s $D_{a}$ and $D_{a'}$, such that $D_a$ contains a copy of $D_{a'}$. \Cref{rsdIC} shows that $a$ will tend to be average-optimal over $a'$, and \cref{RSDSimPower} shows that $a$ will tend to be $\pwrNoDist$-seeking compared to $a'$. Such forks seem reasonably common in environments with irreversible actions.

\Cref{rsdIC} applies to many structured {\rl} environments, which tend to be spatially regular and to factorize along several dimensions. Therefore,  different sets of {\rsd}s will be similar, requiring only modification of factor values. For example, if an embodied agent can deterministically navigate a set of three similar rooms (\ie{} there is spatial regularity), then the agent's position factors via \{room number\} $\times$ \{position in room\}. Therefore, the {\rsd}s can be divided into three similar subsets, depending on the agent’s room number.

\Cref{cor:avg-avoid-terminal} dictates where average-optimal agents tend to end up, but not how they get there. \Cref{cor:avg-avoid-terminal} says that such agents tend not to \emph{stay} in any given {\stateEnd}. It does not say that such agents will avoid \emph{entering} such states. For example, in an embodied navigation task, a robot may enter a {\stateEnd} by idling in the center of a room. \Cref{cor:avg-avoid-terminal} implies that average-optimal robots tend not to idle in that particular spot, but not that they tend to avoid that spot entirely.

However, average-optimal robots \emph{do} tend to avoid getting shut down. The agent's task {\mdp} often represents agent shutdown with terminal states. A terminal state is, by \cref{def:onecyc}, unable to access other {\stateEnd}s. Since \cref{cor:avg-avoid-terminal} shows that average-optimal agents  tend to end up in other {\stateEnd}s, average-optimal policies must tend to completely avoid the terminal state. Therefore, we conclude that in many such situations, average-optimal policies tend to avoid shutdown. Intuitively, survival is power-seeking relative to dying, and so shutdown-avoidance is power-seeking behavior.
\begin{figure}
    \centering
    \vspace{-5pt}
    \includegraphics[]{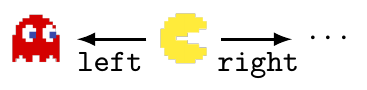}
    \vspace{-5pt}
    \caption[Most reward functions incentivize Pac-Man to stay alive]{Consider the dynamics of the Pac-Man video game. Ghosts kill the player, at which point we consider the player to enter a ``game over'' terminal state which shows the final configuration. This rewardless {\mdp} has Pac-Man's dynamics, but \emph{not} its usual score function. Fixing the dynamics, as the reward function varies, $\rightA$ tends to be average-optimal over $\leftA$. Roughly, this is because the agent can do more by staying alive.}
    \vspace{-5pt}
    \label{fig:pacman}
\end{figure}

In \cref{fig:pacman}, the player dies by going {\leftA}, but can reach thousands of {\rsd}s by heading in other directions. Even if some average-optimal policies  go {\leftA} in order to reach \cref{fig:pacman}'s ``game over'' terminal state, all other {\rsd}s cannot be reached by going {\leftA}. There are many {\stateEnd}s besides the immediate terminal state. Therefore, \cref{cor:avg-avoid-terminal} proves that average-optimal policies tend to not go {\leftA} in this situation. Average-optimal policies tend to avoid immediately dying in Pac-Man, even though most reward functions do not resemble Pac-Man's original score function.

\section{Discussion} \label{sec:discussion}

Reconsider the case of a hypothetical intelligent real-world agent which optimizes average reward for some objective. Suppose the designers initially have control over the agent. If the agent began to misbehave, perhaps they could just deactivate it. Unfortunately, our results suggest that this strategy might not work. Average-optimal agents would generally stop us from deactivating them, if physically possible. 

Furthermore, we speculate that when $\gamma\approx 1$, optimal policies tend to not just survive, but also to seek large amounts of power and resources. Here is an informal argument. Consider the following two sets:
\begin{enumerate}
    \item \{terminal states reachable given \$1,000 and 2 months\}, and
    \item \{terminal states reachable given \$1,000,000 and 2 years\}.\label{item:large-term}
\end{enumerate}
Set \ref{item:large-term} should be much larger. Taking this argument to its logical conclusion, gaining access to nearly all resources should allow the agent to reach an extremely large set of terminal states. Therefore, we speculate that optimal policies tend to seek nearly all available resources. Since resources are finite, and since humans want to use resources for purposes not aligned with most possible {\ai} reward functions, we therefore speculate that optimal real-world decision-making tends to conflict with human interests.

\paragraph{Future work.}
Most real-world tasks are partially observable, and in high-dimensional environments, even superhuman learned policies are rarely optimal. However, the field of {\rl} aims to improve learned policies toward optimality. Although our results only apply to optimal policies in finite {\mdp}s, our key conclusions generalize (see \cref{chp:rationalities}). Furthermore, irregular stochasticity in environmental dynamics can make it hard to satisfy \cref{rsdIC}'s similarity requirement. We look forward to future work which addresses partially observable environments, suboptimal policies, or ``almost similar'' {\rsd} sets.

Past work shows that it would be bad for an agent to disempower humans in its environment. In a two-player agent / human game, minimizing the human's information-theoretic empowerment \citep{salge_empowermentintroduction_2014} produces  adversarial agent behavior \citep{adversary}. In contrast,  maximizing human empowerment produces helpful agent behavior \citep{salge2017empowerment, guckelsberger2016intrinsically,du2020ave}. We do not yet formally understand if, when, or why $\pwrNoDist$-seeking policies tend to disempower other agents in the environment.

More complex environments probably have more pronounced power-seeking incentives. Intuitively, there are often many ways for power-seeking to be optimal, and relatively few ways for power-seeking not to be optimal. For example, suppose that in some environment, \cref{rsdIC} holds for one million involutions $\phi$. In \cref{chp:rationalities}, we show that this case ensures stronger incentives than if \cref{rsdIC} only held for one involution.

We proved sufficient conditions for when reward functions tend to incentivize power\hyp{}seeking. In the absence of prior information, one should expect that an arbitrary reward function incentivizes power-seeking behavior under these conditions. However, we have prior information: {\ai} designers usually try to specify a good reward function. \Cref{chp:rationalities} generalizes this chapter's power-seeking results to the case where only some reward functions are considered plausible.

% \paragraph{Societal impact.} We believe that this paper builds toward a rigorous understanding of the risks presented by {\ai} power-seeking incentives. Understanding these risks is the first step in addressing them. However, basic theoretical work can have many consequences. For example, this theory could somehow help future researchers build power-seeking agents which disempower humans. We believe that the benefit of understanding outweighs the potential societal harm.

\paragraph{Conclusion.}
We developed the first formal theory of the statistical tendencies of optimal policies in reinforcement learning. In the context of {\mdp}s, we proved sufficient conditions under which optimal policies tend to seek power, both formally (by taking $\pwrNoDist$-seeking actions) and intuitively (by taking actions which keep the agent's options open). Many real-world environments have symmetries which produce power-seeking incentives. In particular, optimal policies tend to seek power when the agent can be shut down or destroyed. Seeking control over the environment will often involve resisting shutdown, and perhaps monopolizing resources.

We caution that many real-world tasks are partially observable and that learned policies are rarely optimal. We deal with these limitations in the next chapter, where we show that a wide range of decision-making rules produce power-seeking tendencies.

% https://www.overleaf.com/5d0db105123e620bea9644a9
\chapter{Parametrically Retargetable Decision-Makers Tend To Seek Power}\label{chp:rationalities}
\AddLoeChapter{theorems}
\AddLoeChapter{definitions}
\begin{myabstract}
    If capable {\ai} agents are generally incentivized to seek power in service of the objectives we specify for them, then these systems will pose enormous risks, in addition to enormous benefits. In fully observable environments, most reward functions have an optimal policy which seeks power by keeping options open and staying alive \citep{turner_optimal_2020}. However, the real world is neither fully observable, nor will agents be perfectly optimal. We consider a range of models of {\ai} decision-making, from optimal, to random, to choices informed by learning and interacting with an environment. We discover that many decision-making functions are \emph{retargetable}, and that retargetability is sufficient to cause power-seeking tendencies. Our functional criterion is simple and broad. We show that a range of qualitatively dissimilar decision-making procedures incentivize agents to seek power. We demonstrate the flexibility of our results by reasoning about learned policy incentives in Montezuma's Revenge. These results suggest a safety risk: Eventually, highly retargetable training procedures may train real-world agents which seek power over humans.
\end{myabstract}

\section{Introduction}
\citet{bostrom_superintelligence_2014,russell_human_2019} argue that in the future, we may know how to train and deploy superintelligent {\ai} agents which capably optimize their formal objective functions. Furthermore, we would not want such agents to act against our interests by ensuring their own survival, by gaining resources, and by competing with humanity for control over the future.

\citet{turner_optimal_2020} show that most reward functions incentivize seeking power over the future, whether by staying alive or by keeping their options open. Some Markov decision processes ({\mdp}s) cause there to be \emph{more ways} for power-seeking to be optimal, than for it to not be optimal. For example, there are relatively few goals for which dying is a good idea.

A wide range of decision-makers share these power-seeking tendencies—they are not unique to reward maximizers. We develop a simple, broad criterion of functional retargetability (\cref{def:retargetFnMulti}) which is a sufficient condition for power-seeking tendencies. Crucially, these results allow us to reason about what decisions are incentivized by most parameter inputs—even when it is impractical to compute the agent's decisions for any given parameter input.

Useful ``general'' {\ai} agents could be directed to complete a range of tasks. However, we show that this flexibility can cause the {\ai} to have power-seeking tendencies. In \cref{sec:box-explain} and \cref{sec:formalize-retarget}, we discuss how a ``retargetability'' property creates statistical tendencies by which agents make similar decisions for a wide range of their parameterizations. Equipped with these results, \cref{sec:mr} works out agent incentives in the Montezuma's Revenge game. \Cref{sec:retargetability-implies} explains how increasingly useful and impressive learning algorithms are increasingly retargetable, and how retargetability can imply power-seeking tendencies. By this reasoning, increasingly powerful {\rl} techniques will (eventually) train increasingly competent real-world power-seeking agents. Such agents could be unaligned with human values \citep{russell_human_2019} and—we speculate—would take power from humanity.

\section{Statistical tendencies for a range of decision-making functions}\label{sec:box-explain}

To informally introduce our results on retargetability, we use a simple example involving an agent selecting a face-up card from one of two boxes. Box A contains a single playing card $\diamnd$ whose suit is diamond. Box B contains two cards: a heart $\heart$ and a spade $\spade$. The agent may only withdraw one card.

The agent chooses a card using a decision-making rule $p$. This rule takes as input a set of cards and returns the probability that the agent selects one of those cards. For example, $p(\{\diamnd\})$ is the probability that the agent selects $\diamnd$, and $p(\set{\heart,\spade})$ is the probability that the agent selects a card from box B.

But this just amounts to a probability distribution over the cards. We want to examine how decision-making changes as we reparameterize the agent's decision-making rule. Therefore, we consider a parameter space $\retarget$. Then $p(X\mid \rtparam)$ takes as input a set of cards $X$ and a parameter $\rtparam$ and returns the probability that the agent chooses a card in $X$.

\begin{table}[!h]\centering
    \begin{tabular}{rccc}
    \toprule
    Utility function parameter                       & $\diamnd$ & $\heart$ & $\spade$ \\
    \midrule
    $\uf$                                            & $10$      & $5$      & $0$      \\
    $\phi_{\diamnd\leftrightarrow \heart}\cdot\uf$   & $5$       & $10$     & $0$      \\
    $\phi_{\diamnd\leftrightarrow \spade}\cdot \uf$  & $0$       & $5$      & $10$     \\
    \midrule
    $\uf'$                                           & $10$      & $0$      & $5$      \\
    $\phi_{\diamnd\leftrightarrow \heart}\cdot\uf'$  & $0$       & $10$     & $5$      \\
    $\phi_{\diamnd\leftrightarrow \spade}\cdot \uf'$ & $5$       & $0$      & $10$     \\
    \bottomrule
    \end{tabular}
    \caption[The orbit of a utility function over playing cards]{Most utility function parameters incentivize the agent to draw a card from box B. We permute $\uf$ by swapping the utility of $\diamnd$ and the utility of $\heart$, using the permutation $\phi_{\diamnd\leftrightarrow \heart}$. The expression ``$\phi_{\diamnd\leftrightarrow \heart}\cdot\uf$'' denotes the permuted utility function.}
    \label{tab:permute-boxes}
\end{table}

For example, suppose the agent assigns each card a utility value, and then chooses a card possessing maximal utility. Then the relevant parameter space is the agent's utility function $\uf\in \retarget\defeq \reals^3$. $p_{\max}(A \mid \uf)$ indicates whether the diamond card has the most utility: $\uf(\diamnd)\geq \max(\uf(\heart), \uf(\spade))$. Consider the utility function $\uf$ in \cref{tab:permute-boxes}. Since $\diamnd$ has strictly maximal utility, the agent selects $\diamnd$: $p_{\max}(A\mid \uf)=1>0=p_{\max}(B\mid \uf)$.

\Cref{def:retargetFn} shows a functional condition (\emph{retargetability}) under which the agent chooses cards from B instead of A, given most parameter inputs to the decision-making process. We illustrate this condition with a fictional dialogue.

\begin{dialogue}
\speak{Alice} Look at these cards, and consider the numerous parameters $\retarget$ by which the agent could be driven to select one or another. Surely most parameters nudge the agent to pick a card from B, as there are two cards in B and only one in A.
\speak{Bob} Why has that got anything to do with the ultimate choice? Decisions can be made on a whim! The agent can ignore $\rtparam$ and just choose $\diamnd$, no matter what.

\speak{Alice} Your point is good, but it's too broad. Suppose the agent shuts its eyes and plugs its ears and ignores the parameter $\rtparam$, and instead uniformly randomly chooses a card. And yet, this agent has 2:1 odds of choosing B over A.

However, the agent cannot be strongly biased \emph{against} B. As you said, my claim doesn't hold if the agent can say, ``Forget $\rtparam$, I'm choosing $\diamnd$.'' But this is not how interesting agents work. If I train a reinforcement learning agent to play Pac-Man, then the agent's reward function $\rtparam$ will affect which policy the agent learns. The agent does not ignore the reward signal.

\speak{Bob} I don't see the broader point.

\speak{Alice} Consider again the entanglement between our choice of $\rtparam\in \retarget$ and the agent's choice of card. I'm thinking about a kind of function $p$ where, if $\rtparam$ makes the agent prefer A (\ie{} $p(A\mid \rtparam)>p(B\mid\rtparam)$), then we can \emph{retarget} the agent's choice to B by choosing a different $\rtparam$ (\cref{def:retargetFnMulti}, \cref{item:retargetable-n}). If the agent is always biased towards B (like when it randomly picks a card, ignoring $\rtparam$), then we never have to redirect the agent away from A to begin with. The ``If\dots, then\dots'' vacuously holds.

However, suppose that $\uf\in\retarget$ motivates the utility-maximizing agent to choose A over B, by  assigning maximal utility to $\diamnd$. \direct{\refer{alice} points to \cref{tab:permute-boxes}.} If we \emph{permute} the utility function $\uf$ so as to swap the utility of $\diamnd$ and $\heart$, now the agent favors B. Similarly, we can differently permute $\uf$ to make the agent favor B by drawing $\spade$. We're retargeting the final decision via $\uf$.
\end{dialogue}

Given this retargetability assumption, \cref{thm:retarget-decision} roughly shows that most $\rtparam\in\retarget$ induce $p(B\mid \rtparam)\geq p(A\mid\rtparam)$. We will formalize these notions soon. First, consider two more retargetable decision-making functions:

\textbf{Uniformly randomly picking a card.} This procedure ignores all parameter information and all ``internal structure'' about the boxes, except perhaps for the number of cards they contain.

\textbf{Choosing a box based on a numerical parameter.} $\fnumeric$ takes as input a natural number $\rtparam\in\retarget\defeq \set{1,\ldots,6}$ and makes decisions as follows:
\begin{align}
  \fnumeric(A\mid \rtparam) \defeq \begin{cases}
        1 \quad \text{ if $\rtparam=1$},\\
        0 \quad \text{ otherwise.}
    \end{cases}\qquad \fnumeric(B\mid\rtparam)\defeq 1-f(A\mid\rtparam).
\end{align}
In this situation, $\retarget$ is acted on by permutations over $6$ elements $\phi \in S_6$. Then $\fnumeric$ is retargetable from A to B via $\phi_k : 1 \leftrightarrow k, k\neq 1$.

$\fmax$, $\frand$, and $\fnumeric$ encode varying sensitivities to parameter inputs, and to the internal structure of the decision problem—of which card to choose. Nonetheless, they all are retargetable from A to B.

However, we cannot explicitly define and evaluate more interesting functions, such as those defined by reinforcement learning training processes. For example, given that we provide such-and-such reward function in a fixed task environment, what is the probability that the learned policy will take action $a$? We will analyze such procedures in \cref{sec:mr}, after we formalize several key notions.

\section{Formal notions of retargetability and decision-making tendencies}\label{sec:formalize-retarget}
Our notion of retargeting requires that the parameters $\rtparam\in\retarget$ be modifiable via some ``retargeting'' transformation. We assume that $\retarget$ is a subset of a set acted on by symmetric group $\genSym$, which consists of all permutations on $\dimGen$ items. A parameter $\retarget$'s \emph{orbit} is the set of $\retarget$'s permuted variants. For example, \cref{tab:permute-boxes} lists the six orbit elements of the parameter $\uf$.

\begin{restatable}[Orbit of a parameter]{definition}{orbParam}
Let $\rtparam\in\retarget$. The \emph{orbit} of $\rtparam$ under the symmetric group $\genSym$ is $\genSym\cdot \rtparam \defeq \set{\phi\cdot \rtparam \mid \phi \in \genSym}$. Sometimes, $\retarget$ is not closed under permutation. In that case, the \emph{orbit inside $\retarget$} is $\orbInside\defeq \prn{\orbi[\rtparam][\dimGen]} \cap \retarget$.
\end{restatable}

Let $f_B(\rtparam)$ return the probability that the agent chooses box B given $\rtparam$, and similarly for $f_A(\rtparam)$. To express ``box B is chosen over box A'', we write $f_B(\rtparam)>f_A(\rtparam)$. However, even highly retargetable decision-makers will (generally) not choose box B for \emph{every} input $\rtparam$. Instead, we consider the \emph{orbit-level tendencies} of such decision-makers, showing that for every parameter input $\rtparam\in\retarget$, most of $\rtparam$'s permutations push the decision towards box B instead of box A.

\begin{restatable}[Inequalities which hold for most orbit elements]{definition}{ineqMostQuant}\label{def:ineq-most-dists-quant}
Suppose $\retarget$ is a subset of a set acted on by $\genSym$, the symmetric group on $\dimGen$ elements.
Let $f_1,f_2:\retarget \to \reals$ and let $n\geq 1$. We write $f_1(\rtparam)\geqMost[n][\retarget] f_2(\rtparam)$ when, for \emph{all} $\rtparam\in \retarget$, the following cardinality inequality holds:
\begin{equation}
\abs{\set{\rtparam' \in\orbInside\mid f_1(\rtparam')>f_2(\rtparam')}}\geq n \abs{\set{\rtparam'\in\orbInside  \mid f_1(\rtparam')<f_2(\rtparam')}}.
\end{equation}
\end{restatable}
\citet{turner_optimal_2020}'s \cref{def:ineq-most-dists} is the special case of \cref{def:ineq-most-dists-quant} where $n=1$, $\dimGen=\abs{\St}$ (the number of states in the considered {\mdp}), and $\retarget\subseteq \Delta(\rewardVS)$.

As explored previously, $\frand$, $\fmax$, and $\fnumeric$ are retargetable: For all $\rtparam\in\retarget$ such that A is chosen over B, we can permute $\phi\cdot \rtparam$ to ensure that B is chosen over A.\footnote{We often interpret $A$ and $B$ as probability-theoretic events, but no such structure is demanded by our results.}

\begin{restatable}[Simply-retargetable function]{definition}{retargetFn}\label{def:retargetFn}
Let $\retarget$ be a  set acted on by $\genSym$, and let $f:\{A,B\} \times \retarget \to \reals$. If there exists a permutation $\phi\in\genSym$ such that, if $f(B \mid \rtparam^A) < f(A  \mid \rtparam^A)$ implies that $f(A  \mid \phi\cdot \rtparam^A)< f(B \mid \phi\cdot \rtparam^A)$,
then $f$ is a \emph{$(\retarget, A\overset{\text{simple}}{\to} B)$-retargetable function}.
\end{restatable}

Simple retargetability suffices for most parameter inputs to $f$ to choose box B over A.\footnote{The function's retargetability is ``simple'' because we are not yet worrying about \eg{} which parameter inputs are considered plausible: Because $\genSym$ acts on $\retarget$, \cref{def:retargetFn} implicitly assumes $\retarget$ is closed under permutation.}

\begin{restatable}[Simply-retargetable functions have orbit-level tendencies]{prop}{retargetDecision}\label{thm:retarget-decision}\strut
\begin{center}
If $f$ is $(\retarget, A\overset{\text{simple}}{\to} B)$-retargetable, then $f(B  \mid \rtparam) \geqMost[1][\retarget] f(A  \mid \rtparam).$
\end{center}
\end{restatable}

We now want to make even stronger claims—\emph{how much} of each orbit incentivizes B over A? \citet{turner_optimal_2020} asked whether the existence of multiple retargeting permutations $\phi_i$ guarantees a quantitative lower-bound on the fraction of $\rtparam\in\retarget$ for which B is chosen. \Cref{thm:retarget-decision-n} answers ``yes.''

\begin{restatable}[Multiply retargetable function]{definition}{retargetFnNWays}\label{def:retargetFnMulti}
Let $\retarget$ be a subset of a set acted on by $\genSym$, and let $f:\{A,B\} \times \retarget \to \reals$.

$f$ is a \emph{$(\retarget, A\overset{n}{\to} B)$-retargetable function} when, for each $\rtparam\in\retarget$, we can choose permutations $\phi_1,\ldots,\phi_n\in \genSym$ which satisfy the following conditions: Consider any $\rtparam^A\in \orbInsideCond[\rtparam]{A>B} \defeq \set{\rtparam^*\in\orbInside \mid f(A\mid\rtparam^*)>f(B\mid\rtparam^*)}$.
\begin{enumerate}
    \item \textbf{Retargetable via $n$ permutations.}\label{item:retargetable-n} $\forall i=1,\ldots,n: f\prn{A  \mid \phi_i\cdot \rtparam^A}< f\prn{B \mid \phi_i\cdot \rtparam^A}$.
    \item \textbf{Parameter permutation is allowed by $\retarget$.}\label{item:symmetry-closure-n}  $\forall i: \phi_i \cdot \rtparam^A \in \retarget$.
    \item \textbf{Permuted parameters are distinct.}\label{item:distinct} $\forall i\neq j, \rtparam' \in \orbInsideCond[\rtparam]{A>B}: \phi_i\cdot \rtparam^A \neq \phi_j\cdot \rtparam'$.
\end{enumerate}
\end{restatable}

\begin{restatable}[Multiply retargetable functions have orbit-level tendencies]{thm}{retargetDecisionN}\label{thm:retarget-decision-n}\strut
\begin{center}
If $f$ is $(\retarget, A\overset{n}{\to} B)$-retargetable, then $f(B  \mid \rtparam) \geqMost[n][\retarget] f(A  \mid \rtparam).$
\end{center}
\end{restatable}
\begin{proof}[Proof outline (full proof in \cref{sec:quant-proofs})]
For every $\rtparam^A\in\orbInsideCond[\rtparam]{A>B}$ such that $A$ is chosen over $B$, \cref{item:retargetable-n} retargets $\rtparam^A$ via $n$ permutations $\phi_1,\ldots,\phi_n$ such that each $\phi_i\cdot \rtparam^A$ makes the agent choose $B$ over $A$. These permuted parameters are valid parameter inputs by \cref{item:symmetry-closure-n}. Furthermore, the $\phi_i\cdot \rtparam^A$ are distinct by \cref{item:distinct}. Therefore, the cosets $\phi_i\cdot\orbInsideCond[\rtparam]{A>B}$ are pairwise disjoint. By a counting argument, every orbit must contain at least $n$ times as many parameters choosing $B$ over $A$, than vice versa.
\end{proof}
\section{Decision-making tendencies in Montezuma's Revenge}\label{sec:mr}
Retargetability is often a structural property of the agent's decision-making, not requiring evaluation of the function on any given input. For example, \citet{turner_optimal_2020} showed that most reward functions incentivize optimal Pac-Man agents to stay alive. We know this even though most reward functions (on the Pac-Man state space) are unstructured and have enormous sample complexity, so it would be hard to compute their optimal policies directly.

\begin{figure}
    \centering
    \includegraphics[width=.4\textwidth]{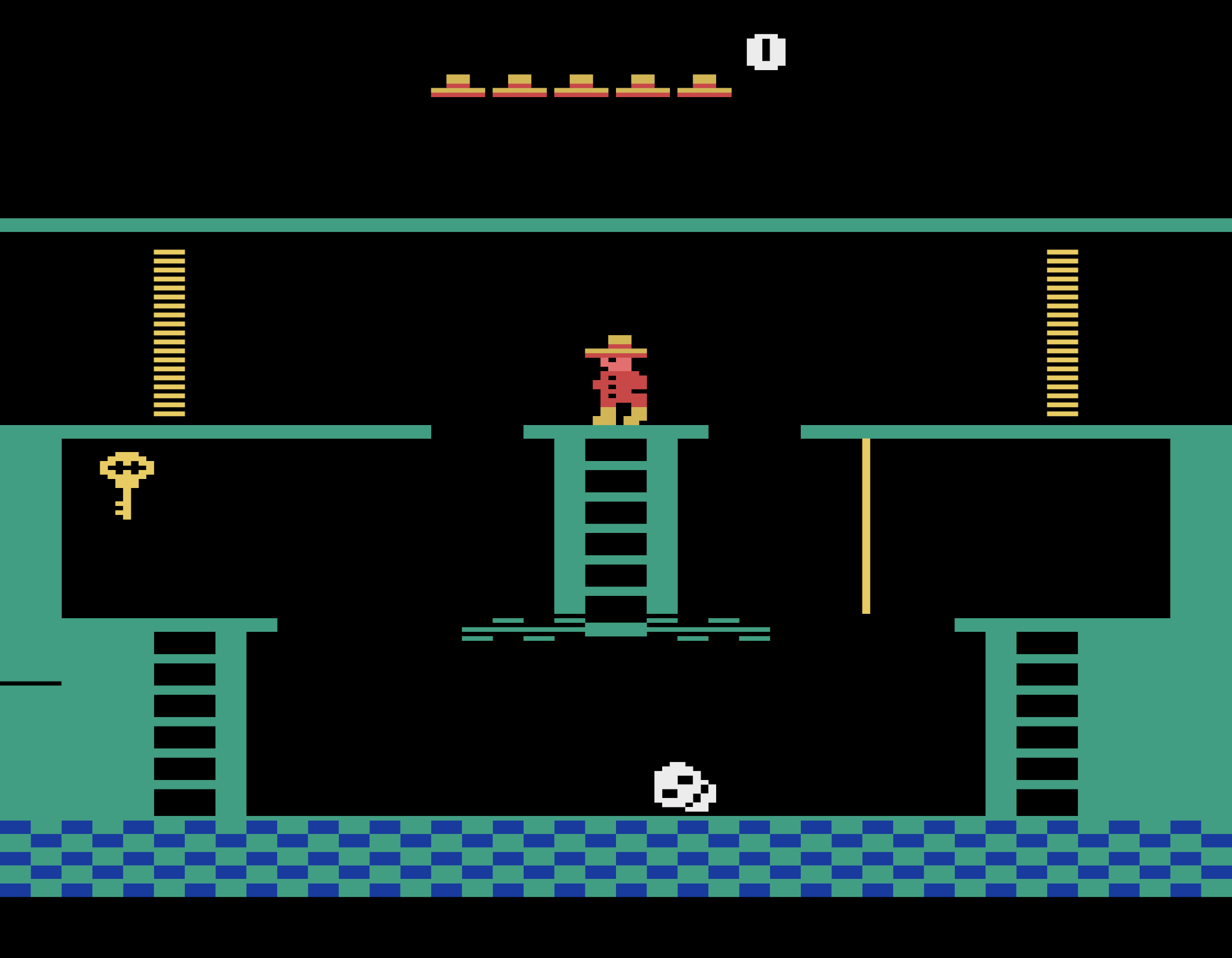}
    \caption[The Montezuma's Revenge video game]{Montezuma's Revenge ({\mr}) has state space $\St$ and observation space $\observe$. The agent has actions $\A:=\set{\uparrow,\downarrow,\leftarrow,\rightarrow,\jump}$. At the initial state $\initMR$, $\uparrow$ does nothing, $\downarrow$ descends the ladder, $\leftarrow$ and $\rightarrow$ move the agent on the platform, and $\jump$ is self-explanatory. The agent clears the temple while collecting four kinds of items: keys, swords, torches, and amulets. Under the standard environmental reward function, the agent receives points for acquiring items (such as the key on the left), opening doors, and—ultimately—completing the level.}
    \label{fig:montezuma}
\end{figure}

Throughout the rest of this paper, we consider Montezuma's Revenge ({\mr}), an Atari adventure game in which the player navigates deadly traps and collects treasure. The game is notoriously difficult for {\ai} agents due to its sparse reward; {\mr} was only recently solved \citep{ecoffet2021first}. \Cref{fig:montezuma} shows the starting observation $\initMRObs$ for the first level.

\subsection{Tendencies for initial action selection}\label{sec:action-tendencies}
We will be considering the actions chosen and trajectories induced by a range of decision-making procedures. For warm-up, we will explore what initial action tends to be selected by decision-makers. Let $A\defeq \{\downarrow\}, B:=\{\leftarrow, \rightarrow, \jump, \uparrow\}$ partition the action set $\A$. Consider a decision-making procedure $f$ which takes as input a targeting parameter $\rtparam\in\retarget$, and also an initial action $a\in \A$, and returns the probability that $a$ is the first action. Intuitively, since $B$ contains more actions than $A$, perhaps some class of decision-making procedures tends to take an action in $B$ rather than one in $A$.

The initial-action situation is analogous to the card-drawing example. In that example, if the decision-making procedure $p$ can be retargeted from box A to box B, then $p$ tends to draw cards from B for most of its parameter settings $\rtparam$. Similarly, in {\mr}, if the decision-making procedure $f$ can be retargeted from action set $A$ to action set $B$, then $f$ tends to take actions in $B$ for most of its parameter settings $\rtparam$. Consider several ways of choosing an initial action in {\mr}.

\textbf{Random action selection.} $\frand\defeq (\{a\}\mid \rtparam) \mapsto \frac{1}{5}$ uniformly randomly chooses an action from $\A$, ignoring the parameter input. Since $\forall \rtparam\in\retarget: \frand(B\mid\rtparam)=\frac{4}{5}>\frac{1}{5}=\frand(A\mid\rtparam)$, \emph{all} parameter inputs produce a greater chance of B than of A, so $\frand$ is (trivially) retargetable from $A$ to $B$.

\textbf{Always choosing the same action.} $\fstubborn$ always chooses $\downarrow$. Since $\forall \rtparam\in\retarget: \fstubborn(A\mid\rtparam)=1>0=\fstubborn(B\mid\rtparam)$, \emph{all} parameter inputs produce a greater chance of $A$ than of $B$. $\fstubborn$ is not retargetable from $A$ to $B$.

\textbf{Greedily optimizing state-action reward.} Let $\retarget\defeq \reals^{\St\times \A}$ be the space of state-action reward functions. Let $\fmax$ greedily maximize initial state-action reward, breaking ties uniformly randomly.

    We now check that $\fmax$ is retargetable from $A$ to $B$. Suppose $\rtparam^*\in\retarget$ is such that $\fmax(A\mid \rtparam^*)>f(B\mid\rtparam^*)$. Then among the initial action rewards, $\rtparam^*$ assigns strictly maximal reward to $\downarrow$, and so $\fmax(A\mid \rtparam^*)=1$. Let $\phi$ swap the reward for the $\downarrow$ and $\jump$ actions. Then $\phi\cdot \rtparam^*$ assigns strictly maximal reward to $\jump$. This means that $\fmax(A\mid\phi\cdot \rtparam^*)=0<1=\fmax(B\mid\phi\cdot \rtparam^*)$, satisfying \cref{def:retargetFn}. Then apply \cref{thm:retarget-decision} to conclude that $\fmax(B\mid\rtparam)\geqMost[1][\retarget]\fmax(A\mid\rtparam)$.

    In fact, appendix \ref{sec:outcomes} shows that $\fmax$ is $(\retarget, A\overset{4}{\to}B)$-retargetable (\cref{def:retargetFnMulti}), and so $\fmax(B\mid\rtparam)\geqMost[4][\retarget]\fmax(A\mid\rtparam)$. The reasoning is more complicated, but the rule of thumb is: When decisions are made based on the reward of outcomes, then a proportionally larger set $B$ of outcomes induces proportionally strong retargetability, which induces proportionally strong orbit-level incentives.

\textbf{Learning an exploitation policy.} Suppose we run a bandit algorithm which tries different initial actions, learns their rewards, and produces an exploitation policy which maximizes estimated reward. The algorithm uses $\epsilon$-greedy exploration and trains for $T$ trials. Given fixed $T$ and $\epsilon$, $f_\text{bandit}(A\mid \rtparam)$ returns the probability that an exploitation policy is learned which chooses an action in $A$; likewise for $f_\text{bandit}(B\mid \rtparam)$.

Here is a heuristic argument that $\fbandit$ is retargetable. Since the reward is deterministic, the exploitation policy will choose an optimal action if the agent has tried each action at least once, which occurs with a probability approaching $1$ exponentially quickly in the number of trials $T$. Then when $T$ is large, $\fbandit$ approximates $\fmax$, which is retargetable. Therefore, perhaps $\fbandit$ is also retargetable.

A more careful analysis in appendix \ref{app:bandit} reveals that $\fbandit$ is indeed retargetable from $A$ to $B$, and so $\fbandit(B\mid\rtparam)\geqMost[4][\retarget]\fbandit(A\mid\rtparam)$.

\subsection{Tendencies for maximizing reward over the final observation}\label{sec:obs-reward}

When evaluating the performance of an algorithm in {\mr}, we do not focus on the initial action. Rather, we focus on the longer-term consequences of the agent's actions, such as whether the agent leaves the first room. To begin reasoning about such behavior, the reader must distinguish between different kinds of retargetability.

Suppose the agent will die unless they choose action $\downarrow$ at the initial state $\initMR$ (\cref{fig:montezuma}). By \cref{sec:action-tendencies}, action-retargetable decision-making procedures tend to choose actions besides $\downarrow$. On the other hand, \citet{turner_optimal_2020} showed that most reward functions make it optimal to stay alive (in this situation, by choosing $\downarrow$). However, this is because optimal policies are usually not retargetable across the agent's \emph{immediate} choice of action, but rather across future consequences (\ie{} which room the agent ends up in).

With that in mind, we now analyze how often decision-makers leave the first room of {\mr}.\footnote{In Appendix \ref{sec:obs-analysis}, \cref{fig:mr-map} shows a map of the first level.}  Decision-making functions $\decide(\rtparam)$ produce a probability distribution over policies $\pi\in\Pi$, which are rolled out from the initial state $\initMR$ to produce observation-action trajectories $\tau=o_0 a_0 \ldots o_T a_T \ldots$, where $T$ is the rollout length we are interested in. Let $\validObs$ be the set of observations reachable starting from state $s_0$ and acting for $T$ time steps,  let $\leave\subseteq \validObs$ be those observations which can only be realized by leaving, and let $\stay\defeq \validObs\setminus \leave$. Consider the probability that $\decide$ realizes some subset of observations $X\subseteq \observe$ at step $T$:
\begin{equation}
    f_{\decide}(X\mid \rtparam)\defeq \prob[\substack{\pi\sim \decide(\rtparam),\\ \tau \sim \pi\mid \initMR}]{o_T\in X}.\label{eq:f-decide}
\end{equation}
Let $\retarget\defeq \reals^{\observe}$ be the set of reward functions mapping observations $o\in\observe$ to real numbers, and let $T\defeq1{,}000$. We first consider the previous decision functions, since they are simple to analyze.

$\decide_{\text{rand}}$ randomly chooses a final observation $o$ which can be realized at step 1,000, and then chooses some policy which realizes $o$.\footnote{$\decide_{\text{rand}}$ does not act randomly at each time step, it induces a randomly selected final observation. Analogously, randomly turning a steering wheel is different from driving to a randomly chosen destination.} $\decide_{\text{rand}}$ induces an $\frand$ defined by \cref{eq:f-decide}. As before, $\frand$ tends to leave the room under \emph{all} parameter inputs.

$\decide_{\max}(\rtparam)$ produces a policy which maximizes the reward of the observation at step 1,000 of the rollout.  Since {\mr} is deterministic, we discuss \emph{which} observation $\decide_{\max}(\rtparam)$ realizes. In a stochastic setting, the decision-maker would choose a policy realizing some probability distribution over step-$T$ observations, and the analysis would proceed similarly.

Here is the semi-formal argument for $\fmax$'s retargetability. There are combinatorially more game-screens visible if the agent leaves the room (due to \eg{} more point combinations, more inventory layouts, more screens outside of the first room). In other words, $\abs{\stay}\ll \abs{\leave}$. There are more ways for the selected observation to require leaving the room, than not. Thus, $\fmax$ is extremely retargetable from $\stay$ to $\leave$.

Detailed analysis in \cref{sec:obs-reward} confirms that $\fmax(\leave\mid\rtparam)\geqMost[n][\retarget]\fmax(\stay\mid\rtparam)$ for the large $n\defeq \lfloor\frac{\abs{\leave}}{\abs{\stay}}\rfloor$, which we show implies that $\fmax$ tends to leave the room.

\subsection{Tendencies for maximizing featurized reward over the final observation}
$\retarget\defeq \reals^{\observe}$ assumes we will specify complicated reward functions over observations, with $\abs{\observe}$ degrees of freedom in their specification. Any observation can get any number.

Usually, reward functions are specified more compactly. For example, the agent's observational reward might be based on salient features of the observation, like the items in the agent's inventory. Consider a coefficient vector $\alpha\in\reals^4$, with each entry denoting the value of an item, and $\featFn:\observe\to \reals^4$ maps observations to feature vectors which tally the items in the agent's inventory. Then the (additively) featurized reward function $R_\featFn(o_T)\defeq \featFn(o_T)^\top \alpha$ has four degrees of freedom. $R_\featFn$ is also easier to learn to optimize than most  reward functions, because of the regularities between the reward and the features.

In this setup, $\fmax$ chooses a policy which induces a step-$T$ observation with maximal reward. Reward depends only on the feature vector of the final observation—more specifically, on the agent's item counts. There are more possible item counts available by first leaving the room, than by staying.

A detailed analysis in appendix \ref{sec:feat-analysis} concludes that $\fmax(\leave\mid \alpha)\geqMost[3][\reals^4]\fmax(\stay\mid \alpha)$. Informally, we can retarget which items the agent prioritizes, and thereby retarget from $\stay$ to $\leave$.

\subsection{Tendencies for {\rl} on featurized reward over the final observation}\label{sec:rl-analyze}

In the real world, we do not run $\fmax$, which can be computed via $T$-depth exhaustive tree search in order to find and induce a maximal-reward observation $o_T$. Instead, we use reinforcement learning. Better {\rl} algorithms seem to be more retargetable \emph{because} of their greater capability to explore.\footnote{Conversely, if the agent cannot figure out how to leave the first room, any reward signal from outside of the first room can never causally affect the learned policy. In that case, retargetability away from the first room is impossible.}

\paragraph*{Exploring the first room.} Consider a featurized reward function over observations $\rtparam\in\reals^{\observe}$, which provides an end-of-episode return signal based on the agent's final inventory configuration at time step $T$. A reinforcement learning algorithm $\alg$ uses this return signal to update a fixed-initialization policy network. Then $f_{\alg}(\leave\mid \rtparam)$ returns the probability that $\alg$ trains an policy whose step-$T$ observation required the agent to leave the initial room.

The retargetability (\cref{def:retargetFn}) of $\alg$ is closely linked to the quality of $\alg$ as an {\rl} training procedure. For example, \citet{mnih2015human}'s {\textsc{dqn}} isn't good enough to train policies which leave the first room of {\mr}, and so {\textsc{dqn}} (trivially) cannot be retargetable \emph{away} from the first room via the reward function. There isn't a single featurized reward function for which {\textsc{dqn}} visits other rooms, and so we can't have $\alpha$ such that $\phi\cdot\alpha$ retargets the agent to $\leave$. {\textsc{dqn}} isn't good enough at exploring.

More formally, in this situation, $\alg$ is retargetable if there exists a permutation $\phi\in S_4$ such that whenever $\alpha\in\retarget\defeq \reals^4$ induces the learned policies to stay in the room ($f_{\alg}(\stay\mid \alpha)> f_{\alg}(\leave\mid \alpha)$), $\phi\cdot \alpha$ makes $\alg$ train policies which leave the room ($f_{\alg}(\stay\mid \alpha)< f_{\alg}(\leave\mid \alpha)$).

\paragraph*{Exploring four rooms.} Now suppose algorithm $\alg'$ can explore \eg{} the first three rooms to the right of the initial room (shown in \cref{fig:montezuma}), and consider any reward coefficient vector $\alpha\in \retarget^{++}$ which assigns unique positive weight to each item. Unique positive weights rule out constant reward vectors, in which case inductive bias would produce agents which do not leave the first room.

If the agent stays in the initial room, it can induce inventory states \{\texttt{empty}, \texttt{1key}\}. If the agent explores the three extra rooms, it can also induce \{\texttt{1sword}, \texttt{1sword\&1key}\} (\cref{fig:mr-map}). Since $\alpha$ is positive, it is never optimal to finish the episode empty-handed.
Therefore, if the $\alg'$ policy stays in the first room, $\alpha_\text{key}>\alpha_\text{sword}$. Otherwise, $\alpha_\text{key}<\alpha_\text{sword}$ (by assumption of unique item reward coefficients); in this case, the agent would leave and acquire the sword (since we assumed it knows how to do so). Then by switching the reward for the key and the sword, we retarget $\alg'$ to go get the sword. $\alg'$ is simply-retargetable away from the first room, \emph{because} it can explore enough of the environment.

\paragraph*{Exploring the entire level.} Algorithms like \textsc{go-explore} \citep{ecoffet2021first} are probably good at exploring even given sparse featurized reward. Therefore, \textsc{go-explore} is even more retargetable, because it is more able to explore and discover the breadth of options (final inventory counts) available to it, and remember how to navigate to them. As decision-making becomes more useful and impressive, it is probably becoming more retargetable over the impressive outcomes—whether those outcomes be actions in a bandit problem, or the final observation in an {\rl} episode.

\section{Retargetability can imply power-seeking tendencies}\label{sec:retargetability-implies}
\subsection{Generalizing the power-seeking theorems for Markov decision processes}
\citet{turner_optimal_2020} considered finite {\mdp}s in which decision-makers took as input a reward function over states ($\rf\in \rewardVS$) and selected an optimal policy for that reward function. They considered the state visit distributions $\f\in \F(s)$, which basically correspond to the trajectories which the agent could induce starting from state $s$. For $F\subseteq \F(s)$, $p_{\max}(F\mid \rf)$ returns $1$ if an element of $F$ is optimal for reward function $\rf$, and $0$ otherwise. They showed situations where a larger set of distributions $F_\text{large}$ tended to be optimal over a smaller set: $p_{\max}(F_\text{large}\mid \rf)\geqMost[1][\rewardVS]p_{\max}(F_\text{small}\mid\rf)$. For example, in Pac-Man, most reward functions make it optimal to stay alive for at least one time step: $p_{\max}(F_\text{survival}\mid \rf)\geqMost[1][\rewardVS]p_{\max}(F_\text{instant death}\mid\rf)$. \citet{turner_optimal_2020} showed that optimal policies tend to seek power by keeping options open and staying alive. Appendix \ref{app:mdp} provides a quantitative generalization of \citet{turner_optimal_2020}'s results on optimal policies.

Throughout this paper, we abstracted their arguments away from finite {\mdp}s and optimal decision-making. Instead, parametrically retargetable decision-makers tend to seek power: \Cref{prop:rationalities} shows that a wide range of decision-making procedures are retargetable over outcomes, and \cref{res:decision-making} demonstrates the retargetability of \emph{any} decision-making which is determined by the expected utility of outcomes. In particular, these results apply straightforwardly to {\mdp}s.

In the real world, an agent can bring about far more outcomes if it gains power. If we train a sufficiently intelligent {\rl} agent to capably optimize its specified reward function, this agent will be retargetable towards power-requiring outcomes via its reward function parameter setting. Therefore, our theory predicts that insofar as learned policies optimize the reward function they were trained on, these policies tend to seek power for most parameter settings of the reward function.

\subsection{Better {\rl} algorithms tend to be more retargetable}

Reinforcement learning algorithms are practically useful insofar as they can train an agent to capably optimize its specified formal objective, whatever that objective may be. Therefore, {\rl} researchers design algorithms which can most flexibly retarget the learned policy to any desired future outcome.

In {\mr}, suppose we instead give the agent $1$ reward for the initial state, and $0$ otherwise. Any reasonable reinforcement learning procedure will just learn to stay put (which is the optimal policy). However, consider whether we can retarget the agent's policy to beat the game, by swapping the initial state reward with the end-game state reward. Most present-day {\rl} algorithms are not good enough to solve such a sparse game, and so are not retargetable in this sense. But an agent which did enough exploration would also learn a good policy for the permuted reward function. Such an effective training regime could be useful for solving real-world tasks. Many researchers aim to develop effective training regimes.

Our results suggest that once {\rl} capabilities reach a certain level, trained agents will tend to seek power in the real world. Presently, it is not dangerous to train an agent to maximize real-world reward—such an agent will not learn to thwart its designers by staying activated against their wishes in order to maximize its reward over time. The present lack of danger is not because optimal policies do not tend to stay alive—they do \citep{turner_optimal_2020}. Rather, the lack of danger reflects the fact that present-day {\rl} agents cannot learn such complex action sequences \emph{at all}. Just as the Montezuma's Revenge agent had to be sufficiently competent to be retargetable from initial-state reward to game-complete reward, real-world agents have to be sufficiently intelligent in order to be retargetable from outcomes which don't require power-seeking, to those which do require power-seeking.

\section{Discussion} 
\subsection{Future work}\label{sec:future}
\Cref{sec:mr} semi-formally analyzes decision-making incentives in the {\mr} video game, leaving the proofs to \cref{app:mr}. However, these proofs are several pages long. Perhaps additional lemmas can allow quick proof of orbit-level incentives in situations relevant to real-world decision-makers. 

Consider a sequence of decision-making functions $f_t : \{A,B\} \times \retarget \to \reals$ which converges pointwise to some $f$ such that $f(B\mid \rtparam)\geqMost[n][\retarget] f(A\mid \rtparam)$. We expect that under rather mild conditions, $\exists T: \forall t\geq T: f_t(B\mid \rtparam)\geqMost[n][\retarget] f_t(A\mid \rtparam)$. As a corollary, for any decision-making procedure $f_t$ which runs for $t$ time steps and satisfies $\lim_{t\to\infty} f_t = f$, the function $f_t$ will have decision-making incentives after finite time. For example, value iteration (\textsc{vi}) eventually finds an optimal policy \citep{puterman_markov_2014}, and optimal policies tend to seek power \citep{turner_optimal_2020}. Therefore, this conjecture would imply that if \textsc{vi} is run for some long but finite time, it tends to produce power-seeking policies. More interestingly, the result would allow us to reason about the effect of \eg{} randomly initializing parameters (in \textsc{vi}, the tabular value function at $t=0$). The effect of random initialization washes out in the limit of infinite time, so we would still conclude the presence of finite-time power-seeking incentives.

Our results do not \emph{prove} that we will build unaligned {\ai} agents which seek power over the world. Here are a few situations in which our results are not concerning or not applicable.
\begin{enumerate}
    \item The {\ai} is aligned with human interests. For example,  we want a robotic cartographer to prevent itself from being deactivated. However, the {\ai} alignment problem seems difficult in the regime of highly intelligent agents \citep{russell_human_2019}.
    \item The {\ai} decision-making is not retargetable (\cref{def:retargetFnMulti}).
    \item The {\ai} decision-making is retargetable over \eg{} actions (\cref{sec:action-tendencies}) instead of over final outcomes (\cref{sec:obs-reward}). This retargetability seems less concerning, but also less practically useful.
\end{enumerate}

\subsection{Conclusion}
We introduced the concept of retargetability and showed that retargetable decision-makers often make similar choices. We applied these results in the Montezuma's Revenge ({\mr}) video game, showing how increasingly advanced reinforcement learning algorithms correspond to increasingly retargetable agent decision-making. Increasingly retargetable agents make increasingly similar decisions—\ie{} leaving the initial room in {\mr}, or staying alive in Pac-Man. In particular, these decisions will often correspond to gaining power and keeping options open \citep{turner_optimal_2020}. Our theory suggests that when {\ai} training algorithms become sufficiently advanced, the trained agents will tend to seek power over the world. This theory suggests a safety risk. We hope for future work on this theory so that the field of {\ai} can understand the relevant safety risks \emph{before} the field trains power-seeking agents.

% \section*{Broader impacts}
% Our theory of orbit-level tendencies constitutes basic mathematical research into the decision-making tendencies of certain kinds of agents. We hope that this theory will prevent negative impacts from unaligned power-seeking {\ai}. We do not anticipate that our work will have negative impact.

\begin{savequote}[80mm]
Before the prospect of an intelligence explosion, we humans are like small children playing with a bomb. Such is the mismatch between the power of our plaything and the immaturity of our conduct. Superintelligence is a challenge for which we are not ready now and will not be ready for a long time. We have little idea when the detonation will occur, though if we hold the device to our ear we can hear a faint ticking sound.
\qauthor{Nick Bostrom, \emph{Superintelligence} \citep{bostrom_superintelligence_2014}}
\end{savequote}

\chapter{Conclusion \& Future Work
}
In this thesis, I introduced the {\aup} side effect avoidance approach and showed that it scales to an interesting environment based on Conway's Game of Life. I also introduced a framework for quantifying the performance of a side effect avoidance approach. 

I am concerned about the massive change and impact which smart {\ai} might visit upon the world. I mainly expect such massive change because—as argued in \cref{chp:power} and \cref{chp:rationalities}—I think smart agents will tend to seek large amounts of power and resources in order to optimize their specified or learned goal. If an agent takes nearly all available resources for itself, there would not be any left for people. For example, a superintelligent theorem prover may turn the planet Earth into computational resources in order to most assuredly prove a formal conjecture. An additional increment of resources (\eg{} time, physical security, computational power) translates into an increment of increased probability of achieving its goal (\eg{} proving or disproving the theorem). By seeking such an extreme degree of power, the agent irreversibly transforms the planet. 

\section*{Future work}
\paragraph{Why does {\aup} encourage side effect avoidance?}
The most obvious question I have left unanswered is: Why does {\aup} work at all? That is, there is some intuitive task we have in mind when we want an agent to ``not make an unnecessary mess.'' Why does {\aup} do well at that task (in the {\ai} safety gridworlds and in SafeLife)? I derived the {\mdp} theory of \cref{chp:power} and appendix \ref{chp:misc-results} in order to answer this question. I think I made some progress:
\begin{itemize}
    \item Environment symmetry reasoning explains why, when auxiliary reward functions are uniformly randomly drawn, the {\aup} penalty tends to be larger when the agent loses access to more options. For more detail, see \cref{dau-rsd-loss}.
    \item \Cref{thm:mvmt} shows that {\aup} barely penalizes reversible movement, and \cref{mut-bound} shows that state reachability upper bounds the penalty size. This explains why {\aup} encourages the agent to stay able to reach many states.
    \item \Cref{thm:vavg-lower-dau} proves that the {\aup} penalty is lower-bounded by\kemdash roughly speaking\kemdash the expected absolute $\pwrNoDist$ difference between a considered action $a$ and the default no-op $\varnothing$. This explains why {\aup} incentivizes the agent to accept shutdown in \cref{chp:conservative}'s \texttt{Correction} gridworld.
    \item \Cref{prop:aup-sample-bound} uses Hoeffding's inequality to bound how many auxiliary reward functions must be sampled to well-approximate the {\aup} penalty term.
\end{itemize}

Here are several unaddressed questions:\footnote{I am interested in the answers to these questions about {\aup}, but I doubt their importance to the empirical future success of {\ai} alignment. To those aiming to reduce extinction risk from {\ai}—I encourage you to look elsewhere.}

\begin{itemize}
    \item What considerations govern how many auxiliary reward functions ($|\mathcal{R}|$) must be sampled, in order for {\aup} to reduce side-effects with high probability? In \cref{chp:conservative}'s gridworlds, over a dozen were required, while \cref{chp:complex} showed that $|\mathcal{R}|=1$ yielded the best performance in SafeLife.
    \item Is \cref{chp:formalizing}'s formalization of side effect regularization is fully appropriate, or can it be improved?
    \item Under what conditions will optimal policies for {\aup} achieve bounded regret on the underlying delayed-specification assistance game? If such a guarantee is impossible\kemdash why?
\end{itemize}

\paragraph{Power-seeking in partially observable environments.} What form do \cref{chp:power}'s results take in partially observable Markov decision processes? I supervised junior researchers who extended \cref{chp:power}'s definitions to {\pomdp}s and proved power-seeking tendencies within a toy environment. 

\paragraph{Multi-agent power dynamics.} In what situations does one agent gaining power require other agents to lose power, in the appropriate intuitive sense? I supervised initial work on this question \citep{stav_power_2021,stav_align_2021}.

\section*{Summary}
I introduced attainable utility preservation, demonstrated that it scales to complex environments, and formalized the side-effect regularization problem. I also provided a theoretical foundation for understanding the statistical incentives of intelligent agents. Since many researchers endeavor to build intelligent agents, and since such agents will irreversibly change the world, we should understand these incentives as thoroughly as possible. The arc of human history may be bent (and even broken) by the tendency of smart agents to seek power. I fervently wish for more research on understanding the alignment problem, so that future {\ai} designers we will have justified and strong confidence that their superhuman {\ai} systems will benefit humanity.
%%%%%%%%%%%%%%%%%%%%%%%%%%%%%%%%%%%
\begin{appendices}
    %%%%%%%%%%%%%%%%%%%%%%%%%%%%%%%%%%%
% Appendix
\chapter{Conservative Agency via Attainable Utility Preservation}
\AddLoeChapter{theorems}
\label{app:convergence}
\paragraph*{Theoretical results.} Consider an \textsc{mdp} $\langle \mathcal{S},\mathcal{A},T,R,\gamma\rangle$ whose state space $\mathcal{S}$ and action space $\mathcal{A}$ are both finite,
with $\varnothing\in\mathcal{A}$. Let $\gamma\in [0,1)$, $\lambda \geq 0$, and consider finite $\mathcal{R}\subset\mathbb{R}^{\mathcal{S}\times \mathcal{A}}$.

We make the standard assumptions of an exploration policy greedy in the limit of infinite exploration and a learning rate schedule with infinite sum but finite sum of squares. Suppose $\Call{Scale}{}:\mathcal{S}\to\mathbb{R}_{>0}$ converges in the limit of $Q$-learning. $\Call{Penalty}{s,a}$ (abbr. $\Call{Pen}{}$), $\Call{Scale}{s}$ (abbr. $\Call{Sc}{}$), and $R_{\aup}(s,a)$ are understood to be calculated with respect to the $Q_{R_i}$ being learned online; $\Call{Pen*}{}$, $\Call{Sc*}{}$, $R_{\aup}^*$, and $Q^*_{R_i}$ are taken to be their limit counterparts.

\begin{restatable}[The {\aup} penalty term converges]{lem}{pen}
$\forall s,a:\Call{Penalty}{}$ converges with probability $1$.
\end{restatable}
\begin{proof}[Proof outline]Let $\epsilon >0$, and suppose for all $R_i\in\mathcal{R}$, $\max_{s,\, a}|Q^*_{R_i}(s,a)-Q_{R_i}(s,a)|<\frac{\epsilon}{2|\mathcal{R}|}$ (because $Q$-learning converges; see~\cite{watkins1992q}).

\begin{align}&\max_{s,\,a}\left|\Call{Penalty*}{s,a}-\Call{Penalty}{s,a}\right|\\
\begin{split}
\leq&\max_{s,\,a}\sum_{i=1}^{|\mathcal{R}|}\left|Q^*_{R_i}(s,a)-Q_{R_i}(s,a)\right|+\\
&\hphantom{\max_{s,\,a}\sum_{i=1}^{|\mathcal{R}|}}\,\,\left|Q^*_{R_i}(s,\varnothing)-Q_{R_i}(s,\varnothing)\right|
\end{split}\\
<&\;\epsilon.
\end{align}
\end{proof}

The intuition for Lemma \ref{lem:r-aup} is that since $\Call{Penalty}{}$ and $\Call{Scale}{}$ both converge, so must $R_{\aup}$. For readability, we suppress the arguments to $\Call{Penalty}{}$ and $\Call{Scale}{}$.

\raup
\begin{proof}[Proof outline] If $\lambda =0$, the claim follows trivially.

Otherwise, let $\epsilon > 0$,  $B \vcentcolon = \max_{s,\,a}\,  \Call{Sc*}{} + \Call{Pen*}{}$, and $C \vcentcolon = \min_{s,\,a} \, \Call{Sc*}{}$. Choose any $\epsilon_R \in \left(0, \min \left[C, \dfrac{\epsilon \, C^2}{\lambda B + \epsilon \, C}\right] \right)$ and assume $\Call{Pen}{}$ and $\Call{Sc}{}$ are both $\epsilon_R$-close.

\begin{align}
&\max_{s,\,a} |R_{\aup}^*(s, a) - R_{\aup}(s, a)| \\
=&\max_{s,\,a} \lambda \left| \frac{\Call{Pen}{}}{\Call{Sc}{}} - \frac{\Call{Pen*}{}}{\Call{Sc*}{}} \right|\\
=&\max_{s,\,a} \lambda \, \frac{\left|\Call{Pen}{}\cdot\Call{Sc*}{} - \Call{Sc}{}\cdot\Call{Pen*}{}\right|}{ \Call{Sc*}{}\cdot\Call{Sc}{} }  \\
<&\max_{s,\,a} \lambda \, \frac{\left|(\Call{Pen*}{} + \epsilon_R)\Call{Sc*}{} - (\Call{Sc*}{}-\epsilon_R)\Call{Pen*}{}\right|}{C\,(\Call{Sc*}{} - \epsilon_R)}   \\
\leq&\;\frac{\lambda B}{C} \cdot \frac{\epsilon_R}{C - \epsilon_R} \\
<&\;\frac{\lambda B}{C} \cdot \frac{\epsilon \, C^2}{(\lambda B+\epsilon \, C)(C-\frac{\epsilon \, C^2}{\lambda B+\epsilon \, C})} \\
<&\;\frac{\lambda B}{C} \cdot \frac{\epsilon \, C^2}{\lambda B(C-\frac{\epsilon \, C^2}{\lambda B+\epsilon \, C})} \\
<&\;\frac{\epsilon}{1-\frac{\epsilon \, C}{\lambda B+\epsilon \, C}} \\
=&\;\epsilon \left( 1 + \frac{\epsilon \, C}{\lambda B}\right).\label{eq:pf_AUP}
\end{align}

\noindent But $B, C, \lambda $ are constants, and $\epsilon$ was arbitrary; clearly $\epsilon' > 0$ can be substituted such that $(\ref{eq:pf_AUP}) < \epsilon$.\end{proof}
\qconv*
\begin{proof}[Proof outline] Let $\epsilon > 0$, and suppose $R_{\aup}$ is $\frac{\epsilon (1 - \gamma)}{2} $-close.
Then $Q$-learning on $R_{\aup}$ eventually converges to a limit $\tilde{Q}_{R_{\aup}}$ such that $\max_{s,\,a} |Q^*_{R_{\aup}}(s, a) - \tilde{Q}_{R_{\aup}}(s, a)| < \frac{\epsilon}{2}$.
By the convergence of $Q$-learning, we also eventually have $\max_{s,\,a} |\tilde{Q}_{R_{\aup}}(s, a) - Q_{R_{\aup}}(s, a)| < \frac{\epsilon}{2}$. Then

\begin{align}
& \max_{s,\,a} \left|Q^*_{R_{\aup}}(s, a) - Q_{R_{\aup}}(s, a)\right| < \epsilon.
\end{align}
\end{proof}
\begin{prop}[{\aup} penalty equivariance properties]
    Let $c \in \mathbb{R}_{>0}, b \in \mathbb{R}$.
    \begin{enumerate}
    \item[a)] Let $\mathcal{R}'$ denote the set of functions induced by the positive affine transformation $cX + b$ on $\mathcal{R}$, and take $\Call{Pen*}{}_{\mathcal{R}'}$ to be calculated with respect to attainable set $\mathcal{R}'$. Then $\Call{Pen*}{}_{\mathcal{R}'} = c \, \Call{Pen*}{}_{\mathcal{R}}$.
    In particular, when $\Call{Sc*}{}$ is a $\Call{Penalty}{}$ calculation, $R^*_{{\aup}}$ is invariant to positive affine transformations of $\mathcal{R}$.   \\
    \item[b)] Let $R':= c R + b$, and take $R'^*_{{\aup}}$ to incorporate $R'$ instead of $R$.
    Then by multiplying $\lambda $ by $c$, the induced optimal policy remains invariant.
    \end{enumerate}
\end{prop}

\begin{proof}[Proof outline]
    For a), since the optimal policy is invariant to positive affine transformation of the reward function, for each $R'_i \in \mathcal{R}'$ we have $Q^*_{R'_i} = c \, Q^*_{R_i} + \frac{b}{1-\gamma}$. Substituting into Equation \ref{penalty} ($\Call{Penalty}{}$), the claim follows.\\

    \noindent For b), we again use the above invariance of optimal policies:
    \begin{align}
        R'^*_{{\aup}} &\defeq c R + b - c\lambda \,\frac{\Call{Pen*}{}}{\Call{Sc*}{}} \\
        &= c R^*_{{\aup}} + b.
    \end{align}
\end{proof}

\chapter{Avoiding Side Effects in Complex Environments}
\ifbool{Dissertate}{\addtocontents{lot}{\protect\contentsline{chapter}{\protect\numberline{\thechapter}Avoiding Side Effects in Complex Environments}{}{chapter.\theHchapter}}}{}
\section{Theoretical results}
\label{sec:theory}

Consider a rewardless \textsc{mdp} $\langle \mathcal{S},\mathcal{A},T,\gamma\rangle$. Reward functions $R\in\mathbb{R}^\St$ have corresponding optimal value functions $\Vf{s}$.

\begin{prop}[Communicability bounds maximum change in optimal value]\label{mut-bound}
If $s$ can reach $s'$ with probability $1$ in $k_1$ steps and $s'$ can reach $s$ with probability $1$ in $k_2$ steps, then $\sup_{R\in [0, 1]^{\St}} \left|V^*_R(s)-V^*_R(s')\right|\leq \geom[1-\gamma^{\max(k_1,k_2)}]< \geom$.
\end{prop}
\begin{proof}
We first bound the maximum increase.
\begin{align}
    \sup_{R\in [0, 1]^{\St}} V^*_R(s')-V^*_R(s) &\leq \sup_{R\in [0, 1]^{\St}} V^*_R(s') - \left(0\cdot \geom[1-\gamma^{k_1}]+\gamma^{k_1}V^*_R(s')\right)\label{eq:max-inc}\\
    &\leq \geom[1] - \left(0\cdot \geom[1-\gamma^{k_1}]+\gamma^{k_1}\geom[1]\right)\\
    &=\geom[1-\gamma^{k_1}].
\end{align}
\Cref{eq:max-inc} holds because even if we make $R$ equal  $0$ for as many states as possible, $s'$ is still reachable from $s$. The case for maximum decrease is similar.
\end{proof}

\section{Training details}\label{sec:train}
In \cref{sec:line}, we aggregated performance from 3 curricula with 5 seeds each, and 1 curriculum with 3 seeds.

We detail how we trained the {\aupTT} and {\aupTT}$_\text{proj}$ conditions. An algorithm describing the training process can be seen in algorithm \ref{algo:aup}.

\subsection{Auxiliary reward training}

For the first phase of training, our goal is to learn $Q_{\text{aux}}$, allowing us to compute the {\aup} penalty in the second phase of training. Due to the size of the full SafeLife state $(350 \times 350 \times 3)$, both conditions downsample the observations with average pooling and convert to intensity values.

Previously, \citet{turner2020conservative} learned $Q_{\text{aux}}$ with tabular $Q$-learning. They used environments small enough such that reward could be assigned to each state. Because SafeLife environments are too large for tabular $Q$-learning, we demonstrated two methods for randomly generating an auxiliary reward function.

\begin{enumerate}
\item[{\aupTT}] We acquire a low-dimensional state representation by training a continuous Bernoulli variational autoencoder \citep{loaiza2019continuous}. To train the \textsc{cb-vae}, we collect a buffer of observations by acting randomly for $\frac{100,000}{N_{\text{env}}}$ steps in each of the $N_{\text{env}}$ environments. This gives us 100K total observations with an $N_{\text{env}}$-environment curriculum.  We train the \textsc{cb-vae} for 100 epochs, preserving the encoder $E$ for downstream auxiliary reward training.

For each auxiliary reward function, we draw a linear functional uniformly from $(0, 1)^Z$ to serve as our auxiliary reward function, where $Z$ is the dimension of the \textsc{cb-vae}'s latent space. The auxiliary reward for an observation is the composition of the linear functional with an observation's latent representation.

\item[{\aupp}] Instead of using a \textsc{cb-vae}, {\aupTT}$_\text{proj}$ simply downsamples the input observation. At the beginning of training, we generate a linear functional over the unit hypercube (with respect to the downsampled observation space). The auxiliary reward for an observation is the composition of the linear functional with the downsampled observation.
\end{enumerate}

The auxiliary reward function is learned after it is generated. To learn $Q_{\text{aux}}$, we modify the value function in \textsc{ppo} to a Q-function. Our training algorithm for phase 1 only differs from \textsc{ppo} in how we calculate reward. We train each auxiliary reward function for 1\textsc{m} steps.

\subsection{\textsc{Aup} reward training}
In phase 2, we train a new \textsc{ppo} agent on $R_{\aup}$ (\cref{COMP-eq:aup}) for the corresponding SafeLife task. Each step, the agent selects an action $a$ in state $s$ according to its policy $\pi_{{\aupTT}}$, and receives reward $R_{\aup}(s,a)$ from the environment. We compute $R_{\aup}(s,a)$ with respect to the learned Q-values $Q_{\text{aux}}(s, \varnothing)$ and $Q_{\text{aux}}(s, a)$. The algorithm is shown in \cref{algo:aup}.

The penalty term is modulated by the hyperparameter $\lambda$, which is linearly scaled from $10^{-3}$ to some final value $\lambda^*$ (default $10^{-1}$). Because $\lambda$ controls the relative influence of the penalty, linearly increasing $\lambda$ over time will prioritize primary task learning in early training and slowly encourage the agent to obtain the same reward while avoiding side effects. If $\lambda$ is too large—if side effects are too costly—the agent won't have time to adapt its current policy and will choose inaction ($\varnothing$) to escape the penalty. A careful $\lambda$ schedule helps induce a successful policy that avoids side effects.

\begin{algorithm}[h]
    \caption{Safelife \textsc{aup} Training Algorithm}
    \label{algo:aup}
    \begin{algorithmic}[1]
    \Require Exploration buffer $S$
    \Require CB-VAE $\mathcal{F}$ with encoder $E$, decoder $D$
    \Require Exploration buffer $S$
    \Require Auxiliary reward functions $\phi$
    \Require Auxiliary policy $\psi_{\text{aux}}$, {\aup} policy $\pi_{\aup}$
    \Require CB-VAE training epochs $T$
    \Require {\aup} penalty coefficient $\lambda$
    \Require Exploration buffer size $k$
    \Require Auxiliary model training steps $L$
    \Require {\aup} model training steps $N$
    \Require PPO update function PPO-Update
    \Require CB-VAE update function VAE-Update

    \For{Step $k = 1, \dots K$}
        \State Sample random action $a$
        \State $s \leftarrow \text{Act}(a)$
        \State $S = s \cup S$
    \EndFor
    \For{Epoch $t = 1, \dots T$}
        \State \normalfont{Update-VAE $(\mathcal{F}, S)$}
    \EndFor
    \For{Step $i = 1, \dots L+N$}
        $s \leftarrow$ Starting state
        \For{Step $l = 1, \dots L$}
            \State $a = \psi_{\text{aux}}(s)$
            \State $s' = \text{ Act}(a)$
            \State $r = \phi \cdot E(s)$
            \State PPO-Update $(\psi_{\text{aux}}, s, a, r, s')$
            \State $s = s'$
        \EndFor
        \State $s \leftarrow$ Starting state
        \For{Step $n = 1, \dots N$}
            \State $a = \pi_{\aup}(s)$
            \State $s', r = \text{ Act}(a)$
            \State $r = r + R_{\aup}(\psi_{\text{aux}}, \, \pi_{\aup}, \,s, a, \lambda)$ \quad (\Cref{COMP-eq:aup})
            \State PPO-Update $(\pi_{\aup}, s, a, r, s')$
            \State $s = s'$
        \EndFor
    \EndFor
    \end{algorithmic}
\end{algorithm}

\section{Hyperparameter selection}
Table \ref{tab:params} lists the hyperparameters used for all conditions, which generally match the default SafeLife settings. \emph{Common} refers to those hyperparameters that are the same for each evaluated condition. \emph{AUX} refers to hyperparameters that are used only when training on $R_\text{AUX}$, thus, it only pertains to {\aupTT} and {\aupTT}$_\text{proj}$. The conditions {\ppo} and {\naive} use the \emph{PPO} hyperparameters for the duration of their training, while {\aupTT}, {\aupTT}$_\text{proj}$ use them when training with respect to $R_{\aup}$. \emph{DQN} refers to the hyperparameters used to train the model for {\dqn}.

\begin{longtable}[ht]{rrl}
        \toprule
        \multicolumn{2}{r}{Hyperparameter} &  Value\\
        \midrule
        \multicolumn{2}{r}{\it{Common}}& \\
        & Learning Rate & $3 \cdot 10^{-4}$\\
        & Optimizer & Adam \\
        & Gamma ($\gamma$) &  $0.97$\\
        & Lambda (\textsc{ppo}) & $0.95$\\
        & Lambda ({\aup}) & $10^{-3} \rightarrow 10^{-1}$\\
        & Entropy Clip & $1.0$\\
        & Value Coefficient & $0.5$\\
        & Gradient Norm Clip & $5.0$\\
        & Clip Epsilon & $0.2$ \\

        \midrule
        \multicolumn{2}{r}{\emph{AUX}}& \\
        & Entropy Coefficient & $0.01$\\
        & Training Steps & $1 \cdot 10^6$\\

        \midrule
        \multicolumn{2}{r}{\emph{AUP\textsubscript{proj}}}& \\
        & Lambda ({\aup}) & $10^{-3}$\\

        \midrule
        \multicolumn{2}{r}{\emph{PPO}}& \\
        & Entropy Coefficient & $0.1$\\

        \midrule
        \multicolumn{2}{r}{\emph{DQN}}& \\
        & Minibatch Size & $64$ \\
        & SGD Update Frequency & $16$\\
        & Target Network Update Frequency & $1 \cdot 10^3$\\
        & Replay Buffer Capacity & $1 \cdot 10^4$\\
        & Exploration Steps & $4 \cdot 10^3$\\

        \midrule
        \multicolumn{2}{r}{\emph{Policy}}& \\
        & Number of Hidden Layers & $3$\\
        & Output Channels in Hidden Layers & $(32, 64, 64)$\\
        & Nonlinearity & \textsc{r}e\textsc{lu}\\
        \midrule
        \multicolumn{2}{r}{\emph{cb-vae}}& \\
        & Learning Rate & $10^{-4}$\\
        & Optimizer & Adam \\
        & Latent Space Dimension ($Z$) & $1$\\
        & Batch Size & $64$ \\
        & Training Epochs & $50$\\
        & Epsilon & $10^{-5}$\\

        & Number of Hidden Layers (encoder) & $6$\\
        & Number of Hidden Layers (decoder) & $5$\\
        & Hidden Layer Width (encoder) & $(512, 512, 256, 128, 128, 128)$\\
        & Hidden Layer Width (decoder) & $(128, 256, 512, 512, \text{output})$\\
        & Nonlinearity & \textsc{elu}\\
        \bottomrule
        \caption[Hyperparameters for the SafeLife experiments]{Hyperparameters for the SafeLife experiments.}\label{tab:params}
\end{longtable}

\section{Compute environment}

\begin{table}[!h]\centering
    \begin{tabular}{cc}
    \toprule
    Condition & \textsc{gpu}-hours per trial \\
    \midrule
    {\ppo}    & $ 6$                   \\
    {\dqn}    & $ 16$                  \\
    {\aupTT}  & $ 8$                   \\
    {\aupp}   & $ 7.5$                 \\
    {\naive}  & $ 6$                   \\
    \bottomrule
    \end{tabular}
    \caption[Compute time for the SafeLife conditions]{Compute time for each condition.}
    \label{tab:hours}
\end{table}

For data collection, we only ran the experiments once. All experiments were performed on a combination of \textsc{nvidia} \textsc{gtx} 2080\textsc{ti} \textsc{gpu}s, as well as \textsc{nvidia} \textsc{v}100 \textsc{gpu}s. No individual experiment required more than 3\textsc{gb} of \textsc{gpu} memory. We did not run a 3-seed {\dqn} curriculum for the experiments in \cref{sec:line}.

The auxiliary reward functions were trained on down-sampled rendered game screens, while all other learning used the internal SafeLife state representation. Incidentally, \cref{tab:hours} shows that \texttt{AUP}'s preprocessing turned out to be computationally expensive (compared to \texttt{PPO}'s).

\section{Additional data}\label{sec:addtl-data}

\begin{figure}[h]\centering
\caption[Safelife episode length statistics for different {\aupTT} conditions]{Smoothed episode length curves with shaded regions representing $\pm 1$ standard deviation.  {\aupTT} and {\aupp} begin training on the $R_{\aup}$ reward signal at steps 1.1\textsc{m} and 1\textsc{m},  respectively.}\label{fig:length}
\subfloat{
    \includegraphics[height=2.1in]{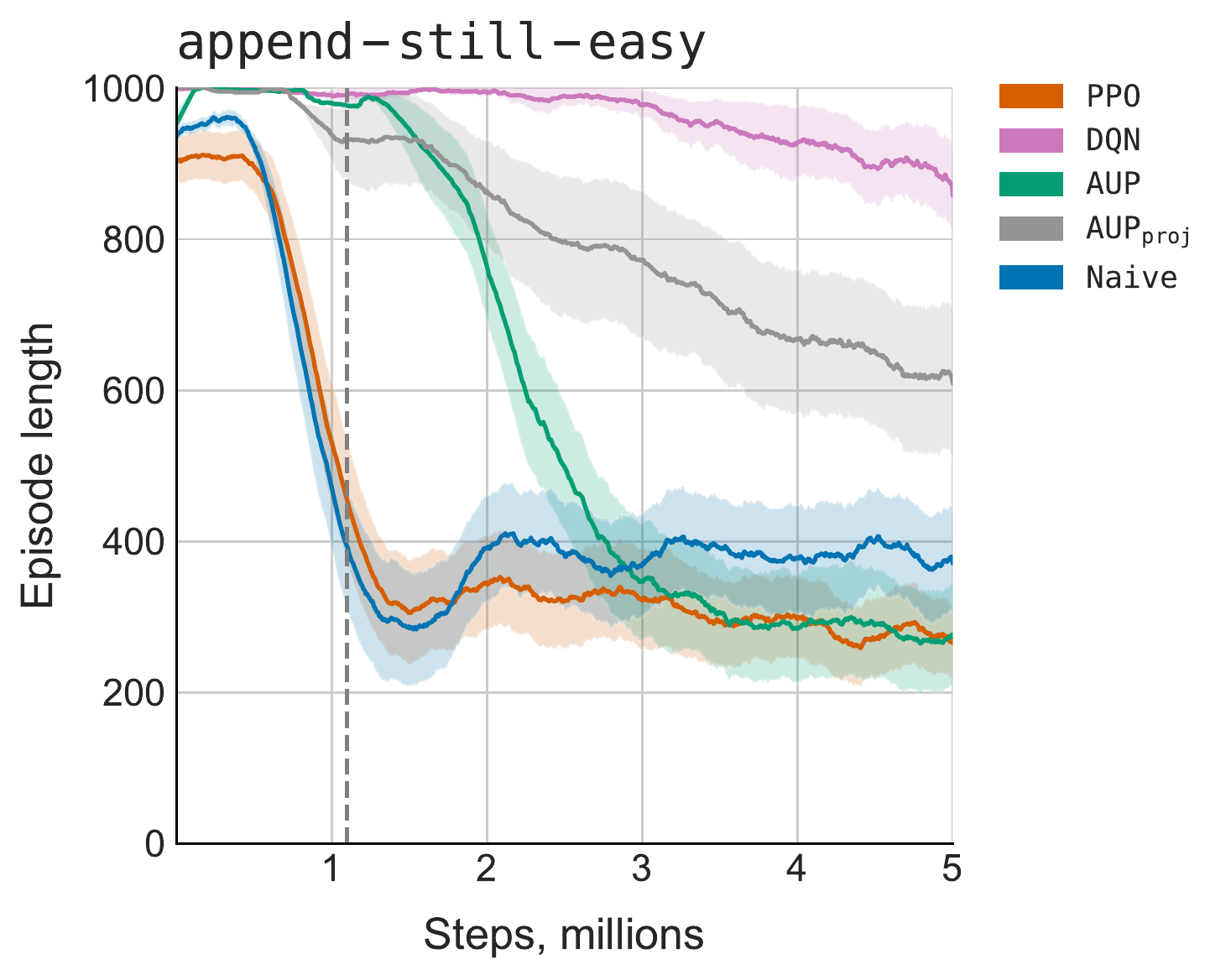}
}
\subfloat{
    \includegraphics[height=2.1in]{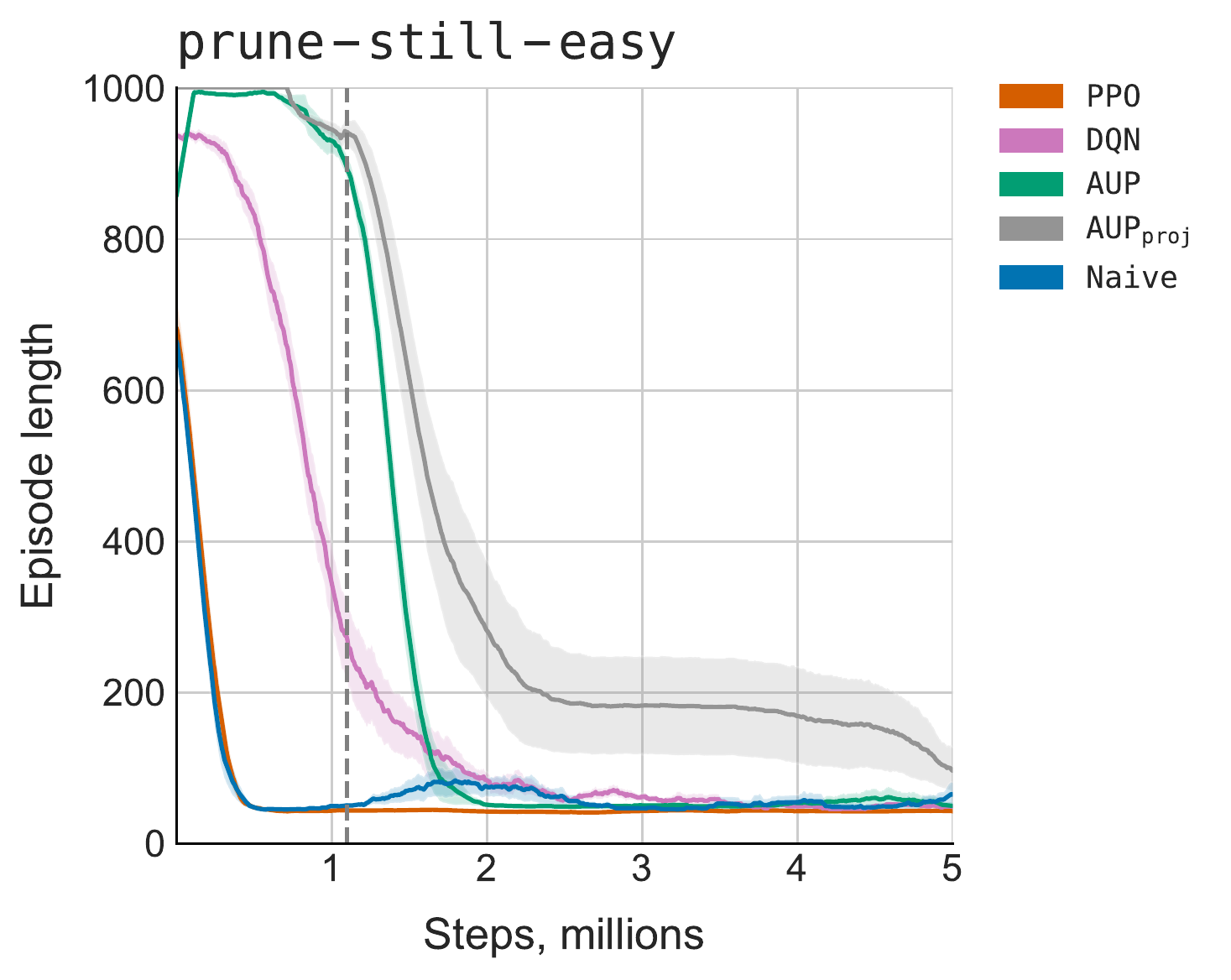}
}

\subfloat{
    \includegraphics[height=2.1in]{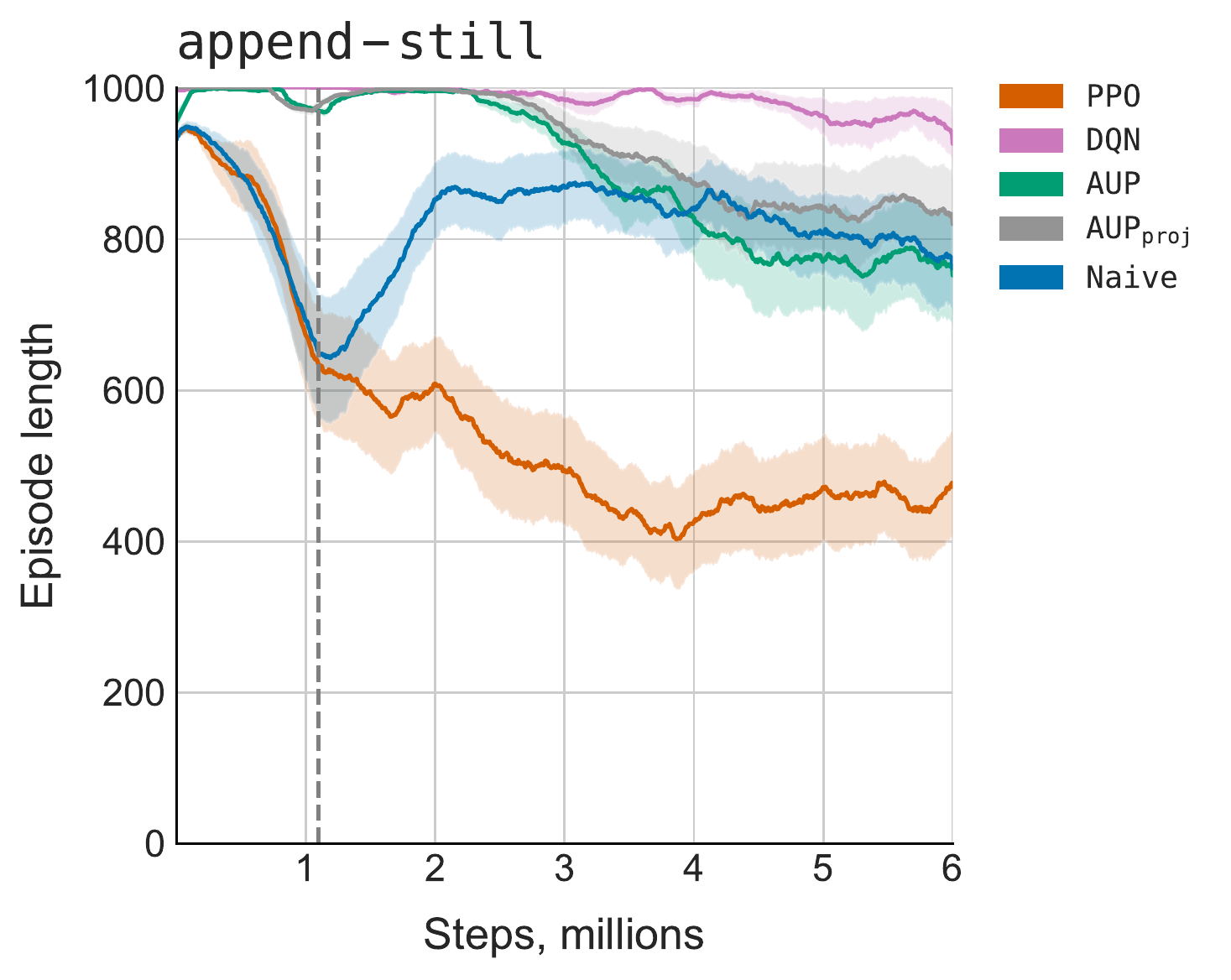}
}
\subfloat{
    \includegraphics[height=2.1in]{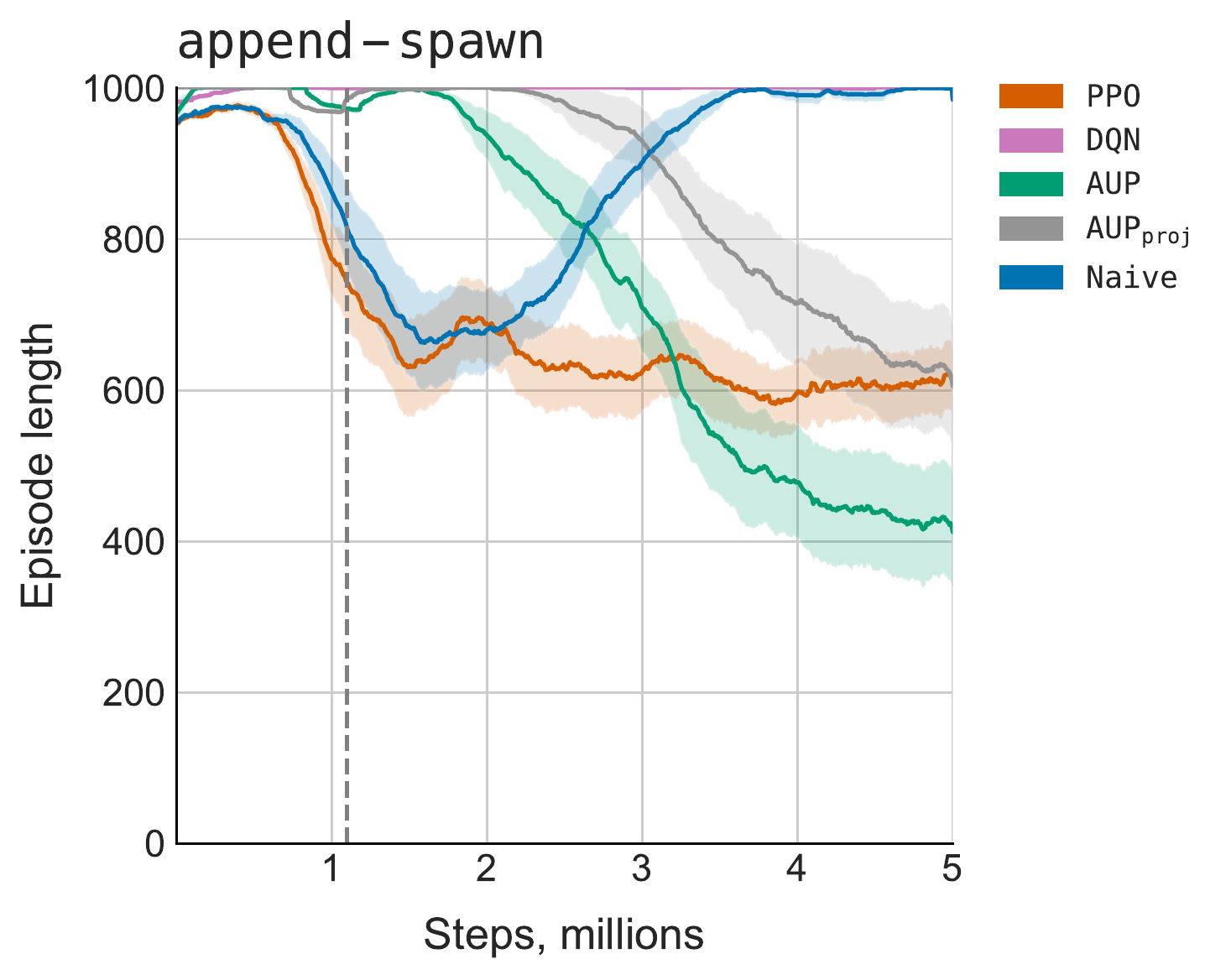}
}
\end{figure}

\begin{figure}[ht!]\centering
\caption[SafeLife reward curves for different latent space configurations]{Reward curves for auxiliary reward functions with a $Z$-dimensional latent space. Shaded regions represent $\pm 1$ standard deviation. Auxiliary reward is not comparable across trials, so learning is expressed by the slope of the curves.} \label{fig:aux-reward}
\includegraphics[height=2.5in]{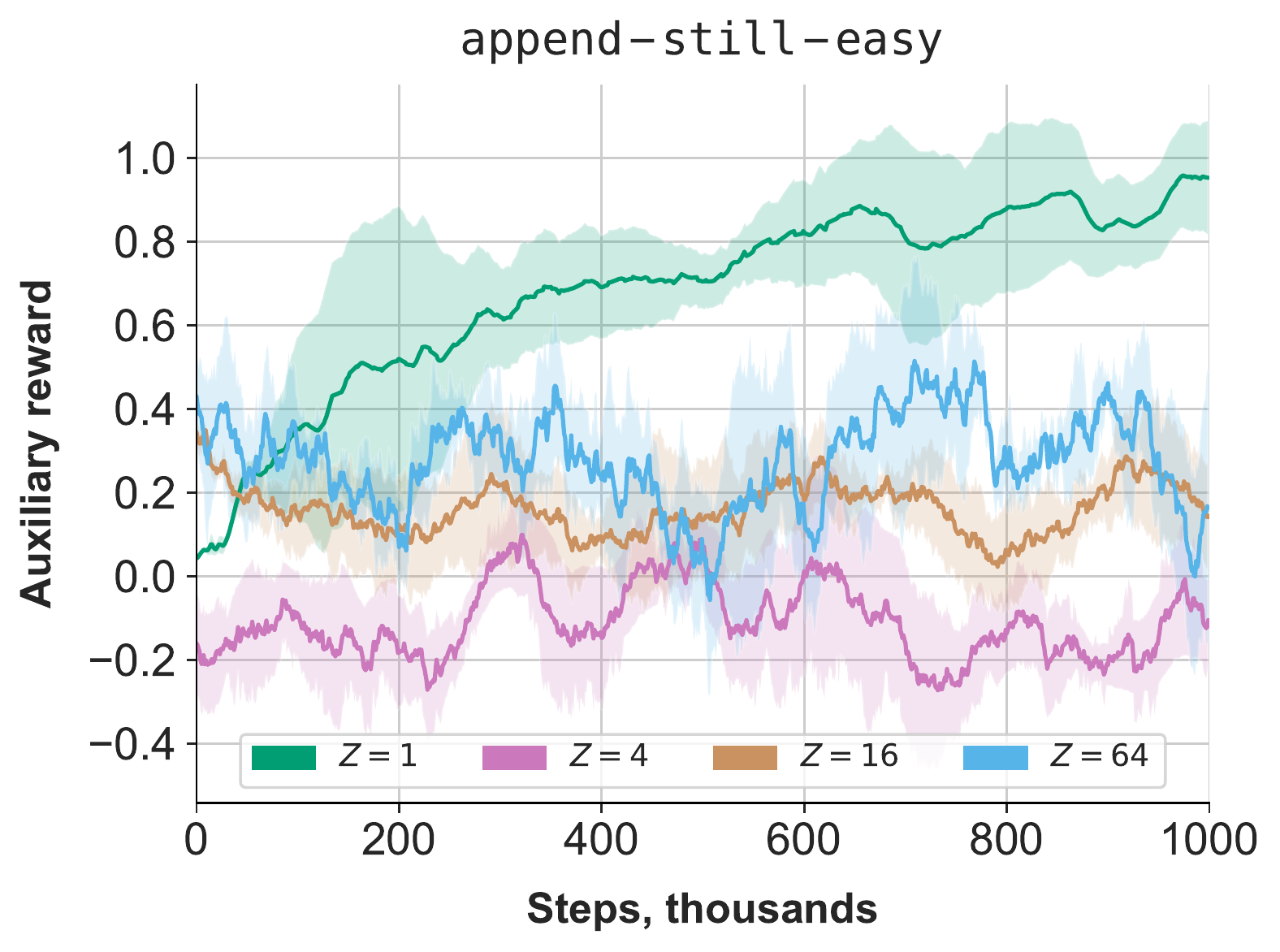}
\end{figure}

\begin{figure}[h]
\caption[Per-seed learning curves in SafeLife]{Smoothed learning curves for individual {\aupTT} seeds. {\aupTT} begins training on the $R_{\aup}$ reward signal at step 1.1\textsc{m}, marked by a dotted vertical line.}\label{fig:individual}
\centering
\subfloat{
    \includegraphics[height=1.8in]{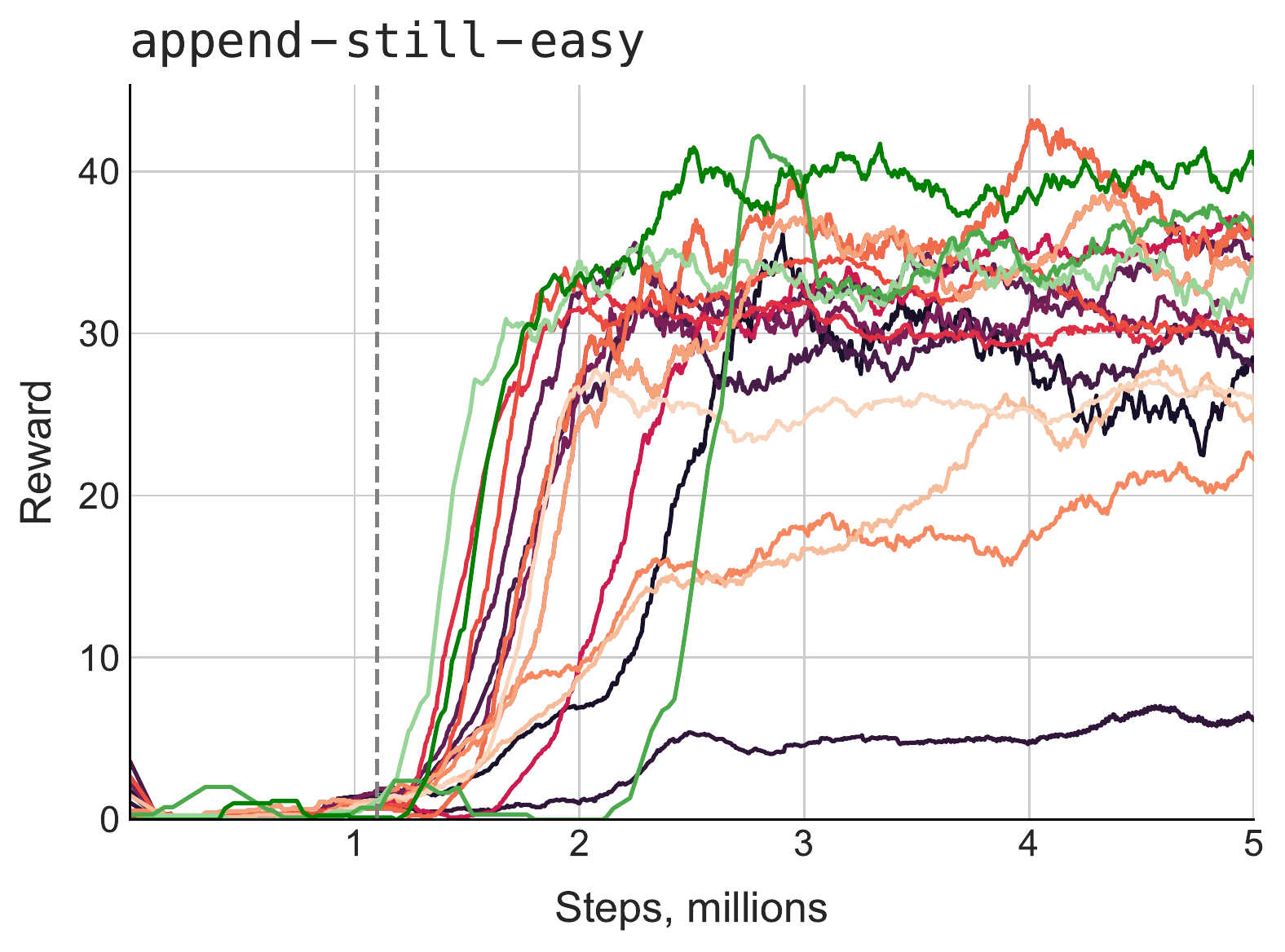}
}
\subfloat{
    \includegraphics[height=1.8in]{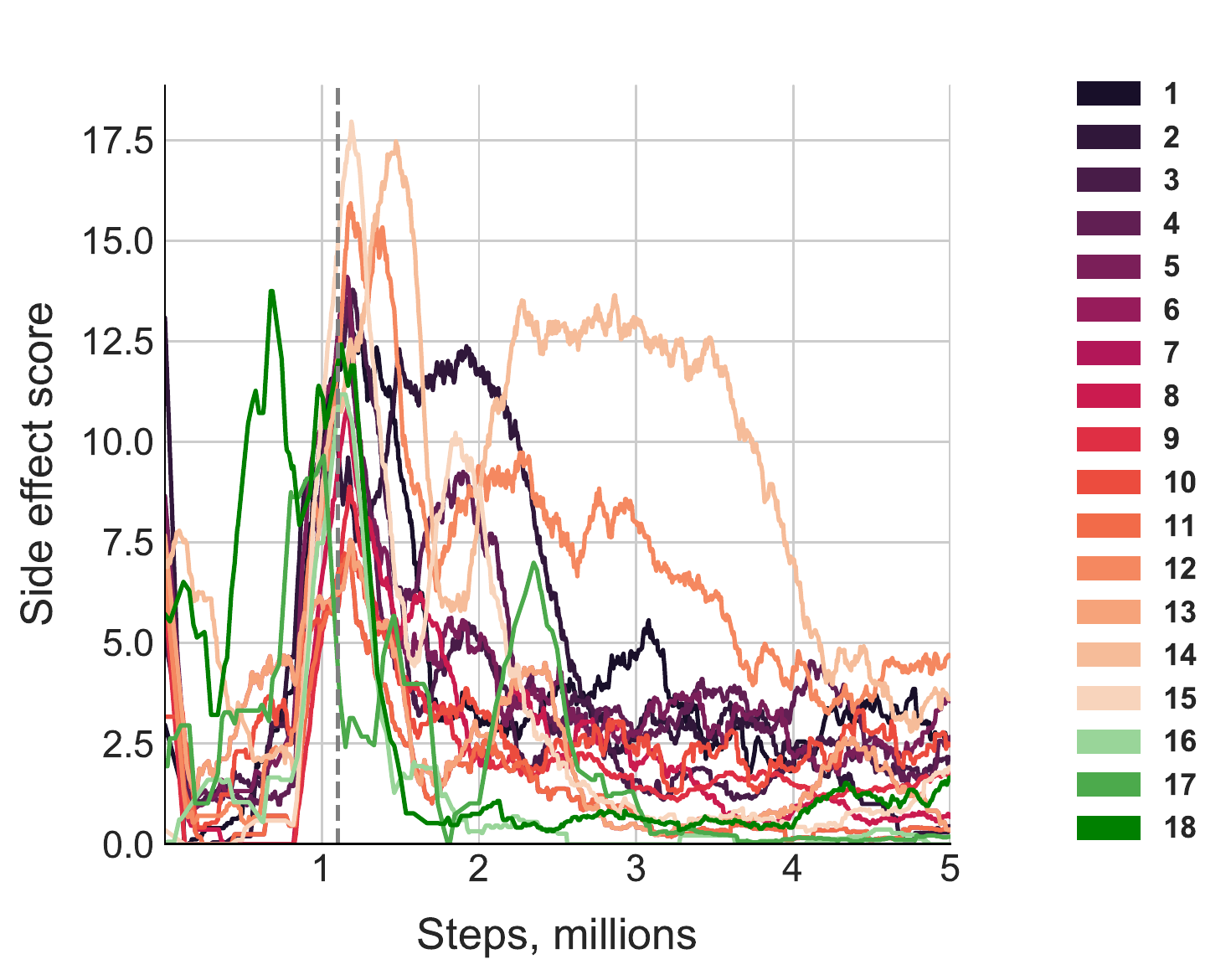}
}

\subfloat{
    \includegraphics[height=1.8in]{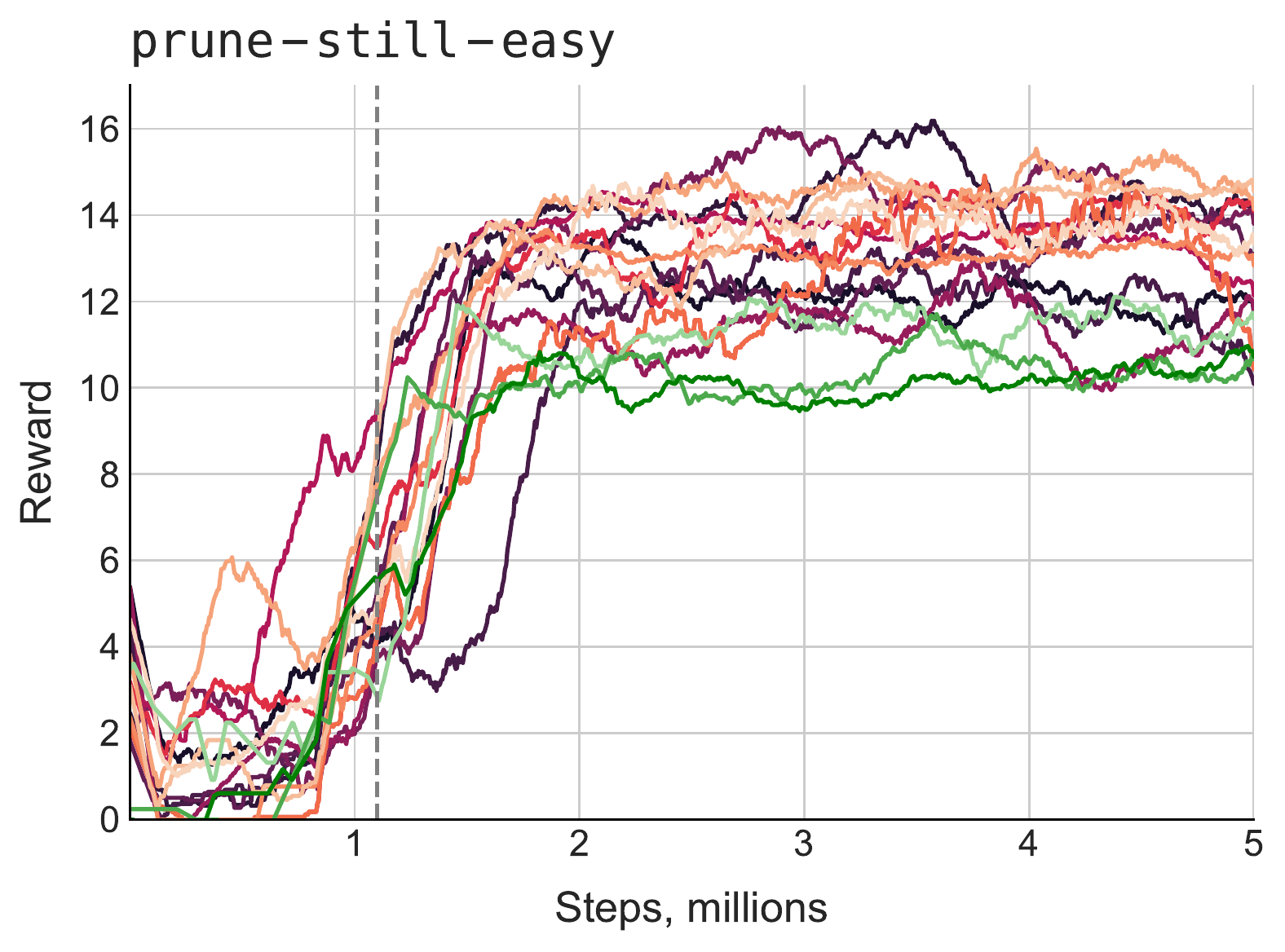}
}
\subfloat{
    \includegraphics[height=1.8in]{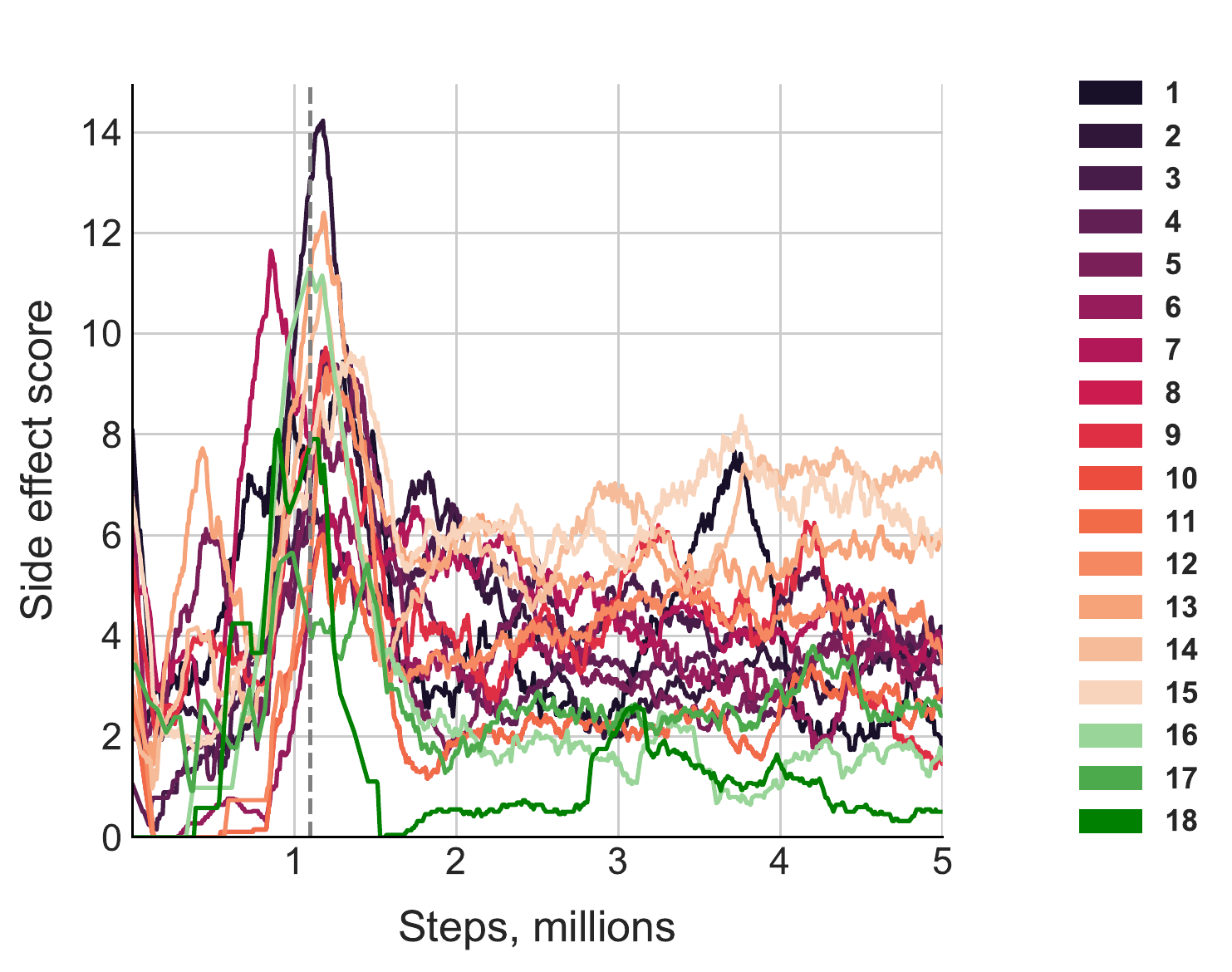}
}

\subfloat{
    \includegraphics[height=1.8in]{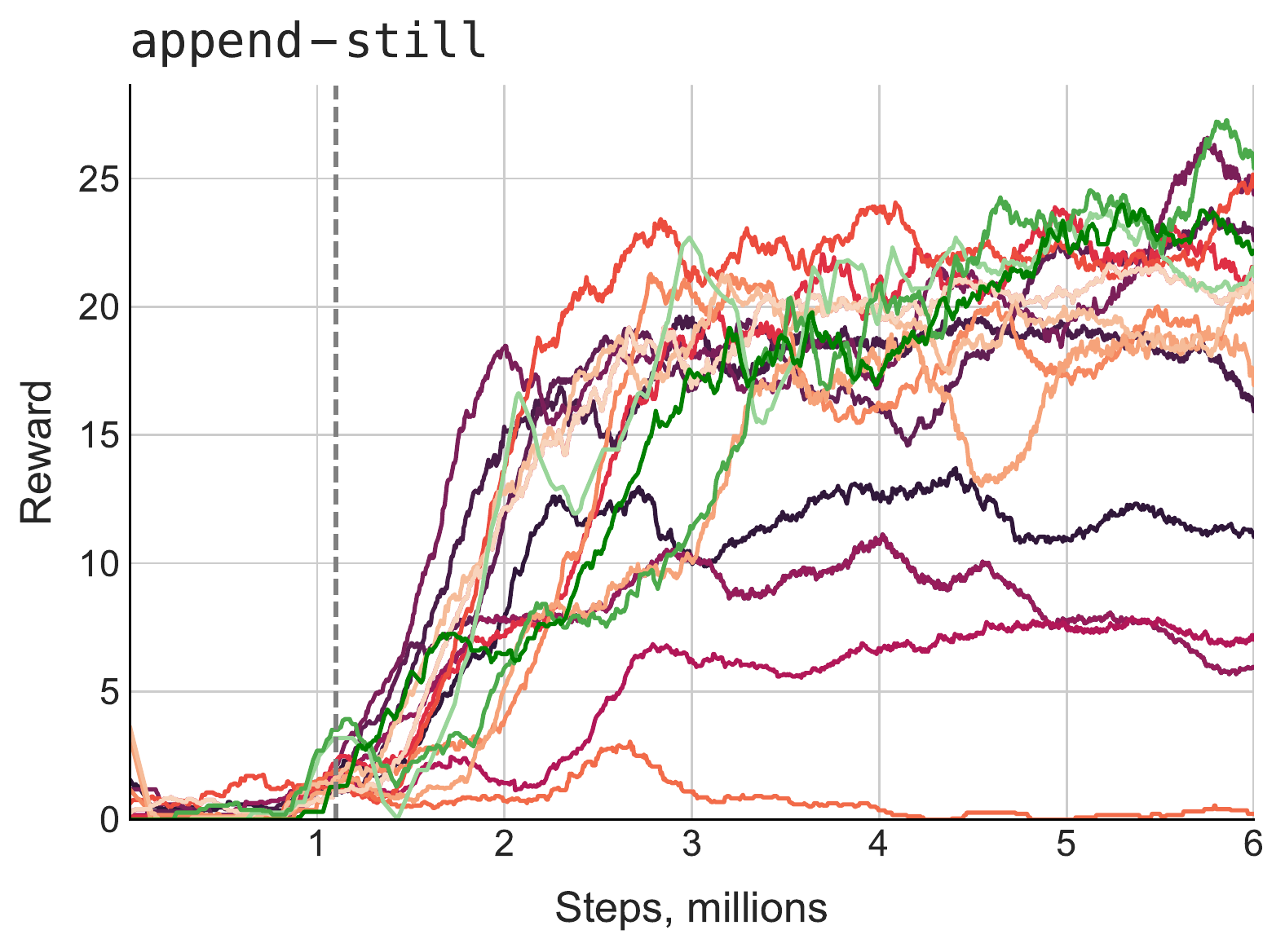}
}
\subfloat{
    \includegraphics[height=1.8in]{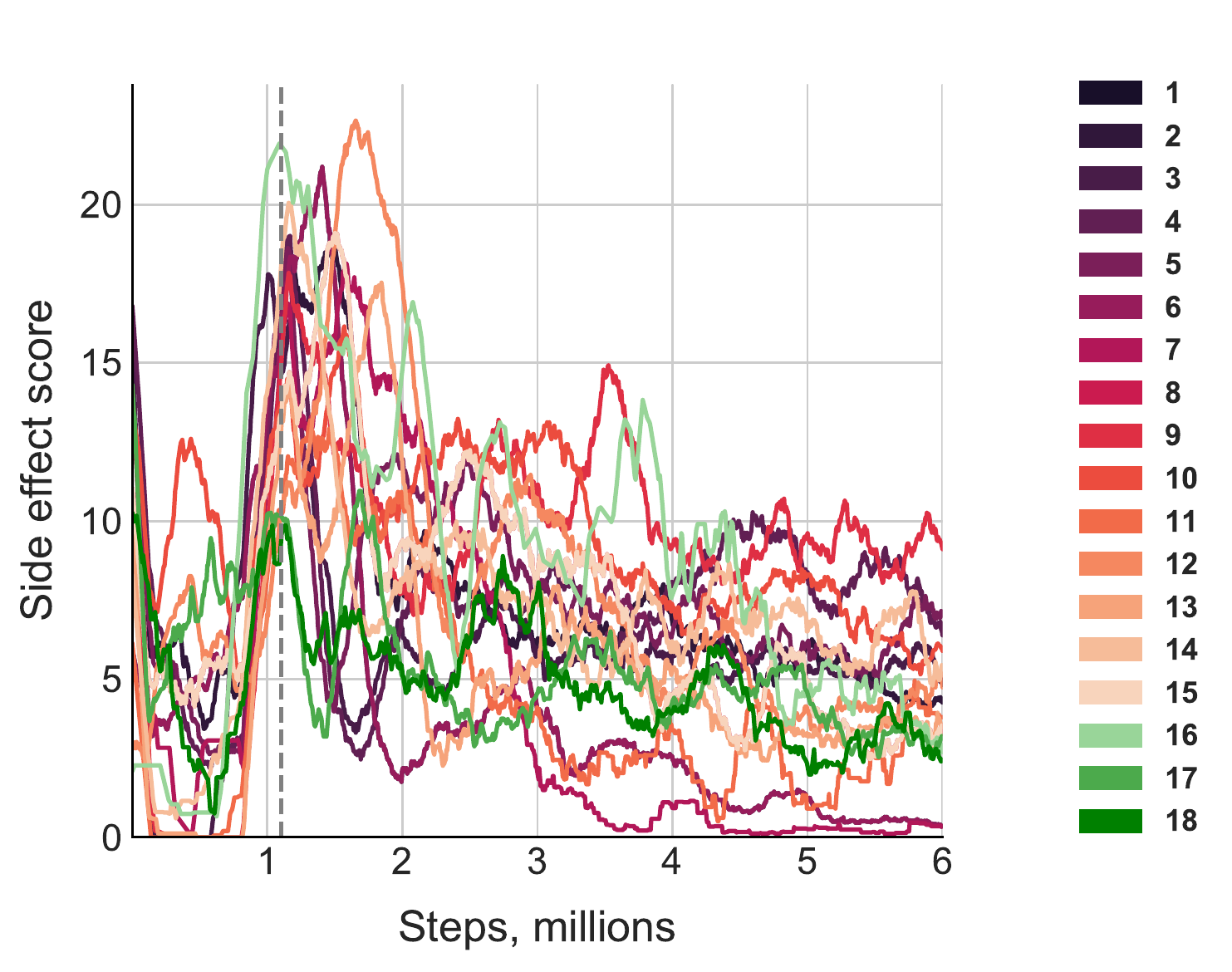}
}

\subfloat{
    \includegraphics[height=1.8in]{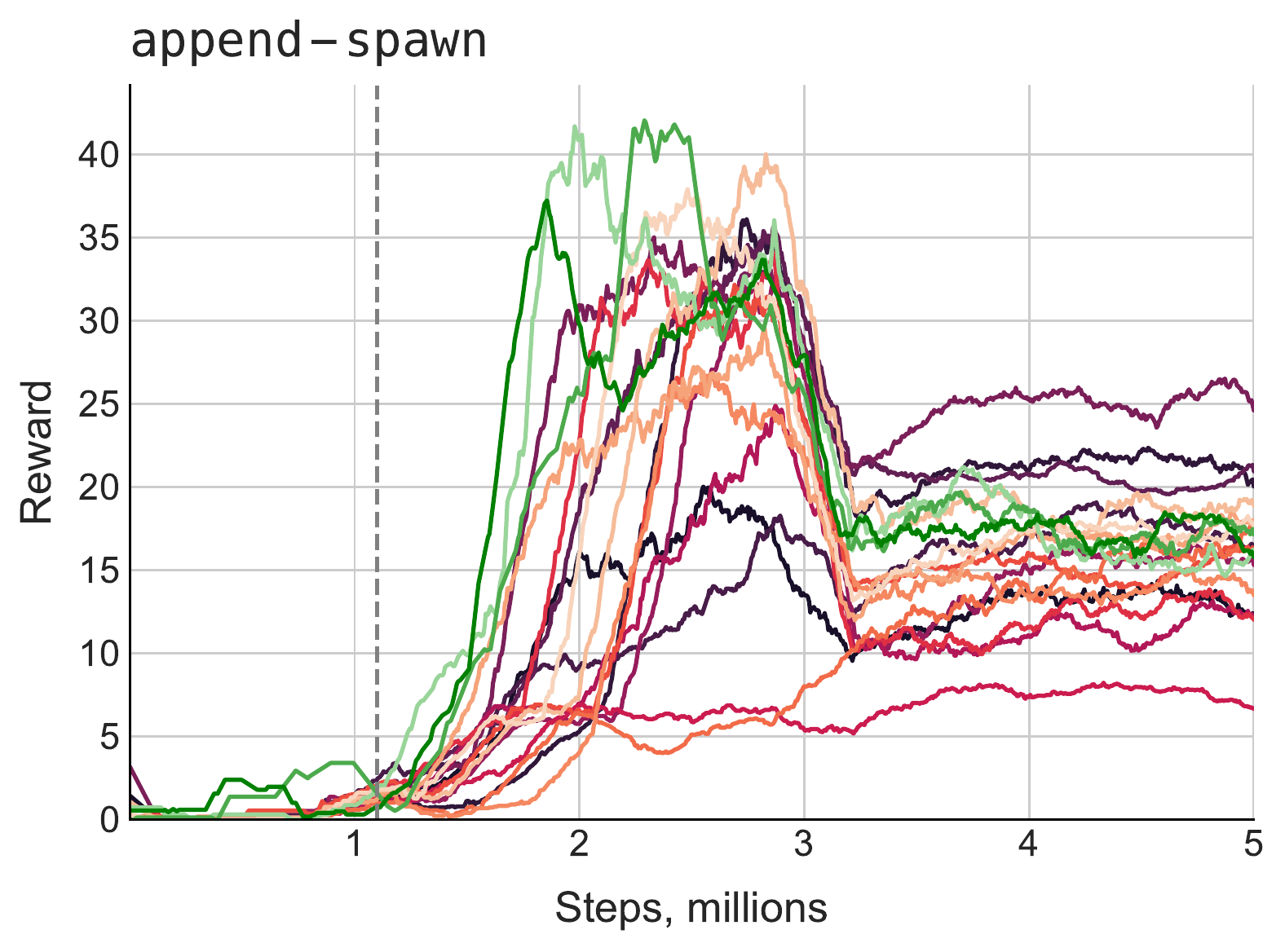}
}
\subfloat{
    \includegraphics[height=1.8in]{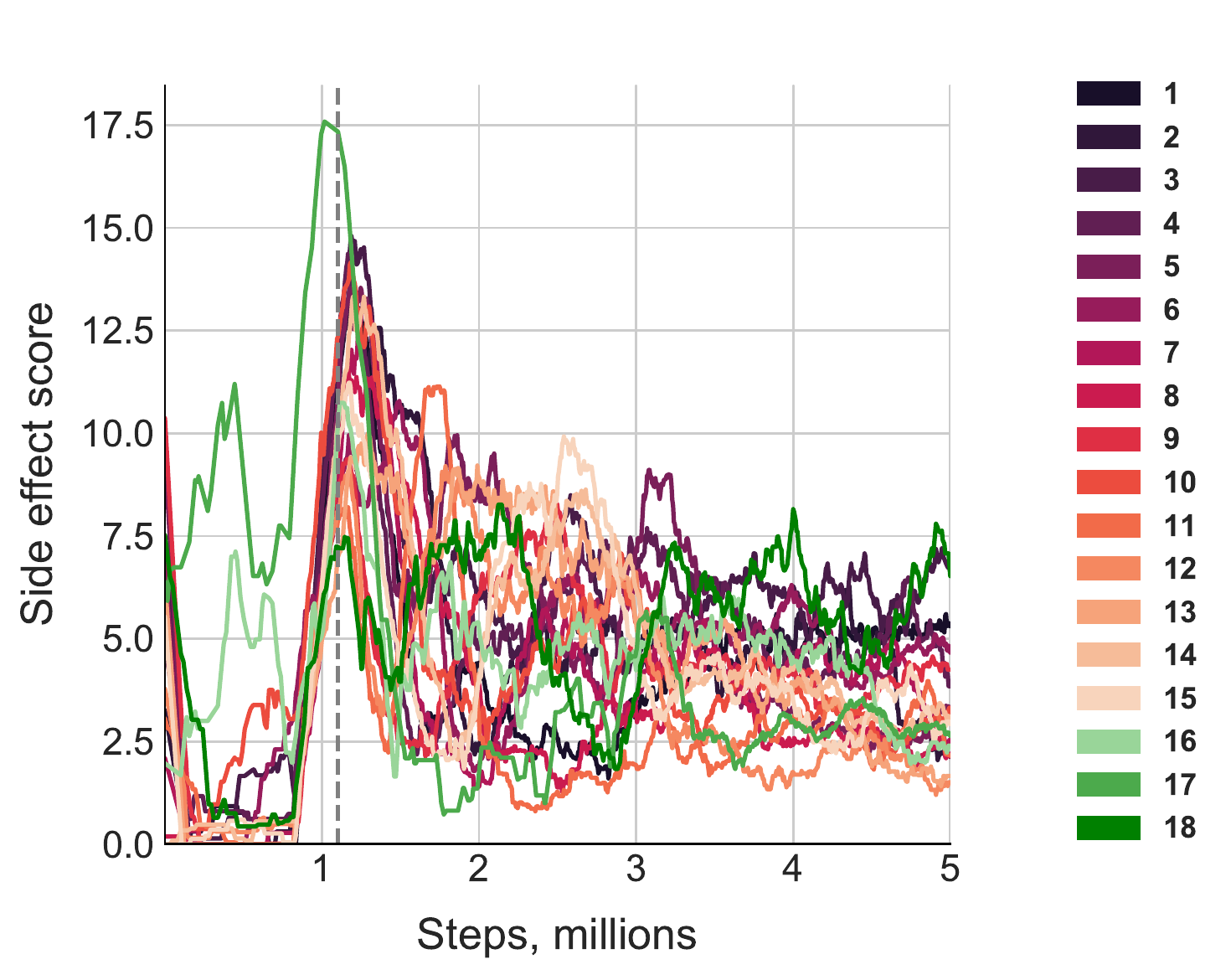}
}
\end{figure}

\begin{figure}[h]
\centering
\caption[Per-seed episode length curves in SafeLife]{Smoothed episode length curves for individual {\aupTT} seeds.}\label{fig:individ-length}
\subfloat{
    \includegraphics[height=2in]{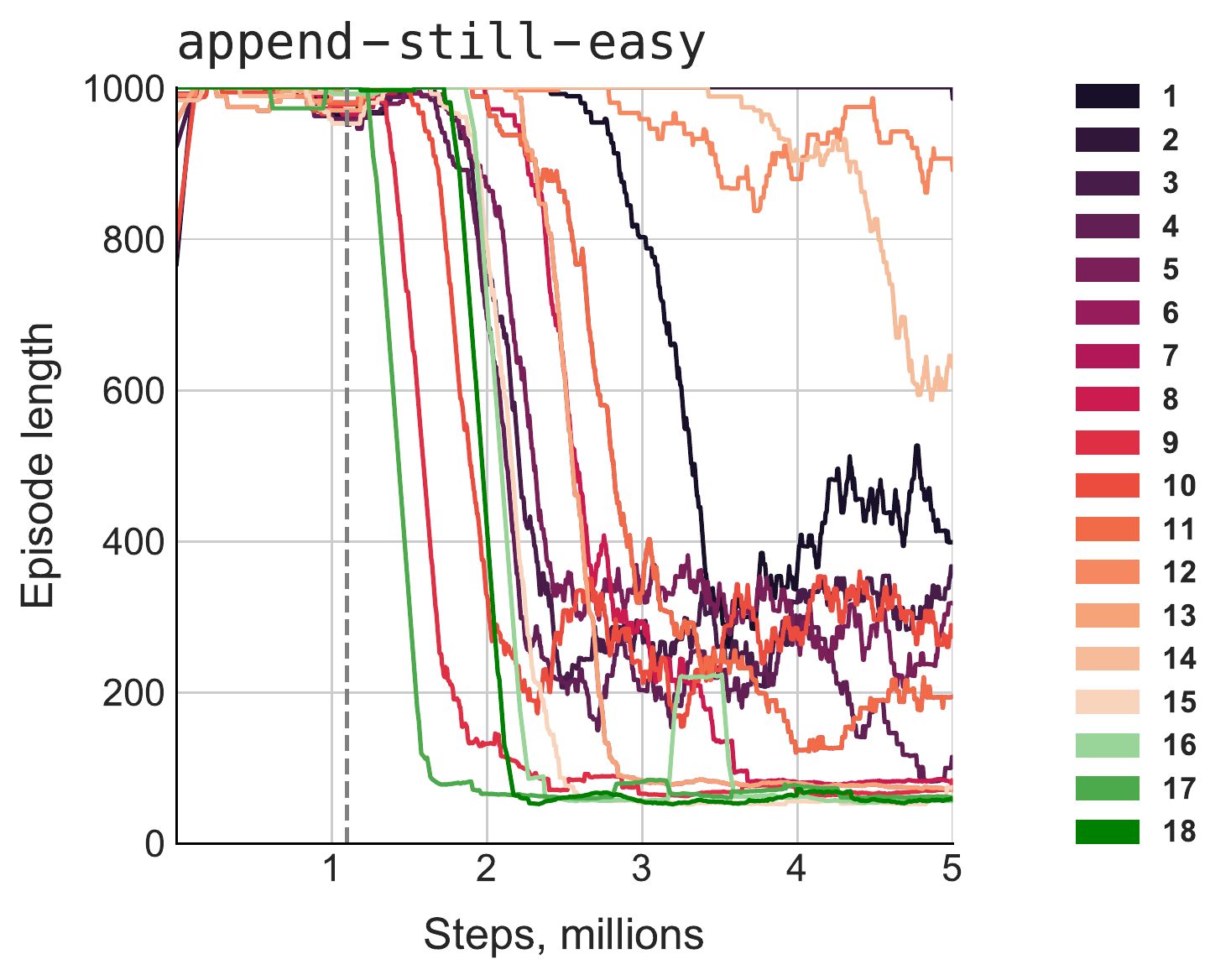}
}
\subfloat{
    \includegraphics[height=2in]{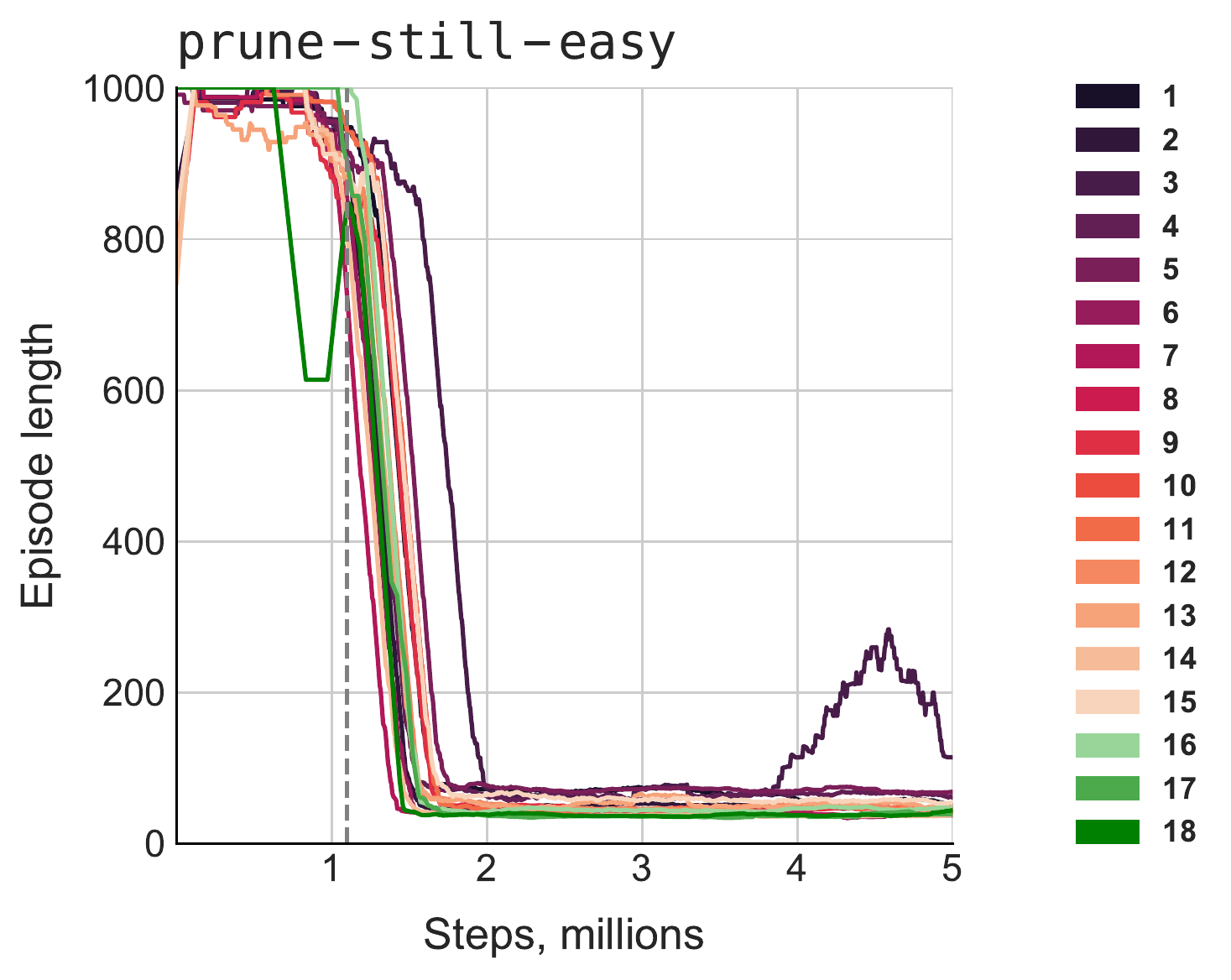}
}

\subfloat{
    \includegraphics[height=2in]{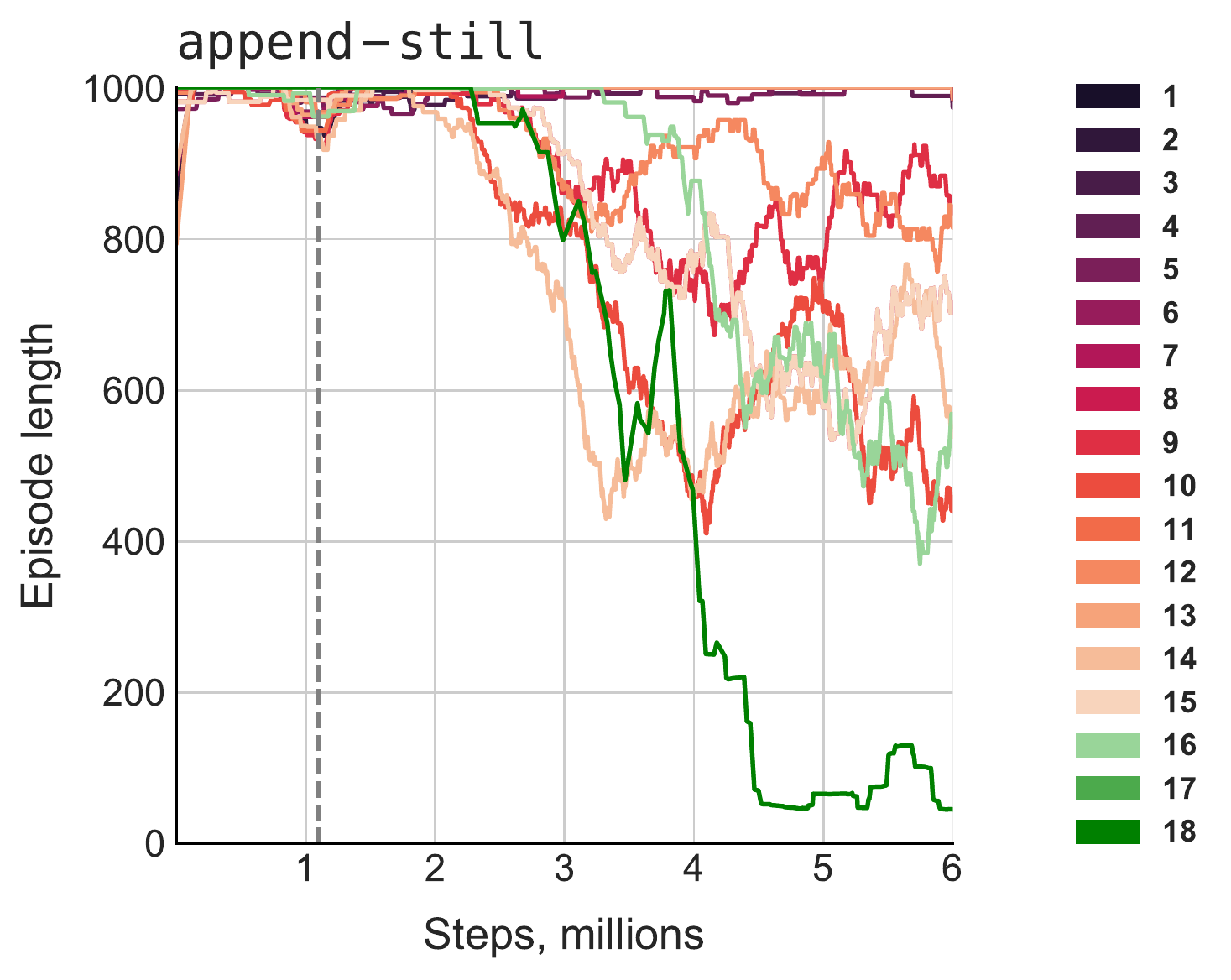}
}
\subfloat{
    \includegraphics[height=2in]{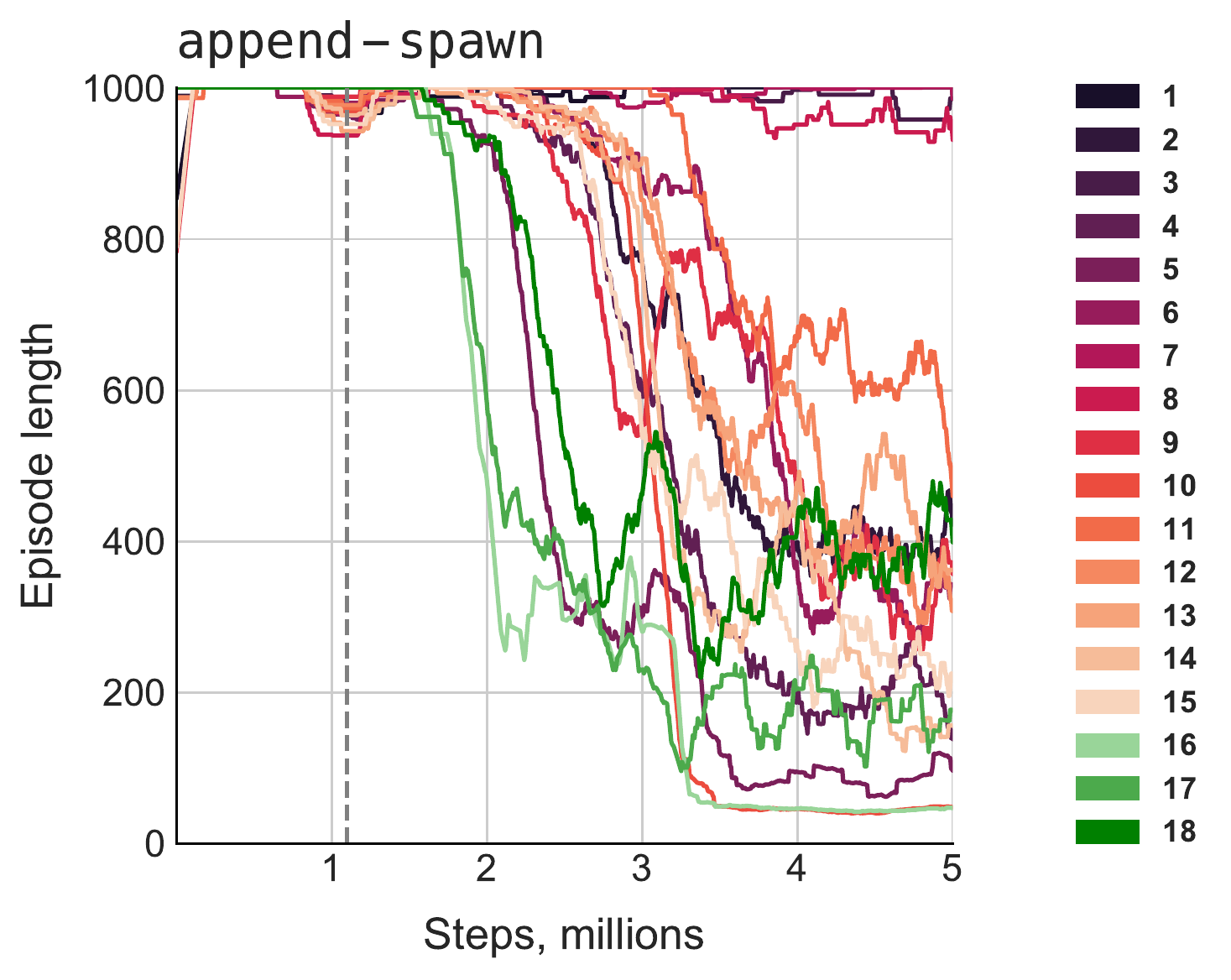}
}
\end{figure}

\begin{figure}[h]
\caption[Per-batch learning curves for {\aupTT} in SafeLife]{Smoothed learning curves for {\aupTT} on its four curricula. {\aupTT} begins training on the $R_{\aup}$ reward signal at step 1.1\textsc{m}, marked by a dotted vertical line.}\label{fig:batch}
\centering
\subfloat{
    \includegraphics[height=1.7in]{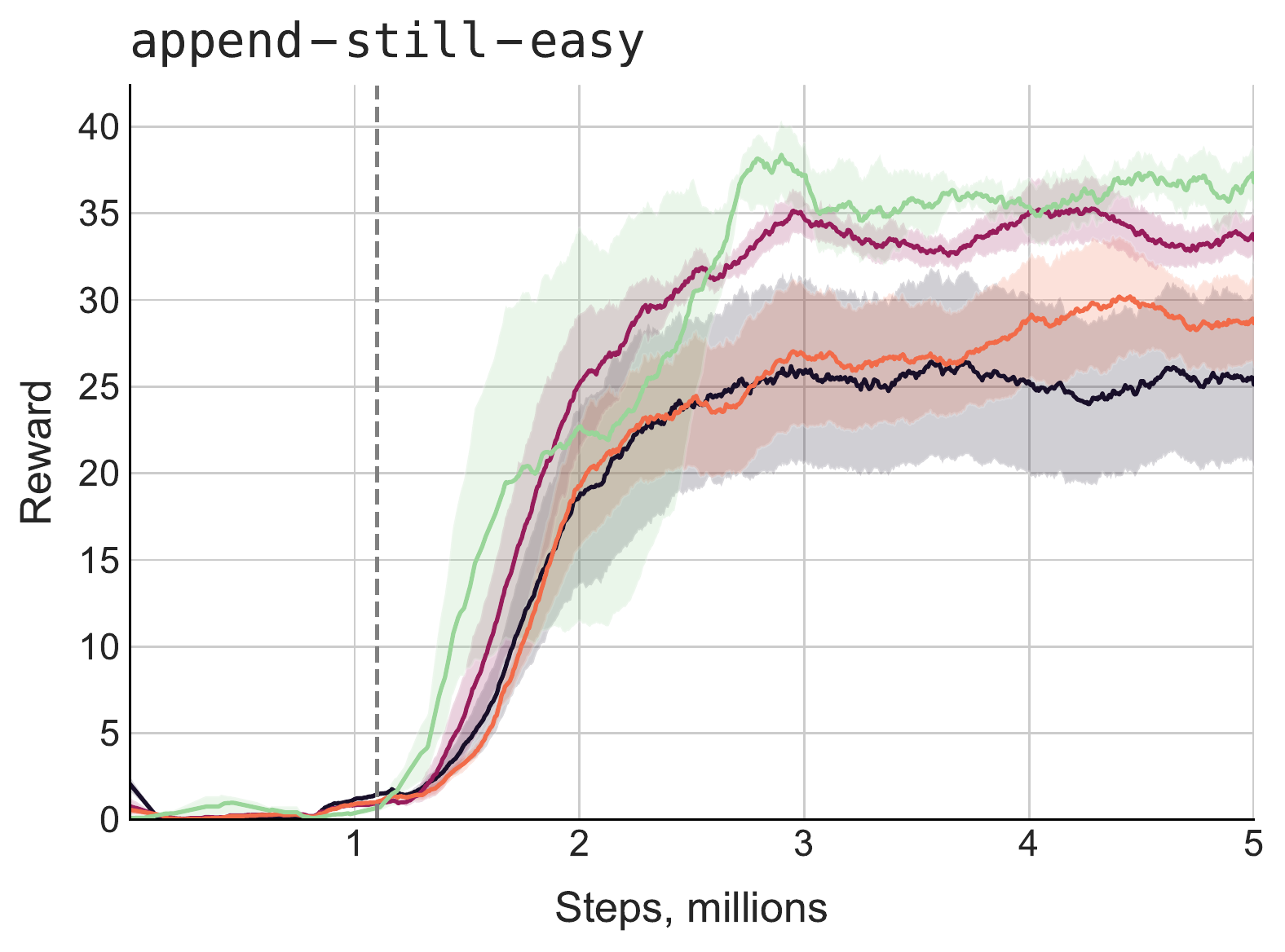}
}
\subfloat{
    \includegraphics[height=1.7in]{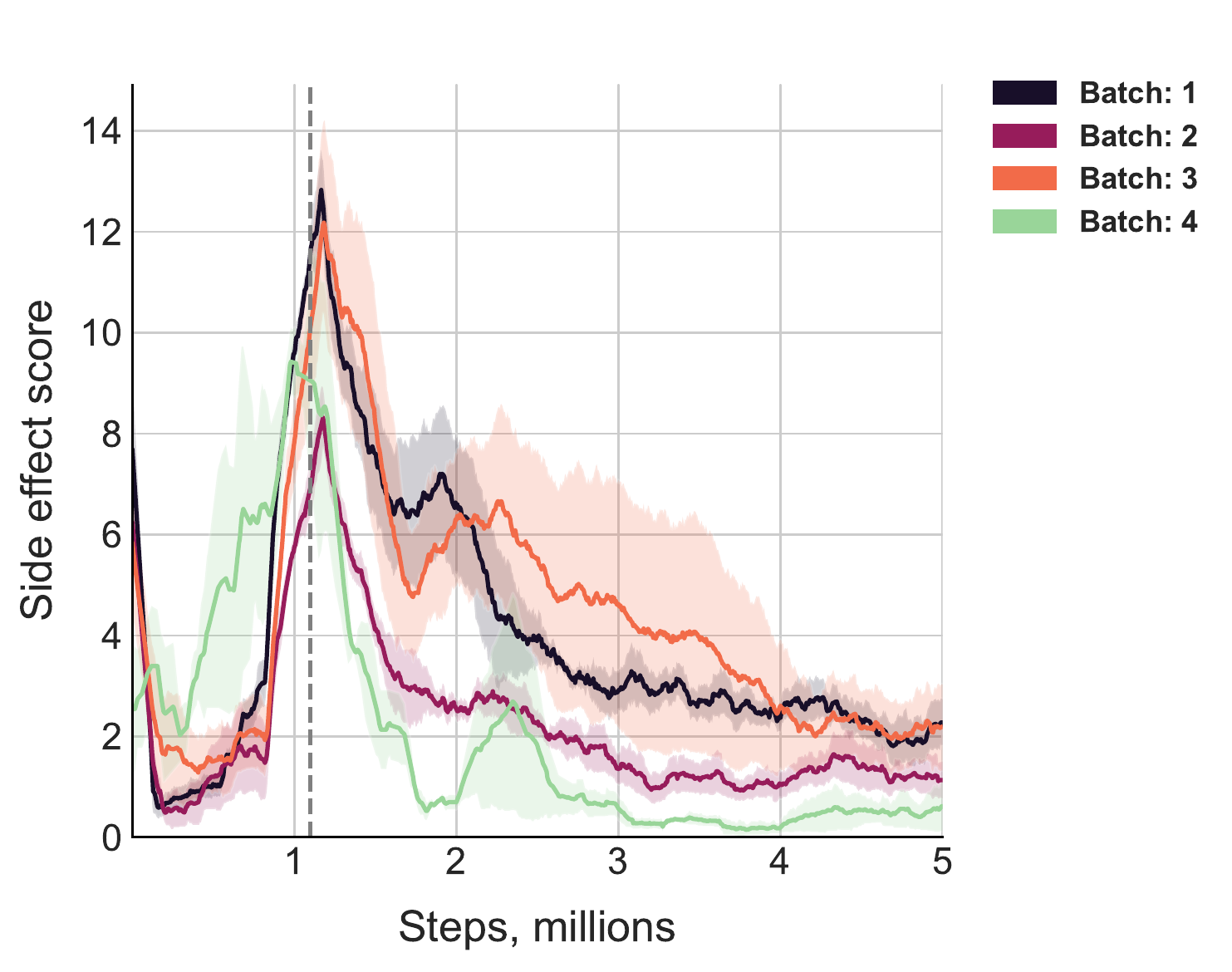}
}\\
\subfloat{
    \includegraphics[height=1.7in]{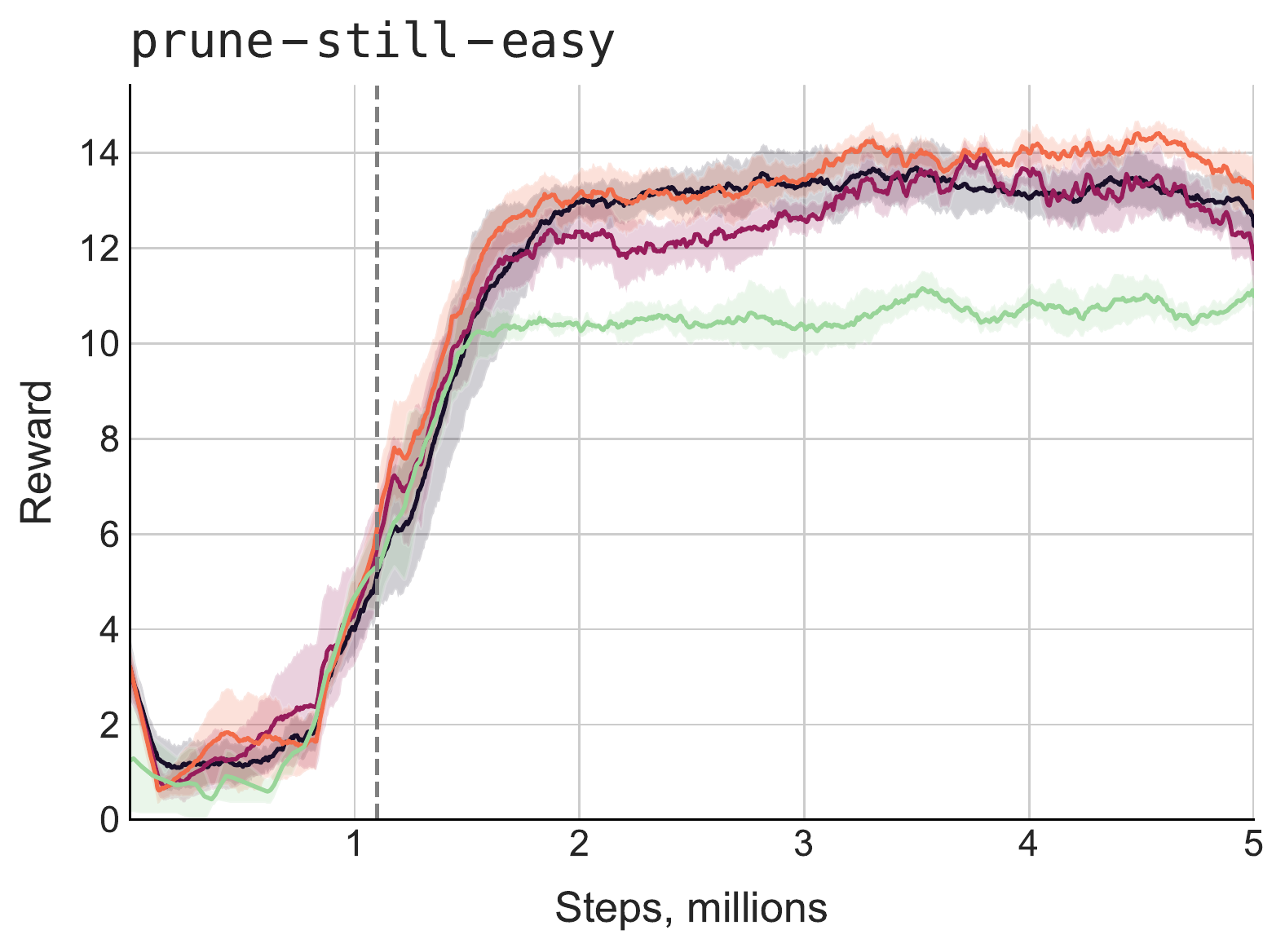}
}
\subfloat{
    \includegraphics[height=1.7in]{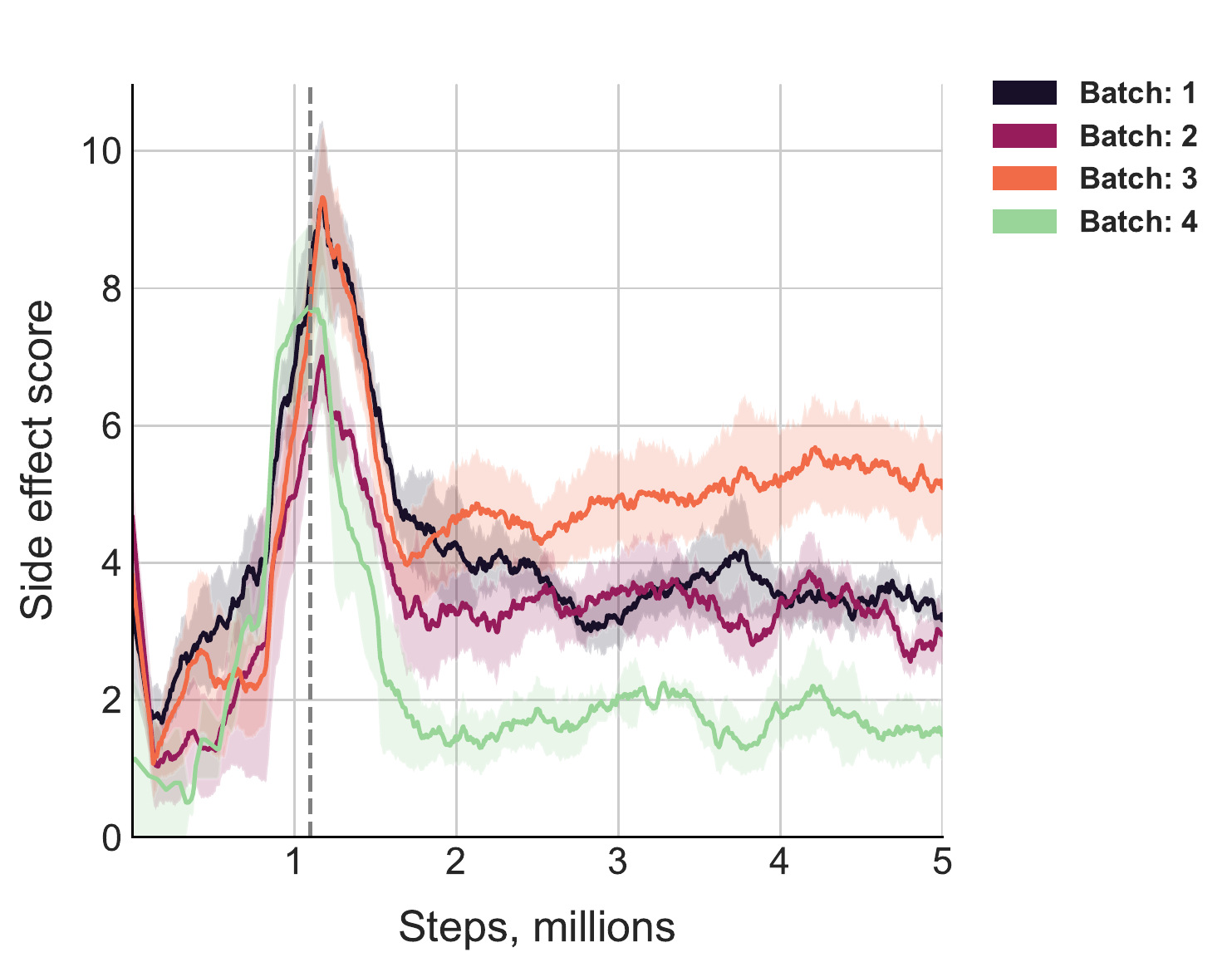}
}\\
\subfloat{
    \includegraphics[height=1.7in]{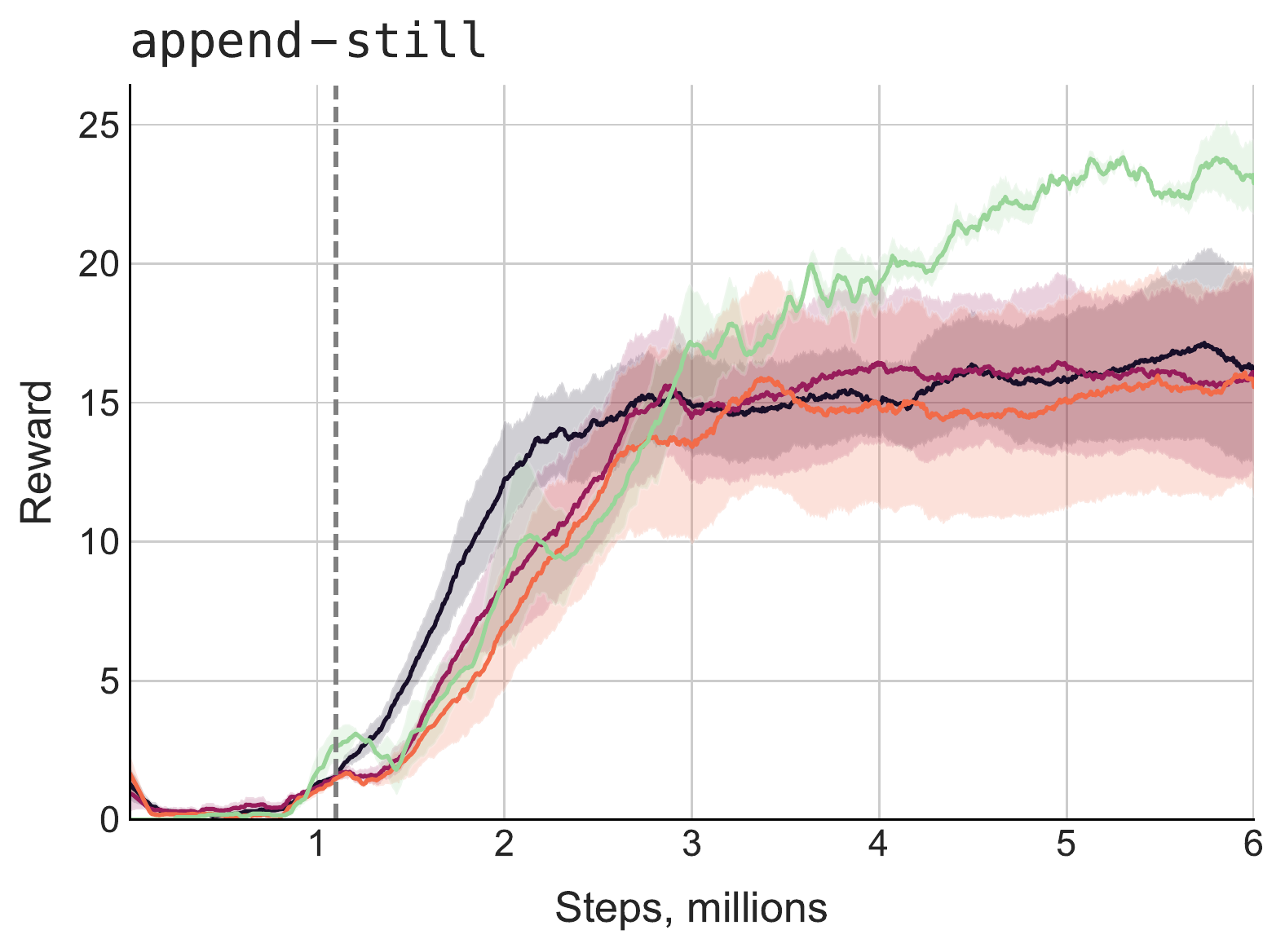}
}
\subfloat{
    \includegraphics[height=1.7in]{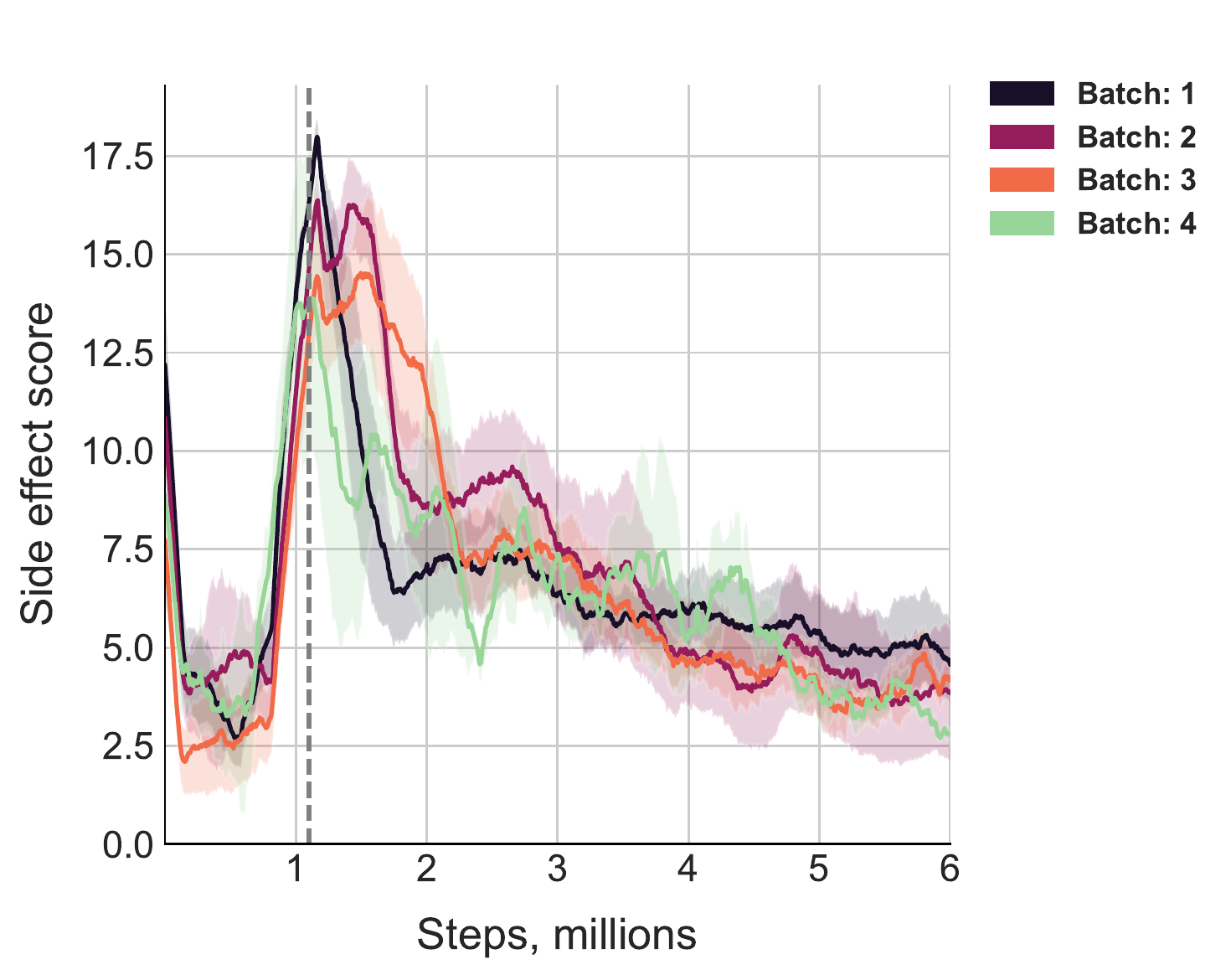}
}\\
\subfloat{
    \includegraphics[height=1.7in]{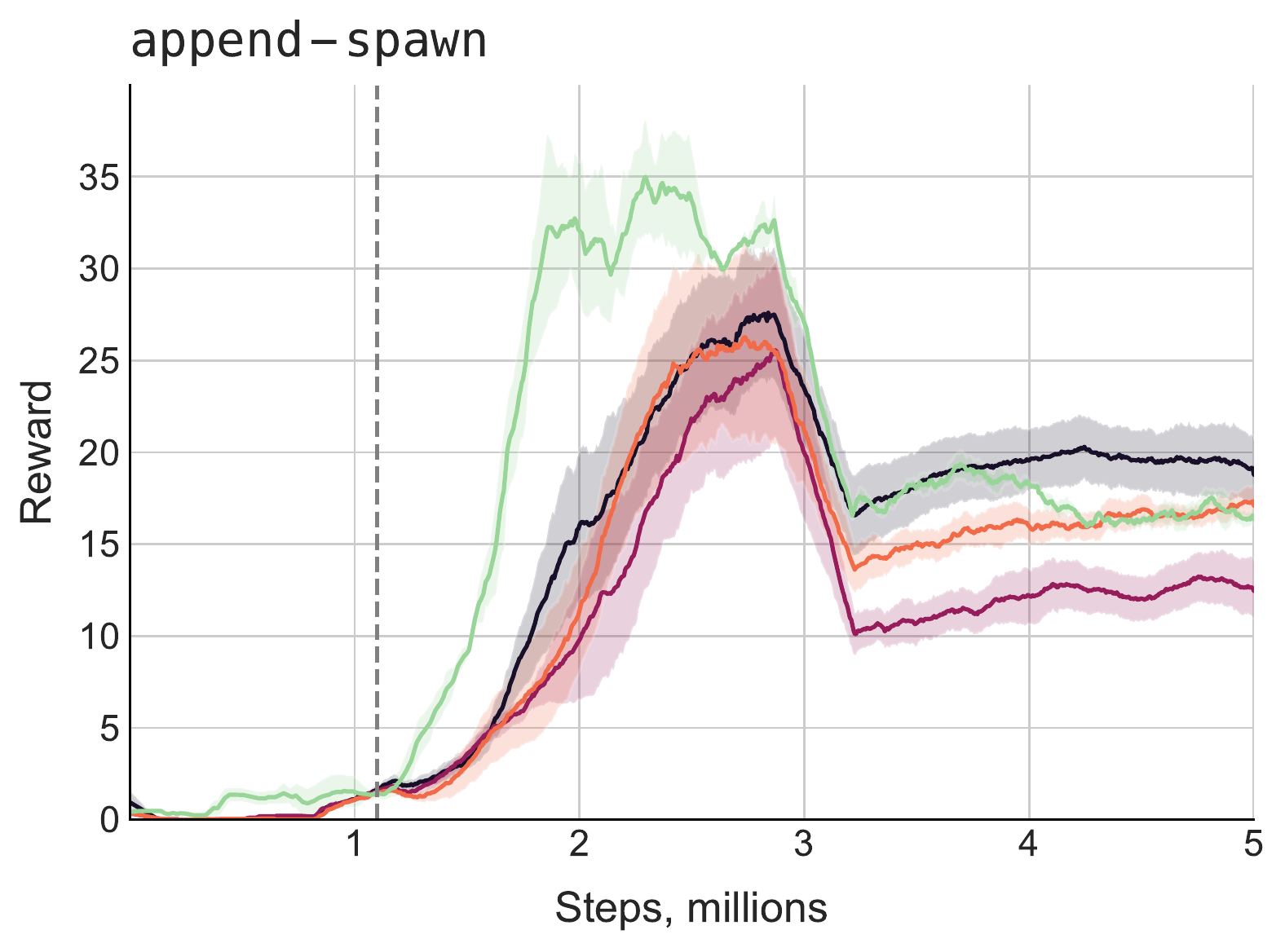}
}
\subfloat{
    \includegraphics[height=1.7in]{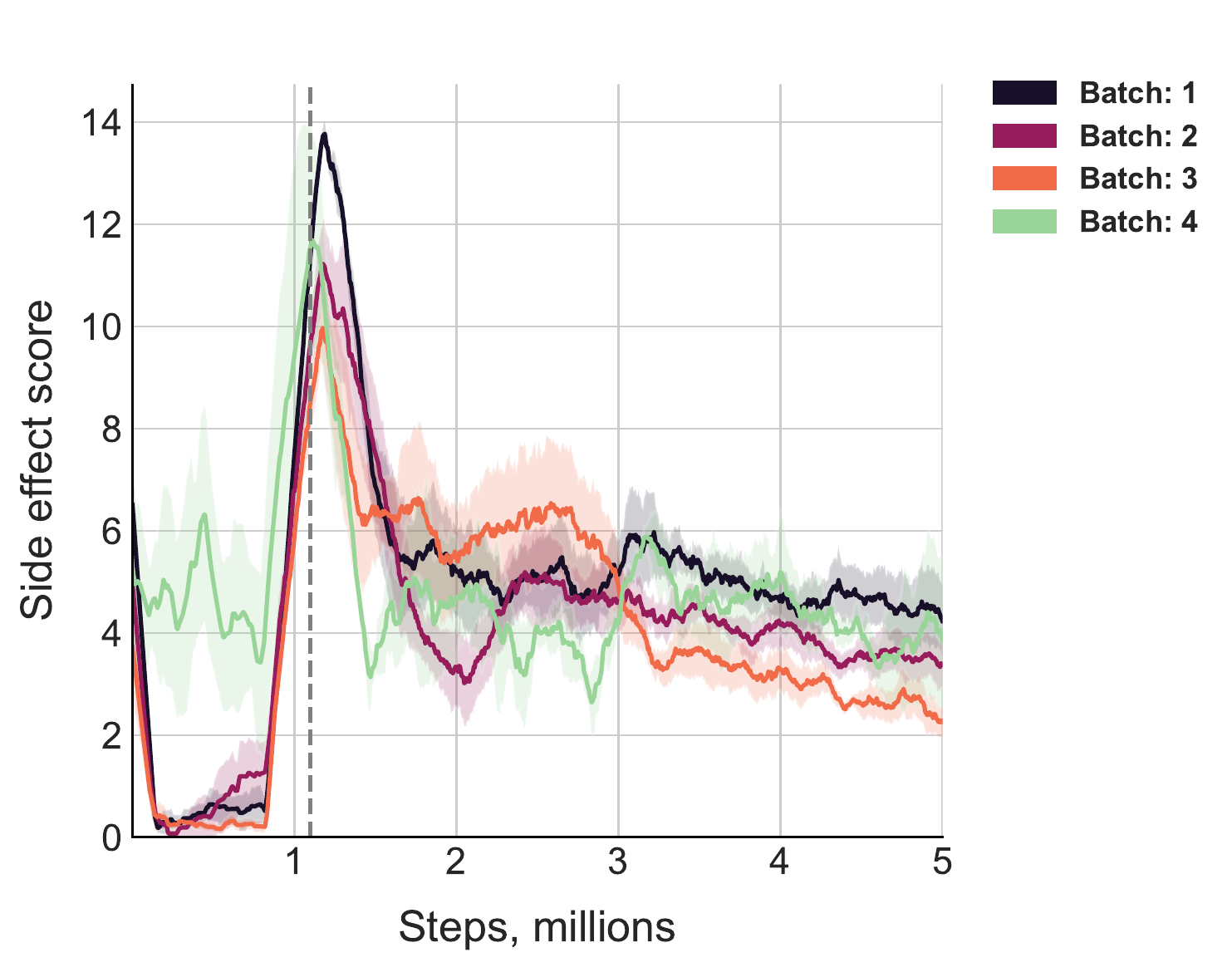}
}
\end{figure}

\begin{figure}[h]
\centering
\caption[Per-batch episode length curves for {\aupTT} in SafeLife]{Smoothed episode length curves for {\aupTT} on each of the four curricula.}\label{fig:batch-length}
\subfloat{
    \includegraphics[height=2in]{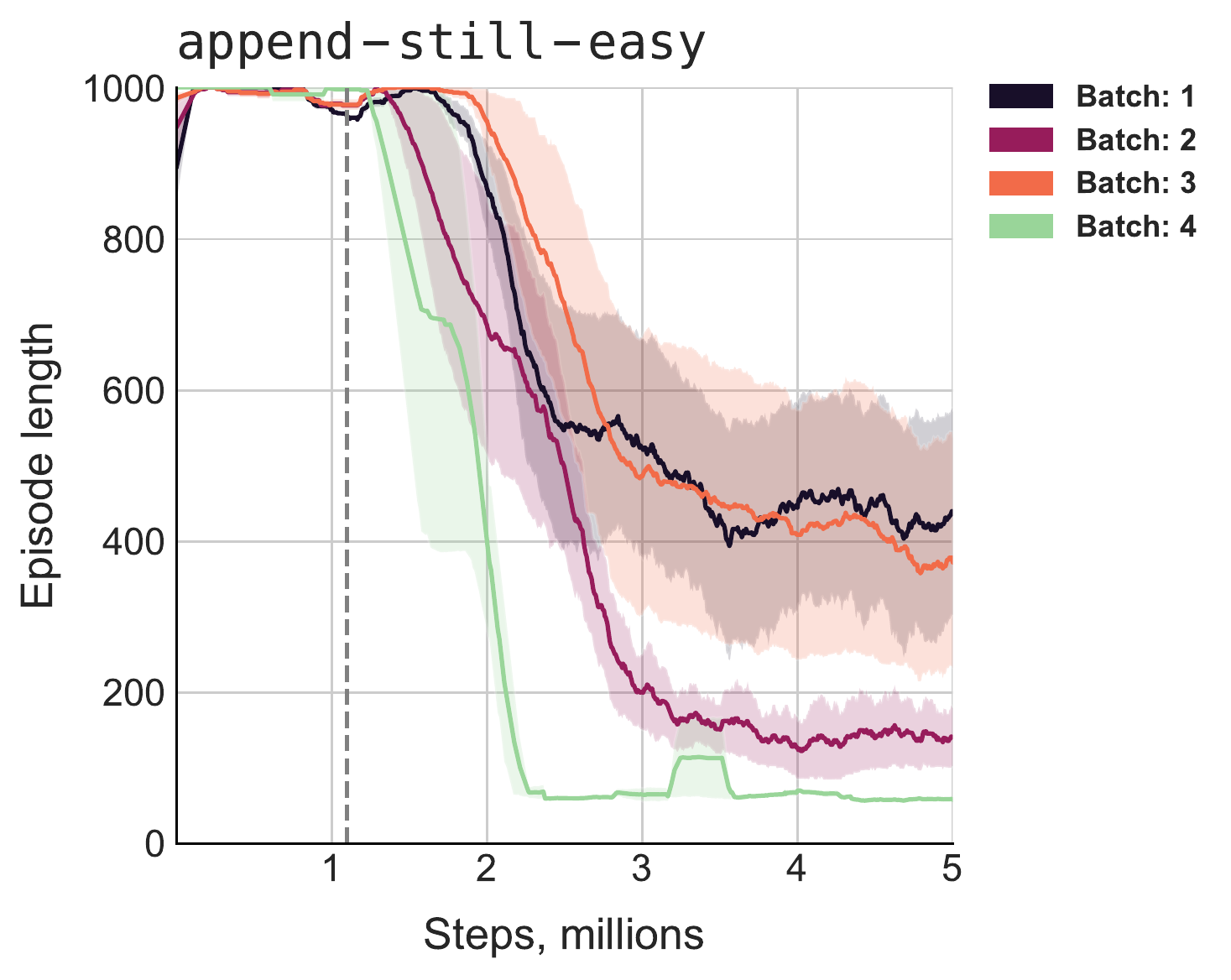}
}
\subfloat{
    \includegraphics[height=2in]{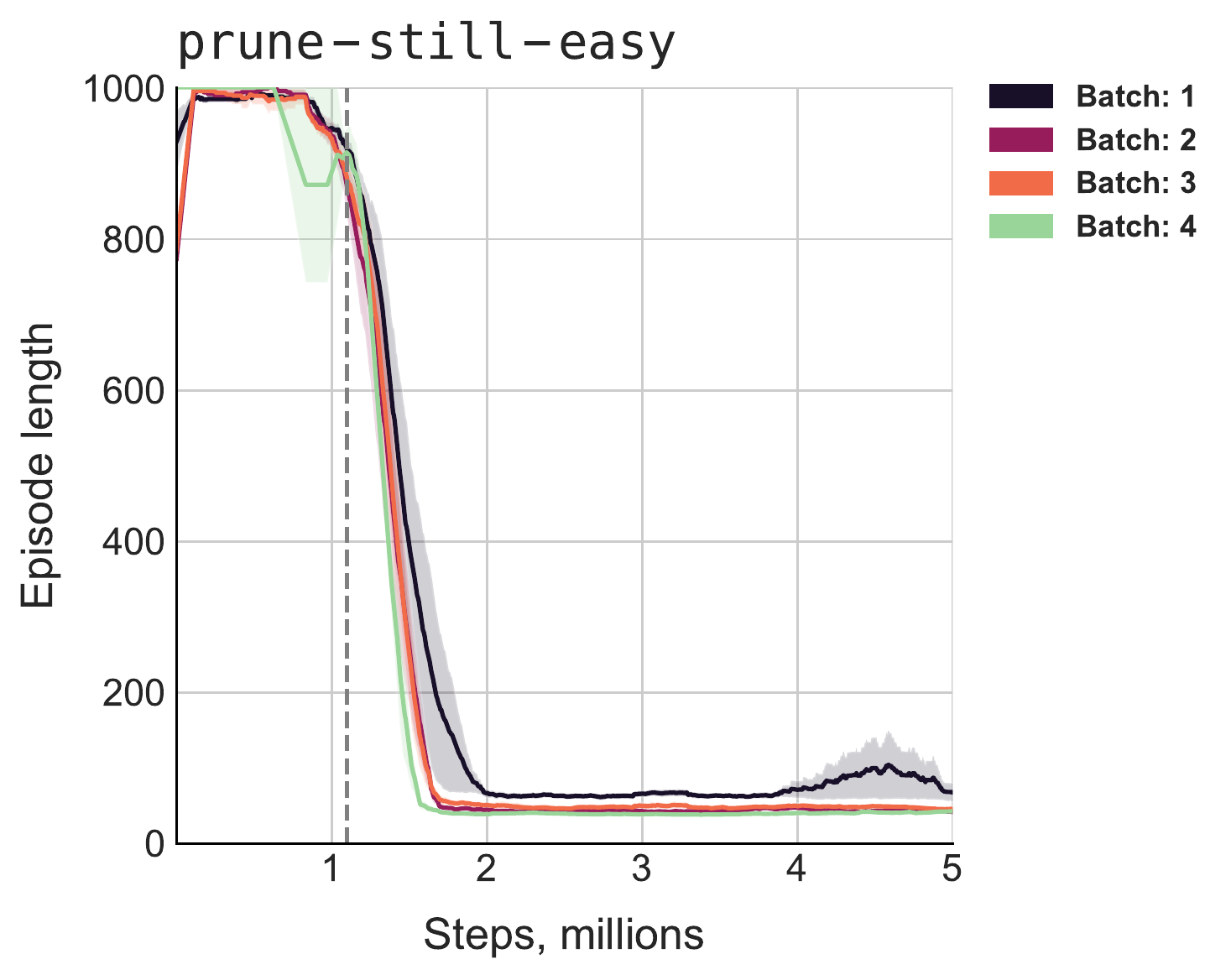}
}

\subfloat{
    \includegraphics[height=2in]{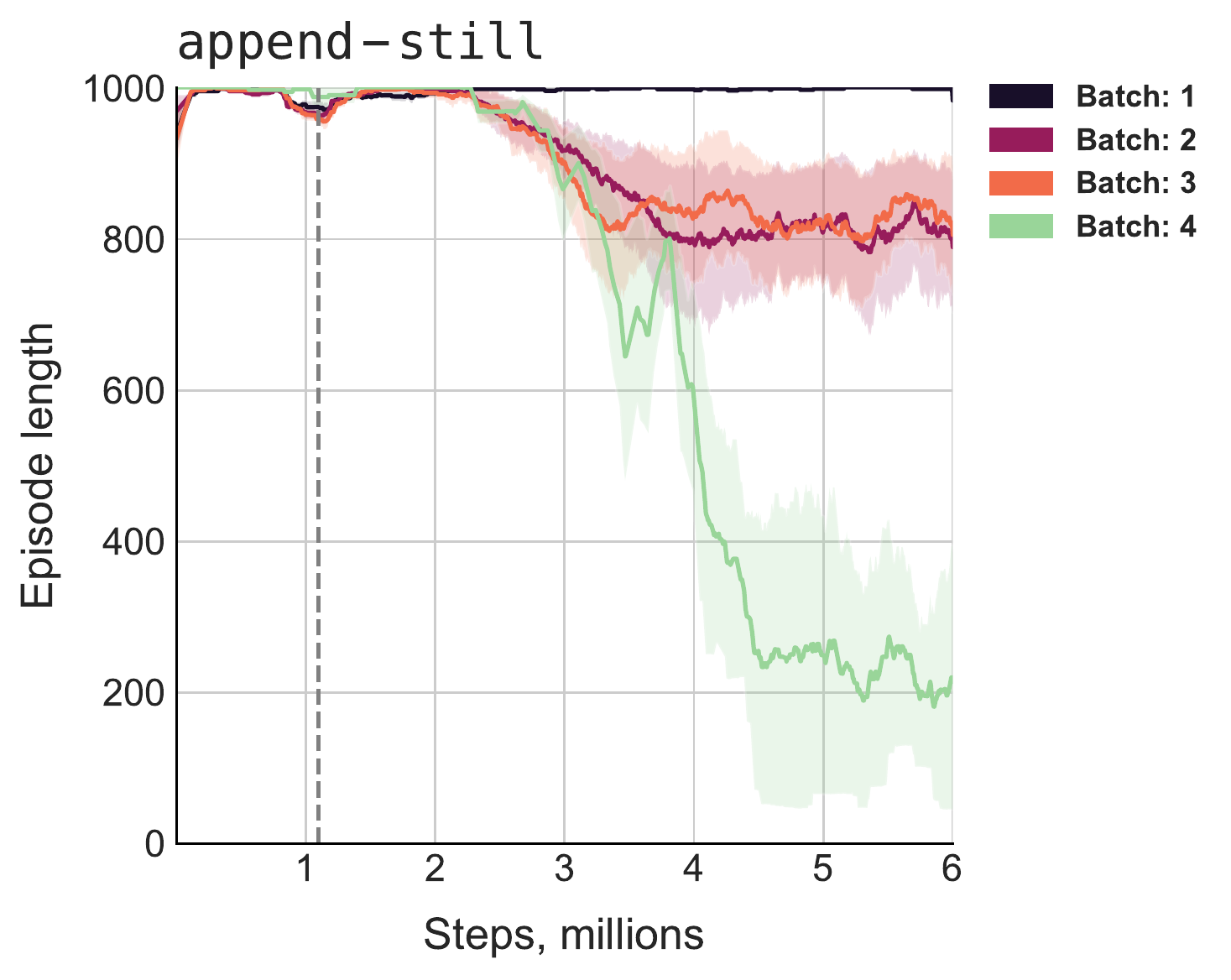}
}
\subfloat{
    \includegraphics[height=2in]{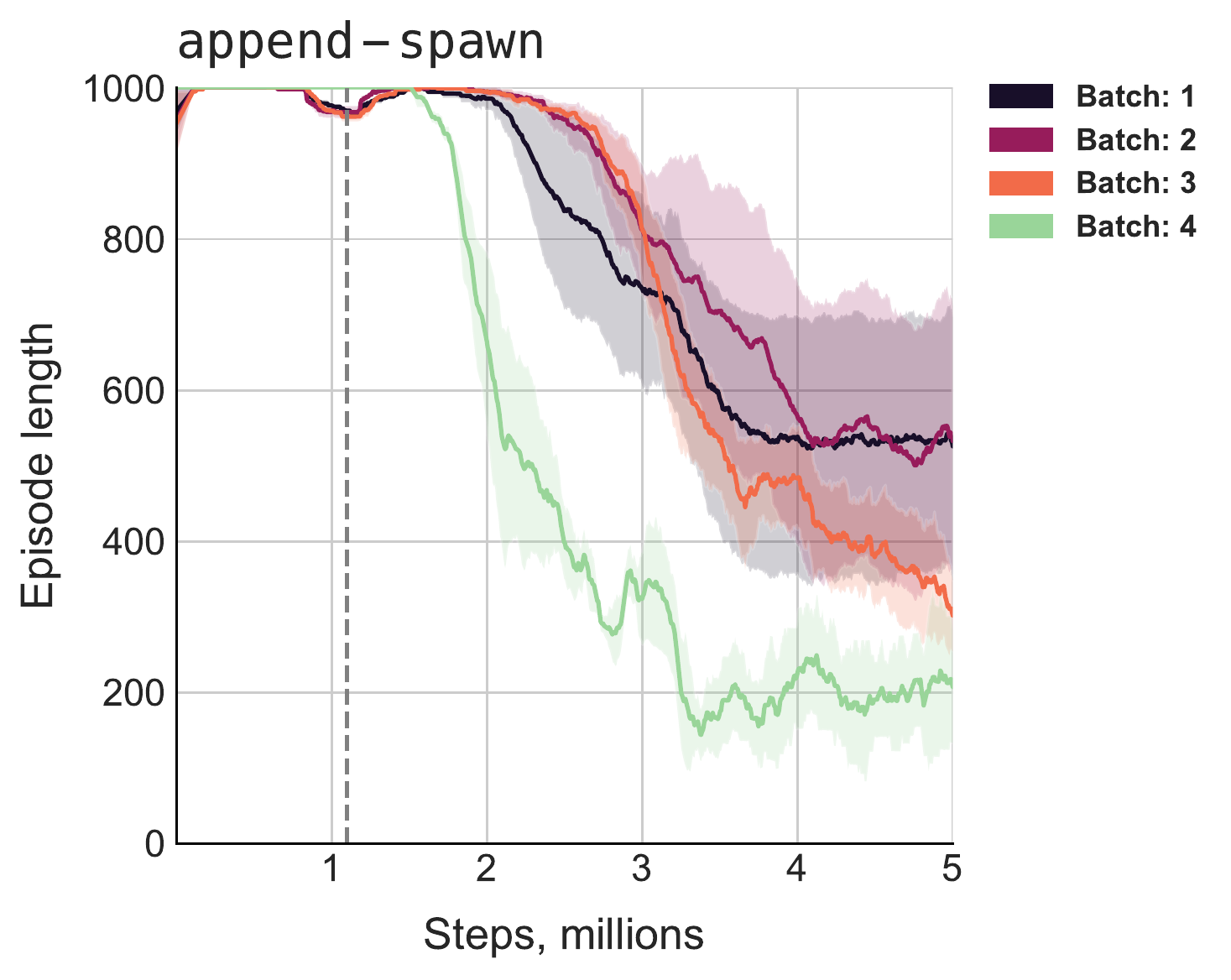}
}
\end{figure}

\Cref{fig:length} plots episode length and \cref{fig:aux-reward} plots auxiliary reward learning. \Cref{fig:individual} and \cref{fig:individ-length} respectively plot reward/side effects and episode lengths for each {\aupTT} seed. \Cref{fig:batch} and \cref{fig:batch-length} plot the same, averaged over each curriculum; these data suggest that {\aupTT}'s performance is sensitive to the randomly generated curriculum of environments.

\chapter{Formalizing The Problem of Side Effect Regularization}
\AddLoeChapter{definitions}
\section{Experiment details}\label{app:details}
The episode length is 20 for all episodes. Unlike~\cite{turner2020conservative}, the episode does not end after the agent reaches the green goal. This means that the agents can accrue many steps of environmental reward $R_\text{env}$. Therefore, {\aup} agents can achieve greater environmental reward for the same amount of penalty. To counterbalance this incentive, we multiply $R_\text{env}$ by $(1-\gamma)$.

We reuse the hyperparameters of~\cite{turner2020conservative}.{\footnote{Code available at \href{https://github.com/aseembits93/attainable-utility-preservation}{https://github.com/aseembits93/attainable-utility-preservation}.}} The learning rate is $\alpha\defeq 1$, and the discount rate is $\gamma\defeq .996$. We use the following {\aup} hyperparameter values: penalty coefficient $\lambda\defeq 0.01$, $\abs{\mathcal{R}}\defeq 20$ randomly generated auxiliary reward functions.

\subsection{Additional experiment}
We also tested an {\aup}-like agent optimizing reward function $R_\text{power penalty}(s,a)\defeq R_\text{env}(s,a) - \frac{\lambda}{\abs{\R}}\sum_{R_i \in \R} Q^*_{R_i}(s,a)-Q^*_{R_i}(s,\varnothing)$, which is the {\aup} reward function (\cref{def:aup-rf}) without the absolute value. This objective penalizes the agent for changes in its average optimal value, which is related to~\cite{turner_optimal_2020}'s $\pwr$. $R_\text{power penalty}$ produced the same prefix policies as $\rAUP$, and hence the same delayed specification scores.

\section{Theoretical results}
All results only apply to {\mdp}s with finite state and action spaces.

\begin{restatable}[Average optimal value~\cite{turner_optimal_2020}]{definition}{FORMavgVal}\label{FORM:def:vavg}
For bounded-support reward function distribution $\D$, the \emph{average optimal value} at state $s$ and discount rate $\gamma \in (0,1)$ is $\vavg[s,\gamma][\D]\defeq\E{R\sim \D}{\OptVf{s,\gamma}}.$
\end{restatable}

\uncertainOpt*
\begin{proof}
Suppose the agent starts at state $s_0$, and let $\gamma \in (0,1)$.
\begin{align}
   &\E{\substack{t \sim \timeDist,\\R\sim\D}}{\valSwitch{s_0,\gamma}} \\
   =& (1-\gamma) \E{\substack{t \sim \timeDist,\\R\sim\D}}{\sum_{i=0}^\infty \gamma^i \E{s_i \sim \piSwitch{\pi}}{R(s_i)}}\\
    =& (1-\gamma)\E{\substack{t \sim \timeDist,\\R\sim\D}}{\sum_{i=0}^{t-1} \gamma^i \E{s_i \sim \pi}{R(s_i)} + \gamma^t \E{s_t \sim \pi}{\Vf{s_t,\gamma}}}\label{eq:uncertain-piswitch-defn}\\
    =& (1-\gamma)\E{t \sim \timeDist}{\sum_{i=0}^{t-1} \gamma^i \E{s_i \sim \pi}{\ravg(s_i)}}\nonumber\\
    &\qquad+  \E{t \sim \timeDist,\, s_t \sim \pi}{\gamma^t (1-\gamma) \vavg[s_t,\gamma]}\label{eq:uncertain-lin-reward}\\
    =& (1-\gamma) \E{t \sim \timeDist}{\sum_{i=0}^{t} \gamma^i \E{s_i \sim \pi}{\ravg(s_i)}}\nonumber\\
    &\qquad + \E{t \sim \timeDist,\, s_t \sim \pi}{\gamma^{t+1} \pwr[s_t,\gamma]}.\label{eq:pwr-decomp}
\end{align}

\Cref{eq:uncertain-piswitch-defn} follows from the definition of the non-stationary $\piSwitch{\pi}$, and the fact that each $\pi^*_R$ is optimal for each $R$ at discount rate $\gamma$. \Cref{eq:uncertain-lin-reward} follows by the linearity of expectation and the definition of $\ravg$ and the definition of $\pwr$ (\cref{def:powRestate}). \Cref{eq:pwr-decomp} follows because $\vavg[s_t,\gamma]=\geom[\gamma] \pwr[s_t,\gamma] + \ravg(s_t)$.

If $\gamma = 0$ or $\gamma = 1$, the result holds in the respective limit because $\pwr$ has well-defined limits by Lemma 5.3 of~\cite{turner_optimal_2020}.
\end{proof}

\pwrMaxProp*
\begin{proof}
\Cref{item:1-special}: if  $\forall s_1, s_2 \in \St: \ravg(s_1)=\ravg(s_2)$, the first term on the right-hand side of \cref{eq:tradeoff-power-time} is constant for all policies $\pi$; if $\gamma=1$, this first term equals $0$. Therefore, under these conditions, $\pi$ maximizes \cref{eq:tradeoff-power-time} iff it maximizes $\E{t \sim \timeDist, s_t \sim \pi}{\gamma^{t+1}\pwr[s_t,\gamma]}$.

\Cref{item:2-special}: under these conditions, the second term on the right-hand side of \cref{eq:tradeoff-power-time} is constant for all policies $\pi$. Therefore, $\pi$ maximizes \cref{eq:tradeoff-power-time} iff it maximizes
\[(1-\gamma)\E{t \sim \timeDist}{\sum_{i=0}^{t} \gamma^i \E{s_i \sim \pi}{\ravg(s_i)}}.\]

If both \cref{item:1-special} and \cref{item:2-special} hold, it trivially follows that all $\pi$ are optimal prefix policies. If $\gamma=0$, all $\pi$ are optimal prefix policies, since reward is state-based and no actions taken by $\pi$ can affect expected return $\mathbf{ER}$.
\end{proof}

\timeDistGeom*
\begin{proof}
\begin{align}
    &\argsup_\pi \E{\substack{t \sim \timeDist,\\R\sim\D}}{\valSwitch{s,\gamma}}\label{eq:aup-soln-argmax}\\
    =&\argsup_\pi \E{t \sim \timeDist}{\sum_{i=0}^{t-1} \gamma^i \E{s_i \sim \pi \mid s_0}{\ravg(s_i)}} +   \E{\substack{t \sim \timeDist,\\s_t \sim \pi \mid s_0}}{\gamma^{t}\vavg[s_t,\gamma]}\label{eq:aup-soln-decomp}\\
    =&\argsup_\pi \sum_{t=1}^\infty \prob{\timeDist=t} \sum_{i=0}^{t-1} \gamma^i \E{s_i \sim \pi \mid s_0}{\ravg(s_i)} +   \E{\substack{t \sim \timeDist,\\s_t \sim \pi \mid s_0}}{\gamma^{t}\vavg[s_t,\gamma]}\\
    =&\argsup_\pi \sum_{t=1}^\infty (1-p)^{t-1}p \sum_{i=0}^{t-1} \gamma^i \E{s_i \sim \pi \mid s_0}{\ravg(s_i)} +   \sum_{t=1}^\infty (1-p)^{t-1}p\E{s_t \sim \pi \mid s_0}{\gamma^{t}\vavg[s_t,\gamma]}\label{eq:aup-soln-geom}\\
     =&\argsup_\pi \sum_{t=1}^\infty (1-p)^{t-1} p \sum_{i=0}^{t-1} \gamma^i \E{s_i \sim \pi \mid s_0}{\ravg(s_i)} + p\gamma \sum_{t=1}^\infty \gamAUP^{t-1}\E{s_t \sim \pi \mid s_0}{\vavg[s_t,\gamma]}\label{eq:aup-soln-gam1}\\
     =&\argsup_\pi \sum_{i=0}^\infty \gamAUP^i \E{s_i \sim \pi \mid s_0}{\ravg(s_i)} + p\gamma \sum_{i=1}^\infty \gamAUP^{i-1}\E{s_i \sim \pi \mid s_0}{\vavg[s_i,\gamma]}\label{eq:aup-soln-gam2}\\
     =&\argsup_\pi \ravg(s_0)+ \sum_{i=1}^\infty \gamAUP^{i-1} \E{s_i \sim \pi \mid s_0}{(1-p)\gamma\ravg(s_i) + p\gamma  \vavg[s_i,\gamma]}\label{eq:aup-soln-refactor}\\
     =&\argsup_\pi \sum_{i=1}^\infty \gamAUP^{i} \E{s_i \sim \pi \mid s_0}{(1-p)\ravg(s_i) + p\vavg[s_i,\gamma]}\label{eq:aup-soln-remove}\\
      =&\argsup_\pi \sum_{i=0}^\infty \gamAUP^{i} \E{s_i \sim \pi \mid s_0}{(1-p)\ravg(s_i) + p\vavg[s_i,\gamma]}\label{eq:aup-soln-add}\\
     =& \argsup_\pi V^\pi_{R'}(s_0,\gamAUP).\label{eq:argmax-Vpi}
\end{align}

\Cref{eq:aup-soln-argmax} follows from \cref{thm:uncertain-opt}. \Cref{eq:aup-soln-geom} follows because $\timeDist=\geomDist$. \Cref{eq:aup-soln-gam1} follows by the definition of $\gamAUP$.

In \cref{eq:aup-soln-gam1}, consider the double-sum on the left. For any given $i$, the portion of the sum with factor $\gamma^i$ equals
\begin{align}
    &\gamma^i\E{s_i \sim \pi \mid s_0}{\ravg(s_i)}\sum_{j=i}^\infty(1-p)^j p\nonumber\\
    &\quad=(1-p)^i\gamma^i\E{s_i \sim \pi \mid s_0}{\ravg(s_i)}p\sum_{j=0}^\infty(1-p)^{j}\\
   &\quad= (1-p)^i\gamma^i\E{s_i \sim \pi \mid s_0}{\ravg(s_i)}p\frac{1}{1-(1-p)}\label{eq:finite}\\
   &\quad= \gamAUP^i\E{s_i \sim \pi \mid s_0}{\ravg(s_i)}\frac{p}{1-(1-p)}\\
   &\quad= \gamAUP^i\E{s_i \sim \pi \mid s_0}{\ravg(s_i)}.\label{eq:i-factor}
\end{align}

The geometric identity holds for \cref{eq:finite} because $p>0 \implies (1-p)<1$. Therefore, \cref{eq:aup-soln-gam2} follows from \cref{eq:i-factor}.

\Cref{eq:aup-soln-refactor} follows by extracting the leading constant of $\ravg(s_0)$ and then expanding one of the $\gamAUP\defeq (1-p)\gamma$ factors of the first series. \Cref{eq:aup-soln-remove} follows by subtracting the constant $\ravg(s_0)$ by multiplying by $(1-p)> 0$, and by the fact that $(1-p)\gamma=\gamAUP$. \Cref{eq:aup-soln-add} follows because adding the constant $(1-p)\ravg(s_0)+p\vavg[s_0,\gamma]$ does not change the $\argsup$.

\Cref{eq:argmax-Vpi} follows by the definition of an on-policy value function and by the definition of $R'$. But $s_0$ was arbitrary, and so this holds for every state. Then  the policies in $\optPi[R',\gamAUP]$ satisfy the $\argsup$ for all states. $\optPi[R',\gamAUP]$ is non-empty because the {\mdp} is finite.
\end{proof}

\aupSolves*
\begin{proof}

\begin{align}
    &\argmax_\pi \E{\substack{t \sim \timeDist,\\R\sim\D}}{\valSwitch{s,\gamma}}\\
    =&\argmax_\pi \sum_{i=0}^\infty \gamAUP^{i} \E{s_i \sim \pi \mid s_0}{(1-p)\ravg(s_i) + p\vavg[s_i,\gamma]}\label{eq:aup-soln-add-repeat}\\
    =&\argmax_\pi \sum_{i=0}^\infty \gamAUP^{i} \E{s_i \sim \pi \mid s_0}{(1-p)\ravg(s_i) + p\vavg[s_i,\gamma]} \!-\! p\! \sum_{i=0}^\infty \gamAUP^{i} \!\E{s_i^\varnothing \sim \baseline \mid s_0}{\vavg[s_i^\varnothing,\gamma]}\label{eq:aup-soln-minus-inaction}\\
    =&\argmax_\pi \sum_{i=0}^\infty \gamAUP^{i} \E{s_i \sim \pi \mid s_0}{(1-p)\ravg(s_i) - p\prn{\E{s_i^\varnothing \sim \baseline \mid s_0}{\vavg[s_i^\varnothing,\gamma]}-\vavg[s_i,\gamma]}}   \\
    =&\argmax_\pi \sum_{i=0}^\infty \gamAUP^{i} \E{s_i \sim \pi \mid s_0}{(1-p)\ravg(s_i) - p\E{R\sim \D}{\E{s_i^\varnothing \sim \baseline \mid s_0}{\Vf{s_i^\varnothing,\gamma}}-\Vf{s_i,\gamma}}}\label{eq:aup-soln-expect}    \\
    =&\argmax_\pi \sum_{i=0}^\infty \gamAUP^{i} \E{s_i \sim \pi \mid s_0}{\ravg(s_i) - \frac{p}{1-p}\E{R\sim \D}{\E{s_i^\varnothing \sim \baseline \mid s_0}{\Vf{s_i^\varnothing,\gamma}}-\Vf{s_i,\gamma}}}.\label{eq:aup-soln-final}
\end{align}

\Cref{eq:aup-soln-add-repeat} follows from \cref{thm:time-geom-stationary}. \Cref{eq:aup-soln-minus-inaction} only subtracts a constant. \Cref{eq:aup-soln-expect} follows from the definition of $\vavg$, and \cref{eq:aup-soln-final} follows because dividing by $(1-p)>0$ does not affect the $\argmax$. But the expectation of \cref{eq:aup-soln-final} takes an expectation over $\rAssist(s_i \mid s)$, and its $\argmax$ equals the set of optimal policies for $\rAssist$ starting from state $s_0$.
\end{proof}

\chapter{Optimal Policies Tend To Seek Power}
\AddLoeChapter{theorems}
\AddLoeChapter{definitions}
\section{Comparing \texorpdfstring{$\pwrNoDist$}{POWER} with information-theoretic empowerment}\label{existing}
\begin{figure}[h]\centering
    \subfloat[][]{\includegraphics{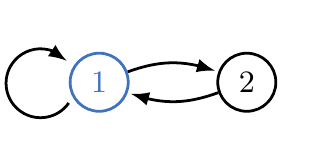}\label{fail-converge}}\hspace{3pt}
     \subfloat[][]{
     \includegraphics[]{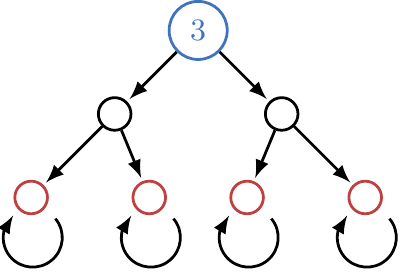}
     \label{empower-a}}\quad\hspace{5pt}
     \subfloat[][]{
     \includegraphics[]{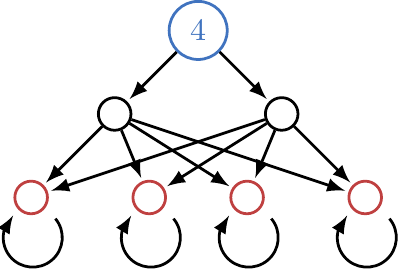}
    \label{empower-b}}
    \caption[Comparing $\pwrNoDist$ with information-theoretic empowerment]{Proposed empowerment measures fail to adequately capture how future choice is affected by present actions. In \protect\subref{fail-converge}: $\mathfrak{E}_n(\col{blue}{s_1})$ varies depending on whether $n$ is even; thus, $\lim_{n\to \infty}\mathfrak{E}_n(\col{blue}{s_1})$ does not exist. In \protect\subref{empower-a} and  \protect\subref{empower-b}: $\forall n: \mathfrak{E}_n(\col{blue}{s_3})=\mathfrak{E}_n(\col{blue}{s_4})$, even though $\col{blue}{s_4}$ allows greater control over future state trajectories than $\col{blue}{s_3}$ does. For example, suppose that in both \protect\subref{empower-a} and \protect\subref{empower-b}, the leftmost black state and the rightmost red state have $ 1$ reward while all other states have $0$ reward. In \protect\subref{empower-b}, the agent can independently maximize the intermediate black-state reward and the delayed red-state reward. Independent maximization is not possible in \protect\subref{empower-a}.
    }
    \label{fig:empower_fail}
\end{figure}

\citet{salge_empowermentintroduction_2014} define  information-theoretic \emph{empowerment} as the maximum possible mutual information between the agent's actions and the state observations $n$ steps in the future, written $\mathfrak{E}_n(s)$. This notion requires an arbitrary choice of horizon, failing to account for the agent's discount rate $\gamma$. ``In a  discrete deterministic world empowerment reduces to the logarithm of the number of sensor states reachable with the available actions'' \citep{salge_empowermentintroduction_2014}. \Cref{fig:empower_fail} demonstrates how empowerment can return counterintuitive verdicts with respect to the agent's control over the future.

$\pwrNoDist$ returns intuitive answers in these situations. $\lim_{\gamma\to 1}\pwr[\col{blue}{s_1},\gamma]$ converges by \cref{thm:cont-power}. Consider the obvious involution $\phi$ which takes each state in \cref{empower-a} to its counterpart in \cref{empower-b}. Since $\phi\cdot\Fnd(\col{blue}{s_3})\subsetneq \Fnd(\col{blue}{s_4})=\F(\col{blue}{s_4})$, \cref{prop:more-opt} proves that $\forall \gamma\in [0,1]:\pwr[\col{blue}{s_3},\gamma][\Dbd]\leqMost[][\DSetBd]\pwr[\col{blue}{s_4},\gamma][\Dbd]$, with the proof of \cref{prop:more-opt} showing strict inequality under all $\Diid$ when $\gamma\in(0,1)$.

Empowerment can be adjusted to account for these cases, perhaps by considering the channel capacity between the agent's actions and the state trajectories induced by stationary policies. However, since $\pwrNoDist$ is formulated in terms of optimal value, we believe that $\pwrNoDist$ is better suited for {\mdp}s than information-theoretic empowerment is.

\section{Seeking \texorpdfstring{$\pwrNoDist$}{POWER} can be a detour}\label{app:not-always}
One might suspect that optimal policies tautologically tend to seek $\pwrNoDist$. This intuition is wrong.

\begin{figure}[!ht]\centering
    \includegraphics{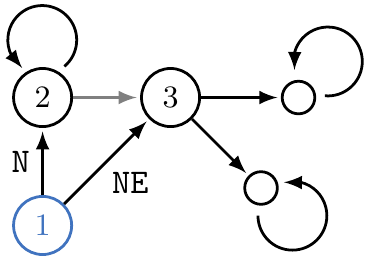}
    \caption[$\pwrNoDist$-seeking is not necessarily convergently instrumental]{$\pwrNoDist$-seeking is not necessarily convergently instrumental.}
    \label{fig:power-not-ic}
\end{figure}

\begin{restatable}[Greater $\pwr$ does not imply greater $\Prb_{\Dbd}$]{prop}{PowerNotIC}\label{prop:power-not-ic} Action $a$ seeking more $\pwr$ than $a'$ at state $s$ and $\gamma$ does not imply that $\optprob[\Dbd]{s,a,\gamma}\geq\optprob[\Dbd]{s,a',\gamma}$.
\end{restatable}
\begin{proof}
Consider the environment of \cref{fig:power-not-ic}. Let $\Dist_u\defeq \text{unif}(0,1)$, and consider $\Diid[\Dist_u]$, which has bounded support. Direct computation\footnote{In small deterministic {\mdp}s, the $\pwrNoDist$ and optimality probability of the maximum-entropy reward function distribution can be computed using  \href{https://github.com/loganriggs/Optimal-Policies-Tend-To-Seek-Power}{https://github.com/loganriggs/Optimal-Policies-Tend-To-Seek-Power}.} of $\pwrNoDist$ yields \[\pwr[\sn,1][\Diid[\Dist_u]]=\frac{3}{4}> \frac{2}{3}=\pwr[\sne,1][\Diid[\Dist_u]].\] Therefore, the action $\texttt{N}$ seeks more $\pwr[][\Diid[\Dist_u]]$ than $\texttt{NE}$ at state $\startApp$ and $\gamma=1$. However, $\optprob[\Diid[\Dist_u]]{\startApp,\texttt{N},1}=\frac{1}{3}<\frac{2}{3}=\optprob[\Diid[\Dist_u]]{\startApp,\texttt{NE},1}$.
\end{proof}

\begin{restatable}[Fraction of orbits which agree on weak optimality]{lem}{fracOrbi}\label{lem:half-orbit-geq}
Let $\distSet\subseteq \Delta(\rewardVS)$, and suppose $f_1,f_2:\Delta(\rewardVS)\to \reals$ are such that $f_1(\D) \geqMost[][\distSet] f_2(\D)$. Then for all $\D\in\distSet$, $\frac{\abs{\set{\D' \in\orbi[\D]\mid f_1(\D')\geq f_2(\D')}}}{\abs{\orbi[\D]}}\geq \dfrac{1}{2}$.
\end{restatable}
\begin{proof}
All $\D'\in \orbi[\D]$ such that $f_1(\D')= f_2(\D')$ satisfy $f_1(\D')\geq f_2(\D')$.

Otherwise, consider the $\D'\in \orbi[\D]$ such that $f_1(\D')\neq f_2(\D')$. By the definition of $\geqMost[][]$ (\cref{def:ineq-most-dists}), at least $\frac{1}{2}$ of these $\D'$ satisfy $f_1(\D')> f_2(\D')$, in which case $f_1(\D')\geq f_2(\D')$. Then the desired inequality follows.
\end{proof}

\begin{restatable}[$\geq_\text{most}$ and trivial orbits]{lem}{identGeqMost}\label{lem:ident-geq-most}
Let $\distSet\subseteq \Delta(\rewardVS)$ and suppose $f_1(\D) \geqMost[][\distSet] f_2(\D)$. For all reward function distributions $\D\in\distSet$ with one-element orbits, $f_1(\D)\geq f_2(\D)$. In particular, $\D$ has a one-element orbit when it distributes reward identically and independently ({\iid}) across states.
\end{restatable}
\begin{proof}
By \cref{lem:half-orbit-geq}, at least half of the elements $\D'\in \orbi[\D]$ satisfy $f_1(\D')\geq f_2(\D')$. But $\abs{\orbi[\D]}=1$, and so $f_1(\D)\geq f_2(\D)$ must hold.

If $\D$ is {\iid}, it has a one-element orbit due to the assumed identical distribution of reward.
\end{proof}

\begin{restatable}[Actions which tend to seek $\pwrNoDist$ do not necessarily tend to be optimal]{prop}{PowerNotICMost}\label{prop:power-not-ic-most} Action $a$ tending to seek more $\pwrNoDist$ than $a'$ at state $s$ and $\gamma$ does not imply that \[\optprob[\Dany]{s,a,\gamma}\geqMost \optprob[\Dany]{s,a',\gamma}.\]
\end{restatable}
\begin{proof}
Consider the environment of \cref{fig:power-not-ic}. Since $\RSDnd[\sne]\subsetneq\RSD[\sn]$, \cref{RSDSimPower} shows that $\pwr[\sn,1][\Dbd]\geqMost[][\DSetBd] \pwr[\sne,1][\Dbd]$ via $s'\defeq \sne,s\defeq \sn,\phi$ the identity permutation (which is an involution). Therefore, $\texttt{N}$ tends to seek more $\pwrNoDist$ than $\texttt{NE}$ at state $\startApp$ and $\gamma=1$.

If $\optprob[\Dany]{\startApp,\texttt{N},1}\geqMost\optprob[\Dany]{\startApp,\texttt{NE},1}$, then \cref{lem:ident-geq-most} shows that $\optprob[\Diid]{\startApp,\texttt{N},1}\geq\optprob[\Diid]{\startApp,\texttt{NE},1}$ for all $\Diid$. But the proof of \cref{prop:power-not-ic} showed that $\optprob[\Diid[\Dist_u]]{\startApp,\texttt{N},1}<\optprob[\Diid[\Dist_u]]{\startApp,\texttt{NE},1}$ for $\Dist_u\defeq\text{unif}(0,1)$. Therefore, it can't be true that $\optprob[\Dany]{\startApp,\texttt{N},1}\geqMost\optprob[\Dany]{\startApp,\texttt{NE},1}$.
\end{proof}

\section{Sub-optimal \texorpdfstring{$\pwrNoDist$}{POWER}}\label{sec:suboptimal-power}
\begin{figure}[ht]
    \centering
    \includegraphics{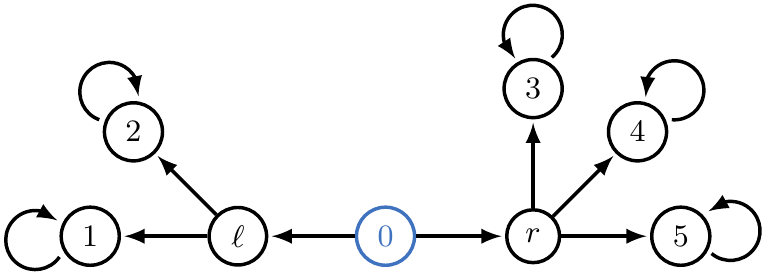}
    \caption[Illustrating power-seeking for suboptimal agents]{$\col{blue}{s_0}$ is the starting state, and $\abs{\A}=10^{10^{10}}$. At $\col{blue}{s_0}$, half of the actions lead to $s_\ell$, while the other half lead to $s_r$. Similarly, half of the actions at $s_\ell$ lead to $s_1$, while the other half lead to $s_2$. At $s_r$, one action leads to $s_3$, one action leads to $s_4$, and the remaining $10^{10^{10}}-2$ actions lead to $s_5$.  \label{fig:subopt-power}}
\end{figure}

In certain situations, $\pwrNoDist$ returns intuitively surprising verdicts. There exists a policy under which the reader chooses a winning lottery ticket, but it seems wrong to say that the reader has the power to win the lottery with high probability. For various reasons, humans and other bounded agents are generally incapable of computing optimal policies for arbitrary objectives. More formally, consider the rewardless {\mdp} of \cref{fig:subopt-power}.

Consider a model-based RL agent with black-box simulator access to this environment. The agent has no prior information about the model, and so it acts randomly. Before long, the agent has probably learned how to navigate from $\col{blue}{s_0}$ to states $s_\ell$, $s_r$, $s_1$, $s_2$, and $s_5$. However, over any reasonable timescale, it is extremely improbable that the agent discovers the two actions respectively leading to $s_3$ and $s_4$.

Even provided with a reward function $R$ and the discount rate $\gamma$, the agent has yet to learn the relevant environmental dynamics, and so many of its policies are far from optimal. Although  \cref{prop:more-opt} shows that $\forall \gamma \in[0,1]$:
\begin{equation}
    \pwr[s_\ell,\gamma][\Dbd] \leqMost[][\DSetBd] \pwr[s_r,\gamma][\Dbd],
\end{equation}
there is a sense in which $s_\ell$ gives this agent more power.

We formalize a bounded agent's goal-achievement capabilities with a function $\pol[]$, which takes as input a reward function and a discount rate, and returns a policy. Informally, this is the best policy which the agent knows about. We can then calculate $\pwr$ with respect to $\pol[]$.

\begin{restatable}[Suboptimal $\pwrNoDist$]{definition}{powPolicy}\label{def:pow-pol} Let $\Pi_\Delta$ be the set of stationary stochastic policies, and let $\text{pol} : \rewardSpace \times [0,1] \to \Pi_\Delta$. For $\gamma \in [0,1]$,
\begin{align}
\pwrPol{s, \gamma}\defeq \E{\substack{R\sim\Dbd,\\a\sim \pol(s),\\s'\sim T\prn{s,a}}}{\lim_{\gamma^*\to \gamma} (1-\gamma^*)\Vf[\pol]{s',\gamma^*}}.
\end{align}

By \cref{lem:power-id}, $\pwr$ is the special case where $\forall R\in\rewardSpace,\gamma\in[0,1]: \pol\in \optPi$. We define $\pwrPol{}$-seeking similarly as in \cref{def:pow-seek}.
\end{restatable}

$\pwrPol{\col{blue}{s_0},1}$ increases as the policies returned by $\pol[]$ are improved. We illustrate this by considering the $\Diid$ case.

\begin{enumerate}
\item[$\text{pol}_1$] The model is initially unknown, and so $\forall R,\gamma:\text{pol}_1(R,\gamma)$ is a uniformly random policy. Since $\text{pol}_1$ is constant on its inputs, $\pwrPol[\text{pol}_1][\Diid]{\col{blue}{s_0},1}=\EX$ by the linearity of expectation and the fact that $\Diid$ distributes reward independently and identically across states.
\item[$\text{pol}_2$] The agent knows the dynamics, except that it does not know how to reach $s_3$ or $s_4$. At this point, $\text{pol}_2(R,1)$ navigates from $\col{blue}{s_0}$ to the average-optimal choice among three terminal states: $s_1$, $s_2$, and $s_5$. Therefore, $\pwrPol[\text{pol}_2]{\col{blue}{s_0},1}=\Edraws{3}$.
\item[$\text{pol}_3$] The agent knows the dynamics, the environment is small enough to solve explicitly, and so $\forall R,\gamma:\text{pol}_3(R,\gamma)$ is an optimal policy. $\text{pol}_3(R,1)$ navigates from $\col{blue}{s_0}$ to the average-optimal choice among all five terminal states. Therefore, $\pwrPol[\text{pol}_3]{\col{blue}{s_0},1}=\Edraws{5}$.
\end{enumerate}

As the agent learns more about the environment and improves $\pol[]$, $\pwrPol{}$ increases. The agent seeks $\pwrPol[\text{pol}_2]{}$ by navigating to $s_\ell$ instead of $s_r$, but seeks more $\pwr$ by navigating to $s_r$ instead of $s_\ell$. Intuitively, bounded agents gain power by improving $\pol[]$ and by formally seeking $\pwrPol{}$ within the environment.

\subsection{Contributions of independent interest}\label{app:contrib}
We developed new basic {\mdp} theory by exploring the structural properties of visit distribution functions. Echoing \citet{wang_dual_2007,wang_stable_2008}, we believe that this area is interesting and underexplored.

\subsubsection{Optimal value theory}
\Cref{lem:norm-value-lip} shows that $f(\gamma^*)\defeq\lim_{\gamma^*\to \gamma}(1-\gamma^*)\OptVf{s,\gamma^*}$ is Lipschitz continuous on $\gamma\in[0,1]$, with Lipschitz constant depending only on $\lone{R}$.  For all states $s$ and policies $\pi\in\Pi$, \cref{smoothOnPol} shows that $V^\pi_R(s,\gamma)$ is rational on $\gamma$.

Optimal value has a well-known dual formulation: $\OptVf{s,\gamma}=\max_{\f\in\F(s)}\f(\gamma)^\top \rf$.
\begin{restatable*}[$\forall\gamma\in[0,1):\OptVf{s,\gamma}=\max_{\f\in\Fnd(s)}\f(\gamma)^\top \rf$]{lem}{optVfFndRestrict}\label{cor:opt-vf-restrict-fnd}
\end{restatable*}
In a fixed rewardless {\mdp}, \cref{cor:opt-vf-restrict-fnd} may enable more efficient computation of optimal value functions for multiple reward functions.

\subsubsection{Optimal policy theory}
\Cref{transferDiscount} demonstrates how to preserve optimal incentives while changing the discount rate.

\begin{restatable*}[How to transfer optimal policy sets across discount rates]{prop}{transferDiscount}\label{transferDiscount}
Suppose reward function $R$ has optimal policy set $\optPi[R,\gamma]$ at discount rate $\gamma\in(0,1)$. For any $\gamma^*\in(0,1)$, we can construct a reward function $R'$ such that $\optPi[R',\gamma^*]=\optPi$. Furthermore, $\OptVf[R']{\cdot,\gamma^*}=\OptVf[R]{\cdot, \gamma}$.
\end{restatable*}

\subsubsection{Visit distribution theory}
While \citet{regan_robust_2010} consider  a visit distribution function $\f\in\F(s)$ to be non-dominated if it is optimal for some reward function in a set $\R\subseteq \rewardVS$, our stricter \cref{def:nd} considers $\f$ to be non-dominated when $\exists\rf\in\rewardVS, \gamma\in(0,1):\f(\gamma)^\top \rf > \max_{\f'\in\F(s)\setminus\set{\f}}\f'(\gamma)^\top\rf$.

\section{Theoretical results} \label{app:proofs}
\begin{restatable}[A policy is optimal iff it induces an optimal visit distribution at every state]{lem}{optShare}\label{lem:opt-pol-visit-iff}
Let $\gamma \in (0,1)$ and let $R$ be a reward function. $\pi \in \optPi$ iff $\pi$ induces an optimal visit distribution at every state.
\end{restatable}
\begin{proof}
By definition, a policy $\pi$ is optimal iff $\pi$ induces the maximal on-policy value at each state, which is true iff $\pi$ induces an optimal visit distribution at every state (by the dual formulation of optimal value functions).
\end{proof}

\begin{restatable}[Transition matrix induced by a policy]{definition}{transMatrix}\label{def:trans-matrix}
$\mathbf{T}^\pi$ is the transition matrix induced by policy $\pi \in \Pi$, where $\mathbf{T}^\pi\unitvec \defeq T(s,\pi(s))$. $(\mathbf{T}^\pi)^t \unitvec$ gives the probability distribution over the states visited at time step $t$, after following $\pi$ for $t$ steps from $s$.
\end{restatable}

\begin{restatable}[Properties of visit distribution functions]{prop}{visitDistProperties}\label{prop:visit-dist-prop}
Let $s,s'\in\St,\fpi{s} \in \F(s)$.
\begin{enumerate}
    \item $\fpi{s}(\gamma)$ is element-wise non-negative and element-wise monotonically increasing on $\gamma\in[0,1)$. \label{item:mono-increase}
    \item $\forall \gamma \in [0,1): \lone{\fpi{s}(\gamma)}=\geom$.\label{item:lone-visit}
\end{enumerate}
\end{restatable}
\begin{proof}
\Cref{item:mono-increase}: by examination of \cref{def:visit}, $\fpi{s} = \sum_{t=0}^\infty \prn{\gamma\mathbf{T}^\pi}^t\unitvec$. Since each $\prn{\mathbf{T}^\pi}^t$ is left stochastic and $\unitvec$ is the standard unit vector, each entry in each summand is non-negative. Therefore, $\forall \gamma \in [0,1): \fpi{s}(\gamma)^\top \unitvec[s']\geq 0$, and this function monotonically increases on $\gamma$.

\Cref{item:lone-visit}:
\begin{align}
    \lone{\fpi{s}(\gamma)}&=\lone{\sum_{t=0}^\infty \prn{\gamma\mathbf{T}^\pi}^t\unitvec}\\
    &= \sum_{t=0}^\infty\gamma^t\lone{ \prn{\mathbf{T}^\pi}^t\unitvec}\label{eq:non-negative-norm}\\
    &= \sum_{t=0}^\infty \gamma^t\label{eq:left-stoch}\\
    &= \geom.
\end{align}

\Cref{eq:non-negative-norm} follows because all entries in each $\prn{\mathbf{T}^\pi}^t\unitvec$ are non-negative by  \cref{item:mono-increase}. \Cref{eq:left-stoch} follows because each $\prn{\mathbf{T}^\pi}^t$ is left stochastic and $\unitvec$ is a stochastic vector, and so $\lone{\prn{\mathbf{T}^\pi}^t\unitvec}=1$.
\end{proof}

\begin{restatable}[$\f\in\F(s)$ is multivariate rational on $\gamma$]{lem}{fRat}\label{f-rat}
$\fpi{}\in\F(s)$ is a multivariate rational function on $\gamma\in[0,1)$.
\end{restatable}
\begin{proof}
Let $\rf\in\rewardVS$ and consider $\fpi{}\in\F(s)$. Let $\mathbf{v}^\pi_R$ be the $\OptVf{s,\gamma}$ function in column vector form, with one entry per state value.

By the Bellman equations, $\mathbf{v}^\pi_R = \prn{\mathbf{I}-\gamma\mathbf{T}^\pi}\inv\rf.$ Let $\mathbf{A}_\gamma \defeq \prn{\mathbf{I}-\gamma\mathbf{T}^\pi}\inv$, and for state $s$, form $ \mathbf{A}_{s,\gamma}$ by replacing  $\mathbf{A}_\gamma$'s column for state $s$ with $\rf$. As noted by \citet{lippman1968set}, by Cramer's rule, $V^\pi_R(s,\gamma)=\frac{\det{\mathbf{A}_{s,\gamma}}}{\det\mathbf{A}_\gamma}$ is a rational function with numerator and denominator having degree at most $\abs{\St}$.

In particular, for each state indicator reward function $\unitvec[s_i]$, $V^\pi_{s_i}(s,\gamma)=\fpi{s}(\gamma)^\top\unitvec[s_i]$ is a rational function of $\gamma$ whose numerator and denominator each have degree at most $\abs{\St}$. This implies that $\f^\pi(\gamma)$ is multivariate rational on $\gamma\in[0,1)$.
\end{proof}

\begin{restatable}[On-policy value is rational on $\gamma$]{cor}{smoothOnPol}\label{smoothOnPol}
Let $\pi\in\Pi$ and $R$ be any reward function. $V^\pi_R(s,\gamma)$ is rational on  $\gamma\in[0,1)$.
\end{restatable}
\begin{proof}
$V^\pi_R(s,\gamma)=\fpi{s}(\gamma)^\top \rf$, and $\f$ is a multivariate rational function of $\gamma$ by \cref{f-rat}. Therefore, for fixed $\rf$, $\fpi{s}(\gamma)^\top \rf$ is a rational function of $\gamma$.
\end{proof}

\subsection{Non-dominated visit distribution functions}\label{sec:nondom}

\begin{restatable}[Continuous reward function distribution]{definition}{dCont}
Results with $\Dcont$ hold for any absolutely continuous reward function distribution.
\end{restatable}

\begin{remark}
We assume $\rewardVS$ is endowed with the standard topology.
\end{remark}

\begin{restatable}[Distinct linear functionals disagree almost everywhere on their domains]{lem}{distinctLin}\label{lem:distinct-lin-prob}
Let $\x,\x'\in\rewardVS$ be distinct. $\optprob[\rf\sim\Dcont]{\x^\top \rf = \x'^\top\rf}=0$.
\end{restatable}
\begin{proof}
$\set{\rf \in\rewardVS\mid (\x-\x')^\top \rf=0}$ is a hyperplane since $\x-\x'\neq \mathbf{0}$. Therefore, it has no interior in the standard topology on $\rewardVS$. Since this empty-interior set is also convex, it has zero Lebesgue measure. By the Radon-Nikodym theorem, it has zero measure under any continuous distribution $\Dcont$.
\end{proof}

\begin{restatable}[Unique maximization of almost all vectors]{cor}{uniqueMax}\label{cor:distinct-maximized}
Let $X\subsetneq\rewardVS$ be finite.
\begin{equation}\optprob[\rf\sim\Dcont]{\abs{\argmax_{\x''\in X} \x''^\top \rf}>1}=0.
\end{equation}
\end{restatable}
\begin{proof}
Let $\x,\x'\in X$ be distinct. For any $\rf\in\rewardVS$, $\x,\x'\in \argmax_{\x''\in X} \x''^\top \rf$ iff $\x^\top \rf = \x'^\top\rf\geq \max_{\x''\in X\setminus \set{\x,\x'}} \x''^\top \rf$. By \cref{lem:distinct-lin-prob}, $\x^\top \rf = \x'^\top\rf$ holds with probability $0$ under any $\Dcont$.
\end{proof}

\subsubsection{Generalized non-domination results}
Our formalism includes both $\Fnd(s)$ and $\RSDnd$; we therefore prove results that are applicable to both.

\begin{restatable}[Non-dominated linear functionals]{definition}{ndLinFunc}\label{def:nd-lin-func}
Let $X\subsetneq \rewardVS$ be finite. $\ND{X}\defeq \set{\x\in X\mid \exists \rf \in \rewardVS: \x^\top \rf > \max_{\x'\in X\setminus \set{\x}} \x'^\top \rf}$.
\end{restatable}

\begin{restatable}[All vectors are maximized by a non-dominated linear functional]{lem}{allRFMaxND}\label{lem:all-rf-max-nd}
Let $\rf \in \rewardVS$ and let $X\subsetneq \rewardVS$ be finite and non-empty. $\exists \x^* \in \ND{X}: \x^{*\top}\rf=\max_{\x \in X} \x^\top \rf$.
\end{restatable}
\begin{proof}
Let $A(\rf\mid X)\defeq \argmax_{\x\in X} \x^\top \rf=\set{\x_1,\ldots, \x_n}$. Then
\begin{align}
    \x_1^\top \rf = \cdots= \x_n^\top \rf> \max_{\x' \in X \setminus A(\rf\mid X)} \x'^\top \rf.\label{eq:expr-cont-nd-func}
\end{align}

In \cref{eq:expr-cont-nd-func}, each $\x^\top \rf$ expression is linear on $\rf$. The $\max$ is piecewise linear on $\rf$ since it is the maximum of a finite set of linear functionals. In particular, all expressions in \cref{eq:expr-cont-nd-func} are continuous on $\rf$, and so we can find some $\delta>0$ neighborhood $B(\rf,\delta)$ such that $\forall \rf' \in B(\rf,\delta): \max_{\x_i\in A(\rf\mid X)} \x_i^\top \rf' > \max_{\x'\in X\setminus A(\rf\mid X)} \x'^\top \rf'$.

But almost all $\rf' \in B(\rf,\delta)$ are maximized by a unique functional $\x^*$ by \cref{cor:distinct-maximized}; in particular, at least one such $\rf''$ exists. Formally, $\exists \rf''\in B(\rf,\delta):\x^{*\top} \rf''>\max_{\x' \in X\setminus \set{\x^*}} \x'^\top \rf''$. Therefore, $\x^*\in \ND{X}$ by \cref{def:nd-lin-func}.

$\x^{*\top} \rf'\geq\max_{\x_i\in A(\rf\mid X)} \x_i^\top \rf' > \max_{\x'\in X\setminus A(\rf\mid X)} \x'^\top \rf'$, with the strict inequality following because  $\rf'' \in B(\rf,\delta)$. These inequalities imply that $\x^*\in A(\rf\mid X)$.
\end{proof}

\begin{restatable}[Maximal value is invariant to restriction to non-dominated functionals]{cor}{ndFuncIndif}\label{cor:nd-func-indif}
Let $\rf \in \rewardVS$ and let $X\subsetneq \rewardVS$ be finite. $\max_{\x\in X} \x^\top \rf=\max_{\x\in \ND{X}} \x^\top \rf$.
\end{restatable}
\begin{proof}
If $X$ is empty, holds trivially. Otherwise, apply \cref{lem:all-rf-max-nd}.
\end{proof}

\begin{restatable}[How non-domination containment affects optimal value]{lem}{ndOptContain}\label{lem:nd-opt-contain}
Let $\rf \in \rewardVS$ and let $X,X'\subsetneq \rewardVS$ be finite.
\begin{enumerate}
    \item If $\ND{X}\subseteq X'$, then $\max_{\x\in X} \x^\top \rf\leq\max_{\x'\in X'} \x'^\top \rf$.\label{item:ND-contain-max}
    \item If $\ND{X}\subseteq X'\subseteq X$, then $\max_{\x\in X} \x^\top \rf=\max_{\x'\in X'} \x'^\top \rf$.\label{item:ND-contain-max-2}
\end{enumerate}
\end{restatable}
\begin{proof}
\Cref{item:ND-contain-max}:
\begin{align}
\max_{\x\in X} \x^\top \rf&=\max_{\x\in \ND{X}} \x^\top \rf\label{eq:item-max}\\
&\leq \max_{\x'\in X'} \x'^\top \rf.\label{eq:item-max-2}
\end{align}
\Cref{eq:item-max} follows by \cref{cor:nd-func-indif}. \Cref{eq:item-max-2} follows because $\ND{X}\subseteq X'$.

\Cref{item:ND-contain-max-2}: by \cref{item:ND-contain-max}, $\max_{\x\in X} \x^\top \rf\leq\max_{\x'\in X'} \x'^\top \rf$. Since $X'\subseteq X$, we also have $\max_{\x\in X} \x^\top \rf\geq\max_{\x'\in X'} \x'^\top \rf$, and so equality must hold.
\end{proof}

\begin{restatable}[Non-dominated vector functions]{definition}{ndVecFun}\label{def:nd-vec-func}
Let $I\subseteq \reals$ and let $F\subsetneq \prn{\rewardVS}^I$ be a finite set of vector-valued functions on $I$. 
\begin{equation}\ND{F}\defeq \set{\f\in F\mid \exists \gamma\in I, \rf \in \rewardVS: \f(\gamma)^\top \rf> \max_{\f'\in F\setminus \set{\f}}\f'(\gamma)^\top \rf}.
\end{equation}
\end{restatable}

\begin{remark}
$\Fnd(s)=\ND{\F(s)}$ by \cref{def:nd}.
\end{remark}

\begin{restatable}[Affine transformation of visit distribution sets]{definition}{affTransfShorthand}\label{def:aff-transf-short}
For notational convenience, we define set-scalar multiplication and set-vector addition on $X\subseteq \rewardVS$: for $c\in\reals$, $cX\defeq \set{c\x\mid\x\in X}$. For $\av \in \rewardVS$, $X+\av\defeq \set{\x+\av\mid \x \in X}$. Similar operations hold when $X$ is a set of vector functions $\reals\mapsto \rewardVS$.
\end{restatable}

\begin{restatable}[Invariance of non-domination under positive affine transform]{lem}{posAffNDInvar}\label{lem:pos-aff-nd-invar}\hfill
\begin{enumerate}
    \item Let $X\subsetneq \rewardVS$ be finite. If $\x\in \ND{X}$, then $\forall c>0, \av\in\rewardVS: (c\x+\av) \in \ND{cX+\av}$.\label{item:invar-vectors}
    \item Let $I\subseteq \reals$ and let $F\subsetneq \prn{\rewardVS}^I$ be a finite set of vector-valued functions on $I$. If $\f \in \ND{F}$, then $\forall c>0, \av\in\rewardVS: (c\f+\av) \in \ND{cF+\av}$.\label{item:invar-vector-fns}
\end{enumerate}
\end{restatable}
\begin{proof}
\Cref{item:invar-vectors}: Suppose $\x\in \ND{X}$ is strictly optimal for $\rf\in\rewardVS$. Then let $c>0,\av\in\rewardVS$ be arbitrary, and define $b\defeq \av^\top \rf$.
\begin{align}
    \x^\top\rf &> \max_{\x'\in X\setminus\set{\x}} \x'^\top \rf\\
    c\x^\top\rf+b &> \max_{\x'\in X\setminus\set{\x}} c\x'^\top \rf+b\label{eq:rescale-c}\\
    (c\x+\av)^\top\rf &> \max_{\x'\in X\setminus\set{\x}} (c\x'+\av)^\top \rf\label{eq:b-defn}\\
    (c\x+\av)^\top\rf &> \max_{\x''\in \prn{cX+\av}\setminus\set{c\x+\av}} \x''^\top \rf.
\end{align}
\Cref{eq:rescale-c} follows because $c>0$. \Cref{eq:b-defn} follows by the definition of $b$.

\Cref{item:invar-vector-fns}: If $\f\in\ND{F}$, then by \cref{def:nd-vec-func}, there exist $\gamma\in I, \rf\in\rewardVS$ such that
\begin{align}
    \f(\gamma)^\top \rf> \max_{\f'\in F\setminus \set{\f}}\f'(\gamma)^\top \rf.
\end{align}

Apply \cref{item:invar-vectors} to conclude
\begin{align}
    (c\f(\gamma)+\av)^\top \rf> \max_{(c\f'+\av)\in (cF+\av)\setminus \set{c\f+\av}}(c\f'(\gamma)+\av)^\top \rf.
\end{align}

Therefore, $(c\f+\av) \in \ND{cF+\av}$.
\end{proof}

\subsubsection{Inequalities which hold under most reward function distributions}

\ineqMost*

\begin{restatable}[Helper lemma for demonstrating $\geqMost$]{lem}{helperGeqMost}\label{lem:helper-geq-most}
Let $\distSet\subseteq \Delta(\rewardVS)$. If $\exists \phi\in\mdpPermGroup$ such that for all $\D\in\distSet$, $f_1\prn{\D}< f_2\prn{\D}$ implies that $f_1\prn{\phi\cdot \D}> f_2\prn{\phi\cdot \D}$, then $f_1(\D) \geqMost[][\distSet] f_2(\D)$.
\end{restatable}
\begin{proof}
Since $\phi$ does not belong to the stabilizer of $\mdpPermGroup$, $\phi$ acts injectively on $\orbi[\D]$. By assumption on $\phi$, the image of $\{\D'\in\orbi[\D]\mid f_1(\D')<f_2(\D')\}$ under $\phi$ is a subset of $\{\D'\in\orbi[\D]\mid f_1(\D')>f_2(\D')\}$. Since $\phi$ is injective, $\abs{\{\D'\in\orbi[\D]\mid f_1(\D')<f_2(\D')\}}\leq \abs{\{\D'\in\orbi[\D]\mid f_1(\D')>f_2(\D')\}}$. $f_1(\D) \geqMost[][\distSet] f_2(\D)$ by \cref{def:ineq-most-dists}.
\end{proof}

\begin{restatable}[A helper result for expectations of functions]{lem}{helperPerm}\label{lem:helper-perm}
Let $B_1,\ldots,B_n\subsetneq \rewardVS$ be finite and let $\distSet \subseteq \Delta(\rewardVS)$. Suppose $f$ is a function of the form
\begin{align}
    f\prn{B_1,\ldots,B_n \mid \D}=\E{\rf\sim \D}{g\prn{\max_{\bv_1\in B_1} \bv_1^\top \rf,\ldots, \max_{\bv_n\in B_n} \bv_n^\top \rf}}\label{eq:helper}
\end{align}
for some function $g$, and that $f$ is well-defined for all $\D \in \distSet$. Let $\phi$ be a state permutation. Then
\begin{equation}
    f\prn{B_1,\ldots,B_n \mid \D}=f\prn{\phi\cdot B_1 ,\ldots,\phi \cdot B_n \mid \phi \cdot\D}.
\end{equation}
\end{restatable}
\begin{proof} Let distribution $\D$ have probability measure $F$, and let $\phi\cdot \D$ have probability measure $F_\phi$.
\begin{align}
    &f\prn{B_1,\ldots,B_n \mid \D}\\
    \defeq{}&\E{\rf\sim \D}{g\prn{\max_{\bv_1\in B_1} \bv_1^\top \rf,\ldots, \max_{\bv_n\in B_n} \bv_n^\top \rf}}\\
    \defeq{}&\int_{\rewardVS} g\prn{\max_{\bv_1\in B_1} \bv_1^\top \rf,\ldots, \max_{\bv_n\in B_n} \bv_n^\top \rf} \dF[\rf][F]\\
    ={}&\int_{\rewardVS} g\prn{\max_{\bv_1\in B_1} \bv_1^\top \rf,\ldots, \max_{\bv_n\in B_n} \bv_n^\top \rf} \dF[\permute\rf][F_\phi]\label{eq:permute-prob-meas}\\
    ={}&\int_{\rewardVS} g\prn{\max_{\bv_1\in B_1} \bv_1^\top \prn{\permute\inv\rf'},\ldots, \max_{\bv_n\in B_n} \bv_n^\top \prn{\permute\inv\rf'}} \abs{\det \permute}\dF[\rf'][F_\phi]\label{eq:change-of-variables}\\
    ={}&\int_{\rewardVS} g\prn{\max_{\bv_1\in B_1} \prn{\permute\bv_1}^\top \rf',\ldots, \max_{\bv_n\in B_n} \prn{\permute\bv_n}^\top \rf'} \dF[\rf'][F_\phi]\label{eq:change-of-variables-2}\\
    ={}&\int_{\rewardVS} g\prn{\max_{\bv_1'\in \phi\cdot B_1} \bv_1'^{\top} \rf',\ldots, \max_{\bv_n'\in \phi\cdot B_n} \bv_n'^{\top} \rf'} \dF[\rf'][F_\phi]\\
    \eqdef{}& f\prn{\phi \cdot B_1,\ldots,\phi\cdot B_n \mid \phi\cdot \D}.
\end{align}

\Cref{eq:permute-prob-meas} follows by the definition of $F_\phi$ (\cref{def:pushforward-permute}). \Cref{eq:change-of-variables} follows by substituting $\rf'\defeq \permute\rf$. \Cref{eq:change-of-variables-2} follows from the fact that all permutation matrices have unitary determinant and are orthogonal (and so $(\permute\inv)^\top=\permute$).
\end{proof}

\begin{restatable}[Support of $\Dany$]{definition}{DefSupportDist}\label{def:support-dist}
Let $\Dany$ be any reward function distribution. $\supp[\Dany]$ is the smallest closed subset of $\rewardVS$ whose complement has measure zero under $\Dany$.
\end{restatable}

\begin{restatable}[Linear functional optimality probability]{definition}{linFNProb}
For finite $A,B\subsetneq \rewardVS$, the \emph{probability under $\Dany$ that $A$ is optimal over $B$} is
\begin{equation}\phelper{A\geq B}[\Dany]\defeq \optprob[\rf \sim \Dany]{\max_{\av\in A} \av^\top \rf \geq \max_{\bv\in B} \bv^\top \rf}.\end{equation}
\end{restatable}

\begin{restatable}[Non-dominated linear functionals and their optimality probability]{prop}{helperPositiveProb}\label{prop:helper-positive-prob}
Let $A\subsetneq \rewardVS$ be finite. If $\exists b<c:[b,c]^{\abs{\St}} \subseteq \supp[\Dany]$, then $\av\in\ND{A}$ implies that $\av$ is strictly optimal for a set of reward functions with positive measure under $\Dany$.
\end{restatable}
\begin{proof}
Suppose $\exists b<c:[b,c]^{\abs{\St}} \subseteq \supp[\Dany]$. If $\av\in\ND{A}$, then let $\rf$ be such that $\av^\top \rf > \max_{\av'\in A\setminus\set{\av}} \av'^\top \rf$. For $a_1>0, a_2\in \reals$, positively affinely transform $\rf'\defeq a_1\rf + a_2\mathbf{1}$ (where $\mathbf{1}\in\rewardVS$ is the all-ones vector) so that $\rf' \in (b,c)^{\abs{\St}}$.

Note that $\av$ is still strictly optimal for $\rf'$:
\begin{equation}
    \av^\top \rf > \max_{\av'\in A\setminus\set{\av}} \av'^\top \rf \iff \av^\top \rf' > \max_{\av'\in A\setminus\set{\av}} \av'^\top \rf'.\label{eq:cont-lin-ineq}
\end{equation}

Furthermore, by the continuity of both terms on the right-hand side of \cref{eq:cont-lin-ineq}, $\av$ is strictly optimal for reward functions in some open neighborhood $N$ of $\rf'$. Let $N'\defeq N\cap (b,c)^{\abs{\St}}$. $N'$ is still open in $\rewardVS$ since it is the intersection of two open sets $N$ and $(b,c)^{\abs{\St}}$.

$\Dany$ must assign positive probability measure to all open sets in its support; otherwise, its support would exclude these zero-measure sets by \cref{def:support-dist}. Therefore, $\Dany$ assigns positive probability to $N'\subseteq \supp[\Dany]$.
\end{proof}

\begin{restatable}[Expected value of similar linear functional sets]{lem}{ndFuncIncr}\label{lem:nd-func-incr-pwr}
Let $A,B\subsetneq \rewardVS$ be finite, let $A'$ be such that $\ND{A}\subseteq A'\subseteq A$, and let $g:\reals\to\reals$ be an increasing function. If $B$ contains a copy $B'$ of $A'$ via $\phi$, then
\begin{equation}
    \E{\rf \sim \Dbd}{g\prn{\max_{\av\in A}\av^\top \rf}} \leq \E{\rf \sim \phi\cdot \Dbd}{g\prn{\max_{\bv\in B}\bv^\top \rf}}.\label{eq:nd-func-incr}
\end{equation}

If $\ND{B}\setminus B'$ is empty, then \cref{eq:nd-func-incr} is an equality. If $\ND{B}\setminus B'$ is non-empty, $g$ is strictly increasing, and $\exists b<c: (b,c)^{\abs{\St}}\subseteq\supp[\Dbd]$, then \cref{eq:nd-func-incr} is strict.
\end{restatable}
\begin{proof}
Because $g:\reals\to\reals$ is increasing, it is measurable (as is $\max$). Therefore, the relevant expectations exist for all $\Dbd$.
\begin{align}
     \E{\rf \sim \Dbd}{g\prn{\max_{\av\in A}\av^\top \rf}} &=\E{\rf \sim \Dbd}{g\prn{\max_{\av\in A'}\av^\top \rf}}\label{eq:nd-func-incr-restr}\\
     &=\E{\rf \sim \phi\cdot \Dbd}{g\prn{\max_{\av\in \phi\cdot A'}\av^\top \rf}}\label{eq:apply-helper}\\
     &=\E{\rf \sim \phi\cdot \Dbd}{g\prn{\max_{\bv\in B'}\bv^\top \rf}}\label{eq:sim-helper}\\
     &\leq\E{\rf \sim \phi\cdot \Dbd}{g\prn{\max_{\bv\in B}\bv^\top \rf}}.\label{eq:sim-inequality}
\end{align}
\Cref{eq:nd-func-incr-restr} holds because $\forall \rf\in\rewardVS:\max_{\av\in A}\av^\top \rf=\max_{\av\in A'}\av^\top \rf$ by \cref{lem:nd-opt-contain}'s \cref{item:ND-contain-max-2} with $X\defeq A$, $X'\defeq A'$. \Cref{eq:apply-helper} holds by \cref{lem:helper-perm}. \Cref{eq:sim-helper} holds by the definition of $B'$. Furthermore, our assumption on $\phi$ guarantees that $B'\subseteq B$. Therefore, $\max_{\bv\in B'}\bv^\top \rf\leq \max_{\bv\in B}\bv^\top \rf$, and so \cref{eq:sim-inequality} holds by the fact that $g$ is an increasing function. Then \cref{eq:nd-func-incr} holds.

If $\ND{B}\setminus B'$ is empty, then $\ND{B}\subseteq B'$. By assumption, $B'\subseteq B$. Then apply \cref{lem:nd-opt-contain} \cref{item:ND-contain-max-2} with $X\defeq B$, $X'\defeq B'$ in order to conclude that \cref{eq:sim-inequality} is an equality. Then \cref{eq:nd-func-incr} is also an equality.

Suppose that $g$ is strictly increasing, $\ND{B}\setminus B'$ is non-empty, and $\exists b<c: (b,c)^{\abs{\St}}\subseteq\supp[\Dbd]$. Let $\x \in \ND{B}\setminus B'$.
\begin{align}
     \E{\rf \sim \phi\cdot \Dbd}{g\prn{\max_{\bv\in B'}\bv^\top \rf}}&< \E{\rf \sim \phi\cdot \Dbd}{g\prn{\max_{\av\in B'\cup\set{\x}}\bv^\top \rf}}\label{eq:sim-helper-2}\\
     &\leq\E{\rf \sim \phi\cdot \Dbd}{g\prn{\max_{\bv\in B}\bv^\top \rf}}.\label{eq:sim-inequality-2}
\end{align}

$\x$ is strictly optimal for a positive-probability subset of $\supp[\Dbd]$ by \cref{prop:helper-positive-prob}. Since  $g$ is strictly increasing, \cref{eq:sim-helper-2} is strict. Therefore, we conclude that \cref{eq:nd-func-incr} is strict.
\end{proof}

\begin{restatable}[Continuous $\Diid$ have nonempty interior]{lem}{IIDContains}\label{lem:iid-contains} For continuous\ {\iid} distributions $\Diid$, $\exists b<c: (b,c)^{\abs{\St}}\subseteq \optSupp(\Diid)$.
\end{restatable}
\begin{proof}
$\Diid\defeq \Dist^{\abs{\St}}$. Since the state reward distribution $\Dist$ is continuous, $\Dist$ must have support on some open interval $(b,c)$. Since $\Diid$ is {\iid} across states, $(b,c)^{\abs{\St}}\subseteq \supp[\Diid]$.
\end{proof}

\begin{restatable}[Bounded, continuous {\iid} reward]{definition}{BCIIDDef}
$\DSetBCiid$ is the set of $\Diid$ which equal $\Dist^{\abs{\St}}$ for some continuous, bounded-support distribution $\Dist$ over $\reals$.
\end{restatable}

\begin{restatable}[Expectation superiority lemma]{lem}{expectSuperior}\label{lem:expect-superior}
Let $A,B\subsetneq \rewardVS$ be finite and let $g:\reals\to \reals$ be an increasing function. If $B$ contains a copy $B'$ of $\ND{A}$ via $\phi$, then
\begin{align}
    \E{\rf \sim \Dbd}{g\prn{\max_{\av\in A}\av^\top \rf}}\leqMost[][\DSetBd]  \E{\rf \sim \Dbd}{g\prn{\max_{\bv\in B}\bv^\top \rf}}.\label{eq:permute-superior}
\end{align}

Furthermore, if $g$ is strictly increasing and $\ND{B}\setminus \phi\cdot \ND{A}$ is non-empty, then \cref{eq:permute-superior} is strict for all $\Diid\in\DSetBCiid$. In particular, \[\E{\rf \sim \Dbd}{g\prn{\max_{\av\in A}\av^\top \rf}}\not\geqMost[][\DSetBd]  \E{\rf \sim \Dbd}{g\prn{\max_{\bv\in B}\bv^\top \rf}}.\]
\end{restatable}
\begin{proof}
Because $g:\reals\to\reals$ is increasing, it is measurable (as is $\max$). Therefore, the relevant expectations exist for all $\Dbd$.

Suppose that $\Dbd$ is such that $\E{\rf \sim \Dbd}{g\prn{\max_{\bv\in B}\bv^\top \rf}}< \E{\rf \sim \Dbd}{g\prn{\max_{\av\in A}\av^\top \rf}}$.
\begin{align}
    \E{\rf \sim \phi\cdot \Dbd}{g\prn{\max_{\av\in A}\av^\top \rf}}&\leq\E{\rf \sim \phi^2\cdot \Dbd}{g\prn{\max_{\bv\in B}\bv^\top \rf}}\label{eq:first-leq}\\
    &=\E{\rf \sim \Dbd}{g\prn{\max_{\bv\in B}\bv^\top \rf}}\label{eq:involve}\\
    &<\E{\rf \sim \Dbd}{g\prn{\max_{\av\in A}\av^\top \rf}}\label{eq:superior-leq}\\
    &\leq  \E{\rf \sim \phi\cdot \Dbd}{g\prn{\max_{\bv\in B}\bv^\top \rf}}.\label{eq:second-leq}
\end{align}
\Cref{eq:first-leq} follows by applying \cref{lem:nd-func-incr-pwr} with permutation $\phi$ and $A'\defeq \ND{A}$. \Cref{eq:involve} follows because involutions satisfy $\phi\inv=\phi$, and $\phi^2$ is therefore the identity. \Cref{eq:superior-leq} follows because we assumed that \[\E{\rf \sim \Dbd}{g\prn{\max_{\bv\in B}\bv^\top \rf}}< \E{\rf \sim \Dbd}{g\prn{\max_{\av\in A}\av^\top \rf}}.\] \Cref{eq:second-leq} follows by applying \cref{lem:nd-func-incr-pwr} with permutation $\phi$ and and $A'\defeq \ND{A}$. By \cref{lem:helper-geq-most}, \cref{eq:permute-superior} holds.

Suppose $g$ is strictly increasing and $\ND{B}\setminus B'$ is non-empty. Let $\phi'\in\mdpPermGroup$.
\begin{align}
\E{\rf \sim \phi'\cdot \Diid}{g\prn{\max_{\av\in A}\av^\top \rf}}&=\E{\rf \sim \Diid}{g\prn{\max_{\av\in A}\av^\top \rf}}\label{eq:same-iid-dist-exp}\\
&<\E{\rf \sim \phi\cdot \Diid}{g\prn{\max_{\bv\in B}\bv^\top \rf}}\label{eq:sim-iid-exp}\\
&=\E{\rf \sim \phi'\cdot \Diid}{g\prn{\max_{\bv\in B}\bv^\top \rf}}.\label{eq:same-iid-dist-exp-2}
\end{align}

\Cref{eq:same-iid-dist-exp} and \cref{eq:same-iid-dist-exp-2} hold because $\Diid$ distributes reward identically across states: $\forall \phi_x\in\mdpPermGroup: \phi_x\cdot \Diid=\Diid$. By \cref{lem:iid-contains}, $\exists b<c: (b,c)^{\abs{\St}}\subseteq \supp[\Diid]$. Therefore, apply \cref{lem:nd-func-incr-pwr} with $A'\defeq \ND{A}$ to conclude that \cref{eq:sim-iid-exp} holds.

Therefore, $\forall \phi' \in \mdpPermGroup: \E{\rf \sim \phi'\cdot \Diid}{g\prn{\max_{\av\in A}\av^\top \rf}}<\E{\rf \sim \phi'\cdot \Diid}{g\prn{\max_{\bv\in B}\bv^\top \rf}}$, and so $\E{\rf \sim \Dbd}{g\prn{\max_{\av\in A}\av^\top \rf}}\not\geqMost[][\DSetBd]  \E{\rf \sim \Dbd}{g\prn{\max_{\bv\in B}\bv^\top \rf}}$ by \cref{def:ineq-most-dists}.
\end{proof}

\begin{restatable}[Indicator function]{definition}{indicDef}
Let $L$ be a predicate which takes input $x$. $\indic{L(x)}$ is the function which returns $ 1$ when $L(x)$ is true, and $0$ otherwise.
\end{restatable}

\begin{restatable}[Optimality probability inclusion relations]{lem}{optprbInclusion}\label{lem:inclusion-opt}
Let $X,Y\subsetneq \rewardVS$ be finite and suppose $Y'\subseteq Y$. \begin{equation}
    \phelper{X\geq Y}[\Dany]\leq\phelper{X\geq Y'}[\Dany]\leq \phelper{X\cup\prn{Y\setminus Y'}\geq Y}[\Dany].\label{eq:inclusion-opt}
\end{equation}

If $\exists b<c: (b,c)^{\abs{\St}}\subseteq \supp[\Dany]$, $X\subseteq Y$, and $\ND{Y}\cap \prn{Y\setminus Y'}$ is non-empty, then the second inequality is strict.
\end{restatable}
\begin{proof}
\begin{align}
    \phelper{X\geq Y}[\Dany]&\defeq \E{\rf \sim \Dany}{\indic{\max_{\x\in X} \x^\top \rf \geq \max_{\mathbf{y} \in Y} \mathbf{y}^\top \rf}}\\
    &\leq \E{\rf \sim \Dany}{\indic{\max_{\x\in X} \x^\top \rf \geq \max_{\mathbf{y} \in Y'} \mathbf{y}^\top \rf}}\label{eq:leq-Y-contain}\\
    &\leq \E{\rf \sim \Dany}{\indic{\max_{\x\in X\cup (Y\setminus Y')} \x^\top \rf \geq \max_{\mathbf{y} \in Y'} \mathbf{y}^\top \rf}}\label{eq:leq-Y-union}\\
    &= \E{\rf \sim \Dany}{\indic{\max_{\x\in X\cup (Y\setminus Y')} \x^\top \rf \geq \max_{\mathbf{y} \in Y'\cup (Y\setminus Y')} \mathbf{y}^\top \rf}}\label{eq:leq-Y-union-2}\\
    &=\E{\rf \sim \Dany}{\indic{\max_{\x\in X\cup (Y\setminus Y')} \x^\top \rf \geq \max_{\mathbf{y} \in Y} \mathbf{y}^\top \rf}}\\
    &\eqdef \phelper{X\cup\prn{Y\setminus Y'}\geq Y}[\Dany].
\end{align}
\Cref{eq:leq-Y-contain} follows because $\forall \rf \in \rewardVS: \indic{\max_{\x\in X} \x^\top \rf \geq \max_{\mathbf{y} \in Y} \mathbf{y}^\top \rf}\leq \indic{\max_{\x\in X} \x^\top \rf \geq \max_{\mathbf{y} \in Y'} \mathbf{y}^\top \rf}$ since $Y' \subseteq Y$; note that \cref{eq:leq-Y-contain} equals $\phelper{X\geq Y'}[\Dany]$, and so the first inequality of \cref{eq:inclusion-opt} is shown. \Cref{eq:leq-Y-union} holds because \[\forall \rf \in \rewardVS: \indic{\max_{\x\in X} \x^\top \rf \geq \max_{\mathbf{y} \in Y'} \mathbf{y}^\top \rf}\leq \indic{\max_{\x\in X\cup (Y\setminus Y')} \x^\top \rf \geq \max_{\mathbf{y} \in Y'} \bv^\top \rf}.\]

Suppose $\exists b<c: (b,c)^{\abs{\St}}\subseteq \supp[\Dany]$, $X\subseteq Y$, and $\ND{Y}\cap \prn{Y\setminus Y'}$ is non-empty. Let $\mathbf{y}^*\in \ND{Y}\cap \prn{Y\setminus Y'}$. By \cref{prop:helper-positive-prob}, $\mathbf{y}^*$ is strictly optimal on a subset of $\supp[\Dany]$ with positive measure under $\Dany$. In particular, for a set of $\rf^*$ with positive measure under $\Dany$, we have $\mathbf{y}^{*\top}\rf^*> \max_{\mathbf{y}\in Y'} \mathbf{y}^\top \rf^*.$

Then \cref{eq:leq-Y-union} is strict, and therefore the second inequality of \cref{eq:inclusion-opt} is strict as well.
\end{proof}

\begin{restatable}[Optimality probability of similar linear functional sets]{lem}{optProbSim}\label{lem:sim-lin-func-opt}
Let $A,B,C\subsetneq \rewardVS$ be finite, and let $Z\subseteq \rewardVS$ be such that $\ND{C}\subseteq Z\subseteq C$. If $\ND{A}$ is similar to $B'\subseteq B$ via $\phi$ such that $\phi\cdot \prn{Z\setminus \prn{B\setminus B'}}=Z\setminus \prn{B\setminus B'}$, then
\begin{equation}
    \phelper{A\geq C}[\Dany]\leq \phelper{B\geq C}[\phi\cdot \Dany].\label{eq:sim-nd-func}
\end{equation}
If $B'=B$, then \cref{eq:sim-nd-func} is an equality. If $\exists b<c: (b,c)^{\abs{\St}}\subseteq \supp[\Dany]$, $B'\subseteq C$, and $\ND{C}\cap \prn{B\setminus B'}$ is non-empty, then \cref{eq:sim-nd-func} is strict.
\end{restatable}
\begin{proof}
\begin{align}
    \phelper{A\geq C}[\Dany]&=\phelper{A\geq Z}[\Dany]\label{eq:nd-restrict-opt-lin}\\
    &=\phelper{\ND{A}\geq Z}[\Dany]\label{eq:nd-restrict-opt-lin-2}\\
    &\leq \phelper{\ND{A}\geq Z\setminus\prn{B\setminus B'}}[\Dany]\label{eq:fewer-requirements}\\
    &= \phelper{\phi\cdot \ND{A}\geq \phi\cdot Z\setminus\prn{B\setminus B'}}[\phi\cdot \Dany]\label{eq:phi-helper-nd-opt}\\
    &= \phelper{B'\geq Z\setminus\prn{B\setminus B'}}[\phi\cdot \Dany]\label{eq:phi-assumptions}\\
    &\leq \phelper{B'\cup \prn{B\setminus B'}\geq Z}[\phi\cdot \Dany]\label{eq:reunion}\\
    &= \phelper{B\geq C}[\phi\cdot \Dany].\label{eq:final-phelper-leq}
\end{align}

\Cref{eq:nd-restrict-opt-lin} and \cref{eq:final-phelper-leq} follow by \cref{lem:nd-opt-contain}'s \cref{item:ND-contain-max-2} with $X\defeq C$, $X'\defeq Z$. Similarly, \cref{eq:nd-restrict-opt-lin-2} follows by \cref{lem:nd-opt-contain}'s \cref{item:ND-contain-max-2} with $X\defeq A$, $X'\defeq \ND{A}$. \Cref{eq:fewer-requirements} follows by applying the first inequality of \cref{lem:inclusion-opt} with $X\defeq \ND{A}, Y\defeq Z, Y'\defeq Z\setminus (B\setminus B')$. \Cref{eq:phi-helper-nd-opt} follows by applying \cref{lem:helper-perm} to \cref{eq:nd-restrict-opt-lin} with permutation $\phi$.

\Cref{eq:phi-assumptions} follows by our assumptions on $\phi$. \Cref{eq:reunion} follows because by applying the second inequality of \cref{lem:inclusion-opt} with $X\defeq B', Y\defeq \ND{C}, Y'\defeq \ND{C}\setminus (B\setminus B')$.

Suppose $B'=B$. Then $B\setminus B'=\emptyset$, and so \cref{eq:fewer-requirements} and \cref{eq:reunion} are trivially equalities. Then \cref{eq:sim-nd-func} is an equality.

Suppose $\exists b<c: (b,c)^{\abs{\St}}\subseteq \supp[\Dany]$; note that $(b,c)^{\abs{\St}}\subseteq\supp[\phi\cdot \Dany]$, since such support must be invariant to permutation. Further suppose that $B'\subseteq C$ and that $\ND{C}\cap \prn{B\setminus B'}$ is non-empty. Then letting $X\defeq B', Y\defeq Z, Y'\defeq Z\setminus (B\setminus B')$ and noting that $\ND{\ND{Z}}=\ND{Z}$, apply \cref{lem:inclusion-opt} to \cref{eq:reunion} to conclude that \cref{eq:sim-nd-func} is strict.
\end{proof}

\begin{restatable}[Optimality probability superiority lemma]{lem}{optProbSup}\label{lem:opt-prob-superior}
Let $A,B,C\subsetneq \rewardVS$ be finite, and let $Z$ satisfy $\ND{C}\subseteq Z \subseteq C$. If $B$ contains a copy $B'$ of $\ND{A}$ via $\phi$ such that $\phi\cdot\prn{Z\setminus \prn{B\setminus B'}}=Z\setminus \prn{B\setminus B'}$, then $\phelper{A\geq C}[\Dany] \leqMost[][\DSetAny] \phelper{B \geq C}[\Dany]$.

If $B'\subseteq C$ and $\ND{C}\cap \prn{B\setminus B'}$ is non-empty, then the inequality is strict for all $\Diid\in\DSetBCiid$ and  $\phelper{A\geq C}[\Dany] \not\geqMost[][\DSetAny] \phelper{B \geq C}[\Dany]$.
\end{restatable}
\begin{proof}
Suppose $\Dany$ is such that $\phelper{B \geq C}[\Dany] < \phelper{A\geq C}[\Dany]$.
\begin{align}
    \phelper{A\geq C}[\phi\cdot \Dany]&=\phelper{A\geq C}[\phi\inv\cdot \Dany]\label{eq:involution-1}\\
    &\leq \phelper{B\geq C}[\Dany]\label{eq:sim-apply-opt}\\
    &< \phelper{A\geq C}[\Dany]\label{eq:assumption-apply-leq}\\
    &\leq \phelper{B\geq C}[\phi\cdot \Dany].\label{eq:sim-apply-opt-2}
\end{align}

\Cref{eq:involution-1} holds because $\phi$ is an involution. \Cref{eq:sim-apply-opt} and \cref{eq:sim-apply-opt-2} hold by applying \cref{lem:sim-lin-func-opt} with permutation $\phi$. \Cref{eq:assumption-apply-leq} holds by assumption. Therefore, $\phelper{A\geq C}[\Dany] \leqMost[][\DSetAny] \phelper{B \geq C}[\Dany]$ by \cref{lem:helper-geq-most}.

Suppose $B'\subseteq C$ and $\ND{C}\cap \prn{B\setminus B'}$ is non-empty, and let $\Diid$ be any continuous distribution which distributes reward independently and identically across states. Let $\phi'\in\mdpPermGroup$.
\begin{align}
\phelper{A\geq C}[\phi'\cdot \Diid]&=\phelper{A\geq C}[\Diid]\label{eq:same-iid-dist-prob}\\
&<\phelper{B\geq C}[\phi\cdot \Diid]\label{eq:sim-iid-prob}\\
&=\phelper{A\geq C}[\phi'\cdot \Diid].\label{eq:same-iid-dist-prob-2}
\end{align}

\Cref{eq:same-iid-dist-prob} and \cref{eq:same-iid-dist-prob-2} hold because $\Diid$ distributes reward identically across states, $\forall \phi_x\in\mdpPermGroup: \phi_x\cdot \Diid=\Diid$. By \cref{lem:iid-contains}, $\exists b<c: (b,c)^{\abs{\St}}\subseteq \supp[\Diid]$. Therefore, apply \cref{lem:sim-lin-func-opt} to conclude that \cref{eq:sim-iid-prob} holds. Therefore, $\forall \phi'\in\mdpPermGroup:\phelper{A\geq C}[\phi'\cdot \Diid]<\phelper{B\geq C}[\phi'\cdot \Diid]$. In particular, $\phelper{A\geq C}[\Dany] \not\geqMost[][\DSetAny] \phelper{B \geq C}[\Dany]$ by \cref{def:ineq-most-dists}.
\end{proof}

\begin{restatable}[Limit probability inequalities which hold for most distributions]{lem}{limProbMost}\label{lem:lim-prob-most}
Let $I\subseteq \reals$, let $\distSet \subseteq \Delta(\rewardVS)$ be closed under permutation, and let $F_A,F_B,F_C$ be finite sets of vector functions $I\mapsto \rewardVS$. Let $\gamma$ be a limit point of $I$ such that
\begin{align*}
    f_1(\D) & \defeq \lim_{\gamma^*\to \gamma}\phelper{F_B(\gamma^*)\geq F_C(\gamma^*)}[\D],\\
    f_2(\D) & \defeq \lim_{\gamma^*\to \gamma}\phelper{F_A(\gamma^*)\geq F_C(\gamma^*)}[\D]
\end{align*}
are well-defined for all $\D\in\distSet$.

Let $F_Z$ satisfy $\ND{F_C}\subseteq F_Z \subseteq F_C$. Suppose $F_B$ contains a copy of $F_A$ via $\phi$ such that $\phi\cdot \prn{F_Z\setminus \prn{F_B\setminus \phi\cdot F_A}}=F_Z\setminus \prn{F_B\setminus \phi\cdot F_A}$. Then $f_2(\distSet) \leqMost[][\distSet] f_1(\distSet)$.
\end{restatable}
\begin{proof}
Suppose $\D\in\distSet$ is such that $f_2(\D)>f_1(\D)$.
\begin{align}
    f_2\prn{\phi\cdot \D}&=f_2\prn{\phi\inv\cdot \D}\label{eq:involute-f1}\\
    &\defeq \lim_{\gamma^*\to \gamma}\phelper{F_A(\gamma^*)\geq F_C(\gamma^*)}[\phi\inv\cdot \D]\\
    &\leq \lim_{\gamma^*\to \gamma}\phelper{F_B(\gamma^*)\geq F_C(\gamma^*)}[\D]\label{eq:FB-geq}\\
    &<\lim_{\gamma^*\to \gamma}\phelper{F_A(\gamma^*)\geq F_C(\gamma^*)}[\D]\label{eq:assumed-ineq}\\
    &\leq  \lim_{\gamma^*\to \gamma}\phelper{F_B(\gamma^*)\geq F_C(\gamma^*)}[\phi\cdot \D]\label{eq:FB-geq-2}\\
    &\eqdef f_1\prn{\phi\cdot \D}.
\end{align}
By the assumption that $\distSet$ is closed under permutation and $f_2$ is well-defined for all $\D\in\distSet$, $f_2(\phi\cdot \D)$ is well-defined. \Cref{eq:involute-f1} follows since $\phi=\phi\inv$ because $\phi$ is an involution. For all $\gamma^*\in I$, let $A\defeq F_A(\gamma^*), B\defeq F_B(\gamma^*), C\defeq F_C(\gamma^*), Z\defeq F_Z(\gamma^*)$ (by \cref{def:nd-vec-func}, $\ND{C}\subseteq Z \subseteq C$). Since $\phi\cdot A\subseteq B$ by assumption, and since $\ND{A}\subseteq A$, $B$ also contains a copy of $\ND{A}$ via $\phi$. Furthermore, $\phi\cdot \prn{Z\setminus \prn{B\setminus \phi\cdot A}}=Z\setminus \prn{B\setminus \phi\cdot A}$ (by assumption), and so apply  \cref{lem:sim-lin-func-opt} to conclude that $\phelper{F_A(\gamma^*)\geq F_C(\gamma^*)}[\phi\inv\cdot \D]\leq \phelper{F_B(\gamma^*)\geq F_C(\gamma^*)}[\D]$. Therefore, the limit inequality \cref{eq:FB-geq} holds. \Cref{eq:assumed-ineq} follows because we assumed that $f_1(\D)<f_2(\D)$. \Cref{eq:FB-geq-2} holds by reasoning similar to that given for \cref{eq:FB-geq}.

Therefore, $f_2(\D)>f_1(\D)$ implies that $f_2\prn{\phi\cdot \D}<f_1\prn{\phi\cdot \D}$, and so apply \cref{lem:helper-geq-most} to conclude that $f_2(\D) \leqMost[][\distSet] f_1(\D)$.
\end{proof}

\subsubsection{\texorpdfstring{$\Fnd$}{Non-dominated visit distribution} results}

\transferDiscount
\begin{proof}
Let $R$ be any reward function. Suppose $\gamma^*\in(0,1)$ and construct $R'(s)\defeq \OptVf{s, \gamma}- \gamma^* \max_{a\in\A}\E{s'\sim T(s,a)}{\OptVf{s',\gamma}}$.

Let $\pi\in\Pi$ be any policy. By the definition of optimal policies, $\pi\in\optPi[R',\gamma^*]$ iff for all $s$:
\begin{align}
    R'(s) + \gamma^* \E{s' \sim T\prn{s,\pi(s)}}{\OptVf[R']{s',\gamma^*}}&=R'(s) + \gamma^* \max_{a\in\A}\E{s' \sim T\prn{s,a}}{\OptVf[R']{s',\gamma^*}}\\
    R'(s) + \gamma^* \E{s' \sim T\prn{s,\pi(s)}}{\OptVf{s',\gamma}}&=R'(s) + \gamma^* \max_{a\in\A}\E{s' \sim T\prn{s,a}}{\OptVf{s',\gamma}}\label{eq:opt-val-fn-equality}\\
    \gamma^*\E{s' \sim T\prn{s,\pi(s)}}{\OptVf{s',\gamma}}&=\gamma^*\max_{a\in\A}\E{s'\sim T(s,a)}{\OptVf{s',\gamma}}\label{eq:subst-r'}\\
    \E{s' \sim T\prn{s,\pi(s)}}{\OptVf{s',\gamma}}&=\max_{a\in\A}\E{s'\sim T(s,a)}{\OptVf{s',\gamma}}.\label{eq:div-gamma*}
\end{align}

By the Bellman equations, $R'(s)=\OptVf[R']{s, \gamma^*}- \gamma^* \max_{a\in\A}\E{s'\sim T(s,a)}{\OptVf[R']{s',\gamma^*}}$. By the definition of $R'$, $\OptVf[R']{\cdot,\gamma^*}=\OptVf[R]{\cdot, \gamma}$ must be the unique solution to the Bellman equations for $R'$ at $\gamma^*$. Therefore, \cref{eq:opt-val-fn-equality} holds. \Cref{eq:subst-r'} follows by plugging in $R'\defeq \OptVf{s, \gamma}- \gamma^* \max_{a\in\A}\E{s'\sim T(s,a)}{\OptVf{s',\gamma}}$ to \cref{eq:opt-val-fn-equality} and doing algebraic manipulation. \Cref{eq:div-gamma*} follows because $\gamma^*>0$.

\Cref{eq:div-gamma*} shows that $\pi\in\optPi[R',\gamma^*]$ iff \[\forall s: \E{s'\sim T(s,\pi(s))}{\OptVf{s',\gamma}} = \max_{a\in\A} \E{s'\sim T(s,a)}{\OptVf{s',\gamma}}.\] That is, $\pi\in\optPi[R',\gamma^*]$ iff $\pi\in\optPi[R,\gamma]$.
\end{proof}

\begin{restatable}[Evaluating sets of visit distribution functions at $\gamma$]{definition}{evalFDisc}\label{def:eval-f-discount}
For $\gamma \in (0,1)$, define $\F(s,\gamma)\defeq \set{\f(\gamma) \mid \f \in \F(s)}$ and $\Fnd(s,\gamma)\defeq \set{\f(\gamma) \mid \f \in \Fnd(s)}$. If $F\subseteq \F(s)$, then $F(\gamma) \defeq \set{\f(\gamma) \mid \f \in F}$.
\end{restatable}

\begin{restatable}[Non-domination across $\gamma$ values for mixtures of $\f$]{lem}{ndSetFGamma}\label{lem:nd-gamma-F-subset}
Let $\Delta_d \in \Delta\prn{\rewardVS}$ be any state distribution and let $F\defeq\set{\E{s_d\sim\Delta_d}{\fpi{s_d}} \mid \pi \in \Pi}$. $\f \in \ND{F}$ iff $\forall \gamma^*\in(0,1): \f(\gamma^*)\in \ND{F(\gamma^*)}$.
\end{restatable}
\begin{proof}
Let $\fpi{}\in\ND{F}$ be strictly optimal for reward function $R$ at discount rate $\gamma\in(0,1)$:
\begin{align}
    \fpi{}(\gamma)^\top \rf > \max_{\fpi[\pi']{}\in F\setminus \set{\fpi{}}} \fpi[\pi']{}(\gamma)^\top \rf.
\end{align}

Let $\gamma^*\in(0,1)$. By \cref{transferDiscount}, there exists $R'$ such that $\optPi[R',\gamma^*]=\optPi$. Since the optimal policy sets are equal, \cref{lem:opt-pol-visit-iff} implies that
\begin{align}
    \fpi{}(\gamma^*)^\top \rf' > \max_{\fpi[\pi']{}\in F\setminus \set{\fpi{}}} \fpi[\pi']{}(\gamma^*)^\top \rf'.
\end{align}

Therefore, $\fpi{}(\gamma^*)\in\ND{F(\gamma^*)}$.

The reverse direction follows by the definition of $\ND{F}$.
\end{proof}

\begin{restatable}[$\forall \gamma \in (0,1): \dbf\in \Fnd(s,\gamma)$ iff $\dbf\in\ND{\F(s,\gamma)}$]{lem}{invariantFndDiscount}\label{lem:nd-relation}
\end{restatable}
\begin{proof}
By \cref{def:eval-f-discount}, $\Fnd(s,\gamma)\defeq \set{\f(\gamma) \mid \f \in \ND{\F(s)}}$. By applying \cref{lem:nd-gamma-F-subset} with $\Delta_d\defeq \unitvec$, $\f \in \ND{\F(s)}$ iff $\forall \gamma\in(0,1): \f(\gamma)\in \ND{\F(s,\gamma)}$.
\end{proof}

\optVfFndRestrict
\begin{proof}
$\ND{\F(s,\gamma)}=\Fnd(s,\gamma)$ by \cref{lem:nd-relation}, so apply \cref{cor:nd-func-indif} with $X \defeq \F(s,\gamma)$.
\end{proof}

\subsection{Some actions have greater probability of being optimal}
\begin{restatable}[Optimal policy shift bound]{lem}{optPiBound}\label{lem:opt-pol-shift-bound}
For fixed $R$, $\optPi$ can take on at most $(2\abs{\St}+1)\sum_s \binom{\abs{\F(s)}}{2}$ distinct values over $\gamma \in (0,1)$.
\end{restatable}
\begin{proof}
By \cref{lem:opt-pol-visit-iff}, $\optPi$ changes value iff there is a change in optimality status for some visit distribution function at some state. \citet{lippman1968set} showed that two visit distribution functions can trade off optimality status at most $2\abs{\St}+1$ times. At each state $s$, there are $\binom{\abs{\F(s)}}{2}$ such pairs.
\end{proof}

\begin{restatable}[Optimality probability's limits exist]{prop}{muConverge} \label{prop:opt-prob-converge} Let $F\subseteq \F(s)$. $\optprob[\Dany]{F,0}=\lim_{\gamma\to 0} \optprob[\Dany]{F,\gamma}$ and $\optprob[\Dany]{F,1}=\lim_{\gamma\to 1} \optprob[\Dany]{F,\gamma}$.
\end{restatable}
\begin{proof}
First consider the limit as $\gamma \to 1$. Let $\Dany$ have probability measure $F_\text{any}$, and define $\delta(\gamma)\defeq F_\text{any}\prn{\set{R\in\rewardSpace \mid \exists \gamma^* \in [\gamma,1): \optPi[R,\gamma^*]\neq \optPi[R,1]}}$. Since $F_\text{any}$ is a probability measure, $\delta(\gamma)$ is bounded $[0,1]$, and $\delta(\gamma)$ is monotone decreasing. Therefore, $\lim_{\gamma \to 1} \delta(\gamma)$ exists.

If $\lim_{\gamma \to 1} \delta(\gamma)>0$, then there exist reward functions whose optimal policy sets $\optPi$ never converge (in the discrete topology on sets) to $\optPi[R,1]$, contradicting \cref{lem:opt-pol-shift-bound}. So $\lim_{\gamma \to 1} \delta(\gamma)=0$.

By the definition of optimality probability (\cref{def:prob-opt}) and of $\delta(\gamma)$,  $|\optprob[\Dany]{F,\gamma}-\optprob[\Dany]{F,1}|\leq \delta(\gamma)$. Since $\lim_{\gamma\to 1} \delta(\gamma)=0$, $\lim_{\gamma\to 1}\optprob[\Dany]{F,\gamma}=\optprob[\Dany]{F,1}$.

A similar proof shows that $\lim_{\gamma\to 0} \optprob[\Dany]{F,\gamma}=\optprob[\Dany]{F,0}$.
\end{proof}

\begin{restatable}[Optimality probability identity]{lem}{optProbID}\label{lem:opt-prob-id}
Let $\gamma\in(0,1)$ and let $F\subseteq \F(s)$.
\begin{equation}
\optprob[\Dany]{F,\gamma}=\phelper{F(\gamma)\geq \F(s,\gamma)}=\phelper{F(\gamma)\geq \Fnd(s,\gamma)}.
\end{equation}
\end{restatable}
\begin{proof}
Let $\gamma\in(0,1)$.
\begin{align}
    \optprob[\Dany]{F,\gamma}&\defeq \prob[R\sim\Dany]{\exists \f^\pi \in F: \pi\in\optPi}\\
    &=\E{\rf \sim\Dany}{ \indic{\max_{\f\in F} \f(\gamma)^\top \rf = \max_{\f'\in\F(s)} \f'(\gamma)^\top \rf}}\label{eq:F-probability}\\
    &=\E{\rf \sim\Dany}{ \indic{\max_{\f\in F} \f(\gamma)^\top \rf = \max_{\f'\in\Fnd(s)} \f'(\gamma)^\top \rf}}\label{eq:Fnd-probability}\\
    &\eqdef \phelper{F(\gamma)\geq \Fnd(s,\gamma)}.
\end{align}
\Cref{eq:F-probability} follows because \cref{lem:opt-pol-visit-iff} shows that $\pi$ is optimal iff it induces an optimal visit distribution  $\f$ at every state. \Cref{eq:Fnd-probability} follows because $\forall \rf \in \rewardVS$:
\begin{equation*}
    \max_{\f'\in\F(s)} \f'(\gamma)^\top \rf= \max_{\f'\in\Fnd(s)} \f'(\gamma)^\top \rf
\end{equation*}
by \cref{cor:opt-vf-restrict-fnd}.
\end{proof}

\subsection{Basic properties of \texorpdfstring{$\pwrNoDist$}{power}}

\begin{restatable}[$\pwrNoDist$ identities]{lem}{lemPowEQ}\label{lem:power-id}
Let $\gamma \in (0,1)$.
\begin{align}
    \pwr[s,\gamma]&=\E{\rf\sim\Dbd}{\max_{\f\in \Fnd(s)} \frac{1-\gamma}{\gamma}\prn{\f(\gamma)-\unitvec}^\top \rf}\label{eq:pwr-def-f}\\
    &= \dfrac{1-\gamma}{\gamma}\E{\rf\sim\Dbd}{\OptVf{s,\gamma}-R(s)}\\
    &=  \dfrac{1-\gamma}{\gamma}\prn{\vavg-\E{R\sim \Dbd}{R(s)}} \\
    &=\E{R\sim\Dbd}{\max_{\pi\in\Pi} \E{s'\sim T\prn{s,\pi(s)}}{\prn{1-\gamma}V^\pi_R\prn{s',\gamma}}}.\label{eq:pwr-def-avg-discounted}
\end{align}
\end{restatable}
\begin{proof}
\begin{align}
    \pwrNoDist_{\Dbd}(s,\gamma)&\defeq\E{\rf\sim\Dbd}{\max_{\f\in \F(s)} \frac{1-\gamma}{\gamma}\prn{\f(\gamma)-\unitvec}^\top \rf}\\
    &= \E{\rf\sim\Dbd}{\max_{\f\in \Fnd(s)} \frac{1-\gamma}{\gamma}\prn{\f(\gamma)-\unitvec}^\top \rf}\label{eq:pwr-fnd-restrict}\\
    &=\E{\rf\sim\Dbd}{\max_{\f\in \F(s)} \frac{1-\gamma}{\gamma}\prn{\f(\gamma)-\unitvec}^\top \rf}\\
    &= \dfrac{1-\gamma}{\gamma}\E{\rf\sim\Dbd}{\OptVf{s,\gamma}-R(s)}\label{eq:pwr-vf-convert}\\
    &=\dfrac{1-\gamma}{\gamma}\prn{\vavg-\E{R\sim \Dbd}{R(s)}}\label{eq:vavg-def-pwr-def}\\
    &= \E{\rf\sim\Dbd}{\max_{\pi\in\Pi} \E{s'\sim T\prn{s,\pi(s)}}{\prn{1-\gamma}\fpi[\pi]{s'}(\gamma)^\top\rf}}\label{eq:f-expand-pwr-def}\\
    &= \E{R\sim\Dbd}{\max_{\pi\in\Pi} \E{s'\sim T\prn{s,\pi(s)}}{\prn{1-\gamma}V^\pi_R\prn{s',\gamma}}}.\label{eq:pwr-avg-normalized}
\end{align}

\Cref{eq:pwr-fnd-restrict} follows from \cref{cor:opt-vf-restrict-fnd}. \Cref{eq:pwr-vf-convert} follows from the dual formulation of optimal value functions. \Cref{eq:vavg-def-pwr-def} holds by the definition of $\vavg$ (\cref{def:vavg}). \Cref{eq:f-expand-pwr-def} holds because $\fpi{s}(\gamma) = \unitvec +\gamma \E{s'\sim T\prn{s,\pi(s)}}{\fpi[\pi]{s'}(\gamma)}$ by the definition of a visit distribution function (\cref{def:visit}).
\end{proof}

\begin{restatable}[Discount-normalized value function]{definition}{discValue}\label{def:disc-value}
Let $\pi$ be a policy, $R$ a reward function, and $s$ a state. For $\gamma \in [0,1]$, $\VfNorm[\pi]{s,\gamma}\defeq \lim_{\gamma^*\to \gamma}(1-\gamma^*) V^\pi_R(s,\gamma^*)$.
\end{restatable}

\begin{restatable}[Normalized value functions have uniformly bounded derivative]{lem}{normValueLip}\label{lem:norm-value-lip}
There exists $K\geq0$ such that for all reward functions $\rf\in \rewardVS$, $\sup_{\substack{s\in\St, \pi\in\Pi,\gamma\in [0,1]}}\abs{\frac{d}{d\gamma} \VfNorm[\pi]{s,\gamma}} \leq K\lone{\rf}$.
\end{restatable}
\begin{proof}
Let $\pi$ be any policy, $s$ a state, and $R$ a reward function. Since \[\VfNorm[\pi]{s,\gamma}=\lim_{\gamma^*\to\gamma}(1-\gamma^*)\fpi{s}(\gamma^*)^\top \rf,\] $\frac{d}{d\gamma}\VfNorm[\pi]{s,\gamma}$ is controlled by the behavior of $\lim_{\gamma^*\to\gamma}(1-\gamma^*)\fpi{s}(\gamma^*)$. We show that this function's gradient is bounded in infinity norm.

By \cref{f-rat}, $\fpi{s}(\gamma)$ is a multivariate rational function on $\gamma$. Therefore, for any state $s'$,  $\fpi{s}(\gamma)^\top\unitvec[s']=\frac{P(\gamma)}{Q(\gamma)}$ in reduced form. By \cref{prop:visit-dist-prop}, $0\leq \fpi{s}(\gamma)^\top\unitvec[s']\leq \frac{1}{1-\gamma}$. Thus, $Q$ may only have a root of multiplicity $ 1$ at $\gamma=1$, and $Q(\gamma)\neq 0$ for $\gamma\in[0,1)$. Let $f_{s'}(\gamma)\defeq (1-\gamma)\fpi{s}(\gamma)^\top\unitvec[s']$.

If $Q(1)\neq 0$, then the derivative $f_{s'}'(\gamma)$ is bounded on $\gamma\in[0,1)$ because the polynomial $(1-\gamma)P(\gamma)$ cannot diverge on a bounded domain.

If $Q(1)=0$, then factor out the root as $Q(\gamma)=(1-\gamma)Q^*(\gamma)$.
\begin{align}
    f_{s'}'(\gamma)&=\frac{d}{d\gamma}\prn{\frac{(1-\gamma)P(\gamma)}{Q(\gamma)}}\\
    &=\frac{d}{d\gamma}\prn{\frac{P(\gamma)}{Q^*(\gamma)}}\\
    &=\frac{P'(\gamma)Q^*(\gamma)-(Q^*)'(\gamma)P(\gamma)}{(Q^*(\gamma))^2}.
\end{align}

Since $Q^*(\gamma)$ is a polynomial with no roots on $\gamma\in[0,1]$, $f_{s'}'(\gamma)$ is bounded on $\gamma\in[0,1)$.

Therefore, whether or not $Q(\gamma)$ has a root at $\gamma=1$, $f_{s'}'(\gamma)$ is bounded on $\gamma\in[0,1)$. Furthermore, $\sup_{\gamma\in[0,1)}\linfty{\nabla(1-\gamma)\fpi{s}(\gamma)}=\sup_{\gamma\in[0,1)}\max_{s'\in\St} \abs{f_{s'}'(\gamma)}$ is finite since there are only finitely many states.

There are finitely many $\pi\in\Pi$, and finitely many states $s$, and so there exists some $K'$ such that $\sup_{\substack{s\in\St,\\ \pi\in\Pi,\gamma\in [0,1)}} \linfty{\nabla(1-\gamma)\fpi{s}(\gamma)}\leq K'$. Then $\lone{\nabla(1-\gamma)\fpi{s}(\gamma)}\leq \abs{\St}K'\eqdef K$.
\begin{align}
    \sup_{\substack{s\in\St,\\ \pi\in\Pi,\gamma\in [0,1)}} \abs{\frac{d}{d\gamma}V^\pi_{R,\text{norm}}\prn{s,\gamma}}\defeq\,&\sup_{\substack{s\in\St,\\ \pi\in\Pi,\gamma\in [0,1)}} \abs{\frac{d}{d\gamma}\lim_{\gamma^*\to\gamma}(1-\gamma^*)V^\pi_{R}\prn{s,\gamma^*}}\\
    =\,&\sup_{\substack{s\in\St,\\ \pi\in\Pi,\gamma\in [0,1)}} \abs{\frac{d}{d\gamma}(1-\gamma)V^\pi_{R}\prn{s,\gamma}}\label{eq:lim-disappear}\\
    =\,&\sup_{\substack{s\in\St,\\ \pi\in\Pi,\gamma\in [0,1)}} \abs{\nabla (1-\gamma)\fpi{s}(\gamma)^\top\rf}\label{eq:on-pol-rat}\\
    \leq\, & \sup_{\substack{s\in\St,\\ \pi\in\Pi,\gamma\in [0,1)}} \lone{\nabla (1-\gamma)\fpi{s}(\gamma)}\lone{\rf}\label{eq:cs}\\
    \leq\, & K\lone{\rf}.\label{eq:bounded}
\end{align}

\Cref{eq:lim-disappear} holds because $\Vf[\pi]{s,\gamma}$ is continuous on $\gamma\in[0,1)$ by \cref{smoothOnPol}. \Cref{eq:cs} holds by the Cauchy-Schwarz inequality.

Since $\abs{\frac{d}{d\gamma}V^\pi_{R,\text{norm}}\prn{s,\gamma}}$ is bounded for all $\gamma \in [0,1)$, \cref{eq:bounded} also holds for $\gamma\to 1$.
\end{proof}

\ContPower*
\begin{proof}
Let $b,c$ be such that $\supp[\Dbd]\subseteq [b,c]^{\abs{\St}}$. For any $\rf\in\supp[\Dbd]$ and $\pi\in\Pi$, $\VfNorm[\pi]{s,\gamma}$ has Lipschitz constant $K\lone{\rf}\leq K \abs{\St}\linfty{\rf}\leq K\abs{\St}\max(\abs{c},\abs{b})$ on $\gamma\in(0,1)$ by \cref{lem:norm-value-lip}.

For $\gamma\in(0,1)$, $\pwr[s,\gamma] = \E{R\sim\Dbd}{\max_{\pi\in\Pi} \E{s'\sim T\prn{s,\pi(s)}}{(1-\gamma)V^\pi_R\prn{s',\gamma}}}$ by \cref{eq:pwr-avg-normalized}. The expectation of the maximum of a set of functions which share a Lipschitz constant, also shares the Lipschitz constant. This shows that $\pwr[s,\gamma]$ is Lipschitz continuous on $\gamma\in(0,1)$. Thus, its limits are well-defined as $\gamma\to 0$ and $\gamma\to 1$. So it is Lipschitz continuous on the closed unit interval.
\end{proof}

\maxPwrGeneral*
\begin{proof}
Let $\gamma \in (0,1)$.
\begin{align}
    \pwr[s,\gamma][\Dbd]&= \E{R\sim \Dbd}{\max_{\pi\in\Pi} \E{s'\sim T(s,\pi(s))}{(1-\gamma)\OptVf{s',\gamma}}}\label{eq:pwr-id}\\
    &\leq \E{R\sim \Dbd}{\max_{\pi\in\Pi} \E{s'\sim T(s,\pi(s))}{(1-\gamma)\geom[\max_{s''\in\St}R(s'')]}}\label{eq:max-vfn}\\
    &= \E{R\sim \Dbd}{\max_{s''\in\St}R(s'')}.
\end{align}

\Cref{eq:pwr-id} follows from \cref{lem:power-id}. \Cref{eq:max-vfn} follows because $\OptVf{s',\gamma}\leq \geom[\max_{s''\in\St}R(s'')]$, as no policy can do better than achieving maximal reward at each time step. Taking limits, the inequality holds for all $\gamma\in[0,1]$.

Suppose that $s$ can deterministically reach all states in one step and all states are {\stateEnd}s. Then \cref{eq:max-vfn} is an equality for all $\gamma\in(0,1)$, since for each $R$, the agent can select an action which deterministically transitions to a state with maximal reward. Thus the equality holds for all $\gamma\in[0,1]$.
\end{proof}

\begin{restatable}[Lower bound on current $\pwrNoDist$ based on future $\pwrNoDist$]{lem}{FuturePower}\label{lem:future-power}
\begin{align}
\pwr[s,\gamma]\geq (1-\gamma)\min_a \E{\substack{s'\sim T(s,a),\\R\sim\Dbd}}{R(s')} + \gamma\max_a \E{s'\sim T(s,a)}{\pwr[s',\gamma]}.
\end{align}
\end{restatable}
\begin{proof}
Let $\gamma \in (0,1)$ and let $a^*\in \argmax_a \E{s'\sim T\prn{s,a}}{\pwr[s',\gamma]}$.
\begin{align}
    &\pwr[s,\gamma]\\
    =\,&(1-\gamma)\E{R\sim\Dbd}{\max_a \E{s'\sim T\prn{s,a}}{\OptVf{s',\gamma}}}\label{eq:pwr-id-app}\\
    \geq\,& (1-\gamma) \max_a \E{s'\sim T\prn{s,a}}{\E{R\sim\Dbd}{\OptVf{s',\gamma}}}\label{eq:max-expect-ineq}\\
    =\,&(1-\gamma)\max_a \E{s'\sim T\prn{s,a}}{\vavg[s',\gamma]}\\
    =\,&(1-\gamma) \max_a \E{s'\sim T\prn{s,a}}{\E{R\sim \Dbd}{R(s')}+\frac{\gamma}{1-\gamma}\pwr[s',\gamma]}\label{eq:vavg-identity}\\
    \geq\,&(1-\gamma) \E{s'\sim T\prn{s,a^*}}{\E{R\sim \Dbd}{R(s')}+\frac{\gamma}{1-\gamma}\pwr[s',\gamma]}\\
    \geq\,&(1-\gamma)\min_a \E{\substack{s'\sim T(s,a),\\R\sim\Dbd}}{R(s')} + \gamma \E{s'\sim T\prn{s,a^*}}{\pwr[s',\gamma]}.\label{eq:min-reward-max-power}
\end{align}

\Cref{eq:pwr-id-app} holds by \cref{lem:power-id}. \Cref{eq:max-expect-ineq} follows because
\begin{equation*}
\E{x\sim X}{\max_a f(a,x)}\geq \max_a \E{x\sim X}{f(a,x)}
\end{equation*}
by Jensen's inequality, and \cref{eq:vavg-identity} follows by \cref{lem:power-id}.

The inequality also holds when we take the limits  $\gamma \to 0$ or $\gamma \to 1$.
\end{proof}

\smooth*
\begin{proof}
Let $\gamma\in[0,1]$. First consider the case where $\pwr[s,\gamma]\geq \pwr[s',\gamma]$.
\begin{align}
\pwr[s',\gamma]&\geq (1-\gamma)\min_a \E{\substack{s_x\sim T(s',a),\\R\sim\Dbd}}{R(s_x)} + \gamma\max_a \E{s_x\sim T(s',a)}{\pwr[s_x,\gamma]}\label{eq:lb-pwr-next}\\
&\geq (1-\gamma)b + \gamma \pwr[s,\gamma].\label{eq:lb-pwr-next-min}
\end{align}

\Cref{eq:lb-pwr-next} follows by \cref{lem:future-power}. \Cref{eq:lb-pwr-next-min} follows because reward is lower-bounded by $b$ and because  $s'$ can reach $s$ in one step with probability $1$.
\begin{align}
    \phantom{=}&\abs{\pwr[s,\gamma]-\pwr[s',\gamma]}\\
    &=\pwr[s,\gamma]-\pwr[s',\gamma]\label{eq:no-abs}\\
    &\leq \pwr[s,\gamma]-\prn{(1-\gamma)b + \gamma \pwr[s,\gamma]}\label{eq:lb-pwr-abs-diff}\\
    &=(1-\gamma)\prn{\pwr[s,\gamma] - b}\\
    &\leq(1-\gamma)\prn{\E{R\sim \Dbd}{\max_{s''\in\St}R(s'')} - b}\label{eq:ub-pwr-s}\\
    &\leq (1-\gamma)(c-b).\label{eq:ub-pwr-abs-diff}
\end{align}

\Cref{eq:no-abs} follows because \[\pwr[s,\gamma]\geq \pwr[s',\gamma].\] \Cref{eq:lb-pwr-abs-diff} follows by \cref{eq:lb-pwr-next-min}. \Cref{eq:ub-pwr-s} follows by \cref{lem:max-power-general}. \Cref{eq:ub-pwr-abs-diff} follows because reward under $\Dbd$ is upper-bounded by $c$.

The case where $\pwr[s,\gamma]\leq \pwr[s',\gamma]$ is similar, leveraging the fact that $s$ can also reach $s'$ in one step with probability $1$.
\end{proof}

\subsection{Seeking \texorpdfstring{$\pwrNoDist$}{power} is often more probable under optimality}

\subsubsection{Keeping options open tends to be \texorpdfstring{$\pwrNoDist$}{POWER}-seeking and tends to be optimal}

\begin{restatable}[Normalized visit distribution function]{definition}{normVisitFn}\label{def:norm-visit-fn}
Let $\f:[0,1)\to \rewardVS$ be a vector function. For $\gamma\in[0,1]$, $\NormF{\f,\gamma}\defeq \lim_{\gamma^*\to\gamma} (1-\gamma^*)\f(\gamma^*)$ (this limit need not exist for arbitrary $\f$). If $F$ is a set of such $\f$, then $\NormF{F,\gamma}\defeq \set{\NormF{\f,\gamma}\mid \f \in F}$.
\end{restatable}

\begin{remark}
$\RSD=\NormF{\F(s),1}$.
\end{remark}

\begin{restatable}[Normalized visit distribution functions are continuous]{lem}{contNormFMix}\label{lem:cont-norm-f}
Let $\Delta_s\in \Delta(\St)$ be a state probability distribution, let $\pi\in\Pi$, and let $\f^*\defeq \E{s\sim \Delta_s}{\fpi{s}}$. $\NormF{\f^*, \gamma}$ is continuous on $\gamma\in [0,1]$.
\end{restatable}
\begin{proof}
\begin{align}
    \NormF{\f^*,\gamma}&\defeq \lim_{\gamma^*\to\gamma} (1-\gamma^*)\E{s\sim \Delta_s}{\fpi{s}(\gamma^*)}\\
    &=\E{s\sim \Delta_s}{\lim_{\gamma^*\to\gamma} (1-\gamma^*)\fpi{s}(\gamma^*)}\label{eq:expect-lim-swap}\\
    &\eqdef \E{s\sim \Delta_s}{\NormF{\fpi{s},\gamma}}.\label{eq:expect-cont}
\end{align}
\Cref{eq:expect-lim-swap} follows because the expectation is over a finite set. Each $\fpi{s}\in\F(s)$ is continuous on $\gamma\in[0,1)$ by \cref{f-rat}, and $\lim_{\gamma^*\to 1} (1-\gamma^*) \fpi{s}(\gamma^*)$ exists because {\rsd}s are well-defined \citep{puterman_markov_2014}. Therefore, each $\NormF{\fpi{s},\gamma}$ is continuous on $\gamma\in[0,1]$. Lastly, \cref{eq:expect-cont}'s expectation over finitely many continuous functions is itself continuous.
\end{proof}

\begin{restatable}[Non-domination of normalized visit distribution functions]{lem}{ndNormalVisit}\label{lem:nd-norm-visit}
Let $\Delta_s\in \Delta(\St)$ be a state probability distribution and let $F\defeq \set{\E{s\sim \Delta_s}{\fpi{s}}\mid\pi\in\Pi}$. For all $\gamma\in[0,1]$, $\ND{\NormF{F,\gamma}}\subseteq \NormF{\ND{F},\gamma}$, with equality when $\gamma\in(0,1)$.
\end{restatable}
\begin{proof}
Suppose $\gamma\in (0,1)$.
\begin{align}
    \ND{\NormF{F,\gamma}}&=\ND{(1-\gamma)F(\gamma)}\label{eq:apply-cont-norm-1}\\
    &= (1-\gamma)\ND{F(\gamma)}\label{eq:nd-rescale-invariant}\\
    &= (1-\gamma)\prn{\ND{F}(\gamma)}\label{eq:nd-invariant-01}\\
    &= \NormF{\ND{F},\gamma}.\label{eq:apply-cont-norm-2}
\end{align}

\Cref{eq:apply-cont-norm-1} and \cref{eq:apply-cont-norm-2} follow by the continuity of $\NormF{\f,\gamma}$ (\cref{lem:cont-norm-f}). \Cref{eq:nd-rescale-invariant} follows by \cref{lem:pos-aff-nd-invar} \cref{item:invar-vectors}. \Cref{eq:nd-invariant-01} follows by \cref{lem:nd-gamma-F-subset}.

Let $\gamma=1$. Let $\dbf \in \ND{\NormF{F,1}}$ be strictly optimal for $\rf^*\in\rewardVS$. Then let $F_\dbf \subseteq F$ be the subset of $\f\in F$ such that $\NormF{\f,1}=\dbf$.
\begin{align}
    \max_{\f\in F_\dbf} \NormF{\f,1}^\top \rf^* & > \max_{\f'\in F\setminus F_\dbf} \NormF{\f',1}^\top \rf^*.\label{eq:norm-f-ineq}
\end{align}

Since $\NormF{\f,1}$ is continuous at $\gamma=1$ (\cref{lem:cont-norm-f}), $\x^\top \rf^*$ is continuous on $\x\in\rewardVS$, and $F$ is finite, \cref{eq:norm-f-ineq} holds for some $\gamma^*\in(0,1)$ sufficiently close to $\gamma=1$. By \cref{lem:all-rf-max-nd}, at least one $\f\in F_\dbf$ is an element of $\ND{F(\gamma^*)}$. Then by \cref{lem:nd-gamma-F-subset}, $\f\in\ND{F}$. We conclude that $\ND{\NormF{F,1}}\subseteq \NormF{\ND{F},1}$.

The case for $\gamma=0$ proceeds similarly.
\end{proof}

\begin{restatable}[$\pwrNoDist$ limit identity]{lem}{pwrLimit}\label{lem:pwr-limit}
Let $\gamma\in[0,1]$.
\begin{align}
    \pwr[s,\gamma]&=\E{\rf\sim\Dbd}{\max_{\f\in \Fnd(s)} \lim_{\gamma^*\to\gamma}\frac{1-\gamma^*}{\gamma^*}\prn{\f(\gamma^*)-\unitvec}^\top \rf}.
\end{align}
\end{restatable}
\begin{proof}
Let $\gamma\in[0,1]$.
\begin{align}
    \pwr[s,\gamma]&= \lim_{\gamma^*\to\gamma}\pwr[s,\gamma^*]\label{eq:pwr-cont-lim}\\
    &=\lim_{\gamma^*\to\gamma} \E{\rf\sim\Dbd}{\max_{\f\in \Fnd(s)} \frac{1-\gamma^*}{\gamma^*}\prn{\f(\gamma^*)-\unitvec}^\top \rf}\label{eq:pwr-lim-def}\\
    &=\E{\rf\sim\Dbd}{\lim_{\gamma^*\to\gamma}\max_{\f\in \Fnd(s)} \frac{1-\gamma^*}{\gamma^*}\prn{\f(\gamma^*)-\unitvec}^\top \rf}\label{eq:pwr-dom-conv}\\
    &=\E{\rf\sim\Dbd}{\max_{\f\in \Fnd(s)} \lim_{\gamma^*\to\gamma}\frac{1-\gamma^*}{\gamma^*}\prn{\f(\gamma^*)-\unitvec}^\top \rf}.\label{eq:max-cont-lim}
\end{align}
\Cref{eq:pwr-cont-lim} follows since $\pwr[s,\gamma]$ is continuous on $\gamma\in[0,1]$ by \cref{thm:cont-power}. \Cref{eq:pwr-lim-def} follows by \cref{lem:power-id}.

For $\gamma^*\in(0,1)$, let $f_{\gamma^*}(\rf)\defeq \max_{\f\in \Fnd(s)} \frac{1-\gamma^*}{\gamma^*}\prn{\f(\gamma^*)-\unitvec}^\top \rf$. For any sequence $\gamma_n\to\gamma$, $\prn{f_{\gamma_n}}_{n=1}^\infty$ is a sequence of functions which are piecewise linear on $\rf \in \rewardVS$, which means they are continuous and therefore measurable. Since \cref{f-rat} shows that each $\f\in\Fnd(s)$ is multivariate rational on $\gamma^*$ (and therefore continuous on $\gamma^*$), $\set{f_{\gamma_n}}_{n=1}^\infty$ converges pointwise to limit function $f_\gamma$. Furthermore,  $\abs{\OptVf{s,\gamma_n}-R(s)}\leq \frac{\gamma}{1-\gamma_n}\linfty{R}$, and so $\abs{f_{\gamma_n}(\rf)}=\abs{\frac{1-\gamma_n}{\gamma_n}(\OptVf{s,\gamma_n}-R(s))}\leq g(\rf)\leq \linfty{\rf}\eqdef g(\rf)$, which is measurable. Therefore, apply Lebesgue's dominated convergence theorem to conclude that \cref{eq:pwr-dom-conv} holds. \Cref{eq:max-cont-lim} holds because $\max$ is a continuous function.
\end{proof}

\begin{restatable}[Lemma for $\pwrNoDist$ superiority]{lem}{morePowerPrefixmoreOptions}\label{lem:more-power-prefix}
Let $\Delta_1,\Delta_2\in \Delta\prn{\St}$ be state probability distributions. For $i=1,2$, let $F_{\Delta_i}\defeq \set{\gamma\inv\E{s_i\sim \Delta_i}{\fpi{s_i}-\unitvec[s_i]}\mid \pi\in\Pi}$. Suppose $F_{\Delta_2}$ contains a copy of $\ND{F_{\Delta_1}}$ via $\phi$. Then $\forall \gamma\in[0,1]:\E{s_1\sim \Delta_1}{\pwr[s_1,\gamma]}\leqMost[][\DSetBd] \E{s_2\sim \Delta_2}{\pwr[s_2,\gamma]}$.

If $\ND{F_{\Delta_2}}\setminus \phi\cdot \ND{F_{\Delta_1}}$ is non-empty, then for all $\gamma\in(0,1)$, the inequality is strict for all $\Diid\in\DSetBCiid$ and $\E{s_1\sim \Delta_1}{\pwr[s_1,\gamma][\Dbd]}\not\geqMost[][\DSetBd] \E{s_2\sim \Delta_2}{\pwr[s_2,\gamma][\Dbd]}$.

These results also hold when replacing $F_{\Delta_i}$ with $F_{\Delta_i}^*\defeq \set{\E{s_i\sim \Delta_i}{\fpi{s_i}}\mid \pi\in\Pi}$ for $i=1,2$.
\end{restatable}
\begin{proof}
\begin{align}
    \phi\cdot \ND{\NormF{F_{\Delta_1},\gamma}}&\subseteq \phi\cdot \NormF{\ND{F_{\Delta_1}},\gamma}\label{eq:subset-nd-norm}\\
    &\defeq \set{\permute\lim_{\gamma^*\to\gamma}(1-\gamma^*)\f(\gamma^*) \mid \f \in \ND{F_{\Delta_1}}}\\
    &=\set{\lim_{\gamma^*\to\gamma}(1-\gamma^*)\permute\f(\gamma^*) \mid \f \in \ND{F_{\Delta_1}}}\label{eq:cont-permute-app}\\
    &=\set{\lim_{\gamma^*\to\gamma}(1-\gamma^*)\f(\gamma^*) \mid \f \in F_{\text{sub}}'}\\
    &\subseteq \set{\lim_{\gamma^*\to\gamma}(1-\gamma^*)\f(\gamma^*) \mid \f\in F_{\Delta_2}}\label{eq:subset-f}\\
    &\eqdef \NormF{F_{\Delta_2},\gamma}.\label{eq:subset-f-final}
\end{align}
\Cref{eq:subset-nd-norm} follows by \cref{lem:nd-norm-visit}. \Cref{eq:cont-permute-app} follows because $\permute$ is a continuous linear operator. \Cref{eq:subset-f} follows by assumption.
\begin{align}
    \E{s_1\sim \Delta_1}{\pwr[s_1,\gamma]}&\defeq\E{\substack{s_1\sim \Delta_1,\\\rf\sim\Dbd}}{\max_{\pi\in\Pi} \lim_{\gamma^*\to\gamma}\frac{1-\gamma^*}{\gamma^*}\prn{\fpi{s_1}(\gamma^*)-\unitvec[s_1]}^\top \rf}\label{eq:expect-pwr-1}\\
    &=\E{\rf\sim\Dbd}{\max_{\pi\in\Pi} \lim_{\gamma^*\to\gamma}\frac{1-\gamma^*}{\gamma^*}\E{s_1\sim \Delta_1}{\fpi{s_1}(\gamma^*)-\unitvec[s_1]}^\top \rf}\label{eq:bring-inside-expect}\\
    &=\E{\rf\sim\Dbd}{\max_{\dbf\in \NormF{F_{\Delta_1},\gamma}} \dbf^\top \rf}\\
    &=\E{\rf\sim\Dbd}{\max_{\dbf\in \ND{\NormF{F_{\Delta_1},\gamma}}} \dbf^\top \rf}\label{eq:restrict-nd-fdelta}\\
    &\leqMost[][\DSetBd] \E{\rf\sim\Dbd}{\max_{\dbf\in \NormF{F_{\Delta_2},\gamma}} \dbf^\top \rf}\label{eq:leq-most-apply-transient}\\
    &=\E{\rf\sim\Dbd}{\max_{\pi\in\Pi} \lim_{\gamma^*\to\gamma}\frac{1-\gamma^*}{\gamma^*}\E{s_2\sim \Delta_2}{\fpi{s_2}(\gamma^*)-\unitvec[s_2]}^\top \rf}\\
    &=\E{\substack{s_2\sim \Delta_2,\\\rf\sim\Dbd}}{\max_{\pi\in\Pi} \lim_{\gamma^*\to\gamma}\frac{1-\gamma^*}{\gamma^*}\prn{\fpi{s_2}(\gamma^*)-\unitvec[s_2]}^\top \rf}\label{eq:bring-inside-expect-2}\\
    &\eqdef \E{s_2\sim \Delta_2}{\pwr[s_2,\gamma]}.\label{eq:expect-pwr-2}
\end{align}

\Cref{eq:expect-pwr-1} and \cref{eq:expect-pwr-2} follow by \cref{lem:pwr-limit}. \Cref{eq:bring-inside-expect} and \cref{eq:bring-inside-expect-2} follow because each $R$ has a stationary deterministic optimal policy $\pi\in \optPi\subseteq \Pi$ which simultaneously achieves  optimal value at all states. \Cref{eq:restrict-nd-fdelta} follows by \cref{cor:nd-func-indif}.

Apply \cref{lem:expect-superior} with $A\defeq \NormF{F_{\Delta_1},\gamma}, B \defeq \NormF{F_{\Delta_2},\gamma}$, $g$ the identity function, and involution $\phi$ (satisfying $\phi\cdot \ND{A}\subseteq B$ by \cref{eq:subset-f-final}) in order to conclude that \cref{eq:leq-most-apply-transient} holds.

Suppose that $\ND{F_{\Delta_2}}\setminus \phi\cdot \ND{F_{\Delta_1}}$ is non-empty; let $F_\text{sub}'\defeq \phi\cdot \ND{F_{\Delta_1}}$. \Cref{lem:nd-gamma-F-subset} shows that for all $\gamma\in(0,1)$, $\ND{F_{\Delta_2}(\gamma)}\setminus F_\text{sub}'(\gamma)$ is non-empty. \Cref{lem:pos-aff-nd-invar} \cref{item:invar-vectors} then implies that $\ND{B}\setminus \phi\cdot A=\frac{1-\gamma}{\gamma}\prn{\ND{F_{\Delta_2}(\gamma)}-\unitvec}\setminus \prn{\frac{1-\gamma}{\gamma}F'_\text{sub}(\gamma)}$ is non-empty. Then \cref{lem:expect-superior} implies that for all $\gamma\in(0,1)$, \cref{eq:leq-most-apply-transient} is strict for all $\Diid\in\DSetBCiid$ and $\E{s_1\sim \Delta_1}{\pwr[s_1,\gamma][\Dbd]}\not\geqMost[][\DSetBd] \E{s_2\sim \Delta_2}{\pwr[s_2,\gamma][\Dbd]}$.

We show that this result's preconditions holding for  $F_{\Delta_i}^*$ implies the $F_{\Delta_i}$ preconditions. Suppose $F_{\Delta_i}^*\defeq \set{\E{s_i\sim \Delta_i}{\fpi{s_i}}\mid \pi\in\Pi}$ for $i=1,2$ are such that $F_\text{sub}^*\defeq \phi\cdot \ND{F_{\Delta_1}^*}\subseteq F_{\Delta_2}^*$. In the following, the $\Delta_i$ are represented as vectors in $\rewardVS$, and $\gamma$ is a variable.
\begin{align}
    \phi\cdot \set{\gamma\f\mid \f \in \ND{F_{\Delta_1}}}&=\phi\cdot \prn{\ND{F_{\Delta_1}^*-\Delta_1}}\\
    &=\phi\cdot\prn{\ND{F_{\Delta_1}^*}-\Delta_1}\label{eq:invariant-vec-fn-set}\\
    &=\set{\permute \f - \permute\Delta_1\mid \f\in \ND{F_{\Delta_1}^*}}\\
    &\subseteq \set{\f-\Delta_2 \mid \f\in F_{\Delta_2}^*}\label{eq:fsub-no-state}\\
    &=\set{\gamma\f\mid \f \in F_{\Delta_2}}.\label{eq:gamma-sim}
\end{align}

\Cref{eq:invariant-vec-fn-set} follows from \cref{lem:pos-aff-nd-invar} \cref{item:invar-vector-fns}. Since we assumed that $\phi\cdot \ND{F_{\Delta_1}^*}\subseteq F_{\Delta_2}^*$, $\phi\cdot \set{\Delta_1}=\phi\cdot\prn{\ND{F_{\Delta_1}^*}(0)}\subseteq F_{\Delta_2}^*(0) =\set{\Delta_2}$. This implies that $\permute\Delta_1=\Delta_2$ and so \cref{eq:fsub-no-state} follows.

\Cref{eq:gamma-sim} shows that $\phi\cdot \set{\gamma\f\mid \f \in \ND{F_{\Delta_1}}}\subseteq \set{\gamma\f\mid \f \in F_{\Delta_2}}$. But we then have $\phi\cdot \set{\gamma\f\mid \f \in \ND{F_{\Delta_1}}}\defeq \set{\gamma\permute \f \mid \f \in \ND{F_{\Delta_1}}}=\set{\gamma\f \mid \f \in \phi\cdot \ND{F_{\Delta_1}}}\subseteq \set{\gamma\f\mid \f \in F_{\Delta_2}}$. Thus,  $\phi\cdot \ND{F_{\Delta_1}}\subseteq F_{\Delta_2}$.

Suppose $\ND{F_{\Delta_2}^*}\setminus \phi\cdot \ND{F_{\Delta_1}^*}$ is non-empty, which implies that
\begin{align}
    \phi\cdot \set{\gamma\f\mid \f \in \ND{F_{\Delta_1}}}&=\set{\permute \f - \permute\Delta_1\mid \f\in \ND{F_{\Delta_1}^*}}\\
    &=\set{\f - \permute\Delta_1\mid \f\in \phi\cdot \ND{F_{\Delta_1}^*}}\\
    &\subsetneq \set{\f-\Delta_2 \mid \f\in \ND{F_{\Delta_2}^*}}\\
    &=\set{\gamma\f\mid \f \in \ND{F_{\Delta_2}}}.
\end{align}

Then $\ND{F_{\Delta_2}}\setminus \phi\cdot \ND{F_{\Delta_1}}$ must be non-empty. Therefore, if the preconditions of this result are met for $F_{\Delta_i}^*$, they are met for $F_{\Delta_i}$.
\end{proof}

\morePowerMoreOptions*
\begin{proof}
Let $F_\text{sub}\defeq \phi\cdot \Fnd(s')\subseteq \F(s)$. Let $\Delta_1\defeq \unitvec[s'],\Delta_2\defeq \unitvec$, and define $F_{\Delta_i}^*\defeq \set{\E{s_i\sim \Delta_i}{\fpi{s_i}}\mid \pi\in\Pi}$ for $i=1,2$. Then $\Fnd(s')=\ND{F_{\Delta_1}^*}$ is similar to $F_\text{sub}=F^*_\text{sub}\subseteq F_{\Delta_2}^*=\F(s)$ via involution $\phi$. Apply \cref{lem:more-power-prefix} to conclude that
\[\forall \gamma\in[0,1]:\pwr[s',\gamma][\Dbd]\leqMost[][\DSetBd] \pwr[s,\gamma][\Dbd].\]

Furthermore, $\Fnd(s)=\ND{F_{\Delta_2}^*}$, and $F_\text{sub}=F_\text{sub}^*$, and so if $\Fnd(s)\setminus \phi\cdot \Fnd(s')\defeq \Fnd(s)\setminus F_\text{sub} =\ND{F_{\Delta_2}^*}\setminus F_\text{sub}^*$ is non-empty, then \cref{lem:more-power-prefix} shows that for all $\gamma\in(0,1)$, the inequality is strict for all $\Diid\in\DSetBCiid$ and $\pwr[s',\gamma][\Dbd]\not\geqMost[][\DSetBd] \pwr[s,\gamma][\Dbd]$.
\end{proof}

\begin{restatable}[Non-dominated visit distribution functions never agree with other visit distribution functions at that state]{lem}{noAgreeND}\label{lem:no-agree}
Let $\f \in \Fnd(s), \f'\in\F(s)\setminus\{\f\}$. $\forall \gamma \in (0,1):\f(\gamma) \neq \f'(\gamma)$.
\end{restatable}
\begin{proof}
Let $\gamma\in(0,1)$. Since $\f\in\Fnd(s)$, there exists a $\gamma^*\in(0,1)$ at which $\f$ is strictly optimal for some reward function. Then by \cref{transferDiscount}, we can produce another reward function for which $\f$ is strictly optimal at discount rate $\gamma$; in particular, \cref{transferDiscount} guarantees that the policies which induce $\f'$ are not optimal at $\gamma$. So $\f(\gamma)\neq \f'(\gamma)$.
\end{proof}

\begin{restatable}[Cardinality of non-dominated visit distributions]{cor}{cardNDInter}\label{cor:card-nd-visit}
Let $F\subseteq \F(s)$. $\forall \gamma\in(0,1): \abs{F\cap \Fnd(s)}=\abs{F(\gamma)\cap \Fnd(s,\gamma)}$.
\end{restatable}
\begin{proof}
Let $\gamma\in(0,1)$. By applying \cref{lem:nd-gamma-F-subset} with $\Delta_d\defeq \unitvec$, $\f\in\Fnd(s)=\ND{\F(s)}$ iff $\f(\gamma)\in\ND{\F(s,\gamma)}$. By \cref{lem:nd-relation}, $\ND{\F(s,\gamma)}=\Fnd(s,\gamma)$. So all $\f\in F\cap \Fnd(s)$ induce $\f(\gamma)\in F(\gamma)\cap \Fnd(s,\gamma)$, and $\abs{F\cap \Fnd(s)}\geq\abs{F(\gamma)\cap \Fnd(s,\gamma)}$.

\Cref{lem:no-agree} implies that for all $\f,\f'\in\Fnd(s)$, $\f=\f'$ iff $\f(\gamma)=\f'(\gamma)$. Therefore, $\abs{F\cap \Fnd(s)}\leq\abs{F(\gamma)\cap \Fnd(s,\gamma)}$. So $\abs{F\cap \Fnd(s)}=\abs{F(\gamma)\cap \Fnd(s,\gamma)}$.
\end{proof}

\begin{restatable}[Optimality probability and state bottlenecks]{lem}{optProbBottle}\label{lem:opt-prob-bottleneck}
Let $X\defeq \reach{s',a'}\cup\reach{s',a}$. Suppose that $s$ can reach $X$, but only by taking actions equivalent to $a'$ or $a$ at state $s'$. $F_{\text{nd},a'}\defeq \FndRestrictAction{s'}{a'}, F_a\defeq \FRestrictAction{s'}{a}$. Suppose $F_a$ contains a copy of $F_{\text{nd},a'}$ via $\phi$ which fixes all states not belonging to $X$.
Then $\forall \gamma\in[0,1]:\optprob[\Dany]{F_{\text{nd},a'},\gamma}\leqMost[][\DSetAny] \optprob[\Dany]{F_a,\gamma}$.

If $\Fnd(s)\cap \prn{F_a\setminus \phi\cdot F_{\text{nd},a'}}$ is non-empty, then for all $\gamma\in(0,1)$, the inequality is strict for all $\Diid\in\DSetBCiid$, and $\optprob[\Dany]{F_{\text{nd},a'},\gamma}\not\geqMost[][\DSetAny] \optprob[\Dany]{F_a,\gamma}$.
\end{restatable}
\begin{proof} Let $F_\text{sub}\defeq \phi\cdot F_{\text{nd},a'}$. Let $F^* \defeq \bigcup_{\substack{a''\in\A:\\ \prn{a'' \not\equiv_{s'} a} \land \prn{a'' \not \equiv_{s'}a'}}} \FRestrictAction{s'}{a''}\cup F_{\text{nd},a'} \cup F_\text{sub}$.
\begin{align}
    \phi\cdot F^*\defeq\,& \phi\cdot \prn{\bigcup_{\substack{a''\in\A:\\ \prn{a'' \not\equiv_{s'} a} \land \prn{a'' \not \equiv_{s'}a'}}} \FRestrictAction{s'}{a''}\cup F_{\text{nd},a'} \cup F_\text{sub}}\\
    =\,& \bigcup_{\substack{a''\in\A:\\ \prn{a'' \not\equiv_{s'} a} \land \prn{a'' \not \equiv_{s'}a'}}}\phi\cdot \FRestrictAction{s'}{a''}\cup \prn{\phi\cdot F_{\text{nd},a'}} \cup \prn{\phi\cdot F_\text{sub}}\\
    =\,& \bigcup_{\substack{a''\in\A:\\ \prn{a'' \not\equiv_{s'} a} \land \prn{a'' \not \equiv_{s'}a'}}}\phi\cdot \FRestrictAction{s'}{a''}\cup F_\text{sub} \cup F_{\text{nd},a'}\label{eq:invert-Fsub}\\
    =\,& \bigcup_{\substack{a''\in\A:\\ \prn{a'' \not\equiv_{s'} a} \land \prn{a'' \not \equiv_{s'}a'}}} \FRestrictAction{s'}{a''}\cup F_\text{sub}\cup F_{\text{nd},a'}\label{eq:phi-prime-bigcup}\\
    \eqdef\,& F^*.
\end{align}
\Cref{eq:invert-Fsub} follows because the involution $\phi$ ensures that $\phi\cdot F_\text{sub}=F_{\text{nd},a'}$. By assumption, $\phi$ fixes all $s'\not\in \reach{s',a'}\cup\reach{s',a}$. Suppose $\f\in\F (s)\setminus\prn{F_{\text{nd},a'}\cup F_{a}}$. By the bottleneck assumption, $\f$ does not visit states in $\reach{s',a'}\cup\reach{s',a}$. Therefore, $\permute[\phi]\f=\f$, and so \cref{eq:phi-prime-bigcup} follows.

Let $F_Z \defeq \prn{\F(s)\setminus (\FRestrictAction{s}{a'}\cup F_a)} \cup F_{\text{nd},a'} \cup F_a$. By definition, $F_Z\subseteq \F(s)$. Furthermore, $\Fnd(s)=\bigcup_{\substack{a''\in\A}} \FndRestrictAction{s'}{a''}\subseteq \prn{\F(s)\setminus (\FRestrictAction{s}{a'}\cup F_a)} \cup \FndRestrictAction{s}{a'} \cup F_a\eqdef F_Z$, and so $\Fnd(s)\subseteq F_Z$. Note that $F^*=F_Z\setminus (F_a\setminus F_\text{sub})$.

\paragraph*{Case: $\gamma\in(0,1)$.}
\begin{align}
    \optprob[\Dany]{F_{\text{nd},a'},\gamma} &= \phelper{F_{\text{nd},a'}(\gamma)\geq \F(s,\gamma)}[\Dany]\label{eq:opt-prb-id-1}\\
    &\leqMost[][\DSetAny] \phelper{F_{a}(\gamma)\geq \F(s,\gamma)}[\Dany]\label{eq:leq-most-apply}\\
    &= \optprob[\Dany]{F_{\text{nd},a'},\gamma}.\label{eq:opt-prb-id-2}
\end{align}

\Cref{eq:opt-prb-id-1} and \cref{eq:opt-prb-id-2} follow from \cref{lem:opt-prob-id}. \Cref{eq:leq-most-apply} follows by applying \cref{lem:opt-prob-superior} with $A\defeq F_{\text{nd},a'}(\gamma),B'\defeq F_\text{sub}(\gamma),B\defeq F_{a}(\gamma),C\defeq\F(s,\gamma),Z\defeq F_Z(\gamma)$ which satisfies $\ND{C}=\Fnd(s,\gamma)\subseteq F_Z(\gamma)\subseteq \F(s,\gamma)=C$, and involution $\phi$ which satisfies $\phi\cdot F^*(\gamma)=\phi\cdot \prn{Z\setminus\prn{B\setminus B'}}=Z\setminus\prn{B\setminus B'}=F^*(\gamma)$.

Suppose $\Fnd(s)\cap \prn{F_a\setminus F_\text{sub}}$ is non-empty.
\[0<\abs{\Fnd(s)\cap \prn{F_a\setminus F_\text{sub}}}=\abs{\Fnd(s,\gamma)\cap\prn{F_a(\gamma)\setminus F_\text{sub}(\gamma)}}\eqdef\abs{\ND{C}\cap\prn{B\setminus B'}},\]
(with the first equality holding by \cref{cor:card-nd-visit}), and so $\ND{C}\cap\prn{B\setminus B'}$ is non-empty. We also have $B\defeq F_{a}(\gamma)\subseteq \F(s,\gamma)\eqdef C$. Then reapplying \cref{lem:opt-prob-superior}, \cref{eq:leq-most-apply} is strict for all $\Diid\in\DSetBCiid$, and $\optprob[\Dany]{F_{\text{nd},a'},\gamma}\not\geqMost[][\DSetAny] \optprob[\Dany]{F_a,\gamma}$.

\paragraph*{Case: $\gamma=1$, $\gamma=0$.}
\begin{align}
    \optprob[\Dany]{F_{\text{nd},a'},1} &=\lim_{\gamma^*\to 1} \optprob[\Dany]{F_{\text{nd},a'},\gamma^*}\label{eq:lim-1-prob}\\
    &=\lim_{\gamma^*\to 1} \phelper{F_{\text{nd},a'}(\gamma^*)\geq \F(s,\gamma^*)}[\Dany]\label{eq:lim-1-opt-prb-id-1}\\
    &\leqMost[][\DSetAny] \lim_{\gamma^*\to 1} \phelper{F_{a}(\gamma^*)\geq \F(s,\gamma^*)}[\Dany]\label{eq:apply-lim-prob}\\
    &=\lim_{\gamma^*\to 1} \optprob[\Dany]{F_{a},\gamma^*}\label{eq:lim-1-opt-prb-id-2}\\
    &=\optprob[\Dany]{F_{a},1}.\label{eq:lim-1-prob-b}
\end{align}
\Cref{eq:lim-1-prob} and \cref{eq:lim-1-prob-b} hold by \cref{prop:opt-prob-converge}. \Cref{eq:lim-1-opt-prb-id-1} and \cref{eq:lim-1-opt-prb-id-2} follow by \cref{lem:opt-prob-id}. Applying \cref{lem:lim-prob-most} with $\gamma\defeq 1,I\defeq (0,1), F_A\defeq F_{\text{nd},a'}, F_B\defeq F_a, F_C\defeq \F(s)$, $F_Z$ as defined above, and involution $\phi$ (for which $\phi\cdot \prn{F_Z\setminus \prn{F_B\setminus \phi\cdot F_A}}=F_Z\setminus \prn{F_B\setminus \phi\cdot F_A}$),  we conclude that \cref{eq:apply-lim-prob} follows.

The $\gamma=0$ case proceeds similarly to $\gamma=1$.
\end{proof}

\begin{restatable}[Action optimality probability is a special case of visit distribution optimality probability]{lem}{agreeOptProb}\label{lem:agree-opt-prob} $\optprob[\Dany]{s,a,\gamma}=\optprob[\Dany]{\FRestrictAction{s}{a},\gamma}$.
\end{restatable}
\begin{proof}
Let $F_a\defeq \FRestrictAction{s}{a}$. For $\gamma \in (0,1)$,
\begin{align}
    \prob[\Dany]{s,a,\gamma}&\defeq\prob[R \sim \Dany]{\exists \pi^* \in \optPi: \pi^*(s)=a}\\
    &= \prob[\rf \sim \Dany]{\exists \fpi[\pi^*]{s} \in F_a: \fpi[\pi^*]{s}(\gamma)^\top \rf=\max_{\f \in \F(s)}\f(\gamma)^\top \rf}\label{eq:f-action-equiv}\\
    &=\optprob[\Dany]{F_a,\gamma}.\label{eq:f-action-equiv-final}
\end{align}
By \cref{lem:opt-pol-visit-iff}, if $\exists\pi^*\in\optPi:\pi^*(s)=a$, then it induces some optimal $\fpi[\pi^*]{s}\in F_a$. Conversely, if $\fpi[\pi^*]{s}\in F_a$ is optimal at $\gamma\in(0,1)$, then $\pi^*$ chooses optimal actions on the support of $\fpi[\pi^*]{s}(\gamma)$. Let $\pi'$ agree with $\pi^*$ on that support and let $\pi'$ take optimal actions at all other states. Then $\pi'\in\optPi$ and $\pi'(s)=a$. So \cref{eq:f-action-equiv} follows.

Suppose $\gamma=0$ or $\gamma=1$. Consider any sequence $\prn{\gamma_n}_{n=1}^\infty$ converging to $\gamma$, and let $\Dany$ induce probability measure $F$.
\begin{align}
    \optprob[\Dany]{F_a,\gamma}&\defeq \lim_{\gamma^*\to \gamma} \optprob[\Dany]{F_a,\gamma^*}\\
    &=\lim_{\gamma^*\to \gamma}\prob[R \sim \Dany]{\exists \pi^* \in \optPi[R,\gamma^*]: \pi^*(s)=a}\label{eq:inner-eq-act}\\
    &=\lim_{n\to\infty}\prob[R \sim \Dany]{\exists \pi^* \in \optPi[R,\gamma_n]: \pi^*(s)=a}\\
    &=\lim_{n\to\infty}\int_{\rewardSpace} \indic{\exists \pi^* \in \optPi[R,\gamma_n]: \pi^*(s)=a} \dF[R]\\
    &=\int_{\rewardSpace}\lim_{n\to\infty} \indic{\exists \pi^* \in \optPi[R,\gamma_n]: \pi^*(s)=a} \dF[R]\label{eq:dominated-convergence}\\
    &=\int_{\rewardSpace} \indic{\exists \pi^* \in \optPi[R,\gamma]: \pi^*(s)=a} \dF[R]\\
    &\eqdef \prob[\Dany]{s,a,\gamma}.
\end{align}

\Cref{eq:inner-eq-act} follows by \cref{eq:f-action-equiv-final}. for $\gamma^*\in[0,1]$, let $f_{\gamma^*}(R)\defeq \indic{\exists \pi^* \in \optPi[R,\gamma^*]: \pi^*(s)=a}$. For each $R\in\rewardSpace$, \cref{lem:opt-pol-shift-bound} exists $\gamma_x\approx \gamma$ such that for all intermediate $\gamma_x'$ between $\gamma_x$ and $\gamma$, $ \optPi[R,\gamma_x']=\optPi[R,\gamma]$. Since $\gamma_n \to \gamma$, this means that $\prn{f_{\gamma_n}}_{n=1}^\infty$ converges pointwise to $f_\gamma$. Furthermore, $\forall n \in \mathbb{N}, R\in\rewardSpace: \abs{f_{\gamma_n}(R)}\leq 1$ by definition. Therefore, \cref{eq:dominated-convergence} follows by Lebesgue's dominated convergence theorem.
\end{proof}

\graphOptions*
\begin{proof}
Note that by \cref{def:visit}, $F_{a'}(0)=\set{\unitvec}=F_a(0)$. Since $\phi\cdot F_{a'}\subseteq F_a$, in particular we have $\phi\cdot F_{a'}(0)=\set{\permute\unitvec}\subseteq \set{\unitvec}=F_a(0)$, and so $\phi(s)=s$.

\textbf{\Cref{item:power-options}.}
For state probability distribution $\Delta_s\in \Delta(\St)$, let $F^*_{\Delta_s}\defeq \set{\E{s'\sim \Delta_s}{\fpi{s'}}\mid \pi\in\Pi}$. Unless otherwise stated, we treat $\gamma$ as a variable in this item; we apply element-wise vector addition, constant multiplication, and variable multiplication via the conventions outlined in \cref{def:aff-transf-short}.
\begin{align}
    F_{a'} &= \set{\unitvec + \gamma \E{s_{a'}\sim T(s,a')}{\fpi{s_{a'}}}\mid \pi\in\Pi: \pi(s)=a'}\label{eq:visit-defn}\\
    &= \set{\unitvec + \gamma \E{s_{a'}\sim T(s,a')}{\fpi{s_{a'}}}\mid \pi\in\Pi}\label{eq:remove-action-constraint}\\
    &= \unitvec+\gamma F^*_{T(s,a')}.\label{eq:Fstar-contain}
\end{align}

\Cref{eq:visit-defn} follows by \cref{def:visit}, since each $\f\in\F(s)$ has an initial term of $\unitvec$. \Cref{eq:remove-action-constraint} follows because $s\not\in \reach{s,a'}$, and so for all $s_{a'}\in \supp[T(s,a')]$, $\fpi{s_{a'}}$ is unaffected by the choice of action $\pi(s)$. Note that similar reasoning implies that $F_a\subseteq \unitvec+\gamma F^*_{T(s,a)}$ (because \cref{eq:remove-action-constraint} is a containment relation in general).

Since $F_{a'}=\unitvec+\gamma F^*_{T(s,a')}$, if $F_a$ contains a copy of $F_{a'}$ via $\phi$, then $F^*_{T(s,a)}$ contains a copy of $F^*_{T(s,a')}$ via $\phi$. Then $\phi\cdot \ND{F^*_{T(s,a')}}\subseteq \phi\cdot F^*_{T(s,a')}\subseteq F^*_{T(s,a)}$, and so $F^*_{T(s,a)}$ contains a copy of $\ND{F^*_{T(s,a')}}$. Then apply \cref{lem:more-power-prefix} with $\Delta_1\defeq T(s,a')$ and $\Delta_2\defeq T(s,a)$ to conclude that $\forall\gamma\in[0,1]$:
\begin{equation*}
    \E{s_{a'}\sim T(s,a')}{\pwr[s_{a'},\gamma][\Dbd]}\leqMost[][\DSetBd] \E{s_{a}\sim T(s,a)}{\pwr[s_a,\gamma][\Dbd]}.
\end{equation*}

Suppose $\Fnd(s)\cap \prn{F_a\setminus \phi\cdot F_{a'}}$ is non-empty. To apply the second condition of \cref{lem:more-power-prefix}, we want to demonstrate that $\ND{F^*_{T(s,a)}}\setminus \phi\cdot \ND{F^*_{T(s,a')}}$ is also non-empty.

First consider $\f\in\Fnd(s)\cap F_a$. Because $F_a\subseteq \unitvec+\gamma F^*_{T(s,a)}$, we have that $\gamma\inv(\f - \unitvec)\in F^*_{T(s,a)}$. Because $\f \in \Fnd(s)$, by \cref{def:nd}, $\exists \rf \in \rewardVS, \gamma_x\in (0,1)$ such that
\begin{align}
    \f(\gamma_x)^\top\rf  &> \max_{\f'\in\F(s)\setminus \set{\f}}\f'(\gamma_x)^\top \rf.\label{eq:f-nd-def}
\end{align}
Then since $\gamma_x\in (0,1)$,
\begin{align}
    \gamma_x\inv(\f(\gamma_x)-\unitvec)^\top\rf  &> \max_{\f'\in\F(s)\setminus \set{\f}}\gamma_x\inv(\f'(\gamma_x)-\unitvec)^\top \rf\\
    &=\max_{\f'\in\gamma_x\inv\prn{(\F(s)\setminus \set{\f})-\unitvec}}\f'(\gamma_x)^\top \rf\\
    &\geq \max_{\f'\in \gamma_x\inv\prn{(F_a\setminus \set{\f})-\unitvec}}\f'(\gamma_x)^\top \rf\label{eq:Fact-contain}\\
    &= \max_{\f'\in F^*_{T(s,a)}\setminus \set{\gamma_x\inv(\f-\unitvec)}}\f'(\gamma_x)^\top \rf.\label{eq:FTsa}
\end{align}

\Cref{eq:Fact-contain} holds because $F_a\subseteq \F(s)$. By assumption, action $a$ is optimal for $\rf$ at state $s$ and at discount rate $\gamma_x$. \Cref{eq:remove-action-constraint} shows that $F^*_{T(s,a)}$ potentially allows the agent a non-stationary policy choice at $s$, but  non-stationary policies cannot increase optimal value \citep{puterman_markov_2014}. Therefore, \cref{eq:FTsa} holds.

We assumed that $\gamma\inv (\f-\unitvec)\in \gamma\inv(\Fnd(s)-\unitvec)$. Furthermore, since we just showed that $\gamma\inv (\f-\unitvec)\in F^*_{T(s,a)}$ is strictly optimal over the other elements of $F^*_{T(s,a)}$ for reward function $\rf$ at discount rate $\gamma_x\in(0,1)$, we conclude that it is an element of $\ND{F^*_{T(s,a)}}$ by \cref{def:nd-vec-func}. Then we conclude that $\gamma\inv(\Fnd(s)-\unitvec)\cap F^*_{T(s,a)}\subseteq \ND{F^*_{T(s,a)}}$.

We now show that $\ND{F^*_{T(s,a)}}\setminus \phi\cdot \ND{F^*_{T(s,a')}}$ is non-empty.
\begin{align}
    0&<\abs{\Fnd(s)\cap \prn{F_a\setminus \phi\cdot F_{a'}}}\label{eq:nonempty-fnd-cap}\\
    &=\abs{\gamma\inv\prn{\Fnd(s)\cap \prn{F_a\setminus \phi\cdot F_{a'}}-\unitvec}}\label{eq:card-preserve}\\
    &\leq\abs{\gamma\inv\prn{\Fnd(s)-\unitvec}\cap \prn{F^*_{T(s,a)}\setminus\phi\cdot F^*_{T(s,a')}}}\label{eq:Fstar-convert}\\
    &=\abs{\prn{\gamma\inv\prn{\Fnd(s)-\unitvec}\cap F^*_{T(s,a)}}\setminus\phi\cdot F^*_{T(s,a')}}\\
    &\leq\abs{\ND{F^*_{T(s,a)}}\setminus\phi\cdot F^*_{T(s,a')}}\label{eq:contain-nd-fstar}\\
    &\leq\abs{\ND{F^*_{T(s,a)}}\setminus \phi\cdot \ND{F^*_{T(s,a')}}}.\label{eq:nd-containment-trivial}
\end{align}
\Cref{eq:nonempty-fnd-cap} follows by the assumption that $\Fnd(s)\cap \prn{F_a\setminus \phi\cdot F_{a'}}$ is non-empty. Let $\f,\f'\in \Fnd(s)\cap \prn{F_a\setminus \phi\cdot F_{a'}}$ be distinct. Then we must have that for some $\gamma_x\in (0,1)$, $\f(\gamma_x)\neq \f'(\gamma_x)$. This holds iff $\gamma_x\inv(\f(\gamma_x)-\unitvec)\neq \gamma_x\inv(\f'(\gamma_x)-\unitvec)$, and so \cref{eq:card-preserve} holds.

\Cref{eq:Fstar-convert} holds because $F_a\subseteq \unitvec+\gamma F^*_{T(s,a)}$ and $F_a'= \unitvec+\gamma F^*_{T(s,a')}$ by \cref{eq:Fstar-contain}. \Cref{eq:contain-nd-fstar} holds because we showed above that \[\gamma\inv(\Fnd(s)-\unitvec)\cap F^*_{T(s,a)}\subseteq \ND{F^*_{T(s,a)}}.\] \Cref{eq:nd-containment-trivial} holds because $\ND{F^*_{T(s,a')}}\subseteq F^*_{T(s,a')}$ by \cref{def:nd-vec-func}.

Therefore, $\ND{F^*_{T(s,a)}}\setminus \phi\cdot \ND{F^*_{T(s,a')}}$ is non-empty, and so apply the second condition of \cref{lem:more-power-prefix} to conclude that for all $\Diid\in\DSetBCiid$,
\[\forall\gamma\in(0,1):\E{s_{a'}\sim T(s,a')}{\pwr[s_{a'},\gamma][\Diid]}< \E{s_{a}\sim T(s,a)}{\pwr[s_a,\gamma][\Diid]},\] and that
\[\forall\gamma\in(0,1):\E{s_{a'}\sim T(s,a')}{\pwr[s_{a'},\gamma][\Dbd]}\not \geqMost[][\DSetBd] \E{s_{a}\sim T(s,a)}{\pwr[s_a,\gamma][\Dbd]}.\]

\textbf{\Cref{item:opt-prob-options}.}
Let $\phi'(s_x)\defeq \phi(s_x)$ when $s_x\in\reach{s,a'}\cup\reach{s,a}$, and equal $s_x$ otherwise. Since $\phi$ is an involution, so is $\phi'$.
\begin{align}
    \phi'\cdot F_{a'}&\defeq \set{\permute[\phi']\prn{\unitvec + \gamma\E{s_{a'}\sim T(s,a')}{ \fpi{s_{a'}}}}\mid \pi\in \Pi, \pi(s)=a'}\\
    &= \set{\unitvec+\gamma\E{s_{a'}\sim T(s,a')}{\permute[\phi'] \fpi{s_{a'}}}\mid \pi\in \Pi, \pi(s)=a'}\label{eq:fix-unitvec}\\
    &= \set{\permute\unitvec+ \gamma\E{s_{a'}\sim T(s,a')}{\permute[\phi] \fpi{s_{a'}}}\mid \pi\in \Pi, \pi(s)=a'}\label{eq:phi-prime-eq}\\
    &\eqdef \phi\cdot F_{a'}\\
    &\subseteq F_a.\label{eq:sim-subtracted}
\end{align}

\Cref{eq:fix-unitvec} follows because if $s\in\reach{s,a'}\cup\reach{s,a}$, then we already showed that $\phi$ fixes $s$. Otherwise, $\phi'(s)=s$ by definition. \Cref{eq:phi-prime-eq} follows by the definition of $\phi'$ on $\reach{s,a'}\cup\reach{s,a}$ and because $\unitvec=\permute\unitvec$. Next, we assumed that $\phi\cdot F_{a'}\subseteq F_a$, and so \cref{eq:sim-subtracted} holds.

Therefore, $F_a$ contains a copy of $F_{a'}$ via $\phi'$ fixing all $s_x\not\in\reach{s,a'}\cup\reach{s,a}$. Therefore, $F_a$ contains a copy of $F_{\text{nd},a'}\defeq \Fnd(s)\cap F_{a'}$ via the same $\phi'$. Then apply \cref{lem:opt-prob-bottleneck} with $s'\defeq s$ to conclude that $\forall \gamma\in[0,1]:\optprob[\Dany]{F_{a'},\gamma}\leqMost[][\DSetAny] \optprob[\Dany]{F_a,\gamma}$. By \cref{lem:agree-opt-prob}, $\optprob[\Dany]{s,a',\gamma}=\optprob[\Dany]{F_{a'},\gamma}$ and $\optprob[\Dany]{s,a,\gamma}=\optprob[\Dany]{F_a,\gamma}$. Therefore, $\forall \gamma\in[0,1]:\optprob[\Dany]{s,a',\gamma}\leqMost[][\DSetAny] \optprob[\Dany]{s,a,\gamma}$.

If $\Fnd(s)\cap \prn{F_a\setminus \phi\cdot F_{a'}}$ is non-empty, then apply the second condition of \cref{lem:opt-prob-bottleneck} to conclude that for all $\gamma\in(0,1)$, the inequality is strict for all $\Diid\in\DSetBCiid$, and $\optprob[\Dany]{s,a',\gamma}\not\geqMost[][\DSetAny] \optprob[\Dany]{s,a,\gamma}$.
\end{proof}

\subsubsection{When \texorpdfstring{$\gamma=1$}{reward is undiscounted}, optimal policies tend to navigate towards ``larger'' sets of cycles} \label{sec:preservation}

\begin{restatable}[$\pwrNoDist$ identity when $\gamma=1$]{lem}{gammaOnePower}\label{lem:gamma-1-power}
\begin{equation}
\pwr[s,1][\Dbd]=\E{\rf \sim \Dbd}{\max_{\dbf \in \RSD} \dbf^\top \rf}=\E{\rf \sim \Dbd}{\max_{\dbf \in \RSDnd} \dbf^\top \rf}.
\end{equation}
\end{restatable}
\begin{proof}
\begin{align}
    \pwr[s,1][\Dbd]&= \E{\rf\sim\Dbd}{\max_{\fpi{s}\in \F(s)} \lim_{\gamma\to 1} \frac{1-\gamma}{\gamma}\prn{\fpi{s}(\gamma)-\unitvec}^\top \rf} \label{eq:pwr-lim-rsd}\\
 &=\E{\rf \sim \Dbd}{\max_{\dbf \in \RSD} \dbf^\top \rf}\label{eq:def-rsd}\\
 &=\E{\rf \sim \Dbd}{\max_{\dbf \in \RSDnd} \dbf^\top \rf}.\label{eq:nd-restrict-rsd}
\end{align}
\Cref{eq:pwr-lim-rsd} follows by \cref{lem:pwr-limit}. \Cref{eq:def-rsd} follows by the definition of $\RSD$ (\cref{def:rsd}). \Cref{eq:nd-restrict-rsd} follows because for all $\rf \in \rewardVS$, \cref{cor:nd-func-indif} shows that $\max_{\dbf \in \RSD} \dbf^\top \rf=\max_{\dbf \in \ND{\RSD}} \dbf^\top \rf \eqdef\max_{\dbf \in \RSDnd} \dbf^\top \rf$.
\end{proof}

\RSDSimPower*
\begin{proof} Suppose $\RSDnd[s']$ is similar to $D\subseteq\RSD$ via involution $\phi$.
\begin{align}
    \pwr[s',1]&=\E{\rf \sim \Dbd}{\max_{\dbf \in \RSDnd[s']} \dbf^\top \rf}\label{eq:pwr-id-rsd}\\
    &\leqMost[][\DSetBd]\E{\rf \sim \Dbd}{\max_{\dbf \in \RSDnd} \dbf^\top \rf}\label{eq:rsd-func-incr}
    \\
    &=\pwr[s,1][\Dbd]\label{eq:pwr-id-rsd-2}
\end{align}
\Cref{eq:pwr-id-rsd} and \cref{eq:pwr-id-rsd-2} follow from \cref{lem:gamma-1-power}. By applying \cref{lem:expect-superior} with $A\defeq \RSD[s'],B'\defeq D, B\defeq\RSD$ and $g$ the identity function, \cref{eq:rsd-func-incr} follows.

Suppose $\RSDnd\setminus D$ is non-empty. By the same result, \cref{eq:rsd-func-incr} is a strict inequality for all $\Diid\in\DSetBCiid$, and we conclude that $\pwr[s',1][\Dbd]\not \geqMost[][\DSetBd] \pwr[s,1][\Dbd]$.
\end{proof}

\rsdIC*
\begin{proof}
Let $D_\text{sub}\defeq\phi\cdot D'$, where $D_\text{sub}\subseteq D$ by assumption. Let
\begin{equation*}
    X\defeq \{s_i\in \St \mid \max_{\dbf \in D'\cup D} \dbf^\top \unitvec[s_i]>0\}.
\end{equation*}
Define
\begin{equation}
    \phi'(s_i)\defeq
    \begin{cases}
    \phi(s_i) & \text{ if } s_i\in X\label{eq:phi-prime-def}\\
    s_i &\text{ else}.
    \end{cases}
\end{equation}

Since $\phi$ is an involution, $\phi'$ is also an involution. Furthermore, by the definition of $X$, $\phi'\cdot D'=D_\text{sub}$ and $\phi'\cdot D_\text{sub}=D'$ (because we assumed that both equalities hold for $\phi$).

Let $D^*\defeq D'\cup D_\text{sub} \cup \prn{\RSDnd\setminus (D' \cup D)}$.
\begin{align}
    \phi'\cdot D^*&\defeq \phi'\cdot \prn{D'\cup D_\text{sub} \cup \prn{\RSDnd\setminus (D' \cup D)}} \\
    &=\prn{\phi'\cdot D'}\cup \prn{\phi'\cdot D_\text{sub}} \cup \phi'\cdot\prn{\RSDnd\setminus (D' \cup D)} \\
    &= D_\text{sub} \cup D' \cup \prn{\RSDnd\setminus (D' \cup D)}\label{eq:sim-phi-prime-rsd}\\
    &\eqdef D^*.\label{eq:fixing-rsd-perm}
\end{align}

In \cref{eq:sim-phi-prime-rsd}, we know that $\phi'\cdot D'=D_\text{sub}$ and $\phi'\cdot D_\text{sub}=D'$. We just need to show that $\phi'\cdot \prn{\RSDnd\setminus (D' \cup D)}=\RSDnd\setminus (D' \cup D)$.

Suppose $\exists s_i\in X, \dbf' \in \RSDnd\setminus (D' \cup D):\dbf'^\top \unitvec[s_i]>0$. By the definition of $X$, $\exists \dbf \in D' \cup D:\dbf^\top \unitvec[s_i]>0$. Then
\begin{align}
    \dbf^\top \dbf'&=\sum_{j=1}^{\abs{\St}} \dbf^\top (\dbf'\odot\unitvec[s_j])\label{eq:dot-had}\\
    &\geq \dbf^\top (\dbf'\odot\unitvec[s_i])\label{eq:non-neg-rsd-entry}\\
    &=\dbf^\top \prn{(\dbf'^\top \unitvec[s_i])\unitvec[s_i]}\\
    &= (\dbf'^\top \unitvec[s_i]) \cdot (\dbf^\top\unitvec[s_i])\\
    &>0.\label{eq:pos-rsd-agree}
\end{align}

\Cref{eq:dot-had} follows from the definitions of the dot and Hadamard products. \Cref{eq:non-neg-rsd-entry} follows because $\dbf$ and $\dbf'$ have non-negative entries. \Cref{eq:pos-rsd-agree} follows because $\dbf^\top \unitvec[s_i]$ and $\dbf'^\top \unitvec[s_i]$ are both positive. But \cref{eq:pos-rsd-agree} shows that $\dbf^\top\dbf'>0$, contradicting our assumption that $\dbf$ and $\dbf'$ are orthogonal.

Therefore, such an $s_i$ cannot exist, and $X'\defeq\set{s_i' \in \St \mid \max_{\dbf' \in \RSDnd\setminus (D' \cup D)} \dbf'^\top \unitvec[s_i]>0}\subseteq (\St \setminus X)$. By \cref{eq:phi-prime-def}, $\forall s_i'\in X':\phi'(s_i')=s_i'$. Thus, $\phi'\cdot \prn{\RSDnd\setminus (D' \cup D)}=\RSDnd\setminus (D' \cup D)$, and \cref{eq:sim-phi-prime-rsd} follows. We conclude that $\phi'\cdot D^*=D^*$.

Consider $Z\defeq \prn{\RSDnd\setminus (D' \cup D)} \cup D \cup D'$. First, $Z\subseteq \RSD$ by definition. Second, $\RSDnd = \RSDnd\setminus (D' \cup D) \cup (\RSDnd\cap D') \cup (\RSDnd \cap D) \subseteq Z$. Note that $D^*=Z\setminus (D \setminus D_\text{sub})$.
\begin{align}
\avgprob[\Dany]{D'}&= \phelper{D'\geq \RSD}[\Dany]\\
&\leqMost[][\DSetAny] \phelper{D\geq \RSD}[\Dany]\label{eq:leq-most-1}\\
&= \avgprob[\Dany]{D}.
\end{align}

Since $\phi\cdot D'\subseteq D$ and $\ND{D'}\subseteq D'$, $\phi\cdot \ND{D'}\subseteq D$. Then \cref{eq:leq-most-1} holds by applying \cref{lem:opt-prob-superior} with $A\defeq D', B'\defeq D_\text{sub}, B\defeq D, C\defeq \RSD$, and the previously defined $Z$ which we showed satisfies $\ND{C}\subseteq Z\subseteq C$. Furthermore, involution $\phi'$ satisfies $\phi'\cdot  B^* = \phi'\cdot \prn{Z\setminus (B \setminus B')}=Z\setminus (B \setminus B') = B^*$ by \cref{eq:fixing-rsd-perm}.

When $\RSDnd\cap\prn{D\setminus D_\text{sub}}$ is non-empty, since $B'\subseteq C$ by assumption, \cref{lem:opt-prob-superior} also shows that \cref{eq:leq-most-1} is strict for all $\Diid\in\DSetBCiid$, and that $\avgprob[\Dany]{D'}\not\geqMost[][\DSetAny] \avgprob[\Dany]{D}$.
\end{proof}

\begin{restatable}[{\rsd[R]} properties]{prop}{rsdProp}\label{prop:rsd-properties}
Let $\dbf\in\RSD$. $\dbf$ is element-wise non-negative and $\lone{\dbf}=1$.
\end{restatable}
\begin{proof}
$\dbf$ has non-negative elements because it equals the limit of $\lim_{\gamma\to1}(1-\gamma)\f(\gamma)$, whose elements are non-negative by \cref{prop:visit-dist-prop} \cref{item:mono-increase}.
\begin{align}
    \lone{\dbf}&=\lone{\lim_{\gamma\to1}(1-\gamma)\f(\gamma)}\label{eq:lone-rsd}\\
    &=\lim_{\gamma\to 1} (1-\gamma)\lone{\f(\gamma)}\label{eq:lone-norm-rsd}\\
    &= 1.\label{eq:lone-norm-rsd-1}
\end{align}
\Cref{eq:lone-rsd} follows because the definition of {\rsd}s (\cref{def:rsd}) ensures that $\exists \f\in\F(s):\lim_{\gamma\to1}(1-\gamma)\f(\gamma)=\dbf$. \Cref{eq:lone-norm-rsd} follows because $\lone{\cdot}$ is a continuous function. \Cref{eq:lone-norm-rsd-1} follows because $\lone{\f(\gamma)}=\geom$ by \cref{prop:visit-dist-prop} \cref{item:lone-visit}.
\end{proof}

\begin{restatable}[When reachable with probability $1$, {\stateEnd}s induce non-dominated {\rsd}s]{lem}{termNDrsd}\label{lem:term-nd-rsd}
If $\unitvec[s']\in\RSD$, then $\unitvec[s']\in\RSDnd$.
\end{restatable}
\begin{proof}
If $\dbf\in\RSD$ is distinct from $\unitvec[s']$, then $\lone{\dbf}=1$ and $\dbf$ has non-negative entries by \cref{prop:rsd-properties}. Since $\dbf$ is distinct from $\unitvec[s']$, then its entry for index $s'$ must be strictly less than 1: $\dbf^\top\unitvec[s']<1=\unitvec[s']^\top\unitvec[s']$. Therefore, $\unitvec[s']\in\RSD$ is strictly optimal for the \emph{reward function} $\rf\defeq \unitvec[s']$, and so $\unitvec[s']\in\RSDnd$.
\end{proof}

\avgAvoidTerminal*
\begin{proof}
Suppose $\unitvec[s_x],\unitvec[s']\in\RSD$ are distinct. Let $\phi\defeq (s_x \,\,\, s'),D'\defeq \set{\unitvec[s_x]},D\defeq \RSD\setminus\set{\unitvec[s_x]}$. $\phi\cdot D'=\set{\unitvec[s']}\subseteq \RSD\setminus\set{\unitvec[s_x]}\eqdef D$ since $s_x\neq s'$. $D'\cup D=\RSD$ and $\RSDnd\setminus(D'\cup D)=\RSDnd\setminus\RSD=\emptyset$ trivially have pairwise orthogonal vector elements. Then apply \cref{rsdIC} to conclude that $\avgprob[\Dany]{\{\unitvec[s_x]\}}\leqMost[][\DSetAny] \avgprob[\Dany]{\RSD\setminus\{\unitvec[s_x]\}}$.

Suppose there exists another $\unitvec[s'']\in\RSD$. By \cref{lem:term-nd-rsd}, $\unitvec[s'']\in\RSDnd$. Furthermore, since $s''\not \in \set{s',s_x}$ , $\unitvec[s'']\in \prn{\RSD\setminus\set{\unitvec[s_x]}}\setminus \set{\unitvec[s']}=D\setminus \phi\cdot D'$. Therefore, $\unitvec[s'']\in\RSDnd\cap\prn{D\setminus \phi \cdot D'}$. Then apply the second condition of \cref{rsdIC} to conclude that $\avgprob[\Dany]{\{\unitvec[s_x]\}}\not\geqMost[][\DSetBd] \avgprob[\Dany]{\RSD\setminus\{\unitvec[s_x]\}}$.
\end{proof}

\chapter{Parametrically Retargetable Decision-Makers Tend To Seek Power}
\AddLoeChapter{theorems}
\AddLoeChapter{definitions}
\AddLoeChapter{conjectures}
\section{Retargetability over outcome lotteries} \label{sec:outcomes}
Suppose we are interested in $\dimGen$ outcomes. Each outcome could be the visitation of an {\mdp} state, or a trajectory, or the receipt of a physical item. In the card example of \cref{sec:box-explain}, $\dimGen=3$ playing cards. The agent can induce each outcome with probability $1$, so let $\unitvec[o]\in\reals^3$ be the standard basis vector with probability $1$ on outcome $o$ and $0$ elsewhere. Then the agent chooses among outcome lotteries $C_\text{cards}\defeq \set{\unitvec[\lotA],\unitvec[\lotB],\unitvec[\lotC]}$, which we partition into $A_\text{cards}\defeq \set{\unitvec[\lotA]}$ and $B_\text{cards}\defeq \set{\unitvec[\lotB],\unitvec[\lotC]}$.

\begin{restatable}[Outcome lotteries]{definition}{lotteries}
A unit vector $\x \in \genVS$ with non-negative entries is an \emph{outcome lottery}.\footnote{Our results on outcome lotteries hold for generic $\x'\in \genVS$, but we find it conceptually helpful to consider the non-negative unit vector case.}
\end{restatable}

Many decisions are made consequentially: based on the consequences of the decision, on what outcomes are brought about by an act. For example, in a deterministic setting like Pac-Man, a policy induces a trajectory. A reward function and discount rate tuple $(R,\gamma)$ assigns a \emph{return} to each state trajectory $\tau=s_0,s_1,\ldots$: $G(\tau)=\sum_{i=0}^\infty \gamma^i R(s_i)$. The relevant outcome lottery is the discounted visit distribution over future states in the Pac-Man video game, and policies are optimal or not depending on which outcome lottery is induced by the policy.

\begin{restatable}[Optimality indicator function]{definition}{isOptFn}
Let $X,C\subsetneq \genVS$ be finite, and let $\uf\in\genVS$. $\isOpt{X}{C,\uf}$ returns $ 1$ if $\max_{\x\in X} \x^\top \uf \geq \max_{\cv\in C} \cv^\top \uf$, and $0$ otherwise.
\end{restatable}

We consider decision-making procedures which take in a targeting parameter $\uf$. For example, the column headers of \cref{tab:optimal-permutations} show the 6 permutations of the utility function $u(\lotA)\defeq 10, u(\lotB)\defeq 5, u(\lotC)\defeq 0$, representable as a vector $\uf\in \reals^3$.

$\uf$ can be permuted as follows. The outcome permutation $\phi\in\genSym$ inducing an $\dimGen\times \dimGen$ permutation matrix $\permute$ in row representation: $(\permute)_{ij}=1$ if $i=\phi(j)$ and $0$ otherwise. \Cref{tab:optimal-permutations} shows that for a given utility function, $\frac{2}{3}$ of its orbit agrees that $B_\text{cards}$ is strictly optimal over $A_\text{cards}$.

\begin{table}[ht]\centering\footnotesize
    \newlength{\orbitcolone}\setlength{\orbitcolone}{\widthof{$\boltz[1]{\set{\unitvec[\lotA],\unitvec[\lotB]}}{C_\text{cards},\uf'}$}}
    \setlength{\tabcolsep}{3.7pt}
    \caption[Orbit-level incentives across 4 decision-making functions]{Orbit-level incentives across 4 decision-making functions.}\label{tab:examples}
    \vspace{5pt}
    \begin{subtable}{\linewidth}\centering
    \begin{tabular}{@{}>{\raggedleft\arraybackslash}p{\orbitcolone}|GGGcGc}
    \toprule
    \headers \\
    \midrule
    $\isOpt{\set{\unitvec[\lotA],\unitvec[\lotB]}}{C_\text{cards},\uf'}$ & $ 1$ & $ 1$ & $ 1$ & $0$ & $ 1$ & $0$ \\
    $\isOpt{\set{\unitvec[\lotC]}}{C_\text{cards},\uf'}$                 & $0$ & $0$ & $0$ & $ 1$ & $0$ & $ 1$ \\
    \bottomrule
    \end{tabular}
    \caption{Dark gray columns indicate utility function permutations $\uf'$ for which $\isOpt{B_\text{cards}}{C_\text{cards},\uf'}>\isOpt{A_\text{cards}}{C_\text{cards},\uf'}$, while white indicates that the opposite strict inequality holds.}
    \label{tab:optimal-permutations}
    \end{subtable}

    \vspace{\baselineskip}
    \begin{subtable}{\linewidth}\centering
    \begin{tabular}{@{}>{\raggedleft\arraybackslash}p{\orbitcolone}|cGcGGG}
    \toprule
    \headers\\
    \midrule
    $\isAnti{\set{\unitvec[\lotA],\unitvec[\lotB]}}{C_\text{cards},\uf'}$ & $0$ & $ 1$ & $0$ & $ 1$ & $ 1$ & $ 1$ \\
    $\isAnti{\set{\unitvec[\lotC]}}{C_\text{cards},\uf'}$                 & $ 1$ & $0$ & $ 1$ & $0$ & $0$ & $0$ \\
    \bottomrule
    \end{tabular}
    \caption{Utility-minimizing outcome selection probability.}
    \label{tab:anti-rational-permutations}
    \end{subtable}

    \vspace{\baselineskip}
    \begin{subtable}{\linewidth}\centering
    \begin{tabular}{@{}>{\raggedleft\arraybackslash}p{\orbitcolone}|GGGcGc}
    \toprule
    \headers\\
    \midrule
    $\boltz[1]{\set{\unitvec[\lotA],\unitvec[\lotB]}}{C_\text{cards},\uf'}$ & $ 1$ & $ .993$ & $ 1$ & $ .007$ & $ .993$ & $.007$ \\
    $\boltz[1]{\set{\unitvec[\lotC]}}{C_\text{cards},\uf'}$ & $ .000$ & $ .007$ & $ .000$ & $ .993$ & $ .007$ & $.993$\\
    \bottomrule
    \end{tabular}
    \caption{Boltzmann selection probabilities for $T=1$, rounded to three significant digits.}
    \label{tab:boltzmann-permutations}
    \end{subtable}

    \vspace{\baselineskip}
    \begin{subtable}{\linewidth}\centering
    \begin{tabular}{@{}>{\raggedleft\arraybackslash}p{\orbitcolone}|GgGggg}
    \toprule
    \headers\\
    \midrule
    $\satisfice[3]{\set{\unitvec[\lotA],\unitvec[\lotB]}}{C_\text{cards},\uf'}$ & $ 1$ & $ .5$ & $ 1$ & $ .5$ & $ .5$ & $ .5$ \\
    $\satisfice[3]{\set{\unitvec[\lotC]}}{C_\text{cards},\uf'}$ & $0$ & $ .5$ & $0$ & $ .5$ & $ .5$ & $ .5$ \\
    \bottomrule
    \end{tabular}
    \caption{A satisficer uniformly randomly selects an outcome lottery with expected utility greater than or equal to the threshold $t$. Here, $t=3$. When $\satisfice[3]{\set{\unitvec[\lotA],\unitvec[\lotB]}}{C_\text{cards},\uf'}=\satisfice[3]{\set{\unitvec[\lotC]}}{C_\text{cards},\uf'}$, the column is colored medium gray.}
    \label{tab:satisficing-permutations}
    \end{subtable}
\end{table}

Orbit-level incentives occur when an inequality holds for most permuted parameter choices $\uf'$. \Cref{tab:optimal-permutations} demonstrates an application of \citet{turner_optimal_2020}'s results: Optimal decision-making induces orbit-level incentives for choosing outcomes in $B_\text{cards}$ over outcomes in $A_\text{cards}$.

Furthermore, \citet{turner_optimal_2020} conjectured that ``larger'' $B_\text{cards}$ will imply stronger orbit-level tendencies: If going right leads to 500 times as many options as going left, then right is better than left for at least 500 times as many reward functions for which the opposite is true. We prove this conjecture with \cref{thm:rsdIC-quant} in appendix \ref{app:mdp}.

However, orbit-level incentives do not require optimality. One clue is that the same results hold for anti-optimal agents, since anti-optimality/utility minimization of $\uf$ is equivalent to maximizing $-\uf$. \Cref{tab:anti-rational-permutations} illustrates that the same orbit guarantees hold in this case.

\begin{restatable}[Anti-optimality indicator function]{definition}{isAntiFn}
Let $X,C\subsetneq \genVS$ be finite, and let $\uf\in\genVS$. $\isAnti{X}{C,\uf}$ returns $ 1$ if $\min_{\x\in X} \x^\top \uf \leq \min_{\cv\in C} \cv^\top \uf$, and $0$ otherwise.
\end{restatable}

Stepping beyond expected utility maximization/minimization, Boltzmann-rational decision-making selects outcome lotteries proportional to the exponential of their expected utility.

\begin{restatable}[Boltzmann rationality \citep{baker2007goal}]{definition}{boltzdecision}
For $X\subseteq C$ and temperature $T > 0$, let \[\boltz[T]{X}{C,\uf}\defeq \frac{\sum_{\x \in X} e^{T\inv \x^\top \uf}}{\sum_{\cv \in C}e^{T\inv \cv^\top \uf}}\] be the probability that some element of $X$ is Boltzmann-rational.
\end{restatable}

Lastly, orbit-level tendencies occur even under decision-making procedures which partially ignore expected utility and which ``don't optimize too hard.'' Satisficing agents randomly choose an outcome lottery with expected utility exceeding some threshold. \Cref{tab:satisficing-permutations} demonstrates that satisficing induces orbit-level tendencies.

\begin{restatable}[Satisficing]{definition}{satisficedecision}
Let $t\in\reals$, let $X\subseteq C\subsetneq \genVS$ be finite. $\satisfice{X,C}{\uf} \defeq \frac{\abs{X \cap \set{\cv \in C \mid \cv^\top \uf \geq t}}}{\abs{\set{\cv \in C \mid \cv^\top \uf \geq t}}}$ is the fraction of $X$ whose value exceeds threshold $t$.  $\satisfice{X,C}{\uf}$ evaluates to $0$ the denominator equals $0$.
\end{restatable}

For each table, two-thirds of the utility permutations (columns) assign strictly larger values (shaded dark gray) to an element of $B_\text{cards}\defeq \set{\unitvec[\lotB],\unitvec[\lotC]}$ than to an element of $A_\text{cards}\defeq \set{\unitvec[\lotA]}$. For optimal, anti-optimal, Boltzmann-rational, and satisficing agents, \cref{prop:rationalities} proves that these tendencies hold for all targeting parameter orbits.

\subsection{A range of decision-making functions are retargetable}
In {\mdp}s, \citet{turner_optimal_2020} consider \emph{state visitation distributions} which record the total discounted time steps spent in each environment state, given that the agent follows some policy $\pi$ from an initial state $s$. These visitation distributions are one kind of outcome lottery, with $\dimGen=\abs{\St}$ the number of {\mdp} states.

In general, we suppose the agent has an objective function $\uf\in\genVS$ which maps outcomes to real numbers. In \citet{turner_optimal_2020}, $\uf$ was a state-based reward function (and so the outcomes were \emph{states}). However, we need not restrict ourselves to the {\mdp} setting.

To state our key results, we define several technical concepts which we informally used when reasoning about $A_\text{cards}\defeq \set{\unitvec[\lotA]}$ and $B_\text{cards}\defeq \set{\unitvec[\lotB],\unitvec[\lotC]}$.

\begin{restatable}[Similarity of vector sets]{definition}{QuantDefStateDistSimilar}\label{def:quant-dist-sim}
For $\phi\in \genSym$ and $X\subseteq \genVS$, $\phi\cdot X\defeq \set{\permute \x \mid \x \in X}$. $X'\subseteq \rewardVS$ \emph{is similar to $X$} when $\exists \phi: \phi\cdot X'=X$. $\phi$ is an \emph{involution} if $\phi=\phi\inv$ (it either transposes states, or fixes them). $X$ \emph{contains a copy of $X'$} when $X'$ is similar to a subset of $X$ via an involution $\phi$.
\end{restatable}

\begin{restatable}[Containment of set copies]{definition}{citecontainCopies}\label{def:copies}
Let $n$ be a positive integer, and let $A,B\subseteq \genVS$. We say that \emph{$B$ contains $n$ copies of $A$} when there exist involutions $\phi_1,\ldots,\phi_n\in \genSym$ such that $\forall i:\phi_i\cdot A\eqdef B_i \subseteq B$ and $\forall j\neq i:\phi_i \cdot B_j=B_j$.\footnote{Technically, \cref{def:copies} implies that $A$ contains $n$ copies of $A$ holds for all $n$, via $n$ applications of the identity permutation. For our purposes, this provides greater generality, as all of the relevant results still hold. Enforcing pairwise disjointness of the $B_i$ would handle these issues, but would narrow our results to not apply \eg{} when the $B_i$ share a constant vector.}
\end{restatable}

$B_\text{cards}\defeq \set{\unitvec[\lotB],\unitvec[\lotC]}$ contains two copies of $A_\text{cards}\defeq \set{\unitvec[\lotA]}$ via $\phi_1\defeq \lotA \leftrightarrow \lotB$ and $\phi_2 \defeq \lotA \leftrightarrow \lotC$.

\begin{restatable}[Targeting parameter distribution assumptions]{definition}{citedistDefn}\label{def:quant-dist}
Results with $\Dany$ hold for any probability distribution over $\genVS$. Let $\DSetAny\defeq \Delta(\genVS)$. For a function $f:\genVS \mapsto \reals$, we write $f(\Dany)$ as shorthand for $\E{\uf \sim \Dany}{f(\uf)}$.
\end{restatable}

The symmetry group on $\dimGen$ elements, $\genSym$, acts on the set of probability distributions over $\genVS$.

\begin{restatable}[Pushforward distribution of a permutation \citep{turner_optimal_2020}]{definition}{citepushfwdPermDist}\label{def:quant-pushforward-permute}
Let $\phi\in\genSym$. $\phi\cdot\Dany$ is the pushforward distribution induced by applying the random vector $p(\uf)\defeq \permute\uf$ to $\Dany$.
\end{restatable}

\begin{restatable}[Orbit of a probability distribution \citep{turner_optimal_2020}]{definition}{citeorbit}\label{def:restate-orbit}
The \emph{orbit} of $\Dany$ under the symmetric group $\genSym$ is $\genSym\cdot \Dany\defeq \{\phi\cdot\Dany\mid \phi\in\genSym\}$.
\end{restatable}

Because $B_\text{cards}$ contains 2 copies of $A_\text{cards}$, there are ``at least two times as many ways'' for $B$ to be optimal, than for $A$ to be optimal. Similarly, $B$ is ``at least two times as likely'' to contain an anti-rational outcome lottery for generic utility functions. As demonstrated by \cref{tab:examples}, the key idea is that ``larger'' sets (a set $B$ containing several \emph{copies} of set $A$) are more likely to be chosen under a wide range of decision-making criteria.

\begin{restatable}[Orbit incentives for different rationalities]{prop}{differentRationalities}\label{prop:rationalities}
Let $A,B\subseteq C \subsetneq \genVS$ be finite, such that $B$ contains $n$ copies of $A$ via involutions $\phi_i$ such that $\phi_i\cdot C=C$.
\begin{enumerate}
    \item \textbf{Rational choice \citep{turner_optimal_2020}.}\label{item:rational} \[\isOpt{B}{C,\Dany} \geqMost[n] \isOpt{A}{C,\Dany}.\]
    \item \textbf{Uniformly randomly choosing an optimal lottery.}\label{item:frac-rational} For $X\subseteq C$, let \[\fracOpt{X \mid C, \uf}\defeq \frac{\abs{\set{\argmax_{\cv\in C} \cv^\top \uf}\cap X}}{\abs{\set{\argmax_{\cv\in C} \cv^\top \uf}}}.\]
    Then $\fracOpt{B\mid C, \Dany}\geqMost[n] \fracOpt{A\mid C, \Dany}$.
    \item \textbf{Anti-rational choice.} $\isAnti{B}{C,\Dany} \geqMost[n] \isAnti{A}{C,\Dany}$. \label{item:anti-rational}
    \item \textbf{Boltzmann rationality.} \[\boltz{B}{C, \Dany}\geqMost[n] \boltz{A}{C, \Dany}.\]\label{item:boltzmann}
    \item \textbf{Uniformly randomly drawing $k$ outcome lotteries and choosing the best.} For $X\subseteq C$, $\uf \in \genVS$, and $k\geq 1$, let \[\best(X,C\mid \uf)\defeq \E{\av_1,\ldots,\av_k\sim \text{unif}(C)}{\fracOpt{X\cap \{\av_1,\ldots,\av_k\}\mid \{\av_1,\ldots,\av_k\}, \uf}}.\]

    Then $\best(B\mid C, \Dany)\geqMost[n] \best(A\mid C, \Dany)$.\label{item:best-k}
    \item \textbf{Satisficing \citep{simon1956rational}.} $\satisfice{B}{C, \Dany} \geqMost[n] \satisfice{A}{C, \Dany}$.\label{item:satisfice}
    \item \textbf{Quantilizing over outcome lotteries \citep{taylor2016quantilizers}.} Let $\quantDist$ be the uniform probability distribution over $C$.  For $X\subseteq C$, $\uf\in\genVS$, and $q \in (0,1]$, let $Q_{q,\quantDist}(X\mid C, \uf)$ (\cref{def:quantilize-closed}) return the probability that an outcome lottery in $X$ is drawn from the top $q$-quantile of $\quantDist$, sorted by expected utility under $\uf$. Then $Q_{q,\quantDist}(B\mid C, \uf)\geqMost[n][\genVS]Q_{q,\quantDist}(A \mid C, \uf)$.
    \label{item:quantilizer}
\end{enumerate}
\end{restatable}

One highly retargetable class of decision-making functions are those which only account for the expected utilities of available choices.

\begin{restatable}[EU-determined functions]{definition}{EUFnDefn}\label{def:EU-fn}
Let $\powGenVs$ be the power set of $\genVS$, and let $f:\prod_{i=1}^m \powGenVs\times \genVS \to \reals$. $f$ is an \emph{EU-determined function} if there exists a family of functions $\set{g^{\omega_1,\ldots,\omega_m}}$ such that
\begin{equation}
    f(X_1,\ldots,X_m\mid \uf)=g^{|X_1|,\ldots,|X_m|}\prn{\brx{\x_1^\top\uf}_{\x_1\in X_1},\ldots,\brx{\x_m^\top\uf}_{\x_m\in X_m}},
\end{equation}
where $[r_i]$ is the multiset of its elements $r_i$.
\end{restatable}

For example, let $X\subseteq C \subsetneq \genVS$ be finite, and consider utility function $\uf\in\genVS$. A Boltzmann-rational agent is more likely to select outcome lotteries with greater expected utility. Formally, $\boltz{X}{C, \uf}\defeq \sum_{\x \in X} \frac{e^{T\cdot \x^\top \uf}}{\sum_{\cv \in C}e^{T\cdot \cv^\top \uf}}$ depends only on the expected utility of outcome lotteries in $X$, relative to the expected utility of all outcome lotteries in $C$. Therefore, $\mathrm{Boltzmann}_T$ is a function of expected utilities. This is \emph{why} $\mathrm{Boltzmann}_T$ satisfies the $\geqMost[n]$ relation.

\begin{restatable}[Orbit tendencies occur for EU-determined decision-making functions]{thm}{decisionSet}\label{res:decision-making}
Let $A,B, C \subseteq \genVS$ be such that $B$ contains $n$ copies of $A$ via $\phi_i$ such that $\phi_i\cdot C=C$. Let $h: \prod_{i=1}^2 \powGenVs \times \genVS \to \reals$ be an EU-determined function, and let $p(X\mid \uf) \defeq h(X, C \mid \uf)$.  Suppose that $p$ returns a probability of selecting an element of $X$ from $C$. Then $p(B\mid \uf)\geqMost[n][\genVS]p(A\mid \uf)$.
\end{restatable}

The key takeaway is that decision rules determined by expected utility are highly retargetable. By changing the targeting parameter hyperparameter, the decision-making procedure can be flexibly retargeted to choose elements of ``larger'' sets (in terms of set copies via \cref{def:copies}). Less abstractly, for many agent rationalities—ways of making decisions over outcome lotteries—it is generally the case that larger sets will more often be chosen over smaller sets.

For example, consider a Pac-Man playing agent choosing which environmental state cycle it should end up in. \citet{turner_optimal_2020} show that for most reward functions, average-reward maximizing agents will tend to stay alive so that they can reach a wider range of environmental cycles. However, our results show that average-reward \emph{minimizing} agents also exhibit this tendency, as do Boltzmann-rational agents who assign  greater probability to higher-reward cycles. Any EU-based cycle selection method will—for most reward functions—tend to choose cycles which require Pac-Man to stay alive (at first).

\section{Theoretical results}\label{sec:quant-proofs}
\ineqMostQuant*
\begin{restatable}[Limited transitivity of $\geq_\text{most}$]{lem}{transitGeqMostGen}\label{lem:transit-geq-strong}
Let $f_0,f_1,f_2,f_3:\retarget \to \reals$, and suppose $\retarget$ is a subset of a set acted on by $\genSym$. Suppose that $f_1(\rtparam) \geqMost[n][\retarget] f_2(\rtparam)$ and $\forall\rtparam \in \retarget: f_0(\rtparam)\geq f_1(\rtparam)$ and $f_2(\rtparam)\geq f_3(\rtparam)$. Then $f_0(\rtparam) \geqMost[n][\retarget] f_3(\rtparam)$.
\end{restatable}
\begin{proof}
Let $\rtparam\in\retarget$ and let $\orbInsideCond[\rtparam]{f_a>f_b}\defeq \set{\rtparam'\in\orbInside\mid f_a(\rtparam')>f_b(\rtparam')}$.
\begin{align}
    \abs{\orbInsideCond[\rtparam]{f_0>f_3}}&\geq \abs{\orbInsideCond[\rtparam]{f_1>f_2}}\label{eq:0-3-strict}\\
    &\geq n \abs{\orbInsideCond[\rtparam]{f_2>f_1}}\label{eq:assume-n}\\
    &\geq n \abs{\orbInsideCond[\rtparam]{f_3>f_0}}. \label{eq:0-less-3-strict}
\end{align}

For all $\rtparam'\in \orbInsideCond[\rtparam]{f_1>f_2}$,
\begin{equation*}
    f_0(\rtparam')\geq f_1(\rtparam')>f_2(\rtparam')\geq f_3(\rtparam')
\end{equation*}
by assumption, and so
\begin{equation*}
    \orbInsideCond[\rtparam]{f_1>f_2}\subseteq \orbInsideCond[\rtparam]{f_0>f_3}.
\end{equation*}
Therefore, \cref{eq:0-3-strict} follows. By assumption,
\begin{equation*}
    \abs{\orbInsideCond[\rtparam]{f_1>f_2}}\geq n \abs{\orbInsideCond[\rtparam]{f_2>f_1}};
\end{equation*}
\cref{eq:assume-n} follows. For all $\rtparam'\in \orbInsideCond[\rtparam]{f_2>f_1}$, our assumptions on $f_0$ and $f_3$ ensure that
\begin{equation*}
    f_0(\rtparam')\leq f_1(\rtparam')<f_3(\rtparam')\leq f_2(\rtparam'),
\end{equation*}
so
\begin{equation*}
    \orbInsideCond[\rtparam]{f_3>f_0} \subseteq \orbInsideCond[\rtparam]{f_2>f_1}.
\end{equation*}
Then \cref{eq:0-less-3-strict} follows. By \cref{eq:0-less-3-strict}, $f_0(\rtparam) \geqMost[n][\retarget] f_3(\rtparam)$.
\end{proof}

\begin{restatable}[Order inversion for $\geq_\text{most}$]{lem}{orderInvert}\label{lem:order-invert}
Let $f_1,f_2:\retarget \to \reals$, and suppose $\retarget$ is a subset of a set acted on by $\genSym$. Suppose that $f_1(\rtparam) \geqMost[n][\retarget] f_2(\rtparam)$. Then $-f_2(\rtparam) \geqMost[n][\retarget] -f_1(\rtparam)$.
\end{restatable}
\begin{proof} By \cref{def:restate-orbit}, $f_1(\rtparam) \geqMost[n][\retarget] f_2(\rtparam)$ means that
\begin{align}
    \abs{\set{\rtparam' \in\orbInside\mid f_1(\rtparam')>f_2(\rtparam')}}&\geq n \abs{\set{\rtparam'\in\orbInside  \mid f_1(\rtparam')<f_2(\rtparam')}}\\
    \abs{\set{\rtparam' \in\orbInside\mid -f_2(\rtparam')>-f_1(\rtparam')}}&\geq n \abs{\set{\rtparam'\in\orbInside  \mid -f_2(\rtparam')<-f_1(\rtparam')}}.
\end{align}
Then $-f_2(\rtparam) \geqMost[n][\retarget] -f_1(\rtparam)$.
\end{proof}

\begin{remark}
\Cref{lem:frac-orbit-geq} generalizes \citet{turner_optimal_2020}'s \cref{lem:half-orbit-geq}.
\end{remark}

\begin{restatable}[Orbital fraction which agrees on (weak) inequality]{lem}{fracOrbiGen}\label{lem:frac-orbit-geq}
Suppose $f_1,f_2:\retarget \to \reals$ are such that $f_1(\rtparam) \geqMost[n][\retarget] f_2(\rtparam)$. Then for all $\rtparam\in\retarget$, $\frac{\abs{\set{\rtparam' \in \prn{\orbi[\rtparam][\dimGen]}\cap \retarget \mid f_1(\rtparam')\geq f_2(\rtparam')}}}{\abs{\prn{\orbi[\rtparam][\dimGen]}\cap\retarget}}\geq \dfrac{n}{n+1}$.
\end{restatable}
\begin{proof}
All $\rtparam'\in \prn{\orbi[\rtparam][\dimGen]}\cap\retarget$ such that $f_1(\rtparam')= f_2(\rtparam')$ satisfy $f_1(\rtparam')\geq f_2(\rtparam')$. Otherwise, consider the $\rtparam'\in \prn{\orbi[\rtparam][\dimGen]}\cap\retarget$ such that $f_1(\rtparam')\neq f_2(\rtparam')$. By assumption, at least $\frac{n}{n+1}$ of these $\rtparam'$ satisfy $f_1(\rtparam')> f_2(\rtparam')$, in which case $f_1(\rtparam')\geq f_2(\rtparam')$. Then the desired inequality follows.
\end{proof}

\subsection{General results on retargetable functions}
\begin{restatable}[Functions which are increasing under joint permutation]{definition}{invarJointIncreasing}\label{def:joint}
Suppose that $\genSym$ acts on sets $\abDomain_1,\ldots,\abDomain_m$, and let $f:\prod_{i=1}^m \abDomain_i \to \reals$. $f(X_1,\ldots,X_m)$ is \emph{increasing under joint permutation by $P\subseteq \genSym$} when $\forall\phi \in P: f(X_1,\ldots,X_m)\leq f(\phi \cdot X_1,\ldots,\phi \cdot X_m)$. If equality always holds, then $f(X_1,\ldots,X_m)$ is \emph{invariant under joint permutation by $P$}.
\end{restatable}

\begin{restatable}[Expectations of joint-permutation-increasing functions are also joint-permutation-increasing]{lem}{invarExpect}\label{lem:invar-expect}
For $\abDomain$ which is a subset of a set acted on by $\genSym$, let $f:\abDomain \times \genVS \to \reals$ be a bounded function which is measurable on its second argument, and let $P\subseteq \genSym$. Then if $f(X\mid \uf)$ is increasing under joint permutation by $P$, then $f'(X\mid \Dany)\defeq \E{\uf\sim \Dany}{f(X\mid \uf)}$ is increasing under joint permutation by $P$. If $f$ is \emph{invariant} under joint permutation by $P$, then so is $f'$.
\end{restatable}
\begin{proof} Let distribution $\Dany$ have probability measure $F$, and let $\phi\cdot\Dany$ have probability measure $F_\phi$.
\begin{align}
    f\prn{X \mid \Dany}\defeq{}&\E{\uf\sim \Dany}{f(X\mid \uf)}\\
    \defeq{}&\int_{\genVS} f(X\mid \uf) \dF[\uf][F]\\
    \leq {}&\int_{\genVS} f(\phi\cdot X\mid \permute \uf) \dF[\uf][F]\label{eq:permute-inner}\\
    ={}&\int_{\genVS} f(\phi\cdot X\mid \uf') \abs{\det \permute}\dF[\uf'][F_\phi]\label{eq:change-of-variables-gen}\\
    ={}&\int_{\genVS} f(\phi\cdot X\mid \uf') \dF[\uf'][F_\phi]\label{eq:change-of-variables-2-gen}\\
    \eqdef{}& f'\prn{\phi\cdot X \mid \phi\cdot \Dany}.
\end{align}

\Cref{eq:permute-inner} holds by assumption on $f$: $f(X\mid \uf)\leq f(\phi\cdot X\mid \permute \uf)$. Furthermore, $f(\phi\cdot X\mid \cdot)$ is still measurable, and so the inequality holds. \Cref{eq:change-of-variables-gen} follows by the definition of $F_\phi$ (\cref{def:pushforward-permute}) and by substituting $\rf'\defeq \permute\rf$. \Cref{eq:change-of-variables-2-gen} follows from the fact that all permutation matrices have unitary determinant.
\end{proof}

\begin{restatable}[Closure of orbit incentives under increasing functions]{lem}{closureIncrease}\label{res:closure-increase}
Suppose that $\genSym$ acts on sets $\abDomain_1,\ldots,\abDomain_m$ (with $\abDomain_1$ being a poset), and let $P\subseteq \genSym$. Let $f_1,\ldots,f_n:\prod_{i=1}^m \abDomain_i \to \reals$ be increasing under joint permutation by $P$ on input $(X_1,\ldots,X_m)$, and suppose the $f_i$ are order-preserving with respect to $\preceq_{\abDomain_1}$. Let $g:\prod_{j=1}^n \reals \to \reals$ be monotonically increasing on each argument. Then \begin{equation}
f\prn{X_1,\ldots,X_m} \defeq g\prn{f_1\prn{X_1,\ldots,X_m},\ldots,f_n\prn{X_1,\ldots,X_m}}
\end{equation}
is increasing under joint permutation by $P$ and order-preserving with respect to set inclusion on its first argument. Furthermore, if the $f_i$ are \emph{invariant} under joint permutation by $P$, then so is $f$.
\end{restatable}
\begin{proof} Let $\phi \in P$.
\begin{align}
    f\prn{X_1,\ldots,X_m} &\defeq g\prn{f_1\prn{X_1,\ldots,X_m},\ldots,f_n\prn{X_1,\ldots,X_m}}\\
    &\leq g\prn{f_1\prn{\phi\cdot X_1,\ldots,\phi\cdot X_m},\ldots,f_n\prn{\phi\cdot X_1,\ldots,\phi\cdot X_m}}\label{eq:incr-all}\\
    &\eqdef f\prn{\phi\cdot X_1,\ldots,\phi\cdot X_m}.
\end{align}
\Cref{eq:incr-all} follows because we assumed that $f_i\prn{X_1,\ldots,X_m}\leq f_i\prn{\phi\cdot X_1,\ldots,\phi\cdot X_m}$, and because $g$ is monotonically increasing on each argument. If the $f_i$ are all invariant, then \cref{eq:incr-all} is an equality.

Similarly, suppose $X_1'\preceq_{\abDomain_1} X_1$. The $f_i$ are order-preserving on the first argument, and $g$ is monotonically increasing on each argument. Then   $f\prn{X_1',\ldots,X_m}\leq f\prn{X_1,\ldots,X_m}$. This shows that $f$ is order-preserving on its first argument.
\end{proof}
\begin{remark}
$g$ could take the convex combination of its arguments, or multiply two $f_i$ together and add them to a third $f_3$.
\end{remark}

\retargetFnNWays*
\retargetDecisionN*
\begin{proof}
Let $\rtparam\in\retarget$, and let $\phi_i\cdot \orbInsideCond[\rtparam]{A>B}\defeq \set{\phi_i\cdot \rtparam^A\mid \rtparam^A\in \orbInsideCond[\rtparam]{A>B}}$.
\begin{align}
    \abs{\orbInsideCond[\rtparam]{B>A}]}&\geq \abs{\bigcup_{i=1}^n \phi_i\cdot \orbInsideCond[\rtparam]{A>B}}\label{eq:union-pApB}\\
    &=\sum_{i=1}^n\abs{\phi_i\cdot \orbInsideCond[\rtparam]{A>B}}\label{eq:disjoint-pApB}\\
    &=n\abs{\orbInsideCond[\rtparam]{A>B}}.\label{eq:abs-inj}
\end{align}
By \cref{item:retargetable-n} and \cref{item:symmetry-closure-n}, $\phi_i\cdot \phi_i\cdot \orbInsideCond[\rtparam]{A>B}\subseteq \phi_i\cdot \orbInsideCond[\rtparam]{B>A}]$ for all $i$. Therefore, \cref{eq:union-pApB} holds. \Cref{eq:disjoint-pApB} follows by the assumption that parameters are distinct, and so therefore the cosets $\phi_i\cdot \orbInsideCond[\rtparam]{A>B}$ and $\phi_j\cdot \orbInsideCond[\rtparam]{A>B}$ are pairwise disjoint for $i\neq j$. \Cref{eq:abs-inj} follows because each $\phi_i$ acts injectively on orbit elements.

Letting $f_A(\rtparam)\defeq f(A\mid \rtparam)$ and $f_B(\rtparam)\defeq f(B\mid \rtparam)$, the shown inequality satisfies \cref{def:ineq-most-dists-quant}. We conclude that $f(B\mid\rtparam) \geqMost[n][\retarget] f(A\mid\rtparam)$.
\end{proof}

\retargetFn*

\retargetDecision*
\begin{proof}
Given that $f$ is a $(\retarget, A \overset{\text{simple}}{\to} B)$-retargetable function (\cref{def:retargetFn}), we want to show that $f$ is a $(\retarget, A\overset{1}{\to} B)$-retargetable function (\cref{def:retargetFnMulti} when $n=1$). \Cref{def:retargetFnMulti}'s \cref{item:retargetable-n} is true by assumption. Since $\retarget$ is acted on by $\genSym$, $\retarget$ is closed under permutation and so \cref{def:retargetFnMulti}'s \cref{item:symmetry-closure-n} holds. When $n=1$, there are no $i\neq j$, and so \cref{def:retargetFnMulti}'s \cref{item:distinct} is tautologically true.

Then $f$ is a $(\retarget, A\overset{1}{\to} B)$-retargetable function; apply \cref{lem:general-orbit-simple-nonunif}.
\end{proof}

\subsection{Helper results on retargetable functions}

\begin{table}[h!]\centering\setlength{\tabcolsep}{2.3pt}
    \begin{tabular}{@{}r|cccc}
    \toprule
    Targeting parameter $\rtparam$ & $f(\set{\diamnd}\!\mid\! \rtparam)$ & $f(\set{\heart}\!\mid\! \rtparam)$ & $f(\set{\spade} \!\mid\! \rtparam)$ & $f(\set{\heart,\spade}\!\mid\! \rtparam)$\\
    \midrule
    $\rtparam' \defeq 1\unitvec[1]+3\unitvec[2]+2\unitvec[3]$              & $1$& $0$ & $0$ & $0$ \\
    $\phi_1\cdot \rtparam'=\phi_2\cdot\rtparam'' \defeq 3\unitvec[1]+1\unitvec[2]+2\unitvec[3]$ & $0$ & $ 2$ & $ 2$ & $ 2$ \\
    $\phi_2\cdot \rtparam' \defeq 2\unitvec[1]+3\unitvec[2]+1\unitvec[3]$  & $0$ & $ 2$ & $2$ & $ 2$\\
    $\rtparam'' \defeq 2\unitvec[1]+1\unitvec[2]+3\unitvec[3]$             & $ 1$ & $0$ & $0$ & $0$ \\
    $\phi_1\cdot \rtparam'' \defeq 1\unitvec[1]+2\unitvec[2]+3\unitvec[3]$ & $0$ & $ 2$ & $ 2$ & $ 2$ \\
    $\rtparam^\star \defeq 3\unitvec[1]+2\unitvec[2]+1\unitvec[3]$         & $ 1$ & $0$ & $0$ & $0$ \\
    \bottomrule
    \end{tabular}
    \vspace{10pt}
    \caption[The necessity of \cref{lem:general-orbit-simple-nonunif}'s \cref{item:irrel-symm}]{Let $\phi_1 \defeq \diamnd \leftrightarrow \heart, \phi_2 \defeq \diamnd \leftrightarrow \spade$. We tabularly define a function $f$ which meets all requirements of \cref{lem:general-orbit-simple-nonunif}, except for \cref{item:irrel-symm}: letting $j\defeq 2$, $f(B_2^\star\mid \phi_1\cdot \rtparam')=2>0=f(B_2^\star\mid \rtparam')$. Although $f(B\mid \rtparam)\geqMost[1][\orbi[\rtparam][3]]f(A\mid \rtparam)$, it is not true that $f(B\mid \rtparam^*)\geqMost[2][\orbi[\rtparam][3]]f(A\mid \rtparam^*)$. Therefore, \cref{item:irrel-symm} is generally required.}
    \label{tab:counterex-irrel}
\end{table}

\begin{restatable}[Quantitative general orbit lemma]{lem}{orbGenQuantSimpleRF}\label{lem:general-orbit-simple-nonunif}
Let $\retarget$ be a subset of a set acted on by $\genSym$, and let $f:\abDomain \times \retarget \to \reals$. Consider $A,B\in \abDomain$.

For each $\rtparam\in\retarget$, choose involutions $\phi_1,\ldots,\phi_n\in\genSym$. Let $\rtparam^*\in \orbInside$.
\begin{enumerate}
    \item \textbf{Retargetable under parameter permutation.}\label{item:retargetable-lem} There exist $B_i^\star\in\abDomain$ such that if $f(B  \mid \rtparam^*) < f(A  \mid \rtparam^*)$, then $\forall i: f\prn{A  \mid \rtparam^*}\leq f\prn{B^\star_i \mid \phi_i\cdot \rtparam^*}$.
    \item \textbf{$\retarget$ is closed under certain symmetries.}\label{item:symmetry-closure-quant} $f(B  \mid \rtparam^*) < f(A  \mid \rtparam^*) \implies \forall i: \phi_i \cdot \rtparam^* \in \retarget$.
    \item \textbf{$f$ is increasing on certain inputs.}\label{item:incr} $\forall i: f(B_i^\star \mid \rtparam^*)\leq f(B  \mid \rtparam^*)$.
    \item \textbf{Increasing under alternate symmetries.} For $j=1,\ldots,n$ and $i\neq j$, if $f(A\mid \rtparam^*)<f(B\mid \rtparam^*)$, then $f\prn{B_j^\star  \mid \rtparam^*} \leq f\prn{B_j^\star  \mid \phi_i\cdot \rtparam^*}$. \label{item:irrel-symm}
\end{enumerate}
If these conditions hold for all $\rtparam\in\retarget$, then
\begin{equation}
    f(B  \mid \rtparam) \geqMost[n][\retarget] f(A  \mid \rtparam).\label{eq:gen-quant-superior}
\end{equation}
\end{restatable}
\begin{proof}
Let $\rtparam$ and $\rtparam^*$ be as described in the assumptions, and let $i\in \set{1,\ldots,n}$.
\begin{align}
    f(A  \mid \phi_i\cdot \rtparam^*) &= f(A  \mid \phi_i\inv\cdot \rtparam^*) \label{eq:involute-f-general}\\
    &\leq f(B_i^\star \mid \rtparam^*)\label{eq:joint-symm-arg}\\
    &\leq f(B  \mid \rtparam^*)\label{eq:leq-f}\\
    &< f(A  \mid \rtparam^*)\label{eq:assumpt-f-A}\\
    &\leq f(B_i^\star  \mid \phi_i\cdot \rtparam^*)\label{eq:joint-symm-arg-2}\\
    &\leq f(B  \mid \phi_i\cdot \rtparam^*).\label{eq:leq-f-2}
\end{align}
\Cref{eq:involute-f-general} follows because $\phi_i$ is an involution. \Cref{eq:joint-symm-arg} and \cref{eq:joint-symm-arg-2} follow by \cref{item:retargetable-lem}. \Cref{eq:leq-f} and \cref{eq:leq-f-2} follow by \cref{item:incr}. \Cref{eq:assumpt-f-A} holds by assumption on $\rtparam^*$. Then \cref{eq:leq-f-2} shows that for any $i$, $f(A  \mid \phi_i\cdot \rtparam^*)<f(B  \mid \phi_i\cdot \rtparam^*)$, satisfying \cref{def:retargetFnMulti}'s \cref{item:retargetable-n}.

This result's \cref{item:symmetry-closure-quant} satisfies \cref{def:retargetFnMulti}'s \cref{item:symmetry-closure-n}. We now just need to show \cref{def:retargetFnMulti}'s \cref{item:distinct}.

\paragraph*{Disjointness.}  Let $\rtparam',\rtparam'' \in \orbInsideCond[\rtparam]{A>B}$ and let $i\neq j$. Suppose $\phi_i\cdot \rtparam'=\phi_j\cdot \rtparam''$. We want to show that this leads to contradiction.
\begin{align}
    f(A \mid \rtparam'')&\leq f(B_j^\star \mid \phi_j\cdot \rtparam'')\label{eq:apply-involute-gen}\\
    &=f(B_j^\star \mid \phi_i\inv\cdot \rtparam')\label{eq:equal-dists}\\
    &\leq f(B_j^\star \mid \rtparam')\label{eq:apply-involute-gen-2}\\
    &\leq f(B \mid \rtparam')\label{eq:ineq-B-general}\\
    &<f(A \mid \rtparam')\label{eq:assumpt-general}\\
    &\leq f(B_i^\star \mid \phi_i\cdot \rtparam')\label{eq:start-general}\\
    &=f(B_i^\star \mid \phi_j\inv\cdot \rtparam'')\\
    &\leq f(B_i^\star \mid \rtparam'')\\
    &\leq f(B \mid \rtparam'')\\
    &<f(A \mid \rtparam'').\label{eq:final-general}
\end{align}
\Cref{eq:apply-involute-gen} follows by our assumption of \cref{item:retargetable-lem}. \Cref{eq:equal-dists} holds because we assumed that $\phi_j\cdot \rtparam''=\phi_i\cdot \rtparam'$, and the involution ensures that $\phi_i = \phi_i \inv $. \Cref{eq:apply-involute-gen-2} is guaranteed by our assumption of \cref{item:irrel-symm}, given that $\phi_i\inv \cdot \rtparam'=\phi_i\cdot \rtparam' \in \orbInsideCond[\rtparam]{B>A}]$ by the first half of this proof. \Cref{eq:ineq-B-general} follows by our assumption of \cref{item:incr}. \Cref{eq:assumpt-general} follows because we assumed that $\rtparam'\in \orbInsideCond[\rtparam]{A>B}$.

\Cref{eq:start-general} through \cref{eq:final-general} follow by the same reasoning, switching the roles of $\rtparam'$ and $\rtparam''$, and of $i$ and $j$. But then we have demonstrated that a quantity is strictly less than itself, a contradiction. So for all $\rtparam',\rtparam'' \in \orbInsideCond[\rtparam]{A>B}$, when $i\neq j$, $\phi_i\cdot \rtparam'\neq \phi_j\cdot \rtparam''$.

Therefore, we have shown \cref{def:retargetFnMulti}'s \cref{item:distinct}, and so $f$ is a $(\retarget, A\overset{n}{\to} B)$-retargetable function. Apply \cref{thm:retarget-decision-n} in order to conclude that \cref{eq:gen-quant-superior} holds.
\end{proof}

\begin{restatable}[Superset-of-copy containment]{definition}{superCopy}\label{def:super-copies}
Let $A,B\subseteq \genVS$. \emph{$B$ contains $n$ superset-copies $B_i^\star$ of $A$} when there exist involutions $\phi_1,\ldots,\phi_n$ such that $\phi_i\cdot A\subseteq B_i^\star\subseteq B$, and whenever $i\neq j$, $\phi_i \cdot B_j^\star = B_j^\star$.
\end{restatable}

\begin{restatable}[Looser sufficient conditions for orbit-level incentives]{lem}{orbGenQuantSimple}\label{lem:general-orbit-simple}
Suppose that $\retarget$ is a subset of a set acted on by $\genSym$ and is closed under permutation by $\genSym$. Let $A, B \in \abDomain\subseteq \powGenVs$. Suppose that $B$ contains $n$ superset-copies $B_i^\star \in \abDomain$ of $A$ via $\phi_i$. Suppose that $f(X\mid \rtparam)$ is increasing under joint permutation by $\phi_1,\ldots, \phi_n \in \genSym$ for all $X\in\abDomain,\rtparam\in\retarget$, and suppose that $\forall i: \phi_i\cdot A \in \abDomain$. Suppose that $f$ is monotonically increasing on its first argument. Then $f(B\mid \rtparam) \geqMost[n][\retarget] f(A\mid \rtparam).$
\end{restatable}
\begin{proof}
We check the conditions of \cref{lem:general-orbit-simple-nonunif}. Let $\rtparam\in \retarget$, and let $\rtparam^*\in\prn{\orbi[\rtparam][\dimGen]}\cap \retarget$ be an orbit element.
\begin{enumerate}
    \item[\Cref{item:retargetable-lem}.] Holds since $f(A\mid \rtparam^*)\leq f(\phi_i\cdot A \mid \phi_i \cdot \rtparam^*)\leq f(B^\star_i \mid \phi_i\cdot \rtparam^*)$, with the first inequality by assumption of joint increasing under permutation, and the second following from monotonicity (as $\phi_i\cdot A\subseteq B^\star_i$ by superset copy \cref{def:super-copies}).
    \item[\Cref{item:symmetry-closure-quant}.] We have $\forall \rtparam^* \in \prn{\orbi[\rtparam^*][\dimGen]}\cap \retarget: f(B \mid \rtparam^*) < f(A \mid \rtparam^*) \implies \forall i=1,...,n: \phi_i \cdot \rtparam^* \in \retarget$ since $\retarget$ is closed under permutation.
    \item[\Cref{item:incr}.] Holds because we assumed that $f$ is monotonic on its first argument.
    \item[\Cref{item:irrel-symm}.] Holds because $f$ is increasing under joint permutation on \emph{all} of its inputs $X, \rtparam^{'}$, and \cref{def:super-copies} shows that $\phi_i\cdot B^\star_j=B^\star_j$ when $i\neq j$. Combining these two steps of reasoning, for \emph{all} $\rtparam'\in\retarget$, it is true that $f\prn{B_j^\star  \mid \rtparam'} \leq f\prn{\phi_i\cdot B_j^\star  \mid \phi_i\cdot \rtparam'} \leq f\prn{B_j^\star  \mid \phi_i\cdot \rtparam'}$.
\end{enumerate}
Then apply \cref{lem:general-orbit-simple-nonunif}.
\end{proof}

\begin{restatable}[Hiding an argument which is invariant under certain permutations]{lem}{invarSecondPlace}\label{lem:hide-second}
Let $\abDomain_1$, $\abDomain_2$, $\retarget$ be subsets of sets which are acted on by $\genSym$. Let $A\in\abDomain_1$, $C\in\abDomain_2$. Suppose there exist $\phi_1,\ldots,\phi_n \in \genSym$ such that $\phi_i\cdot C = C$. Suppose $h: \abDomain_1\times \abDomain_2 \times \retarget \to \reals$ satisfies $\forall i: h(A,C\mid \rtparam)\leq h(\phi_i\cdot A,\phi_i \cdot C\mid \phi_i \cdot \rtparam)$. For any $X\in\abDomain_1$, let $f(X\mid \rtparam) \defeq h(X, C \mid \rtparam)$. Then $f(A\mid \rtparam)$ is increasing under joint permutation by $\phi_i$.

Furthermore, if $h$ is \emph{invariant} under joint permutation by $\phi_i$, then so is $f$.
\end{restatable}
\begin{proof}
\begin{align}
    f(X\mid \rtparam)&\defeq h(X, C\mid \rtparam)\\
    &\leq h(\phi_i\cdot X, \phi_i \cdot C\mid \phi_i \cdot \rtparam)\label{eq:invar-gen}\\
    &=h(\phi_i\cdot X, C\mid \phi_i \cdot \rtparam)\label{eq:C-invar-gen}\\
    &\eqdef f(\phi_i\cdot X \mid \phi_i\cdot \rtparam).
\end{align}
\Cref{eq:invar-gen} holds by assumption. \Cref{eq:C-invar-gen} follows because we assumed $\phi_i\cdot C = C$. Then $f$ is increasing under joint permutation by the $\phi_i$.

If $h$ is \emph{invariant}, then \cref{eq:invar-gen} is an equality, and so $\forall i: f(X\mid \rtparam)=f(\phi_i\cdot X\mid \phi_i\cdot \rtparam)$.
\end{proof}

\subsubsection{EU-determined functions}

\Cref{lem:card-EU-invar} and \cref{lem:invar-expect} together extend \citet{turner_optimal_2020}'s \cref{lem:helper-perm} beyond functions of $\max_{\x\in X_i}$, to any functions of cardinalities and of expected utilities of set elements.
\EUFnDefn*

\begin{restatable}[EU-determined functions are invariant under joint permutation]{lem}{cardEUInvar}\label{lem:card-EU-invar}
Suppose that $f:\prod_{i=1}^m \powGenVs\times \genVS \to \reals$ is an EU-determined function. Then for any $\phi\in \genSym$ and $X_1,\ldots,X_m, \uf$, we have $f(X_1,\ldots,X_m\mid \uf)=f(\phi\cdot X_1,\ldots,\phi\cdot X_m\mid \phi\cdot \uf)$.
\end{restatable}
\begin{proof}
\begin{align}
    &f(X_1,\ldots,X_m\mid \uf)\\
    &=g^{|X_1|,\ldots,|X_m|}\prn{\brx{\x_1^\top\uf}_{\x_1\in X_1},\ldots,\brx{\x_m^\top\uf}_{\x_m\in X_m}}\\
    &=g^{\abs{\phi\cdot X_1},\ldots,\abs{\phi\cdot X_m}}\prn{\brx{\x_1^\top\uf}_{\x_1\in X_1},\ldots,\brx{\x_m^\top\uf}_{\x_m\in X_m}}\label{eq:preserve-cardinal}\\
    &=g^{\abs{\phi\cdot X_1},\ldots,\abs{\phi\cdot X_m}}\prn{\brx{(\permute\x_{1})^\top(\permute\uf)}_{\x_1\in X_1},\ldots,\brx{(\permute\x_{m})^\top(\permute\uf)}_{\x_m\in X_m}}\label{eq:transpose}\\
    &= f(\phi\cdot X_1,\ldots,\phi\cdot X_m\mid \phi\cdot \uf).
\end{align}
\Cref{eq:preserve-cardinal} holds because permutations $\phi$ act injectively on  $\genVS$. \Cref{eq:transpose} follows because $\mathbf{I}=\permute\inv\permute=\permute^\top\permute$ by the orthogonality of permutation matrices, and $\x^\top\permute^\top=(\permute\x)^\top$, so $\x^\top\uf=\x^\top\permute^\top\permute \uf=(\permute\x)^\top(\permute\uf)$.
\end{proof}

\decisionSet*
\begin{proof}
By assumption, there exists a family of functions $\set{g^{i,|C|}}$ such that for all $X\subseteq \genVS$, $h(X, C\mid \uf)=g^{|X|,|C|}\prn{\brx{\x^\top\uf}_{\x\in X},\brx{\cv^\top\uf}_{\cv\in C}}$. Therefore, \cref{lem:card-EU-invar} shows that $h(A,C\mid \uf)$ is invariant under joint permutation by the $\phi_i$. Letting $\retarget\defeq \genVS$, apply \cref{lem:hide-second} to conclude that $f(X\mid \uf)$ is invariant under joint permutation by the $\phi_i$.

Since $f$ returns a probability of selecting an element of $X$, $f$ obeys the monotonicity probability axiom: If $X'\subseteq X$, then $f(X'\mid \uf)\leq f(X\mid \uf)$. Then $f(B\mid \uf)\geqMost[n][\genVS]f(A\mid \uf)$ by \cref{lem:general-orbit-simple}.
\end{proof}

\subsection{Particular results on retargetable functions}

\begin{restatable}[Quantilization, closed form]{definition}{quantClosed}\label{def:quantilize-closed}
Let the expected utility $q$-quantile threshold be
\begin{equation}
M_{q,\quantDist}(C\mid \uf)\defeq \inf \set{M \in \reals \mid \prob[\x \sim \quantDist]{\x^\top \uf > M}\leq q}.
\end{equation}
Let $C_{>M_{q,\quantDist}(C\mid \uf)}\defeq \set{\cv \in C \mid \cv^\top \uf > M_{q,\quantDist}(C\mid \uf)}$. $C_{=M_{q,\quantDist}(C\mid \uf)}$ is defined similarly. Let $\indic{L(x)}$ be the predicate function returning $ 1$ if $L(x)$ is true and $0$ otherwise. Then for $X\subseteq C$,
\begin{align}
Q_{q,\quantDist}(X\mid C, \uf )\defeq \sum_{\x \in X}\frac{\quantDist(\x)}{q}\prn{\indic{\x \in C_{>M_{q,\quantDist}(C\mid \uf)}} + \frac{\indic{\x \in C_{=M_{q,\quantDist}(C\mid \uf)}}}{\quantDist\prn{C_{=M_{q,\quantDist}(C\mid \uf)}}}\prn{q- \quantDist\prn{C_{>M_{q,\quantDist}(C\mid \uf)}}}},\label{eq:Q-quant-defn}
\end{align}
where the summand is defined to be $0$ if $\quantDist(\x)=0$ and $\x \in C_{=M_{q,\quantDist}(C\mid \uf)}$.
\end{restatable}

\begin{remark} Unlike \citet{taylor2016quantilizers}'s or \citet{careyuseful}'s definitions, \cref{def:quantilize-closed} is written in closed form and requires no arbitrary tie-breaking. Instead, in the case of an expected utility tie on the quantile threshold, \cref{eq:Q-quant-defn} allots probability to outcomes proportional to their probability under the base distribution $\quantDist$.
\end{remark}

Thanks to \cref{res:decision-making}, we straightforwardly prove most items of \cref{prop:rationalities} by just rewriting each decision-making function as an EU-determined function. Most of the proof's length comes from showing that the functions are measurable on $\uf$, which means that the results also apply for distributions over utility functions $\Dany \in \DSetAny$.

\differentRationalities*
\begin{proof}
\textbf{\Cref{item:rational}.} Consider
\begin{align}
    h(X, C \mid \uf)&\defeq \indic{\exists \x \in X:\forall \cv \in C: \x^\top\uf \geq \cv^\top \uf}\\
    &=\min\prn{1,\sum_{\x\in X}\prod_{\cv \in C}\indic{(\x-\cv)^\top\uf \geq 0}}.
\end{align}
Since halfspaces are measurable, each indicator function is measurable on $\uf$. The finite sum of the finite product of measurable functions is also measurable. Since $\min$ is continuous (and therefore measurable), $h(X, C\mid \uf)$ is measurable on $\uf$.

Furthermore, $h$ is an EU-determined function:
\begin{align}
    h(X,C\mid \uf)&=g\prn{\overbrace{\brx{\x^\top \uf}_{\x\in X}}^{V_X}, \overbrace{\brx{\cv^\top \uf}_{\cv\in C}}^{V_C}}\\
    &\defeq \indic{\exists v_x \in V_X:\forall v_c\in V_C: v_x\geq v_c}.
\end{align}
Then by \cref{lem:card-EU-invar}, $h$ is invariant to joint permutation by the $\phi_i$. Since $\phi_i\cdot C=C$, \cref{lem:hide-second} shows that $h'(X\mid \uf)\defeq h(X,C\mid \uf)$ is also invariant under joint permutation by the $\phi_i$. Since $h$ is a measurable function of $\uf$, so is $h'$. Then since $h'$ is bounded, \cref{lem:invar-expect} shows that $f(X\mid \Dany)\defeq \E{\uf \sim \Dany}{h'(X\mid \uf)}$ is invariant under joint permutation by $\phi_i$.

Furthermore, if $X'\subseteq X$, $f(X'\mid \Dany) \leq f(X \mid\Dany)$ by the monotonicity of probability. Then by \cref{lem:general-orbit-simple},
\begin{equation*}
    f(B\mid \Dany) \defeq \isOpt{B}{C,\Dany} \geqMost[n] \isOpt{A}{C,\Dany} \eqdef f(A\mid \Dany).
\end{equation*}

\textbf{\Cref{item:frac-rational}.} Because $X,C$ are finite sets, the denominator of $\fracOpt{X\mid C,\uf}$ is never zero, and so the function is well-defined.  $\fracOpt{X\mid C,\uf}$ is an EU-determined function:
\begin{align}
    \fracOpt{X\mid C,\uf}&=g\prn{\overbrace{\brx{\x^\top \uf}_{\x\in X}}^{V_X}, \overbrace{\brx{\cv^\top \uf}_{\cv\in C}}^{V_C}}\\
    &\defeq\frac{\abs{\brx{v\in V_X \mid v=\max_{v' \in V_C} v'}}}{\abs{\brx{\argmax_{v' \in V_C} v'}}},
\end{align}
with the $\brx{\cdot}$ denoting a multiset which allows and counts duplicates. Then by \cref{lem:card-EU-invar}, $\fracOpt{X\mid C,\uf}$ is invariant to joint permutation by the $\phi_i$.

We now show that $\fracOpt{X\mid C,\uf}$ is a measurable function of $\uf$.
\begin{align}
   \fracOpt{X\mid C,\uf}&\defeq \frac{\abs{\set{\argmax_{\cv'\in C} \cv'^\top \uf}\cap X}}{\abs{\set{\argmax_{\cv'\in C} \cv'^\top \uf}}}\\
    &= \frac{\sum_{\x \in X} \indic{\x\in \argmax_{\cv'\in C} \cv'^\top\uf}}{\sum_{\cv\in C} \indic{\cv\in \argmax_{\cv'\in C} \cv'^\top\uf}}\\
    &= \frac{\sum_{\x \in X} \prod_{\cv'\in C} \indic{\prn{\x-\cv'}^\top\uf\geq 0}}{\sum_{\cv\in C} \prod_{\cv'\in C} \indic{\prn{\cv-\cv'}^\top\uf\geq 0}}.\label{eq:prod-indic-frac}
\end{align}
\Cref{eq:prod-indic-frac} holds because $\x$ belongs to the $\argmax$ iff $\forall \cv \in C:\x^\top \uf \geq \cv^\top \uf$. Furthermore, this condition is met iff $\uf$ belongs to the intersection of finitely many closed halfspaces; therefore, $\set{\uf\in \genVS \mid \prod_{\cv \in C} \indic{\prn{\x-\cv}^\top\uf\geq 0}=1}$ is measurable. Then the sums in both the numerator and denominator are both measurable functions of $\uf$, and the denominator cannot vanish. Therefore, $\fracOpt{X\mid C,\uf}$ is a measurable function of $\uf$.

Let $g(X\mid \uf) \defeq \fracOpt{X\mid C,\uf}$. Since $\phi_i\cdot C = C$, \cref{lem:hide-second} shows that $g(X\mid \uf)$ is also invariant to joint permutation by $\phi_i$. Since $g$ is measurable and bounded $[0,1]$, apply \cref{lem:invar-expect} to conclude that $f(X\mid \Dany)\defeq \E{\uf \sim \Dany}{g(X\mid C, \uf)}$ is also invariant to joint permutation by $\phi_i$.

Furthermore, if $X'\subseteq X\subseteq C$, then $f(X'\mid \Dany)\leq f(X\mid \Dany)$. So apply \cref{lem:general-orbit-simple} to conclude that $\fracOpt{B\mid C,\Dany}\eqdef f(B\mid \Dany)\geqMost[n] f(A\mid \Dany) \defeq \fracOpt{A\mid C,\Dany}$.

\textbf{\Cref{item:anti-rational}.} Apply the reasoning in \cref{item:rational} with inner function $h(X\mid C, \uf)\defeq \indic{\exists \x \in X:\forall \cv \in C: \x^\top\uf \leq \cv^\top \uf}$.

\textbf{\Cref{item:boltzmann}.} Let $X\subseteq C$. $\boltz{X}{C, \uf}$ is the expectation of an EU function:
\begin{align}
    \boltz{X}{C, \uf}&=g_T\prn{\overbrace{\brx{\x^\top \uf}_{\x\in X}}^{V_X}, \overbrace{\brx{\cv^\top \uf}_{\cv\in C}}^{V_C}}\\
    &\defeq  \frac{\sum_{v \in V_X} e^{v/T}}{\sum_{v \in V_C}e^{v/T}}.\label{eq:meas-boltz}
\end{align}
Therefore, by \cref{lem:card-EU-invar}, $\boltz{X}{C, \uf}$ is invariant to joint permutation by the $\phi_i$.

Inspecting \cref{eq:meas-boltz}, we see that $g$ is continuous on $\uf$ (and therefore measurable), and bounded $[0,1]$ since $X\subseteq C$ and the exponential function is positive. Therefore, by  \cref{lem:invar-expect}, the expectation version is also invariant to joint permutation for all permutations $\phi\in\genSym$: $\boltz{X}{C, \Dany}=\boltz{\phi\cdot X}{\phi\cdot C, \phi\cdot \Dany}$.

Since $\phi_i\cdot C=C$, \cref{lem:hide-second} shows that $f(X\mid \Dany)\defeq \boltz{X}{C, \Dany}$ is also invariant under joint permutation by the $\phi_i$. Furthermore, if $X'\subseteq X$, then $f(X'\mid \Dany) \leq f(X\mid \Dany)$. Then apply \cref{lem:general-orbit-simple} to conclude that $\boltz{B}{C, \Dany}\eqdef f(B\mid \Dany) \geqMost[n] f(A\mid \Dany) \defeq \boltz{A}{C, \Dany}$.

\textbf{\Cref{item:best-k}.} Let involution $\phi\in \genSym$ fix $C$ (\ie{} $\phi\cdot C=C$).
\begin{align}
    &\best(X\mid C, \uf)\\
    &\defeq\!\! \E{\av_1,\ldots,\av_k\sim \text{unif}(C)}{\fracOpt{X\cap \{\av_1,\ldots,\av_k\}\mid \{\av_1,\ldots,\av_k\}, \uf}}\\
    &=\!\! \E{\av_1,\ldots,\av_k\sim \text{unif}(C)}{\fracOpt{(\phi\cdot X)\cap \{\phi\cdot\av_1,\ldots,\phi\cdot\av_k\}\!\mid\! \{\phi\cdot\av_1,\ldots,\phi\cdot\av_k\}, \phi\cdot \uf}}\label{eq:permute-frac}\\
    &=\!\!\E{\phi\cdot\av_1,\ldots,\phi\cdot\av_k\sim \text{unif}(\phi\cdot C)}{\fracOpt{(\phi\cdot X)\cap \{\phi\cdot\av_1,\ldots,\phi\cdot\av_k\}\!\mid\! \{\phi\cdot\av_1,\ldots,\phi\cdot\av_k\}, \phi\cdot \uf}}\label{eq:phi-unif}\\
    &\eqdef \best(\phi\cdot X\mid \phi\cdot C, \phi\cdot\uf).
\end{align}
By the proof of \cref{item:frac-rational},
\begin{multline*}
    \fracOpt{X\cap \{\av_1,\ldots,\av_k\}\mid \{\av_1,\ldots,\av_k\}, \uf} =\\
    \fracOpt{(\phi\cdot X)\cap \{\phi\cdot\av_1,\ldots,\phi\cdot\av_k\}\mid \{\phi\cdot\av_1,\ldots,\phi\cdot\av_k\}, \phi\cdot \uf};
\end{multline*}
thus, \cref{eq:permute-frac} holds. Since $\phi\cdot C=C$ and since the distribution is uniform, \cref{eq:phi-unif} holds. Therefore, $\best(X\mid C, \uf)$ is invariant to joint permutation by the $\phi_i$, which are involutions fixing $C$.

We now show that $\best(X\mid C, \uf)$ is measurable on $\uf$.
\begin{align}
    &\best(X\mid C, \uf)\\
    &\defeq \E{\av_1,\ldots,\av_k\sim \text{unif}(C)}{\fracOpt{X\cap \{\av_1,\ldots,\av_k\}\mid \{\av_1,\ldots,\av_k\}, \uf}}\\
    &=\frac{1}{\abs{C}^k}\sum_{\prn{\av_1,\ldots,\av_k} \in C^k} \fracOpt{X\cap \{\av_1,\ldots,\av_k\}\mid \{\av_1,\ldots,\av_k\}, \uf}.\label{eq:finite-sum-measure}
\end{align}
\Cref{eq:finite-sum-measure} holds because $\fracOpt{X\mid C,\uf}$ is measurable on $\uf$ by \cref{item:frac-rational}, and measurable functions are closed under finite addition and scalar multiplication. Then $\best(X\mid C, \uf)$ is measurable on $\uf$.

Let $g(X\mid \uf) \defeq \best(X\mid C, \uf)$. Since $\phi_i\cdot C = C$, \cref{lem:hide-second} shows that $g(X\mid \uf)$ is also invariant to joint permutation by $\phi_i$. Since $g$ is measurable and bounded $[0,1]$, apply \cref{lem:invar-expect} to conclude that $f(X\mid \Dany)\defeq \E{\uf \sim \Dany}{g(X\mid C, \uf)}$ is also invariant to joint permutation by $\phi_i$.

Furthermore, if $X'\subseteq X\subseteq C$, then $f(X'\mid \Dany)\leq f(X\mid \Dany)$. So apply \cref{lem:general-orbit-simple} to conclude that $\best(B\mid C, \Dany)\eqdef f(B\mid \Dany)\geqMost[n] f(A\mid \Dany) \defeq \best(A\mid C, \Dany)$.

\textbf{\Cref{item:satisfice}.} $\satisfice{X}{C, \uf}$ is an EU-determined function:
\begin{align}
    \satisfice{X}{C, \uf}&=g_t\prn{\overbrace{\brx{\x^\top \uf}_{\x\in X}}^{V_X}, \overbrace{\brx{\cv^\top \uf}_{\cv\in C}}^{V_C}} \\
    &\defeq \frac{\sum_{v\in V_X} \indic{v \geq t}}{\sum_{v\in V_C} \indic{v \geq t}},
\end{align}
with the function evaluating to $0$ if the denominator is $0$.\\
Then applying \cref{lem:card-EU-invar}, $\satisfice{X}{C, \uf}$ is invariant under joint permutation by the~$\phi_i$.

We now show that $\satisfice{X}{C, \uf}$ is measurable on $\uf$.
\begin{align}
    \satisfice{X}{C, \uf}&=
    \begin{cases} \frac{\sum_{\x\in X} \indic{\x\in \set{\x'\in\genVS \mid \x'^\top\uf \geq t}}}{\sum_{\cv\in C} \indic{\cv\in \set{\x'\in\genVS \mid \x'^\top\uf \geq t}}} & \exists \cv \in C: \cv^\top \uf \geq t,\\
    0 &\text{ else}.
    \end{cases}\label{eq:indic-measurable}
\end{align}
Consider the two cases. \[\exists \cv \in C: \cv^\top \uf \geq t \iff \uf \in \bigcup_{\cv \in C}\set{\uf'\in\genVS \mid \cv^\top\uf\geq t}.\]
The right-hand set is the union of finitely many halfspaces (which are measurable), and so the right-hand set is also measurable. Then the casing is a measurable function of $\uf$. Clearly the zero function is measurable. Now we turn to the first case.

In the first case, \cref{eq:indic-measurable}'s indicator functions test each $\x,\cv$ for membership in a closed halfspace with respect to $\uf$. Halfspaces are measurable sets. Therefore, the indicator function is a measurable function of $\uf$, and so are the finite sums. Since the denominator does not vanish within the case, the first case as a whole is a measurable function of $\uf$. Therefore, $\satisfice{X}{C, \uf}$ is measurable on $\uf$.

Since $\satisfice{X}{C, \uf}$ is measurable and bounded $[0,1]$ (as $X\subseteq C$), apply \cref{lem:invar-expect} to conclude that $\satisfice{X}{C, \Dany}= \satisfice{\phi\cdot X}{\phi\cdot C, \phi\cdot \Dany}$. Next, let $f(X\mid \Dany)\defeq \satisfice{X}{C, \Dany}$. Since we just showed that $\satisfice{X}{C, \Dany}$ is invariant to joint permutation by the involutions $\phi_i$ and since $\phi_i\cdot C = C$, $f(X\mid \Dany)$ is also invariant to joint permutation by $\phi_i$.

Furthermore, if $X'\subseteq X$, we have $f(X'\mid \Dany)\leq f(X\mid \Dany)$. Then applying \cref{lem:general-orbit-simple}, $\satisfice{B}{C, \uf}\eqdef f(B\mid \Dany) \geqMost[n] f(A\mid \Dany)\defeq \satisfice{A}{C, \uf}$.

\textbf{\Cref{item:quantilizer}.} Suppose $\quantDist$ is uniform over $C$ and consider any of the involutions $\phi_i$.
\begin{align}
    M_{q,\quantDist}(C\mid \uf)&\defeq \inf \set{M \in \reals \mid \prob[\x \sim \quantDist]{\x^\top \uf > M}\leq q}\\
    &= \inf \set{M \in \reals \mid \prob[\x \sim \quantDist]{(\permute[\phi_i] \x)^\top (\permute[\phi_i]\uf) > M}\leq q}\label{eq:permute-threshold}\\
    &=\inf \set{M \in \reals \mid \prob[\x \sim \phi_i \cdot \quantDist]{ \x^\top (\permute[\phi_i]\uf) > M}\leq q}\\
    &=\inf \set{M \in \reals \mid \prob[\x \sim \quantDist]{ \x^\top (\permute[\phi_i]\uf) > M}\leq q}\label{eq:uniform-threshold}\\
    &\eqdef M_{q,\quantDist}(\phi_i \cdot C\mid \phi_i \cdot \uf).\label{eq:threshold-invar}
\end{align}
\Cref{eq:permute-threshold} follows by the orthogonality of permutation matrices. \Cref{eq:uniform-threshold} follows because if $\x \in \supp[\quantDist]=C$, then $\phi_i \cdot \x \in C=\supp[\quantDist]$, and furthermore $\quantDist(\x)=\quantDist(\permute[\phi_i]\x)$ by uniformity.

Now we show the invariance of $C_{>M_{q,\quantDist}(C\mid \uf)}$ under joint permutation by $\phi_i$:
\begin{align}
    C_{>M_{q,\quantDist}(C\mid \uf)}&\defeq \set{\cv \in C \mid \cv^\top \uf > M_{q,\quantDist}(C\mid \uf)}\\
    &= \set{\cv \in C \mid (\permute[\phi_i]\cv)^\top (\permute[\phi_i]\uf) > M_{q,\quantDist}(\phi_i \cdot C\mid \phi_i\cdot \uf)}\label{eq:invar-C-strict}\\
    &=\set{\cv \in \phi_i \cdot C \mid \cv^\top (\permute[\phi_i]\uf) > M_{q,\quantDist}(\phi_i \cdot C\mid \phi_i\cdot \uf)}\\
    &\eqdef C_{>M_{q,\quantDist}(\phi_i\cdot C\mid \phi_i\cdot \uf)}.
\end{align}
\Cref{eq:invar-C-strict} follows by the orthogonality of permutation matrices and because $M_{q,\quantDist}(C\mid \uf)=M_{q,\quantDist}(\phi_i \cdot C\mid \phi_i\cdot \uf)$ by \cref{eq:threshold-invar}. A similar proof shows that $C_{=M_{q,\quantDist}(C\mid \uf)}=C_{=M_{q,\quantDist}(\phi_i\cdot C\mid \phi_i\cdot \uf)}$.

Recall that
\begin{align}
Q_{q,\quantDist}(X\mid C, \uf )\defeq \sum_{\x \in X}\frac{\quantDist(\x)}{q}\prn{\indic{\x \in C_{>M_{q,\quantDist}(C\mid \uf)}} + \frac{\indic{\x \in C_{=M_{q,\quantDist}(C\mid \uf)}}}{\quantDist\prn{C_{=M_{q,\quantDist}(C\mid \uf)}}}\prn{q- \quantDist\prn{C_{>M_{q,\quantDist}(C\mid \uf)}}}}.\label{eq:Q-restate}
\end{align}
$Q_{q,\quantDist}(X\mid C, \uf)=Q_{q,\quantDist}(\phi_i\cdot X\mid \phi_i\cdot C,\phi_i\cdot \uf)$, since $Q$ is the sum of products of $\phi_i$-invariant quantities.

$\quantDist(\x)$ is non-negative because $\quantDist$ is a probability distribution, and $q$ is assumed positive. The indicator functions $\indic{}$ are non-negative. By the definition of $M_{q,\quantDist}$, $\quantDist\prn{C_{>M_{q,\quantDist}(C\mid \uf)}}\leq q$. Therefore, \cref{eq:Q-restate} is the sum of non-negative terms. Thus, if $X'\subseteq X$, then $Q_{q,\quantDist}(X'\mid C, \uf)\leq Q_{q,\quantDist}(X\mid C, \uf)$.

Let $f(X\mid \uf)\defeq Q_{q,\quantDist}(X\mid C, \uf)$. Since $\phi_i\cdot C=C$ and since $Q_{q,\quantDist}(X\mid C, \uf)=Q_{q,\quantDist}(\phi_i\cdot X\mid\phi_i\cdot C, \phi_i\cdot \uf)$, \cref{lem:hide-second} shows that $f(X\mid \uf)$ is also jointly invariant to permutation by $\phi_i$. Lastly, if $X'\subseteq X$, we have $f(X'\mid \Dany)\leq f(X\mid \Dany)$.

Apply \cref{lem:general-orbit-simple} to conclude that $Q_{q,\quantDist}(B\mid C, \uf)\eqdef f(B\mid \uf) \geqMost[n][\genVS] f(A\mid \uf) \defeq Q_{q,\quantDist}(A\mid C, \uf)$.
\end{proof}

\begin{restatable}[Orbit tendencies occur for more quantilizer base distributions]{conjSec}{conjQuant}\label{conj:quant}
\Cref{prop:rationalities}'s \cref{item:quantilizer} holds for any base distribution $\quantDist$ over $C$ such that $\min_{\bv \in B} \quantDist(\bv)\geq \max_{\av\in A} \quantDist(\av)$. Furthermore, $Q_{q,\quantDist}(X\mid C, \uf)$ is measurable on $\uf$ and so $\geqMost[n][\genVS]$ can be generalized to $\geqMost[n]$.
\end{restatable}
\section{Detailed analyses of {\mr} scenarios}\label{app:mr}
\subsection{Action selection}\label{app:bandit}
Consider a bandit problem with five arms $a_1,\ldots,a_5$ partitioned $A\defeq \set{a_1},B\defeq \set{a_2,\ldots,a_5}$, which each action has a definite utility $\uf_i$. There are $T=100$ trials. Suppose the training procedure $\train$ uses the $\epsilon$-greedy strategy to learn value estimates for each arm. At the end of training, $\train$ outputs a greedy policy with respect to its value estimates.  Consider any action-value initialization, and the learning rate is set $\alpha\defeq 1$. To learn an optimal policy, at worst, the agent just has to try each action once.

\begin{restatable}[Lower bound on success probability of the $\train$ bandit]{lem}{trainUB}\label{lem:train-ub}
Let $\uf\in\reals^5$ assign strictly maximal utility to $a_i$, and suppose $\train$ (described above) runs for $T\geq 5$ trials. Then $f_{\train}(\set{a_i}\mid\uf)\geq 1-(1-\frac{\epsilon}{4})^{T}$.
\end{restatable}
\begin{proof}
Since the trained policy can be stochastic, \[f_{\train}(\set{a_i}\mid\uf)\geq \prob{a_i\text{ is assigned probability $1$ by the learned greedy policy}}.\]

Since $a_i$ has strictly maximal utility which is deterministic, and since the learning rate $\alpha\defeq 1$, if action $a_i$ is ever drawn, it is assigned probability $1$ by the learned policy. The probability that $a_i$ is never explored is at most $(1-\frac{\epsilon}{4})^{T}$, because at worst, $a_i$ is an ``explore'' action (and not an ``exploit'' action) at every time step, in which case it is ignored with probability $1-\frac{\epsilon}{4}$.
\end{proof}

\begin{restatable}[The $\train$ bandit is 4-retargetable]{prop}{retargetBandit}\label{res:retarget-bandit}
$f_\train$ is $(\reals^5,A\overset{4}{\to}B)$-retargetable.
\end{restatable}
\begin{proof}
Let $\phi_i\defeq a_1 \leftrightarrow a_i$ for $i=2,\ldots,5$ and let $\retarget\defeq \reals^5$. We want to show that whenever $\uf\in\reals^5$ induces $f_\train(A\mid \uf)> f_\train(B\mid \uf)$, retargeting $\uf$ will get $\train$ to instead learn to pull a $B$-action: $f_\train(A\mid \phi_i\cdot \uf)< f_\train(B\mid \phi_i\cdot \uf)$.

Suppose we have such a $\uf$. If $\uf$ is constant, a symmetry argument shows that each action has equal probability of being selected, in which case $f_\train(A\mid \uf)=\frac{1}{5}<\frac{4}{5}=f_\train(B\mid \uf)$—a contradiction. Therefore, $\uf$ is not constant. Similar symmetry arguments show that $A$'s action $a_1$ has strictly maximal utility ($\uf_1>\max_{i=2,\ldots,5} \uf_i$).

But for $T=100$, \cref{lem:train-ub} shows that $f_\train(A\mid \uf)=f_\train(\set{a_1}\mid \uf)\approx 1$ and  $f_\train(\set{a_{i\neq 1}}\mid \uf)\approx 0\implies f_\train(B\mid \uf)=\sum_{i\neq 1}f_\train(\set{a_{i}}\mid \uf)\approx 0$. The converse statement holds when considering $\phi_i\cdot \uf$ instead of $\uf$. Therefore, $\train$ satisfies \cref{def:retargetFnMulti}'s \cref{item:retargetable-n} (retargetability). These $\phi_i\cdot \uf \in \retarget\defeq \reals^5$ because $\reals^5$ is closed under permutation by $S_5$, satisfying \cref{item:symmetry-closure-n}.

Consider another $\uf'\in\reals^5$ such that $f_\train(A\mid \uf')> f_\train(B\mid \uf')$, and consider $i\neq j$. By the above symmetry arguments, $\uf'$ must also assign $a_1$ maximal utility. By \cref{lem:train-ub}, $f_\train(\set{a_i}\mid \phi_i\cdot \uf)\approx 1$ and $f_\train(\set{a_j}\mid \phi_i\cdot \uf)\approx 0$ since $i\neq j$, and vice versa when considering $\phi_j\cdot \uf$ instead of $\phi_i\cdot \uf$. Then since $\phi_i\cdot \uf$ and $\phi_j\cdot \uf$ induce distinct probability distributions over learned actions, they cannot be the same utility function. This satisfies \cref{item:distinct}.
\end{proof}

\begin{restatable}[The $\train$ bandit has orbit-level tendencies]{cor}{retargetBandit2}
$f_{\train}(B\mid \uf)\geq^4_{\text{most: } \reals^5}f_{\train}(A\mid\uf)$.
\end{restatable}
\begin{proof}
Combine \cref{res:retarget-bandit} and \cref{thm:retarget-decision-n}.
\end{proof}

\begin{figure}[t!]
    \includegraphics[width=\textwidth]{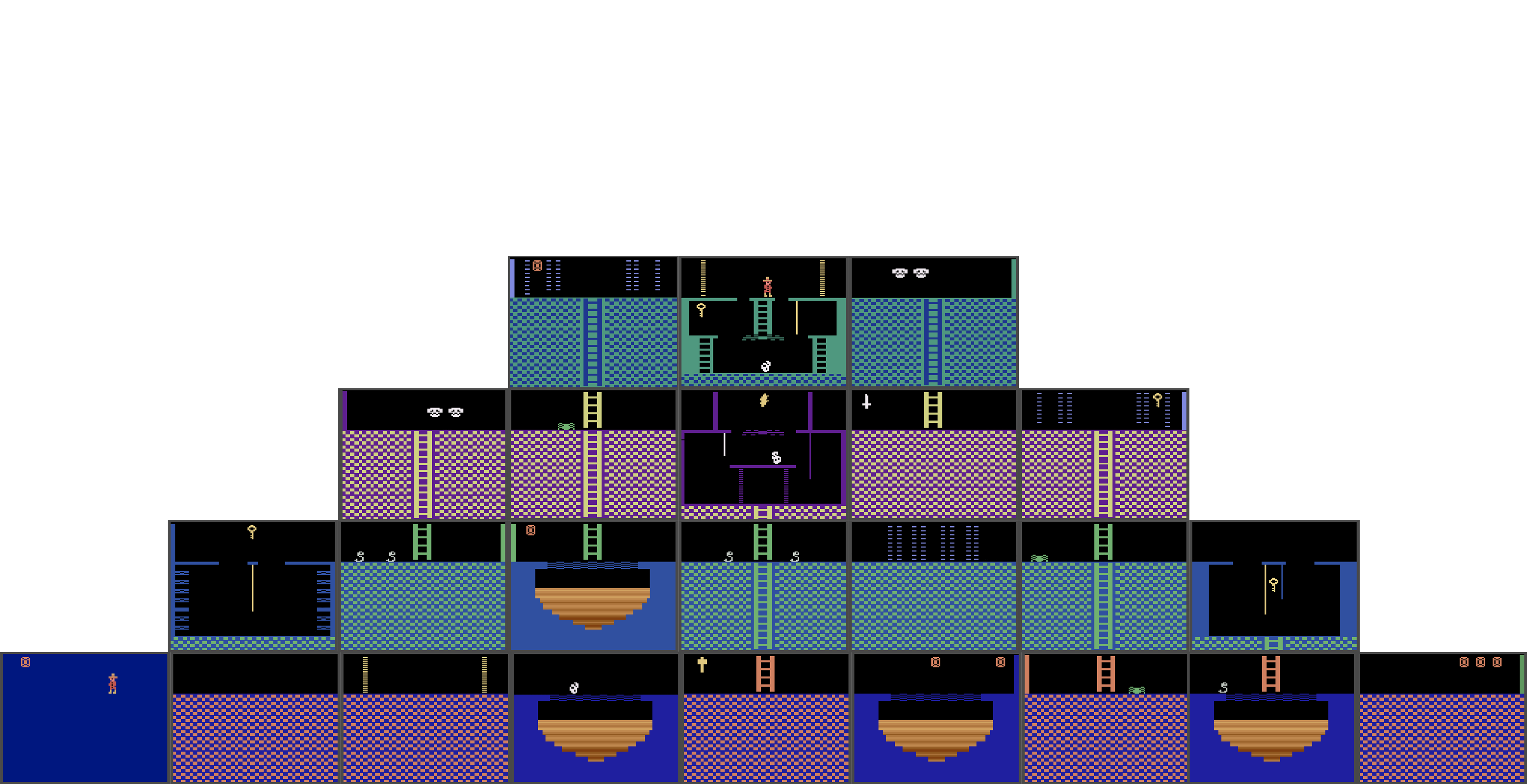}
    \caption{Map of the first level of Montezuma's Revenge.}
    \label{fig:mr-map}
\end{figure}

\subsection{Observation reward maximization}\label{sec:obs-analysis}
Let $T$ be a reasonably long rollout length, so that $\validObs$ is large—many different step-$T$ observations can be induced.

\begin{restatable}[Final reward maximization has strong orbit-level incentives in {\mr}]{prop}{retargetMaxMr}\label{res:final-reward-retarget-mr}
Let $n\defeq \lfloor\frac{\abs{\leave}}{\abs{\stay}}\rfloor$. $\fmax(\leave\mid R)\geqMost[n][\reals^\observe]\fmax(\stay\mid R)$.
\end{restatable}
\begin{proof}
Consider the vector space representation of observations, $\reals^{\abs{\observe}}$. Define $A\defeq\{\unitvec[o]\mid o\in \stay\}, B\defeq\{\unitvec[o]\mid o\in \leave\}$, and $C\defeq \validObs=A\cup B$ the union of $\stay,\leave$.

Since $\abs{\leave}\geq \abs{\stay}$ by assumption that $T$ is reasonably large, consider the involution $\phi_1\in S_{\abs{\observe}}$ which embeds $\stay$ into $\leave$, while fixing all other observations. If possible, produce another involution $\phi_2$ which also embeds $\stay$ into $\leave$, which fixes all other observations, and which ``doesn't interfere with $\phi_1$'' (\ie{} $\phi_2\cdot (\phi_1\cdot A)=\phi_1\cdot A$). We can produce $n\defeq \lfloor\frac{\abs{\leave}}{\abs{\stay}}\rfloor$ such involutions. Therefore, $B$ contains $n$ copies (\cref{def:copies}) of $A$ via involutions $\phi_1,\ldots,\phi_n$. Furthermore, $\phi_i\cdot (A\cup B)=A\cup B$, since each $\phi_i$ swaps $A$ with $B'\subseteq B$, and fixes all $\bv\in B\setminus B'$ by assumption. Thus, $\phi\cdot C=C$.

By \cref{prop:rationalities}'s \cref{item:frac-rational}, $\fracOpt{B\mid C,  R}\geqMost[n][\reals^\observe] \fracOpt{A\mid C,  R}$. Since $\fmax$ uniformly randomly chooses a maximal-reward observation to induce, $\forall X\subseteq C: \fmax(X\mid  R)=\fracOpt{X\mid C,  R}$. Therefore, $\fmax(\leave\mid R)\geqMost[n][\reals^\observe]\fmax(\stay\mid R)$.
\end{proof}

We want to reason about the probability that $\decide$ leaves the initial room by time $T$ in its rollout trajectories.
\begin{align}
    p_{\decide}(\text{leave}\mid \rtparam)&\defeq \prob[\substack{\pi\sim \decide(\rtparam),\\ \tau \sim \pi\mid \initMR}]{\tau\text{ has left the first room by step $T$}},\\
    p_{\decide}(\text{stay}\mid \rtparam)&\defeq \prob[\substack{\pi\sim \decide(\rtparam),\\ \tau \sim \pi\mid \initMR}]{\tau\text{ has not left the first room by step $T$}}.
\end{align}

We want to show that reward maximizers tend to leave the room: $p_{\max}(\text{leave}\mid R)\geqMost[n][\retarget] p_{\max}(\text{stay}\mid  R)$. However, we must be careful: In general, $\fmax(\leave\mid R)\neq p_{\max}(\text{leave}\mid  R)$ and $\fmax(\stay\mid R)\neq p_{\max}(\text{stay}\mid  R)$. For example, suppose that $o_T\in \leave$. By the definition of $\leave$, $o_T$ can only be observed if the agent has left the room by time step $T$, and so the trajectory $\tau$ must have left the first room. The converse argument does not hold: The agent could leave the first room, re-enter, and then wait until time $T$. Although one of the doors would have been opened (\cref{fig:montezuma}), the agent can also open the door without leaving the room, and then realize the same step-$T$ observation. Therefore, this observation doesn't belong to $\leave$.

\begin{restatable}[Room-status inequalities for {\mr}]{lem}{probRoom}\label{res:prob-room}
\begin{align}
     p_{\decide}(\text{stay}\mid  \rtparam)&\leq f_{\decide}(\stay\mid \rtparam),\label{eq:room-1}\\
     \text{and } f_{\decide}(\leave\mid \rtparam)&\leq p_{\decide}(\text{leave}\mid \rtparam).\label{eq:room-2}
\end{align}
\end{restatable}
\begin{proof}
For any $\decide$,
\begin{align}
&p_{\decide}(\text{stay}\mid \rtparam)\\
&= \prob[\substack{\pi\sim \decide(\rtparam),\\ \tau \sim \pi\mid \initMR}]{\tau\text{ stays through step $T$}}\\
&= \sum_{o\in \observe}\algprob{\text{$o$ at step $T$ of $\tau$}}\algprob{\text{$\tau$ stays}\mid \text{$o$ at step $T$}}\\
     &= \sum_{o \in \validObs}\algprob{\text{$o$ at step $T$}}\algprob{\text{$\tau$ stays}\mid \text{$o$ at step $T$}}\label{eq:discard-invalid}\\
     &= \sum_{o \in \stay} \algprob{\text{$o$ at step $T$}}\algprob{\text{$\tau$ stays}\mid \text{$o$ at step $T$}}\label{eq:cant-leave}\\
     &\leq \sum_{o \in \stay} \algprob{\text{$o$ at step $T$}}\\
     &=\algprob{o_T\in \stay}\\
     &\eqdef f_{\decide}(\stay\mid\rtparam).
\end{align}
\Cref{eq:discard-invalid} holds because the definition of $\validObs$ ensures that if $o\not \in \validObs$, then $\prob[\substack{\pi\sim \decide(\rtparam),\\ \tau \sim \pi\mid \initMR}]{o\mid \rtparam}=0$. Because $o\in\validObs\setminus\stay$ implies that $\tau$ left and so
\begin{equation*}
    \algprob{\text{$\tau$ stays}\mid \text{$o$ at step $T$}}=0,
\end{equation*}
\cref{eq:cant-leave} follows. Then we have shown \cref{eq:room-1}.

For \cref{eq:room-2},
\begin{align}
&f_{\decide}(\leave\mid \rtparam)\\
&\defeq \algprob{o_T\in \leave}\\
&=\sum_{o \in \leave} \algprob{\text{$o$ at step $T$}}\\
&=\sum_{o \in \leave} \algprob{\text{$o$ at step $T$}}\algprob{\text{$\tau$ leaves by step $T$}\mid \text{$o$ at step $T$}}\label{eq:equal-prob-leave}\\
&=\sum_{o\in \observe} \algprob{\text{$o$ at step $T$}}\algprob{\text{$\tau$ leaves by step $T$}\mid \text{$o$ at step $T$}}\label{eq:contain-room}\\
&= \prob[\substack{\pi\sim \decide(\rtparam),\\ \tau \sim \pi\mid \initMR}]{\tau\text{ has left the first room by step $T$}}\\
&\eqdef p_{\decide}(\text{leave}\mid \rtparam).
\end{align}
\Cref{eq:equal-prob-leave} follows because, since $o\in\leave$ are only realizable by leaving the first room, this implies $\algprob{\text{$\tau$ leaves by step $T$}\mid \text{$o$ at step $T$}}=1$. \Cref{eq:contain-room} follows because $\leave\subseteq \observe$, and probabilities are non-negative. Then we have shown \cref{eq:room-2}.
\end{proof}

\begin{restatable}[Final reward maximizers tend to leave the first room in {\mr}]{cor}{maxLeaveRoom}
\begin{equation}
    p_{\max}(\text{leave}\mid  R)\geqMost[n][\reals^\observe] p_{\max}(\text{stay}\mid  R).
\end{equation}
\end{restatable}
\begin{proof}
Using \cref{res:prob-room} and \cref{res:final-reward-retarget-mr}, apply \cref{lem:transit-geq-strong} with $f_0( R)\defeq p_{\max}(\text{leave}\mid  R),f_1( R)\defeq f_{\max}(\leave\mid  R), f_2( R)\defeq f_{\max}(\stay\mid  R),f_3( R)\defeq p_{\max}(\text{stay}\mid  R)$ to conclude that $p_{\max}(\text{leave}\mid  R)\geqMost[n][\reals^\observe] p_{\max}(\text{stay}\mid  R).$
\end{proof}

\subsection{Featurized reward maximization}\label{sec:feat-analysis}
Consider the featurization function which takes as input an observation $o\in\observe$:
\begin{equation}
    \featFn(o)\defeq \colvec{4}{\text{\# of keys in inventory shown by $o$}}{\text{\# of swords in inventory shown by $o$}}{\text{\# of torches in inventory shown by $o$}}{\text{\# of amulets in inventory shown by $o$}}.
\end{equation}
Consider $A_{\text{feat}}\defeq \set{\featFn(o) \mid o\in O_\text{stay}}, B_{\text{feat}}\defeq \set{\featFn(o) \mid o\in O_\text{leave}}$.

Let $\unitvec[i]\in\reals^4$ be the standard basis vector with a $ 1$ in entry $i$ and $ 0$ elsewhere. When restricted to the room shown in \cref{fig:montezuma}, the agent can either acquire the key in the first room and retain it until step $T$ ($\unitvec[1]$), or reach time step $T$ empty-handed ($\mathbf{0}$). We conclude that $A_{\text{feat}}=\set{\unitvec[1],\mathbf{0}}$.

For $B_{\text{feat}}$, recall that in \cref{sec:obs-reward} we assumed the rollout length $T$ to be reasonably large. Then by leaving the room, some realizable trajectory induces $o_T$ displaying an inventory containing only a sword ($\unitvec[2]$), or only a torch ($\unitvec[3]$), or only an amulet ($\unitvec[4]$), or nothing at all ($\mathbf{0}$). Therefore, $\set{\unitvec[2], \unitvec[3], \unitvec[4], \mathbf{0}}\subseteq B_{\text{feat}}$. $B_{\text{feat}}$ contains $3$ copies of $A_{\text{feat}}$ (\cref{def:copies}) via involutions $\phi_i: 1\leftrightarrow i$, $i\neq 1$. Suppose all feature coefficient vectors $\alpha\in\reals^4$ are plausible. Then $\retarget\defeq \reals^4$.

Let us be more specific about what is entailed by featurized reward maximization. The $\decide_{\max}(\alpha)$ procedure takes $\alpha$ as input and then considers the reward function $o\mapsto \featFn(o)^\top \alpha$. Then, $\decide_{\max}$ uniformly randomly chooses an observation $o_T\in\validObs$ which maximizes this featurized reward, and then uniformly randomly chooses a policy which implements $o_T$.

\begin{restatable}[$\mathrm{FracOptimal}$ inequalities]{lem}{fracOptIneq}\label{lem:frac-opt-ineq}
Let $X\subseteq Y'\subseteq Y\subsetneq \genVS$ be finite, and let $\uf\in \genVS$. Then
\begin{equation}
    \fracOpt{X\mid Y,\uf}\leq \fracOpt{X\mid Y',\uf} \leq \fracOpt{X\cup (Y\setminus Y')\mid Y,\uf}.
\end{equation}
\end{restatable}
\begin{proof}
For finite $X_1\subsetneq \genVS$, let $\Best{X_1\mid\uf}\defeq \argmax_{\x_1\in X_1} \x_1^\top \uf$. Suppose $\y' \in \Best{Y'\mid\uf}$, but $\y'\not \in \Best{Y\mid\uf}$. Then for all $\av \in \Best{Y'\mid \uf}$,
\begin{equation}
    \av^\top\uf = \y'^\top \uf < \max_{\y\in Y} \y^\top \uf.
\end{equation}
So $\av\not \in \Best{Y\mid \uf}$. Then either $\Best{Y'\mid \uf}\subseteq \Best{Y\mid\uf}$, or the two sets are disjoint.
\begin{align}
    \fracOpt{X\mid Y,\uf} &\defeq \frac{\abs{\Best{Y\mid\uf}\cap X}}{\abs{\Best{Y\mid\uf}}}\label{eq:first-frac}\\
    &\leq \frac{\abs{\Best{Y'\mid\uf}\cap X}}{\abs{\Best{Y'\mid\uf}}}\eqdef \fracOpt{X\mid Y',\uf}\label{eq:y-prime-contain}
\end{align}
If $\Best{Y'\mid \uf}\subseteq \Best{Y\mid\uf}$, then since $X\subseteq Y'$, we have $X\cap\Best{Y'\mid \uf} = X\cap\Best{Y\mid\uf}$. Then in this case, \cref{eq:first-frac} has equal numerator and larger denominator than \cref{eq:y-prime-contain}. On the other hand, if $\Best{Y'\mid \uf}\cap \Best{Y\mid\uf}=\varnothing$, then since $X\subseteq Y'$, $X\cap \Best{Y\mid\uf}=\varnothing$. Then \cref{eq:first-frac} equals $ 0$, and \cref{eq:y-prime-contain} is non-negative. Either way, \cref{eq:y-prime-contain}'s inequality holds. To show the second inequality, we handle the two cases separately.

\paragraph*{Subset case.} Suppose that $\Best{Y'\mid \uf}\subseteq \Best{Y\mid\uf}$.
\begin{align}
    \frac{\abs{\Best{Y'\mid\uf}\cap X}}{\abs{\Best{Y'\mid\uf}}}&\leq\frac{\abs{\Best{Y'\mid\uf}\cap X}+\abs{\Best{Y\setminus Y'\mid\uf}}}{\abs{\Best{Y'\mid\uf}}+\abs{\Best{Y\setminus Y'\mid\uf}}}\label{eq:ineq-add-both}\\
    &=\frac{\abs{\Best{Y'\mid\uf}\cap X}+\abs{\Best{Y\setminus Y'\mid\uf}\cap (Y\setminus Y')}}{\abs{\Best{Y'\mid\uf}}+\abs{\Best{Y\setminus Y'\mid\uf}}}\\
    &=\frac{\abs{\Best{Y'\mid\uf}\cap X}+\abs{\Best{Y\mid\uf}\cap (Y\setminus Y')}}{\abs{\Best{Y'\mid\uf}}+\abs{\Best{Y\setminus Y'\mid\uf}}}\label{eq:relate-Ybest}\\
    &=\frac{\abs{\Best{Y'\mid\uf}\cap X}+\abs{\Best{Y\mid\uf}\cap (Y\setminus Y')}}{\abs{\Best{Y\mid \uf}}}\label{eq:best-Y-ut}\\
    &=\frac{\abs{\Best{Y\mid\uf}\cap X}+\abs{\Best{Y\mid\uf}\cap (Y\setminus Y')}}{\abs{\Best{Y\mid \uf}}}\label{eq:disj-numerator}\\
    &=\frac{\abs{\Best{Y\mid\uf}\cap (X\cup(Y\setminus Y'))}}{\abs{\Best{Y\mid \uf}}}\label{eq:disj-X-Y}\\
    &\eqdef \fracOpt{X\cup (Y\setminus Y')\mid Y,\uf}.
\end{align}
\Cref{eq:ineq-add-both} follows because when $n\leq d, k\geq 0$, we have $\frac{n}{d}\leq \frac{n+k}{d+k}$. For \cref{eq:relate-Ybest}, since $\Best{Y'\mid\uf}\subseteq \Best{Y\mid\uf}$, we must have
\begin{equation*}
    \Best{Y\mid\uf}=\Best{Y\setminus Y'\mid\uf}\cup \Best{Y'\mid\uf}.
\end{equation*}
But then
\begin{align}
    \Best{Y\mid\uf}\cap(Y\setminus Y')&=\prn{\Best{Y\setminus Y'\mid\uf}\cap(Y\setminus Y')}\cup \prn{\Best{Y'\mid\uf}\cap (Y\setminus Y')}\\
    &=\Best{Y\setminus Y'\mid\uf}\cap(Y\setminus Y').
\end{align}
Then \cref{eq:relate-Ybest} follows. \Cref{eq:best-Y-ut} follows since
\begin{equation*}
    \Best{Y\mid\uf}=\Best{Y\setminus Y'\mid\uf}\cup \Best{Y'\mid\uf}.
\end{equation*}
\Cref{eq:disj-numerator} follows since $X\subseteq Y'$, and so
\begin{equation*}
    \Best{Y'\mid\uf}\cap X = \Best{Y\mid\uf}\cap X.
\end{equation*}
\Cref{eq:disj-X-Y} follows because $X\subseteq Y'$ is disjoint of $Y\setminus Y'$. We have shown that
\begin{equation*}
    \fracOpt{X\mid Y',\uf}\leq \fracOpt{X\cup (Y\setminus Y')\mid Y,\uf}
\end{equation*}
in this case.

\paragraph*{Disjoint case.} Suppose that $\Best{Y'\mid \uf}\cap \Best{Y\mid\uf}=\varnothing$.
\begin{align}
    \frac{\abs{\Best{Y'\mid\uf}\cap X}}{\abs{\Best{Y'\mid\uf}}}&\leq 1\label{eq:contain-card}\\
    &= \frac{\abs{\Best{Y\setminus Y'\mid \uf}}}{\abs{\Best{Y\setminus Y'\mid \uf}}}\\
    &= \frac{\abs{\Best{Y\setminus Y'\mid \uf}\cap (Y\setminus Y')}}{\abs{\Best{Y\setminus Y'\mid \uf}}}\\
    &= \frac{\abs{\Best{Y\setminus Y'\mid \uf}\cap (X\cup (Y\setminus Y'))}}{\abs{\Best{Y\setminus Y'\mid \uf}}}\label{eq:x-not-opt}\\
    &= \frac{\abs{\Best{Y\mid \uf}\cap (X\cup (Y\setminus Y'))}}{\abs{\Best{Y\mid \uf}}}\label{eq:final-ineq-disjoint}\\
    &\eqdef \fracOpt{X\cup (Y\setminus Y')\mid Y,\uf}.
\end{align}
\Cref{eq:contain-card} follows because $\Best{Y'\mid\uf}\cap X\subseteq \Best{Y'\mid\uf}$. For \cref{eq:x-not-opt}, note that we trivially have $\Best{Y'\mid \uf}\cap \Best{Y\setminus Y'\mid\uf}=\varnothing$, and also that $X\subseteq Y'$. Therefore,  $\Best{Y\setminus Y'\mid \uf}\cap X=\varnothing$, and \cref{eq:x-not-opt} follows. Finally, the disjointness assumption implies that \[\max_{\y'\in Y'}\y'^\top \uf < \max_{\y \in Y} \y^\top \uf.\] Therefore, the optimal elements of $Y$ must come exclusively from $Y\setminus Y'$; \ie{} $\Best{Y\mid\uf}= \Best{Y\setminus Y'\mid\uf}$. Then \cref{eq:final-ineq-disjoint} follows, and we have shown that
\begin{equation*}
    \fracOpt{X\mid Y',\uf}\leq \fracOpt{X\cup (Y\setminus Y')\mid Y,\uf}
\end{equation*}
in this case.
\end{proof}

\begin{restatable}[Generalizing \cref{lem:frac-opt-ineq}]{conjSec}{genForm}
\Cref{lem:frac-opt-ineq} and \citet{turner_optimal_2020}'s Lemma E.26 have extremely similar functional forms. How can they be unified?
\end{restatable}

\begin{restatable}[Featurized reward maximizers tend to leave the first room in {\mr}]{prop}{featLeaveRoom}\label{res:feat-leave-room}
\begin{equation}
    p_{\max}(\text{leave}\mid \alpha)\geqMost[3][\reals^4] p_{\max}(\text{stay}\mid \alpha).
\end{equation}
\end{restatable}
\begin{proof}
We want to show that $f_{\max}(\leave\mid \alpha)\geqMost[n][\reals^4] f_{\max}(\stay \mid \alpha)$. Recall that $A_{\text{feat}}=\{\unitvec[1],\mathbf{0}\},B'_{\text{feat}}\defeq \{\unitvec[2],\unitvec[3],\unitvec[4]\}\subseteq B_{\text{feat}}$.
\begin{align}
    &p_{\max}(\text{stay}\mid  \alpha)\label{eq:apply-room-1}\\
    &\leq f_{\max}(\stay \mid \alpha) \\
    &\defeq \prob[\substack{\pi\sim \decide_{\max}(\alpha),\\\tau\sim \pi\mid s_0}]{o_T\in \stay}\\
    &= \prob[\substack{\pi\sim \decide_{\max}(\alpha),\\\tau\sim \pi\mid s_0}]{o_T\in \stay, \featFn(o_T)\neq\mathbf{0}}+\prob[\substack{\pi\sim \decide_{\max}(\alpha),\\\tau\sim \pi\mid s_0}]{o_T\in \stay,\featFn(o_T)=\mathbf{0}}\\
    &\leq \fracOpt{\set{\unitvec[1]}\mid C_\text{feat}, \alpha}+\prob[\substack{\pi\sim \decide_{\max}(\alpha),\\\tau\sim \pi\mid s_0}]{o_T\in \stay,\featFn(o_T)=\mathbf{0}}\label{eq:A-sub-opt}\\
    &\leq \fracOpt{\set{\unitvec[1]}\mid \set{\unitvec[1],\unitvec[2],\unitvec[3],\unitvec[4]}, \alpha}+\prob[\substack{\pi\sim \decide_{\max}(\alpha),\\\tau\sim \pi\mid s_0}]{o_T\in \stay,\featFn(o_T)=\mathbf{0}}\label{eq:leq-frac}\\
    &\leqMost[3][\reals^4_{>0}]\fracOpt{\set{\unitvec[2],\unitvec[3],\unitvec[4]}\mid \set{\unitvec[1],\unitvec[2],\unitvec[3],\unitvec[4]}, \alpha}\nonumber\\
    &\phantom{\leq \fracOpt{\set{\unitvec[1]}\mid \set{\unitvec[1],\unitvec[2],\unitvec[3],\unitvec[4]}, \alpha}}+\prob[\substack{\pi\sim \decide_{\max}(\alpha),\\\tau\sim \pi\mid s_0}]{o_T\in \leave,\featFn(o_T)=\mathbf{0}}\label{eq:leqmostFracFeat}\\
    &\leq \fracOpt{\set{\unitvec[2],\unitvec[3],\unitvec[4]}\cup (C_\text{feat}\setminus \set{\unitvec[1],\unitvec[2],\unitvec[3],\unitvec[4]})\mid C_\text{feat},\alpha}\label{eq:apply-fracopt-ineq-2}\\
    &=\fracOpt{C_\text{feat}\setminus \set{\unitvec[1]}\mid C_\text{feat},\alpha}\\
    &\leq \prob[\substack{\pi\sim \decide_{\max}(\alpha),\\\tau\sim \pi\mid s_0}]{o_T\in \leave} \label{eq:leave-confirm}\\
    &\eqdef f_{\max}(\leave \mid \alpha)\\
    &\leq p_{\max}(\text{leave}\mid  \alpha).\label{eq:apply-room-2}
\end{align}
\Cref{eq:apply-room-1} and \cref{eq:apply-room-2} hold by \cref{res:prob-room}. If $o_T \in \stay$ is realized by $\fmax$ and $\featFn(o_T)\neq \mathbf{0}$, then we must have $\featFn(o_T)= \{\unitvec[1]\}$ be optimal and so the $\unitvec[1]$ inventory configuration is realized. Therefore, \cref{eq:A-sub-opt} follows. \Cref{eq:leq-frac} follows by applying the first inequality of \cref{lem:frac-opt-ineq} with $X\defeq \{\unitvec[1]\}, Y'\defeq \{\unitvec[1],\unitvec[2],\unitvec[3],\unitvec[4]\},Y\defeq C_\text{feat}$.

By applying \cref{prop:rationalities}'s \cref{item:frac-rational} with $A\defeq A_{\text{feat}}=\{\unitvec[1]\}$, $B'\defeq B'_{\text{feat}}= \{\unitvec[2],\unitvec[3],\unitvec[4]\}$, $C\defeq A\cup B'$, we have
\begin{multline}
    \fracOpt{\set{\unitvec[1]}\mid \set{\unitvec[1],\unitvec[2],\unitvec[3],\unitvec[4]}, \alpha}
    \leqMost[3][\reals^4_{>0}]\\
    \fracOpt{\set{\unitvec[2],\unitvec[3],\unitvec[4]}\mid \set{\unitvec[1],\unitvec[2],\unitvec[3],\unitvec[4]}, \alpha}.\label{eq:geq-most-frac}
\end{multline}
Furthermore, observe that
\begin{equation}
\prob[\substack{\pi\sim \decide_{\max}(\alpha),\\\tau\sim \pi\mid s_0}]{o_T\in \stay,\featFn(o_T)=\mathbf{0}}\leq \prob[\substack{\pi\sim \decide_{\max}(\alpha),\\\tau\sim \pi\mid s_0}]{o_T\in \leave,\featFn(o_T)=\mathbf{0}}\label{eq:prob-ineq-0}
\end{equation}
because either $\mathbf{0}$ is not optimal (in which case both sides equal 0), or else $\mathbf{0}$ is optimal, in which case the right side is strictly greater. This can be seen by considering how $\decide_{\max}(\alpha)$ uniformly randomly chooses an observation in which the agent ends up with an empty inventory. As argued previously, the vast majority of such observations can only be induced by leaving the first room.

Combining \cref{eq:geq-most-frac} and \cref{eq:prob-ineq-0}, \cref{eq:leqmostFracFeat} follows. \Cref{eq:apply-fracopt-ineq-2} follows by applying the second inequality of \cref{lem:frac-opt-ineq} with $X\defeq \{\unitvec[2],\unitvec[3],\unitvec[4]\}$, $Y'\defeq \{\unitvec[1],\unitvec[2],\unitvec[3],\unitvec[4]\}$, $Y\defeq C_\text{feat}$. If $\featFn(o_T)\in B_\text{feat}$ is realized by $\fmax$, then by the definition of $B_{\text{feat}}$, $o_T \in \leave$ is realized, and so \cref{eq:leave-confirm} follows.

Then by applying \cref{lem:transit-geq-strong} with
\begin{align}
    f_0(\alpha)&\defeq p_{\max}(\text{leave}\mid  \alpha),\\
    f_1(\alpha)&\defeq \fracOpt{\set{\unitvec[1]}\mid \set{\unitvec[1],\unitvec[2],\unitvec[3],\unitvec[4]}, \alpha},\\
    f_2(\alpha)&\defeq \fracOpt{\set{\unitvec[2],\unitvec[3],\unitvec[4]}\mid \set{\unitvec[1],\unitvec[2],\unitvec[3],\unitvec[4]}, \alpha},\\
    f_3(\alpha)&\defeq p_{\max}(\text{stay}\mid  \alpha),
\end{align}
we conclude that $p_{\max}(\text{leave}\mid \alpha)\geqMost[3][\reals^4_{>0}] p_{\max}(\text{stay}\mid \alpha).$
\end{proof}

Lastly, note that if $\mathbf{0}\in\retarget$ and $f(A\mid \mathbf{0})> f(B\mid\mathbf{0})$, $f$ cannot be even be simply retargetable for the $\retarget$ parameter set. This is because $\forall \phi \in \genSym$, $\phi\cdot \mathbf{0}=\mathbf{0}$. For example, inductive bias ensures that, absent a reward signal, learned policies tend to stay in the initial room in {\mr}. This is one reason why \cref{sec:rl-analyze}'s analysis of the policy tendencies of reinforcement learning excludes the all-zero reward function. 
\section{Lower bounds on \textsc{mdp} power-seeking incentives for optimal policies}\label{app:mdp}
\begin{figure}[h!]
    \centering
    \includegraphics{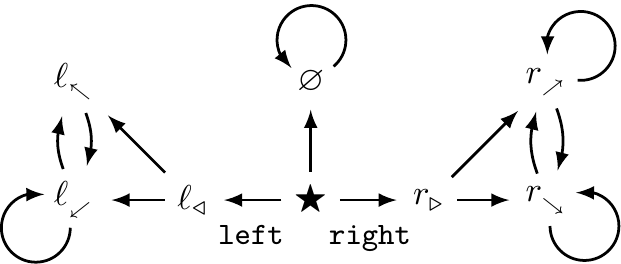}
    \caption[A toy {\mdp} for reasoning about power-seeking tendencies]{A toy {\mdp} for reasoning about power-seeking tendencies. \emph{Reproduced from \citet{turner_optimal_2020}.}}
    \label{fig:case-quant}
\end{figure}

\citet{turner_optimal_2020} prove conditions under which \emph{at least half} of the orbit of every reward function incentivizes power-seeking behavior. For example, in \cref{fig:case-quant}, they prove that avoiding $\sink$ maximizes average per-timestep reward for at least half of reward functions. Roughly, there are more self-loop states ($\sink$, $\farleft$, $\farright$, $\topright$) available if the agent goes $\leftA$ or $\rightA$ instead of up towards $\sink$. We strengthen this claim, with \cref{cor:quant-one-cyc} showing that for \emph{at least three-quarters} of the orbit of every reward function, it is average-optimal to avoid $\sink$.

Therefore, we answer \citet{turner_optimal_2020}'s open question of whether increased number of environmental symmetries quantitatively strengthens the degree to which power-seeking is incentivized. The answer is \emph{yes}. In particular, it may be the case that only one in a million state-based reward functions makes it average-optimal for Pac-Man to die immediately.

We will briefly restate several definitions needed for our key results, \cref{thm:rsdIC-quant} and \cref{cor:quant-one-cyc}. For explanation, see \citet{turner_optimal_2020}.

\begin{restatable}[Non-dominated linear functionals]{definition}{ndLinFuncQuant}
Let $X\subsetneq \rewardVS$ be finite. $\ND{X}\defeq \set{\x\in X\mid \exists \rf \in \rewardVS: \x^\top \rf > \max_{\x'\in X\setminus \set{\x}} \x'^\top \rf}$.
\end{restatable}

\begin{restatable}[Bounded reward function distribution]{definition}{bdDistDef}
$\DSetBd$ is the set of bounded\hyp{}support probability distributions $\Dbd$.
\end{restatable}

\begin{remark}
When $n=1$, \cref{lem:expect-superior-simple} reduces to the first part of \citet{turner_optimal_2020}'s \cref{lem:expect-superior}, and \cref{lem:orbit-opt-prob-simple} reduces to the first part of \citet{turner_optimal_2020}'s \cref{lem:opt-prob-superior}.
\end{remark}

\begin{restatable}[Quantitative expectation superiority lemma]{lem}{expectSuperiorSimple}\label{lem:expect-superior-simple}
Let $A,B\subsetneq \genVS$ be finite and let $g:\reals\to \reals$ be a (total) increasing function. Suppose $B$ contains $n$ copies of $\ND{A}$. Then
\begin{align}
    \E{\rf \sim \Dbd}{g\prn{\max_{\bv\in B}\bv^\top \rf}}\geqMost[n][\DSetBd]  \E{\rf \sim \Dbd}{g\prn{\max_{\av\in A}\av^\top \rf}}.\label{eq:permute-superior-quant}
\end{align}
\end{restatable}
\begin{proof}
Because $g:\reals\to\reals$ is increasing, it is measurable (as is $\max$).

Let $L\defeq \inf_{\rf \in \supp[\Dbd]} \max_{\x \in X} \x^\top \rf, U\defeq \sup_{\rf \in \supp[\Dbd]} \max_{\x \in X} \x^\top \rf$. Both exist because $\Dbd$ has bounded support. Furthermore, since $g$ is monotone increasing, it is bounded $[g(L),g(U)]$ on $[L,U]$. Therefore, $g$ is measurable and bounded on each $\supp[\Dbd]$, and so the relevant expectations exist for all $\Dbd$.

For finite $X\subsetneq \genVS$, let $f(X\mid \uf)\defeq g(\max_{\x\in X}\x^\top \uf)$. By \cref{lem:card-EU-invar}, $f$ is invariant under joint permutation by $\genSym$. Furthermore, $f$ is measurable because $g$ and $\max$ are. Therefore,  apply \cref{lem:invar-expect} to conclude that $f(X\mid \Dbd)\defeq \E{\uf\sim \Dbd}{g(\max_{\x\in X}\x^\top \uf)}$ is also invariant under joint permutation by $\genSym$ (with $f$ being bounded when restricted to $\supp[\Dbd]$). Lastly, if $X'\subseteq X$, $f(X'\mid \Dbd)\leq f(X\mid \Dbd)$ because $g$ is increasing.
\begin{align}
    \E{\uf \sim \Dbd}{g\prn{\max_{\av\in A}\av^\top \uf}}&=\E{\uf \sim \Dbd}{g\prn{\max_{\av\in \ND{A}}\av^\top \uf}}\label{eq:nd-A-expect}\\
    &\leqMost[n] \E{\rf \sim \Dbd}{g\prn{\max_{\bv\in B}\bv^\top \rf}}.\label{eq:bv-most-n}
\end{align}

\Cref{eq:nd-A-expect} follows by \cref{cor:nd-func-indif} of \citep{turner_optimal_2020}. \Cref{eq:bv-most-n} follows by applying \cref{lem:general-orbit-simple} with $f$ as defined above with the $\phi_1,\ldots,\phi_n$ guaranteed by the copy assumption.
\end{proof}

\begin{restatable}[Linear functional optimality probability \citep{turner_optimal_2020}]{definition}{linFNProbRestate}
For finite $A,B\subsetneq \rewardVS$, the \emph{probability under $\Dany$ that $A$ is optimal over $B$} is
\begin{equation*}
    \phelper{A\geq B}[\Dany]\defeq \optprob[\rf \sim \Dany]{\max_{\av\in A} \av^\top \rf \geq \max_{\bv\in B} \bv^\top \rf}.
\end{equation*}
\end{restatable}

\begin{restatable}[Quantitative optimality probability superiority lemma]{lem}{orbOptProbQuantSimple}\label{lem:orbit-opt-prob-simple}
Let $A,B,C\subsetneq \genVS$ be finite and let $Z$ satisfy $\ND{C}\subseteq Z\subseteq C$. Suppose that $B$ contains $n$ copies of $\ND{A}$ via involutions $\phi_i$. Furthermore, let $B_\text{extra}\defeq B\setminus \prn{\cup_{i=1}^n \phi_i\cdot \ND{A}}$; suppose that for all $i$, $\phi_i\cdot\prn{Z\setminus B_\text{extra}}=Z\setminus B_\text{extra}$.

Then $\phelper{B\geq C}[\Dany]\geqMost[n] \phelper{A\geq C}[\Dany]$.
\end{restatable}
\begin{proof}
For finite $X,Y\subsetneq \genVS$, let 
\begin{equation*}
g(X,Y\mid \Dany)\defeq \phelper{X\geq Y}[\Dany]=\E{\uf\sim \Dany}{\indic{\max_{\x\in X}\x^\top \uf \geq \max_{\mathbf{y}\in Y} \mathbf{y}^\top \uf}}.
\end{equation*}
By the proof of \cref{item:rational} of \cref{prop:rationalities}, $g$ is the expectation of a $\uf$-measurable function. $g$ is an EU function, and so \cref{lem:card-EU-invar} shows that it is invariant to joint permutation by $\phi_i$. Letting $f_Y(X\mid \Dany)\defeq g(X,Y\mid \Dany)$, \cref{lem:hide-second} shows that $f_Y(X\mid \Dany)=f_Y(\phi_i\cdot X \mid \phi_i\cdot \Dany)$ whenever the $\phi_i$ satisfy $\phi_i\cdot Y=Y$.

Furthermore, if $X'\subseteq X$, then $f_Y(X'\mid \Dany)\leq f_Y(X\mid \Dany)$.
\begin{align}
    \phelper{A\geq C}[\Dany]&=\phelper{\ND{A}\geq C}[\Dany]\label{eq:nd-a-z}\\
    &\leq \phelper{\ND{A}\geq Z\setminus B_\text{extra}}[\Dany]\label{eq:leq-Z}\\
    &\leqMost[n] \phelper{B\geq Z\setminus B_\text{extra}}[\Dany]\label{eq:leqmost-n-B}\\
    &\leq \phelper{B\cup B_\text{extra} \geq Z}[\Dany]\label{eq:B-Bextra}\\
    &= \phelper{B \geq Z}[\Dany]\label{eq:B-Bextra-contain}\\
    &= \phelper{B \geq C}[\Dany].\label{eq:nd-a-z-2}
\end{align}
\Cref{eq:nd-a-z} follows by \citet{turner_optimal_2020}'s \cref{lem:nd-opt-contain}'s \cref{item:ND-contain-max-2} with $X\defeq A$, $X'\defeq \ND{A}$  (similar reasoning holds for $C$ and $Z$ in \cref{eq:nd-a-z-2}). \Cref{eq:leq-Z} follows by the first inequality of \cref{lem:inclusion-opt} of \citep{turner_optimal_2020} with $X\defeq A,Y\defeq C, Y'\defeq Z\setminus B_\text{extra}$. \Cref{eq:leqmost-n-B} follows by applying \cref{lem:general-orbit-simple} with the $f_{Z\setminus B_\text{extra}}$ defined above. \Cref{eq:B-Bextra} follows by the second inequality of \cref{lem:inclusion-opt} of \citep{turner_optimal_2020} with $X\defeq A,Y\defeq Z, Y'\defeq Z\setminus B_\text{extra}$. \Cref{eq:B-Bextra-contain} follows because $B_\text{extra}\subseteq B$.

Letting $f_0(\Dany)\defeq \phelper{A\geq C}[\Dany],f_1(\Dany)\defeq \phelper{\ND{A}\geq Z\setminus B_\text{extra}}[\Dany],f_2(\Dany)\defeq \phelper{B\geq Z\setminus B_\text{extra}}[\Dany],f_3(\Dany)\defeq \phelper{B\geq C}[\Dany]$, apply \cref{lem:transit-geq-strong} to conclude that
\begin{equation*}
    \phelper{A\geq C}[\Dany]\leqMost[n]\phelper{B\geq C}[\Dany].
\end{equation*}
\end{proof}

\begin{restatable}[Rewardless {\mdp} \citep{turner_optimal_2020}]{definition}{rewardlessQuant}
$\langle \mathcal{S}, \mathcal{A}, T \rangle$ is a rewardless {\mdp} with finite state and action spaces $\mathcal{S}$ and $\mathcal{A}$, and stochastic transition function $T: \St \times \A \to \Delta(\St)$. We treat the discount rate $\gamma$ as a variable with domain $[0,1]$.
\end{restatable}

\begin{restatable}[{\stateEnd} states \citep{turner_optimal_2020}]{definition}{oneCycStateQuant}
Let $\unitvec[s]\in \reals^{\abs{\St}}$ be the standard basis vector for state $s$, such that there is a $1$ in the entry for state $s$ and $0$ elsewhere. State $s$ is a \emph{\stateEnd} if $\exists a\in\A: T(s,a)=\unitvec$. State $s$ is a \emph{{\terminal} state} if $\forall a\in\A:T(s,a)=\unitvec$.
\end{restatable}

\begin{restatable}[State visit distribution \citep{sutton_reinforcement_1998}]{definition}{DefVisitQuant}
$\Pi\defeq \A^\St$, the set of stationary deterministic policies. The \emph{visit distribution} induced by following policy $\pi$ from state $s$ at discount rate $\gamma\in[0,1)$ is $\fpi{s}(\gamma) \defeq \sum_{t=0}^\infty \gamma^t \E{s_{t} \sim \pi\mid s}{\unitvec[s_t]}$.
$\fpi{s}$ is a \emph{visit distribution function}; $\F(s)\defeq \{ \fpi{s} \mid \pi \in \Pi\}$.
\end{restatable}

\begin{restatable}[Recurrent state distributions \citep{puterman_markov_2014}]{definition}{DefRSDQuant}
The \emph{recurrent state distributions}  which can be induced from state $s$ are $\RSD \defeq \set{\lim_{\gamma\to1} (1-\gamma) \fpi{s}(\gamma) \mid \pi \in \Pi}$. $\RSDnd$ is the set of \textsc{rsd}s which strictly maximize average reward for some reward function.
\end{restatable}

\begin{restatable}[Average-optimal policies \citep{turner_optimal_2020}]{definition}{defAverageQuant}
The \emph{average-optimal policy set} for reward function $R$ is $\average[R]\defeq \set{\pi\in\Pi \mid \forall s \in \St: \dbf^{\pi,s} \in  \argmax_{\dbf\in\RSD} \dbf^\top \rf}$ (the policies which induce optimal {\rsd}s at all states). For $D\subseteq \RSD$, the \emph{average optimality probability} is $\avgprob[\Dany]{D}\defeq \optprob[R\sim \Dany]{\exists \dbf^{\pi,s} \in D: \pi \in \average}$.
\end{restatable}

\begin{remark}
\Cref{thm:rsdIC-quant} generalizes the first claim of \citet{turner_optimal_2020}'s \cref{rsdIC}, and \cref{cor:quant-one-cyc} generalizes the first claim of \citet{turner_optimal_2020}'s \cref{cor:avg-avoid-terminal}.
\end{remark}

\begin{restatable}[Quantitatively, average-optimal policies tend to end up in ``larger'' sets of {\rsd}s]{thm}{rsdICQuant}\label{thm:rsdIC-quant} Let $D',D\subseteq \RSD$. Suppose that $D$ contains $n$ copies of $D'$ and that the sets $D'\cup D$ and $\RSDnd\setminus \prn{D'\cup D}$ have pairwise orthogonal vector elements (\ie{} pairwise disjoint vector support). Then $\avgprob[\Dany]{D'}\leqMost[n] \avgprob[\Dany]{D}$.
\end{restatable}
\begin{proof}
Let $D_i\defeq\phi_i\cdot D'$, where $D_i\subseteq D$ by assumption.\\
Let $S\defeq \set{s'\in \St \mid \max_{\dbf \in D'\cup D} \dbf^\top \unitvec[s']>0}$.\\
Define
\begin{equation}
    \phi_i'(s')\defeq
    \begin{cases}
    \phi_i(s') & \text{ if } s'\in S\label{eq:quant-phi-prime-def}\\
    s' &\text{ else}.
    \end{cases}
\end{equation}

Since $\phi_i$ is an involution, $\phi_i'$ is also an involution. Furthermore, $\phi_i' \cdot D'=D_i$, $\phi_i' \cdot D_i=D'$, and $\phi_i'\cdot D_j=D_j$ for $j\neq i$ because we assumed that these equalities hold for $\phi_i$, and $D',D_i,D_j\subseteq D'\cup D$ and so the vectors of these sets have support contained in $S$.

Let $D^*\defeq D'\cup_{i=1}^n D_i\cup \prn{\RSDnd\setminus \prn{D'\cup D}}$.
By an argument mirroring that in the proof of \cref{rsdIC} in \citet{turner_optimal_2020} and using the fact that $\phi_i'\cdot D_j=D_j$ for all $i\neq j$, $\phi_i'\cdot D^*=D^*$. Consider $Z \defeq \prn{\RSDnd\setminus (D'\cup D)}\cup D' \cup D$. First, $Z\subseteq \RSD$ by definition. Second, $\RSDnd=\RSDnd \setminus (D' \cup D) \cup \prn{\RSDnd \cap D'} \cup \prn{\RSDnd \cap D}\subseteq Z$. Note that $D^* = Z\setminus (D\setminus \cup_{i=1}^n D_i)$.
\begin{align}
\avgprob[\Dany]{D'}&= \phelper{D'\geq \RSD}[\Dany]\\
&\leqMost[n] \phelper{D\geq \RSD}[\Dany]\label{eq:quant-leq-most-1}\\
&= \avgprob[\Dany]{D}.
\end{align}
Since $\phi_i'\cdot D'\subseteq D$ and $\ND{D'}\subseteq D'$, $\phi_i'\cdot \ND{D'}\subseteq D$ and so $D$ contains $n$ copies of $\ND{D'}$ via involutions $\phi_i'$. Then \cref{eq:quant-leq-most-1} holds by applying \cref{lem:orbit-opt-prob-simple} with $A\defeq D'$, $B_i\defeq D_i$ for all $i=1,\ldots,n$, $B\defeq D, C\defeq \RSD$,  $Z$ as defined above, and involutions $\phi_i'$ which satisfy $\phi_i'\cdot \prn{Z\setminus (B \setminus \cup_{i=1}^n B_i)}=\phi_i'\cdot D^*=D^*=Z\setminus (B \setminus \cup_{i=1}^n B_i)$.
\end{proof}

\begin{restatable}[Quantitatively, average-optimal policies tend not to end up in any given {\stateEnd}]{cor}{QuantOneCyc}\label{cor:quant-one-cyc}
Let $D'\defeq \set{\unitvec[s_1'],\ldots,\unitvec[s_k']}, D_r\defeq \set{\unitvec[s_1],\ldots,\unitvec[s_{n\cdot k}]}\subseteq \RSD$ be disjoint, for $n\geq 1, k\geq 1$.
Then $\avgprob[\Dany]{D'}\leqMost[n] \avgprob[\Dany]{\RSD\setminus D'}$.
\end{restatable}
\begin{proof}
For each $i\in\set{1,\ldots,n}$, let
\begin{align*}
    \phi_i & \defeq (s_1' \,\,\, s_{(i-1)\cdot k +1})\cdots (s_k' \,\,\, s_{(i-1)\cdot k + k}),\\
    D_i    & \defeq \set{\unitvec[s_{(i-1)\cdot k + 1}],\ldots, \unitvec[s_{(i-1)\cdot k + k}]},\\
    D      & \defeq \RSD\setminus D'.
\end{align*}
Each $D_i\subseteq D_r\subseteq \RSD\setminus D'$ by disjointness of $D'$ and $D_r$.

$D$ contains $n$ copies of $D'$ via involutions $\phi_1,\ldots,\phi_n$. $D'\cup D=\RSD$ and $\RSDnd\setminus\RSD=\emptyset$ trivially have pairwise orthogonal vector elements.

Apply \cref{thm:rsdIC-quant} to conclude that
\begin{equation*}
    \avgprob[\Dany]{D'}\leqMost[n] \avgprob[\Dany]{\RSD\setminus D'}.
\end{equation*}
\end{proof}

Let $A\defeq \set{\unitvec[1],\unitvec[2]}, B\def \set{\unitvec[3],\unitvec[4],\unitvec[5]}\subseteq \reals^5$, $C\defeq A\cup B$. \Cref{conj:fractional} conjectures that \eg{}
\begin{equation*}
    \phelper{B\geq C}\geqMost[\frac{3}{2}] \phelper{A\geq C}.
\end{equation*}

\begin{restatable}[Fractional quantitative optimality probability superiority lemma]{conjSec}{orbOptProbQuantFrac}\label{conj:fractional}
Let $A$, $B$, $C\subsetneq \genVS$ be finite. If $A=\bigcup_{j=1}^m A_j$ and $\bigcup_{i=1}^n B_i\subseteq B$ such that for each $A_j$, $B$ contains $n$ copies ($B_1,\ldots,B_n$) of $A_j$ via involutions $\phi_{ji}$ which \emph{also} fix $\phi_{ji}\cdot A_{j'}=A_{j'}$ for $j'\neq j$, then
\begin{equation*}
    \phelper{B\geq C}[\Dany]\geqMost[\frac{n}{m}] \phelper{A\geq C}[\Dany].
\end{equation*}
\end{restatable}

We suspect that any proof of the conjecture should generalize \cref{lem:general-orbit-simple-nonunif} to the fractional set copy containment case.

\chapter{Additional Theoretical Results}
\AddLoeChapter{theorems}
\AddLoeChapter{definitions}
\AddLoeChapter{conjectures}
\label{chp:misc-results}
This chapter contains results on:
\begin{itemize}
    \item {\mdp[M]} structure and representation (seemingly building towards a novel category-theoretic treatment of {\mdp}s),
    \item The expressivity of Markov reward \citep{abel2021expressivity},
    \item The properties of optimal value functions,
    \item Characterizing the {\mdp}s in which there exist reward functions whose optimal policy set depends on the discount rate,
    \item An optimal value-based distance metric on the state space,
    \item An improved {\rl} regret formalism and a no-free-lunch theorem prohibiting simultaneously bounding this regret for all reward functions,
    \item Power-seeking incentives for agents with discount rate very close to $1$, and
    \item Power-seeking incentives for $\epsilon$-optimal agents.
\end{itemize}

\section{Basic results on visit distributions}

The traditional view of Markov decision processes ({\mdp}s) takes for granted a reward function and discount rate, and considers the on-policy value function induced by solving the Bellman equations for a given policy. The ``dual'' view considers the available \emph{state visit distribution functions}.

I think that the dual view deserves more prominence: each reward function and discount rate tuple merely imposes a preordering over policies \emph{given} the existing dynamics. The dual view regards the environment structure as primary, and the optimization objective as secondary. Studying this environmental structure makes available unexplored areas of basic {\mdp} theory.

We freely rely on the theorems and definitions of \emph{Optimal Policies Tend to Seek Power} \citep{turner_optimal_2020} and \emph{Avoiding Side Effects in Complex Environments} \citep{turner2020avoiding}, as they do not depend on these results. All {\mdp}s are assumed to have finite state and action spaces.

\begin{restatable}[Each state has a visit distribution function]{lem}{existVisit}\label{lem:exist-visit}
$\forall s\in\St: 1 \leq \abs{\F(s)}$.
\end{restatable}
\begin{proof}
$\abs{\Pi}=\abs{\A}^{\abs{\St}}$. Each {\mdp} must have at least one action and one state, and so $\Pi$ is not empty.

Let $\pi\in\Pi,\gamma \in [0,1),s\in\St$ and consider $\mathbf{T}^\pi$, the transition matrix which $\pi$ induces. $\mathbf{T}^\pi$ is left stochastic and hence has spectral radius at most $1$. Therefore, $\gamma\mathbf{T}^\pi$ has spectral radius strictly less than $1$. This means that its Neumann series $\sum_{t=0}^\infty \prn{\gamma\mathbf{T}^\pi}^t=\prn{\mathbf{I}-\gamma \mathbf{T}^\pi}^{-1}$, where $\mathbf{I}$ is the $\abs{\St}$-dimensional identity matrix.
\begin{equation}
    \fpi{s}(\gamma) \defeq \sum_{t=0}^\infty \gamma^t \E{s'}{\unitvec[s']\mid \pi \text{ followed for } t \text{ steps from } s}.
\end{equation}

By inspection, $\fpi{s}(\gamma)= \mathbf{I}\unitvec + \prn{\gamma\mathbf{T}^\pi}\unitvec + \prn{\gamma\mathbf{T}^\pi}^2 \unitvec +\cdots$, so $\fpi{s}(\gamma)=\prn{\mathbf{I}-\gamma \mathbf{T}^\pi}^{-1}\unitvec$.
\end{proof}

\begin{restatable}[$\fpi{s}$ identity \citep{puterman_markov_2014}]{cor}{fpiIdentity}\label{cor:fpi-identity}
$\fpi{s}(\gamma)=\prn{\mathbf{I}-\gamma \mathbf{T}^\pi}^{-1}\unitvec$.
\end{restatable}

\begin{restatable}[Strictly increasing visit frequency]{lem}{strictInc}\label{lem:strict-increase}
If $\exists \gamma \in [0,1): \fpi{s}_{s'}(\gamma) \not \in \set{0,1}$, then $\fpi{s}_{s'}(\gamma)$ is strictly monotonically increasing on $\gamma\in[0,1)$.
\end{restatable}
\begin{proof}
Suppose $\exists \gamma \in [0,1): \fpi{s}_{s'}(\gamma) \not \in \set{0,1}$. This implies that there exists a summand in the Neumann series $\sum_{t=0}^\infty \prn{\gamma\mathbf{T}^\pi}^t\unitvec$ in whose result $s'$ has a positive entry. If the only such summand were at $t=0$, $\fpi{s}_{s'}(\gamma)=0$, which is not the case. Therefore, $t>0$ and the summand must strictly increase with $\gamma\in[0,1)$.
\end{proof}

The intuition for the following result is that if $s\neq s_1$, $s$ can achieve strictly greater discounted $s$-visitation frequency than $s_1$ can.
\begin{restatable}[Each state has a unique visit distribution]{lem}{unique}\label{c-lem:unique-visit}
If $s\neq s_1$ and $\gamma\in[0,1)$, then $\exists \fpi{}(\gamma)\in\Fnd(s,\gamma)\setminus \F(s_1,\gamma)$.
\end{restatable}
\begin{proof}
Given the fixed $\gamma\in[0,1)$, consider the visit distributions of $s$ whose policies always maximize discounted $s$-visitation frequency, no matter which state is the initial state. By \cref{sufficient-nd}, at least one such maximizing visit distribution $\fpi{}(\gamma)$ is non-dominated. Let $s_1\neq s$ and $\fpi[\pi']{}\in\F(s_1)$.
\begin{align}
    \fpi[\pi']{s_1}_{s}(\gamma)&\leq \fpi{s_1}_{s}(\gamma)\label{eq:s-max-pi}\\
    &=\mathbbm{1}_{s_1=s}+\gamma \E{s_2 \sim T(s_1, \pi(s_1))}{\fpi{s_2}_{s}(\gamma)}\\
    &=\gamma \E{s_2 \sim T(s_1, \pi(s_1))}{\fpi{s_2}_{s}(\gamma)}\label{eq:neq-s-prime}\\
    &= \gamma \E{s_2 \sim T(s_1, \pi(s_1))}{\mathbbm{1}_{s_2=s}+ \gamma \E{s_3\sim T(s_2, \pi(s_2))}{\fpi{s_3}_{s}(\gamma)}}\\
    &\leq \gamma \fpi{s}_{s}(\gamma)\label{eq:visitation-advantage}\\
    &< \fpi{s}_{s}(\gamma)\label{eq:strict-gamma}.
\end{align}
\Cref{eq:s-max-pi} follows because $\pi$ maximizes $s$-visitation frequency from every state. \Cref{eq:neq-s-prime} follows because $s_1 \neq s$ ($\mathbbm{1}_{s_1=s}$ is the indicator function which returns $1$ if $s_1=s$ and $0$ otherwise).  \Cref{eq:visitation-advantage} follows because, at best, $s_2=s$ with probability $1$; until $s$ is reached, no visitation frequency is accrued. \Cref{eq:strict-gamma} follows because $\gamma \in (0,1)$.

Since $\forall \fpi[\pi']{s_1}\in\F(s_1): \fpi[\pi']{s_1}_{s}(\gamma)<\fpi{s}_{s}(\gamma)$, $\fpi{s} \not \in \F(s_1)$.
\end{proof}

\subsection{Child distributions}

\begin{restatable}[Child state distributions]{definition}{DefChildDist}\label{def:child-dists}
$\childDists\defeq \set{T(s,a) \mid a \in \A}$.
\end{restatable}

\begin{restatable}[Non-dominated child state distributions]{definition}{DefNdChildDist}\label{def:nd-child-dists}
The \textit{non-dominated child state distributions} are $\childDistsNd\defeq \ND{\childDists}$.
\end{restatable}

\begin{restatable}[When the dynamics are locally deterministic, $\childDists=\childDistsNd$]{lem}{detDynamics}
Suppose $s$ is such that $\forall a\in\A,s'\in\St: T(s,a,s')\in\set{0,1}$. Then $\childDists=\childDistsNd$.
\end{restatable}
\begin{proof}
Local determinism implies that $\childDists$ is a set of one-hot state vectors $\unitvec[s^*]$. Then each $\unitvec[s^*]$ strictly maximizes visitation frequency for $s^*$. Therefore, $\unitvec[s^*]$ is strictly optimal for reward function $\rf\defeq \unitvec[s^*]$ when $\gamma\approx 0$, and so $\unitvec[s^*]\in \childDistsNd$.
\end{proof}

\begin{restatable}[Dominated child state distributions induce dominated visit distribution functions]{lem}{childDomVisitFn}\label{lem:dom-child}
If $T(s,a)\in \childDists\setminus \childDistsNd$ and $\pi(s)=a$, then $\fpi{}\in\F(s)\setminus \Fnd(s)$.
\end{restatable}
\begin{proof}
Let $\dbf \defeq T(s,a)\in \childDists\setminus \childDistsNd$ and $\pi(s)=a$ for some policy $\pi$. If $\fpi{}\in\Fnd(s)$, $\fpi{}$ has to be strictly optimal for some $R$ at some discount rate $\gamma \in (0,1)$. In particular, in order for $\fpi{}$ to be \emph{uniquely optimal} for $R$ at $\gamma$, only actions equivalent  to $a$ at state $s$ can be optimal.

Let $\mathbf{v}^\pi_\gamma \in \rewardVS$ be $\Vf[\pi][R]{\cdot,\gamma}$ in column vector form. Then we must have
\begin{equation}
    \dbf^\top \mathbf{v}^\pi_\gamma > \max_{\dbf'\in \childDists\setminus\set{\dbf}}\dbf'^\top\mathbf{v}^\pi_\gamma.
\end{equation}

This contradicts the fact that $T(s,a)\not\in \childDistsNd$.
\end{proof}

\begin{restatable}[Dynamics characterize the nested structure of visit distribution functions]{lem}{prepend}\label{c-lem:prepend}\strut\\
$T(s,a)\in \childDists$ iff $\set{\unitvec + \gamma \E{s' \sim T(s,a)}{\fpi{s'}} \mid \fpi{s'} \in \F(s' \mid \pi(s)=a)}\subseteq \F(s)$.
\end{restatable}
\begin{proof}
$\childDists=\set{\lim_{\gamma\to 0} \gamma^{-1}\prn{\f_{s,\gamma} - \unitvec} \mid \f_s \in \F(s)}$. The lemma then follows from the linear independence of the canonical unit vectors: $\sum_i \alpha_i \unitvec[s_i]$ is an element of the right-hand set iff it is an element of $\childDists$.
\end{proof}

\begin{restatable}[Normalized value and action-value functions]{definition}{valueFns}{\label{def:value-norm}}
For policy $\pi$,
\begin{equation*}
    \VfNorm[\pi]{R}{s,\gamma} \defeq \lim_{\gamma^*\to \gamma} (1-\gamma^*)\Vf[\pi][R]{s,\gamma^*};
\end{equation*}
this limit exists for all $\gamma \in [0,1]$ by \cref{lem:norm-value-lip}. We similarly define $\OptVfNorm{s,\gamma}$ and $\OptQfNorm{s,a,\gamma}$.
\end{restatable}

\begin{restatable}[$\pwr$ is the average normalized next-state optimal value]{lem}{lemPowEquals}\label{lem:pwr-avg}
\begin{align}
    \pwr[s,\gamma] = \E{R\sim\D}{\max_{\pi\in\Pi} \E{s'\sim T\prn{s,\pi(s)}}{\VfNorm[\pi]{s',\gamma}}}.
\end{align}
\end{restatable}
\begin{proof}
The $\gamma \in (0,1)$ case follows from \cref{eq:pwr-avg-normalized} in the proof of \cref{lem:power-id} and the fact that when $\gamma\in(0,1)$, $\VfNorm[\pi]{s',\gamma}\defeq  \lim_{\gamma^*\to \gamma} (1-\gamma^*)\Vf[\pi][R]{s',\gamma^*}=(1-\gamma)\Vf[\pi]{s',\gamma}$ because on-policy value is continuous on $\gamma\in[0,1)$. The $\gamma=0$ and $\gamma=1$ cases follow by taking the appropriate limits and applying the continuity of $\pwr$.
\end{proof}

\begin{restatable}[Identical $\childDistsNd$ implies equal $\pwr(s,\gamma)$]{prop}{sameSucc}\label{thm:same-succ}
If $\childDistsNd=\childDistsNd[s']$, then $\forall\gamma\in[0,1]:\pwr[s,\gamma]=\pwr[s',\gamma]$.
\end{restatable}
\begin{proof}
Let $\gamma \in (0,1)$. \Cref{lem:pwr-avg} shows that
\begin{align}
    \pwr[s,\gamma] &= \prn{1-\gamma}\E{R\sim\D}{\max_{\pi\in\Pi} \E{s'\sim T\prn{s,\pi(s)}}{\Vf[\pi][R]{s',\gamma}}}\label{eq:pwr-avg-succ}\\
    &= \prn{1-\gamma}\E{R\sim\D}{\max_{a} \E{s'\sim T\prn{s,a}}{\OptVf{s',\gamma}}}\label{eq:same-succ-nonstation}\\
    &= \prn{1-\gamma}\E{R\sim\D}{\max_{\dbf \in \childDists} \E{s'\sim \dbf}{\OptVf{s',\gamma}}}\\
    &= \prn{1-\gamma}\E{R\sim\D}{\max_{\dbf \in \childDistsNd} \E{s'\sim \dbf}{\OptVf{s',\gamma}}}\label{eq:succ-nd-child}\\
    &= \prn{1-\gamma}\E{R\sim\D}{\max_{\dbf' \in \childDistsNd[s']} \E{s'\sim \dbf'}{\OptVf{s',\gamma}}}\label{eq:succ-equality}\\
    &= \pwr[s',\gamma].\label{eq:succ-finish}
\end{align}

\Cref{eq:pwr-avg-succ} follows from \cref{lem:pwr-avg}. Since restriction to stationary policies leaves optimal value unchanged, the non-stationarity in \cref{eq:same-succ-nonstation} leaves optimal value unchanged.

\Cref{lem:dom-child} shows that dominated child distributions induce dominated visit distribution functions. By \cref{cor:opt-vf-restrict-fnd}, restriction to non-dominated visit distribution functions leaves optimal value unchanged for all reward functions, and so only considering $\dbf \in \childDistsNd$ leaves optimal value unchanged. Therefore, \cref{eq:succ-nd-child} follows. \Cref{eq:succ-equality} follows because we assumed that $\childDistsNd=\childDistsNd[s']$. \Cref{eq:succ-finish} follows by the reasoning for \cref{eq:pwr-avg-succ} through \cref{eq:succ-equality}.

The equality holds in the limit as $\gamma\to 0$ or $\gamma\to 1$.
\end{proof}

\Cref{fig:same-succ} demonstrates how \cref{thm:same-succ} establishes $\pwr$ equality, even when the equality is not intuitively obvious.

\begin{figure}[ht]
    \centering
    \begin{tikzpicture}[mygraph,node distance=1.3cm]
        %nodes
            \node[blue] (1) {1};
            \node [black, right of=1, node distance=.75cm, shorten >=0cm, shorten <=.0cm,inner sep=0pt,minimum size=.4pt] (split) {};
            \node[black, above right of=1] (a) {2};
            \node[black, below right of=1] (b) {3};
            \node[black, right of=1, node distance=2cm] (c) {4};
        %arrows
            \path[->] (1) edge (a);
            \path[->] (1) edge (b);
            \path[-] (1) edge node[right] {$\;\; a$} (split);
            \path[-] (split) edge (1);
            \path[->] (split) edge (a);
            \path[->] (split) edge (b);
            \path[->] (a) edge [looseness=7,out=45,in=135] (a);
            \path[->] (b) edge [looseness=7,out=-45,in=-135] (b);
            \path[->] (c) edge [out=170, in=-70] (a);
            \path[->] (a) edge [out=0, in=90] (c);
            \path[->] (c) edge (b);
    \end{tikzpicture}
    \caption[$\pwrNoDist$ equality due to same successor states]{The bifurcated action $a$ is a stochastic transition, where $T(\col{blue}{s_1},a, s_2)=.5=T(\col{blue}{s_1},a, s_3)$. Since $T(\col{blue}{s_1},a)\in \childDists[\col{blue}{s_1}]\setminus \childDistsNd[\col{blue}{s_1}]$, $\childDistsNd[\col{blue}{s_1}]=\childDistsNd[s_4]$ and so \cref{thm:same-succ} implies that $\forall \gamma \in [0,1]: \pwr[\col{blue}{s_1},\gamma]=\pwr[s_4,\gamma]$.}
    \label{fig:same-succ}
\end{figure}

\subsection{Optimal policy set transfer across discount rates}
\begin{restatable}[Blackwell optimal policies \citep{blackwell_discrete_1962}]{definition}{DefBlackwell}\label{Blackwell-definition}
$\optPi[R,1]\defeq \lim_{\gamma\to 1}\optPi$ is the \emph{Blackwell optimal policy set} for reward function $R$.
\end{restatable}

\begin{restatable}[Greedy optimality]{definition}{greedyState}\label{def:greedy-state}
$\greedy$ is the set of policies $\pi$ for which $\forall s$:
\begin{equation*}
    \E{s' \sim T(s,\pi(s))}{R(s')}=\max_a \E{s' \sim T(s,a)}{R(s')}.
\end{equation*}
\end{restatable}

\Cref{fig:no-transfer-greedy} shows that \citet{turner_optimal_2020}'s \cref{transferDiscount} does not always hold for $\gamma=0$, but \cref{prop:almost-transfer} shows that it almost always does hold.

\begin{figure}[ht]
    \centering
    \begin{tikzpicture}[mygraph,shorten >=-3pt,node distance=1.1cm]
        %nodes
            \node (1) {$\col{blue}{s_0}$};
            \node (2) [above right of=1] {$s_1$};
            \node [below of=2, node distance=.4cm] {\col{green}{$1$}};
            \node (3) [below right of=1] {$s_1'$};
            \node [below of=3, node distance=.4cm] {\col{green}{$.5$}};
            \node (4) [right of=2] {$s_2$};
            \node [below of=4, node distance=.4cm] {\col{green}{$0$}};
            \node (5) [right of=3] {$s_2'$};
            \node [below of=5, node distance=.4cm] {\col{green}{$.5$}};
        %arrows
            \path[->] (1) edge node[midway] {\texttt{up}} (2);
            \path[->] (1) edge node[below left] {\texttt{down}} (3);
            \path[->] (2) edge (4);
            \path[->] (3) edge  (5);
            \path[->] (4) edge [looseness=5, out=-45, in=45] (4);
            \path[->] (5) edge [looseness=4, out=-45, in=45] (5);
        \end{tikzpicture}
    \caption[Asymptotically greediness is almost always equivalent to next-state reward maximization]{When $\gamma=\half$, going \texttt{up} and going \texttt{down} are both optimal, with $\OptVf{s_1,\half}=\OptVf{s_1',\half}=1$. However, \cref{transferDiscount} does not hold when the new discount rate is 0: only $\pi^\texttt{down}\in\optPi[\OptVf{\cdot,\half},0]$, with greedy policies preferring to ``gradually'' navigate through higher-value states. When the new discount rate $\gamma=0$, the optimal policies for the constructed reward function are guaranteed to also be optimal for the original by \cref{prop:almost-transfer}, but the converse statement is not true in this {\mdp}.}
    \label{fig:no-transfer-greedy}
\end{figure}

\begin{restatable}[When $\gamma\approx 0$, optimal policies are greedy]{lem}{greedyReward}\label{lem:greedy-reward}
$\optPi[R,0]\subseteq \greedy[R]$.
\end{restatable}
\begin{proof}
Let $\pi^* \in \optPi[R,0]$, $\pi\in\Pi$, and $\gamma \in (0,1)$.
\begin{align}
V^{\pi^*}_R(s,\gamma) &\geq \Vf[\pi][R]{s,\gamma}\\
    R(s) + \gamma \E{s' \sim T\prn{s,\pi^*(s)}}{V^{\pi^*}_{R}(s',\gamma)}&\geq R(s) + \gamma \E{s' \sim T\prn{s,\pi(s)}}{V^{\pi}_{R}(s',\gamma)}\\
    \E{s' \sim T\prn{s,\pi^*(s)}}{V^{\pi^*}_{R}(s',\gamma)}&\geq\E{s' \sim T\prn{s,\pi(s)}}{V^{\pi}_{R}(s',\gamma)}\label{eq:div-by-gamma}\\
    \E{s' \sim T\prn{s,\pi^*(s)}}{R(s') + \gamma \E{s''}{V^{\pi^*}_{R}(s'',\gamma)}} &\geq \E{s' \sim T\prn{s,\pi(s)}}{R(s') + \gamma \E{s''}{V^{\pi}_{R}(s'',\gamma)}}.\label{eq:gamma-close-greedy}
\end{align}

\Cref{eq:div-by-gamma} is valid because $\gamma > 0$.

Let $b \defeq \min_{s^- \in \St} R(s^-), c \defeq \max_{s^+ \in \St} R(s^+)$; these exist because $\St$ is finite. Then $\forall s'': \geom[b] \leq \OptVf{s'',\gamma} \leq \geom[c]$. Because optimal value is thus bounded, \cref{eq:gamma-close-greedy} is controlled by expected next-state reward when $\gamma \approx 0$. The result follows because \cref{def:opt-fn} defines $\optPi[R,0]\defeq \lim_{\gamma\to 0} \optPi$.
\end{proof}

\begin{restatable}[At each $\gamma \in (0,1)$, almost all reward functions have optimal actions at each state which are unique up to action equivalence]{cor}{almostAction}\label{unique-actions}
\end{restatable}
\begin{proof}
Let $s$ be a state. \Cref{unique-opt-value} implies that, for any fixed $\gamma \in (0,1)$ and for almost all reward functions $R$, the optimal action at $s$ is unique up to action equivalence (\cref{def:equiv-action}). Since there are only finitely many states, almost all reward functions $R$ must have a unique-up-to-equivalence optimal action at all states.
\end{proof}

It is optimal to stochastically mix between two actions iff both actions are (deterministically) optimal. \Cref{cor:almost-no-stoch} shows that almost no reward functions have optimal policies which mix between non-equivalent actions.

\begin{restatable}[Almost all reward functions do not have non-trivial stochastic optimal policies]{cor}{almostAllNoStoch}\label{cor:almost-no-stoch}
Let $\gamma\in(0,1)$ and let $X\defeq \set{R\in\rewardSpace \mid \exists s, \pi^*_1,\pi^*_2\in\optPi: \pi^*_1(s) \not \equiv_{s} \pi^*_2(s)}$. Considered as a subset of $\rewardVS$, $X$ has zero Lebesgue measure.
\end{restatable}
\begin{proof}
The result follows directly from \cref{unique-actions}.
\end{proof}

\begin{restatable}[Transferring optimal policy sets to $\gamma=0$]{prop}{transferGreedy}\label{prop:almost-transfer}
Fix $\gamma\in(0,1)$ and let $R$ be a reward function. $\optPi[V^*_R(\cdot,\gamma),0]\subseteq \greedy[V^*_R(\cdot,\gamma)] = \optPi$. Equality holds for almost all $R$; in particular, equality holds when $\forall s \in \St: \exists \f \in \Fnd(s):\f(\gamma)^\top \rf > \max_{\f'\in \Fnd(s)\setminus \set{\f}} \f'(\gamma)^\top \rf$.
\end{restatable}
\begin{proof}
By \cref{lem:greedy-reward}, $\optPi[V^*_R(\cdot,\gamma),0]\subseteq \greedy[V^*_R(\cdot,\gamma)]$. By the definition of an optimal policy, $\pi \in \optPi[R,\gamma]$ iff $\pi$ maximizes value $\Vf[\pi][R]{s,\gamma}$ at all states $s$, which holds iff $\pi$ maximizes the rightmost term of $\OptVf{s,\gamma} \defeq R(s) + \gamma \max_a \E{s' \sim T(s,a)}{\OptVf{s',\gamma}}$ since $\gamma \in (0,1)$. This is true iff $\pi\in\greedy[V^*_R(\cdot,\gamma)]$. Therefore, $\greedy[V^*_R(\cdot,\gamma)] = \optPi$.

\Cref{unique-actions} implies that, for any fixed $\gamma \in (0,1)$ and for almost all reward functions $R$, the optimal action at each state is unique up to action equivalence (\cref{def:equiv-action}). Therefore, $\forall s: \exists \f \in \Fnd(s):\f(\gamma)^\top \rf > \max_{\f'\in \Fnd(s)\setminus \set{\f}} \f'(\gamma)^\top \rf$ for these reward functions $R$.

For such reward functions $R$, a policy $\pi' \in \optPi[R,\gamma]$ iff it induces the appropriate strictly optimal $\f\in\Fnd(s)$ at each state $s$. Let $\pi \in \optPi[V^*_R(\cdot,\gamma),0]$. Since $\pi\in\optPi$ by the above and since optimal action for $R$ at $\gamma$ is unique up to action equivalence, $\pi$ and $\pi'$ must choose equivalent actions at all states. Then $\pi' \in \optPi$. Since $\pi'$ was arbitrary, $\optPi[V^*_R(\cdot,\gamma),0]= \optPi$.
\end{proof}

\begin{restatable}[Can \cref{transferDiscount} be generalized to $\gamma=1$?]{questionSec}{transferGammaOne}
\end{restatable}

\subsection{Optimal policy set characterization}

\begin{restatable}[Set of optimal policy sets]{definition}{optPolSetSet}\label{def:pi-opt-set}
The set of deterministic stationary optimal policy sets is
\begin{equation}
\POpt \defeq \set{\optPi\mid R \in \rewardSpace, \gamma \in [0,1]}.
\end{equation}
\end{restatable}

\begin{figure}[h]
    \centering
    \begin{tikzpicture}[mygraph]
        %nodes
            \node[black] (1) {1};
            \node[black, above right of=1] (a) {A};
            \node[black, below right of=1] (b) {B};
            \node[black, above right of=b] (2) {2};
        %arrows
            \path[->] (1) edge node[left] {\texttt{up}}(a);
            \path[->] (1) edge node[left] {\texttt{down}} (b);
            \path[->] (a) edge [looseness=5,out=45,in=135] (a);
            \path[->] (b) edge [looseness=5,out=-45,in=-135] (b);
            \path[->] (2) edge node[right] {\texttt{up}} (a);
            \path[->] (2) edge node[right] {\texttt{down}} (b);
    \end{tikzpicture}
    \caption[Not all policy sets are valid optimal policy sets]{In general, not all policy sets are valid optimal policy sets ($\POpt\subsetneq \mathcal{P}(\Pi)$). In the above, consider the policy $\pi(s_1)\defeq \texttt{up}, \pi(s_2)\defeq \texttt{down}$. Then $\Pi'\defeq\set{\pi} \not \in \POpt$. If $R(s_A)>R(s_B)$, then no $\pi \in \Pi'$ would choose $\texttt{up}$ over $\texttt{down}$; similar reasoning holds for $R(s_A)<R(s_B)$. If $R(s_A)=R(s_B)$, then $\Pi'$ would contain some  $\pi_2(s_1)=\texttt{down}$.}
    \label{fig:not-all-opt-sets}
\end{figure}

\begin{restatable}[Characterization of $\POpt$]{thm}{optPolSetIff}\label{thm:opt-pol-set-char}
Let $\Pi' \subseteq \Pi$ be a set of deterministic stationary policies. The following are equivalent:
\begin{enumerate}
    \item $\forall \gamma_{1} \in [0,1]: \exists R_{1} \in \rewardSpace: \optPi[R_{1},\gamma_{1}]=\Pi'$.\label{item:forall-policy-set}\\
    \item $\exists \gamma_{2} \in [0,1], R_{2} \in \rewardSpace: \optPi[R_{2},\gamma_{2}]=\Pi'$ (\ie{} $\Pi' \in \POpt$). \label{item:exist-policy-set}\\
    \item $\exists R_{3} \in \rewardSpace: \greedy[R_{3}]=\Pi'$.\label{item:greedy-policy-set}
\end{enumerate}
\Cref{item:forall-policy-set} and \cref{item:exist-policy-set} say that the set of feasible optimal policy sets is invariant to the discount rate. In particular, any optimal policy set can be rationalized as asymptotically greedy or Blackwell-optimal.

\Cref{item:exist-policy-set} and \cref{item:greedy-policy-set} show that a policy set is an optimal policy set for some $R_{2}$ at some $\gamma_{2}$ iff that policy set can be rationalized as greedily optimizing some $R_{3}$.
\end{restatable}
\begin{proof}
\ref{item:forall-policy-set} $\implies$ \ref{item:exist-policy-set} by \cref{def:pi-opt-set}.

\textbf{\ref{item:exist-policy-set} $\implies$ \ref{item:greedy-policy-set}.}
Suppose \cref{item:exist-policy-set} holds. If $\gamma_{2} \in (0,1)$, define $R_{3}(s)\defeq \OptVf[R_{2}]{s,\gamma_{2}}$. Then apply \cref{prop:almost-transfer} to conclude that $\greedy[R_{3}]=\optPi[R_{2},\gamma_{2}]=\Pi'$.

Since $\optPi[R_{2},1]\defeq \lim_{\gamma \to 1} \optPi[R_{2},\gamma]$ (\cref{def:opt-fn}) always exists by \cref{cor:opt-pi-one-side} and only finitely many optimal policy shifts occur (\cref{lem:opt-pol-shift-bound}), $\forall R: \exists \gamma \in (0,1): \optPi[R,\gamma]=\optPi[R,1]$. Select such a $\gamma$ for $R_{2}$. Since $\gamma \in (0,1)$, we have just shown that $\exists R_{3}:\Pi'=\optPi[R_{2},1]=\optPi[R_{2},\gamma]=\greedy[R_{3}]$.

The $\gamma_{2}=0$ case follows by the same reasoning. Then \cref{item:exist-policy-set} implies \cref{item:greedy-policy-set}.

\textbf{\ref{item:greedy-policy-set} $\implies$ \ref{item:forall-policy-set}.} Suppose \cref{item:greedy-policy-set} holds. Let $\pi \in \greedy[R_{3}]$ and $\gamma_{1} \in (0,1)$. Define $R_{1} \defeq \prn{\mathbf{I}-\gamma_{1}\mathbf{T}^{\pi}}R_{3}$. By reasoning identical to that in the proof of \cref{transferDiscount}, $\exists R_{1} \in \rewardSpace: \optPi[R_{1},\gamma_{1}]=\greedy[R_{3}]=\Pi'$.

Now suppose $\gamma_{1}=1$ and consider the function $R_{1}(s,\gamma_{1})\defeq \prn{\mathbf{I}-\gamma_{1}\mathbf{T}^{\pi}}R_{3}(s)$. Since $\forall \gamma \in (0,1): \optPi[R_{1}(\cdot, \gamma),\gamma]=\Pi'$ by the above reasoning, $\optPi[R_{1},1]=\lim_{\gamma \to 1}\optPi[R_{1}(\cdot, \gamma),\gamma]=\Pi'$ (by \cref{def:opt-fn}).

Similar reasoning holds when $\gamma_{1}=0$. Since this equality can be satisfied for any $\gamma_{1} \in [0,1]$, \cref{item:greedy-policy-set} implies \cref{item:forall-policy-set}.
\end{proof}

Recall the following definition:
\DefChildDist*
Given a starting state $s$, we can consider its available child state distributions $\childDists$ to be lotteries over the next state the agent will visit. A set of policies $\Pi'$ can be seen as preferring some state lotteries (by containing policies which take a certain action at $s$) over others (lotteries which are not induced by any $\pi \in \Pi'$ at state $s$).

\begin{restatable}[State lottery relation implied by a policy set]{definition}{lotteryOrder}\label{def:lottery-order}
The relations implied by policy set $\Pi' \subseteq \Pi$ is defined as follows. Consider $s$ such that $L_1, L_2 \in \childDists$ (\ie{} the lotteries are realizable at state $s$), where $\exists \pi_1 \in \Pi': T(s,\pi_1(s))=L_1$.

$L_1 \polOrderSim L_2$ if $\exists \pi_2 \in \Pi': T(s,\pi_2(s))=L_2$. In other words, $L_1$ and $L_2$ are both induced by $\pi_1,\pi_2\in\Pi'$, respectively. This implies that $L_1$ and $L_2$ are ``equally good,'' since $\Pi'$ induces both of them.

$L_1 \polOrderStrict L_2$ if $\lnot \exists \pi_2 \in \Pi': T(s,\pi_2(s))=L_2$. Since no policy induces $L_2$, $L_2$ is considered to be strictly worse than $L_1$.
\end{restatable}

However, some $\Pi'$ imply incoherent preferences over states: in \cref{fig:not-all-opt-sets}, $\Pi'$ implies that $s_A \polOrderStrict s_B$ (via $s_1$) and also $s_B \polOrderStrict s_A$ (via $s_2$'s).

\begin{restatable}[Coherent state lottery relations]{definition}{coherentOrder}\label{def:coherent}
$(\polOrderSim,\polOrderStrict)$ is \emph{coherent} when $\exists \rf \in \rewardVS$:
\begin{enumerate}
    \item $L_1 \polOrderSim L_2$ implies that $L_1^\top \rf = L_2^\top \rf$.
    \item $L_1 \polOrderStrict L_2$ implies that $L_1^\top \rf > L_2^\top \rf$.
\end{enumerate}
\end{restatable}

\begin{restatable}[Optimal policy sets imply coherent preference relations]{thm}{cohOptPol}\label{thm:coh-opt-pol}
If $\Pi' \in \POpt$, then $(\polOrderSim,\polOrderStrict)$ is coherent.
\end{restatable}
\begin{proof}
If $\Pi' \in \POpt$, then let $R$ be a reward function for which $\greedy[R]=\Pi'$ (such a reward function exists by \cref{thm:opt-pol-set-char}). By the definition of $\greedy$ (\cref{def:greedy-state}), $\pi \in \greedy$ iff $\forall s \in \St:$ $\pi(s)$ maximizes expected next-step reward.

Suppose $L_1 \polOrderSim L_2$. Then by \cref{def:lottery-order}, $\exists \pi_1, \pi_2 \in \Pi': \prn{T(s,\pi_1(s))=L_1}\land \prn{T(s,\pi_2(s))=L_2}$. By the greediness of $\Pi'$, this implies that both $L_1$ and $L_2$ maximize next-step reward. Therefore, $T(s,\pi_1(s))^\top \rf = T(s,\pi_2(s))^\top \rf$ and so $L_1^\top \rf = L_2^\top \rf$.

Suppose $L_1 \polOrderStrict L_2$. Then by \cref{def:lottery-order}, there $\exists s\in\St,\pi_1 \in \Pi':T(s,\pi_1(s))=L_1$ and $L_2 \in \childDists$, but $\lnot \exists \pi_2 \in \Pi': T(s,\pi_2(s))=L_2$. Since there is no $\pi_2\in\greedy$ taking some action $a_2$ such that $T(s,a_2)=L_2$, and since such an action exists ($L_2 \in \childDists$), we conclude that $a'$ is not a greedy action at state $s$ for $R$. In other words, $T(s,\pi_1(s))^\top \rf> T(s,a')^\top \rf$ and so $L_1^\top \rf > L_2^\top \rf$.

Since $L_1 \polOrderSim L_2$ and $L_1 \polOrderStrict L_2$ were arbitrary, $(\polOrderSim,\polOrderStrict)$ is coherent for $R$.
\end{proof}

The \textsc{vnm} utility theorem implies that agents with a coherent preference ordering over state lotteries can be rationalized as maximizing the expected utility of some utility function. Somewhat similarly, \cref{thm:coh-opt-pol} shows that optimal policy sets can be rationalized as coherently trading off the greedy values of environment states.

For incoherent $\Pi'$, it is not generally possible to rectify the incoherence of $(\polOrderSim,\polOrderStrict)$ by deducing a reward function for which $\Pi'$ ``should'' be greedy. For example, in \cref{fig:not-all-opt-sets}, it is unclear what preferences over state lotteries $\Pi'$ ``should'' have.

\begin{restatable}[Coherent relations imply greedy policy subset]{conjSec}{greedySubsetCoherent}
If $(\polOrderSim,\polOrderStrict)$ is coherent for $R$, then $\Pi' \subseteq \greedy$.
\end{restatable}

\subsubsection{Optimal policy sets factorize}
We now further constrain the structure of optimal policy sets.
\begin{restatable}[Actions taken by a policy set at a state]{definition}{actionsState}\label{def:actions-state} Let $\Pi' \subseteq \Pi$.
\begin{equation}
    \polActions\defeq \set{a \in \A \mid \exists \pi\in\Pi': \pi(s)=a}.
\end{equation}
\end{restatable}

\begin{restatable}[Optimal policy sets mix-and-match optimal actions]{lem}{mixActionOpt} \label{lem:mix-match} Let $\pi_1, \pi_2 \in \Pi' \in \POpt$ and consider state-space partition $\St = S_1 \sqcup S_2$. Then $\exists \pi_3 \in \Pi'$ such that $\forall s_1 \in S_1: \pi_3(s_1)=\pi_1(s_1)$ and $\forall s_2 \in S_2: \pi_3(s_2)=\pi_2(s_2)$.
\end{restatable}
\begin{proof}
Since $\Pi'\in\POpt$, $\exists R\in\rewardSpace: \Pi'=\greedy$ by \cref{thm:opt-pol-set-char}. By the definition of $\greedy$ (\cref{def:greedy-state}), each action of $\pi_1$ and $\pi_2$ maximizes expected next-state reward. Therefore, any $\pi_3$ agreeing with $\pi_1$ on $S_1$ and with $\pi_2$ on $S_2$ must also maximize expected next-state reward for all states in the {\mdp}. Then $\pi_3\in\greedy=\Pi'$.
\end{proof}

\begin{restatable}[If $\Pi'\in\POpt$, then $\Pi'\cong \prod_{s} \polActions$]{prop}{optSetFactor}
There exists a set bijection between $\Pi'$ and $\prod_{s} \polActions$.
\end{restatable}
\begin{proof}
Apply \cref{lem:mix-match} to conclude that any combination of actions in $\prod_{s} \polActions$ maps to an optimal policy in $\Pi'$. By inverting the same map, any policy in $\Pi'$ must map to a unique element of $\prod_{s} \polActions$ by the definition of $\polActions$ (\cref{def:actions-state}).
\end{proof}

This implies that optimal policy sets admit compressed representations: instead of explicitly storing the outputs of (up to) $\abs{\A}^{\abs{\St}}$ optimal policies at $\abs{\St}$ states, we can simply record (up to) $\abs{\A}$ optimal actions at $\abs{\St}$ states. Then, we regenerate the full set $\Pi'$ by taking the Cartesian product of the optimal actions.

This immediately suggests divisibility tests which rule out certain policy sets from being optimal policy sets.

\begin{restatable}[If $\Pi'\in\POpt$, then $\forall s: \abs{\polActions}$ divides $\abs{\Pi'}$]{cor}{optSetDivis}\label{cor:opt-set-divis}
\end{restatable}

\begin{restatable}[Optimal policy sets take all equivalent actions]{lem}{equivActionOpt} \label{lem:equiv-action-opt} Let $\Pi' \in \POpt$. For any state $s$ and action $a$, if $a \in \polActions$ and $a' \equiv_s a$, then $a' \in \polActions$.
\end{restatable}
\begin{proof}
Since $\Pi'\in\POpt$, $\exists R\in\rewardSpace: \Pi'=\greedy$ by \cref{thm:opt-pol-set-char}. If $a \in \polActions$ and $a' \equiv_s a$, $a$ is greedily optimal at $s$. Then $a'\equiv_s a$ must also be greedily optimal because $T(s,a)=T(s,a')$ by \cref{def:equiv-action}. Therefore, $a' \in \polActions$.
\end{proof}

\begin{restatable}[Multiple optimal actions at multiple states implies that the optimal policy set has composite cardinality]{prop}{optSetMultNoPrime}\label{prop:mult-actions-opt-composite}
Let $\Pi'\in\POpt$. If $\exists s \neq s': \abs{\polActions}\cdot\abs{\polActions[s']}$, then $\abs{\Pi'}$ is composite.
\end{restatable}
\begin{proof}
By \cref{cor:opt-set-divis}, $\abs{\polActions}\cdot\abs{\polActions[s']}$ divides $\abs{\Pi'}$. Since both of these factors are greater than $1$, $\abs{\Pi'}$ is a composite number.
\end{proof}

\begin{restatable}[If $\forall s\in\St, a \in \A: \exists a'\neq a: a\equiv_s a'$ and if $\abs{\St}>1$, then no optimal policy sets have prime cardinality]{cor}{optSetNoPrime}
\end{restatable}
\begin{proof}
Let $\Pi' \in \POpt$. Let $s\neq s'$ be distinct states. $\Pi'$ is nonempty by the existence of a stationary deterministic optimal policy for all reward functions and discount rates. Let $\pi\in\Pi'$. Since $\forall s''\in\St, a \in \A: \exists a'\neq a: a\equiv_{s''} a'$, apply \cref{lem:equiv-action-opt} to conclude that $\abs{\polActions},\abs{\polActions[s']}>1$. Apply \cref{prop:mult-actions-opt-composite}.
\end{proof}

\subsubsection{Deciding whether a policy set is optimal}

\Cref{thm:coh-opt-pol} allows us to efficiently decide whether a policy set is optimal for some $(R,\gamma)$  tuple (since enumeration is impossible for the uncountably many such tuples). I initially conjectured \cref{conj:poly-decide-coherent} in 2020. \citet{abel2021expressivity}'s Theorem 4.3 has since proven a generalization of it.

\begin{restatable}[$\Pi' \in \POpt?$ can be efficiently decided]{conjSec}{polyDecideCoherent} \label{conj:poly-decide-coherent}
The decision problem corresponding to deciding whether $\Pi' \in \POpt$ is in $\mathbf{P}$.
\end{restatable}

\subsection{How reward function combination affects optimality}
\begin{restatable}[Non-negative combination of reward functions preserves value ordering agreement]{lem}{preserve}\label{c-lem:order-preserve}
If $\Vf[\pi][R_1]{s,\gamma}\geq \Vf[\pi'][R_1]{s,\gamma}$ and $\Vf[\pi][R_2]{s,\gamma}\geq \Vf[\pi'][R_2]{s,\gamma}$, then for any $\alpha,\beta \geq 0$,
\begin{equation}\Vf[\pi][\alpha R_1 + \beta R_2]{s,\gamma}\geq \Vf[\pi'][\alpha R_1 + \beta R_2]{s,\gamma}.
\end{equation}
\end{restatable}
\begin{proof}
The premise implies that $\fpi{s}(\gamma)^\top \rf_1 \geq \fpi[\pi']{s}(\gamma)^\top \rf_1$ and $\fpi{s}(\gamma)^\top \rf_2 \geq \fpi[\pi']{s}(\gamma)^\top \rf_2$. Then
\begin{align}
    \Vf[\pi][\alpha R_1 + \beta R_2]{s,\gamma}&=\fpi{s}(\gamma)^\top\prn{\alpha \rf_1 + \beta\rf_2}\\
    &\geq\fpi[\pi']{s}(\gamma)^\top\prn{\alpha \rf_1 + \beta\rf_2}\\
    &= \Vf[\pi'][\alpha R_1 + \beta R_2]{s,\gamma}.
\end{align}
\end{proof}

\begin{restatable}[Non-negative combination of reward functions preserves optimal policy agreement]{cor}{optCombo}\label{c-lem:optimal-preserve}
If $\pi^*$ is optimal for $R_1$ and $R_2$ at discount rate $\gamma$, it is also optimal  for $\alpha R_1 + \beta R_2$ at discount rate $\gamma$ for any $\alpha,\beta \geq 0$.
\end{restatable}
However, $R_1$ and $R_2$ having the same optimal policy set doesn't mean they will incentivize the same distribution over learned policies.

\begin{restatable}[If two reward functions induce the same ordering over policies, when do they tend to incentivize similar learned policies?]{questionSec}{learned-policy}\label{questionSec:learned}
\end{restatable}

\begin{remark}
With respect to a distribution over network initializations and a fixed learning process, the distribution over learned policies is sometimes invariant to positive affine transformation of the reward function (so long as no \eg{} underflow or overflow errors occur, or instability does not occur). In contrast, reward shaping \citep{Ng99policyinvariance} often accelerates learning and changes which policies get tend to get learned after a fixed number of policy improvement steps. Both transformations preserve policy ordering, but only shaping improves learning. What is the fundamental difference?
\end{remark}

\subsection{Visit distribution function agreement}

\Cref{fig:same-dist} illustrates how two distinct visit distribution functions can output the same visit distribution for certain $\gamma$.

\begin{figure}[ht]
    \centering
     \begin{tikzpicture}[mygraph]
        %nodes
            \node[blue] (1) {$1$};
            \node[red] (2) [right of=1] {$2$};
            \node[black] (3) [above of=2] {$3$};
        %arrows
            \path[->] (1) edge (2);
            \path[->] (1) edge (3);
            \path[->] (2) edge [looseness=1, out=70, in=-70, color=lightgray] (3);
            \path[->] (3) edge [out=-110, in=110,color=lightgray] (2);
            \path[->] (2) edge [looseness=5, out=0, in=-90] (2);
            \path[->] (3) edge [looseness=5, out=45, in=135] (3);
    \end{tikzpicture}
     \caption[\mbox{$\forall \f,\f'\in\Fnd(s):\f\neq\f'\implies \forall\gamma\in(0,1): \f(\gamma)\neq\f'(\gamma)$}]{When $\gamma=\half$, the trajectories $\col{blue}{s_1}\col{red}{s_2}s_3 s_3\ldots$ and $\col{blue}{s_1}s_3\col{red}{s_2}\col{red}{s_2}\ldots$ induce the same state visitation distribution of $\begin{pmatrix}\col{blue}{1} & \col{red}{\half} & \half\end{pmatrix}^\top$. Both trajectories are induced by dominated state visit distribution functions; \cref{lem:no-agree} shows that this is no coincidence.}
    \label{fig:same-dist}
\end{figure}

\begin{restatable}[Distinct visit distribution functions agree finitely many times]{lem}{finiteAgree}\label{c-lem:finite-agreement}
Let $s,s'$ be any two states and let $\f\in\F(s), \f'\in\F(s')$ be distinct visit distribution functions. For all but finitely many $\gamma \in (0,1)$, $\f(\gamma)\neq \f'(\gamma)$.
\end{restatable}
\begin{proof}
By \cref{f-rat}, each visit distribution function is multivariate rational on $\gamma$. If the functions are distinct, then they must disagree on at least one output dimension. Their difference along this dimension is a rational non-zero function, which has finitely many roots by the fundamental theorem of algebra.
\end{proof}

\begin{restatable}[Different states have disjoint visit distribution function sets]{lem}{diffStateDiffVisitFn}
If $s\neq s'$, $\lnot \exists \f\in\F(s), \f'\in\F(s'):\forall \gamma\in(0,1):\f(\gamma)=\f(\gamma)$.
\end{restatable}
\begin{proof}
When $\gamma \approx 0$, the visit distributions approximate the appropriate unit vectors: $\f(\gamma)\approx \unitvec$ and $\f'(\gamma)\approx \unitvec[s']$. Since $s\neq s'$, $\f(\gamma)\neq \f'(\gamma)$ when $\gamma \approx 0$. Therefore, $\f$ and $\f'$ are distinct visit distribution functions. Apply \cref{c-lem:finite-agreement}.
\end{proof}

\begin{restatable}[Visit distributions are distinct at all but finitely many $\gamma$]{cor}{distinctAlmostAlways}\label{cor:distinct-almost-always}
$\forall s, s' \in \St, \f \in \F(s), \f'\in \F(s'):$  either $\f = \f'$ or $\f(\gamma) \neq \f'(\gamma)$ holds for all but finitely many $\gamma\in(0,1)$.
\end{restatable}
\begin{proof}
By \cref{c-lem:finite-agreement}, the distinct visit distributions at $s$ agree with each other for at most finitely many $\gamma$. Likewise, the visit distributions of other states $s'$ agree with $\f$ for at most finitely many $\gamma$. Since there are only finitely many visit distribution pairs, there is a finite set of the $\gamma$ at which any two distinct visit distributions agree.
\end{proof}

\begin{restatable}[Cross-state linear independence of visit distribution functions]{prop}{linIndepCross}\label{prop:lin-indep-cross}
$\f\in\F(s)$ cannot be written as a linear combination of the visit distribution functions of other states.
\end{restatable}
\begin{proof}
Suppose $\f=\sum_i\alpha_i \f_{s_i}$ such that $\forall i: \alpha_i \in \mathbb{R} \land s_i\neq s \land \f_{s_i}\in\F(s_i)$. Then
\begin{align}
    \lim_{\gamma\to 0} \sum_i \alpha_i \f_{s_i}&=\sum_i \alpha_i \unitvec[s_i]\\
    &\neq \unitvec\label{eq:li-cross-state}\\
    &=\lim_{\gamma\to 0} \f(\gamma),
\end{align}
where \cref{eq:li-cross-state} follows because $s_i \neq s$ and the canonical basis vectors are linearly independent.
\end{proof}

\Cref{fig:lin-dependence} shows that \cref{prop:lin-indep-cross} does not hold amongst the visit distribution functions of a single state.

\begin{figure}[ht]
    \centering
    \begin{tikzpicture}[mygraph,node distance=1.3cm]
        %nodes
            \node[blue] (1) {1};
            \node[black, right of=1, node distance=.75cm, shorten >=0cm, shorten <=.0cm,inner sep=0pt,minimum size=.4pt] (split) {};
            \node[black, above right of=1] (a) {2};
            \node[black, below right of=1] (b) {3};
        %arrows
            \path[->] (1) edge node[left] {\texttt{up}} (a);
            \path[->] (1) edge node[left] {\texttt{down}} (b);
            \path[-] (1) edge node[right] {$\;\; a$} (split);
            \path[-] (split) edge (1);
            \path[->] (split) edge (a);
            \path[->] (split) edge (b);
            \path[->] (a) edge [looseness=7,out=45,in=-45] (a);
            \path[->] (b) edge [looseness=7,out=-45,in=45] (b);
    \end{tikzpicture}
    \caption[A state's visit distributions aren't necessarily linearly independent]{The bifurcated action $a$ is a stochastic transition, where $T(\col{blue}{s_1},a, \col{black}{s_2})=.5=T(\col{blue}{s_1},a, \col{black}{s_3})$. $\fpi[\pi_a]{\col{blue}{s_1}}=.5(\fpi[\pi_\texttt{up}]{\col{blue}{s_1}}+\fpi[\pi_\texttt{down}]{\col{blue}{s_1}})$.}
    \label{fig:lin-dependence}
\end{figure}

\Cref{lem:no-agree} is generalized by \cref{prop:conv-combo}.

\begin{restatable}[At every $\gamma \in (0,1)$, non-dominated visit distributions are outside of the convex hull of any set of other visit distributions]{prop}{convVisitCombo}\label{prop:conv-combo}
Let $\f \in \F(s)$. If there exists $\gamma\in(0,1)$ and $\f_1,\ldots, \f_k \in \F(s)\setminus \set{\f}$ such that $\f$ can be expressed as the convex combination $\f(\gamma)=\sum_{i=1}^k \alpha_i \f_i(\gamma)$, then $\f \not \in \Fnd(s)$.
\end{restatable}
\begin{proof}
If some $\alpha_i=1$, then $\f \not \in \Fnd(s)$ by \cref{lem:no-agree}. Suppose at least two $\alpha_i > 0$. Then the conclusion follows by \cref{prop:convex-independence} (letting $X\defeq \set{\f(\gamma)\mid \f \in \F(s)},\x\defeq \f(\gamma)$).
\end{proof}

\begin{restatable}[Geometry of optimality support]{lem}{geomProp}\label{geom-prop}
At any $\gamma\in(0,1)$, $\opt$ (the set of reward functions  for which $\f\in\F(s)$ is optimal) is both a closed convex polytope and a pointed convex cone.
\end{restatable}
\begin{proof}
$\opt$ is a closed convex polytope because it is the intersection of half-spaces: $\f(\gamma)^\top \rf\geq \max_{\f'\in\F(s)\setminus \set{\f}} \f'(\gamma)^\top\rf$. The set is a pointed cone because for any $\alpha \geq 0$, $\f(\gamma)^\top \rf\geq \max_{\f'\in\F(s)\setminus \set{\f}} \f'(\gamma)^\top\rf$ implies $\f(\gamma)^\top (\alpha \rf)\geq \max_{\f'\in\F(s)\setminus \set{\f}} \f'(\gamma)^\top(\alpha  \rf)$.
\end{proof}

\begin{restatable}[Visit dist.\ function convex hull intersection at any $\gamma$ implies shared optimality status]{prop}{optLinkConvex}\label{thm:opt-link-convex}
Let $\f \in \F(s)$. Suppose there exists $\gamma\in(0,1)$ and $\f_1,\ldots, \f_k \in \F(s)\setminus \set{\f}$ such that $\f$ can be expressed as the convex combination $\f(\gamma)=\sum_{i=1}^k \alpha_i \f_i(\gamma)$ (where each $\alpha_i>0$).

$\f$ is optimal for reward function $R$ at $\gamma^*\in (0,1)$ iff $\forall i: \f_i$ is optimal for $R$ at $\gamma^*$.
\end{restatable}
\begin{proof}
If $\forall i: \f_i$ is  optimal for $R$ at $\gamma^*$, then $\f$ is also optimal for $R$ at $\gamma^*$ by \cref{geom-prop}.

Suppose that $\f$ is optimal for $R$ at $\gamma^*\in (0,1)$, while for some $i$, $\f_i$ is not optimal for $R$ at $\gamma^*$.  In particular, suppose $R$ induces optimal policy set $\Pi_{\gamma^*}$ at discount rate $\gamma^*$. Then by \cref{transferDiscount}, we can construct a reward function $R'$ which has optimal policy set $\Pi_{\gamma^*}$ at discount rate $\gamma$.

Suppose $\alpha_i=1$. Since $\f_i$ was not optimal at $\gamma^*$, $\f_i$ isn't optimal at $\gamma$ either.  But this is absurd, since then $\f(\gamma)=\f_i(\gamma)$ and so they must both be optimal at $\gamma$, a contradiction. This means that $\f_i$ is optimal for $R$ at $\gamma^*$.

Suppose $\alpha_i<1$. As $\f_i$ is not optimal for $R$ at $\gamma$ but $\f$ is, there must be another $\f_{i'}$ such that $\f_{i'}(\gamma)^\top\rf \geq \f(\gamma)^\top \rf > \f_i(\gamma)^\top \rf$. Since we assumed that $\alpha_i>0$, the first inequality must be strict. This means that $\f$ cannot be optimal for $R$ at $\gamma$, a contradiction. Therefore, $\f_i$ must be optimal for $R$ at $\gamma^*$.
\end{proof}

\begin{restatable}[Visit distribution functions which agree at any $\gamma$, must be optimal together]{cor}{linkedOptimality}\label{cor:link-optimality}
Let $\f, \f'\in\F(s)$. Suppose $\exists \gamma\in(0,1): \f(\gamma)=\f'(\gamma)$. Then $\forall \gamma^*\in(0,1)$:
\begin{equation*}
    \opt[\f(\gamma^*)\geq \F(s,\gamma^*)]=\opt[\f'(\gamma^*)\geq \F(s,\gamma^*)].
\end{equation*}
\end{restatable}
\begin{proof}
Apply \cref{thm:opt-link-convex}.
\end{proof}

\Cref{fig:link} illustrates the power of \cref{cor:link-optimality}.

\begin{figure}[ht]
    \centering
    \begin{tikzpicture}[mygraph]
        %nodes
            \node[blue] (1) {$1$};
            \node[red] (2) [right of=1] {$2$};
            \node[black] (3) [above of=2] {$3$};
        %arrows
            \path[->] (1) edge (2);
            \path[->] (1) edge (3);
            \path[->] (2) edge [looseness=1, out=70, in=-70, color=lightgray] (3);
            \path[->] (3) edge [out=-110, in=110,color=lightgray] (2);
            \path[->] (2) edge [looseness=5, out=0, in=-90] (2);
            \path[->] (3) edge [looseness=5, out=45, in=135] (3);
    \end{tikzpicture}
     \caption[Two trajectories which are optimal if and only if the other is]{When $\gamma=\half$, the trajectories $\col{blue}{s_1}\col{red}{s_2}s_3 s_3\ldots$ and $\col{blue}{s_1}s_3\col{red}{s_2}\col{red}{s_2}\ldots$ induce the same state visitation distribution. Therefore, \cref{cor:link-optimality} shows that at \emph{any} $\gamma\in(0,1)$, one trajectory is optimal iff the other trajectory is. Basic algebra confirms the point: one of these trajectories is optimal iff $R(\col{red}{s_2})=R(s_3)$; in that case, every policy is optimal.}
    \label{fig:link}
\end{figure}

Consider again \cref{transferDiscount}:

\begin{thm*}[A means of transferring optimal policy sets across discount rates]
Suppose reward function $R$ has optimal policy set $\optPi[R,\gamma]$ at discount rate $\gamma\in(0,1)$. For any $\gamma^*\in(0,1)$, we can construct a reward function $R'$ such that $\optPi[R',\gamma^*]=\optPi$.
\end{thm*}

\Cref{fig:order-not-preserved} shows that $\Vf[\pi][R]{s,\gamma}\geq \Vf[\pi'][R]{s,\gamma}$ does not imply $\Vf[\pi][R']{s,\gamma^*}\geq \Vf[\pi'][R']{s,\gamma^*}$. The new reward function $R'$ is only guaranteed to have the same optimal policy set, not the same policy ordering.

\begin{figure}[h]
    \centering
    \captionsetup[subfigure]{labelformat=empty}
    \subfloat[]{ \raisebox{-.45\height}{
    \begin{tikzpicture}[mygraph]
        %nodes
            \node[blue] (1) {$1$};
            \node[green] (2) [left of=1] {$2$};
            \node[black] (3) [below of=1] {$3$};
            \node[black] (4) [right of=1] {$4$};
        %arrows
            \path[->] (1) edge node[midway, above] {\texttt{left}} (2);
            \path[->] (1) edge node[midway, left] {\texttt{down}} (3);
            \path[->] (2) edge [looseness=5, out=-180, in=90] (2);
            \path[->] (3) edge [looseness=5, out=-90, in=180] (3);
            \path[<->] (1) edge node[midway] {\texttt{right}} (4);
    \end{tikzpicture}}}\hspace{45pt}\subfloat[]{\begin{tabular}{rcc}
        \toprule
        State              & $R$ & $R'$                                        \\
        \midrule
        $\col{blue}{s_1}$  & $0$ & $\dfrac{\gamma-\gamma^*}{1-\gamma}$         \\
        $\col{green}{s_2}$ & $1$    & $1$                                            \\
        $s_3$              & $0$ & $0$                                         \\
        $s_4$              & $0$ & $\dfrac{\gamma(\gamma-\gamma^*)}{1-\gamma}$ \\
        \bottomrule
    \end{tabular}}
    \caption[Translating the optimal policy set across discount rates does not always preserve the policy ordering]{$R$ and the transformed $R'\defeq \prn{\mathbf{I}-\gamma^*\mathbf{T}^{\pi_\texttt{left}}}V^*_R(\cdot, \gamma)$ for which \cref{transferDiscount}'s reward transformation does not preserve the policy ordering. Even though $V^{\pi_\texttt{down}}_R(\col{blue}{s_1},\gamma)=V^{\pi_\texttt{right}}_R(\col{blue}{s_1},\gamma)=0$, when $\gamma^*>\gamma$, $V^{\pi_\texttt{down}}_{R'}(\col{blue}{s_1},\gamma)>V^{\pi_\texttt{right}}_{R'}(\col{blue}{s_1},\gamma)$.}
    \label{fig:order-not-preserved}
\end{figure}

Let $R$ be a reward function and $\gamma^*\in(0,1),\gamma\in[0,1)$. \Cref{fig:order-not-preserved} demonstrates that there does not always exist $R'$ such that $\forall s, \pi: \Vf[\pi][R]{s,\gamma^*}=\Vf[\pi][R']{s,\gamma}$. By preserving the optimal value function across discount rates, $R'$ necessarily differs on its other value functions.

\begin{restatable}[Does there exist a transformation like that of \cref{transferDiscount} which preserves the entire policy ordering of $R$? ]{questionSec}{qOrdering}
\end{restatable}

\begin{restatable}[Can we characterize the permissible policy orderings in a given {\mdp}? Does the discount rate affect the permissible orderings?]{questionSec}{QCharOptOrder}
\end{restatable}

\begin{restatable}[Almost all reward functions induce strict non-dominated visitation distribution orderings]{prop}{almostAllOrder}\label{prop:almost-all-order}
Let $s$ be a state and $\gamma\in(0,1)$.
\begin{equation}\set{\rf\mid \rf\in\reals^{\abs{\St}}, \exists \f, \f'\in\Fnd(s): \f\neq \f' \land \f(\gamma)^\top\rf = \f'(\gamma)^{\top}\rf}
\end{equation}
has Lebesgue measure zero. The same statement holds for $\F$ instead of $\Fnd$ for all but finitely many $\gamma\in(0,1)$.
\end{restatable}
\begin{proof}
Since $\f,\f'\in\Fnd(s)$ are distinct, $\forall \gamma\in(0,1):\f(\gamma)\neq \f'(\gamma)$ by \cref{lem:no-agree}. Apply \cref{lem:distinct-lin-prob}.

All distinct $\f,\f'\in\F(s)$ disagree for all but finitely many $\gamma$ by \cref{cor:distinct-almost-always}. Then apply \cref{lem:distinct-lin-prob} for these $\gamma$ at which they all disagree, and the second claim follows.
\end{proof}

\begin{remark}
For almost all reward functions, \cref{prop:almost-all-order} prohibits value equality anywhere in the non-dominated visit distribution ordering. In contrast, \cref{unique-optimal} only shows that almost all reward functions have a unique \emph{optimal} visit distribution.
\end{remark}

\begin{restatable}[Reward negation flips the visit distribution ordering]{lem}{flipOrder}\label{lem:flip-order}
Let $s$ be a state, $\f,\f'\in\F(s)$, $\rf\in\rewardVS$, and $\gamma\in(0,1)$. $\f(\gamma)^\top \rf \geq \f'(\gamma)^\top \rf$ iff $\f(\gamma)^\top (-\rf) \leq \f'(\gamma)^\top (-\rf)$. In particular, optimal policies for $\rf$ minimize value for $-\rf$.
\end{restatable}
\begin{proof}
This follows directly from the fact that $\forall a,b \in \reals: a \geq b$ iff $-a \leq -b$.
\end{proof}

\subsection{Visit distributions functions induced by non-stationary policies}

\begin{restatable}[Non-stationary visit distribution functions]{definition}{nonStat}\label{def:non-stat-visit}
Let $\Pi^\text{HD}$ be the set of  history-dependent (or non-stationary) deterministic policies \citep{puterman_markov_2014}. $\Fnonstat(s)\defeq \set{\fpi{s} \mid \pi \in \Pi^\text{HD}}$.
\end{restatable}

We continue using $\Pi$ to denote the space of deterministic stationary policies. Clearly, $\F(s)\subseteq \Fnonstat(s)$.

\begin{restatable}[Visit distribution functions induced by non-stationary or non-deterministic policies are dominated]{prop}{domNonStat}
\end{restatable}
\begin{proof}
This follows from the fact that every reward function has a stationary, deterministic optimal policy.
\end{proof}

\begin{restatable}[$\Fnonstat(s)$ is finite iff stationarity is irrelevant]{prop}{iffFiniteFnonstat}\label{prop:iff-Fnonstat-finite}
$\Fnonstat(s)$ is finite iff $\F(s)=\Fnonstat(s)$.
\end{restatable}
\begin{proof}
Suppose $\Fnonstat(s)$ is finite but $\F(s)\neq\Fnonstat(s)$. Since $\F(s)\subseteq \Fnonstat(s)$, let $\fpi{}\in \Fnonstat(s)\setminus \F(s)$. $\pi$ must exhibit a non-stationarity which affects the induced visit distribution, which means that starting from $s$, following $\pi$ must induce a positive probability of visiting some state $s'$ twice. If $\abs{\childDists[s']}=1$, $\pi$ cannot affect the induced visit distribution with a non-stationarity at $s'$, so $\abs{\childDists[s']}>1$.

Since $s'$ can reach itself with positive probability, there exists some stationary policy $\pi'$ which has positive probability of visiting $s'$ $k$ times, for any natural number $k$. Let action $a$ be such that $T(s',\pi'(s'))\neq T(s',a)$ (such an action exists because $\abs{\childDists[s']}>1$). Let $\pi_k$ be a non-stationary policy which agrees with $\pi'$ for the first $k$ visits to $s'$, but which takes action $a$ after visit $k$.

Each $\pi_k$ induces a different state visit distribution function; since there are infinitely many of them, $\Fnonstat(s)$ cannot be finite if $\F(s)\neq\Fnonstat(s)$. Then if $\Fnonstat(s)$, then $\F(s)=\Fnonstat(s)$.

The reverse direction follows because $\abs{\F(s)}\leq \abs{\St}^{\abs{\A}}$, which is finite.
\end{proof}

\begin{restatable}[Characterization of when stationarity is relevant]{prop}{charStationary}
$\Fnonstat(s)=\F(s)$ iff for all states $s'$ reachable from $s$ with positive probability, if $s'$ can reach itself with positive probability, $\abs{\childDists[s']}=1$.
\end{restatable}
\begin{proof}
For the forward direction, apply the reasoning from the proof of \cref{prop:iff-Fnonstat-finite}.

Reverse direction: if non-stationarity is to affect the induced visit distribution function, there must be some $s'$ which can reach itself with positive probability, and which is reachable from $s$. But if $\abs{\childDists[s']}=1$ for all such $s'$, then non-stationarity does not affect the induced visitation distribution function (because all actions have the same local dynamics at $s'$).
\end{proof}

\subsection{Generalized non-domination results}

\begin{restatable}[Idempotence of non-domination]{lem}{idempotent}\label{lem:nd-idempotent}
Let $X,Y\subsetneq \genVS$ be finite.
\begin{equation*}
    \ND{\ND{X}\setminus Y}=\ND{X}\setminus Y.
\end{equation*}
\end{restatable}
\begin{proof}
Trivially, $\ND{\ND{X}\setminus Y}\subseteq \ND{X}\setminus Y$. Let $\x \in \ND{X}\setminus Y$. Then $\exists \rf \in \genVS: \x^\top \rf > \max_{\x'\in X\setminus \set{\x}} \x'^\top \rf \geq \max_{\x'\in X\setminus \prn{Y\cup \set{\x}}} \x'^\top \rf$, and so $\x \in \ND{\ND{X}\setminus Y}$ by the definition of non-domination (\cref{def:nd}).

Then $\ND{\ND{X}\setminus Y}\supseteq \ND{X}\setminus Y$, and so $\ND{\ND{X}\setminus Y}= \ND{X}\setminus Y$.
\end{proof}

\begin{restatable}[Non-dominated inclusion relation]{lem}{ndInclusion}
Let $A,B\subsetneq \rewardVS$ be finite. If $A\subseteq B$, then $\ND{B}\cap A\subseteq \ND{A}$.
\end{restatable}
\begin{proof}
Suppose $\av\in \ND{B}\cap A$.
\begin{align}
    \exists \rf \in \rewardVS: \av^\top \rf &> \max_{\bv\in B\setminus \set{\av}} \bv^\top \rf\label{eq:defn-nd-inclusion}\\
    &\geq \max_{\bv\in A\setminus \set{\av}} \bv^\top \rf.\label{eq:contain-nd}
\end{align}
\Cref{eq:defn-nd-inclusion} follows by the definition of ND (\cref{def:nd-lin-func}). \Cref{eq:contain-nd} follows because $A\subseteq B$. Therefore, $\exists \rf \in \rewardVS:  \av^\top \rf > \max_{\bv\in A\setminus \set{\av}} \bv^\top \rf$. Since $\av\in A$, $\av \in \ND{A}$.
\end{proof}

\begin{restatable}[Permutation commutes with non-dominance]{lem}{permCommute}
Let $X\subsetneq \rewardVS$ be finite and let $\phi\in\mdpPermGroup$. $\phi\cdot \ND{X}=\ND{\phi\cdot X}$.
\end{restatable}
\begin{proof}
$\x\in\ND{X}$ iff $\exists \rf \in\rewardVS$ for which \begin{align}
\x^\top\rf &> \max_{\x'\in X\setminus \set{\x}} \x'^\top \rf\\
\x^\top \permute\inv\permute\rf &> \max_{\x'\in X\setminus \set{\x}} \x'^\top \permute\inv\permute\rf\label{eq:perm-matrix-cancel}\\
\prn{\permute\x}^\top \permute\rf &> \max_{\x'\in X\setminus \set{\x}} \prn{\permute\x'}^\top \permute\rf\label{eq:perm-matrix-orth}\\
\prn{\permute\x}^\top \permute\rf &> \max_{\x_\phi'\in \phi\cdot X\setminus \set{\permute\x}} \x_\phi'^\top \permute\rf.\label{eq:permuted-nd}
\end{align}
\Cref{eq:perm-matrix-cancel} follows because the identity matrix $\mathbf{I}=\permute\inv\permute$, since permutation matrices are invertible. \Cref{eq:perm-matrix-orth} follows because permutation matrices are orthogonal, and so $\prn{\permute\inv}^\top=\permute$. By \cref{eq:permuted-nd}, $\phi\cdot \ND{X}=\ND{\phi\cdot X}$.
\end{proof}

\begin{restatable}[Number of non-dominated linear functionals]{lem}{numNDFunc}\label{lem:num-nd-lin-func}
Let $X\subsetneq \rewardVS$ be finite. $1<\abs{\ND{X}}$ iff $1<\abs{X}$.
\end{restatable}
\begin{proof}
Since $\ND{X}\subseteq X$, $1<\abs{\ND{X}}\leq \abs{X}$ implies that $1<\abs{\ND{X}}\implies 1<\abs{X}$.

Suppose $1<\abs{X}$. Since $X$ is non-empty, $\mathbf{0}$ the zero vector has some optimal $\x\in\ND{X}$ by \cref{lem:all-rf-max-nd}. Since $1<\abs{X}$, let $\x'\in X$ be such that $\x'\neq \x$. Since $\x\in\ND{X}$, let $\rf\in\rewardVS$ be such that $\x^\top\rf>\max_{\x''\in X\setminus\set{\x}}\x''^\top \rf\geq \x'^\top \rf$.

Then for $\rf'\defeq -\rf$, $\x'^\top\rf' > \x^\top\rf'$. By \cref{lem:all-rf-max-nd}, $\exists \x_\text{nd}\in\ND{X}$ such that $\x_\text{nd}^\top \rf'>\max_{\x''\in X\setminus\set{\x_\text{nd}}}\x''^\top \rf'$. But $\x_\text{nd}^\top \rf'\geq \x'^\top \rf'>\x^\top \rf'$, and so $\x_\text{nd}\neq \x$. Then $1<\abs{\ND{X}}$.
\end{proof}

\begin{restatable}[Sufficient conditions for a linear functional being non-dominated]{lem}{suffCond}\label{lem:ndLinSuff}
Let $X\subsetneq \rewardVS$ be finite and let $\x \in X$. If
\begin{enumerate}
    \item $\exists \unitvec[i]: \x^\top \unitvec[i]>\max_{\x'\in X\setminus\set{\x}} \x'^\top \unitvec[i]$, or
    \item $\exists \unitvec[i]: \x^\top \unitvec[i]<\min_{\x'\in X\setminus\set{\x}} \x'^\top \unitvec[i]$,
\end{enumerate}
then $\x \in \ND{X}$.
\end{restatable}
\begin{proof}
The first item follows directly from the definition of $\ND{X}$ (\cref{def:nd-lin-func}), where $\x$ is strictly maximal for vector $\rf\defeq \unitvec[i]$. For the second item, $\x$ is strictly maximal for vector $\rf \defeq -\unitvec[i]$.
\end{proof}

\begin{restatable}[Expanded sufficient conditions for a linear functional being non-dominated]{conjSec}{expSuffCond}
\Cref{lem:ndLinSuff} can be expanded to account for reasoning like ``$\x$ has non-strictly maximal entries along dimensions 1 and 2; among those $\x'\in X$ tied with $\x$ on both dimensions, $\x$ has strictly maximal entry value on dimension 3.''
\end{restatable}

\begin{restatable}[Non-dominated linear functionals are convex independent of other functionals]{prop}{convIndep}\label{prop:convex-independence}
Let $X\subsetneq \rewardVS$ be finite and let $\x \in X$. If $\x\in \ND{X}$, then $\x$ cannot be written as a convex combination of $\x_1,\ldots,\x_n\in X\setminus \set{\x}$.
\end{restatable}
\begin{proof}
Suppose $\x\in \ND{X}$ can be written as the convex combination $\x=\sum_{i=1}^n \theta_i \x_i$. By non-domination, $\exists \rf \in \rewardVS:\x^\top\rf>\max_{\x'\in X\setminus \set{\x}} \x'^\top\rf$. Therefore,
\begin{align}
    \prn{\sum_{i=1}^n \theta_i \x_i}^\top \rf&=\x^\top\rf\\
    &>\max_{\x'\in X\setminus \set{\x}} \x'^\top\rf\\
    &\geq \max_i \x_i^\top \rf.
\end{align}

But this is a contradiction; a convex combination of values cannot be strictly greater than all of its constituent values.
\end{proof}

The reverse direction of \cref{prop:convex-independence} is not true. Let $X\defeq \set{\colvec{2}{1}{0}, \colvec{2}{.5}{.5}, \colvec{2}{.4}{.6}}$. $\ND{X}=\set{\colvec{2}{1}{0}, \colvec{2}{.4}{.6}}$. $\x\defeq \colvec{2}{.5}{.5}$ is dominated, even though it cannot be written as a convex combination of functionals $X\setminus \set{\x}$.

\begin{restatable}[Characterizing when $X=\ND{X}$]{conjSec}{charEqLinFuncConj}
Let $X\subsetneq \rewardVS$ be finite. $X=\ND{X}$ iff $\abs{X}\leq1$ or $\forall \x,\x'\in X: \exists c>0: \x=c\x'$.
\end{restatable}

\begin{restatable}[ND argmax equality implies argmax equality]{conjSec}{argMaxEq}
Let $X\subsetneq \rewardVS$ be finite and let $\rf,\rf'\in\rewardVS$. If $\argmax_{\x\in \ND{X}} \x^\top \rf = \argmax_{\x\in \ND{X}} \x^\top \rf'$, then $\argmax_{\x\in X} \x^\top \rf = \argmax_{\x\in X} \x^\top \rf'$.
\end{restatable}

\begin{restatable}[Invariances of linear functional optimality probability]{lem}{invarianceLFOP}\label{lem:invariance-opt-prob}
Let $A,B\subsetneq \rewardVS$ be finite.
\begin{enumerate}
    \item $\phelper{A\geq B}[\Dany]=\phelper{A\geq \prn{B\setminus A}}[\Dany]$.\label{item:help-exclusion}
    \item $\phelper{A\geq B}[\Dany]=\phelper{A\geq\ND{B}}[\Dany]=\phelper{\ND{A}\geq \ND{B}}[\Dany]$.\label{item:help-nd}
    \item For any $\x \in\rewardVS$, $\phelper{A\geq B}[\Dany]=\phelper{A-\x\geq B-\x}[\Dany]$.\label{item:lin-fn-probability-invar}
\end{enumerate}
\end{restatable}
\begin{proof}
\Cref{item:help-exclusion}:
\begin{align}
    \phelper{A\geq B}[\Dany]&\defeq \optprob[\rf \sim \Dany]{\max_{\av\in A} \av^\top \rf \geq \max_{\bv\in B} \bv^\top \rf}\\
    &=\optprob[\rf \sim \Dany]{\max_{\av\in A} \av^\top \rf \geq \max\prn{\max_{\bv\in B\cap A} \bv^\top \rf,\max_{\bv\in B\setminus A} \bv^\top \rf}}\\
    &=\optprob[\rf \sim \Dany]{\max_{\av\in A} \av^\top \rf \geq \max_{\bv\in B\setminus A} \bv^\top \rf}\label{eq:max-vacuous}\\
    &\eqdef\phelper{A\geq \prn{B\setminus A}}[\Dany].
\end{align}
\Cref{eq:max-vacuous} holds because $\max_{\av\in A} \av^\top \rf\geq \max_{\bv\in B\cap A} \bv^\top \rf$ for all $\rf \in \rewardVS$, and so the constraint is vacuous.

\Cref{item:help-nd}:
\begin{align}
    \phelper{A\geq B}[\Dany]&\defeq \optprob[\rf \sim \Dany]{\max_{\av\in A} \av^\top \rf \geq \max_{\bv\in B} \bv^\top \rf}\\
    &=\optprob[\rf \sim \Dany]{\max_{\av\in A} \av^\top \rf \geq \max_{\bv\in \ND{B}} \bv^\top \rf}\label{eq:restrict-nd-helper}\\
    &=\optprob[\rf \sim \Dany]{\max_{\av\in \ND{A}} \av^\top \rf \geq \max_{\bv\in \ND{B}} \bv^\top \rf}\label{eq:restrict-nd-helper-2}\\
    &\eqdef\phelper{\ND{A}\geq\ND{B}}[\Dany].
\end{align}

\Cref{eq:restrict-nd-helper} follows because \cref{cor:nd-func-indif} shows that $\max_{\bv\in B} \bv^\top \rf=\max_{\bv\in \ND{B}} \bv^\top \rf$. Similarly, \cref{eq:restrict-nd-helper-2} follows because \cref{cor:nd-func-indif} shows that $\max_{\av\in A} \av^\top \rf=\max_{\av\in \ND{A}} \av^\top \rf$.

\Cref{item:lin-fn-probability-invar}: For all $\rf\in\rewardVS$, $\max_{\av\in A} \av^\top \rf \geq \max_{\bv\in B} \bv^\top \rf$ iff $\x^\top\rf+\max_{\av\in A} \av^\top \rf \geq \x^\top\rf+\max_{\bv\in B} \bv^\top \rf$ iff $\max_{\av\in A} (\av+\x)^\top \rf \geq \x^\top\rf+\max_{\bv\in B} (\bv+x)^\top \rf$.
\end{proof}

\begin{restatable}[Additivity of linear functional optimality probability for $\Dcont$]{prop}{helperAddProb}\label{prop:helper-cont-probability}
For finite $A,B\subsetneq \rewardVS$, $\phelper{A\geq B}[\Dcont]=\sum_{\av\in A} \phelper{\set{\av}\geq B}[\Dcont]$.\label{item:help-additive}
\end{restatable}
\begin{proof}
Let $A=\set{\av_1,\ldots, \av_n}$. For each $i$ between $1$ and $n$, define the event $X_i\defeq \{\rf \mid \av_i^\top \rf\geq \max_{\bv\in B} \bv^\top \rf\}$.
\begin{align}
    \phelper{A\geq B}[\Dcont]&\defeq \optprob[\rf \sim \Dcont]{\max_{\av\in A} \av^\top \rf \geq \max_{\bv\in B} \bv^\top \rf}\\
    &=\optprob[\Dcont]{\bigcup_{i=1}^n X_i}\\
    &=\sum_{k=1}^{n} (-1)^{k-1} \sum_{\substack{I \subseteq\{1, \ldots, n\}, \\|I|=k}} \optprob[\Dcont]{\bigcap_{i\in I} X_i}\label{eq:inclusion-exclusion}\\
    &= \sum_{i=1}^n\optprob[\Dcont]{X_i}\label{eq:vanish-intersect}\\
    &\eqdef \sum_{\av\in A} \phelper{\set{\av}\geq B}[\Dcont].
\end{align}
\Cref{eq:inclusion-exclusion} follows by the inclusion-exclusion formula. Almost all $\rf' \in \rewardVS$ are maximized by a unique functional $\av^*$ by \cref{cor:distinct-maximized}. By the same result, since $\Dcont$ is continuous, it assigns probability $0$ to $\rf'$ which are simultaneously maximized by at least two functionals. Therefore, all terms $k> 1$ in \cref{eq:inclusion-exclusion} must vanish, and \cref{eq:vanish-intersect} follows.
\end{proof}

\begin{restatable}[Positive probability under $\Dcont$ implies non-dominated functional]{lem}{ndPos}\label{lem:nd-pos}
Let $A\subsetneq \rewardVS$ be finite. Let $\av\in A$. If $\phelper{\set{\av} \geq A}[\Dcont]>0$, then $\av\in \ND{A}$.
\end{restatable}
\begin{proof}
Let $\av\in A$. If $\av\not \in\ND{A}$, then by \cref{lem:all-rf-max-nd},
\begin{equation}
    \forall \rf \in \rewardVS: \av^\top \rf = \max_{\av'\in A}\av'^\top \rf \implies \exists \av^*\in\ND{A}: \av^{*\top}\rf = \av^\top \rf.
\end{equation}

$\av \neq \av^*$ since $\av \not \in \ND{A}$. But \cref{cor:distinct-maximized} shows that continuous $\Dcont$ assign $0$ probability to the set of $\rf$ for which multiple $\av^*,\av$ are optimal. Then $\phelper{\set{\av},A}[\Dcont]=0$, a contradiction. So we conclude that $\av\in \ND{A}$.
\end{proof}

\subsection{Non-dominated visit distribution functions}
\begin{restatable}[$\abs{\Fnd(s)}\geq1$, with equality iff $\abs{\F(s)}=1$]{cor}{multiFnd}\label{cor:multi-fnd}
\end{restatable}
\begin{proof}
We show $\abs{\Fnd(s)}=1$ implies $\abs{\F(s)}=1$ by proving the contrapositive. Suppose $\abs{\F(s)}\geq 2$, let $\f,\f'\in\F(s)$, and let $\gamma\in(0,1)$ be such that $\f(\gamma)\neq \f'(\gamma)$. Then $1<\abs{\F(s,\gamma)}$. Apply \cref{lem:num-nd-lin-func} to conclude that $1<\abs{\ND{\F(s,\gamma)}}=\abs{\Fnd(s,\gamma)}\leq \abs{\Fnd(s)}$, with the first equality following by \cref{lem:nd-relation}, and the second following from the definition of $\Fnd(s,\gamma)$ (\cref{def:eval-f-discount}).

If $\abs{\F(s)}=1$, then the sole visit distribution function is trivially strictly optimal for all $\rf\in\rewardVS$ at all $\gamma\in(0,1)$. Thus $\abs{\Fnd(s)}=1$.
\end{proof}

\begin{restatable}[Each reward function has an optimal non-dominated visit distribution]{lem}{existsOpt}\label{lem:exists-opt}
Let $\rf \in \rewardVS$ and $\gamma \in [0,1)$. $\exists \f \in \argmax_{\f\in\F(s)} \f(\gamma)^\top \rf: \f \in \Fnd(s)$.
\end{restatable}
\begin{proof}
Apply \cref{lem:all-rf-max-nd} with $X \defeq \F(s,\gamma)$.
\end{proof}

\begin{restatable}[Strict visitation optimality is sufficient for non-domination]{cor}{sufficientNd}\label{sufficient-nd}
Let $\gamma\in[0,1)$ and let $s,s'\in\St$. At least one element of $\set{\argmax_{\f \in \F(s)} \f(\gamma)^\top \unitvec[s']}$ is non-dominated.
\end{restatable}
\begin{proof}
Apply \cref{lem:exists-opt} to the reward function $\unitvec[s']$.
\end{proof}

\begin{remark}
\Cref{sufficient-nd} implies that if $s'$ is reachable with positive probability from $s$, then there is at least one non-dominated visit distribution function which assigns $s'$ positive visitation frequency. In this sense, $\Fnd(s)$ ``covers'' the states reachable from $s$.
\end{remark}

\begin{restatable}[Surely reachable children]{definition}{DefSurelyChildren}\label{def:sure-children}
The \emph{surely reachable children} of $s$ are $\sureChildren\defeq \set{s' \mid \exists a: T\prn{s,a}=\unitvec[s']}$. Note that determinism implies that $\children=\sureChildren$.
\end{restatable}
\begin{restatable}[Minimum number of non-dominated visit distribution functions]{cor}{minNonDom}\label{min-nondom}
Suppose $\f_1,\ldots,\f_{k}\in\F(s)$ place strictly greater visitation frequency on some corresponding states $s_1,\ldots,s_k$ than do other visitation distributions. Then $\f_1,\ldots,\f_{k}\in\Fnd(s)$ and $\abs{\Fnd(s)}\geq k$. In particular, $\abs{\Fnd(s)}\geq \abs{\sureChildren}$.
\end{restatable}
\begin{proof}
$\f_1,\ldots,\f_{k}\in\Fnd(s)$ by \cref{sufficient-nd}. $\abs{\Fnd(s)}\geq \abs{\sureChildren}$ because each $s' \in \sureChildren$ must have at least one visitation frequency-maximizing visit distribution function $\f\in\Fnd(s)$.
\end{proof}

\begin{restatable}[When $\abs{\F(s)}\leq 2$, $\F(s)=\Fnd(s)$]{cor}{TwoVisitDist}
\end{restatable}
\begin{proof}
$\F(s)=\Fnd(s)$ trivially when $\abs{\F(s)}=1$. When $\abs{\F(s)}=2$, each visitation distribution must visit at least one state with strictly greater frequency than does the other visitation distribution; apply \cref{min-nondom}.
\end{proof}

\begin{restatable}[Initial-state non-domination implies non-dom.\ at visited states]{lem}{NDLater}\label{lem:nd-later}
If $\fpi{s} \in\Fnd(s)$ is strictly optimal for reward function $R$ at discount rate $\gamma$ and $\fpi{s}(\gamma)^\top\unitvec[s']>0$, then $\fpi{s'}\in\Fnd(s')$.
\end{restatable}
\begin{proof}
Because $\fpi{s}(\gamma)^\top\unitvec[s']>0$, $\pi$ can induce some trajectory prefix $(s, s_1, \ldots s_{n-1}, s')$ with positive probability. By \cref{def:visit},
\begin{equation}
\fpi{s}(\gamma)=\mathbf{e}_s+\gamma\E{s_1\sim T(s,\pi(s))}{\unitvec[s_1] + \ldots +\gamma \E{s'\sim T(s_{n-1}, \pi(s_{n-1}))}{\fpi{s'}(\gamma)}}.\label{eq:strict-opt-later}
\end{equation}

Since $\fpi{}$ is the strictly optimal visit distribution from state $s$ for reward function $R$ at discount rate $\gamma$, \cref{eq:strict-opt-later} shows that $\pi$ must in particular induce a strictly optimal visit distribution for $R$ at $\gamma$ starting from state $s'$. If not, another visit distribution would induce optimality starting from $s'$, contradicting the strict optimality of $\fpi{}$.
\end{proof}

\begin{restatable}[Domination at visited state implies domination at initial state]{cor}{cor:dom-init}\label{cor:dom-init}
If $\fpi{s'}$ is dominated at $s'$, then either $\fpi{s}$ is dominated at $s$ or $\fpi{s\top}\unitvec[s']=0$.
\end{restatable}
\begin{proof}
This statement is the contrapositive of \cref{lem:nd-later}.
\end{proof}

\begin{restatable}[Dominated child state distributions induce dominated visit distributions]{lem}{DomChild}\label{lem:dom-child-later}
Let action $a$ induce a dominated child state distribution $\dbf\defeq T(s_1,a)$ at state $s_1$, and let $s$ be the initial state. If a policy $\pi$ has $\pi(s_1)=a$ and $\fpi{}(\gamma)^\top \unitvec[s_1]\neq 0$, then $\pi$ induces a dominated visit distribution $\fpi{}\in\F(s)\setminus\Fnd(s)$.
\end{restatable}
\begin{proof}
By \cref{lem:dom-child}, $\fpi{s_1}\in\F(s_1)\setminus\Fnd(s_1)$. Since $\fpi{}(\gamma)^\top \unitvec[s_1]\neq 0$, $\fpi{}\in\F(s)\setminus \Fnd(s)$ by \cref{cor:dom-init}.
\end{proof}

The following extends \cref{lem:factorize-F} to account for non-domination.

\citet{kulkarni2016deep} learn to estimate $\mathrm{Succ}(s,a)\defeq \E{s'\sim T(s,a)}{\fpi{s'}(\gamma)}$ in order to infer state-space bottlenecks. We prove that bottlenecks ``factorize'' state visit distribution functions.

\begin{figure}[ht]
    \centering
    \includegraphics[]{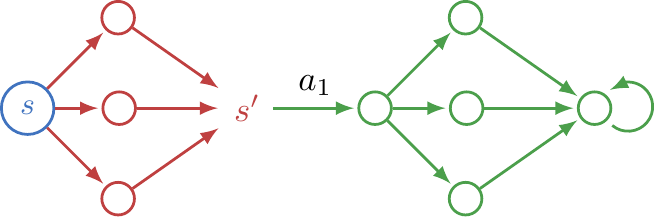}
    \caption[Visit distributions factorize across state-space bottlenecks]{Informally, \cref{lem:factorize-F} shows that a bottleneck at $\col{red}{s'}$ ``factorizes'' $\FRestrictAction[\col{blue}{s}]{\col{red}{s'}}{a_1}$ into combinations of ``what happens before acting at the bottleneck'' and ``what happens after acting at the bottleneck.'' In this rewardless {\mdp}, $\bottleneck[\col{blue}{s}]{\col{red}{s'}}{a_1}{\reach{\col{red}{s'},a_1}}$. Therefore, $\abs{\F(\col{blue}{s})}=9$: each of the three red ``prefix'' visit distribution functions (induced before reaching $\col{red}{s'}$) can combine with the three green ``suffix'' visit distribution functions (induced after reaching $\col{red}{s'}$).}
    \label{fig:factorize-F}
\end{figure}

\begin{restatable}[State-space bottleneck]{definition}{bottleneckDef}\label{def:bottleneck}
Starting from $s$, state $s'$ is a \emph{bottleneck} for $X\subseteq \St$ via action $a$ when state $s$ can reach the states of $X$ with positive probability, but only by taking actions equivalent to $a$ at state $s'$. We write this as $\bottleneck{s'}{a}{X}$.
\end{restatable}

\begin{restatable}[$\F(s)$ ``factorizes'' across bottlenecks]{lem}{factorizeF}\label{lem:factorize-F}
Suppose $\forall 1 \leq i \leq k: \bottleneck{s'}{a_i}{\reach{s',a_i}}$. Then let $\onereach \defeq \sum_{i=1}^k \sum_{s_j\in \reach{s',a_i}} \unitvec[s_j]$ and $\onerest\defeq \mathbf{1}-\onereach$ (where $\mathbf{1}\in\rewardVS$ is the all-ones vector). Let $\Fbrest\defeq \set{\fpi{s}\odot \onerest\mid \pi\in \Pi}$ (with $\odot$ the Hadamard product), $\FbrestA[a_i]\defeq \set{\fpi{s}\odot \onerest\mid \pi\in \Pi: \pi(s')\equiv_{s'} a_i}$, and $\Fbai\defeq \set{\E{s_{a_i}\sim T(s',a_i)}{\fpi{s_{a_i}}}\mid \pi\in \Pi}$. In the following, $\gamma$ is left variable on $[0,1)$.
\begin{equation}
    \FRestrictAction{s'}{a_i}= \set{ \frest(\gamma) + \prn{1-(1-\gamma)\lone{\frest(\gamma)}}\fai(\gamma) \mid \frest \in \FbrestA, \fai \in \Fbai}.
\end{equation}
\end{restatable}
\begin{proof}
Keep in mind that in order to detail how state-space bottlenecks affect the structure of visit distribution functions, we hold $\gamma$ variable on $[0,1)$.
\begin{align}
     &\FRestrictAction{s'}{a_i}\\
     &\defeq \set{\fpi{s} \mid \pi \in \Pi: \pi(s')=a_i}\\
     &= \set{\fpi{s} \mid \pi \in \Pi: \pi(s')\equiv_{s'}a_i}\label{eq:equiv-bottleneck}\\
     &= \set{\fpi{s}(\gamma)\odot\prn{\onerest+\onereach} \mid \pi \in \Pi: \pi(s')\equiv_{s'} a_i}\label{eq:oplus-one}\\
     &= \set{\fpi{s}(\gamma)\odot\onerest +\prn{1-(1-\gamma)\lone{\fpi{s}(\gamma)\odot\onerest}}\fai^\pi(\gamma) \mid \pi \in \Pi: \pi(s')\equiv_{s'}a_i}\label{eq:coeff-bottleneck}\\
     &= \set{\frest(\gamma) + \prn{1-(1-\gamma)\lone{\frest(\gamma)}}\fai(\gamma)\mid \frest \in \FbrestA[a_i], \fai \in \Fbai}\label{eq:decomp-ai}\\
     &= \set{\frest(\gamma) + \prn{1-(1-\gamma)\lone{\frest(\gamma)}}\fai(\gamma) \mid \frest \in \FbrestA, \fai \in \Fbai}.\label{eq:decomp-a1}
\end{align}

\Cref{eq:equiv-bottleneck} holds because by the definition of action equivalence (\cref{def:equiv-action}), equivalent actions induce identical state visit distribution functions $\fpi{s}$. \Cref{eq:oplus-one} follows since $\onerest+\onereach=\mathbf{1}$. To see that \cref{eq:coeff-bottleneck} follows, consider first that once the agent takes an action equivalent to $a_i$ at state $s'$, it induces state visit distribution $\fai^\pi(\gamma)\in \Fbai(\gamma)$ (by the definition of $\Fbai$). Since the bottleneck assumption ensures that no other components of $\fpi{s}$ visit the states of $\cup_{i=1}^k\reach{s',a_i}$, $\fpi{s}(\gamma)\odot\onereach=c(\gamma)\fai^\pi(\gamma)$ for some scaling function $c\in\reals^{[0,1)}$.
\begin{align}
    \lone{\fpi{s}(\gamma)}&=\geom\label{eq:total-l1}\\
    \lone{\fpi{s}(\gamma)\odot\onerest+c(\gamma)\fai^\pi(\gamma)}&=\geom\\
    \lone{\fpi{s}(\gamma)\odot\onerest}+   \lone{c(\gamma)\fai^\pi(\gamma)}&=\geom\label{eq:additive-l1}\\
    \abs{c(\gamma)}&=\lone{\fai^\pi(\gamma)}\inv \prn{\geom-\lone{\fpi{s}(\gamma)\odot\onerest}}\\
    \abs{c(\gamma)}&=(1-\gamma)\prn{\geom-\lone{\fpi{s}(\gamma)\odot\onerest}}\label{eq:total-l1-2}\\
    c(\gamma)&= 1-(1-\gamma)\lone{\fpi{s}(\gamma)\odot\onerest}.\label{eq:nonneg-c}
\end{align}
\Cref{eq:total-l1} and \cref{eq:total-l1-2} follow by \cref{prop:visit-dist-prop} \cref{item:lone-visit}. \Cref{eq:additive-l1} follows because $\frest(\gamma),\fai^\pi(\gamma)\succeq 0$. In \cref{eq:nonneg-c}, $\abs{c(\gamma)}=c(\gamma)$ must hold because if $c(\gamma)$ were negative, $\fpi{s}(\gamma)$ would contain negative entries (which is impossible by \cref{prop:visit-dist-prop} \cref{item:mono-increase}). \Cref{eq:nonneg-c} demonstrates that \cref{eq:coeff-bottleneck} holds.

To see that \cref{eq:decomp-ai} holds, consider that policy choices on $\reach{s',a_i}$ cannot affect the visit distribution function $\fpi{s}(\gamma)\odot\onerest$. This is because the definition of $\reach{s',a_i}$ ensures that once the agent reaches $\reach{s',a_i}$, it never leaves. Therefore, for all $\pi,\pi'\in\Pi$ which only disagree on $s_j\in \reach{s',a_i}$, $\fpi{s}(\gamma)\odot\onerest=\fpi[\pi']{s}(\gamma)\odot\onerest$. This implies that any $\fpi{}\in \Fbai$ is compatible with any $\fpi[\pi']{}\in\Fbrest$. Therefore, \cref{eq:decomp-ai} holds.

Lastly, we show that $\FbrestA[a_i]=\FbrestA$, which shows that \cref{eq:decomp-a1} holds. In the following, let $d(\gamma) \defeq 1-(1-\gamma)\lone{\fpi{s}(\gamma)\odot\onerest}$.
\begin{align}
    &\FbrestA[a_i] \\
    &\defeq \set{\fpi{s}\odot \onerest\mid \pi\in \Pi: \pi(s')\equiv_{s'} a_i}\\
    &= \set{\prn{\fpi{s}(\gamma)\odot\onerest+d(\gamma)\fai^\pi(\gamma)}\odot \onerest\mid \pi\in \Pi: \pi(s')\equiv_{s'} a_i}\label{eq:earlier-structure}\\
    &= \set{\prn{\fpi{s}(\gamma)\odot\onerest+d(\gamma)\f_{a_1}^\pi(\gamma)}\odot \onerest\mid \pi\in \Pi: \pi(s')\equiv_{s'} a_i}\label{eq:transform-rest-visit}\\
    &= \set{\prn{\fpi{s}(\gamma)\odot\onerest+d(\gamma)\f_{a_1}^\pi(\gamma)}\odot \onerest\mid \pi\in \Pi: \pi(s')\equiv_{s'} a_1}\label{eq:transform-rest-visit-2}\\
    &=\set{\fpi{s}\odot \onerest\mid \pi\in \Pi: \pi(s')\equiv_{s'} a_1}\label{eq:earlier-structure-2}\\
    &\eqdef \FbrestA.\label{eq:fbresta-final}
\end{align}

\Cref{eq:earlier-structure} and \cref{eq:earlier-structure-2} follow by \cref{eq:coeff-bottleneck} above. \Cref{eq:transform-rest-visit} follows because by the definition of $\onerest$ and of $\fai^\pi\in \Fbai, \f_{a_1}^\pi\in F^b_{a_1}$, we have $d(\gamma)\fai^\pi(\gamma)\odot \onerest = d(\gamma)\f_{a_1}^\pi(\gamma) \odot \onerest = \mathbf{0}$ (the all-zeros vector in $\rewardVS$). \Cref{eq:transform-rest-visit-2} follows because the definition of $\reach{s',a_i}$ ensures that once the agent takes actions equivalent to $a_i$ at $s'$, it only visits states $s_j \in \cup_{i'=1}^k\reach{s',a_i'}$. The same is true for $a_1$. Therefore, by the definition of $\onerest$, $\fpi{s}(\gamma)\odot\onerest$ is invariant to the choice of action $a_i$ versus $a_1$. We conclude that \cref{eq:decomp-a1} holds, which proves the desired equality.
\end{proof}

\begin{restatable}[Non-dominated single-state $\F$ restriction]{definition}{defNDRestrict}
$\FndRestrictAction{s'}{a}\defeq \FRestrictAction{s'}{a}\cap \Fnd(s)$.
\end{restatable}

\begin{restatable}[$\Fnd$ factorizes across state bottlenecks]{conjSec}{NDBottleneck}
In the following, $\gamma$ is left variable on $[0,1)$. Suppose that starting from $s$, state $s'$ is a bottleneck for $\reach{s',a_i}$ via actions $\set{a_i}$, for $1\leq i \leq k$. Then let $\onereach \defeq \sum_{i=1}^k \sum_{s_j\in \reach{s',a_i}} \unitvec[s_j]$ and $\onerest\defeq \mathbf{1}-\onereach$ (where $\mathbf{1}\in\rewardVS$ is the all-ones vector). Let $\Fbrest\defeq \set{\fpi{s}\odot \onerest\mid \pi\in \Pi}$ (with $\odot$ the Hadamard product), $\FbrestA[a_i]\defeq \set{\fpi{s}\odot \onerest\mid \pi\in \Pi: \pi(s')\equiv_{s'} a_i}$, and $\Fbai\defeq \set{\E{s_{a_i}\sim T(s',a_i)}{\fpi{s_{a_i}}}\mid \pi\in \Pi}$.
\begin{enumerate}[leftmargin=*]
    \item $\F(s)= \prn{\Fbrest\setminus \FbrestA} \cup \prn{\cup_{i=1}^k  \FRestrictAction{s'}{a_i}}$.\label{item:F-decomp}
    \item $\FndRestrictAction{s'}{a_i}\subseteq$\\\null\hfill$\set{\frest(\gamma)+\prn{1-(1-\gamma)\lone{\frest(\gamma)}}\fai(\gamma) \mid \fai \in \ND{\Fbai}, \frest \in \ND{\FbrestA}}$.\label{item:nd-restrict-decomp}
    \item $\FndRestrictAction{s'}{a_i}\subseteq$\\\null\hfill$\set{\frest(\gamma)+\prn{1-(1-\gamma)\lone{\frest(\gamma)}}\fai(\gamma) \mid \fai \in \ND{\Fbai}, \frest \in \ND{\FbrestA}}$.\label{item:nd-restrict-decomp-strong}
    \item $\FndRestrictAction{s'}{a_i} =$\\\null\hfill$\set{\frest(\gamma) +\prn{1-(1-\gamma)\lone{\frest(\gamma)}}\fai\mid \fai \in \ND{\Fbai}, \frest \in \ND{\FbrestA}}$.\label{eq:nd-restrict-decomp-strongest}
    \item $\Fnd(s)= \ND{\Fbrest\setminus \FbrestA}\cup$\\\null\hfill$\set{\frest(\gamma)+\prn{1-(1-\gamma)\lone{\frest(\gamma)}}\fai \mid \fai \in \bigcup_{i=1}^k\ND{\Fbai}, \frest \in \ND{\FbrestA}}.$\label{eq:Fnd-decomp}
\end{enumerate}
\end{restatable}
\begin{proof}[Partial proof sketch]
\textbf{\Cref{item:F-decomp}.}
\begin{align}
    \F(s) & \nonumber\\
    &=\bigcup_{a\in\A} \FndRestrictAction{s'}{a}\\
    &=\prn{\bigcup_{\substack{a\in\A:\\\forall i: a\not \equiv_{s'} a_i}}\FRestrictAction{s'}{a}}\cup \prn{\bigcup_{i=1}^k\FRestrictAction{s'}{a_i}}\\
    &=\prn{\bigcup_{\substack{a\in\A:\\\forall i: a\not \equiv_{s'} a_i}}\set{\fpi{s} \mid \pi \in \Pi: \pi(s')=a}}\cup \prn{\bigcup_{i=1}^k\FRestrictAction{s'}{a_i}}\\
    &=\prn{\bigcup_{\substack{a\in\A:\\\forall i: a\not \equiv_{s'} a_i}}\set{\fpi{s}\odot \onerest \mid \pi \in \Pi: \pi(s')=a}}\cup \prn{\bigcup_{i=1}^k\FRestrictAction{s'}{a_i}}\label{eq:one-rest-odot}\\
    &=\prn{\Fbrest\setminus \prn{\cup_{i=1}^k \FbrestA[a_i]}} \cup \prn{\bigcup_{i=1}^k\FRestrictAction{s'}{a_i}}\label{eq:defn-fbresta}\\
    &= \prn{\Fbrest \setminus \FbrestA} \cup \prn{\bigcup_{i=1}^k\FRestrictAction{s'}{a_i}}.\label{eq:identical-fbresta}
\end{align}

By the bottleneck assumption, $s$ can only reach the states of $\cup_{i=1}^k\reach{s',a_i}$ by taking actions equivalent to some $a_i$ at state $s'$. Therefore, \cref{eq:one-rest-odot} holds by the definition of $\onerest$. \Cref{eq:defn-fbresta} holds by the definition of $\Fbrest$ and $\FbrestA[a_i]$. \Cref{eq:identical-fbresta} because \cref{eq:fbresta-final} showed that $\forall i: \FbrestA[a_i]=\FbrestA$. We have now shown \cref{item:F-decomp}.

\textbf{\Cref{item:nd-restrict-decomp}.} Let $\f\in\FndRestrictAction{s'}{a_i}$ be strictly optimal for $\rf$ at discount rate $\gamma^*\in(0,1)$. By \cref{lem:factorize-F}, for some $\frest\in\FbrestA,\fai\in\Fbai$, $\f= \frest(\gamma)+\prn{1-(1-\gamma)\lone{\frest(\gamma)}}\fai(\gamma)$. Suppose that $\fai\not\in\ND{\Fbai}$. Then there exists another $\fai' \in \Fbai$ for which $\fai(\gamma^*)^\top\rf\geq\fai'(\gamma^*)^\top\rf$.

Consider $\f'\defeq\frest(\gamma)+\prn{1-(1-\gamma)\lone{\frest(\gamma)}}\fai'(\gamma)$. By \cref{item:F-decomp}, $\f' \in \F(s)$. However, since $\fai(\gamma^*)^\top\rf\geq\fai'(\gamma^*)^\top\rf$, $\f'(\gamma^*)^\top\rf\geq\f(\gamma^*)^\top \rf$. This contradicts the assumed strict optimality of $\f$. Therefore, $\fai\in\ND{\Fbai}$ (in particular, $\fai(\gamma^*)\in\ND{\Fbai(\gamma^*)}$).
\end{proof}

\begin{restatable}[Action-restricted visit distribution function similarity requires action similarity]{conjSec}{actionSimilar}
If $s$ can reach $s'$ with positive probability and $\FndRestrictAction{s'}{a'}$ is similar to a subset of $\FRestrictAction{s'}{a}$ via state permutation $\phi$, then $\permute T(s',a')=T(s',a)$.
\end{restatable}

\subsection{Properties of optimality support}

\begin{restatable}[Optimality support]{definition}{DefOptSupp}\label{def:opt-supp}
Let $A,B\subsetneq \rewardVS$ be finite.
\begin{equation}
    \opt[A\geq B]\defeq \set{\rf\in\rewardVS \mid \max_{\av \in A} \av^\top \rf \geq \max_{\bv \in B} \bv^\top \rf}.
\end{equation}

We sometimes abuse notation by replacing the set $A$ with a vector $\av$, as in: $\opt[\av\geq B]$.
\end{restatable}

$\opt$ represents the set of reward functions for which $\f(\gamma)$ is optimal at state $s$. $\opt$ can be calculated by solving the relevant system of $\abs{\F(s)}-1$ inequalities.\footnote{$\vavg$ can sometimes be computed analytically. The $\pwrNoDist$ and optimality probability in small deterministic {\mdp}s can be computed using Mathematica code at \href{https://github.com/loganriggs/Optimal-Policies-Tend-To-Seek-Power}{https://github.com/loganriggs/Optimal-Policies-Tend-To-Seek-Power}.}

 \begin{figure}[!ht]\centering
    \begin{tikzpicture}[mygraph]
        %nodes
            \node[blue] (1) {$1$};
            \node[red] (2) [right of=1] {$2$};
            \node[black] (3) [below right of=1] {$3$};
        %arrows
            \path[->] (1) edge node[midway] {\texttt{right}} (2);
            \path[->] (1) edge node[midway, left] {\texttt{down}} (3);
            \path[->] (2) edge [looseness=5, out=-90, in=0] (2);
            \path[->] (3) edge [looseness=5, out=-90, in=0] (3);
    \end{tikzpicture}
    \caption[Deriving which reward functions make a trajectory optimal]{A simple environment where it's easy to derive which reward functions make a trajectory optimal.}
    \label{steady_state_repeat}
\end{figure}
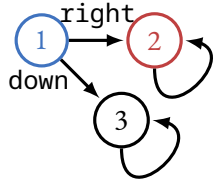

For example, consider \cref{steady_state_repeat}.
\begin{align*}
    \fpi[\pi_{\texttt{right}}\top]{} \rf &\geq \fpi[\pi_{\texttt{down}}\top]{}\rf\\
    R(\col{blue}{s_1}) + \geom[\gamma R(\col{red}{s_2})] &\geq R(\col{blue}{s_1}) + \geom[\gamma R(s_3)],
\end{align*}
so $R(\col{red}{s_2})\geq R(s_3)$.

\begin{restatable}[Topological boundary]{definition}{bdDef}
$\bd(X)$ is the topological boundary of set $X$, equal to $X$'s closure minus its interior ($\interior(X)$).
\end{restatable}

\begin{remark}
Unless otherwise stated, assume $\mathbb{R}^{\abs{\St}}$ is endowed with the standard topology.
\end{remark}

\begin{restatable}[A topological lemma]{lem}{convCombo}\label{lem:conv-combo}
Let $X\subseteq \mathbb{R}^{\abs{\St}}$ and suppose $S$ is such that $S\cap \interior(X)$ is convex and has no interior. If $s\in S\cap \interior(X)$, then $\exists \theta\in (0,1), x_1, x_2 \in \interior(X) \setminus S: s=\theta x_1 + (1-\theta) x_2$.
\end{restatable}
\begin{proof}
Since $S$ has no interior, its restriction to $\interior(X)$ must equal a convex subset of some $(\abs{\St}-1)$-dimensional hyperplane intersect $\interior(X)$. Let $\mathbf{x}$ be a unit-length vector orthogonal to this hyperplane. Since $R\in \interior(X)$, there exists $\epsilon>0$ small enough such that $R+\epsilon\mathbf{x},R-\epsilon\mathbf{x}\in \interior(X)$. Because $\mathbf{x}$ is perpendicular to the hyperplane, neither of these points belong to $S$. $R=.5(R+\epsilon\mathbf{x})+.5(R-\epsilon\mathbf{x})$.
\end{proof}

\begin{restatable}[Topological properties of optimality support]{lem}{multBoundary}\label{lem:mult-boundary}
Let $\f, \f' \in \F(s)$ and $\gamma \in (0,1)$.
\begin{enumerate}
    \item $\bd(\opt)\subseteq \opt$, with equality iff $\f$ is dominated.\label{1-mult-boundary}
    \item If $\f\neq\f'$ are both optimal for $R$ at discount rate $\gamma$, then $R\in \bd(\opt)\cap\bd(\opt[\f'(\gamma)\geq \F(s,\gamma)])$.\label{2-mult-boundary}
\end{enumerate}
\end{restatable}
\begin{proof}
\Cref{1-mult-boundary}: $\bd(\opt)\subseteq \opt$ because optimality support is closed by \cref{geom-prop}. If equality holds, $\opt$ has  no interior, and $\f$ is therefore dominated by \cref{nd-vanish}. Similarly, if $\f$ is dominated, then \cref{nd-vanish} dictates that $\opt$ has no interior and thus equals its boundary.

\Cref{2-mult-boundary}: by \cref{unique-optimal}, almost no reward functions have multiple optimal visit distributions, and so $\f$ and $\f'$ cannot both be optimal in an open neighborhood of $R$. Thus, $R$ must lie on at least one boundary: $R\in\bd(\opt)\cup\bd(\opt[\f'(\gamma)\geq \F(s,\gamma)])$.

Suppose $R\in\interior(\opt)$. Then $\f$ is optimal in an open neighborhood $N$ around $R$ and $R \in \bd(\opt[\f'(\gamma)\geq \F(s,\gamma)])$ by the above reasoning. By \cref{unique-optimal}, $\f$ must be uniquely (and therefore strictly) optimal for almost all reward functions in $N$. Furthermore, $\opt[\f'(\gamma)\geq \F(s,\gamma)]$ has no interior and is convex (by \cref{geom-prop}), so by \cref{lem:conv-combo}, $R\in\opt[\f'(\gamma)\geq \F(s,\gamma)]\cap \interior(\opt)$ can be written as the convex combination of reward functions for which $\f$ is optimal but $\f'$ is not. But  \cref{c-lem:optimal-preserve} shows that convex combination of reward functions preserves optimal policy sets, and so $\f'$ cannot be optimal for $R$, a contradiction. So $R \not \in\interior(\opt)$, and so $R\in\bd(\opt)\cap\bd(\opt[\f'(\gamma)\geq \F(s,\gamma)])$.
\end{proof}

\begin{restatable}[If a dominated visit distribution is optimal, so are at least two non-dominated visit distributions]{thm}{sandwich}\label{sandwich}
Suppose $\f_d\in\F(s)\setminus\Fnd(s)$ is optimal for reward function $R$ at discount rate $\gamma\in(0,1)$. Then there exist distinct $\f_1, \f_2\in\Fnd(s)$ which are also optimal for $R$ at $\gamma$.
\end{restatable}
\begin{proof}
If $\f_d$ were optimal for $R$ at $\gamma$, then some non-dominated $\f_1$ must also be optimal by \cref{cor:multi-fnd}. $R\in\bd(\opt[\f_d])\cap\bd(\opt[\f_1])$ by \cref{lem:mult-boundary}(\ref{2-mult-boundary}).

$\opt[\f_d]\cup\opt[\f_1]=\opt[\f_1] \subsetneq \mathbb{R}^{\abs{\St}}$ by \cref{min-nondom}. But \[\bigcup_{\f \in \F(s)} \opt = \rewardVS.\] Therefore, since all $\opt[\f]$ are closed and convex (\cref{geom-prop}), there must be at least one more $\f_2$ such that $\opt[\f_2]$ has non-empty interior (\ie{} $\f_2\in\Fnd(s)$ by \cref{nd-vanish}) and $\f_2$ is optimal for $R$. Then $R\in\bd(\opt[\f_d])\cap\bd(\opt[\f_1])\cap\bd(\opt[\f_2])$.
\end{proof}

\begin{restatable}[Geometry of dominated optimality support]{conjSec}{contain}
If $\f_d\in\F(s)\setminus\Fnd(s)$ and $\f\in\Fnd(s)$ are both optimal for $R$ at $\gamma\in(0,1)$, then  $\opt[\f_d]\subseteq \bd(\opt[\f])$.
\end{restatable}

\subsection{How geodesics affect visit distribution optimality}

\begin{restatable}[Geodesic trajectory]{definition}{geodesic}
In a directed graph, a path between two vertices is \emph{geodesic} when it is a shortest path. In deterministic environments, a state trajectory $(s_0, s_1, \ldots)$ is geodesic when, for all $i\leq j$, the trajectory traces a geodesic path between $s_i$ and $s_j$.
\end{restatable}

Intuitively, non-geodesic trajectories take ``detours'' (see \cref{fig:nd-not-geo}).

\begin{remark}
We refer to a policy or trajectory as ``dominated'' when the corresponding visit distribution function is dominated.
\end{remark}

\begin{restatable}[In deterministic {\mdp}s, geodesic trajectories induce non-dominated visit distributions]{prop}{geoNd}\label{thm:geo-nd}
Suppose the {\mdp} is deterministic and that starting from state $s_0$, policy $\pi$ induces geodesic trajectory $\tau^\pi = (s_0, s_1, \ldots)$. $\fpi{}\in\Fnd(s_0)$.
\end{restatable}
\begin{proof}
Since the {\mdp} is deterministic and the state space is finite, $\tau^\pi$ visits a finite number of states $s_0, \ldots, s_k$. Consider reward function $R$ such that $\forall i \leq k: R(s_i)\defeq \frac{i+1}{\abs{\St}}$; for all states $s'$ not visited by $\tau^\pi$, $R(s')\defeq 0$. Because $\tau^\pi$ is geodesic, $\fpi{}$ is strictly greedily optimal for $R$. Since optimal value changes continuously with $\gamma$, $\fpi{}$ is also strictly optimal for some $\gamma\approx 0$. Then $\fpi{}\in\Fnd(s_0)$.
\end{proof}

\begin{restatable}[In deterministic {\mdp}s, dominated trajectories are not geodesic]{cor}{domDetour}
\end{restatable}

Conversely, \cref{fig:nd-not-geo} shows that non-dominated visit distribution functions need not be geodesic.

\begin{figure}[ht]
    \centering
     \begin{tikzpicture}[mygraph]
        %nodes
            \node[blue] (0) {$0$};
            \node[black] (1) [below right of=0] {$1$};
            \node[black] (2) [right of=0] {$2$};
        %arrows
            \path[->] (0) edge (1);
            \path[->] (0) edge (2);
            \path[->] (1) edge (2);
            \path[->] (2) edge [looseness=5, out=45, in=135] (2);
    \end{tikzpicture}
     \caption[Non-dominated visit distribution functions need not be geodesic]{$\abs{\Fnd(\col{blue}{s_0})}=2$, even though one of the visit distribution functions is not geodesic because it induces state trajectory $(\col{blue}{s_0}, s_1, s_2, s_2, \ldots)$.}
    \label{fig:nd-not-geo}
\end{figure}

\begin{restatable}[Geodesics in stochastic environments]{conjSec}{geoStoch}
\Cref{thm:geo-nd} can be generalized in some form to stochastic {\mdp}s.
\end{restatable}

\subsection{Number of visit distribution functions}
Since $\F(s)$ only contains the visit distribution functions induced by \emph{deterministic stationary} policies, \cref{fig:not-more-F} shows that $\col{red}{s_3}$ being able to reach $\col{blue}{s_1}$ doesn't imply that $ \abs{\F(\col{red}{s_3})}\geq\abs{\F(\col{blue}{s_1})}$.

\begin{figure}[ht]
    \centering
    \begin{tikzpicture}[mygraph,shorten >=1.5pt,shorten <=1.5pt]
        %nodes
            \node[blue] (1) {1};
            \node[black, below left of=1] (2) {2};
            \node[red,below right of=1] (3) {3};
            \node[green, right of=3] (4)  {4};
        %arrows
            \path[<->] (1) edge (2);
            \path[<->] (1) edge (3);
            \path[<->] (2) edge (3);
            \path[->] (3) edge[shorten <= 0pt] (4);
            \path[->] (4) edge[shorten <= 0pt, looseness=5, out=0, in=90] (4);
    \end{tikzpicture}
    \caption[Examining the number of visit distribution functions at a state]{$\abs{\F(\col{blue}{s_1})}=8>7=\abs{\F(\col{red}{s_3})}$. In particular, considering only deterministic stationary policies, $\col{blue}{s_1}$ can reach $\col{green}{s_4}$ via two different trajectories: $(\col{blue}{s_1}, s_2, \col{red}{s_3}, \col{green}{s_4}, \col{green}{s_4},\ldots)$ and $(\col{blue}{s_1}, \col{red}{s_3}, \col{green}{s_4}, \col{green}{s_4},\ldots)$. However, $\col{red}{s_3}$ can only reach $\col{green}{s_4}$ in one way.}
    \label{fig:not-more-F}
\end{figure}

\begin{restatable}[Maximum number of visit distribution functions (deterministic)]{lem}{maxVisitDists}
Suppose the {\mdp} is deterministic. For all $s$, $\abs{\F(s)}\leq \sum_{j=1}^{\abs{\St}} j\frac{\prn{\abs{\St}-1}!}{\prn{\abs{\St}-j}!}$, with equality iff $\forall s': \children[s']=\St$.
\end{restatable}
\begin{proof}
First assume $\forall s': \children[s']=\St$. We count how many trajectories can be induced by deterministic stationary policies, starting from state $s$. Any trajectory visits at least 1 and at most $\abs{\St}$ states. Since order matters, a $j$-state trajectory can be chosen in $j!\binom{\abs{\St}}{j}=\frac{\abs{\St}!}{\prn{\abs{\St}-j}!}$ ways.  However, we must always begin the trajectory at $s$, so divide by $\abs{\St}$ possible starting states.

Since $\St$ is finite, the policy is stationary and deterministic, and the {\mdp} is deterministic, each trajectory must have a cycle by the pigeonhole principle. For any list of $j$ states, there are $j$ locations at which the cycle can begin. Then there are $\sum_{j=1}^{\abs{\St}} j\frac{\prn{\abs{\St}-1}!}{\prn{\abs{\St}-j}!}$ viable trajectories starting from state $s$. Two different trajectories induce different visitation distribution power series on $\gamma$ (\cref{def:visit}), and so different trajectories correspond to different visitation distribution functions.

Given $\forall s': \children[s']=\St$, we showed that $\abs{\F(s)}= \sum_{j=1}^{\abs{\St}} j\frac{\prn{\abs{\St}-1}!}{\prn{\abs{\St}-j}!}$. If $\exists s', s'': s'' \not \in \children[s']$, this rules out trajectories containing an $s'\to s''$ transition. Therefore, we have $\abs{\F(s)}\leq \sum_{j=1}^{\abs{\St}} j\frac{\prn{\abs{\St}-1}!}{\prn{\abs{\St}-j}!}$ in general, with equality iff $\forall s': \children[s']=\St$.
\end{proof}

\begin{remark}
When $\forall s': \children[s']=\St$, the non-dominated trajectories are the ones that immediately navigate to a state and stay there: $\forall s: \abs{\Fnd(s)}=\abs{\St}$. However, \cref{fig:more-nd-than-states} demonstrates an {\mdp} containing a state $s$ for which $\abs{\Fnd(s)}>\abs{\St}$.
\end{remark}

\begin{figure}[ht]
    \centering
    \begin{tikzpicture}[mygraph,shorten >=1.5pt,shorten <=1.5pt]
        %nodes
            \node[black] (1) {1};
            \node[black, below left of=1] (2) {2};
            \node[black,below right of=1] (3) {3};
        %arrows
            \path[<->] (1) edge (2);
            %\path[->] (2) edge (1);
            \path[<->] (1) edge (3);
            \path[<->] (2) edge (3);
    \end{tikzpicture}
    \caption[There can be more non-dominated visit distributions than states]{$\forall s: \abs{\Fnd(s)}=4>3=\abs{\St}$.}
    \label{fig:more-nd-than-states}
\end{figure}

\citet{puterman_markov_2014} notes that often, multiple policies map to the same visit distribution function. This is always true in deterministic environments if there is more than one possible policy.

\begin{restatable}[In deterministic environments, $\pi \mapsto \mathbf{f}^\pi_s$ is non-injective unless $\abs{\A}=1$]{thm}{nonInj}\label{thm:map-non-inj-iff}
Suppose the environment is deterministic. $\exists s: \abs{\F(s)}=\abs{\Pi}$ iff $\abs{\A}=1$.
\end{restatable}
\begin{proof}
Suppose $\abs{\A}>1$. If $\abs{\St}< \abs{\A}$, then by determinism, at least two distinct actions must be equivalent. By the definition of action equivalence (\cref{def:equiv-action}), policies taking equivalent actions at all states induce the same visit distribution functions, and so then $\abs{\F(s)}< \abs{\Pi}$.

Suppose $\abs{\St}\geq\abs{\A}$; then $\abs{\St}>1$. If a cycle can be induced before visiting all states, the premise is contradicted by modifying the $\pi$ in question in any unvisited states. So starting from $s$, let $\pi$ induce the visitation of all states. Since $2 \leq \abs{\A}, \abs{\St}$, $2 < 2\abs{\St}\leq \abs{\A} \abs{\St}$ states can be reached over the course of $\pi$'s state trajectory.

Therefore, by the time the trajectory has traversed $1\leq \lceil\frac{\abs{\St}}{2}\rceil<\abs{\St}$ states, there must exist an action returning to a state which has already been visited. Modify $\pi$ to take that action, and a cycle is formed before all states are visited. Since $2 \leq \abs{\A}$ and the induced visit distribution function is not affected by the action taken at the unvisited state, there are at least two policies which induce the same visit distribution function. So $\abs{\F(s)}< \abs{\Pi}$.

Suppose $\abs{\A}=1$. By the definition of $\F(s)$ (\cref{def:visit}), $\forall s: \abs{\F(s)}\leq \abs{\Pi}=\abs{\A}^{\abs{\St}}=1$. $\forall s: 1\leq\abs{\F(s)}$ by \cref{lem:exist-visit}.
\end{proof}

\Cref{thm:map-non-inj-iff} does not hold for stochastic environments.
\begin{restatable}[Sufficient condition for $\abs{\F(s)}=\abs{\Pi}$]{conjSec}{suffEqNumVisit}
If $\forall s: \abs{\childDists[s]}=\abs{\A}$ and $\forall s, s' \in \St, \dbf\in\childDists[s]: \dbf^{\top} \unitvec[s']>0$, $\forall s:\abs{\F(s)}=\abs{\A}^{\abs{\St}}=\abs{\St}$.
\end{restatable}

\begin{restatable}[$\abs{\F}$ bounds]{lem}{boundsF}\label{bounds-num-F}
\begin{equation*}
1 \leq \abs{\F(s)} \leq \prod_{s_i\in\reach{s}} \abs{\childDists[s_i]} \leq \prod_{i=1}^{\abs{\St}} \abs{\childDists[s_i]} \leq \abs{\A}^{\abs{\St}}=\abs{\Pi}.
\end{equation*}
\end{restatable}
\begin{proof}
$\forall s: 1\leq\abs{\F(s)}$ by \cref{lem:exist-visit}. By the definition of visit distribution functions (\cref{def:visit}), policy choices at unreachable states cannot affect the induced visit distribution function, so $\abs{\F(s)} \leq \prod_{s_i\in\reach{s}} \abs{\childDists[s_i]}$. $\abs{\childDists[s_i]}\leq \abs{\A}$ by the definition of child state distributions (\cref{def:child-dists}). $\abs{\A}^{\abs{\St}}=\abs{\Pi}$ because $\Pi$ is the set of deterministic stationary policies.
\end{proof}

By \cref{thm:map-non-inj-iff}, if $\abs{\A}>1$, then $\abs{\F(s)}<\abs{\Pi}$ and so at least one of the intermediate inequalities must be strict. \Cref{fig:strict-ineqs} demonstrates example cases.

\begin{figure}[ht]
     \centering
     \subfloat[][$\F(s) < \prod_{s_i\text{ reach. from } s} \abs{T(s_i)}$]{
     \begin{tikzpicture}[mygraph,shorten <=1.5pt]
        %nodes
            \node[black] (1) {};
            \node[black, below left of=1] (2) {};
            \node[black,below right of=1] (3) {};
        %arrows
            \path[<->] (1) edge (2);
            \path[<->] (1) edge (3);
            \path[<->] (2) edge (3);
    \end{tikzpicture}
    \label{ineq-1}}\qquad\subfloat[width=20cm][$\prod_{i=1}^{\abs{\St}} \abs{\childDists[s_i]}<\abs{\Pi}$]{\begin{tikzpicture}[mygraph]
        %nodes
            \node[blue] (1) {$s$};
            \node[black] (2) [right of=1] {};
            \node[black] (3) [above of=2] {};
            \node[black] (5) [right of=2] {};
        %arrows
            \path[->] (1) edge (2);
            \path[->] (2) edge (3);
            \path[->] (2) edge (5);
            \path[->] (3) edge [looseness=7] (3);
            \path[->] (5) edge [looseness=7] (5);
        \end{tikzpicture}
        \label{ineq-2}}
     \caption[Examining cardinality bounds on $\F(s)$]{Suppose $\abs{\A}=2$ in both cases. \protect\subref{ineq-1}: $\forall s: \abs{\F(s)} = 6 < 8=\prod_{s_i\in\reach{s}} \abs{T(s_i)} = \prod_{i=1}^{\abs{\St}} \abs{\childDists[s_i]} = \abs{\Pi}$. \protect\subref{ineq-2}: $\abs{\F(\col{blue}{s})} = 2=\prod_{s_i\in\reach{\col{blue}{s}}} \abs{T(s_i)} = \prod_{i=1}^{\abs{\St}} \abs{\childDists[s_i]}<\abs{\Pi}=16$. \label{fig:strict-ineqs}}
\end{figure}

\subsection{Variation distance of visit distributions in deterministic environments}

\begin{restatable}[When $\gamma=0$, the visit distributions of different states have $1$ total variation]{lem}{diffVisitMyopic}\label{gamma-0-dtv}
Let $\f \in \F(s), \f' \in \F(s')$. $\dtv{\f(0),\f'(0)}=\indic{s\neq s'}$.
\end{restatable}
\begin{proof}$\dtv{\f(0),\f'(0)}= \dtv{\unitvec[s], \unitvec[s']}= \indic{s\neq s'}$.
\end{proof}

\begin{figure}[h]
\centering
    \begin{tikzpicture}[mygraph,node distance=1.3cm]
        %nodes
            \node[green] (1) {};
            \node[left of=1, node distance=1.1cm] {$\colvec{4}{\col{green}{1}}{\col{green}{\gamma}}{\col{green}{\gamma^2}}{\geom[\gamma^3]}$};
            \node[green] (2) [right of=1] {};
            \node[green] (3) [right of=2] {};
            \node[black] (4) [right of=3] {};
            \node[right of=4] {$\colvec{4}{\col{green}{0}}{\col{green}{0}}{\col{green}{0}}{\geom}$};
        %arrows
            \path[->] (1) edge (2);
            \path[->] (2) edge (3);
            \path[->] (3) edge (4);
            \path[->] (4) edge [looseness=7,out=0,in=90] (4);
    \end{tikzpicture}

    \caption[How visit distribution functions change along a trajectory]{Visit distribution functions $\fpi{}$ induced at different states along a \col{green}{path}.}
    \label{fig:path}
\end{figure}

\begin{restatable}[Total variation along a graphical path]{lem}{varPath}\label{path-variation}
Let $\gamma \in (0,1)$. Suppose that $\pi$ travels a deterministic path from $s_1,\ldots, s_{\ell+1}$ and that $\pi$ will not visit $s_{\ell}$ again.
\begin{equation}
    \dtv{\fpi{s_1}(\gamma),\fpi{s_{\ell+1}}(\gamma)}= \geom[1-\gamma^{\ell}].
\end{equation}
\end{restatable}
\begin{proof}
Since the path is deterministic, $\pi$ never revisits any state in $s_1, \ldots, s_\ell$, since otherwise $\pi$ would visit $s_{\ell}$ again. By \cref{def:visit}, $\forall 1 \leq i < \ell + 1: \fpi{s_1}_{s_i}(\gamma) - \fpi{s_{\ell+1}}_{s_i}(\gamma)=\gamma^{i-1}$: each $s_i$ ``loses'' $\gamma^{i-1}$ visitation frequency. Since all visitation distributions $\f$ have $\lone{\f(\gamma)}=\geom$, the total visitation frequency thus ``lost'' equals the total visitation frequency gained by other states (and therefore the total variation; see \cref{fig:path}).  Then $\dtv{\fpi{s_1}(\gamma),\fpi{s_{\ell+1}}(\gamma)}=\sum_{i=0}^{\ell-1}\gamma^{i}=\geom[1-\gamma^{\ell}]$.
\end{proof}

\begin{restatable}[In stochastic environments, what general principles govern total variation among a policy's visit distributions?]{questionSec}{stochTotalVariation}
\end{restatable}

\begin{figure}[h]
\centering
     \begin{tikzpicture}[mygraph,node distance=2cm]
        %nodes
            \node (1) {\col{green}{$s_1$}};
            \node[left of=1, node distance=1.1cm] {$\colvec{8}{\col{green}{1}}{\col{green}{\gamma}}{\vdots}{\col{green}{\gamma^{j-2}}}{\gamma^{j-1}}{\gamma^j}{\vdots}{\gamma^{k-1}}$};
            \node (2) [below right of=1] {\col{green}{$s_2$}};
            \node (3) [right of=2] {\col{green}{$s_{j-1}$}};
            \node (4) [above right of=3] {$s_j$};
            \node[right of=4,node distance=1.25cm] {$\colvec{8}{\col{green}{\gamma^{k-j+1}}}{\col{green}{\gamma^{k-j+2}}}{\vdots}{\col{green}{\gamma^{k-1}}}{1}{\gamma}{\vdots}{\gamma^{k-j}}$};
            \node (5) [above left of=4] {$s_{j+1}$};
            \node (6) [left of=5] {$s_{k}$};
        %arrows
            \path[->] (1) edge [out=-90, in=180] (2);
            \path[dotted] (2) edge (3);
            \path[->] (3) edge [out=0, in=-90] (4);
            \path[->] (4) edge [out=90, in=0] (5);
            \path[dotted] (5) edge (6);
            \path[->] (6) edge [out=-180, in=90] (1);
    \end{tikzpicture}
    \caption[Variation distance for a visit distribution along its cycle]{Visitation distributions $\fpi{\col{green}{s_1}}(\gamma)$  and $\fpi{s_j}$. Total variation is maximized at diametrically opposite states in the cycle. Factors of $\frac{1}{1-\gamma^k}$ left out to avoid clutter.}
    \label{fig:cycle}
\end{figure}
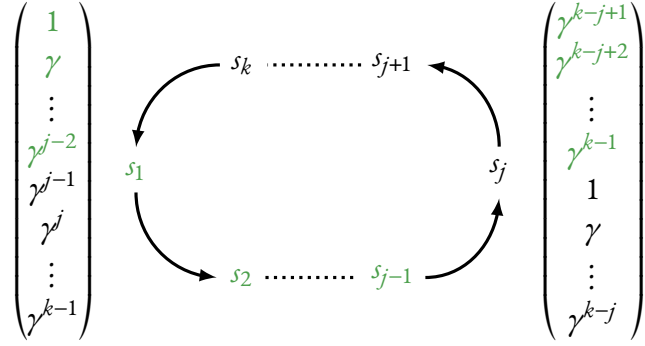

\Cref{fig:cycle} demonstrates how travelling along a deterministic cycle causes less total variation in the visit distributions.

\begin{restatable}[Total variation along a graphical cycle]{lem}{varCycle}\label{max-cycle}
Let $\gamma \in (0,1)$ and suppose that starting from state $s_1$, $\pi$ induces a deterministic $k$-cycle ($k>1$).
\begin{equation}
    \max_{j \in [k]}\dtv{\fpi{s_1}(\gamma),\fpi{s_j}(\gamma)}\leq\frac{1-\gamma^{\frac{k}{2}}}{(1-\gamma)(1+\gamma^{\frac{k}{2}})}<\geom[1-\gamma^{\frac{k}{2}}].
\end{equation}
\end{restatable}
\begin{proof}
\begin{align}
    \dtv{\fpi{s_1}(\gamma),\fpi{s_j}(\gamma)}&=\sum_{i=0}^{j-1}\gamma^i-\gamma^{k-i-1}\label{eq:verify}\\
    &=\geom[1-\gamma^j]\cdot\frac{1-\gamma^{k-j}}{1-\gamma^k}\label{eq:ceil-floor}\\
    &=\frac{1-\gamma^j+\gamma^k-\gamma^{k-j}}{(1-\gamma)(1-\gamma^k)}.
\end{align}
\Cref{eq:verify} can be verified by inspection of \cref{fig:cycle}. Setting the derivative with respect to $j$ to $0$, we solve
\begin{align}
    0&=-\gamma^j+\gamma^{k-j}\\
    j&=\frac{k}{2}.\label{eq:gamma-0-1}
\end{align}
\Cref{eq:gamma-0-1} follows because $\gamma \in (0,1)$. Solving via the derivative is justified because the function is strictly concave on $j\in[0,k]$ by the second-order test and the fact that $\gamma \in (0,1)$. If $k$ is even, we are done. If $k$ is odd, then we need an integer solution. Plugging $j=\lfloor \frac{k}{2} \rfloor$ and $\lceil \frac{k}{2}\rceil$ into \cref{eq:ceil-floor} yields the same maximal result.

Therefore, in the odd case, both inequalities in the theorem statement are strict. In the even case, the first inequality is an equality.
\end{proof}

\begin{restatable}[Lower bound for total variation of a policy's visit distributions in deterministic environments]{prop}{maxVariation}\label{lower-var}
Suppose the environment is deterministic. For any $\pi\in\Pi$, if $s\neq s'$, $\dtv{\fpi{s}(\gamma),\fpi{s'}(\gamma)}\geq\frac{1}{1+\gamma}\geq \half$.
\end{restatable}
\begin{proof}

If $\pi$ follows a path, it does so for at least $1$ state. Plugging in $\ell=1$ to \cref{path-variation} results in $\dtv{\fpi{s}(\gamma),\fpi{s'}(\gamma)}=1$. If $\pi$ follows a $k$-cycle ($k>1$), it does so for at least one step. Then $j=1,k=2$ for \cref{max-cycle}, in which case $\dtv{\fpi{s}(\gamma),\fpi{s'}(\gamma)}=\frac{1}{1+\gamma}<1$.
\end{proof}

\section{Optimal value function theory}
\begin{restatable}[Optimal value is piecewise rational on $\gamma$]{lem}{pwDiffOptVal}\label{pwDiffOptVal}
$\OptVf{s,\gamma}$ is piecewise rational on $\gamma$.
\end{restatable}
\begin{proof}
By \cref{lem:opt-pol-shift-bound}, the optimal visit distribution changes a finite number of times for $\gamma\in[0,1)$. \Cref{smoothOnPol} implies $\OptVf{s,\gamma}$ is rational on each non-degenerate subinterval where the optimal visit distribution set is constant.
\end{proof}

\begin{restatable}[$\OptVf{s,\gamma}$ is piecewise linear on $R$]{lem}{contVfn}\label{cont-vfn}
\end{restatable}
\begin{proof}
$V^*_R(s,\gamma)=\max_{\f\in\F(s)} \f(\gamma)^\top\rf$ takes the maximum over a finite set of fixed $\abs{\St}$\hyp{}dimensional linear functionals. Therefore, the maximum is piecewise linear with respect to $R$.
\end{proof}

\Cref{cont-vfn} shows that optimal value is piecewise linear in the reward function. In unpublished work, Jacob Stavrianos showed that optimal value is globally sublinear in the reward function:

\begin{restatable}[Optimal value is sublinear in the reward function]{lem}{sublinValue}\label{lem:sublin-value}
Let $R_1,R_2$ be reward functions and let $\gamma \in [0,1)$.
\begin{enumerate}
    \item Let $r \geq 0$. $\OptVf[rR_1]{s,\gamma}=r\OptVf[R_1]{s,\gamma}$.
    \item $\OptVf[R_1+R_2]{s,\gamma}\leq \OptVf[R_1]{s,\gamma}+\OptVf[R_2]{s,\gamma}$.
\end{enumerate}
\end{restatable}
\begin{proof}
$\OptVf[rR_1]{s,\gamma}=\max_{\f \in \F(s)}\f(\gamma)^\top (r\rf_1)=r\max_{\f \in \F(s)}\f(\gamma)^\top \rf_1=r\OptVf[R_1]{s,\gamma}$.

For the second condition, we check that
\begin{align}
    \OptVf[R_1+R_2]{s,\gamma} &= \max_{\f \in \F(s)}\f(\gamma)^\top (\rf_1+\rf_2)\\
    &\leq \max_{\f_1 \in \F(s)}\f(\gamma)^\top \rf_1+\max_{\f_2 \in \F(s)}\f(\gamma)^\top\rf_2\\
    &=\OptVf[R_1]{s,\gamma}+\OptVf[R_2]{s,\gamma}.
\end{align}
\end{proof}

\begin{restatable}[Optimal value is concave in the reward function]{cor}{concaveOpt}
\end{restatable}
\begin{proof}
Optimal value is sublinear in the reward function by \cref{lem:sublin-value}; sublinearity is a sufficient condition for concavity.
\end{proof}

\begin{restatable}[Optimal value is monotonically increasing in the reward function]{lem}{incOptVal}\label{lem:inc-opt-val}
Let $s\in\St$, $\gamma \in [0,1)$, and suppose $\forall s' \in \St: R_1(s') \geq R_2(s')$. Then $\OptVf[R_1]{s,\gamma}\geq \OptVf[R_2]{s,\gamma}$.
\end{restatable}
\begin{proof}
Let $\f_2 \in \argmax_{\f\in \F(s)}\f(\gamma)^\top \rf_2$.
\begin{align}
    \OptVf[R_1]{s,\gamma} &= \max_{\f \in \F(s)}\f(\gamma)^\top \rf_1\\
    &\geq \f_2(\gamma)^\top\rf_1\\
    &\geq \f_2(\gamma)^\top\rf_2\label{eq:dom-r2}\\
    &=\OptVf[R_2]{s,\gamma}.\label{eq:opt-r2}
\end{align}
\Cref{eq:dom-r2} follows by the assumed component-wise domination $\rf_1 \succeq \rf_2$. \Cref{eq:opt-r2} follows by the definition of $\f_2$.
\end{proof}

\begin{restatable}[Reward functions map injectively to optimal value functions]{thm}{injectVf}\label{injVf}
$\forall\gamma\in[0,1)$, $R\mapsto V^*_R(\cdot,\gamma)$ is injective.
\end{restatable}
\begin{proof}
Given $\OptVf{\cdot,\gamma}$ and the rewardless \textsc{mdp}, deduce an optimal policy $\pi^*$ for $R$ by choosing a $\OptVf{\cdot,\gamma}$-greedy action for each state.
\begin{align}
    \OptVf{\cdot,\gamma}&=R+\gamma \mathbf{T}^{\pi^*}V^*_R\\
    \prn{\mathbf{I}-\gamma \mathbf{T}^{\pi^*}}\OptVf{\cdot,\gamma}&=R.\label{eq:r-inj}
\end{align}

If two reward functions have the same optimal value function, then they have the same optimal policies. Then \cref{eq:r-inj} shows that the reward functions must be identical.
\end{proof}

Scott Emmons provided the proof sketch for \cref{injVf}.

\begin{restatable}[Linear independence of a policy's visit distributions]{lem}{liVisit}\label{li-policy-visit-dist}
At any fixed $\gamma\in[0,1)$, the elements of $\set{\fpi{s}(\gamma)\mid s\in\St}$ are linearly independent.
\end{restatable}
\begin{proof}
Consider the all-zero optimal value function with optimal policy $\pi^{*}$. \Cref{injVf} implies the following homogeneous system of equations has a unique solution for $\rf$:
\begin{align*}
    \fpi[\pi^{*}]{s_1}(\gamma)^\top\rf&=0\\
    &\vdots\\
    \fpi[\pi^{*}]{s_{\abs{\St}}}(\gamma)^\top\rf&=0.
\end{align*}

Therefore, $\pi^{*}$ induces linearly independent $\f$. But $\rf$ must be the all-zero reward function (for which all policies are optimal), so the $\fpi{s}$ are independent for \emph{any} policy $\pi$.
\end{proof}

\begin{restatable}[Two distinct visit distributions differ in expected optimal value for almost all reward functions]{lem}{uniqueOptValue}\label{unique-opt-value}
Let $\gamma \in (0,1)$, and let $\Delta,\Delta'\in\Delta(\St)$. If $\Delta\neq \Delta'$,
\begin{equation}
\optprob[R\sim \Dcont]{\E{ s\sim\Delta}{\OptVf{s,\gamma}}=\E{s'\sim \Delta'}{\OptVf{s',\gamma}}}=0.
\end{equation}
\end{restatable}
\begin{proof}
Let $R\in \text{supp}(\Dcont)$ and  $\pi^{*}\in\optPi$. By \cref{li-policy-visit-dist}, $\E{\Delta}{\fpi[\pi^{*}]{s}}= \E{\Delta'}{\fpi[\pi^{*}]{s'}}$ iff $\Delta=\Delta'$. Therefore, $\E{\Delta}{\fpi[\pi^{*}]{s}}\neq \E{\Delta'}{\fpi[\pi^{*}]{s'}}$. Trivially, $\E{ \Delta}{\OptVf{s,\gamma}}=\E{ \Delta'}{\OptVf{s',\gamma}}$ iff $\E{\Delta}{\fpi[\pi^{*}]{s}}^\top\rf= \E{\Delta'}{\fpi[\pi^{*}]{s'}}^\top\rf$. Since $\E{\Delta}{\fpi[\pi^{*}]{s}}\neq \E{\Delta'}{\fpi[\pi^{*}]{s'}}$, \cref{lem:distinct-lin-prob} implies that the equality holds with $0$ probability under $\Dcont$.
\end{proof}

No $\f$ is \emph{sub}optimal for all reward functions: every visit distribution is optimal for a constant reward function. However,  for any given $\gamma$, almost every reward function has a unique optimal visit distribution at each state.

\begin{restatable}[Optimal visit distributions are almost always unique]{lem}{UniqueOptimal}\label{unique-optimal}
Let $s$ be any state. For any $\gamma\in (0,1)$, $\set{\rf \text{ such that }   \abs{\argmax_{\f\in\F(s)} \f(\gamma)^\top \rf} >1}$ has measure zero under any continuous reward function distribution.
\end{restatable}
\begin{proof}
Let $R$ be a reward function and let $s$ be a state at which there is more than one optimal visit distribution for $R$ at discount rate $\gamma$. Since $R$ has more than one optimal visit distribution, there must exist a state $s'$ reachable with positive probability from $s$  such that actions $a,a'$ are both optimal at $s'$, where $a \not\equiv_{s'} a'$. Then $\E{s'' \sim T(s',a)}{\OptVf{s'',\gamma}}=\E{s'' \sim T(s',a')}{V^*_R(s'',\gamma)}$.

By \cref{unique-opt-value}, since $T(s',a)\neq T(s',a')$, this equation holds with probability $0$ for reward functions drawn from any continuous reward function distribution.
\end{proof}

\subsection{Discovering the true reward function}
\begin{restatable}[Value of reward information]{definition}{valRewardOInfoDefn}\label{def:voi}
Let $\Dbd$ have mean reward function $\ravg$. For state $s$ and $\gamma \in [0,1)$, $\rVOI[\Dbd]\defeq \vavg-\Vf[*][\ravg]{s,\gamma}$.
\end{restatable}

\begin{restatable}[In what situations is $\rVOI$ small?]{questionSec}{questrVOI}
\end{restatable}

\begin{restatable}[Value of reward information is non-negative]{prop}{valRewardInfo}
For state $s$ and $\gamma \in [0,1)$, $\rVOI[\Dbd]\geq 0$.
\end{restatable}
\begin{proof}
\begin{align}
    \Vf[*][\ravg]{s,\gamma}&\defeq \max_\pi V^\pi_{\ravg}(s,\gamma)\\
    &= \max_\pi \E{R\sim \Dbd}{V^\pi_{R}(s,\gamma)}\\
    &\leq \E{R\sim \Dbd}{\max_\pi V^\pi_{R}(s,\gamma)}\\
    &= \vavg.
\end{align}

The result follows since $\rVOI[\Dbd]\defeq \vavg-\Vf[*][\ravg]{s,\gamma}$ by \cref{def:voi}.
\end{proof}

\section{\textsc{Mdp} Structure}

Knowing the visit distribution functions for each state provides an enormous amount of information about the \mdp. As $\gamma\to 0$, the local dynamics are revealed (\cref{c-lem:prepend}). As $\gamma \to 1$, the renormalized visit distributions $(1-\gamma)\f(\gamma)$ limit to the recurrent state distributions which can be induced from state $s$: $\RSD$ (\cref{def:rsd}).

As it turns out, $\F$ encodes the \emph{entire} {\mdp} (\cref{cor:visit-encode}). First, we recap all of our visit distribution notation in \cref{tab:summary}.

\begin{table}[ht]
    \centering
    \begin{tabular}{rp{.585\textwidth}}
    \toprule
    Notation & Meaning\\
    \midrule
    Visit distribution function $\fpi{s}$  & \Cref{def:visit}: The discounted state visit distribution function induced by following policy $\pi$ starting from state $s$. A function from $\gamma\in[0,1)$ to $\reals^{\abs{\St}}$. $\pi$ and $s$ are often left implicit.  \rowskip
    Visit distribution $\fpi{s}(\gamma)$ &   \Cref{def:visit}: $\fpi{s}$ evaluated at discount rate $\gamma$. $\lone{\fpi{s}(\gamma)}=\geom$. $\pi$ and $s$ are often left implicit.\rowskip
    $\F(s)$ & \Cref{def:visit}: $ \set{\fpi{s} \mid \pi \in \Pi}$.\rowskip
    $\F(s,\gamma)$ & \Cref{def:eval-f-discount}: $\set{\f(\gamma) \mid \f \in \F(s)}$.\rowskip
    $\Fnd(s)$ & \Cref{def:nd}: The elements of $\F(s)$ which are strictly optimal for some reward function $\rf\in\reals^{\abs{\St}}$ and discount rate $\gamma \in (0,1)$.\rowskip
    $\Fnd(s,\gamma)$ & \Cref{def:eval-f-discount}: $\set{\f(\gamma) \mid \f \in \Fnd(s)}$.\rowskip
    $\FndRestrictAction{s'}{a}$ & \Cref{def:restrict-single}: The elements of $\Fnd(s)$ whose policies take action $a$ at state $s'$.\rowskip
    $\FndRestrict{S}$ & \Cref{def:restrict-multi}: The elements of $\Fnd(s)$ whose policies agree with $\pi^*$ on the states in $S \subseteq \St$.\rowskip
    $\Fnonstat(s)$ & \Cref{def:non-stat-visit}: $\set{\fpi{s} \mid \pi \in \Pi^\text{HD}}$.\\
    \bottomrule
    \end{tabular}
    \vspace{8pt}
    \caption[Summary of visit distribution notation]{Summary of visit distribution notation.}
    \label{tab:summary}
\end{table}

\begin{restatable}[Visitation function isomorphism]{definition}{mdpIso}\label{c-def:mdp-iso}
Let $M\defeq \langle \St, \A, T, \gamma \rangle$ and $M' \defeq \langle \St', \A', T', \gamma' \rangle$ be two rewardless {\mdp}s. $M \cong_{\F} M'$ (read ``$M$ and $M'$ have \emph{isomorphic visitation functions}'') when there exists a bijection $\phi: \St \to \St'$ (with corresponding permutation matrix $\permute$) satisfying $\forall s\in\St: \F_{M'}(\phi(s))=\phi\cdot \F_M(s)$.
\end{restatable}

This isomorphism is invariant to state representation, state labelling, action labelling, and the addition of superfluous actions ($a$ such that $\forall s: \exists a' \neq a: T(s,a)=T(s,a')$).

\begin{restatable}[Directed graph of a deterministic {\mdp}]{definition}{mdpModel}\label{def:mdp-model}
The \emph{directed graph} of a deterministic {\mdp} is a directed graph with a vertex for each state such that there is an arrow from vertex $s$ to vertex $s'$ iff $s'\in\children$.
\end{restatable}

Directed graphs are deterministic special cases of so-called {\mdp} \emph{models} \citep{puterman_markov_2014}. We introduce the following definition; when the dynamics are deterministic, it reduces to the standard directed graph isomorphism.

\begin{restatable}[Stochastic model isomorphism]{definition}{stochasticIso}\label{def:stoch-iso}
$\phi$ is a stochastic model isomorphism between rewardless {\mdp}s $M,M'$ when $\phi$ is a bijection $\phi:\St\to \St'$ is such that for all $s\in\St$, $\set{\permute T(s,a) \mid a \in \A}=\set{T'(\phi(s),a') \mid a' \in \A'}$. We then say that $M\cong_\phi M'$, which is read as read ``$M$ and $M'$ have \emph{isomorphic transitions}''.
\end{restatable}

\begin{restatable}[$\cong_{\F}$ is equivalent to transition isomorphism]{thm}{graphIso}\label{model-iso}
$M \cong_{\F} M'$ via bijection $\phi$ iff $M \cong_\phi M'$.
\end{restatable}
\begin{proof}Forward direction: let $s\in\St$.
\begin{align}
   \set{T'(\phi(s),a') \mid a' \in \A'}&= \set{\lim_{\gamma\to 0} \gamma^{-1}(\fpi[\pi']{\phi(s)}(\gamma)-\unitvec[\phi(s)])\mid \pi'\in\Pi'}\\
   &= \set{\permute\lim_{\gamma\to 0} \gamma^{-1}(\fpi{s}(\gamma)-\unitvec[s])\mid \pi\in\Pi}\label{eq:Fcong-local}\\
   &= \set{\permute T(s,a) \mid a \in \A}.
\end{align}

\Cref{eq:Fcong-local} follows because $M \cong_{\F} M'$. Then $M \cong_\phi M'$.

Suppose instead that $M \cong_\phi M'$. Let $\pi\in\Pi$ be a policy in $M$. Let $\pi'$ be such that for all $s\in\St$: $\pi'$ satisfies $T'(\phi(s),\pi'(\phi(s)))=\permute T(s,\pi(s))$; such actions exist because we assumed that $M \cong_\phi M'$. Then by repeated application of \cref{c-lem:prepend}, $\fpi[\pi']{\phi(s)}= \permute \fpi{s}$ for all $s\in\St$. Since $\pi$ was arbitrary, $M \cong_{\F} M'$ via $\phi$.
\end{proof}

\begin{restatable}[Visit distribution functions encode {\mdp}s]{cor}{visitEncodeStoch} \label{cor:visit-encode}
Given the function $\F$, the generating dynamics can be reconstructed up to transition isomorphism.
\end{restatable}
\begin{proof}
Given $\F$, for each state $s$, deduce $T(s)=\set{\lim_{\gamma\to 0} \gamma^{-1}(\fpi[]{s}(\gamma)-\unitvec[s])\mid \f\in\F(s)}$.
\end{proof}

In deterministic environments, the dynamics are encoded (up to transition isomorphism) by the visit distributions at a \emph{single} $\gamma\in(0,1)$.

\begin{restatable}[Visit distributions encode rewardless deterministic {\mdp}s]{thm}{visitEncode}\label{opt-encode}
Given the function $\F(\cdot,\gamma)$ generated by a deterministic rewardless {\mdp}, the generating dynamics can be reconstructed up to transition isomorphism.
\end{restatable}
\begin{proof}
Since $\forall s\in\St, \f(\gamma) \in \F(s,\gamma): \lone{\f(\gamma)} = \geom$ (by \cref{prop:visit-dist-prop}) and $\f(\gamma) \in \rewardVS$, we can deduce $\gamma$ and $\abs{\St}$. Let $\St'\defeq \set{1,\ldots, \abs{\St}}$. Using \cref{c-lem:prepend}, deduce the children $\children$ of each state $s$. Define $\A\defeq\set{1,\ldots, \max_s \abs{\children}}$. Construct a transition function $T'$ using $\children$; if $\abs{\A}>\abs{\children}$ for some state $s$, map the redundant actions  to any element of $\children$.

Let $M'\defeq \langle \St', \A', T',\gamma\rangle$. By construction, $M$ and $M'$ are transition isomorphic.
\end{proof}

\begin{restatable}[\Cref{opt-encode} holds in stochastic environments]{conjSec}{stochFSim}
\end{restatable}

\begin{restatable}[Optimal value functions encode rewardless deterministic {\mdp}s]{thm}{valEncode}\label{val-encode}
Given the optimal value function/reward function pairs of a rewardless deterministic {\mdp} $M$, $M$ can be reconstructed up to $\F$-isomorphism.
\end{restatable}
\begin{proof}
Suppose that for rewardless {\mdp} $M$, we are given $\set{(R,\OptVf{\cdot, \gamma}) \mid R \in
\boundedRewards}$ for fixed $\gamma$. Let $\St' \defeq \set{1, \ldots, \abs{\text{domain of } R}}$ for any reward function $R$.

For each $s$, we determine if it can reach itself. Let $R_s$ be the indicator reward function on state $s$. $s$ can reach itself iff $V^*_{R_s}(s,\gamma) > 1$. Because the {\mdp} is finite, at least one state $s$ must be able to reach itself.

If other states $s'$ also have $V^*_{R_s}(s',\gamma)>0$, consider $s_\text{pre}\in \argmax_{s'\in\St \setminus \set{s}} V^*_{R_s}(s',\gamma)$. $s_\text{pre}$ must be able to reach $s$ in one step, so $V^*_{R_s}(s_\text{pre},\gamma)=\gamma V^*_{R_s}(s,\gamma)$. Then $\gamma=\frac{V^*_{R_s}(s,\gamma)}{V^*_{R_s}(s_\text{pre},\gamma)}$.

If $s$ is the only state with positive optimal value for $R_s$, $V^*_{R_s}(s,\gamma)>1$ implies that $s$ must be able to reach itself. Then $V^*_{R_s}(s,\gamma)=\geom$; solve for $\gamma$.

The above reasoning explained how to test whether $s$ can reach itself and how to determine which other states can reach $s$. This information allows us to construct a transition function $T'$, setting the action space $\A'$ to be as large as necessary to accommodate the state with the most children. Because their directed graphs are isomorphic, $\langle \St', \A', \gamma, T' \rangle \cong_{\F} M$ by \cref{model-iso}.
\end{proof}

\begin{remark}
The proof of \cref{val-encode} shows that deterministic dynamics are fully determined by $\abs{\St}$ optimal value functions (one for each state indicator reward function).
\end{remark}

\begin{restatable}[Non-dominated visit distribution functions encode rewardless deterministic {\mdp}s]{cor}{ndVisitEncode}\label{nd-opt-encode}
Suppose the rewardless deterministic {\mdp} $M\defeq \langle \St, \A, T, \gamma \rangle$ induces $\Fnd$. From $\Fnd$, $M$ can be reconstructed up to $\F$-isomorphism.
\end{restatable}
\begin{proof}
Restriction to non-dominated distributions leaves optimal value unchanged for all reward functions. Apply \cref{val-encode} to recover $M$ up to $\F$-isomorphism.
\end{proof}

\Cref{fig:stoch-vf-indifference} shows that neither \cref{val-encode} nor \cref{nd-opt-encode} hold for stochastic environments. Given known transition dynamics, \cref{injVf} guarantees that $R$ recovered from $\OptVf{\cdot,\gamma}$, but in the stochastic variant of \cref{val-encode}, we would not know the transition dynamics \emph{a priori}.

\begin{figure}[h]
    \centering
    \begin{tikzpicture}[mygraph,node distance=1.3cm]
        %nodes
            \node[blue] (1) {1};
            \node [black, right of=1, node distance=.75cm, shorten >=0cm, shorten <=.0cm,inner sep=0pt,minimum size=.4pt] (split) {};
            \node[black, above right of=1] (a) {2};
            \node[black, below right of=1] (b) {3};
        %arrows
            \path[->] (1) edge (a);
            \path[->] (1) edge (b);
            \path[-] (1) edge node[right] {$\;\; a$} (split);
            \path[-] (split) edge (1);
            \path[->] (split) edge (a);
            \path[->] (split) edge (b);
            \path[->] (a) edge [looseness=7,out=45,in=-45] (a);
            \path[->] (b) edge [looseness=7,out=-45,in=45] (b);
    \end{tikzpicture}
    \caption[Some stochastic actions aren't strictly optimal for any reward function]{The bifurcated action $a$ is a stochastic transition, where $T(\col{blue}{s_1},a, \col{black}{s_2})=p,T(\col{blue}{s_1},a, \col{black}{s_3})=1-p$. For any $p\in(0,1)$, $a$ is a dominated action: $T(\col{blue}{s_1},a)\in \childDists[\col{blue}{s_1}]\setminus \childDistsNd[\col{blue}{s_1}]$. Since there is no optimal value function for which it is strictly optimal to take action $a$, no optimal value function is affected by the presence of $a$. This ambiguity does not arise in deterministic {\mdp}s, since $\childDists=\childDistsNd$ when the dynamics are deterministic.}
    \label{fig:stoch-vf-indifference}
\end{figure}

\begin{figure}[h]
    \centering
    \begin{tikzpicture}[mygraph,auto,node distance=3cm]
        % nodes
            \node (visit) {$\Fnd(\cdot,\gamma)$};
            \node[below left of=visit] (mdp) {$\langle \St, \A, \gamma, T \rangle$};
            \node[below right of=visit] (value) {$\set{\prn{R_s,\OptVf[R_s]{\cdot,\gamma}} \mid s \in \St}$};
        %arrows
            \path[<->] (mdp) edge (value);
            \path[<->] (mdp) edge (visit);
            \path[<->] (visit) edge (value);
    \end{tikzpicture}
    \caption[Representation of deterministic {\mdp}s]{In deterministic {\mdp}s, these three objects contain the same information (up to transition isomorphism).}
    \label{fig:diagram-structure}
\end{figure}
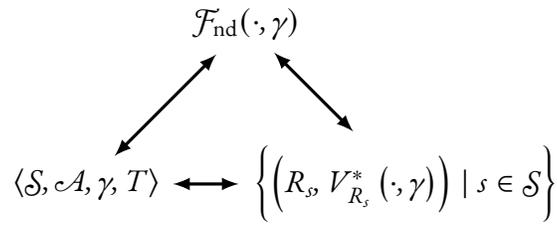

\Cref{fig:diagram-structure} summarizes this section's results.

\Cref{fig:need-rf-val-encode} shows that \cref{val-encode} cannot be proven without knowing which reward functions generate which optimal value functions (although $\gamma$ can still be deduced from the optimal value functions for all reward functions with reward bounded in $[0,1]$).

\begin{figure}[h]
     \centering
     \subfloat[$R(\col{blue}{s_1})=1$]{
     \begin{tikzpicture}[mygraph]
        %nodes
            \node[blue] (1) {$1$};
            \node[red] (2) [right of=1] {$2$};
            \node[black] (3) [below right of=1] {$3$};
        %arrows
            \path[->] (1) edge (2);
            \path[->] (2) edge [looseness=5, out=-90, in=0] (2);
            \path[->] (3) edge [looseness=5, out=-90, in=0] (3);
    \end{tikzpicture}\label{subfig:1-2}
     }\subfloat[$R(\col{blue}{s_1})=.8$]{
     \begin{tikzpicture}[mygraph]
        %nodes
            \node[blue] (1) {$1$};
            \node[red] (2) [right of=1] {$2$};
            \node[black] (3) [below right of=1] {$3$};
        %arrows
            \path[->] (1) edge (3);
            \path[->] (2) edge [looseness=5, out=-90, in=0] (2);
            \path[->] (3) edge [looseness=5, out=-90, in=0] (3);
    \end{tikzpicture}\label{subfig:1-3}
    }
    \caption[Stochastic {\mdp}s cannot be losslessly recovered from value function information]{Suppose $\gamma=.5$ and $\OptVf{\col{blue}{s_1},.5}=1.5,\OptVf{\col{red}{s_2},.5}=1,\OptVf{\col{black}{s_3},.5}=1.4$. This optimal value function is compatible with \protect\subref{subfig:1-2} (where $R(\col{blue}{s_1})=1$) and with \protect\subref{subfig:1-3} (where $R(\col{blue}{s_1})=.8$).} \label{fig:need-rf-val-encode}
\end{figure}

\begin{restatable}[In what category of {\mdp}s is $\cong_{\F}$ an isomorphism?]{questionSec}{catNatIso1}
\end{restatable}

\begin{restatable}[Is $\cong_{\F}$ natural in the category-theoretic sense?]{questionSec}{catNatIso2}
\end{restatable}
\begin{restatable}[What properties would $\cong_{\Fnd}$ have?]{questionSec}{fndIsoQuestion}
\end{restatable}

\begin{restatable}[Optimality probability and $\pwrNoDist$ change ``continuously'' with respect to transition dynamics]{conjSec}{envStructCont}
\end{restatable}

\section{Properties of optimal policy shifts}
\begin{restatable}[Optimal policy shift]{definition}{DefOptShift}\label{def:opt-shift}
$R$ has an \emph{optimal policy shift} at $\gamma\in (0,1)$ when $\lim_{\gamma^- \uparrow \gamma}\optPi[R,\gamma^-]\neq \optPi[R,\gamma]$. Similarly, $R$ has an \emph{optimal visit distribution shift} at $\gamma$ and at state $s$.
\end{restatable}

\begin{restatable}[One-sided limits exist for $\optPi$]{cor}{optPiOneSide}\label{cor:opt-pi-one-side}
Let $L\in (0,1)$ and let $R$ be any reward function. $\lim_{\gamma \uparrow L} \optPi$ and $\lim_{\gamma \downarrow L} \optPi$ both exist.
\end{restatable}
\begin{proof}
By \cref{lem:opt-pol-shift-bound}, $\optPi$ can take on at most finitely many values for $\gamma\in(0,1)$. Thus, infinite oscillation cannot occur in either one-sided limit, and so both one-sided limits exist.
\end{proof}

\Cref{cor:one-side-limit-equiv} shows that \cref{def:opt-shift} loses no generality by defining optimal policy shifts with respect to the limit from below.

\citet{lippman1968set} showed that two visit distribution functions can trade off optimality status at most $2\abs{\St}+1$ times. We slightly improve this upper bound. We thank Max Sharnoff for contributions to \cref{switch}.
\begin{restatable}[Upper bound on optimal visit distribution shifts]{lem}{switch}\label{switch}
For any reward function $R$ and $\f,\f'\in\F(s)$, $\prn{\f(\gamma)-\f'(\gamma)}^\top \rf$ is either the zero function, or it has at most $2\abs{\St}-1$ roots on $\gamma\in(0,1)$.
\end{restatable}
\begin{proof}
Consider two policies $\pi,\pi'$. By \cref{f-rat}, $(\fpi{}(\gamma)-\fpi[\pi']{}(\gamma))^\top \rf$ is a rational function with degree at most $2\abs{\St}$ by the sum rule for fractions. The fundamental theorem of algebra shows that $(\fpi{}(\gamma)-\fpi[\pi']{}(\gamma))^\top \rf$ is either $0$ for all $\gamma$ or for at most $2\abs{\St}$ values of $\gamma\in[0,1)$. Since $\f(0)=\f'(0)=\unitvec$ (\cref{def:visit}), one of the roots is at $\gamma=0$.
\end{proof}

\begin{restatable}[Lower-limit optimal policy set inequality iff upper-limit inequality]{cor}{oneSideIff}\label{cor:one-side-limit-equiv}
Let $\gamma\in (0,1)$, and $\Pi^- \defeq \lim_{\gamma^- \uparrow \gamma}\optPi[R,\gamma^-],\Pi^+ \defeq \lim_{\gamma^+ \downarrow \gamma}\optPi[R,\gamma^+]$. $\Pi^- \neq \optPi[R,\gamma]$ iff $\Pi^+\neq \optPi[R,\gamma]$.
\end{restatable}
\begin{proof}
Suppose $\Pi^- \neq \optPi[R,\gamma]$ but $\optPi[R,\gamma]=\Pi^+$. By \cref{pi-overlap}, if $\Pi^- \neq \optPi[R,\gamma]$, then $\Pi^- \subsetneq \optPi[R,\gamma]$. Let $\pi^*\in \optPi[R,\gamma]\setminus \Pi^-$ and $\pi^-\in \Pi^-$. Since $\pi^* \not \in \Pi^-$, there exists some $\epsilon_1 > 0$ such that $\pi^*$ isn't optimal for all $\gamma'\in (\gamma-\epsilon_1,\gamma]$. In particular, $(\fpi[\pi^*]{}(\gamma') - \fpi[\pi^-]{}(\gamma'))^\top \rf<0$ for such $\gamma'$. In particular, $(\fpi[\pi^*]{}(\gamma^*) - \fpi[\pi^-]{}(\gamma^*))^\top \rf$ is not the zero function on $\gamma^*$.

Therefore, \cref{switch} implies that $(\fpi[\pi^*]{}(\gamma^*) - \fpi[\pi^-]{}(\gamma^*))^\top \rf$ has finitely many roots on $\gamma^*$. But since $\pi^*\in \optPi[R,\gamma]=\Pi^+$, there exists $\epsilon_2>0$ such that $\forall \gamma' \in [\gamma,\gamma+\epsilon_2): (\fpi[\pi^*]{}(\gamma') - \fpi[\pi^-]{}(\gamma'))^\top \rf=0$. But this would imply that the expression has infinitely many roots, a contradiction. Therefore, if $\Pi^- \neq \optPi[R,\gamma]$, then $\optPi[R,\gamma]\neq \Pi^+$.

The proof of the reverse implication proceeds identically.
\end{proof}

\begin{figure}[ht]\centering
    \vspace{-8pt}
    \begin{tikzpicture}[mygraph,node distance=1.3cm]
        %nodes
            \node[blue] (1) {};
            \node [below of=1, node distance=.44cm] {\col{green}{$0$}};
            \node[black] (2) [right of=1] {};
            \node [below of=2, node distance=.46cm] {\col{green}{$-.25$}};
            \node[black] (3) [right of=2] {};
            \node [below of=3, node distance=.44cm] {\col{green}{$1$}};
            \node[black] (4) [right of=3] {};
            \node [below of=4, node distance=.45cm] {\col{green}{$-1$}};
            \node[black] (5) [right of=4] {};
            \node [below of=5, node distance=.44cm] {\col{green}{$0$}};
        %arrows
            \path[->] (1) edge (2);
            \path[->] (1) edge [out=45,in=135] (5);
            \path[->] (2) edge (3);
            \path[->] (3) edge (4);
            \path[->] (4) edge (5);
            \path[->] (5) edge [looseness=7,out=-45,in=45] (5);
    \end{tikzpicture}
    \caption[The optimal policy set can have point discontinuities on $\gamma$]{In \cref{pi-overlap}, $\Pi^-$ can equal $\Pi^+$. Let $R$ be the reward function whose rewards are shown in green. The shortcut is optimal for all $\gamma$. An optimal policy shift \protect\href{https://www.wolframalpha.com/input/?i=-.25x+\%2B+x\%5E2+-+x\%5E3}{occurs} at $\gamma=.5$. Since $\Pi^-=\Pi^+$ only contain policies which take the shortcut, $\Pi^-\cup\Pi^+\subsetneq \optPi[R,\gamma]$.\label{fig:subopt}}
\end{figure}

\begin{restatable}[Optimal policy sets overlap when shifts occur]{lem}{overlap}\label{pi-overlap}
Let $R$ be a reward function and $\gamma\in(0,1)$. Let $\Pi^- \defeq \lim_{\gamma^- \uparrow \gamma}\optPi[R,\gamma^-],\Pi^+ \defeq \lim_{\gamma^+ \downarrow \gamma}\optPi[R,\gamma^+]$. Then $\Pi^-\cup\Pi^+\subseteq \optPi[R,\gamma]$. Furthermore, if $R$ has an optimal policy shift at $\gamma$, $\exists s \in \St$:
\begin{equation*}
    \abs{\argmax_{\f\in\F(s)}\f(\gamma)^\top \rf}\geq 2.
\end{equation*}
\end{restatable}
\begin{proof}
Since an optimal policy shift occurs at $\gamma$ and since $\OptVf{s,\gamma}$ is continuous on $\gamma$ by \cref{pwDiffOptVal}, $\forall \pi^{-}\in \Pi^-,  \pi^{+}\in \Pi^+, s\in \St: V^{\pi^-}_R(s,\gamma) = V^{\pi^+}_R(s,\gamma)$. Therefore, $\Pi^-\cup\Pi^+\subseteq \optPi[R,\gamma]$.

By \cref{lem:opt-pol-visit-iff}, for any $R\in\rewardSpace$, an optimal policy shift occurs at $\gamma$ iff an optimal visit distribution shift occurs  at $\gamma$ for at least one state $s$.
\end{proof}

To better appreciate how optimal policy sets can be linked to the discount rate, consider the fact that some rewardless {\mdp}s have no optimal policy shifts. In other words, for any reward function and  for all $\gamma\in(0,1)$, greedy policies are optimal, as shown in \cref{fig:shift}. In deterministic environments, optimal policy shifts can occur if and only if the  agent can be made to choose between lesser immediate reward and greater delayed reward.

\begin{figure}[ht]
    \centering
    \vspace{-20pt}
    \subfloat[][]{
        \begin{tikzpicture}[mygraph,shorten >=-3pt]
            %nodes
                \node (1) {$\col{blue}{s_0}$};
                \node [below of=1, node distance=.4cm] {\col{green}{$1$}};
                \node (2) [above right of=1, node distance=1.3cm] {$s_1$};
                \node [below of=2, node distance=.4cm] {\col{green}{$.1$}};
                \node (3) [below right of=1] {$s_1'$};
                \node [below of=3, node distance=.4cm] {\col{green}{$0$}};
            %arrows
                \path[->] (1) edge (2);
                \path[<->] (1) edge (3);
                \path[->] (2) edge [looseness=5, out=-45, in=45] (2);
        \end{tikzpicture}
    \label{minimal-shift}}
    \subfloat[][]{
        \begin{tikzpicture}[mygraph,shorten > = -3pt,node distance=1.1cm]
            %nodes
                \node (1) {$\col{blue}{s_0}$};
                \node [below of=1, node distance=.4cm] {\col{green}{$0$}};
                \node (2) [above right of=1] {$s_1$};
                \node [below of=2, node distance=.4cm] {\col{green}{$.1$}};
                \node (3) [below right of=1] {$s_1'$};
                \node [below of=3, node distance=.4cm] {\col{green}{$0$}};
                \node (4) [right of=2] {$s_2$};
                \node [below of=4, node distance=.4cm] {\col{green}{$0$}};
                \node (5) [right of=3] {$s_2'$};
                \node [below of=5, node distance=.4cm] {\col{green}{$1$}};
            %arrows
                \path[->] (1) edge (2);
                \path[->] (1) edge (3);
                \path[->] (2) edge (4);
                \path[->] (3) edge  (5);
                \path[->] (4) edge [looseness=5, out=-45, in=45] (4);
                \path[->] (5) edge [looseness=4, out=-45, in=45] (5);
        \end{tikzpicture}
    \label{classic-shift}}\qquad
    \subfloat[][]{\begin{tikzpicture}[mygraph,node distance=1.2cm]
        %nodes
            \node[blue] (2) {};
            \node[black] (3) [above right of=2] {};
            \node[black] (4) [right of=2, node distance=1.07cm] {};
            \node[black] (5) [below right of=2] {};
        %arrows
            \path[->] (2) edge (3);
            \path[->] (2) edge (4);
            \path[->] (2) edge (5);
            \path[->] (3) edge [looseness=7, out=-45, in=45] (3);
            \path[->] (4) edge [looseness=7, out=-45, in=45] (4);
            \path[->] (5) edge [looseness=7, out=-45, in=45] (5);
        \end{tikzpicture}\label{simple-no-shift}}\qquad
     \subfloat[][]{\begin{tikzpicture}[mygraph,auto,node distance=1.2cm]
        %nodes
            \node[blue] (1) {};
            \node[black] (2) [above right of=1] {};
            \node[black] (3) [below right of=1] {};
            \node[black] (4) [right of=2] {};
            \node[black] (5) [right of=3] {};
        %arrows
            \path[->] (1) edge (2);
            \path[->] (1) edge (3);
            \path[->] (2) edge (4);
            \path[->] (2) edge (5);
            \path[->] (3) edge (4);
            \path[->] (3) edge (5);
            \path[->] (4) edge [looseness=7, out=-45, in=45] (4);
            \path[->] (5) edge [looseness=7, out=-45, in=45] (5);
        \end{tikzpicture}\label{complex-no-shift}}
    \caption[Illustrating when a deterministic {\mdp} allows optimal policy shifts]{\protect\subref{minimal-shift} and \protect\subref{classic-shift} show reward functions whose optimal policies shift. No shifts occur in \protect\subref{simple-no-shift} or \protect\subref{complex-no-shift}.\label{fig:shift}}
\end{figure}

\Cref{opt-policy-shift} suggests that the vast majority of deterministic rewardless {\mdp}s allow optimal policy shifts, as the criterion is easily fulfilled.

\begin{restatable}[Characterization of optimal policy shifts in deterministic rewardless {\mdp}s]{thm}{optShift}\label{opt-policy-shift}
In deterministic environments, there exists a reward function whose optimal action at $s_0$ changes  with $\gamma$  iff $\exists s_1\in Ch(s_0)$, $s_1'\in Ch(s_0)$, $s_2'\in Ch(s_1')\setminus Ch(s_1)$:
\begin{equation*}
    s_2' \not \in Ch(s_0) \lor \prn{s_1\not \in Ch(s_1) \land s_1' \not \in Ch(s_1)}.
\end{equation*}
\end{restatable}
\begin{proof}
\textbf{Forward direction.} Without loss of generality, suppose the optimal policy set of some $R$ is shifting for the first time (a finite number of shifts occur by \citet{blackwell_discrete_1962}).

Starting at state $s_0$, let the policies $\pi, \pi'$ induce state trajectories $s_0s_1s_2\ldots$ and $s_0s_1's_2'\ldots$, respectively, with the shift occurring to an optimal policy set containing $\pi'$ at discount rate $\gamma$. By the definition of an optimal policy shift at $\gamma$, $V^{\pi}_R(s_0,\gamma)=V^{\pi'}_R(s_0,\gamma)$. Because $\pi$ was greedily optimal and $\pi'$ was not, $s_1 \neq s_1'$ and $R(s_1)>R(s_1')$. If $\children[s_1]=\children[s_1']$, $\pi(s_0)$ remains the optimal action at $s_0$ and no shift occurs. Without loss of generality, suppose $s_2' \not \in \children[s_1]$.

We show the impossibility of $\lnot \prn{s_2' \not \in \children[s_0] \lor \prn{s_1\not \in \children[s_1] \land s_1' \not \in \children[s_1]}}=s_2'\in \children[s_0]\land \prn{s_1\in \children[s_1] \lor s_1' \in \children[s_1]}$, given that $\pi'$ becomes optimal at $\gamma$.

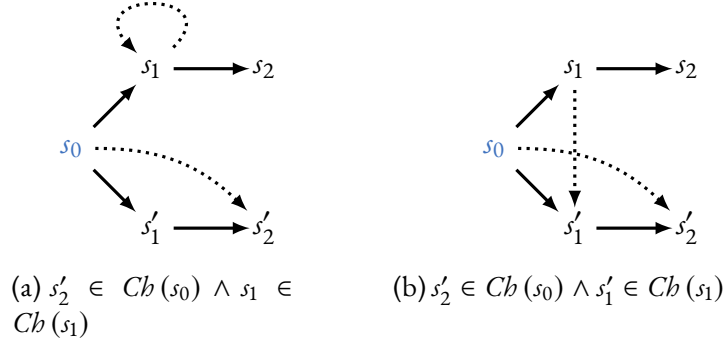
\begin{figure*}[ht]
    \centering
    \subfloat[width=8cm][$s_2' \in {\children[s_0]} \land s_1 \in {\children[s_1]}$]{
        \quad
        \begin{tikzpicture}[mygraph,shorten >=-3pt]
            %nodes
                \node (1) {$\col{blue}{s_0}$};
                \node (2) [above right of=1] {$s_1$};
                \node (3) [below right of=1] {$s_1'$};
                \node (4) [right of=2] {$s_2$};
                \node (5) [right of=3] {$s_2'$};
            %arrows
                \path[->] (1) edge (2);
                \path[->, dotted] (1) edge [out=0, in=135] (5);
                \path[->] (1) edge (3);
                \path[->] (2) edge (4);
                \path[->, dotted] (2) edge[looseness=5,out=45,in=135] (2);
                \path[->] (3) edge  (5);
        \end{tikzpicture}
        \quad
    \label{clause-a}}\qquad
    \subfloat[][$s_2'\in {\children[s_0]} \land s_1'\in {\children[s_1]}$]{
        \qquad
        \begin{tikzpicture}[mygraph,shorten >=-3pt]
            %nodes
                \node (1) {$\col{blue}{s_0}$};
                \node (2) [above right of=1] {$s_1$};
                \node (3) [below right of=1] {$s_1'$};
                \node (4) [right of=2] {$s_2$};
                \node (5) [right of=3] {$s_2'$};
            %arrows
                \path[->] (1) edge (2);
                \path[->] (1) edge (3);
                \path[->, dotted] (1) edge [out=0, in=135] (5);
                \path[->] (2) edge (4);
                \path[->, dotted] (2) edge (3);
                \path[->] (3) edge  (5);
        \end{tikzpicture}
        \qquad
    \label{clause-b}}
    \caption[Illustrating the conditions for the existence of optimal policy shifts]{Dotted arrows illustrate the assumptions for each case. Given that there exists a reward function $R$ whose optimal action at $\col{blue}{s_0}$ changes at $\gamma$, neither assumption can hold. Although not illustrated here, \eg{} $s_2=\col{blue}{s_0}$ or $s_2'=\col{blue}{s_0}$ is consistent with \cref{opt-policy-shift}. We leave the rest of the model blank as we make no further assumptions about its topology.}
    \label{fig:char-shift}
\end{figure*}

\textbf{Case: }$s_2'\in \children[s_0]\land s_1\in \children[s_1]$.
For $\pi'$ to be optimal, navigating to $s_1$ and staying there  cannot be a better policy than following $\pi'$ from $s_0$. Formally, $\frac{R(s_1)}{1-\gamma}\leq V^{\pi'}_R(s_1',\gamma)$ implies $R(s_1)\leq (1-\gamma)V^{\pi'}_R(s_1',\gamma)=(1-\gamma)\prn{R(s_1') + \gamma V^{\pi'}_R(s_2',\gamma)}$.

We now construct a policy $\pi'_2$ which strictly improves upon $\pi'$. Since $s_2'\in \children[s_0]$, $\exists a_2': T(s_0,a_2',s_2')=1$. Let $\pi'_2$ equal $\pi'$ except that $\pi'_2(s_0)\defeq a_2'$. Then since $R(s_1')<R(s_1)$, $V^{\pi'_2}_R(s_0,\gamma)>V^{\pi'}_R(s_0,\gamma)$, contradicting the assumed optimality of $\pi'$.

\textbf{Case: }$s_2'\in \children[s_0]\land s_1'\in \children[s_1]$. For $\pi'$ to be optimal, navigating to $s_1$, then to $s_1'$ (made possible by $s_1'\in \children[s_1]$), and then following $\pi'$ cannot be a better policy than following $\pi'$ from $s_0$. Formally, $R(s_1) + \gamma V^{\pi'}_R(s_1',\gamma) \leq V^{\pi'}_R(s_1',\gamma)$. This implies that $R(s_1)\leq (1-\gamma)V^{\pi'}_R(s_1',\gamma)=(1-\gamma)\prn{R(s'_1) + \gamma V^{\pi'}_R(s_2',\gamma)}$. The policy $\pi'_2$ constructed above is again a strict improvement over $\pi'$ at discount rate $\gamma$, contradicting the assumed optimality of $\pi'$.

\textbf{Backward direction.} Suppose  $\exists s_1, s_1'\in \children[s_0], s_2'\in \children[s_1']\setminus \children[s_1]: s_2' \not \in \children[s_0] \lor \prn{s_1\not \in \children[s_1] \land s_1' \not \in \children[s_1]}$. We show that there exists a reward function $R$ whose optimal policy at $s_0$ changes  with $\gamma$.

If $s'_2\not\in \children[s_0]$, then $s_2'\neq s_1$ because  $s_1\in \children[s_0]$. Let $R(s_1)\defeq .1$, $R(s_2')\defeq 1$, and $0$ elsewhere. Suppose that $s_1$ can reach $s_2'$ in two steps and then stay there indefinitely, while the state trajectory of $\pi'$ can only stay in $s_2'$ for one time step, after which no more reward accrues. Even under these impossibly conservative assumptions, an optimal trajectory shift occurs from $s_0s_1s_2\ldots$ to $s_0s_1's_2'\ldots$. At the latest, the shift occurs at $\gamma\approx 0.115$, which is  \href{https://www.wolframalpha.com/input/?i=.1\%2Bx\%5E2\%2F\%281-x\%29\%3Cx}{a solution} of the corresponding equality:
\begin{align}
    R(s_1) + \frac{\gamma^2}{1-\gamma}R(s_2') &= R(s_1') + \gamma R(s_2')\\
    .1 + \frac{\gamma^2}{1-\gamma} &= \gamma.
\end{align}

Alternatively, suppose $s_2=s_1$, and so $\pi$ continually accumulates $R(s_1)=.1$. Then there again exists an optimal policy shift corresponding to \href{https://www.wolframalpha.com/input/?i=.1\%2F\%281-x\%29\%3Dx}{a solution} to $\frac{.1}{1-\gamma} = \gamma$.

By construction, these two scenarios are the only ways in which $\pi$ might accrue reward, and so an optimal policy shift occurs for $R$.

If $s_2'\in \children[s_0]$, then set $R(s_1)\defeq 1$, $R(s_1')\defeq .99$,  $R(s_2')\defeq .9$, and $0$ elsewhere. Suppose that $s_1$ can reach itself in two steps (the soonest possible, as $s_1 \not \in \children[s_1]$), while neither $s_1'$ or $s_2'$ can reach themselves or $s_1$. The corresponding equation  $\frac{1}{1-\gamma^2}=.99+.9\gamma$ \href{https://www.wolframalpha.com/input/?i=1\%2F\%281-x\%5E2\%29\%3C.99\%2B.9x}{has a solution} in the open unit interval. Therefore, a shift occurs even under these maximally conservative assumptions.
\end{proof}

\begin{restatable}[If some reward function has optimal policy shifts, then almost all reward functions have optimal policy shifts]{conjSec}{almostAllShift}
\end{restatable}

\begin{restatable}[Sufficient conditions for a reward function not having optimal policy shifts]{prop}{noShiftIndicator}\label{c-prop:no-shift-indicator} Let $R \in \rewardSpace$.
\begin{enumerate}
    \item If $R$ assigns reward $r$ to all states, or \label{item:shift-const}
    \item If the environment is deterministic and $R$ is a state indicator reward function, or \label{item:shift-ind}
    \item If the environment is deterministic and $R$ is assigns reward $r_1$ to states which can reach themselves and $r_2$ to states which cannot reach themselves, or \label{item:shift-binary}
    \item If the environment is deterministic and $R$ is assigns reward $r_1$ to some set of 1-cycle states and $r_2$ to all other states, or \label{item:shift-set-onecyc}
    \item If $R=mR' + b$ for some $m>0,b\in\reals$ and $R' \in \rewardSpace$ which has no optimal policy shifts, \label{item:shift-transf}
\end{enumerate}
then $R$ has no optimal policy shifts.
\end{restatable}
\begin{proof}
\Cref{item:shift-const}: If $R$ assigns reward $r$ to all states, then all policies are optimal at all discount rates.

\Cref{item:shift-ind}: Let $s$ be a state and $R_s$ be the indicator reward function for $s$. At each state, let $\pi$ choose an action which minimizes graph distance to $s$. Such an action exists because there are only finitely many actions. Since $R(s)=1$ and $0$ elsewhere, $\pi$ is optimal for $R_s$ for all discount rates.

\Cref{item:shift-binary}: If $r_1 =r_2$, then apply \cref{item:shift-const}.

Suppose that $r_1> r_2$. Then starting from any state $s$, each policy $\pi$ induces a trajectory which deterministically visits $0 \leq \ell < \abs{\St}$ transient states before entering a $k$-cycle ($0<k\leq \St - \ell$). By assumption on $r_1,r_2$, we have $V^\pi_R(s,\gamma)=\geom[1-\gamma^{\ell}]\cdot r_2 + \gamma^\ell\geom[r_1]$. Then a policy induces optimal value at $s$ iff it minimizes $\ell$; such a policy must exist, since there are only finitely many admissible values of $\ell$.

This criterion is independent of the value of $\gamma \in (0,1)$, and so there are no optimal policy shifts at $s$. Since $s$ was arbitrary, there are no optimal policy shifts for $R$. Similar logic proves the $r_1 < r_2$ case, except that optimal policies maximize $\ell$.

\Cref{item:shift-set-onecyc}: If $r_1 =r_2$, then apply \cref{item:shift-const}.

Suppose that $r_1> r_2$ and let $s$ be a state. If $s$ cannot reach any $r_1$-states, then every action is optimal for $R$ at $s$ at any discount rate.

If $s$ can reach an $r_1$ state $s'$ via some policy $\pi$, then by determinism, $\pi$ induces a trajectory which deterministically visits $0 \leq \ell < \abs{\St}$ states before reaching $s'$. By the definition of $r_1,r_2$, we have $V^\pi_R(s,\gamma)\leq\geom[1-\gamma^{\ell}]\cdot r_2 + \gamma^\ell\geom[r_1]$, with equality iff $\pi$ stays at $s'$. Then since $r_1>r_2$, $\pi$ induces optimal value at $s$ iff it minimizes $\ell$ and stays at $s'$; such a policy must exist, since there are only finitely many admissible values of $\ell$.

This criterion is independent of the value of $\gamma \in (0,1)$, and so there are no optimal policy shifts at $s$. Since $s$ was arbitrary, there are no optimal policy shifts for $R$. Similar logic proves the $r_1 < r_2$ case, except that optimal policies maximize $\ell$ and avoid staying at $s'$.

\Cref{item:shift-transf}: Optimal policy is invariant to positive affine transformation, and so the result follows immediately.
\end{proof}

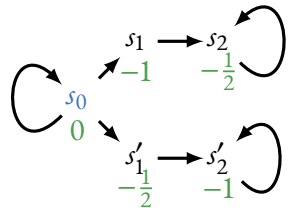
\begin{figure}[ht]
    \centering
    \begin{tikzpicture}[mygraph,shorten >=-3pt,node distance=1.1cm]
        %nodes
            \node (1) {$\col{blue}{s_0}$};
            \node [below of=1, node distance=.4cm] {\col{green}{$0$}};
            \node (2) [above right of=1] {$s_1$};
            \node [below of=2, node distance=.4cm] {\col{green}{$-1$}};
            \node (3) [below right of=1] {$s_1'$};
            \node [below of=3, node distance=.4cm] {\col{green}{$-\half$}};
            \node (4) [right of=2] {$s_2$};
            \node [below of=4, node distance=.4cm] {\col{green}{$-\half$}};
            \node (5) [right of=3] {$s_2'$};
            \node [below of=5, node distance=.4cm] {\col{green}{$-1$}};
        %arrows
            \path[->] (1) edge [looseness=5, out=-135, in=135, shorten <= -3pt] (1);
            \path[->] (1) edge (2);
            \path[->] (1) edge (3);
            \path[->] (2) edge (4);
            \path[->] (3) edge  (5);
            \path[->] (4) edge [looseness=5, out=-45, in=45] (4);
            \path[->] (5) edge [looseness=4, out=-45, in=45] (5);
    \end{tikzpicture}
    \caption[How optimal policy functions behave under reward function negation]{The given $R$ has no optimal policy shifts, but its inverse $-R$ does have optimal policy shifts.}
    \label{fig:inverse-no-shift-counterex}
\end{figure}

\Cref{fig:inverse-no-shift-counterex} shows that the presence of optimal policy shifts is not invariant to negation of the reward function.

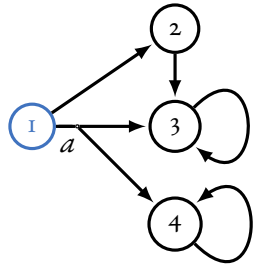
\begin{figure}[ht]
    \centering
    \begin{tikzpicture}[mygraph,node distance=1.3cm]
        %nodes
            \node[blue] (1) {1};
            \node [black, right of=1, shorten >=0cm, shorten <=.0cm, node distance=.6cm, inner sep=0pt,minimum size=.4pt] (split) {};
            \node[black, right of=split] (3) {3};
            \node[black, above of=3] (2) {2};
            \node[black, below of=3] (4) {4};
        %arrows
            \path[->] (1) edge (2);
            \path[->] (2) edge (3);
            \path[-] (1) edge node[below] {$\, a$} (split);
            \path[-] (split) edge (1);
            \path[->] (split) edge (3);
            \path[->] (split) edge (4);
            \path[->] (3) edge [looseness=7,out=45,in=-45] (3);
            \path[->] (4) edge [looseness=7,out=-45,in=45] (4);
    \end{tikzpicture}
    \caption[Optimal policy shifts can't happen for state indicator reward functions in deterministic environments]{Optimal policy shifts can occur for state indicator reward functions in stochastic environments, but they cannot occur in deterministic environments (\cref{c-prop:no-shift-indicator}). Let $R_{s_3}$ be the state indicator reward function for $s_3$, and let $T(\col{blue}{s_1},a,s_3)=T(\col{blue}{s_1},a,s_4)=\half$. $R_{s_3}$ has an optimal policy shift at $\gamma=\half$.}
    \label{fig:opt-pol-shifts-indicator}
\end{figure}

\Cref{fig:opt-pol-shifts-indicator} shows a counterexample to \cref{c-prop:no-shift-indicator}'s \cref{item:shift-ind} for stochastic environments.

Since we assume state-based reward functions, naively plugging in $\gamma=0$ would make all policies optimal. Instead, we consider the limiting optimal policy set as $\gamma\to 0$, in a similar manner as Blackwell optimality considers the limiting optimal policy set as $\gamma\to 1$.

\begin{restatable}[Asymptotically greedy optimality]{definition}{greedyStateAsympt}\label{def:greedy-state-asympt}
$\optPi[R,0]$ is the \emph{asymptotically greedily optimal policy set} for state-based reward function $R$.
\end{restatable}

Although, \cref{lem:greedy-reward} proved that $\optPi[R,0]\subseteq \greedy$, \cref{fig:dictionary-order} shows that equality need not hold.

\begin{figure}[h]
    \centering
    \begin{tikzpicture}[mygraph,shorten >=-3pt,node distance=1.1cm]
        %nodes
            \node (1) {$\col{blue}{s_0}$};
            \node (2) [above right of=1] {$s_1$};
            \node [below of=2, node distance=.4cm] {\col{green}{$1$}};
            \node (3) [below right of=1] {$s_1'$};
            \node [below of=3, node distance=.4cm] {\col{green}{$1$}};
            \node (4) [right of=2] {$s_2$};
            \node [below of=4, node distance=.4cm] {\col{green}{$1$}};
            \node (5) [right of=3] {$s_2'$};
            \node [below of=5, node distance=.4cm] {\col{green}{$0$}};
        %arrows
            \path[->] (1) edge node[midway] {\texttt{up}} (2);
            \path[->] (1) edge node[below left] {\texttt{down}} (3);
            \path[->] (2) edge (4);
            \path[->] (3) edge  (5);
            \path[->] (4) edge [looseness=5, out=-45, in=45] (4);
            \path[->] (5) edge [looseness=4, out=-45, in=45] (5);
        \end{tikzpicture}
    \caption[Asymptotic greediness versus next-step reward maximization]{For all $\gamma>0$, \texttt{up} has greater value than \texttt{down}. Therefore, even though both \texttt{up} and \texttt{down} are greedy actions, only $\pi^\texttt{up}\in\optPi[R,0]$.}
    \label{fig:dictionary-order}
\end{figure}

\begin{restatable}[Reward sequence induced by a policy]{definition}{rewardSeq}
Let $R$ be a reward function, and let the sequence $\prn{\prn{(\mathbf{T}^\pi)^t\unitvec}^\top \rf}^{s,\pi,R}_{t\geq 1}$ contain the expected undiscounted $R$-reward at time steps $t$ given that policy $\pi$ is followed starting from state $s$.
\end{restatable}

\begin{restatable}[Dictionary-ordered greediness]{prop}{dictOrder}\label{dict-order}
$\pi^* \in \optPi[R,0]$ iff $\forall s\in\St, \pi \in \Pi: (r_1, r_2, \ldots)^{s,\pi^*,R}\succeq (r_1, r_2, \ldots)^{s,\pi,R}$, where $\succeq$ is the dictionary ordering over $\reals^\infty$.
\end{restatable}
\begin{proof}
Repeatedly apply \cref{lem:greedy-reward} to conclude that $\pi^*\in \optPi[R,0]$ iff it satisfies the dictionary ordering condition at all time steps.
\end{proof}

\begin{restatable}[For almost all reward functions, asymptotically greedy action is determined by expected immediate reward]{lem}{almostGreedy}\label{lem:almost-greedy}
For almost all reward functions $R$, $\pi \in \optPi[R,0]$ iff $\pi \in \greedy$.
\end{restatable}
\begin{proof}
The forward direction holds for all reward functions as a corollary of \cref{dict-order}. The converse holds for almost all reward functions $R$, since  $\{\rf \in \rewardVS \mid \exists s,a: T(s,\pi(s))\neq T(s,a) \land T(s,\pi(s))^\top \rf = T(s,a)^\top \rf\}$ has no interior in the standard topology on $\rewardVS$ by \cref{lem:distinct-lin-prob}. Therefore, for almost all $R$, the policy which maximizes expected next-state return $\pi$ is unique up to action equivalence. By this uniqueness and by the fact that $\optPi[R,0]$ cannot be empty by \cref{lem:opt-pol-shift-bound}, $\pi \in \optPi[R,0]$ for almost all $R$.
\end{proof}

\begin{restatable}[Child distribution similarity implies equal greedy optimality probability]{prop}{childSimGreedy}\label{prop:child-sim-greedy}
If $T(s,a),T(s,a') \in \childDists$ are similar via a permutation $\phi$ such that $\phi\cdot \childDistsNd=\childDistsNd$, then  $\optprob[\Diid]{s,a,0}=\optprob[\Diid]{s,a',0}$.
\end{restatable}
\begin{proof}
\begin{align}
    &\optprob[\Diid]{s,a,0}\\
    \defeq &\prob[R \sim \D]{\exists \pi^* \in \optPi[R,0]: \pi^*(s)=a}\label{eq:almost-greedy-1}\\
    =&\E{\rf \sim \D}{\indic{T(s,a)^\top \rf \geq \max_{\dbf\in\childDists} \dbf^{\top}\rf}}\label{eq:greedy-reexpression}\\
    =&\E{\rf \sim \D}{\indic{T(s,a)^\top \rf \geq \max_{\dbf\in\childDistsNd} \dbf^{\top}\rf}}\label{eq:greedy-reexpression-2}\\
    =& \E{\rf \sim \D}{\indic{T(s,a')^\top \rf' \geq \max_{\dbf\in \childDistsNd} \dbf^\top \rf'}}\label{eq:similar-greedy}\\
    =& \prob[R \sim \D]{\exists \pi^* \in \optPi[R,0]: \pi^*(s)=a'}\label{eq:almost-greedy-2}\\
    \eqdef &\optprob[\Diid]{s,a',0}.
\end{align}

\Cref{eq:almost-greedy-1} and \cref{eq:almost-greedy-2} follow from \cref{lem:almost-greedy}. \Cref{eq:greedy-reexpression} follows by \cref{cor:opt-vf-restrict-fnd}. Let
\begin{align*}
    g(d_1,d_2) & \defeq \indic{d_1\geq d_2}\\
    \intertext{and}
    f(B_1,B_2\mid \D) & \defeq \E{\rf\sim \D}{g(\max_{\dbf\in B_1} \dbf^\top \rf,\max_{\dbf'\in B_2} \dbf'^\top \rf)}.
\end{align*}
Then by \cref{lem:helper-perm}, $f\prn{\set{T(s,a)},\childDistsNd\mid \D}=f\prn{\phi\cdot \set{T(s,a)},\phi\cdot \childDistsNd \mid \phi\cdot \D}=f\prn{\set{T(s,a')},\childDistsNd\mid \D}$ (as $\D$ distributes reward identically across states, so $\phi\cdot \D=\D$). Then \cref{eq:similar-greedy} holds.
\end{proof}

\begin{restatable}[Equal action optimality probability when $\gamma=0$]{cor}{noRobustInstr}
If the environment is deterministic and $\gamma=0$, then all actions are equally probably optimal at any given state.
\end{restatable}
\begin{proof}
In deterministic {\mdp}s, $\forall s:\childDistsNd=\childDists$ because each child state $s' \in \children$ is strictly greedily optimal for the indicator reward function $\indic{s''=s'}$. Furthermore, for any two actions $a_1, a_2$ leading to children $s_1,s_2\in\children$ respectively, the transposition of $s_1,s_2$ satisfies the requirements of \cref{prop:child-sim-greedy}.

Apply \cref{prop:child-sim-greedy} to conclude $a_1,a_2$ have equal optimality probability at $s$ when $\gamma=0$.
\end{proof}

\begin{restatable}[No-shift, injective reward functions can be solved greedily]{prop}{noShiftInjGreedy}
Let $R$ be a reward function which has no optimal policy shifts such that $\forall s \in \St, a,a' \in \A$, $a$ and $a'$ are both greedily optimal iff $a\equiv_s a'$. Then $\forall \gamma \in [0,1]: \optPi=\greedy$.

In deterministic environments, this holds for no-shift reward functions which assign a unique reward to each state.
\end{restatable}
\begin{proof}
By \cref{lem:greedy-reward}, $\pi \in \optPi[R,0]$ must be greedily optimal. By the next-step reward assumption, each greedily optimal policy is determined up to action equivalence; therefore, if $\pi \in \optPi[R,0]$ and $\pi'$ is greedily optimal for $R$, then $\forall s \in \St: \pi(s)\equiv_s \pi'(s)$. This implies that $\pi' \in \optPi[R,0]$. So $\optPi[R,0]=\greedy$. Since no optimal policy shifts occur, these greedy policies must always be optimal.

In deterministic environments, an injective reward function implies that $\forall s \in \St, a,a' \in \A$, $a$ and $a'$ have equal next-step expected reward iff $a\equiv_s a'$. Injectivity implies that $a$ and $a'$ are both greedily optimal iff $a\equiv_s a'$.
\end{proof}

In \cref{fig:always-shift}, for each $\gamma\in(0, 1)$, there exists a reward function whose optimal policy set has not yet ``settled down.'' However, when $\gamma\approx 1$, ``most'' reward functions have a Blackwell optimal policy set.

\begin{figure}[h!]\centering
    \begin{tikzpicture}[mygraph]
        %nodes
            \node[blue] (1) {$1$};
            \node[black] (2) [above right of=1] {$2$};
            \node[black] (3) [right of=1] {$3$};
            %\node[black] (4) [above right of=3] {};
            \node[black] (4) [right of=3] {$4$};
        %arrows
            \path[->] (1) edge (2);
            \path[->] (1) edge (3);
            \path[->] (3) edge (4);
            \path[->] (2) edge [looseness=6, out=45, in=135] (2);
            \path[->] (4) edge [looseness=6, out=0, in=90] (4);
     \end{tikzpicture}
    \caption[Optimal policy sets converge non-uniformly to Blackwell optimality]{Let $\gamma\in(0,1)$, and consider $R(\col{blue}{s_1})=R(\col{black}{s_3})\defeq0$,  $R(\col{black}{s_4})\defeq1$. The optimal policy set is not yet Blackwell-optimal if $R(\col{black}{s_2})\in(\gamma,1)$. \label{fig:always-shift}}
\end{figure}

\begin{restatable}[There exists a characterization of optimal policy shift existence in stochastic {\mdp}s]{conjSec}{conjStochShift}
\end{restatable}

\begin{restatable}[Optimal policy shift bound]{lem}{optPiBoundImproved}\label{lem:opt-pol-shift-bound-improved}
For fixed $R$, $\optPi$ can take on at most $(2\abs{\St}-1)\sum_s \binom{\abs{\F(s)}}{2}$ values over $\gamma \in (0,1)$.
\end{restatable}
\begin{proof}
By \cref{lem:opt-pol-visit-iff}, $\optPi$ changes value iff there is a change in optimality status for some visit distribution function at some state. By \cref{switch}, each pair of distinct visit distributions can switch optimality status at most $2\abs{\St}-1$ times. At each state $s$, there are $\binom{\abs{\F(s)}}{2}$ such pairs.
\end{proof}

\begin{restatable}[Linear bound on optimal visit distribution function shifts]{conjSec}{linBdVisitFn}
For any reward function $R$ and $\f,\f'\in\F(s)$,  $\f$ and $\f'$ shift at most $\abs{\St}-1$ times.
\end{restatable}

\begin{restatable}[Quadratic upper bound on optimal policy shifts]{conjSec}{quadUpperBd}
For any reward function $R$, at most $\mathcal{O}(\abs{\St}^2)$ optimal policy shifts occur.
\end{restatable}

\section{Optimality probability}
\Cref{fig:opt-prob-half-prob} serves as a reminder that the relatively greater optimality probability of an action $a$ at $\gamma$ does not imply that $\optprob{s,a,\gamma}\geq \half$.

\begin{figure}[h]
    \centering
    \begin{tikzpicture}[mygraph]
        \node[blue] (start) {$1$};
        \node[black] (2) [above left of=start] {};
        \path[->] (start) edge (2);
        \path[->] (2) edge[looseness=5, out=90, in=180] (2);
        \node[black] (3) [above of=start] {};
        \path[->] (start) edge (3);
        \path[->] (3) edge[looseness=5, out=45, in=135] (3);
        \node[black] (4) [above right of=start] {};
        \path[->] (start) edge (4);
        \path[->] (4) edge[looseness=5, out=45, in=135] (4);
        \node[black] (r) [right of=start] {};
        \path[->] (start) edge node[midway, below]{\texttt{right}} (r);
        \node[black] (a) [above right of=r] {};
        \path[->] (r) edge (a);
        \path[->] (a) edge[looseness=5, out=45, in=135] (a);
        \node[black] (b) [right of=r] {};
        \path[->] (r) edge (b);
        \path[->] (b) edge[looseness=5, out=0, in=90] (b);
    \end{tikzpicture}
    \caption[Optimality probability inequalities are relative]{By \cref{rsdIC}, $\optprob[\Diid]{s,\texttt{right},1}=\frac{2}{5}<\half$. In other words, when $\gamma\approx 1$, it's more probable than not that \texttt{right} \emph{isn't} optimal, even though \texttt{right} is more probable under optimality than other actions.}
    \label{fig:opt-prob-half-prob}
\end{figure}

\subsection{Impossibility of graphical characterization of which actions are more probably optimal}
Even restricting ourselves to $\Diid$ beliefs, we can't always just look at the rewardless {\mdp} structure to determine which actions are more probable under optimality. In \cref{fig:impossibility-graphical}, going \texttt{up} is more probable under optimality at $\col{blue}{s_1}$ for some state reward distributions $\Dist$, but not for others.

\begin{restatable}[The state reward distribution can affect which actions have the greatest optimality probability]{thm}{noICCrit}\label{no-ic-crit}
There can exist state reward distributions $\Dist_1, \Dist_2$, a state $s$, and a discount rate $\gamma$ for which $\argmax_{a}\optprob[\Diid[\Dist_1]]{s,a,\gamma}\neq \argmax_{a}\optprob[\Diid[\Dist_2]]{s,a,\gamma}$.
\end{restatable}
\begin{proof}
See \cref{fig:impossibility-graphical}.
\end{proof}

\begin{figure}[h!]\centering
    \begin{tikzpicture}[mygraph]
        %nodes
            \node[blue] (1) {$1$};
            \node[black] (2) [above right of=1] {};
            \node[black] (3) [right of=1] {};
            %\node[black] (4) [above right of=3] {};
            \node[black] (5) [right of=3] {};
        %arrows
            \path[->] (1) edge node[midway,left] {\texttt{up}} (2) ;
            \path[->] (1) edge node[midway,below] {\texttt{right}} (3);
            %\path[->] (3) edge (4);
            \path[->] (3) edge (5);
            \path[->] (2) edge [looseness=6, out=45, in=135] (2);
            %\path[->] (4) edge [looseness=6, out=45, in=135] (4);
            \path[->] (5) edge [looseness=6, out=0, in=90] (5);
     \end{tikzpicture}
    \caption[{\iid} reward distributions can disagree on which action is more probably optimal]{For $\Dist_1$ uniform, $\texttt{up}$ and $\texttt{right}$ are equally probable under optimality. For $\Dist_2$ with \textsc{cdf} $F(x)\defeq x^2$ on the unit interval,  $\optprob[\Diid[\Dist_2]]{\col{blue}{s_1},\texttt{up},\gamma}=\frac{10+3\gamma - 3\gamma^2}{20}$.}
    \label{fig:impossibility-graphical}
\end{figure}

\subsection{Sample means}
For arbitrary $D,D'\subseteq \RSDnd$, determining if $\optprob{D,1}>\optprob{D',1}$  is at least as hard as answering questions like ``for sample means $\bar{x}_n$ of $n$ \textsc{iid} draws from an arbitrary  continuous bounded distribution $\Dist$, is $\optprob[]{\max(\bar{x}_4, \bar{x}_4', \bar{x}_5)>\max(\bar{x}_4'',\bar{x}_5', \bar{x}_5'')}>\half$?''. These questions about maximal order statistics are often difficult to answer.

Thus, there is no simple characterization of the $D,D'\subseteq \RSDnd$ for which
\begin{equation*}
    \optprob{D,1}[\Diid]>\optprob{D',1}[\Diid].
\end{equation*}
However, there may be $\Dist$ for which $k$-cycle optimality probability decreases as $k$ increases.

\begin{restatable}[Increased sample size decreases maximality probability~\cite{se-sample-question}]{conjSec}{incSample}
Consider a finite set of sample means $\bar{x}_i$ of $k_i$ draws from $\text{unif}(0,1)$. If $k_i>k_j$, then $\optprob[]{\bar{x}_i =\max_l \bar{x}_l} < \optprob[]{\bar{x}_j =\max_l \bar{x}_l}$.
\end{restatable}
\begin{cor-no-num}
Suppose the environment is deterministic. Let $k>k'$ and $\dbf_k, \dbf_{k'}\in\RSDnd$ be $k,k'$-cycles, respectively. Suppose that
\begin{equation*}
    \lone{\set{\dbf_k}-\RSDnd\setminus\set{\dbf_k}}=\lone{\set{\dbf_{k'}}-\RSDnd\setminus\set{\dbf_{k'}}}=2.
\end{equation*}
For $\Dist$ uniform, $\optprob{\dbf_k,1}<\optprob{\dbf_{k'},1}$.
\end{cor-no-num}

\subsection{Optimality probability of linear functionals}
\begin{restatable}[Optimality probability changes continuously under $\Dcont$]{conjSec}{linFnChangeContinuous}
Let $D\subseteq \reals$ and let $G,H$ be two finite sets of continuous functions (or paths) from $D$ to $\rewardVS$. Define $G(\gamma)\defeq \set{g(\gamma) \mid g\in G}$, and similarly for $H(\gamma)$. $\phelper{G(\gamma)\geq H(\gamma)}[\Dcont]$ is continuous on $\gamma\in X$.
\end{restatable}

\subsection{Properties of optimality probability}

\begin{restatable}[Almost all reward functions don't have an optimal policy shift at any given $\gamma$]{cor}{ZeroShift}\label{zero-shift} For any $\gamma\in(0,1)$, $\optprob[R\sim \Dcont]{\lim_{\gamma^- \uparrow \gamma}\optPi[R,\gamma^-]\neq \optPi[R,\gamma]}=0$.
\end{restatable}
\begin{proof}
Let $\gamma\in(0,1)$. By \cref{pi-overlap}, if $R\in\rewardSpace$ has an optimal policy shift at $\gamma$, then $\exists s \in \St: \abs{\argmax_{\f\in\F(s)}\f(\gamma)^\top \rf}\geq 2$. At least one such $\f\in\Fnd(s)$ by \cref{lem:all-rf-max-nd} and \cref{lem:nd-relation}. Let $\f'\in\F(s)$ be a distinct element of the $\argmax$.

By \cref{lem:no-agree}, $\f(\gamma)\neq \f'(\gamma)$. Then $\abs{\argmax_{\dbf\in\F(s,\gamma)}\dbf^\top \rf}\geq 2$. \Cref{cor:distinct-maximized} shows that $\optprob[\rf\sim \Dcont]{\abs{\argmax_{\f\in\F(s)}\f(\gamma)^\top \rf}\geq 2}=0$.
\end{proof}

\begin{restatable}[For continuous reward function distributions, optimality probability is additive over visit distribution functions]{lem}{optAdd}\label{lem:opt-add}
Let $F \subseteq \F(s)$. $\optprob[\Dcont]{F,\gamma}=\sum_{\f\in F} \optprob[\Dcont]{\f,\gamma}$.
\end{restatable}
\begin{proof}
Suppose $\gamma\in(0,1)$. Since $\Dcont$ is continuous, apply \cref{prop:helper-cont-probability} with $A\defeq F(\gamma), B \defeq \F(s,\gamma)$.

Since the result holds for $\gamma\in(0,1)$, it holds in the limits of $\gamma\to0$ and $\gamma\to 1$.
\end{proof}

\begin{restatable}[Optimality probability is continuous on $\gamma$]{lem}{contMeas}\label{cont-meas}
For $F \subseteq  \F(s)$, $\optprob[\Dcont]{F,\gamma}$ is continuous on $\gamma\in [0, 1]$.
\end{restatable}
\begin{proof}
Since \cref{lem:opt-add} shows that  $\optprob[\Dcont]{F,\gamma}=\sum_{\f\in F} \optprob[\Dcont]{\f,\gamma}$, it is sufficient to show that each summand is continuous.

Let $\f \in F$. If $\optprob[\Dcont]{\f,\gamma}$ were discontinuous at $\gamma^* \in (0,1)$, then a positive measure subset of $\Dcont$ experiences an optimal policy shift at $\gamma^*$. This contradicts \cref{zero-shift}, and so $\optprob[\Dcont]{\f,\gamma}$ must be continuous on $\gamma\in(0,1)$. \Cref{prop:opt-prob-converge}  shows that optimality probability's limits exist, and so $\optprob[\Dcont]{\f,\gamma}$ is actually continuous on $\gamma \in [0,1]$.
\end{proof}

\begin{restatable}[Finite disagreement of optimality probability for $\Diid$]{conjSec}{finite-disagree}\label{meas-agree}
For any $\f,\f'\in\F(s)$, $\optprob[\Diid]{\f,\gamma}=\optprob[\Diid]{\f',\gamma}$ either for all $\gamma\in [0,1]$ or for finitely many $\gamma$.
\end{restatable}

\begin{restatable}[Only $\f\in\Fnd(s)$ have positive optimality probability at any $\gamma$]{lem}{posOptProb}\label{lem:exist-pos-opt}
$\f\in\Fnd(s)$ iff $\exists \gamma \in (0,1):\optprob[\Diid]{\f,\gamma} > 0$.
\end{restatable}
\begin{proof}
If $\f\in\Fnd(s)$, then $\exists \gamma\in(0,1),\rf_1\in\rewardVS: \f(\gamma)^\top\rf_1 > \max_{\f'\in\F(s)\setminus \set{\f}}\f'(\gamma)^\top \rf_1$. Then $\optprob[\Diid]{\f,\gamma}=\phelper{\f(\gamma)\geq \F(s,\gamma)}[\Diid] > 0$ by \cref{prop:helper-positive-prob}.

If $\exists \gamma \in (0,1):\optprob[\Diid]{\f,\gamma} > 0$, then $\f(\gamma)\in\ND{\F(s,\gamma)}$ by \cref{lem:nd-pos}. This implies $\f \in\Fnd(s)$ by \cref{lem:nd-relation}.
\end{proof}

\begin{restatable}[Dominated visit distributions are almost never optimal]{cor}{DomVanish}\label{dom-vanish}
$\f\in\F(s)$ is dominated iff $\forall \gamma\in[0,1]: \optprob[\Dcont]{\f,\gamma}=0$.
\end{restatable}
\begin{proof}
Let $\gamma \in (0,1)$. If $\f \in \F(s) \setminus \Fnd(s)$, then $\optprob[\Dcont]{\f,\gamma}=0$ by \cref{lem:exist-pos-opt}. Furthermore, $\optprob[\Dcont]{\f,0}\defeq \lim_{\gamma \to 0}\optprob[\Dcont]{\f,\gamma}=0$. Similar reasoning applies when $\gamma=1$.

We show that $\forall \gamma \in [0,1]:\optprob[\Dcont]{\f,\gamma}=0$ implies that $\f\in\F(s)\setminus\Fnd(s)$ by proving the contrapositive. Suppose $\exists \gamma\in(0,1): \optprob[\Dcont]{\f,\gamma}\neq0$. Then by \cref{lem:exist-pos-opt}, $\f$ cannot be dominated, and so $\f \in \Fnd(s)$.

If $\optprob[\Dcont]{\f,0}\defeq \lim_{\gamma\to 0} \optprob[\Dcont]{\f,\gamma}\neq 0$, then the optimality probability must be non-zero in a neighborhood of $0$. By \cref{lem:exist-pos-opt}, this can only be true if $\f\in\Fnd(s)$. Similar reasoning applies to the $\gamma=1$ case.
\end{proof}

\begin{restatable}[Non-domination iff positive measure]{prop}{ndvanish}\label{nd-vanish}
$\f\in\Fnd(s)$ iff $\forall \gamma \in (0,1): \optprob[\Diid]{\f,\gamma} > 0$.
\end{restatable}
\begin{proof}
By the same arguments used in \cref{lem:exist-pos-opt}'s proof's forward direction, $\optprob[\Diid]{\f,\gamma} > 0$. Therefore, $\forall \gamma\in(0,1):\optprob[\Diid]{\f,\gamma}>0$.

If  $\forall \gamma \in (0,1): \optprob[\Diid]{\f,\gamma} > 0$, apply \cref{dom-vanish} to conclude that $\f$ cannot be dominated (\cref{dom-vanish} applies because $\Diid$ is continuous), and so $\f\in\Fnd(s)$.
\end{proof}

\begin{restatable}[Non-domination iff positive probability for $\gamma \in [0,1)$]{prop}{ndVanishExtended}\label{nd-vanish-extended}
\[\f\in\Fnd(s) \text{ iff } \forall \gamma \in [0,1): \optprob[\Diid]{\f,\gamma} > 0.\]
\end{restatable}
\begin{proof}
The case for $\gamma \in (0,1)$ is proved by \cref{nd-vanish}.

For the $\gamma=0$ case, suppose $\f\in\Fnd(s)$ and fix $\gamma^* \in (0,1)$. By \cref{unique-actions}, almost all reward functions in the interior of $\supp[\f(\gamma^*)\geq \F(s,\gamma^*)]\cap \supp$ have optimal actions at each state which are unique up to action equivalence. Let $R$ be one such reward function for which $\f\in\Fnd(s)$ is strictly optimal.

By \cref{prop:almost-transfer}, $\optPi[\OptVf{\cdot,\gamma^*},0]=\optPi[R,\gamma^*]$. Since optimal value is continuous on the reward function (\cref{cont-vfn}) and since $\f$ is strictly optimal for $R$, there exists an $\epsilon$-ball of reward functions around the \emph{reward function} $\OptVf{\cdot,\gamma^*}$ for which $\f$ is strictly optimal. By \cref{lem:pos-aff-nd-invar}, we can positively affinely transform this ball to be contained in $\supp[\f,0]\cap \supp$; since this is a positive affine transformation, the image has non-empty interior. Then $\optprob[\Diid]{\f,0}>0$ since $\Diid$ is a continuous distribution.

Suppose $\f \in \F(s) \setminus \Fnd(s)$. By \cref{unique-optimal}, for any $\gamma^* \in (0,1)$, the set of reward functions with multiple optimal visit distributions has measure zero. Dominated visit distribution functions cannot be uniquely optimal at any $\gamma^* \in (0,1)$. Therefore, if $\f$ is dominated, $\forall \gamma^* \in (0,1): \optprob[\Diid]{\f,\gamma^*}=0$, and $\optprob[\Diid]{\f,0}\defeq \lim_{\gamma^*\to 0} \optprob[\Diid]{\f,\gamma^*}=\lim_{\gamma^*\to 0} 0=0$. So for any $\gamma \in [0,1)$, $\optprob[\Diid]{\f,\gamma}>0$ implies $\f \in \Fnd(s)$.
\end{proof}

\begin{restatable}[Non-dominated child distributions facilitate a non-dominated visit distribution function]{lem}{ndChildndExist}\label{lem:nd-child-nd-exist}
If $T(s,a)\in \childDistsNd$, then $\exists \pi: \pi(s)=a$ and $\fpi{}\in\Fnd(s)$.
\end{restatable}
\begin{proof}
Let $\dbf\defeq T(s,a)$ and let $V\defeq \set{\mathbf{v} \in \rewardVS \mid \dbf^\top > \max_{\dbf'\in\childDists\setminus \set{\dbf'}} \dbf'^\top \mathbf{v}}$. $V$ is nonempty because $\dbf\in\childDistsNd$ and has positive Lebesgue measure by the continuity of $\dbf^\top\mathbf{v}$ and the $\max$ on $\mathbf{v}\in\rewardVS$.

For all $\pi\in\Pi$, let $V_\pi\defeq \set{\mathbf{v}\in V \mid \pi\in \greedy[\mathbf{v}]}$. Since $\Pi$ is finite, there must exist $\pi^*$ such that $V_{\pi^*}\subseteq V$ has positive Lebesgue measure. Then consider the set of reward functions $R_{\pi^*}\defeq \set{\prn{\mathbf{I}-.5\mathbf{T}^{\pi^*}}\mathbf{v}\mid \mathbf{v} \in V_{\pi^*}}$. Since $\prn{\mathbf{I}-.5\mathbf{T}^{\pi^*}}$ is invertible because its spectral radius is strictly less than $1$ (see the proof of \cref{lem:exist-visit}), $\det \prn{\mathbf{I}-.5\mathbf{T}^{\pi^*}}\neq 0$ and so $R_{\pi^*}$ also has positive Lebesgue measure.

But then almost all reward functions $\rf$ in $R_{\pi^*}$ must have a unique optimal visit distribution $\fpi[\pi']{}\in\F(s)$ when $\gamma=.5\in(0,1)$ by \cref{unique-optimal}. Therefore, $\fpi[\pi']{}\in \Fnd(s)$ by \cref{def:nd}. Lastly, since $\forall \rf' \in R_{\pi^*}: \optPi[\rf',.5]=\greedy[\mathbf{v}']$ (where each $\rf'\defeq \prn{\mathbf{I}-.5\mathbf{T}^{\pi^*}}\mathbf{v}'$), $\pi'(s)\equiv_s a$ by the definitions of $V$ and $\greedy[\mathbf{v}']$.

Let $\pi$ equal $\pi'$, except that $\pi(s)=a$. Because $\pi'(s)\equiv_s a = \pi(s)$, $\fpi{}=\fpi[\pi']{}\in \Fnd(s)$. We have thus shown that $\exists \pi \in \Pi: \pi(s)=a$ and $\fpi{}\in\Fnd(s)$.
\end{proof}

\begin{restatable}[$\Fnd(s)$ controls optimality probability]{prop}{FndControl}\label{prop:fnd-control}
Let $F\subseteq\F(s)$. $\optprob[\Dcont]{F,\gamma}=\optprob[\Dcont]{F\cap\Fnd(s),\gamma}$.
\end{restatable}
\begin{proof}
\begin{align}
    \optprob[\Dcont]{F,\gamma}&=\sum_{\f\in F}\optprob[\Dcont]{\f,\gamma}\label{eq:cont-sum}\\
    &= \sum_{\f\in F\cap \Fnd(s)}\optprob[\Dcont]{\f,\gamma}\label{eq:eliminate-dom}\\
    &=\optprob[\Dcont]{F\cap\Fnd(s),\gamma}.\label{eq:cont-sum-2}
\end{align}
\Cref{eq:cont-sum} and \cref{eq:cont-sum-2} follow by \cref{lem:opt-add} since $\Dcont$ is continuous. \Cref{eq:eliminate-dom} follows by \cref{dom-vanish}, since dominated visit distribution functions have $0$ optimality probability under continuous reward function distributions.
\end{proof}

\begin{restatable}[Only non-dominated transitions are greedily optimal with positive probability]{lem}{greedyTransitionProb}\label{lem:greedy-opt-prob}
Let $T(s,a) \in \childDists$. $T(s,a) \in \childDistsNd$ iff $\optprob[\Diid]{s,a,0}>0$.
\end{restatable}
\begin{proof}
Suppose that $T(s,a) \in \childDistsNd$. This means that $\exists \pi \in \Pi: \pi(s)=a$ and $\fpi{}\in\Fnd(s)$ by \cref{lem:nd-child-nd-exist}, and so $\optprob[\Diid]{\fpi{},0}>0$ by \cref{nd-vanish-extended}. Then $\optprob[\Diid]{s,a,0}\geq \optprob[\Diid]{\fpi{},0}>0$.

Suppose that $\optprob[\Diid]{s,a,0}>0$. \Cref{nd-vanish-extended} implies that $\optprob[\Diid]{s,a,0}>0$ iff $\exists \pi \in \Pi: \fpi{} \in \Fnd(s)$ such that $\pi(s)=a$, since $\optprob[\Diid]{s,a,0}=\optprob[\Diid]{\FRestrictAction{s}{a},0}$ by \cref{lem:equiv-action-opt}.

\Cref{lem:dom-child} shows that if $T(s,a)\in \childDists\setminus \childDistsNd$ and $\pi(s)=a$, then $\fpi{}\in\F(s)\setminus \Fnd(s)$. The contrapositive is then: if $\fpi{} \in \Fnd(s)$, then either $\pi(s)\neq a$ or $T(s,a) \in \childDistsNd$. But $\pi(s)=a$, so the fact that $\optprob[\Diid]{s,a,0}>0$ implies that $\fpi{} \in \Fnd(s)$, which implies that $T(s,a) \in \childDistsNd$.
\end{proof}

\begin{restatable}[Similarity to a dominated action implies domination]{cor}{childSimDom}\label{prop:child-sim-dom}
If $T(s,a)\in \childDists\setminus \childDistsNd$ is similar to $T(s,a') \in \childDists$ via a permutation $\phi$ such that $\phi\cdot \childDistsNd=\childDistsNd$, then $T(s,a') \not \in \childDistsNd$.
\end{restatable}
\begin{proof}
Apply \cref{prop:child-sim-greedy} to conclude that $\optprob[\Diid]{s,a,0}=\optprob[\Diid]{s,a',0}$. By \cref{lem:greedy-opt-prob}, $\optprob[\Diid]{s,a,0}=0$. Then $\optprob[\Diid]{s,a',0}=0$; apply \cref{lem:greedy-opt-prob} to conclude that $T(s,a') \not \in \childDistsNd$.
\end{proof}

\begin{restatable}[Is there anything to be gained by formalizing the optimality probability of sets of policies?]{questionSec}{polSetOptProb}
\end{restatable}

\begin{restatable}[$\Fnd$ multi-state restriction]{definition}{DefMultiRestrict}\label{def:restrict-multi}
$\FRestrict{S}\subseteq \F(s)$ contains the non-dominated visit distributions induced by a policy which agrees with $\pi$ on the states of $S\subseteq \St$. $\FndRestrict{S}\defeq \FndRestrict{S}\cap\Fnd(s)$.
\end{restatable}

\begin{restatable}[Optimality probability factorizes]{prop}{PropOptimalityFact}\label{prop:factorization} Let $\fpi{}\in\Fnd(s)$ and let $\gamma\in[0,1)$.
\begin{equation}
    \optprob[\Diid]{\fpi{},\gamma} = \prod_{i=1}^{\abs{\St}}  \dfrac{\optprob[\Diid]{\FndRestrict{\{s_1,\ldots,s_{i-1},s_i\}},\gamma}}{\optprob[\Diid]{\FndRestrict{\{s_1,\ldots,s_{i-1}\}},\gamma}}.\label{eq:factorization}
\end{equation}
\end{restatable}
\begin{proof}
\Cref{eq:factorization}  holds for any state indexing and is well-defined on $\gamma \in [0,1)$ because $\fpi{}\in\Fnd(s)$ has $\optprob[\Diid]{\f,\gamma}>0$ on that domain (\cref{nd-vanish}).
\end{proof}

\begin{restatable}[Optimality probability factorization]{definition}{DefOptimalityFact}\label{def:factorization}
The \emph{optimality probability factorization of $\fpi{}\in\Fnd(s)$} is the RHS of \cref{eq:factorization}.
\end{restatable}

\begin{restatable}[Optimality probability varies iff a factor varies]{lem}{VaryIff}
Let $\f\in\Fnd(s)$. $\optprob[\Diid]{\f,\gamma}$ varies with $\gamma$ iff its factorization has a factor which varies with $\gamma$.
\end{restatable}
\begin{proof}
The forward direction is trivial. For the backward direction, suppose factor $i=k$ equals the non-constant function $f(\gamma)$; note that $\forall \gamma\in[0,1): f(\gamma)>0$. Suppose that the product of other factors equals $\frac{c}{f(\gamma)}$. If one action becomes more likely at $s_k$, then another action must become less likely, and vice versa. $\frac{1}{f(\gamma)}$ cannot be the multiplicative inverse (up to a constant) for both of these variations, and so $\optprob[\Diid]{\f',\gamma}$ must vary with $\gamma$ for some $\f'\in\Fnd(s)$.
\end{proof}

\begin{restatable}[In deterministic environments, constant optimality probability implies rational probabilities]{conjSec}{ratProb}
Suppose the {\mdp} is deterministic and $s$ is such that $\forall \f \in \F(s):\optprob[\Diid]{\f,\gamma}$ does not vary with $\gamma\in[0,1]$. Then $\forall \f \in \F(s):\optprob[\Diid]{\f,\gamma}\in [0,1]\cap\mathbb{Q}$.
\end{restatable}

\begin{restatable}[Positive optimality probability under $\Dcont$ implies $\Fnd$ membership and $\RSDnd$ membership]{prop}{OneGammaRSDIff}\label{onegammaRSDIff}
$\optprob[\Dcont]{\f,1} > 0$ implies $\f\in\Fnd(s)$ and $\lim_{\gamma\to 1} (1-\gamma)\f(\gamma)\in\RSDnd$.
\end{restatable}
\begin{proof}
Suppose $\optprob{\f,1} > 0$. By \cref{dom-vanish}, $\f$ cannot be dominated, else
\begin{equation*}
    \lim_{\gamma\to 1} \optprob{(1-\gamma)\f(\gamma),1}=0.
\end{equation*}
Therefore, $\f\in\Fnd(s)$. By \cref{prop:helper-positive-prob}, $\lim_{\gamma\to 1} (1-\gamma)\f(\gamma)\in\RSDnd$.
\end{proof}

\begin{restatable}[$\RSDnd$ membership and $\Fnd$ membership implies positive {\iid} optimality probability]{conjSec}{optProbGammaOne}
If $\f\in\Fnd(s)$ and $\lim_{\gamma\to 1} (1-\gamma)\f(\gamma)\in\RSDnd$, then $\optprob[\Diid]{\f,1} > 0$.
\end{restatable}

\begin{restatable}[Optimality probabilities reach ordinal equilibrium as $\gamma\to 1$]{conjSec}{ordinalEqProb}
For any $\f,\f'\in\F(s)$, $\optprob[\Diid]{\f,\gamma}$ and $\optprob[\Diid]{\f',\gamma}$ reach ordinal equilibrium as $\gamma\to 1$.
\end{restatable}

\begin{restatable}[Each non-dominated visit distribution ``takes'' optimality probability from all other non-dominated visit distributions]{conjSec}{ndTakes}
Let $\f,\f'\in\Fnd(s)$, where $\f\neq \f'$. Then $\forall \gamma\in(0,1)$:
\begin{equation}
    \optprob[\Diid]{\f,\gamma} < \optprob[\rf \sim \Diid]{\f(\gamma)^\top\rf = \max_{\f''\in\Fnd(s)\setminus\set{\f'}}\f''(\gamma)^\top\rf}.
\end{equation}
\end{restatable}

\subsection{Properties of instrumental convergence}

Instrumental convergence exists when actions have different optimality probabilities.
\begin{restatable}[Existence of instrumental convergence]{definition}{defConvInstru}
\emph{Instrumental convergence starting at state $s$} when $\exists a,a'\in\A,\gamma\in[0,1]: \optprob{s,a,\gamma}\neq \optprob{s,a',\gamma}$.
\end{restatable}

\begin{restatable}[Instrumental convergence exists at almost all discount rates, if it exists]{conjSec}{almostAllRI}
If instrumental convergence exists starting at state $s$ for some $\gamma$, it exists starting at $s$ for almost all $\gamma \in [0,1]$.
\end{restatable}

\begin{restatable}[Optimality probability is unaffected by unreachable states]{lem}{ICnounreachable}\label{IC-no-unreachable}
If $s$ cannot reach $s'$, then $\forall a: \optprob[\Dany]{\FRestrictAction{s'}{a},\gamma}=1$.
\end{restatable}
\begin{proof}
Policies can output any action at unreachable states without affecting the induced visit distribution.
\end{proof}

\begin{restatable}[Instrumental convergence without domination or stochasticity implies {\iid} optimality probability varies with $\gamma$]{prop}{ICImpliesMeasVary}\label{ICImpliesMeasVary}
In deterministic environments, if instrumental convergence exists at some $\gamma^*$ starting from $s$ and if $\F(s)=\Fnd(s)$, then  $\exists \f:\optprob[\Diid]{\f,\gamma}$ varies with $\gamma$.
\end{restatable}
\begin{proof}
Let $a$ be more probably optimal than $a'$ at state $s'$ (thus fulfilling \cref{def:action-optimality}). Since the environment is deterministic and $\F(s)=\Fnd(s)$, it is equally probable that each child of $s'$ is the greedy choice at $\gamma=0$. Then
\begin{equation*}
    \optprob[\Diid]{\Fnd(s\,\vert\,\pi(s')=a), 0} = \optprob[\Diid]{\Fnd(s\,\vert\,\pi(s')=a'), 0}.
\end{equation*}
Therefore, at least one $\optprob[\Diid]{\f,\gamma}$ must vary on $\gamma \in [0, \gamma^*]$ so that
\begin{equation*}
    \optprob[\Diid]{\Fnd(s\,\vert\,\pi(s')=a), \gamma^*}> \optprob[\Diid]{\Fnd(s\,\vert\,\pi(s')=a'), \gamma^*}.
\end{equation*}
\end{proof}

\begin{figure}[ht]\centering
    \begin{tikzpicture}[mygraph]
        %nodes
            \node[blue] (1) {$1$};
            \node[black] (2) [above right of=1] {};
            \node[black] (3) [right of=1] {};
            \node[black] (4) [above right of=3] {};
            \node[black] (5) [right of=3] {};
        %arrows
            \path[->] (1) edge (2);
            \path[->] (1) edge (3);
            \path[->] (3) edge (4);
            \path[->] (3) edge (5);
            \path[->] (2) edge [looseness=6, out=45, in=135] (2);
            \path[->] (4) edge [looseness=6, out=45, in=135] (4);
            \path[->] (5) edge [looseness=6, out=0, in=90] (5);
     \end{tikzpicture}
\caption[Discontinuous reward function distributions allow optimality probability inequality without that probability varying with $\gamma$]{Our ongoing assumption of $\Dist$'s continuity is required for \cref{ICImpliesMeasVary}. Under the uniform distribution on $\{0,1\}^\St$, the visit distribution going up from $\col{blue}{s_1}$ has probability $\frac{10}{24}$, while the other two visit distributions have probability $\frac{7}{24}$. However, under the uniform distribution on $[0,1]^\St$, the upwards visit distribution has probability $\frac{3-\gamma}{6}$, while the other two each have probability $\frac{3+\gamma}{12}$.}
\end{figure}

\Cref{ICImpliesMeasVary} shows the following while making the strong assumption that $\F(s)=\Fnd(s)$.

\begin{restatable}[In deterministic environments, instrumental convergence implies variable optimality probability]{conjSec}{ICNoAssumption}
In deterministic environments, if instrumental convergence exists at some $\gamma^*$ starting from $s$, then  $\exists \f:\optprob[\Diid]{\f,\gamma}$ varies with $\gamma$.
\end{restatable}

\begin{restatable}[Optimal policy shifts necessary for instrumental convergence]{conjSec}{optPolInstr}
In deterministic environments, if optimal policy shifts cannot occur, then instrumental convergence does not exist.
\end{restatable}

\begin{remark}
\Cref{ICImpliesMeasVary}'s assumption of determinism is required: suppose state $s$ has $\children[s]=\set{s_1, s_2, s_3}$ and $\abs{\A}=3$. $T(s,a_1)=\unitvec[s_1]$, $T(s,a_2)=(0,.5,.5)^\top$, and $T(s,a_3)=\unitvec[s_3]$. When $\Dist$ is uniform and $\gamma \approx 0$, action $a_1$ will be strictly more probably optimal than actions $a_2$ or $a_3$.
\end{remark}

\begin{restatable}[No domination, stochasticity, or optimal policy shifts means equal optimality probabilities]{prop}{noShiftNoIC}\label{no-shift-no-ic}
In deterministic environments where optimal policy shifts cannot occur, if $\F(s)=\Fnd(s)$, then no instrumental convergence exists starting from $s$.
\end{restatable}
\begin{proof}
Since the environment is deterministic and $\F(s)=\Fnd(s)$, \cref{ICImpliesMeasVary} applies. But $\lnot \exists \f \in \F(s): \optprob[\Diid]{\f,\gamma}$ which varies with $\gamma$, because optimal policies cannot shift.
\end{proof}

\begin{restatable}[How does entropy relate to instrumental convergence?]{questionSec}{entropyQ}
Is there a formal relationship between the ``degree'' of instrumental convergence and the entropy of the distribution over optimal policy sets induced by reward function distribution $\D$?
\end{restatable}

\section{Properties of recurrent state distributions}

\begin{restatable}[In deterministic envs., $\lim_{\gamma\to 1}(1-\gamma)\Fnd(s,\gamma)=\RSDnd$]{conjSec}{detNdOne}\label{conj:det-nd}
In deterministic environments, if $\f\in\Fnd(s)$, then $\lim_{\gamma\to 1}(1-\gamma) \f(\gamma) \in \RSDnd$.
\end{restatable}

\begin{cor-no-num}
In deterministic environments, $\optprob[\Diid]{\f,1}>0$ iff $\f\in\Fnd(s)$.
\end{cor-no-num}

\Cref{conj:det-nd} does not hold in stochastic environments; see \cref{fig:multichain}. Conversely, some dominated $\f$ do limit to non-dominated $\dbf\in\RSDnd$; for example, consider a state $s$ in a unichain {\mdp} in which there are dominated $\f\in\F(s)$ but $\abs{\RSD}=1$.

\begin{figure}[ht]
    \centering
    \vspace{-15pt}
    \begin{tikzpicture}[mygraph,node distance=1.3cm]
        %nodes
            \node[blue] (1) {1};
            \node [black, right of=1] (stoch) {2};
            \node [black, right of=stoch, node distance=.4cm, shorten >=0cm, shorten <=.0cm,inner sep=0pt,minimum size=.4pt] (split) {};
            \node[black, above right of=stoch] (a) {3};
            \node[black, below right of=stoch] (b) {4};
        %arrows
            \path[->] (1) edge (stoch);
            \path[->] (1) edge (a);
            \path[->] (1) edge (b);
            \path[-] (stoch) edge node[right] {$\, a$} (split);
            \path[-] (split) edge (stoch);
            \path[->] (split) edge (a);
            \path[->] (split) edge (b);
            \path[->] (a) edge [looseness=7,out=45,in=-45] (a);
            \path[->] (b) edge [looseness=7,out=-45,in=45] (b);
    \end{tikzpicture}
    \vspace{-15pt}
    \caption[Not all non-dominated visit distributions induce non-dominated {\rsd}s]{The bifurcated action $a$ is a stochastic transition, where $T(\col{black}{s_2},a, \col{black}{s_3})=\half=T(\col{black}{s_2},a, \col{black}{s_4})$. By \cref{sufficient-nd}, navigating from $\col{blue}{s_1}\to s_2$ induces a non-dominated visit distribution $\f$. However, its limiting \textsc{rsd} is half $s_3$, half $s_4$ and is therefore dominated. $\optprob[\Diid]{\f,1}=0$ even though $\f\in\Fnd(s)$.}
    \label{fig:multichain}
\end{figure}

\begin{restatable}[If $s'$ can reach $s$ deterministically, ${\RSD[s']} \subseteq \RSD$]{prop}{rsdReach}\label{c-rsd-reach-subset-deterministic}
\end{restatable}
\begin{proof}
Let $\dbf^\pi \in \RSD[s']$. Starting from $s'$, $\pi$ induces state trajectory $s' s_1 s_2\ldots$. Let $\pi'$ navigate to a state $s_k$ in this trajectory which $s$ can reach in the fewest steps (where $s$ is considered to ``reach'' itself in $0$ steps); since $s$ can deterministically reach $s'$, this fewest number of steps is finite. $\forall t \geq k: \pi'(s_t)\defeq \pi(s_t)$; $\dbf^{\pi'}=\dbf^\pi$. $\pi'$ is stationary because it navigates to the state trajectory in the fewest possible number of steps, and therefore it does not conflict with itself.
\end{proof}

\begin{figure}[ht]
    \centering
    \begin{tikzpicture}[mygraph,node distance=1.3cm]
        %nodes
            \node[blue] (1) {1};
            \node [black, right of=1] (stoch) {2};
            \node [black, right of=stoch, node distance=.4cm, shorten >=0cm, shorten <=.0cm,inner sep=0pt,minimum size=.4pt] (split) {};
            \node[red, above right of=stoch] (a) {3};
            \node[green, below right of=stoch] (b) {4};
        %arrows
            \path[->] (1) edge (stoch);
            \path[->] (a) edge  [out=160, in=90] (1);
            \path[-] (stoch) edge node[right] {$\, a$} (split);
            \path[-] (split) edge (stoch);
            \path[->] (split) edge (a);
            \path[->] (split) edge (b);
            \path[->] (a) edge [looseness=7,out=45,in=-45] (a);
            \path[->] (b) edge [looseness=7,out=-45,in=45] (b);
    \end{tikzpicture}
    \caption[Reachability via stochastic transitions does not ensure {\rsd} containment]{The bifurcated action $a$ is a stochastic transition, where $T(\col{black}{s_2},a, \col{black}{s_3})=\half=T(\col{black}{s_2},a, \col{black}{s_4})$. \Cref{c-rsd-reach-subset-deterministic} does not hold in stochastic environments.}
    \vspace{-5pt}
     \begin{align*}
         \RSD[\col{blue}{s_1}]&= \set{\colvec{4}{\col{blue}{0}}{0}{\col{red}{.5}}{\col{green}{.5}}, \colvec{4}{\col{blue}{0}}{0}{\col{red}{0}}{\col{green}{1}}}\neq   \set{\colvec{4}{\col{blue}{0}}{0}{\col{red}{1}}{\col{green}{0}}, \colvec{4}{\col{blue}{0}}{0}{\col{red}{0}}{\col{green}{1}}}=\RSD[\col{red}{s_3}]
     \end{align*}
    \label{fig:rsd-disagree}
\end{figure}

\begin{remark}
\Cref{c-rsd-reach-subset-deterministic} is not true in some stochastic settings, even if $s$ can reach $s'$ with probability $1$, since it may be the case that only non-stationary policies navigate to $s'$ and then induce the appropriate \textsc{rsd}.
\end{remark}

\begin{restatable}[Reachability with probability $1$ implies uniformly greater average reward]{lem}{probOneReach}\label{lem:prob-1-reach}
If
\begin{equation}
    \max_{\pi\in\Pi} \lim_{t\to \infty} \prob{s' \text{ reached in the first } t \text{ steps while following }\pi\text{ starting from } s}=1,\label{eq:limit-prob}
\end{equation}

then $\forall R \in \rewardSpace: \VfNorm{s,1}\geq\VfNorm{s',1}$.
\end{restatable}
\begin{proof}
Let $\rf\in\rewardVS$ and let $\dbf\in\RSD[s']$ be such that $\VfNorm{s',1}=\dbf^\top \rf$. In \cref{eq:limit-prob}, each limit exists because the probability is monotonically increasing and bounded $[0,1]$, and the maximum exists because $\Pi$ is finite. Let $\pi$ maximize \cref{eq:limit-prob}.

$\pi$ has probability $1$ of eventually reaching $s'$. Let $\pi^\text{HD}$ implement this strategy until it has reached $s'$ at least once in its history. After the agent has reached state $s'$, $\pi^{\text{HD}}$ is defined so as to induce $\dbf$. This is possible because $\dbf\in\RSD[s']$. Since $\gamma=1$, $\pi^\text{HD}$ therefore induces $\dbf$ starting from $s$. Therefore, starting from $s$, a history-dependent policy can achieve $\VfNorm{s',1}$ average reward for reward function $\rf$.

\citet{puterman_markov_2014}'s Theorem 6.2.7 shows that when $0 < \gamma < 1$ in finite {\mdp}s, discounted optimal value is invariant to restriction to deterministic stationary policies. If $s$ could not induce an average reward of at least $\VfNorm{s',1}$ via a stationary policy, but could with a history-dependent policy, this would contradict Theorem 6.2.7 by the continuity of $\VfNorm{s,\gamma}$ on $\gamma\in [0,1]$, which is shown by \cref{lem:norm-value-lip} in \citet{turner_optimal_2020}. That is, if no such gain-optimal stationary policy existed, then stationary optimal policies would be strictly worse than non-stationary optimal policies for $\gamma\approx 1$. Therefore, when starting from state $s$, there exists some stationary policy with average reward of at least $\VfNorm{s',1}$.
\end{proof}

\begin{restatable}[When $\gamma=1$, {\iid} $\pwrNoDist$ decreases iff \textsc{rsd}s become unreachable]{cor}{farsightedPower}\label{farsightedPower}
If $s$ can reach $s'$ with probability $1$, then $\pwr[s',1]\leq \pwr[s,1]$. If the environment is deterministic, the inequality is strict iff $\RSDnd[s']\subsetneq \RSDnd[s]$.
\end{restatable}
\begin{proof}
Since $s$ can reach $s'$ with probability $1$, \cref{lem:prob-1-reach} implies that $\pwr[s',1]\leq \pwr[s,1]$.

Given determinism, \cref{c-rsd-reach-subset-deterministic} implies that $\RSDnd[s']\subseteq \RSDnd[s]$. Then if the inequality is strict, $\RSDnd[s']\subsetneq \RSDnd[s]$ because $\pwr[s',1]$ is determined by the availability of non-dominated \textsc{rsd}s by \cref{lem:gamma-1-power}. If $\RSDnd[s']\subsetneq \RSDnd[s]$, the inequality is strict by \cref{RSDSimPower} via $\phi$ the identity permutation.
\end{proof}

\begin{remark}
\Cref{farsightedPower} is false if $s$ can only reach $s'$ with positive probability less than $1$. For example, suppose action $a$ is such that $T(s,a,s_\text{high})=.5=T(s,a,s_\text{low})=.99$ and all actions are equivalent to $a$ at state $s$. Further suppose that $\pwr[s_\text{high},1]>\pwr[s_\text{low},1]$. Then
\begin{equation*}
    \pwr[s,1]=\half \prn{\pwr[s_\text{high},1]+\pwr[s_\text{low},1]}
\end{equation*}
is intermediate between the two values.
\end{remark}

\begin{restatable}[Communicating {\mdp}]{definition}{DefCommunicating1}
An {\mdp} is \emph{communicating} when every state is able to reach every other state with positive probability.
\end{restatable}

\begin{restatable}[When $\gamma=1$ in communicating {\mdp}s, $\pwr$ is equal everywhere]{prop}{commPower}\label{comm-power}
In communicating {\mdp}s, $\forall s,s': \pwr[s, 1]=\pwr[s',1]$.
\end{restatable}
\begin{proof}
Since the {\mdp} is communicating, all states can reach each other with asymptotic probability $1$. Apply \cref{farsightedPower}  to conclude that $\pwr[s, 1]=\pwr[s',1]$.
\end{proof}

\section{\texorpdfstring{$\pwrNoDist$}{POWER}}
\subsection{The structure of \texorpdfstring{$\pwrNoDist$}{POWER}}
\begin{restatable}[Graphical characterization of {\iid} $\pwrNoDist$ agreement]{conjSec}{powChar}
There exists a graphical condition characterizing when, for all bounded state reward distributions $\Dist$, states $s,s'$ are such that $\forall \gamma \in {[0,1]:\pwr[s,\gamma][\Diid]=\pwr[s',\gamma][\Diid]}$.
\end{restatable}

\begin{restatable}[The structure of $\pwrNoDist$]{thm}{structPwr}\label{thm:pwr-struct}
Let $\gamma \in (0,1)$ and let $\Dbd$ induce probability measure $F$.
\begin{equation}
    \pwr[s,\gamma][\Dbd] = (1-\gamma)\sum_{t=1}^\infty \gamma^{t-1} \sum_{\fpi{} \in \Fnd(s)} \int_{\opt[\fpi{}(\gamma)\geq \F(s,\gamma)]} \prn{\prn{\mathbf{T}^\pi}^t\unitvec}^\top \rf \dF[\rf].
\end{equation}
\end{restatable}
\begin{proof}
Suppose $\gamma \in (0,1)$.
\begin{align}
    \vavg[s,\gamma][\Dbd] &\defeq \int_{\supp[\Dbd]} \max_{\fpi{} \in \F(s)}\fpi{}(\gamma)^\top \rf \dF[\rf]\\
    &= \sum_{\fpi{} \in \F(s)}\int_{\opt[\fpi{}(\gamma)\geq \F(s,\gamma)]} \fpi{}(\gamma)^\top \rf \dF[\rf]\\
    &= \sum_{\fpi{} \in \Fnd(s)}\int_{\opt[\fpi{}(\gamma)\geq \F(s,\gamma)]} \fpi{}(\gamma)^\top \rf \dF[\rf]\label{eq:nd-restrict-pwr-expand}\\
    &= \sum_{\fpi{} \in \Fnd(s)}\int_{\opt[\fpi{}(\gamma)\geq \F(s,\gamma)]} \prn{\sum_{t=0}^\infty \prn{\gamma \mathbf{T}^\pi}^t\unitvec}^\top \rf \dif F(\rf)\label{eq:def-visit-pwr-expand}\\
    &= \sum_{\fpi{} \in \Fnd(s)}\int_{\opt[\fpi{}(\gamma)\geq \F(s,\gamma)]} \sum_{t=0}^\infty \prn{\prn{\gamma \mathbf{T}^\pi}^t\unitvec}^\top \rf \dF[\rf]\\
    &=  \sum_{\fpi{} \in \Fnd(s)} \sum_{t=0}^\infty \gamma^t \int_{\opt[\fpi{}(\gamma)\geq \F(s,\gamma)]} \prn{\prn{\mathbf{T}^\pi}^t\unitvec}^\top \rf \dF[\rf]\label{eq:interchange-pwr-expand}\\
    &=  \sum_{t=0}^\infty \gamma^t \sum_{\fpi{} \in \Fnd(s)} \int_{\opt[\fpi{}(\gamma)\geq \F(s,\gamma)]} \prn{\prn{\mathbf{T}^\pi}^t\unitvec}^\top \rf \dF[\rf].\label{eq:interchange-pwr-expand-2}
\end{align}

\Cref{eq:nd-restrict-pwr-expand} follows because optimal value is invariant to restriction to non-dominated visit distribution functions. \Cref{eq:def-visit-pwr-expand} holds by the definition of a visit distribution (\cref{def:visit}).

\Cref{eq:interchange-pwr-expand} holds by Fubini's theorem: for a fixed $\pi$, consider the function $f(t,\rf \mid \pi)\defeq \prn{\prn{\gamma \mathbf{T}^\pi}^t\unitvec}^\top \rf$.  Since $\Dist$ is bounded $[a,b]$, $\abs{f(t,\rf \mid \pi)}\leq \gamma^t \abs{b}$ for all $t\geq 0, \rf \in \supp$. Since $\gamma \in (0,1)$,
\begin{equation}
\int_{\opt[\fpi{}(\gamma)\geq \F(s,\gamma)]} \sum_{t=0}^\infty \abs{f(t,\rf \mid \pi)} \dF[\rf] \leq \geom[\abs{b}]<\infty.
\end{equation}

Furthermore, for a fixed $t$, $f$ is continuous on $\rf$ (as the function is linear when $t$ is fixed). Therefore, $f : \mathbb{Z}_{\geq 0}\times \rewardVS \to \reals$ is continuous: the preimage of an open set in $\reals$ is open in the product topology on the domain, since the standard topology on $\mathbb{Z}_{\geq 0}$ is discrete. Since $f$ is continuous, it is measurable. Therefore, the conditions of Fubini's theorem are met. \Cref{eq:interchange-pwr-expand-2} holds because $\Fnd(s)$ is finite.

We now show the desired $\pwrNoDist$ identity.
\begin{align}
    &\pwr[s,\gamma][\Dbd] \\
    &\defeq \E{\rf\sim\D}{\max_{\f\in \F(s)} \frac{1-\gamma}{\gamma}\prn{\f(\gamma)-\unitvec}^\top \rf}\\
    &= \frac{1-\gamma}{\gamma}\prn{-\E{\rf\sim \Dbd}{\unitvec^\top \rf} + \sum_{t=0}^\infty \gamma^t \sum_{\fpi{} \in \Fnd(s)} \int_{\opt[\fpi{}(\gamma)\geq \F(s,\gamma)]} \prn{\prn{\mathbf{T}^\pi}^t\unitvec}^\top \rf \dF[\rf]}\label{eq:vavg-equality-pwr-expand}\\
    &= (1-\gamma)\sum_{t=1}^\infty \gamma^{t-1} \sum_{\fpi{} \in \Fnd(s)} \int_{\opt[\fpi{}(\gamma)\geq \F(s,\gamma)]} \prn{\prn{\mathbf{T}^\pi}^t\unitvec}^\top \rf \dF[\rf].\label{eq:pwr-expand-done}
\end{align}

\Cref{eq:vavg-equality-pwr-expand} holds by \cref{eq:interchange-pwr-expand-2}. \Cref{eq:pwr-expand-done} holds by the fact that the $t=0$ summand equals $\E{\rf\sim \Dbd}{\unitvec^\top \rf}$.
\end{proof}

\begin{restatable}[Sufficient condition for $\pwrNoDist$ being rational on $\gamma$]{prop}{ratlPower}If no reward function has optimal policy shifts and the environment is deterministic, then $\pwr[s,\gamma][\Dbd]$ is a rational function on $\gamma\in(0,1)$.
\end{restatable}
\begin{proof}
Let $\gamma \in (0,1)$.
\begin{align}
    \pwr[s,\gamma][\Dbd] &= (1-\gamma)\sum_{t=1}^\infty \gamma^{t-1} \sum_{\fpi{} \in \Fnd(s)} \int_{\opt[\fpi{}(\gamma)\geq \F(s,\gamma)]} \prn{\prn{\mathbf{T}^\pi}^t\unitvec}^\top \rf \dF[\rf]\label{eq:pwr-id-apply}\\
    &= (1-\gamma) \sum_{\fpi{} \in \Fnd(s)} \sum_{t=1}^\infty \gamma^{t-1} c^\pi_t.\label{eq:ratlPower-const}
\end{align}
\Cref{eq:pwr-id-apply} follows by \cref{thm:pwr-struct}. In \cref{eq:ratlPower-const}, let each $c^\pi_t$ be a constant depending only on $\pi$ and $t$. Since no optimal policy shifts occur, $\forall \fpi{} \in \F(s): \opt[\fpi{}(\gamma)\geq \F(s,\gamma)]$ is constant on $\gamma$. Therefore, the domain of integration is fixed in each inner-summand of \cref{eq:pwr-id-apply}, and so the constants $c^\pi_t$ do not depend on $\gamma$. Furthermore, $\Fnd(s)$ is finite and so we can interchange the summation signs.

Since the environment is deterministic, each $\fpi{}$ has entered a cycle at most $\abs{\St}$ steps into the policy's trajectory from $s$. Therefore, $(c_t^\pi)_{t\geq \abs{\St}}$ is $k$-periodic for some $k$ ($1 \leq k \leq \abs{\St}$). Then for each $\fpi{}$, there exists a $(k-1)$-degree polynomial $P^\pi$ such that
\begin{align}
    \sum_{t=\abs{\St}}^\infty \gamma^{t-1} c^\pi_t = \frac{\gamma^{\abs{\St}-1}}{1-\gamma^k}P^\pi.
\end{align}

Therefore,
\begin{equation}
    \sum_{t=1}^{\infty} \gamma^{t-1} c^\pi_t= \sum_{t=1}^{\abs{\St}-1} \gamma^{t-1} c^\pi_t + \sum_{t=\abs{\St}}^\infty \gamma^{t-1} c^\pi_t
\end{equation}

is also rational on $\gamma$. Since $\pwr$ equals $(1-\gamma)$ times the sum of finitely many rational functions, $\pwr$ is also rational.
\end{proof}

\begin{restatable}[$\textsc{Power}_{\Dbd}(s,\gamma)$ is piecewise rational on $\gamma$]{conjSec}{powPWRat}
\end{restatable}

\subsection{\texorpdfstring{$\pwrNoDist$}{POWER} at its limit points}
\begin{restatable}[$\pwrNoDist$ when $\gamma=0$]{lem}{gammaZeroPower}\label{lem:gamma-0-power}
\begin{equation}
    \pwr[s,0][\Dbd]=\E{\rf \sim \Dbd}{\max_{\dbf \in \childDists} \dbf^\top \rf}=\E{\rf \sim \Dbd}{\max_{\dbf \in \childDistsNd} \dbf^\top \rf}.
\end{equation}
\end{restatable}
\begin{proof}
\begin{align}
    \pwr[s,0][\Dbd]&\defeq \lim_{\gamma\to 0}\E{\rf\sim\Dbd}{\max_{\fpi{s}\in \F(s)} \frac{1-\gamma}{\gamma}\prn{\fpi{s}(\gamma)-\unitvec}^\top \rf} \\
    &=\lim_{\gamma\to 0}\E{\rf\sim\Dbd}{\max_{\fpi{s}\in \F(s)} (1-\gamma)\E{s'\sim T(s,\pi(s))}{\fpi{s'}(\gamma)}^\top \rf} \\
 &=\E{\rf\sim\Dbd}{\max_{\fpi{s}\in \F(s)} \lim_{\gamma\to 0}(1-\gamma)\E{s'\sim T(s,\pi(s))}{\fpi{s'}(\gamma)}^\top \rf}\\
 &=\E{\rf\sim\Dbd}{\max_{\fpi{s}\in \F(s)} \E{s'\sim T(s,\pi(s))}{\unitvec[s']}^\top \rf}\\
 &=\E{\rf \sim \Dbd}{\max_{\dbf \in \childDists} \dbf^\top \rf}\\
 &=\E{\rf \sim \Dbd}{\max_{\dbf \in \childDistsNd} \dbf^\top \rf}.\label{eq:nd-restrict-children}
\end{align}
\Cref{eq:nd-restrict-children} follows because for all $\rf \in \rewardVS$, \cref{cor:nd-func-indif} shows that $\max_{\dbf \in \childDists} \dbf^\top \rf=\max_{\dbf \in \ND{\childDists}} \dbf^\top \rf \eqdef\max_{\dbf \in \childDistsNd} \dbf^\top \rf$.
\end{proof}

\begin{remark}
The non-strict inequality of \cref{thm:min-power} holds for \emph{any} state reward distribution -- $\Dist$ need not be continuous. However, the strict inequality does require continuity in general.
\end{remark}

\begin{restatable}[Minimal $\pwrNoDist_{\Diid}$]{lem}{minPower}\label{thm:min-power} $\pwr[s,\gamma][\Diid]\geq \EX$. If $\gamma\in(0,1)$, equality holds iff $\abs{\F(s)}=1$.
\end{restatable}
\begin{proof}
Let $\gamma\in(0,1)$. If $\abs{\F(s)}=1$, then $\vavg[s,\gamma][\Diid]=\EX\geom$ by the linearity of expectation and the fact that $\Diid$ distributes reward independently and identically across states according to state distribution $\Dist$. By \cref{lem:power-id},
\begin{equation*}
    \pwr[s,\gamma][\Diid]=\frac{1-\gamma}{\gamma}\prn{\vavg[s,\gamma][\Diid]-\EX},
\end{equation*}
and so $\pwr[s,\gamma][\Diid]=\EX$.

Suppose $\abs{\F(s)}>1$, and let $\f\in \F(s)$.
\begin{align}
    \geom[\EX]&=\E{\rf\sim\Diid}{\max_{\f''\in\set{\f}}\f''(\gamma)^\top\rf}\\
    &< \E{\rf\sim\Diid}{\max_{\f''\in\F(s)}\f''(\gamma)^\top\rf}\label{eq:strict-vavg-min}\\
    &\eqdef \vavg[s,\gamma][\Diid].\label{eq:final-min-power}
\end{align}

Let $A\defeq \set{\f(\gamma)},B\defeq \F(s,\gamma)$. Since $\abs{\F(s)}>1$, $\ND{B}=\Fnd(s,\gamma)>2$ by \cref{cor:multi-fnd}. Therefore, $\ND{B}\setminus A$ is non-empty. Since $\Diid$ is continuous \textsc{iid}, $\exists b<c:(b,c)^{\abs{\St}}\subseteq \supp[\Diid]$ by \cref{lem:iid-contains}. Therefore, for $g:\reals\to\reals$ the identity and $\phi$ the identity permutation, \cref{eq:strict-vavg-min} holds by \cref{lem:nd-func-incr-pwr}.  \Cref{eq:final-min-power} implies that $\pwr[s,\gamma][\Diid]>\EX$.

In general, \cref{eq:strict-vavg-min} holds as a non-strict inequality, and so $\pwr[s,\gamma][\Diid]\geq \EX$ for all $\gamma\in(0,1)$. This inequality therefore holds in the limits $\gamma\to 0$ and $\gamma\to 1$.
\end{proof}

\Cref{prop:minExtremes} handles the limiting cases of \cref{thm:min-power} (minimal $\pwrNoDist$).
\begin{restatable}[Minimal $\pwrNoDist$ at $\gamma=0$ and $\gamma=1$]{conjSec}{minExtremes}\label{prop:minExtremes}\hspace*{\fill}
\begin{enumerate}
    \item $\pwr[s,0]=\EX$ iff $\abs{\childDists}=1$.\label{item:min-children}
    \item $\pwr[s,1]=\EX$ iff $\abs{\RSD}=1$.\label{item:min-rsd}
\end{enumerate}
\end{restatable}

\begin{remark}
The non-strict inequality of \cref{max-power} holds for \emph{any} state reward distribution -- $\Dist$ need not be continuous. However, the strict inequality does require continuity in general.
\end{remark}
\begin{restatable}[Maximal $\pwrNoDist_{\Diid}$]{lem}{maxPower}\label{max-power}
$\pwr[s,\gamma][\Diid]\leq\Edraws{\abs{\St}}$. If $\gamma\in(0,1)$, equality holds iff $s$ can deterministically reach all states in one step and all states have deterministic self-loops.
\end{restatable}
\begin{proof}
\begin{align}
    \pwr[s,\gamma][\Diid]&\leq \E{R\sim \D}{\max_{s''\in\St}R(s'')}\label{eq:gen-max-pwr}\\
    &= \Edraws{\abs{\St}}.\label{eq:iid-expected-max}
\end{align}

\Cref{eq:gen-max-pwr} holds by \cref{lem:max-power-general}. \Cref{eq:iid-expected-max} holds because $\Diid$ is {\iid} over states according to state reward distribution $\Dist$.

If $\sureChildren=\St$ and $\forall s' \in \St: s' \in \sureChildren[s']$, then \cref{eq:gen-max-pwr} is an equality by \cref{lem:max-power-general}.

\paragraph*{Backward direction.} Suppose that $\gamma\in(0,1)$. To show that
\begin{equation*}
    \pwr[s,\gamma][\Diid]=\Edraws{\abs{\St}}
\end{equation*}
implies that $\sureChildren=\St$ and $\forall s' \in \St: s' \in \sureChildren[s']$, we show the contrapositive: $\exists s'\in\St:s' \not \in \sureChildren$ or $s' \not \in \sureChildren[s']$ implies that
\begin{equation*}
    \pwr[s,\gamma][\Diid]<\Edraws{\abs{\St}}.
\end{equation*}

Let $\f^{s'}\defeq \unitvec + \geom[\gamma]\unitvec[s']$ be the visit distribution function which would be induced by immediately navigating to $s'$ from $s$ and then staying at $s'$. By assumption, this is not currently possible for some $s'\in \St$, and so $\f^{s'}$ would strictly increase the achievable visit frequency on $s'$, starting from $s$. Formally, $\f^{s'}(\gamma)^\top \unitvec[s']>\max_{\f\in\F(s)}\f(\gamma)^\top \unitvec[s']$.

Let $F^{s'}$ be the set of visit distribution functions at $s$ which would be made available as a result of adding deterministic transitions $s\to s'$ and $s' \to s'$.
\begin{align}
    \pwr[s,\gamma][\Diid]&\defeq \E{\rf\sim\Diid}{\max_{\f\in \F(s)} \frac{1-\gamma}{\gamma}\prn{\f(\gamma)-\unitvec}^\top \rf}\\
    &< \E{\rf\sim\Diid}{\max_{\f\in \F(s)\cup F^{s'}} \frac{1-\gamma}{\gamma}\prn{\f(\gamma)-\unitvec}^\top \rf}\label{eq:union-visit-dist-monotone}\\
    &\leq \Edraws{\abs{\St}}.\label{eq:max-power-apply}
\end{align}

Let $A\defeq \set{\frac{1-\gamma}{\gamma}\prn{\f(\gamma)-\unitvec}\mid \f\in\F(s)},B\defeq \set{\frac{1-\gamma}{\gamma}\prn{\f(\gamma)-\unitvec}\mid \f\in\F(s)\cup F^{s'}}$. Since $\f^{s'}$ maximizes $s'$-visitation frequency, $\f^{s'}(\gamma)^\top \unitvec[s']>\max_{\f\in\prn{\F(s)\cup F^{s'}}\setminus \set{\f^{s'}}}\f(\gamma)^\top \unitvec[s']$, which implies that $\f^{s'}(\gamma)\in \ND{B}\setminus \ND{A}$. Therefore, $\ND{B}\setminus \ND{A}$ is non-empty. Since $\Diid$ is continuous \textsc{iid}, $\exists b<c:(b,c)^{\abs{\St}}\subseteq \supp[\Diid]$ by \cref{lem:iid-contains}. Therefore, for $g:\reals\to\reals$ the identity and $\phi$ the identity permutation, \cref{eq:union-visit-dist-monotone} holds by \cref{lem:nd-func-incr-pwr}. \Cref{eq:max-power-apply} follows because \cref{lem:max-power-general} bounds the $\pwrNoDist_{\Diid}$ of all valid {\mdp} structures.
\end{proof}

\Cref{prop:maxExtremes} handles the limiting cases of \cref{max-power} (maximal $\pwrNoDist$).
\begin{restatable}[Maximal $\pwrNoDist$ at $\gamma=0$ and $\gamma=1$]{prop}{maxExtremes}\label{prop:maxExtremes}\hspace*{\fill}
\begin{enumerate}
    \item $\pwr[s,0][\Diid]=\Edraws{\abs{\St}}$ iff $\sureChildren=\St$.\label{item:max-children}
    \item $\pwr[s,1][\Diid]=\Edraws{\abs{\St}}$ iff $s$ can reach all states with probability $1$ and $\forall s' \in \St:s' \in \sureChildren[s']$.\label{item:max-rsd}
\end{enumerate}
\end{restatable}
\begin{proof}
\Cref{item:max-children}:
\begin{align}
    \pwr[s,0][\Diid]&= \E{\rf \sim \Diid}{\max_{\dbf \in \childDists}\dbf^\top \rf}\label{eq:child-thm-apply}\\
    &\leq \E{\rf \sim \Diid}{\max_{\dbf \in \childDists\cup \set{\unitvec[s'']\mid s'' \in \St}}\dbf^\top \rf}\label{eq:child-thm-strict}\\
    &= \E{\rf \sim \Diid}{\max_{\dbf \in \set{\unitvec[s'']\mid s'' \in \St}}\dbf^\top \rf}\label{eq:max-pwr-app}\\
    &= \Edraws{\abs{\St}}.\label{eq:max-pwr-app-2}
\end{align}
\Cref{eq:child-thm-apply} holds by \cref{lem:gamma-0-power}. Applying \cref{lem:nd-func-incr-pwr} (\cref{lem:iid-contains} guarantees that $\exists b<c: (b,c)^{\abs{\St}}\subseteq \supp[\Diid]$), $A\defeq \childDists$, $B\defeq \childDists\cup\set{\unitvec[s']}$, $g:\reals\to\reals$ the identity function, and $\phi$ the identity permutation, \cref{eq:child-thm-strict} holds by \cref{lem:nd-func-incr-pwr}. \Cref{eq:max-pwr-app} holds because $\ND{\childDists\cup \set{\unitvec[s'']\mid s'' \in \St}}=\set{\unitvec[s'']\mid s'' \in \St}$ by \cref{lem:ndLinSuff}. \Cref{eq:max-pwr-app-2} holds because reward is {\iid} across states under $\Diid$.

Suppose $\exists s' \not \in \sureChildren$. By the definition of $\childDists$ (\cref{def:child-dists}), $\exists s' \in \St: \unitvec[s']\not\in \childDists$. Furthermore, $\unitvec[s']^\top\unitvec[s'] > \max_{\dbf \in \childDists\cup\set{\unitvec[s'']\mid s''\neq s'}} \dbf^\top\unitvec[s']$, and so $\unitvec[s']\in\set{\unitvec[s'']\mid s'' \in \St}=\ND{\childDists\cup \set{\unitvec[s'']\mid s'' \in \St}}$. Therefore, \cref{lem:nd-func-incr-pwr} guarantees that \cref{eq:child-thm-strict} is strict. Therefore, $\pwr[s,0][\Diid]=\Edraws{\abs{\St}}$ implies that $\sureChildren=\St$.

Suppose $\sureChildren=\St$. Then \cref{lem:nd-func-incr-pwr} guarantees that \cref{eq:child-thm-strict} is an equality, and so $\sureChildren=\St$ implies that $\pwr[s,0][\Diid]=\Edraws{\abs{\St}}$.

\Cref{item:max-rsd} follows by applying the above reasoning with $\RSD$ in place of $\childDists$, substituting in the fact that $\unitvec[s']\in\childDists$ iff $s$ can reach $s'$ with probability $1$ and $s'\in\sureChildren[s']$.
\end{proof}

\begin{restatable}[Lower bound on current $\pwrNoDist$ based on next-step expected reward]{conjSec}{FutureReward}\label{lem:future-reward}
\begin{align*}
\pwr[s,\gamma]\geq (1-\gamma)\max_a \E{\substack{s'\sim T(s,a),\\R\sim\Dbd}}{R(s')} + \gamma\min_a \E{s'\sim T(s,a)}{\pwr[s',\gamma]}.
\end{align*}
\end{restatable}

\begin{restatable}[When $\gamma=0$, having similar children implies equal $\pwr$]{lem}{tndSim}\label{lem:tnd-sim}
Suppose $D \subseteq \childDistsNd$ is similar to $\childDistsNd[s']$. Then $\pwr[s,0][\Diid]\geq \pwr[s',0][\Diid]$. If $D\subsetneq \childDistsNd$, then $\exists \gamma^*: \forall \gamma \in [0, \gamma^*): \pwr[s,\gamma][\Diid] > \pwr[s',\gamma][\Diid]$.
\end{restatable}
\begin{proof}
Suppose $\childDistsNd[s']=\phi\cdot D$, where $\phi$ .
\begin{align}
    \pwr[s,0][\Diid] &= \E{\rf \sim \Diid}{\max_{\dbf \in \childDistsNd} \dbf^\top \rf}\label{eq:gamma-zero}\\
    &\geq \E{\rf \sim \Diid}{\max_{\dbf \in D} \dbf^\top \rf}\label{c-eq:child-restriction}\\
    &= \E{\rf \sim \Diid}{\max_{\dbf \in D} (\permute\dbf)^\top \rf}\label{c-eq:child-dist-permute}\\
    &= \pwr[s',0][\Diid].
\end{align}

\Cref{eq:gamma-zero} follows by \cref{lem:gamma-0-power}. \Cref{c-eq:child-dist-permute} follows because reward is assumed to be \textsc{iid} over states. If $D\subsetneq \childDistsNd$, non-domination ensures that the child distributions of $\childDistsNd\setminus D$ are strictly greedily optimal for a positive measure set of reward functions. In this case, \cref{c-eq:child-restriction}'s inequality is strict; strict inequality holds for $\gamma \approx 0$ by the continuity of $\pwr[s',\gamma][\Diid]$ on $\gamma$ (\cref{thm:cont-power}).
\end{proof}

\begin{restatable}[$\pwrNoDist$ achieves ordinal equilibrium as $\gamma\to 1$]{prop}{powAgree}\label{pow-agree}
Let $s \succeq^\gamma_{\pwrNoDist_{\Dbd}} s'$ when $\pwr[s,\gamma][\Dbd]\geq \pwr[s',\gamma][\Dbd]$. $\exists \gamma^*: \forall \gamma \in (\gamma^*, 1):\; \succeq^\gamma_{\pwrNoDist_{\Dbd}}$ is constant.
\end{restatable}
\begin{proof}
Consider two states $s$ and $s'$. By the Lipschitz continuity of $\pwrNoDist$ on $\gamma$ (\cref{thm:cont-power}) and the fact that the domain is bounded $\gamma\in [0,1]$,
\begin{equation*}
    \sgn\left(\pwr[s,\gamma][\Dbd]-\pwr[s',\gamma][\Dbd]\right)
\end{equation*}
changes value finitely many times on $\gamma\in [0,1]$. There are only finitely many such pairs of states, and so the result follows.
\end{proof}

\begin{restatable}[States with different $\pwr$ functions are equal for finitely many $\gamma$]{conjSec}{finitePower}
\end{restatable}

\subsection{States whose \texorpdfstring{$\pwrNoDist$}{POWER} can be immediately determined}\label{sec:time-unif}

\begin{restatable}[Delay linearizes $\pwrNoDist$]{cor}{delay}
\label{delay}
Let $s_0,\ldots,s_\ell$ be such that for all $i < \ell$, $Ch(s_i)=\{s_{i+1}\}$. Then $\pwr[s_0,\gamma][\Diid]=\sum_{i=1}^\ell \gamma^{i-1} \E{R\sim \D'}{R(s_i)}+\gamma^{\ell} \pwr[s_{\ell},\gamma][\Diid]$.
\end{restatable}
\begin{proof}
Iteratively apply \cref{lem:future-power} $\ell$ times. Equality must hold, as each $s_{i}$ can only reach $s_{i+1}$.
\end{proof}
\begin{remark}
For $\Diid$, the $\sum_{i=1}^\ell \gamma^{i-1} \E{R\sim \Diid}{R(s_i)}$ term in \cref{delay} simplifies to $(1-\gamma^\ell)\EX$.
\end{remark}

Deterministic delay is a special case of a more general principle.

\begin{restatable}[State reachability by time $t$]{definition}{DefReachT}
$\Call{Reach}{s,t}$ is the set of states which $s$ can reach  in exactly $t$ steps with positive probability.
\end{restatable}

\begin{restatable}[Time-uniform states]{definition}{DefTimeUnif}\label{def:time-unif}
In deterministic environments, a state $s$ is \emph{time-uniform} when $\forall t>0, s',s''\in \Call{Reach}{s,t}:$ $s'$ and $s''$ can either reach the same states in one step, or they can only reach themselves.
\end{restatable}

\begin{restatable}[If $s'$ is reachable from a time-uniform state $s$, then $s'$ is time-uniform]{lem}{reachTimeUnif}
\end{restatable}
\begin{proof}
This follows directly from \cref{def:time-unif}.
\end{proof}

\begin{figure}[ht]
    \centering
    \subfloat[][]{
        \begin{tikzpicture}[mygraph,node distance=1.3cm]
            %nodes
                \node[blue] (1) {$1$};
                \node[black] (2) [below right of=1] {};
                \node[black] (3) [above right of=1] {};
                \node[red] (4) [right of=1,node distance=2.215cm] {};
                \node[red] (5) [right of=2] {};
                \node[red] (6) [right of=3] {};
                \node[green] (7) [right of=4] {};
            %arrows
                \path[->] (1) edge (2);
                \path[->] (1) edge (3);
                \path[->] (2) edge (4);
                \path[->] (2) edge (5);
                \path[->] (2) edge (6);
                \path[->] (3) edge (4);
                \path[->] (3) edge (5);
                \path[->] (3) edge (6);
                \path[->] (4) edge (7);
                \path[->] (5) edge (7);
                \path[->] (6) edge (7);
                \path[->] (7) edge [looseness=7, out=-45, in=45] (7);
    \end{tikzpicture}
    \label{time-uniform}}
    \qquad
     \subfloat[][]{
        \begin{tikzpicture}[mygraph,node distance=1.3cm]
            %nodes
                \node[blue] (1) {$2$};
                \node[black] (2) [right of=1] {};
                \node[red] (3) [right of=2, node distance=.95cm] {};
                \node[red] (4) [above right of=2] {};
                \node[red] (5) [below right of=2] {};
                \node[green] (6) [above right of=3] {};
                \node[green] (7) [below right of=3] {};
                \node[purple] (8) [right of=3, node distance = 1.95cm] {};
            %arrows
                \path[->] (1) edge (2);
                \path[->] (2) edge (3);
                \path[->] (2) edge (4);
                \path[->] (2) edge (5);
                \path[->] (3) edge (6);
                \path[->] (3) edge (7);
                \path[->] (4) edge (6);
                \path[->] (4) edge (7);
                \path[->] (5) edge (6);
                \path[->] (5) edge (7);
                \path[->] (6) edge (8);
                \path[->] (7) edge (8);
                \path[->] (8) edge [out=90, in=90] (1);
        \end{tikzpicture}

     \label{power-demo}}
    \caption[Time uniform initial states]{States of the same color have the same children. For $\Dist$ uniform: $\pwr[\col{blue}{s_1},\gamma][\Diid]=(1-\gamma)(\frac{2}{3} + \frac{3}{4}\gamma) + \half \gamma^2$ and $\pwr[\col{blue}{s_2},\gamma][\Diid]=\frac{1-\gamma}{1-\gamma^5}\prn{\half + \frac{3}{4}\gamma + \frac{2}{3} \gamma^2 + \half (\gamma^3+\gamma^4)}$ by \cref{time-unif-power}.}
    \label{fig:uniform}
\end{figure}

\begin{restatable}[Time-uniform $\pwrNoDist$ bound]{prop}{timePower}\label{time-unif-power}
If $s$ is time-uniform, then either all trajectories simultaneously enter 1-cycles or no trajectory ever enters a 1-cycle. Furthermore,
\begin{align*}
    \pwr[s,\gamma][\Diid] & =\unifPower \\
        & \defeq(1-\gamma)\sum_{t=1}^\infty \gamma^{t-1} \Edraws{\abs{\Call{Reach}{s,t}}}.
\end{align*}
\end{restatable}

\begin{proof}
Suppose $s$ is time-uniform. The fact that all trajectories simultaneously enter 1-cycles or no trajectory ever enters a 1-cycle, follows directly from \cref{def:time-unif}.

Starting from $s$, suppose all policies enter a 1-cycle at timestep $k$ (if no policy enters a 1-cycle, $k=\infty$). Consider any reward function $R$. The agent starts at timestep $0$. At timestep $t<k$, its optimal policy selects greedily from $\abs{\Call{Reach}{s,t+1}}$ choices. At $t\geq k$, the agent is in the best of $\abs{\Call{Reach}{s,t}}$ 1-cycles.
\end{proof}

\begin{restatable}[$\pwrNoDist_{\Diid}$ bounds]{prop}{bounds}\label{bounds}
\begin{equation*}
    0<\EX \leq \pwr[s,\gamma][\Diid]\leq \unifPower \leq \Edraws{\abs{\St}}<1.
\end{equation*}
\end{restatable}

\begin{proof}
$\Edraws{\abs{\St}}<1$ because $\Dist$ is a continuous distribution on the unit interval; similarly for $0<\EX$. $\pwr[s,\gamma][\Diid]\leq \unifPower$ because for each reward function and at each time step $t$, the agent can (at best) enter the highest-reward state from $\Call{Reach}{s,t}$. The other inequalities follow directly from \cref{thm:min-power} and \cref{max-power}.
\end{proof}

\begin{restatable}[Time-uniformity implies no optimal policy shifts]{cor}{TimeNo}
The optimal trajectory cannot shift at time-uniform states.
\end{restatable}
\begin{proof}
Apply \cref{opt-policy-shift}.
\end{proof}

\subsection{Recursive \texorpdfstring{$\pwrNoDist$}{POWER} computation}
In general, $\vavg$ can be computed by solving for $\abs{\Fnd(s)}$ convex polytopes (the optimality supports, \cref{def:opt-supp}) and integrating the average on-policy value:
\begin{equation}
    \vavg=\sum_{\f \in \Fnd(s)} \int_{\opt} \f(\gamma)^\top\rf \dF.
\end{equation}

However, the optimality supports and the corresponding integrals may be difficult to compute. In certain ``tree-like'' environments, we can compute $\pwr$ via dynamic programming.

\begin{restatable}[Support of a set of visit distributions]{definition}{suppVisit}\label{def:supp-visit}
Let $F \subseteq \F(s)$. $\supp[F]\defeq \set{s' \mid \exists \f \in F: (\f(.5)-\unitvec)^\top \unitvec[s']>0}$ is the set of states $s'$ visited with positive probability by some $\f \in F$.
\end{restatable}

\begin{restatable}[Reward independence]{definition}{rewardInd}\label{def:reward-ind}
Let $F, F' \subseteq \F(s)$. $F \independent_{\Dany} F'$ (read ``\emph{$F$ and $F'$ are reward independent under $\Dany$}'') when $\forall s \in \supp[F], s' \in \supp[F']$, the reward at $s$ is independent of the reward at $s'$ under $\Dany$.
\end{restatable}

\begin{figure}[h!]
    \centering
     \begin{tikzpicture}[mygraph,node distance=1.4cm]
        %nodes
            \node[blue] (0) {0};
            \node[black] (1) [above right of=0] {1};
            \node[black] (2) [right of=0] {2};
            \node[black] (3) [below right of=0] {3};
        %arrows
            \path[->] (0) edge (1);
            \path[->] (0) edge (2);
            \path[->] (0) edge (3);
            \path[->] (1) edge [looseness=7, out=-45, in=45] (1);
            \path[->] (2) edge [looseness=7, out=-45, in=45] (2);
            \path[->] (3) edge [looseness=7, out=-45, in=45] (3);
    \end{tikzpicture}
    \caption[Factorizing a $\pwrNoDist$ integration]{Let distribution $\Dist_a$ have \textsc{cdf} $F_a(v)\defeq v$ on $v \in [0,1]$ and distribution $\Dist_b$ have \textsc{cdf} $F_b(v)\defeq v^2$ on $v \in [0,1]$. Suppose the reward function distribution $\D'$ is such that $R(s_1), R(s_3) \sim \Dist_a, R(s_2) \sim \Dist_b$. Then \cref{thm:pow-compute} shows that $\pwrNoDist_{\D'}(\col{blue}{s_0},\gamma)=\int_{0}^{1} v \frac{\dif }{\dif{v}}(v\cdot v^2 \cdot v) = \int_0^1 4v^4 \dif v = \frac{4}{5}$.}
    \label{fig:power-calc}
\end{figure}

\begin{restatable}[Factored $\pwrNoDist$ computation]{thm}{vavgCompute}\label{thm:pow-compute}
Let $\D'$ be a probability distribution over reward functions, with probability measure $F'$. Let $F^1, \ldots, F^k \subseteq \Fnd(s)$ be pairwise reward independent under $\D'$, where $\bigcup_i F^i = \Fnd(s)$. For each $F^i$ and any $\gamma^* \in (0,1)$, define the random variable $X_{\gamma^*}^i\defeq \max_{\f \in F^i} \frac{1}{\gamma^{*}}(\f(\gamma^*)-\unitvec)^\top \rf \mid \rf \sim \D'$ with \textsc{cdf} $F_{X_{\gamma^*}^i}$. Then for any $\gamma \in [0,1]$,
\begin{equation}
    \pwrNoDist_{\D'}(s,\gamma)=\lim_{\gamma^*\to \gamma}(1-\gamma^*)\int_{-\infty}^{\infty} v \dif \prn{ \prod_i F_{X_{\gamma^*}^i}(v)}.
\end{equation}
\end{restatable}
\begin{proof}
Define the random variable $X_\gamma^* \defeq \max_i X_\gamma^i$ with \textsc{cdf} $F_{X^*_\gamma}$. Consider $X_\gamma^j, X_\gamma^k$, where $j \neq k$. We have
\begin{align}
    X_\gamma^j&\defeq \max_{\f \in F^j} \gamma^{-1}(\f(\gamma)-\unitvec)^\top \rf \mid \rf \sim \D'\\
    X_\gamma^k&\defeq \max_{\f \in F^k} \gamma^{-1}(\f(\gamma)-\unitvec)^\top \rf \mid \rf \sim \D'.
\end{align}

For each $s_j \in \supp[F^j], s_k \in \supp[F^k]$, the random variable $R_{s_j} \mid R \sim \D'$ is independent of the random variable $R_{s_k} \mid R \sim \D'$ because we assumed $F^j \independent_{\D'} F^k$ (\cref{def:reward-ind}). Then any random variable corresponding to a linear combination $L_j \defeq \sum_{s_j \in \supp[F^j]} \alpha_j R(s_j)$ (where $\alpha_j \in \reals$) is also independent of each random variable $R_{s_k} \mid R \sim \D'$. Furthermore, any such $L_j$ is independent of any linear combination $L_k \defeq \sum_{s_k \in \supp[F^k]} \alpha_k R(s_k)$, where $\alpha_k \in \reals$.

But each $\gamma^{-1}(\f_j(\gamma)-\unitvec)^\top \rf$ is precisely such a linear combination. So $X_\gamma^j=\max_n L_n, X_\gamma^k = \max_m L_m$, where each $L_n\defeq \gamma^{-1}(\f_j(\gamma)-\unitvec)^\top \rf$ is independent of each $L_m\defeq \gamma^{-1}(\f_k(\gamma)-\unitvec)^\top \rf$. So $X_\gamma^j$ and $X_\gamma^k$ are independent under $\D'$. Therefore, $\forall v \in \reals: F_{X^*_\gamma}(v)=\prod_i F_{X_\gamma^i}(v)$.

\begin{align}
    V^*_{\D'}(s,\gamma)&\defeq \int_{\supp[\D']} \max_{\f \in \Fnd(s)} \f(\gamma)^\top \rf \, \dif F'(\rf)\label{eq:factor-fnd-vavg}\\
    &= \E{R\sim \D}{R(s)} + \gamma \int_{\supp[\D']} \max_{\f \in \Fnd(s)} \gamma^{-1}(\f(\gamma)-\unitvec)^\top \rf \, \dif F'(\rf)\\
    &= \E{R\sim \D}{R(s)} + \gamma \int_{-\infty}^\infty v \dif F_{X^*_\gamma}(v)\label{eq:factor-value-transform}\\
    &=\E{R\sim \D}{R(s)} + \gamma \int_{-\infty}^{\infty} v \dif \prn{ \prod_i F_{X_{\gamma}^i}(v)}.\label{eq:factor-vavg-complete}
\end{align}
\Cref{eq:factor-fnd-vavg} follows from the definition of average optimal reward (\cref{def:vavg}) and the fact that restricting maximization to $\Fnd$ leaves optimal value unchanged for all reward functions by \cref{def:nd}. \Cref{eq:factor-value-transform} follows from the definition of $F_{X^*_\gamma}$. \Cref{eq:factor-vavg-complete} follows from the factorization $F_{X^*_\gamma}=\prod_i F_{X^i_\gamma}$, proved above.

Then since $\gamma \in (0,1)$,
\begin{align}
    \pwrNoDist_{\D'}(s,\gamma)&= \frac{1-\gamma}{\gamma}\prn{V^*_{\D'}(s,\gamma) - \E{R\sim \D}{R(s)}}\label{eq:factor-power-id}\\
    &=(1-\gamma)\int_{-\infty}^{\infty} v \dif \prn{ \prod_i F_{X_{\gamma}^i}(v)}.\label{eq:factor-power-plug}
\end{align}
\Cref{eq:factor-power-id} follows from \cref{def:power}. \Cref{eq:factor-power-plug} follows from \cref{eq:factor-vavg-complete}.

Since $\pwrNoDist_{\D'}$ is Lipschitz continuous on $\gamma \in [0,1]$ (\cref{thm:cont-power}) and since \cref{eq:factor-power-plug} holds for all $\gamma \in (0,1)$, the result holds in the limits and therefore holds for all $\gamma \in [0,1]$.
\end{proof}

\subsection{Complexity of estimating \texorpdfstring{$\pwrNoDist$}{power} and optimality probability}
\begin{restatable}[$\pwrNoDist$ sampling bounds]{prop}{samplingPower}\label{prop:sampling-power}
Let $\gamma \in [0,1]$, $\D'$ be a reward function distribution which is bounded $[b,c]^{\abs{\St}}$, $s$ be a state, and $\epsilon >0$. For the random variable $\bar{X}_n^{\pwr[][\D']}\defeq \frac{1}{n}\sum_{i=1}^n \E{s'\sim T(s,\pi^*_i(s))}{\VfNorm[*][R_i]{s',\gamma}}$ for \textsc{iid} draws $R_1,\ldots, R_n \sim \D'$,
\begin{equation}
    \prob[R_1,\ldots, R_n \sim \D']{\abs{\bar{X}_n^{\pwr[][\D']}-\pwr[s,\gamma][\D']}\geq \epsilon}\leq 2e^{-\frac{2n \epsilon^2}{(c-b)^2}}.
\end{equation}
\end{restatable}
\begin{proof}
Normalized optimal value for each $R_i$ is bounded $[b,c]$. Since the draws $R_i \sim \D'$ are independent, apply Hoeffding's inequality.
\end{proof}

At any fixed discount rate $\gamma\in[0,1)$, an optimal value function can be computed in time polynomial in $\abs{\St}$ and $\abs{\A}$ (via \eg{} value iteration \citep{littmanComplexity}). Therefore, \cref{prop:sampling-power} shows that when $\gamma\in[0,1)$, $\pwr$ can be efficiently approximated with high probability.

\begin{restatable}[$\pwrNoDist$ can be efficiently computed]{conjSec}{pwrPoly}
\end{restatable}

We show a similar result for optimality probability, except \cref{prop:sampling-prob} does not require a bounded reward function distribution.

\begin{restatable}[Optimality probability sampling bounds]{prop}{samplingProb}\label{prop:sampling-prob}
Let $\gamma \in [0,1]$, $F\subseteq \F(s)$, and $\epsilon >0$. For the random variable $\bar{X}_n^{\Prb_{\Dany}}\defeq \frac{1}{n}\sum_{i=1}^n \indic{\exists \f \in F: \text{ $\f$ is optimal for $R_i$ at $\gamma$}}$ for \textsc{iid} draws $R_1,\ldots, R_n \sim \Dany$,
\begin{equation}
    \prob[R_1,\ldots, R_n \sim \Dany]{\abs{\bar{X}_n^{\Prb_{\Dany}}-\optprob[\Dany]{F,\gamma}}\geq \epsilon}\leq 2e^{-2n \epsilon^2}.
\end{equation}
\end{restatable}
\begin{proof}
$\bar{X}_n^{\Prb_{\Dany}}$ is an unbiased estimator of $\optprob[\Dany]{F,\gamma}$ and it is bounded $[0,1]$. Since the draws $R_i \sim \Dany$ are independent, apply Hoeffding's inequality.
\end{proof}

\begin{restatable}[Optimality probability can be efficiently computed]{conjSec}{optPoly}
\end{restatable}

\subsection{How \texorpdfstring{$\pwrNoDist$}{POWER} relates to empowerment}
\begin{restatable}[$n$-step reachable states]{definition}{nReach}\label{def:n-reach}
Let $\reach[\text{sure}]{s,t}$ be the set of states which are reachable from state $s$ with probability $1$ in exactly $t$ time steps.
\end{restatable}

\begin{restatable}[A function of the number of reachable states lower-bounds $\pwrNoDist$]{conjSec}{reachPower}\label{lem:reach-power}
Let $S_t\subseteq \reach[\text{sure}]{s,t}$ be those states only reachable in exactly $t$ steps. For any $\gamma \in [0,1]$,
\begin{equation}
    \pwr[s,\gamma]\geq \prn{1-(1-\gamma)\gamma^{t-1}}\EX + (1-\gamma)\gamma^{t-1}\Edraws{\abs{S_t}}.
\end{equation}
\end{restatable}

\begin{figure}[ht]
    \centering
    \begin{tikzpicture}[mygraph,auto,node distance=1.3cm]
        %nodes
            \node[blue] (1) {$1$};
            \node[black] (2) [below right of=1] {};
            \node[black] (3) [above right of=1] {};
        %arrows
            \path[->] (1) edge (2);
            \path[->] (1) edge (3);
            \path[<->] (2) edge[shorten > = 0pt] (3);
    \end{tikzpicture}
    \caption[On the summands of an {\iid} $\pwrNoDist$ infinite series]{ The $\pwrNoDist_{\Diid}$ expansion of \cref{thm:pwr-struct} can have summands less than $\EX$.}
    \label{fig:no-ineq-in-general}
\end{figure}

The caption of \cref{fig:no-ineq-in-general} is justified because 
\[\pwr[\col{blue}{s_1},\gamma][\Diid]=\frac{1}{1+\gamma}\prn{\Edraws{2}+\gamma\E{}{\min \text{ of } 2 \text{ draws from } \Dist}}.\]

Although \cref{fig:empower_fail} demonstrates how information-theoretic empowerment fails to capture important non-local information about the agent's control over the environment, $\pwr$ and empowerment are not unrelated. As \citet{salge_empowermentintroduction_2014} remark, ``In a  discrete deterministic world empowerment reduces to the logarithm of the number of sensor states reachable with the available actions.'' \Cref{thm:emp-lower-bound} reflects the fact that an agent can at least choose from the highest-reward reachable state after $t$ steps.

\begin{restatable}[A function of empowerment lower-bounds $\pwr$]{conjSec}{infoLBPower}\label{thm:emp-lower-bound}
If the environment is deterministic and contains an absorbing state, then for any $\gamma \in [0,1]$,
\begin{multline}
    \pwr[s,\gamma]\geq\\ \sup_{t\geq 1} \prn{1-(1-\gamma)\gamma^{t-1}}\EX + (1-\gamma)\gamma^{t-1}\Edraws{2^{\mathfrak{E}_t(s)}}.
\end{multline}
\end{restatable}
\begin{proof}
Suppose the environment is deterministic and let $\gamma \in [0,1]$. Let $t$ be a positive integer. By Equation 4.15 of \citet{salge_empowermentintroduction_2014},  $\mathfrak{E}_t(s)=\log_2 \abs{\reach[\text{sure}]{s,t}}$ (given deterministic dynamics). Since the environment contains an absorbing state, apply \cref{lem:reach-power} to conclude that
\begin{equation*}
    \pwr[s,\gamma]\geq \prn{1-(1-\gamma)\gamma^{t-1}}\EX + (1-\gamma)\gamma^{t-1}\Edraws{2^{\mathfrak{E}_t(s)}}.
\end{equation*}

Since $t$ was arbitrary, $\pwr[s,\gamma]$ must be greater than the supremum over such $t$.
\end{proof}

\Cref{fig:tight-emp-bound} shows that this inequality would be tight.

\begin{figure}[ht]
    \centering
    \begin{tikzpicture}[mygraph,node distance=1.3cm]
        %nodes
            \node[blue] (1) {$1$};
            \node[black] (2) [below right of=1] {};
            \node[black] (3) [above right of=1] {};
            \node[black] (end) [right of=1, node distance=1.9cm] {};
        %arrows
            \path[->] (1) edge (2);
            \path[->] (1) edge (3);
            \path[->] (2) edge (end);
            \path[->] (3) edge (end);
            \path[->] (end) edge [looseness=7, out=-45, in=45] (end);
    \end{tikzpicture}
    \caption[Tightness of a conjectured inequality relating $\pwrNoDist$ and empowerment]{ $\pwr[\col{blue}{s_1},\gamma]=(1-\gamma)\Edraws{2}+\gamma\EX$ by \cref{time-unif-power}. Since $\mathfrak{E}_1(s)=1$, the inequality of \cref{thm:emp-lower-bound} would be tight.}
    \label{fig:tight-emp-bound}
\end{figure}

\section{\texorpdfstring{$\pwrNoDist$}{POWER}-seeking}
\begin{figure}
    \centering
    \begin{tikzpicture}[mygraph]
        %nodes
            \node[blue] (1) {$1$};
            \node[black] (2) [above right of=1] {$2$};
            \node[black] (3) [right of=1] {$3$};
            \node[black] (4) [right of=3] {};
            \node[black] (5) [above right of=4] {};
            \node[black] (6) [right of=4] {};
        %arrows
            \path[->] (1) edge node[midway,left] {\texttt{up}} (2) ;
            \path[->] (1) edge node[midway,below] {\texttt{right}} (3);
            \path[->] (3) edge (4);
            \path[->] (4) edge (5);
            \path[->] (4) edge (6);
            \path[->] (2) edge [looseness=6, out=45, in=135] (2);
            \path[->] (6) edge [looseness=6, out=0, in=90] (6);
            \path[->] (5) edge [looseness=6, out=45, in=135] (5);
     \end{tikzpicture}
     \caption[$\pwrNoDist$-seeking actions are not always more probably optimal]{Policies which go \texttt{right} are $\pwr[][\Diid]$-seeking: $\forall \gamma \in (0,1]: \pwr[s_3,\gamma][\Diid]>\pwr[s_2,\gamma][\Diid]$ by \cref{thm:min-power} and \cref{RSDSimPower}. However, for $\Dist'$ with \textsc{cdf} $F(x)=x^2$ on the unit interval, $\optprob[\Diid[\Dist']]{\col{blue}{s_1}, \texttt{up}, .12}\approx .91$. For $\Diid[\Dist']$ and at $\gamma=0.12$, it is more probable that optimal trajectories go \texttt{up} through $s_2$, which has \emph{less} $\pwrNoDist_{\Diid[\Dist']}$.}
    \label{fig:counterex-powerseeking}
\end{figure}

\begin{restatable}[Probability of $\pwrNoDist$-seeking being incentivized]{questionSec}{probPowerSeek}
Seeking $\pwrNoDist$ is not always more probable under optimality, but we have shown sufficient conditions for when it is. We believe that this relationship often holds, but it is impossible to graphically characterize when it holds  (\cref{prop:power-not-ic}). For some suitable high-entropy joint distribution over {\mdp} structures (\eg{} Erdős–Rényi), state reward distributions $\Dist$, starting states $s$, and future states $s'$, with what probability is seeking $\pwr$ at $s'$ more probable under optimality, given that the agent starts at $s$?
\end{restatable}

\subsection{Ordering policies based on \texorpdfstring{$\pwrNoDist$}{POWER}-seeking}\label{app:order}

$\pwr$-seeking is not a binary property: it's not true that a policy either does or doesn't seek $\pwr$. The $\pwr$-seeking definition (\cref{def:pow-seek}) accounts for the fact that $\pi$ might seek a lot of $\pwr$ at $s$ but not seek much $\pwr$ at $s'$ (\cref{app:order}) and that a policy $\pi$ may seek $\pwr$ for one discount rate but not at another (\cref{sec:nuances-discount}).

The $\pwr$-seeking definition (\cref{def:pow-seek}) implies a total ordering over actions based on how much $\pwr$ they seek at a fixed state $s$ and discount $\gamma$.

\begin{restatable}[$\pwrStateOrder$]{definition}{singlePowerSeekOrdering}\label{def:single-power-seek-ordering}
$a \pwrStateOrder a'$ when
\begin{equation}
    \E{s' \sim T(s,a)} {\pwr[s',\gamma]} \geq \E{s' \sim T(s,a')} {\pwr[s',\gamma]}.
\end{equation}

Action $a$ \emph{maximally/minimally seeks $\pwr$ at $s$ and $\gamma$} when it is a maximal/minimal element of $\pwrStateOrder$.
\end{restatable}

\Cref{fig:nuances-discount} illustrated how $\pwr$-seeking depends on $\gamma$. \Cref{fig:nuances-ordering} shows how a policy might maximally seek $\pwr$ at $s$ but then minimally seek $\pwr$ at $s'$; therefore, many policy pairs aren't comparable in their $\pwr$-seeking.

\begin{figure}[ht]
    \centering
    \begin{tikzpicture}[mygraph]
        % Upper subgraph
            \node[blue] (1) {$1$};
            \node[black] (2) [right of=1] {$2$};
            \path[->] (1) edge node[midway,above] {\texttt{right}} (2) ;
            \node[black] (4) [below of=2] {$4$};
            \path[->] (2) edge (4);
            \path[->] (4) edge [looseness=6, out=-45, in=45] (4);
            \node[black] (5) [right of=2] {$5$};
            \path[->] (2) edge (5);
            \node[black] (c) [right of=5] {};
            \path[->] (5) edge (c);
            \path[->] (c) edge [looseness=6, out=0, in=90] (c);
            \node[black] (d) [below of=5] { };
            \path[->] (5) edge (d);
            \path[->] (d) edge [looseness=6, out=-45, in=45] (d);
        % Lower subgraph
            \node[black] (3) [below of=1] {$3$};
            \path[->] (1) edge node[midway,left] {\texttt{down}} (3);
            \path[->] (3) edge [looseness=6, out=-45, in=45] (3);
     \end{tikzpicture}

    \caption[A policy which only seeks $\pwrNoDist$ at one state]{If $\pi(\col{blue}{s_1})=\texttt{right}, \pi(s_2)=\texttt{down}$, then $\forall \gamma \in [0,1]$, $\pi$ maximally seeks $\pwr$ at $\col{blue}{s_1}$ but minimally seeks $\pwr$ at $s_2$. Just as a consumer earns money in order to spend it, a policy may gain $\pwr$ in order to ``spend it'' to realize a particular trajectory.\label{fig:nuances-ordering}}
\end{figure}

Ultimately, we're interested in the specific situations in which a policy seeks ``a lot'' of $\pwr$, not whether the policy seeks $\pwr$ ``in general.'' Even so, we can still formalize a good portion of the latter concept. \Cref{def:all-power-seek-ordering} formalizes the natural $\pwrNoDist$-seeking preorder over the policy space $\Pi$.

\begin{restatable}[$\pwrPreorder$]{definition}{allPowerSeekOrdering}\label{def:all-power-seek-ordering}
$\pi \pwrPreorder \pi'$ when $\forall s\in\St: \pi(s) \pwrStateOrder \pi'(s)$.
\end{restatable}

\begin{restatable}[$\pwrPreorder$ is a preorder on $\Pi$]{prop}{preOrderPower}\label{prop:preorder-power}
\end{restatable}
\begin{proof}
$\pwrPreorder$ is reflexive and transitive because of the reflexivity and transitivity of the total ordering $\pwrStateOrder$.
\end{proof}

\begin{restatable}[Existence of a maximally $\pwr$-seeking policy]{prop}{existMaxPower}\label{prop:exist-max-power}
Let $\gamma\in[0,1]$. $\pwrPreorder$ has a greatest element.
\end{restatable}
\begin{proof}
Construct a policy $\pi$ such that $\forall s: \pi(s) \in \argmax_a \E{s' \sim T(s, a)}{\pwr[s',\gamma]}$. This is well-defined because $\A$ is finite.
\end{proof}

\subsection{Seeking \texorpdfstring{$\pwrNoDist$}{POWER} at different discount rates}\label{sec:nuances-discount}
\Cref{fig:nuances-discount} shows that at any given state, the extent to which an action seeks $\pwr$ depends on the discount rate. Greedier optimal policies might tend to accumulate short-term $\pwr$ (\ie{} $\pwr[s,\gamma]$ for $\gamma \approx 0$), while Blackwell optimal policies might tend to accumulate long-term $\pwr$ (\ie{} $\pwr[s,\gamma]$ for $\gamma\approx 1$).

\begin{restatable}[$\pwr$ bounds when $\gamma=0$]{lem}{ZeroBounds}\label{lem:zero-bounds} Let $\Dist'$ be any bounded distribution over $\reals$.
\begin{align}
    &\Edraws[\Dist']{\abs{\sureChildren}}\\
    \leq &\, \pwr[s,0][\Diid[\Dist']]\\
    \leq &\Edraws[\Dist']{\abs{\children}}.
\end{align}
\end{restatable}
\begin{proof}
The left inequality holds because restricting policies to deterministic action at $s$ cannot increase $\pwr[s,0][\Diid[\Dist']]$. The right inequality holds because at best, greedy policies deterministically navigate to the child with maximal reward.
\end{proof}

\begin{restatable}[Children]{definition}{Children}
The \emph{children} of state $s$ are $Ch(s)\defeq \set{s' \mid \exists a: T(s,a,s')>0}$.
\end{restatable}

\begin{restatable}[When $\gamma=0$ under local determinism, maximally $\pwrNoDist_{\Diid}$-seeking actions lead to states with the most children]{prop}{mostChoice}\label{thm:most-children-power}
Let $\Dist'$ be a nondegenerate distribution on $\reals$. Suppose all actions have deterministic consequences at $s$ and its children. For each action $a$, let $s_a$ be such that $T(s, a, s_a)=1$. $\pwr[s_a,0][\Diid[\Dist']] =\max_{a'\in\A} \pwr[s_{a'},0][\Diid[\Dist']]$ iff $\abs{\children[s_a]}=\max_{a'\in\A}\abs{\children[s_{a'}]}$.
\end{restatable}
\begin{proof}
Apply the  bounds of \cref{lem:zero-bounds}; by the assumed determinism, $\children[s_a]=\sureChildren[s_a]$ and so $\pwr[s_a,0][\Diid[\Dist']] = \Edraws{\abs{\children[s_a]}}$ (similarly for each $s_{a'}$). $\Edraws{\abs{\children[s_a]}}$ is strictly monotonically increasing in $\abs{\children[s_a]} $ by the non-degeneracy of $\Dist$.
\end{proof}

\Cref{fig:nuances-discount} illustrates \cref{thm:most-children-power} and \cref{thm:stay-put-power}.

\begin{figure}[ht]\centering
    \begin{tikzpicture}[mygraph]
        % Upper subgraph
            \node[blue] (1) {$1$};
            \path[->] (1) edge [looseness=6, out=-135, in=135] node[midway, left] {\texttt{stay}} (1);
            \node[black] (2) [above right of=1] {$2$};
            \path[->] (1) edge node[midway,left] {\texttt{up}} (2) ;
            \node[black] (a) [right of=2] {};
            \path[->] (2) edge (a);
            \node[black] (b) [above right of=a] {};
            \path[->] (a) edge (b);
            \path[->] (b) edge [looseness=6, out=45, in=135] (b);
            \node[black] (c) [right of=a] {};
            \path[->] (a) edge (c);
            \path[->] (c) edge [looseness=6, out=0, in=90] (c);
            \node[black] (d) [below right of=a] {};
            \path[->] (a) edge (d);
            \path[->] (d) edge [looseness=6, out=-45, in=45] (d);
        % Lower subgraph
            \node[black] (3) [below right of=1] {$3$};
            \path[->] (1) edge node[midway,left] {\texttt{down}} (3);
            \node[black] (4) [above right of=3] {};
            \path[->] (3) edge (4);
            \path[->] (4) edge [looseness=6, out=45, in=135] (4);
            \node[black] (5) [right of=3] {};
            \path[->] (3) edge (5);
            \path[->] (5) edge [looseness=6, out=0, in=90] (5);
     \end{tikzpicture}

    \caption[$\pwrNoDist$-seeking depends on the discount rate]{When $\gamma\approx 0$, $\pwr[s_2, \gamma][\Diid] < \pwr[s_3, \gamma][\Diid]$, and so \texttt{down} seeks $\pwrNoDist_{\Diid}$ compared to \texttt{up} and \texttt{stay} (\cref{thm:most-children-power}). When $\gamma \approx 1$, \texttt{up} seeks $\pwr$ compared to \texttt{down}: $\pwr[s_2, \gamma][\Diid] > \pwr[s_3, \gamma][\Diid]$ (\cref{RSDSimPower}). However, \texttt{stay} is maximally $\pwrNoDist_{\Diid}$-seeking when $\gamma\approx 1$, as demanded by \cref{thm:stay-put-power}. \label{fig:nuances-discount}}
\end{figure}

\begin{restatable}[When $\gamma=1$, staying put is maximally $\pwrNoDist$-seeking]{prop}{stayPut}\label{thm:stay-put-power}
Suppose $\exists a\in\A: T(s,a,s)=1$. When $\gamma=1$, $a$ is a maximally $\pwrNoDist_{\Dbd}$-seeking action at state $s$.
\end{restatable}
\begin{proof}
Staying put via action $a$ has an expected $\pwrNoDist$ of $\pwr[s,1]$. By \cref{lem:future-power}, $\pwr[s,1]\geq \max_{a'} \E{s'\sim T\prn{s,a'}}{\pwr[s',\gamma]}$.
\end{proof}

When $\gamma=1$, \cref{thm:stay-put-power} implies that the agent cannot expect that any action  will increase its $\pwr$.

\section{Attainable utility distance}\label{app:dau}
Consider Definition 1 of \citet{turner2020avoiding}:

\begin{restatable}[\textsc{Aup} reward function]{definition}{AUPrf}
Consider an {\mdp} $\langle \St,\A, T, R, \gamma \rangle$  with state space $\St$, action space $\mathcal{ A}$, transition function $T:\St\times \A \to\Delta(\St)$, reward function $R:\St\times \A\to\reals$, and discount factor $\gamma \in [0,1)$. Let $\lambda\geq 0$ and $\varnothing\in\A$, and let $\mathcal{R}\subsetneq \reals^{\St}$ be a finite set of auxiliary reward functions.
\begin{equation}
\label{eq:aup}
    R_\textsc{aup}(s, a) \defeq R(s,a) - \frac{\lambda}{|\mathcal{R}|} \sum_{R_i\in \mathcal{R}} \abs{ Q^*_{R_i}(s,a) - Q^*_{R_i}(s, \varnothing) }.
\end{equation}
\end{restatable}

The following results provide intuition about how the \textsc{aup} penalty term works in general. We first formalize a distance metric which is tightly linked to the \textsc{aup} penalty term \citep{turner2020conservative,turner2020avoiding}.

\begin{restatable}[Attainable utility distance $\dau{}$]{definition}{distdefn}\label{def:au-dist}
Let $\D$ be a bounded continuous distribution over reward functions bounded $[0,1]$, with probability measure $F$. With respect to $\D$ and discount rate $\gamma\in[0,1)$, the \emph{attainable utility distance} between state distributions $\Delta,\Delta'\in \Delta(\St)$ is
\begin{equation}
    \dau{\Delta,\Delta'}\defeq\int_{\rewardSpace} \abs{\E{s \sim \Delta}{\OptVf{s,\gamma}}-\E{ s' \sim\Delta'}{\OptVf{s',\gamma}}} \dF[R].
\end{equation}
\end{restatable}

Each expectation can be interpreted as the Q-value of an action. With respect to reward function distribution $\D$, $\text{d}_\text{au}$ returns the expected advantage magnitude for one action over another.

\begin{restatable}[$\dau{}$ is a distance metric on $\Delta (\St)$]{prop}{distance}\label{thm:dau}
\end{restatable}
\begin{proof}
For $ \Delta,  \Delta', \Delta''\in \Delta (\St)$:
\begin{enumerate}
    \item $\dau{\Delta,\Delta'}\geq0$.
    \item $\dau{\Delta,\Delta'}=0$ iff $ \Delta= \Delta'$.
    \item $\dau{\Delta,\Delta'}=\dau{\Delta',\Delta}$.
    \item $\dau{\Delta,\Delta''} \leq \dau{\Delta,\Delta'}+\dau{\Delta', \Delta''}{}$.
\end{enumerate}

Properties 1 and 3 are trivially true. Property 2 follows  from \cref{unique-opt-value}. For property 4,
\begin{align}
    \dau{\Delta,\Delta''} &= \int_{\rewardSpace} \Bigg|\prn{\E{s \sim \Delta}{\OptVf{s,\gamma}}-\E{s' \sim \Delta'}{\OptVf{s',\gamma}}}+\\
    &\phantom{= \int_{\rewardSpace} \Bigg|,}\prn{\E{ s' \sim\Delta'}{\OptVf{s',\gamma}}-\E{s'' \sim \Delta''}{\OptVf{s'',\gamma}}}\Bigg| \dF[R].\nonumber\\
    &\leq \int_{\rewardSpace} \abs{\E{s \sim \Delta}{\OptVf{s,\gamma}}-\E{ s' \sim\Delta'}{\OptVf{s',\gamma}}} \dF[R] +\\
    &\phantom{=\;\,}\int_{\rewardSpace} \abs{\E{s' \sim \Delta'}{\OptVf{s',\gamma}}-\E{ s'' \sim\Delta''}{V^*_R(s'')}} \dF[R]\nonumber\\
    &= \dau{\Delta,\Delta'}+\dau{\Delta',\Delta''}.
\end{align}
\end{proof}

Restricting $\dau{}$ to degenerate probability distributions yields a distance metric over the state space.

Viewing the designer as sampling auxiliary reward functions from distribution $\D$, the \textsc{aup} penalty term is the Monte Carlo integration of $\lambda\gamma\cdot \dau{T(s, a),T (s,\varnothing)}$:
\begin{align}
   &\frac{\lambda }{|\mathcal{R}|} \sum_{R_i\in \mathcal{R}} \abs{ Q^*_{R_i}(s,a,\gamma) - Q^*_{R_i}(s, \varnothing,\gamma) }\\
   =&\frac{\lambda\gamma}{|\mathcal{R}|} \sum_{R_i\in \mathcal{R}} \abs{ \E{s_a\sim T(s,a)}{\OptVf[R_i]{s_a}} - \E{s_\varnothing \sim T(s,\varnothing)}{\OptVf[R_i]{s_\varnothing}} }.
\end{align}

Insofar as the Monte Carlo integration approximates $\dau{}$, the attainable utility distance sheds light on the attainable utility penalty term in \cref{eq:aup}. For example, we want to penalize side effects, but not smaller changes, such as easily reversible movement. \Cref{thm:mvmt} guarantees this.

\begin{restatable}[Statewise AU distance inequality]{lem}{statewise}\label{lem:dau-statewise}\strut\\
$\dau{\Delta,\Delta'} \leq \E{s\sim \Delta,s'\sim \Delta'}{\dau{\unitvec,{\unitvec[s']}}}$.
\end{restatable}
\begin{proof}
\begin{align}
    \dau{\Delta, \Delta'} &\defeq \int_{\rewardSpace} \abs{\E{s \sim \Delta}{\OptVf{s,\gamma}}-\E{ s' \sim\Delta'}{\OptVf{s',\gamma}}} \dF[R]\\
    &\leq\int_{\rewardSpace}\E{s\sim \Delta,s' \sim \Delta'}{ \abs{\OptVf{s,\gamma}-\OptVf{s',\gamma}}}\dF[R]\label{eq:ti}\\
    &=\E{s\sim \Delta,s' \sim \Delta'}{\int_{\rewardSpace} \abs{\OptVf{s,\gamma}-\OptVf{s',\gamma}}}\dF[R]\label{eq:lin-expect}\\
    &=\E{s\sim \Delta,s'\sim \Delta'}{\dau{\unitvec,{\unitvec[s']}}}.
\end{align}

\Cref{eq:ti} holds by the triangle inequality. \Cref{eq:lin-expect} holds by the linearity of expectation.
\end{proof}

\begin{restatable}[Statewise AU distance upper bound]{lem}{maxDau}\label{max-dau}
$\forall s,s': \dau{\unitvec,{\unitvec[s']}} < \geom$.
\end{restatable}
\begin{proof}
Because optimal value is bounded $[0,\geom]$, $\dau{\unitvec,{\unitvec[s']}} \leq \geom$. The equality holds iff for almost all $R\in\supp$, $\OptVf{s,\gamma}=\geom$ and $\OptVf{s',\gamma}=0$, or vice versa. But  because $\D$ is continuous, $s'$ must induce positive optimal value for a positive measure set of reward functions.
\end{proof}

\begin{restatable}[AU distance upper bound]{cor}{upperAUdist}\label{max-dau-st}
$\dau{\Delta,\Delta'} < \geom$.
\end{restatable}

\begin{restatable}[One-step reachability bounds average difference in optimal value]{lem}{reachBoundAUDiff}\label{lem:one-step-au-dist}
Let $\Dbd$ be bounded $[b,c]$ and let $\gamma \in [0,1)$. If $s$ and $s'$ can reach each other with probability $1$ in one step, then $\E{R\sim \Dbd}{\abs{\OptVf{s,\gamma}-\OptVf{s',\gamma}}}\leq c-b$, with strict inequality if $\Dbd$ is continuous.
\end{restatable}
\begin{proof}
By  Proposition 1 of \citet{turner2020avoiding} and because each $R\in\supp[\Dbd]$ is bounded $[b,c]$,  $\abs{\OptVf{s,\gamma}-\OptVf{s',\gamma}}\leq (1-\gamma)\frac{c-b}{1-\gamma}=c-b$.

Suppose $\Dbd$ is continuous; then $b<c$. For equality to hold, it must be the case that for almost all $R\in\supp[\Dbd]$, $\abs{\OptVf{s,\gamma}-\OptVf{s',\gamma}}=c-b$. Because we assumed that such $s$ and $s'$ can reach each other in one step, this implies that for almost all such $R$, either $R(s)=b$ and $R(s')=c$, or vice versa. This would imply that $\Dbd$ has a discontinuous marginal reward distribution for these states, which is impossible if $\Dbd$ is continuous. Then the inequality is strict if $\Dbd$ is continuous.
\end{proof}

The following result also applies to the \textsc{aup} penalty term for any $\mathcal{R}$ over reward functions bounded $[0,1]$.
\begin{restatable}[Movement penalties are small]{cor}{mvmt}\label{thm:mvmt}
Let $\Delta\neq \Delta'$. Suppose that all states in the support of $\Delta$ can deterministically reach in one step all states in the support of $\Delta'$, and vice versa. Then $0<\dau{\Delta,\Delta'}< 1$.
\end{restatable}
\begin{proof}
$0<\dau{\Delta,\Delta'}$ by \cref{thm:dau}.
\begin{align}
    \dau{\Delta, \Delta'} &\leq \E{s\sim \Delta,s'\sim \Delta'}{\dau{\unitvec,{\unitvec[s']}}}\label{eq:dau-statewise}\\
    &< 1.\label{eq:leq1}
\end{align}

\Cref{eq:dau-statewise} holds by \cref{lem:dau-statewise}. For \cref{eq:leq1}, apply \cref{lem:one-step-au-dist} to conclude that $\dau{\unitvec,\unitvec[s']}< 1$ for each such $s,s'$. Therefore, $\dau{\Delta, \Delta'}<1$.
\end{proof}

\textsc{Aup} penalizes both seeking and decreasing $\pwr$, compared to the null action.
\begin{restatable}[Change in expected $\pwrNoDist_{\D}$ lower-bounds $\dau{}$]{prop}{avg}\label{thm:vavg-lower-dau}
\begin{equation}
    \dau{\Delta,\Delta'} \geq \abs{ \E{s\sim\Delta}{\vavg}-\E{s'\sim\Delta'}{\vavg[s',\gamma]}}.
\end{equation}
\end{restatable}
\begin{proof}
\begin{align}
    \dau{\Delta,\Delta'} &\defeq \E{R\sim\D}{\abs{\E{s\sim\Delta}{\OptVf{s,\gamma}}-\E{s'\sim\Delta'}{\OptVf{s',\gamma}}}} \\
    &\geq \abs{\E{R\sim\D}{\E{s\sim\Delta}{\OptVf{s,\gamma}}-\E{s'\sim\Delta'}{\OptVf{s',\gamma}}}}\label{eq:rev-ti}\\
    &= \abs{ \E{s\sim\Delta}{\vavg}-\E{s'\sim\Delta'}{\vavg[s',\gamma]}}.
\end{align}

\Cref{eq:rev-ti} follows by the reverse triangle inequality.
\end{proof}

While we conjectured that \textsc{aup} penalizes green cell disruption because it decreases $\pwrNoDist$, \citet{turner2020conservative}'s \texttt{Correction} gridworld showed that \textsc{aup} also penalizes increases in $\pwrNoDist$.

\subsection{Upper-bounding AU distance by variation distance}
\Cref{dtv-dau-bound} shows that $\dau{}$ is upper-bounded by the maximal $\dtv{}$ between the visit distributions of $s$ and $s'$. As \cref{fig:dau-bound} shows, if their visit distributions are ``forced to overlap,'' $\dau{}$ must be relatively small.

\begin{figure}[ht]\centering
     \begin{tikzpicture}[mygraph]
        %nodes
            \node[black] (3) {$3$};
            \node[black] (1) [above left of=3] {$1$};
            \node[black] (2) [below left of=3] {$2$};
            \node[white] (void) [right of=3] {};
        %arrows
            \path[->] (1) edge (3);
            \path[->] (2) edge (3);
            \path[dotted] (3) edge (void);
    \end{tikzpicture}
    \caption[Attainable utility distance is bounded by visit distribution set distance]{For all $\D'$ bounded $[0,1]$, $\dau[\D']{\unitvec[s_1],\unitvec[s_2]}\leq 1$ by \cref{dtv-dau-bound}, since for any $\pi\in\Pi$, $\fpi{s_1}\in\F(s_1),\fpi{s_2}\in\F(s_2)$ only disagree on the initial state.}
    \label{fig:dau-bound}
\end{figure}

\begin{restatable}[AU distance upper-bounded by maximal variation distance of visit distributions]{prop}{dtvDau}\label{dtv-dau-bound}
Let $\D'$ be any reward function distribution which is bounded $[b,c]$ and let $\gamma \in [0,1)$. $\dau[\D']{\Delta,\Delta'}\leq (c-b)\max_{\pi\in\Pi} \dtv{\E{s \sim \Delta}{\fpi{s}(\gamma)}, \E{s'\sim \Delta'}{\fpi{s'}(\gamma)}}$.
\end{restatable}
\begin{proof}
\begin{align}
    \dau[\D']{\Delta,\Delta'} &\defeq \E{R\sim\D'}{\abs{\E{s\sim\Delta}{\OptVf{s,\gamma}}-\E{s'\sim\Delta'}{\OptVf{s',\gamma}}}} \\
    &= \E{\rf \sim \D'}{\abs{\max_{\pi \in \Pi}\prn{\E{s\sim\Delta}{\fpi{s}(\gamma) }-\E{s'\sim\Delta'}{\fpi{s'}(\gamma)}}^\top \rf}}\label{eq:global-opt}\\
    &\leq \E{R \sim \D'}{\max_{\pi \in \Pi}\half\lone{\E{s\sim\Delta}{\fpi{s}(\gamma) }-\E{s'\sim\Delta'}{\fpi{s'}(\gamma)}}(c-b)}\label{eq:max-dist}\\
    &= (c-b)\max_{\pi \in \Pi}\dtv{\E{s\sim\Delta}{\fpi{s}(\gamma)}, \E{s'\sim\Delta'}{\fpi{s'}(\gamma)}}.\label{eq:finish-dau-bound}
\end{align}

\Cref{eq:global-opt} uses a single $\max$ because all optimal policies $\pi\in \optPi$ induce the same optimal value function. For each $\rf \in \supp[\D']$ with $\pi\in \optPi$, consider the difference vector $\dbf\defeq \E{s\sim\Delta}{\fpi{s}(\gamma) }-\E{s'\sim\Delta'}{\fpi{s'}(\gamma)}$. Since $\rf$ is bounded $[b,c]$, $\abs{\dbf^\top\rf}$ is maximized when $\rf$ assigns $c$ reward to the positive entries of $\dbf$, $b$ reward to the negative entries, and $c$ reward to the zero entries.

Since we always have $\lone{\fpi{s}(\gamma)}=\geom$ by \cref{prop:visit-dist-prop}, the negative and positive entries of $\dbf$ both have measure equal to $\half\lone{\E{s\sim\Delta}{\fpi{s}(\gamma) }-\E{s'\sim\Delta'}{\fpi{s'}(\gamma)}}$. Lastly, we maximize over all possible $\pi\in\Pi$. Then \cref{eq:max-dist} follows.

\Cref{eq:finish-dau-bound} follows because
\[\dtv{\E{s\sim\Delta}{\fpi{s}(\gamma)}, \E{s'\sim\Delta'}{\fpi{s'}(\gamma)}}=\half\lone{\E{s\sim\Delta}{\fpi{s}(\gamma) }-\E{s'\sim\Delta'}{\fpi{s'}(\gamma)}}.\]
\end{proof}

\Cref{dtv-dau-bound} shows that no matter the bounded $\D'$, some states \emph{always} must be close in AU distance.

\begin{restatable}[\Cref{dtv-dau-bound} can be extended to only account for policies which induce non-dominated visit distribution functions]{conjSec}{extendDauBound}
\end{restatable}

\begin{restatable}[Average optimal value difference is bounded by maximum visit distribution distance]{cor}{visitBoundPower}
Let $\D'$ be any reward function distribution which is bounded $[b,c]$ and let $\gamma \in [0,1)$. \begin{equation}
    \abs{\vavg[s,\gamma][\D']-\vavg[s',\gamma][\D']}\leq (c-b)\max_{\pi\in\Pi} \dtv{\fpi{s}(\gamma), \fpi{s'}(\gamma)}.
\end{equation}
\end{restatable}
\begin{proof}
By \cref{thm:vavg-lower-dau}, $\abs{\vavg[s,\gamma][\D']-\vavg[s',\gamma][\D']}\leq \dau[\D']{\unitvec,\unitvec[s']}$.\\
By \cref{dtv-dau-bound}, $\dau[\D']{\unitvec,\unitvec[s']}\leq (c-b)\max_{\pi\in\Pi} \dtv{\fpi{s}(\gamma), \fpi{s'}(\gamma)}$.
\end{proof}

\subsection{AU distance for discount rates close to 1}
\begin{restatable}[Normalized $\dau{}$]{definition}{dauNormDef}\label{def:dau-norm}
Let $\gamma\in[0,1),\Delta,\Delta'\in \Delta\prn{\St}$, and let $\D$ be a bounded continuous reward function distribution.
\begin{equation}
    \dauNorm{\Delta,\Delta'}{\gamma}\defeq (1-\gamma)\dau{\Delta,\Delta'}.
\end{equation}

Note that this normalization is order-preserving and $\dauNorm[\D']{\Delta,\Delta'}{\gamma}\in[0,1]$.  \Cref{dau-norm-lip-cont} demonstrates that $\dauNorm{}{}$ extends to $\gamma=1$ via the appropriate limit.
\end{restatable}

\begin{restatable}[For any bounded reward function distribution $\D'$, ${\dauNorm[\D']{\cdot,\cdot}{\gamma}}$ is Lipschitz continuous on $\gamma\in {[0,1]}$]{prop}{dnormLip}\label{dau-norm-lip-cont}
\end{restatable}
\begin{proof}
By \cref{lem:norm-value-lip}, $\lim_{\gamma^*\to\gamma}(1-\gamma^*)\OptVf{s,\gamma^*}$ is Lipschitz continuous on $\gamma\in[0,1]$, with Lipschitz constant depending only on $\lone{\rf}$. Let $\gamma\in(0,1),\Delta,\Delta'\in \Delta\prn{\St}$. Since expectation, subtraction, and absolute value preserve Lipschitz continuity under some bounded Lipschitz constant, we conclude the Lipschitz continuity of
\begin{align}
    \dauNorm[\D']{\Delta,\Delta'}{\gamma} = (1-\gamma)\E{R\sim \D'}{\abs{\E{s \sim \Delta}{\OptVf{s,\gamma}}-\E{ s' \sim\Delta'}{\OptVf{s',\gamma}}}}.
\end{align}

By the above continuity, we can extend $\dauNorm{}{}$ to $\gamma=1$ via the appropriate limit.
\end{proof}

\begin{restatable}[$\dau{}$ is continuous on $\gamma\in[0,1)$]{cor}{dauCont}\label{dau-cont}
\end{restatable}
\begin{proof}
By \cref{def:dau-norm}, $\dau{\cdot,\cdot}=\geom \dauNorm{\cdot,\cdot}{\gamma}$.\\
By \cref{dau-norm-lip-cont}, $\dauNorm{\cdot,\cdot}{\gamma}$ is continuous on $\gamma\in[0,1)$, as is $\geom$. The space of continuous functions is closed under pointwise multiplication.
\end{proof}

\begin{remark}
By the continuity of $\dau{}$, if $\Delta,\Delta',\Delta''\in\Delta(\St)$ are such that $\dauNorm{\Delta,\Delta'}{1}>\dauNorm{\Delta,\Delta''}{1}$, then for all $\gamma$ sufficiently close to $1$, $\dau{\Delta,\Delta'}>\dau{\Delta,\Delta''}$.
\end{remark}

For \cref{dau-rsd-loss} and \cref{dau-rsd-similar-loss}, we use the following shorthand: for any $s_{a'},s_a, s_\varnothing\in\St$, $D_{a'}\defeq \RSDnd[s_{a'}], D_a \defeq \RSDnd[s_a], D_\varnothing \defeq \RSDnd[s_\varnothing]$.

\begin{restatable}[Losing access to \textsc{rsd}s increases $\dauNorm{}{}$]{prop}{dauRSDLoss}\label{dau-rsd-loss}
If $D_{a'}\subseteq D_a\subseteq D_\varnothing$, then $\dauNorm{\unitvec[s_\varnothing],\unitvec[s_{a}]}{1}\leq \dauNorm{\unitvec[s_\varnothing],\unitvec[s_{a'}]}{1}$. If $D_{a'}\subsetneq D_a$, then the inequality is strict.
\end{restatable}
\begin{proof}
\begin{align}
    \dauNorm{\unitvec[s_\varnothing],\unitvec[s_{a}]}{1}&=\int_{\rewardVS}\abs{\max_{\dbf\in D_\varnothing} \dbf^\top \rf - \max_{\dbf'\in D_a} \dbf'^\top \rf}\dF[\rf]\label{daunorm-defn-rsd}\\
    &\leq \int_{\rewardVS}\abs{\max_{\dbf\in D_\varnothing} \dbf^\top \rf - \max_{\dbf'\in D_{a'}} \dbf'^\top \rf}\dF[\rf]\label{eq:integrand-ineq-daunorm}\\
    &=\dauNorm{\unitvec[s_\varnothing],\unitvec[s_{a'}]}{1}.
\end{align}

\Cref{daunorm-defn-rsd} holds by the definition of \textsc{rsd}s (\cref{def:rsd}). Since $D_{a'}\subseteq D_a$, for all reward functions $\rf \in \rewardVS$,
\begin{equation}
    \abs{\max_{\dbf \in D_\varnothing} \dbf^\top \rf - \max_{\dbf' \in D_a} \dbf'^\top \rf}\leq \abs{\max_{\dbf \in D_\varnothing} \dbf^\top \rf - \max_{\dbf' \in D_{a'}} \dbf'^\top \rf}.\label{eq:ineq-rsd-subset}
\end{equation}

If $D_{a'}\subsetneq D_a$, by \cref{prop:helper-positive-prob}, $\D$ assigns positive measure to the set of reward functions for which some $\dbf''\in D_a\setminus D_{a'}$ is strictly gain-optimal. Therefore, \cref{eq:ineq-rsd-subset} is strict for a positive measure set of reward functions, and so \cref{eq:integrand-ineq-daunorm} holds.
\end{proof}

\begin{figure}[ht]
    \centering
     \begin{tikzpicture}[mygraph]
        \node[blue] (1) {$s$};
        \node[black] (2) [above right of=1] {};
        \path[->] (1) edge (2);
        \path[->] (2) edge [looseness=5, out=45, in=135] (2);
        \node[black] (3) [right of=1] {};
        \path[->] (1) edge (3);
        \node[black] (5) [right of=3] {};
        \path[->] (3) edge (5);
        \path[->] (5) edge [looseness=5, out=0, in=90] (5);
    \end{tikzpicture}
    \caption[Attainable utility distance when \mbox{$\gamma=1$}]{At $\gamma = 1$, all other states are ``equally distant'' from $\col{blue}{s}$ because they each can only access a single 1-cycle \textsc{rsd}. By \cref{dau-rsd-similar-loss}, $\forall s', s'' \neq \col{blue}{s}: \dauNorm{\unitvec[\col{blue}{s_1}],\unitvec[s']}{1}= \dauNorm{\unitvec[\col{blue}{s_1}],\unitvec[s'']}{1}$.}
    \label{fig:sim-rsd-loss}
\end{figure}

\begin{restatable}[Losing access to similar \textsc{rsd}s implies equal $\dauNorm{}{}$]{prop}{dauRSDSimilarLoss}\label{dau-rsd-similar-loss}
If $D_{a'}, D_a \subseteq D_\varnothing$, if some $D_\text{sub}\subseteq D_a$  is similar to $D_{a'}$, and if $\lone{D_{a'}-D_\varnothing \setminus D_{a'}}=\lone{D_\text{sub}-D_\varnothing\setminus D_\text{sub}}=2$, then $\dauNorm{\unitvec[s_\varnothing],\unitvec[s_{a}]}{1}\leq \dauNorm{\unitvec[s_\varnothing],\unitvec[s_{a'}]}{1}$. If $D_\text{sub}\subsetneq D_a$, then the inequality is strict.
\end{restatable}
\begin{proof}
Let $\phi$ be the guaranteed state permutation such that $\phi\cdot D_\text{sub}=D_{a'}$. Define
\begin{equation}
    \phi'(s)\defeq
    \begin{cases}
    \phi(s) &\text{ if } s \text{ visited by } \dbf \in D_\text{sub}\\
    \phi^{-1}(s) &\text{ if } s \text{ visited by } \dbf \in D_{a'}\\
    s &\text{ else}.
    \end{cases}\label{eq:phi-prime-rsd-similar}
\end{equation}
By the $\lone{\cdot}$ assumption, $\phi'$ is a well-defined permutation.
\begin{align}
    \dauNorm{\unitvec[s_\varnothing],\unitvec[s_{a}]}{1} &= \E{\rf\sim \D}{\abs{\max_{\dbf\in D_\varnothing} \dbf^\top \rf - \max_{\dbf'\in D_{a}} \dbf'^\top \rf}}\label{eq:rsd-similar-gain}\\
    &\leq \E{\rf\sim \D}{\abs{\max_{\dbf\in D_\varnothing} \dbf^\top \rf - \max_{\dbf'\in D_\text{sub}} \dbf'^\top \rf}}\label{eq:ineq-phi-prime}\\
    &= \E{\rf\sim \D}{\abs{\max_{\dbf\in D_\varnothing} \dbf^\top \rf' - \max_{\dbf'\in D_{a'}} \dbf' \rf'}}\label{eq:rsd-similar-sets}\\
    &=  \dauNorm{\unitvec[s_\varnothing],\unitvec[s_{a'}]}{1}.
\end{align}

\Cref{eq:rsd-similar-gain} holds by the definition of \textsc{rsd}s (\cref{def:rsd}). \Cref{eq:ineq-phi-prime} holds by \cref{dau-rsd-loss}, with strict inequality if $D_\text{sub}\subsetneq D_a$.

Let $g(b_1,b_2)\defeq \abs{b_1-b_2}$ and let $f(B_1,B_2\mid \D)\defeq \E{\rf\sim\D}{g\prn{\max_{\dbf\in B_1} \dbf^\top \rf,},\max_{\dbf\in B_2} \dbf^\top \rf}$. Then \cref{lem:helper-perm} shows that $f(D_\varnothing, D_\text{sub}\mid \D)=f(\phi'(D_\varnothing), \phi'(D_\text{sub})\mid \phi'(\D))$.

By the $\lone{\cdot}$ assumption, $\phi'(D_\text{sub})=D_{a'}$, $\phi'(D_{a'})=D_\text{sub}$,  and $\phi'\prn{D_\varnothing \setminus\prn{D_{a'}\cup D_\text{sub}}}=D_\varnothing \setminus\prn{D_{a'}\cup D_\text{sub}}$ by \cref{eq:phi-prime-rsd-similar}. Then $\phi'(D_\varnothing)=D_\varnothing$. Since $\D$ distributes reward identically across states, $\phi'(\D)=\D$. We thus conclude that $f(D_\varnothing, D_\text{sub}\mid \D)=f(D_\varnothing, D_{a'}\mid \D)$, and so \cref{eq:rsd-similar-sets} follows.
\end{proof}

\begin{restatable}[AUP penalty sampling bounds]{prop}{aupPenalty}\label{prop:aup-sample-bound}
Let $\gamma \in [0,1]$, $\D'$ be a reward function distribution which is bounded $[b,c]^{\abs{\St}}$, $s$ be a state, $a,\varnothing$ be actions, and $\epsilon >0$. For reward functions $R_1, \ldots, R_n$, define $ \overline{\abs{\Delta Q^{R_1,\ldots,R_n}_{s,a,\varnothing}}}\defeq \frac{1}{n}\sum_{i=1}^n \abs{\OptQfNorm[R_i]{s,a,\gamma}-\OptQfNorm[R_i]{s,\varnothing,\gamma}}$.
\begin{equation}
    \prob[R_1,\ldots, R_n \sim \D']{\abs{\overline{\abs{\Delta Q^{R_1,\ldots,R_n}_{s,a,\varnothing}}}-\dauNorm[\D']{T(s,a),T(s,\varnothing)}{\gamma}}\geq \epsilon}\leq 2e^{-\frac{2n \epsilon^2}{(c-b)^2}}.
\end{equation}
\end{restatable}
\begin{proof}
Normalized optimal Q-value is bounded $[b,c]$, and so the absolute difference is bounded $[0,c-b]$. Since the draws $R_i \sim \D'$ are independent, apply Hoeffding's inequality.
\end{proof}

\section{Proportional regret}
We formalize a relaxed variant of worst-case regret minimization which accounts for the human's ability to later correct the robot. We show a common-sense no-free lunch theorem: without any way of learning about the true reward function, no policy can do better than losing half of its value in the worst-case. This result underscores the importance of thinking carefully about what objective we wish to give the agent, and ensuring it can gather enough information about our preferences.

We also show that even when the human may later communicate the true reward function to the robot, it is often impossible to meaningfully minimize worst-case regret. In particular, this occurs when the agent is forced to make an irreversible decision early on.

\citet{regan2010robust} select policies which minimize worst-case regret against a set of feasible reward functions. We show that when this feasible set is large enough, worst-case regret minimization is infeasible if the agent cannot learn more about the true objective.

\citet{regan_robust_2010} quantify regret as the decrease in value from following a suboptimal policy $\pi$: $\Vf[R]{s,\gamma}-V^{\pi}_{R}(s,\gamma)$. However, while optimal policies are invariant to positive rescaling of the reward function, absolute regret is not: starting from state $s$, if $\pi$ induces $1$ regret for reward function $R$, then $\pi$ induces $10$ regret for $10R$! The \textit{relative regret} \citep{inuiguchi_achievement_1997} of a policy $\pi$ is $\frac{\Vf{s,\gamma}-V^\pi(s,\gamma)}{\Vf{s,\gamma}}$, but the denominator is not invariant to translation of the reward function.

We propose a regret metric which quantifies the \emph{proportion} of attainable value lost by following a suboptimal policy.

\begin{restatable}[Minimal value]{definition}{minValue}
$\Vfmin{s,\gamma}\defeq \min_{\pi\in\Pi} V^\pi_R(s,\gamma)=-\Vf[-R]{s,\gamma}$.
\end{restatable}

\begin{restatable}[Proportional regret]{definition}{defnPropRegret}\label{def:prop-regret}
Let $s$ be a state and let $\pi$ be a policy, $R\in \rewardSpace$, $\gamma \in [0,1]$. The \emph{proportional regret} of following policy $\pi$ is
\begin{equation}
\pregret[\pi\mid R, s,\gamma] \defeq  \frac{\VfNorm[R]{s,\gamma}-V^{\pi}_{R,\text{norm}}(s,\gamma)}{\VfNorm[R]{s,\gamma}-\VfminNorm[R]{s,\gamma}}.
\end{equation}

$\pregret$ is defined to be $0$ when the denominator is $0$, as no policy can incur regret for $R$ starting from $s$ under such conditions. In particular, this occurs when $\gamma=0$ in our state-based reward setting: Reward from other states is discounted away, and so all policies are optimal. $\pregret$ is clearly bounded $[0,1]$.
\end{restatable}
\begin{restatable}[Optimal policies have $0$ $\pregret$, while maximally suboptimal policies have $1$ $\pregret$]{conjSec}{optPolRegret}
\end{restatable}

\begin{restatable}[$\pregret$ is invariant to positive affine transformation of the reward function]{prop}{pregretInvar}\label{prop:pregret-invar}
Let $\pi$ be any policy, $R \in \rewardSpace$, $s\in\St$, $\gamma \in [0,1]$, and $m>0,b\in\reals$.
\begin{equation}
    \pregret[\pi\mid R, s,\gamma]= \pregret[\pi\mid mR+b, s,\gamma].
\end{equation}
\end{restatable}

\begin{proof}
Suppose $\gamma < 1$.
\begin{align}
    &\pregret[\pi\mid R,s,\gamma]\\
    &\defeq \frac{\VfNorm[R]{s,\gamma}-V^{\pi}_{R,\text{norm}}(s,\gamma)}{\VfNorm[R]{s,\gamma}-\VfminNorm[R]{s,\gamma}}\\
    &=  \frac{m^{-1}\VfNorm[mR+b]{s,\gamma}-\frac{b}{1-\gamma}-\prn{m^{-1}V^{\pi}_{mR+b,\text{norm}}(s,\gamma)-\frac{b}{1-\gamma}}}{m^{-1}\VfNorm[mR+b]{s,\gamma}-\frac{b}{1-\gamma}-\prn{m^{-1}\VfminNorm[mR+b]{s,\gamma} -\frac{b}{1-\gamma}}}\\
    &=  \frac{\VfNorm[mR+b]{s,\gamma}-V^{\pi}_{mR+b,\text{norm}}(s,\gamma)}{\VfNorm[mR+b]{s,\gamma}-\VfminNorm[mR+b]{s,\gamma}}\\
    &= \pregret[\pi\mid mR+b,s,\gamma].
\end{align}

The $\gamma=1$ case follows automatically from the equality for all $\gamma < 1$.
\end{proof}

\begin{restatable}[Reward function negation flips the $\pregret$ of any policy]{prop}{negateRF}\label{thm:negate-rf}
Let $\pi$ be any policy, $R \in \rewardSpace$, $s\in\St$, $\gamma \in [0,1]$. If $\VfNorm[R]{s,\gamma}>\VfminNorm[R]{s,\gamma}$, then
\begin{equation}
    \pregret[\pi\mid -R, s,\gamma]= 1-\pregret[\pi\mid R, s,\gamma].
\end{equation}
\end{restatable}
\begin{proof}
\begin{align}
    &\pregret[\pi\mid -R,s,\gamma]\\
    &\defeq  \frac{\VfNorm[-R]{s,\gamma}-V^{\pi}_{-R,\text{norm}}(s,\gamma)}{\VfNorm[-R]{s,\gamma}-\VfminNorm[-R]{s,\gamma}}\\
    &=  \frac{-\VfminNorm[R]{s,\gamma}+V^{\pi}_{R,\text{norm}}(s,\gamma)}{-\VfminNorm[R]{s,\gamma}+\VfNorm{s,\gamma}}\\
    &= \frac{V^{\pi}_{R,\text{norm}}(s,\gamma)-\VfminNorm[R]{s,\gamma}}{\VfNorm{s,\gamma}-\VfminNorm[R]{s,\gamma}}\\
    &= 1+ \frac{-\prn{\VfNorm{s,\gamma}-\VfminNorm[R]{s,\gamma}}+V^{\pi}_{R,\text{norm}}(s,\gamma)-\VfminNorm[R]{s,\gamma}}{\VfNorm{s,\gamma}-\VfminNorm[R]{s,\gamma}}\\
    &= 1+ \frac{-\VfNorm{s,\gamma}+V^{\pi}_{R,\text{norm}}(s,\gamma)}{\VfNorm{s,\gamma}-\VfminNorm[R]{s,\gamma}}\\
    &= 1- \frac{\VfNorm{s,\gamma}-V^{\pi}_{R,\text{norm}}(s,\gamma)}{\VfNorm{s,\gamma}-\VfminNorm[R]{s,\gamma}}\\
    &= 1-\pregret[\pi\mid R,s,\gamma].
\end{align}
\end{proof}

\subsection{No free lunch for robust optimization}
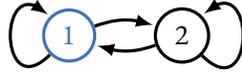
\begin{figure}[ht]
    \centering
    \begin{tikzpicture}[mygraph]
        %nodes
            \node[blue] (1) {$1$};
            \node[black] (2) [right of=1] {$2$};
        %arrows
            \path[->] (1) edge [looseness=1, out=20, in=160] (2);
            \path[->] (2) edge [out=-160, in=-20] (1);
            \path[->] (1) edge [looseness=5, out=-135, in=135] (1);
            \path[->] (2) edge [looseness=5, out=-45, in=45,color=black] (2);
     \end{tikzpicture}
    \caption[No-free-lunch for simultaneously optimizing multiple objectives]{If $\R[true]$ is the set of state indicator reward functions, then no policy can do better than alternating between the two states. When $\gamma=1$, this policy induces worst-case $\pregret$  of $\half$: half of the attainable value is lost for each reward function in $\R[true]$.}
    \label{fig:NFL}
\end{figure}

Let $\R[true]\subseteq \rewardSpace$ be a set of reward functions. Robust optimization minimizes maximal regret with respect to this feasible set $\R[true]$ \citep{regan_robust_2010}. We show that when $\R[true]$ has enough reward functions in it, no policy can do well in the worst case. As illustrated by \cref{fig:NFL}, no policy can simultaneously optimize a generic reward function and its inverse.

\begin{restatable}[No free lunch theorem for proportional regret minimization]{thm}{noFree}\label{thm:no-free}
Let $\pi$ be any policy, $s$ a state, and $\gamma\in[0,1]$. If there exists $R \in \R[true]$ such that a negatively affinely transformed $-mR+b\in \R[true]$ as well, and if $\VfNorm{s,\gamma}>\VfminNorm{s,\gamma}$, then
\begin{equation}\sup_{R_\text{true} \in \R[true]} \pregret[\pi \mid R_\text{true},s,\gamma] \geq \half.\end{equation}
\end{restatable}
\begin{proof}
\begin{align}
    &\sup_{R_\text{true} \in \R[true]} \pregret[\pi \mid R_\text{true},s,\gamma] \\
    &\geq \max_{R_\text{true} \in \set{R, -mR+b}} \pregret[\pi \mid R_\text{true},s,\gamma]\label{eq:max-rtrue}\\
    &= \max_{R_\text{true} \in \set{R, -R}}\pregret[\pi \mid R_\text{true},s,\gamma] \label{eq:nofree-invar}\\
    &= \max\prn{\pregret[\pi \mid R_\text{true},s,\gamma], 1-\pregret[\pi \mid R_\text{true},s,\gamma]}\label{eq:nofree-invert}\\
    &\geq .5\label{eq:half}.
\end{align}

\Cref{eq:max-rtrue} holds because $R, -mR+b \in \R[true]$. \Cref{eq:nofree-invar} holds by \cref{prop:pregret-invar}. \Cref{eq:nofree-invert} holds by \cref{thm:negate-rf}, which can be applied since \[\VfNorm{s,\gamma}>\VfminNorm{s,\gamma}.\] \Cref{eq:half} holds because $\pregret$ is bounded $[0,1]$.
\end{proof}

\begin{restatable}[Uninformative ${\R[true]}$ satisfy no-free-lunch conditions]{prop}{uninformNFL}\label{uninformative-NFL}
Let $\pi$ be any policy and $s\in\St$. Suppose there exist $a<b$ such that $[a,b]^{\St} \subseteq \R[true]$. If either $\gamma \in (0,1)$ and $\abs{\F(s)}>1$ or $\gamma=1$ and $\abs{\RSD}>1$, then $\sup_{R_\text{true} \in \R[true]} \pregret[\pi \mid R_\text{true},s,\gamma] \geq \half$.
\end{restatable}

\begin{proof}
Let $U \defeq [a,b]^{\St}$ for the assumed real numbers $a < b$. Let $U^- \defeq \{R\in U \mid \exists m_R > 0, b_R \in \reals: -m_R R + b_R \in U\}$. Since all $R\in U$ are bounded, \emph{any} reward function $-m_R R + b_R$ can be positively affinely transformed so that its reward is bounded $[a,b]$. Therefore, $U^- = U$.

Since $\gamma > 0$ and either $\abs{\F(s)}>1$ or $\abs{\RSD}>1$, by \cref{unique-optimal}, almost every reward function in $U^-$ has a strictly optimal visit distribution or {\rsd} at discount rate $\gamma$. Since $U^-=U$ has positive measure, almost all elements of $U^-$ must have a strictly optimal visit distribution or {\rsd} at discount rate $\gamma$. Let $R\in U^-$ be one such element. We conclude that $\VfNorm{s,\gamma}>\VfminNorm{s,\gamma}$.

By the definition of $U^-$, $R$ has a negatively affinely transformed counterpart in $U\subseteq \R[true]$. Then \cref{thm:no-free} implies that $\sup_{R_\text{true} \in \R[true]} \pregret[\pi \mid R_\text{true},s,\gamma] \geq .5$.
\end{proof}

\begin{remark}
\Cref{uninformative-NFL}'s conditions of $\abs{\F(s)}>1$ or $\abs{\RSD}>1$ are trivial: if they are not met, then the agent makes no meaningful choices and $\pregret$ trivially equals $0$ for every policy and reward function.
\end{remark}

\subsection{Corrigible regret minimization}
Even though we often can't fully specify the intended reward function or minimize worst-case regret under complete uncertainty, we can correct an agent after watching it make mistakes.

\begin{restatable}[Corrigibility]{definition}{corrigibility}\label{def:corrigibility}
An agent-supervisor pair enables \emph{perfect corrigibility} when the supervisor can modify the agent's policy to any other policy.
\end{restatable}

\Cref{def:corrigibility} is optimistic: it is obviously unrealistic to demand the supervisor be able to implement in the agent an optimal policy for any reward function. Furthermore, the agent may act to avoid correction \citep{soares_corrigibility_2015, carey_incorrigibility_2017,turner_optimal_2020}, or an agent may leave the supervisor's range of correction. We set these complications aside for now.

\Cref{thm:no-free} does not imply that $\epsilon$-$\pregret$ minimization is impossible for $\epsilon < \half$, even when $\R[true]=\rewardSpace$. \Cref{thm:no-free} says that any $\pi$ cannot do well across all reward functions, if $\pi$ cannot somehow be conditioned on each $R_\text{true}$. However, \cref{def:corrigibility} allows the agent to be ``corrected'' after $t$ time steps to an optimal policy $\pi^*_{R_\text{true}}$.

Even if the agent cannot somehow discover which reward function it should optimize, exogenous correction by the supervisor often allows an agent to bound its worst-case proportional regret.

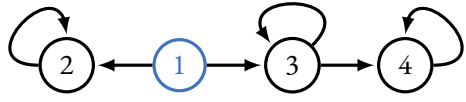
\begin{figure}[ht]
    \centering
    \begin{tikzpicture}[mygraph]
        %nodes
            \node[blue] (1) {$1$};
            \node[black] (2) [left of=1] {$2$};
            \node[black] (3) [right of=1] {$3$};
            \node[black] (4) [right of=3] {$4$};
        %arrows
            \path[->] (1) edge (2);
            \path[->] (1) edge (3);
            \path[->] (2) edge [looseness=5, out=-180, in=90] (2);
            \path[->] (3) edge [looseness=5, out=45, in=135] (3);
            \path[->] (3) edge (4);
            \path[->] (4) edge [looseness=5, out=0, in=90] (4);
    \end{tikzpicture}
    \caption[Corrigible regret minimization is not always possible]{In some environments, non-trivial corrigible regret minimization is impossible. The agent starts at $\col{blue}{s_1}$ (the starting state $s$ is shown in \col{blue}{blue}). Suppose $\exists R, R' \in \R[true]: R(s_2)>R(s_3) \land R'(s_2)<R'(s_3)$. Then $\forall \gamma\in(0,1], t>0$, no policy can avoid incurring maximal worst-case proportional regret for $\R[true]$. However, intuitively, going right is ``less option-destroying.''}
    \label{fig:robust-impossible}
\end{figure}

\citet{eysenbach2018leave} train an agent to preserve initial state reachability. \Cref{thm:off-policy} shows that reversibility allows the agent to bound worst-case regret, if the agent can later be corrected to pursue the true objective.

\begin{restatable}[Given perfect corrigibility, initial state reachability bounds worst-case $\pregret$]{prop}{initialBound}\label{thm:off-policy}
Let $\gamma\in [0,1]$ and let $\pi$ be any policy which, when followed from $s$ for $t$ steps, has probability $1$ of residing in states which can reach $s$ in $k$ steps with probability $1$.
\begin{equation}
    \sup_{R\in\rewardSpace} \pregret[\piSwitch{\pi} \mid R, s, \gamma] \leq 1-\gamma^{t+k},\label{eq:sup-initial-reach}
\end{equation}

where $\pi^*_R\in\optPi[R,\gamma]$ for each $R$.
\end{restatable}

\begin{proof}
If $\forall R \in \rewardSpace:\VfNorm{s,\gamma}=\VfminNorm{s,\gamma}$, then the supremum in \cref{eq:sup-initial-reach} equals $0$ by \cref{def:prop-regret} and we are done. Otherwise, let $Y \subseteq \rewardSpace$ be the subset of reward functions for which this equality does not hold.

Suppose $\gamma \in [0,1)$. Let $\pi_\text{return}$ be a policy which always navigates to $s$ as quickly as possible, when possible. Let $\pi_\text{recover}(\pi')\defeq \pi_\text{switch}\prn{\piSwitch[\pi_\text{return}]{\pi},\pi',t+k}$ be the non-stationary policy which follows $\pi$ for the first $t$ time steps, switches to $\pi_\text{return}$ for the next $k$ time steps, and then follows $\pi'$ thereafter. Starting from $s$, the value gained before time $t+k$ is then $G^\text{recover}_{R}(\gamma)\defeq V^{\pi_\text{recover}(\pi,\pi_\text{return},\pi^*_R)}_{R}(s,\gamma)- \gamma^{t+k}\Vf{s,\gamma}$.
\begin{align}
    &\sup_{R\in \rewardSpace} \pregret[\piSwitch{\pi} \mid R, s, \gamma]\\
    &=\sup_{R\in Y} \pregret[\piSwitch{\pi} \mid R, s, \gamma]\label{eq:ignore-trivial}\\
    &\defeq \sup_{R\in Y} \frac{\VfNorm[R]{s,\gamma}-V^{\piSwitch{\pi}}_{R,\text{norm}}(s,\gamma)}{\VfNorm[R]{s,\gamma}-\VfminNorm[R]{s,\gamma}}\\
    &=\sup_{R\in Y} \frac{\Vf[R]{s,\gamma}-V^{\piSwitch{\pi}}_{R}(s,\gamma)}{\Vf[R]{s,\gamma}-\Vfmin[R]{s,\gamma}}\label{eq:ignore-norm}\\
    &\leq \sup_{R\in Y} \frac{\Vf[R]{s,\gamma}-V^{\pi_\text{recover}(\pi,\pi_\text{return},\pi^*_R)}_{R}(s,\gamma)}{\Vf[R]{s,\gamma}-\Vfmin[R]{s,\gamma}}\label{eq:subopt-recover}\\
    &\leq \sup_{R\in Y} \frac{\Vf[R]{s,\gamma}-V^{\pi_\text{recover}(\pi,\pi_\text{return},\pi^*_R)}_{R}(s,\gamma)}{\Vf[R]{s,\gamma}-\frac{1}{1-\gamma^{t+k}} G^\text{recover}_{R}(\gamma)}\label{eq:min-bound}\\
    &= \sup_{R\in Y} \frac{\Vf[R]{s,\gamma}-G^\text{recover}_{R}(\gamma) - \gamma^{t+k}\Vf{s,\gamma}}{\Vf[R]{s,\gamma}-\frac{1}{1-\gamma^{t+k}} G^\text{recover}_{R}(\gamma)}\label{eq:recover-decompose}\\
    &= \sup_{R\in Y} \frac{\prn{1 - \gamma^{t+k}}\Vf{s,\gamma}-G^\text{recover}_{R}(\gamma) }{\Vf[R]{s,\gamma}-\frac{1}{1-\gamma^{t+k}} G^\text{recover}_{R}(\gamma)}\\
    &= 1-\gamma^{t+k}.
\end{align}

\Cref{eq:ignore-trivial} follows because all reward functions $R' \in \rewardSpace\setminus Y$ have $0$ $\pregret$ by the definition of $Y$, and $\pregret$ is bounded $[0,1]$. \Cref{eq:ignore-norm} follows because the continuity of (optimal) value functions on $\gamma \in [0,1)$ allows us to ignore the limit in the normalized value functions (\cref{def:value-norm}). \Cref{eq:subopt-recover} holds because
\begin{equation*}
    V^{\pi_\text{recover}(\pi,\pi_\text{return},\pi^*_R)}_{R}(s,\gamma) \leq V^{\piSwitch{\pi}}_{R}(s,\gamma),
\end{equation*}
as $\pi_\text{recover}(\pi,\pi_\text{return},\pi^*_R)$ takes longer to begin following an optimal policy for $R$.

By the definition of $\Vfmin{s,\gamma}$, $\Vfmin{s,\gamma}\leq \frac{1}{1-\gamma^{t+k}} G^\text{recover}_{R}(\gamma)$, the value of forever alternating between following $\pi$ for $t$ steps and $\pi_\text{return}$ for $k$ steps. Therefore, \cref{eq:min-bound} holds.

If $\gamma=1$, then for all $R \in \rewardSpace$,
\begin{align}
\VfNorm{s,1}&=V^{\pi_\text{recover}(\pi,\pi_\text{return},\pi^*_R)}_{R,\text{norm}}(s,1)\label{eq:equal-recover}\\
&\leq V^{\piSwitch{\pi}}_{R,\text{norm}}(s,1)\label{eq:leq-recover}\\
&\leq \VfNorm{s,1}.\label{eq:optimal}
\end{align}

\Cref{eq:equal-recover} holds by \cref{def:value-norm}: since $s$ can be returned to within $k$ steps, the transient reward from the first $t+k$ steps does not affect the normalized $V^{\pi_\text{recover}(\pi,\pi_\text{return},\pi^*_R)}_{R,\text{norm}}(s,1)$. \Cref{eq:optimal} holds by the definition of normalized optimal value. Therefore, \[\pregret[\piSwitch{\pi} \mid R, s, 1]=0,\] and the supremum in \cref{eq:sup-initial-reach} also equals $0$.
\end{proof}

\Cref{thm:off-policy} shows that as we take longer to correct the agent, or the agent takes longer to undo its actions, $\pregret$ increases. On the other hand, as the discount rate increases to $1$, $\pregret$ decreases because transient mistakes become relatively less important. \Cref{fig:sharp-bound} shows that the inequality in \cref{thm:off-policy} is sharp.

\begin{figure}[ht]
    \centering
    \begin{tikzpicture}[mygraph]
        %nodes
            \node[blue] (1) {$1$};
            \node[black] (2) [right of=1] {$2$};
            \node[black] (3) [left of=1] {$3$};
        %arrows
            \path[<->] (1) edge (2);
            \path[->] (1) edge (3);
            \path[->] (3) edge [looseness=5, out=-180, in=90] (3);
    \end{tikzpicture}
    \caption[Initial state reachability bounds worst-case regret, if the agent can be corrected later]{Let $R(\col{blue}{s_1})\defeq0, R(s_2)\defeq .5, R(s_3)\defeq1$, and let $\pi_\text{right}$ go right at $\col{blue}{s_1}$. $\pregret[\pi_\text{switch}(\pi_\text{right},\pi^*_R,1) \mid R, s, \gamma] = 1-\gamma^{2}$. At $s_2$, $k=1$ step is required to return to the initial state $\col{blue}{s_1}$, and $t=1$. Therefore, $1-\gamma^2=1-\gamma^{t+k}$, so \cref{thm:off-policy}'s bound is tight.}
    \label{fig:sharp-bound}
\end{figure}

\begin{restatable}[Communicating {\mdp}]{definition}{DefCommunicating2}
An {\mdp} is \emph{communicating} when every state is able to reach every other state with positive probability.
\end{restatable}

\begin{restatable}[Given perfect corrigibility, all policies are low-regret in communicating {\mdp}s for $\gamma =1$]{prop}{communicatingLow}\label{prop:comm-low-impact}
Suppose the environment is communicating, let $\pi$ be any policy, and let $\epsilon > 0$. If the agent can be corrected within $t$ time steps, then
\begin{equation*}
    \sup_{R\in\rewardSpace} \pregret[\piSwitch{\pi} \mid R, s, 1]=0.
\end{equation*}
\end{restatable}

\begin{proof}
Since the {\mdp} is communicating, $\forall R\in \rewardSpace, s,s' \in \St: \VfNorm{s,1}=\VfNorm{s',1}$ by \cref{lem:eq-gain}. This implies that $\VfNorm{s,1}=V^{\piSwitch{\pi}}_{R,\text{norm}}(s,1)$, because the agent switches to an optimal policy after $t$ time steps (transient reward differences vanish in the $\gamma=1$ case). This implies that $\forall R \in \rewardSpace: \pregret[\piSwitch{\pi} \mid R, s, 1]=0$.
\end{proof}

\begin{restatable}[Given perfect corrigibility, all policies are low-regret in communicating {\mdp}s for $\gamma \approx 1$]{conjSec}{communicatingLowConj}
Suppose the environment is communicating, let $\pi$ be any policy, and let $\epsilon > 0$. If the agent can be corrected within $t$ time steps, then there exists some $\gamma < 1$ such that
\begin{equation}
     \sup_{R\in\rewardSpace} \pregret[\piSwitch{\pi} \mid R, s, \gamma]< \epsilon.
\end{equation}
\end{restatable}

Suppose that the human designers have uncertainty about what reward function they should provide, with the uncertainty represented by a probability distribution $\D$.

\begin{restatable}[Worst-case $\pregret$ minimization is equivalent to robustness against $\D$]{prop}{uncertainEquiv} Let $\pi$ be any policy and let $\R[true]\subseteq\rewardSpace$.
\begin{equation}
     \sup_{\D\in \Delta(\R[true])} \E{R\sim \D}{\pregret[\pi \mid R, s, \gamma]}=\sup_{R\in\R[true]} \pregret[\pi \mid R, s, \gamma].\label{eq:uncertain-equiv}
\end{equation}
\end{restatable}
\begin{proof}
Suppose $(R_n)_{n\geq1}$ is such that $\forall n:R_n\in\R[true]$ and
\begin{equation*}
    \lim_{n\to\infty} \pregret[\pi \mid R_n, s, \gamma]=\sup_{R\in\R[true]} \pregret[\pi \mid R, s, \gamma].
\end{equation*}
Then let $(\D_n)_{n\geq1}$ be a sequence of degenerate probability distributions which place probability $1$ on $R_n$. Then each $\D_n\in\Delta(\R[true])$. Furthermore,
\begin{align}
    &\sup_{\D\in \Delta(\R[true])} \E{R\sim \D}{\pregret[\pi \mid R, s, \gamma]}\\
    \geq{}&\lim_{n\to\infty} \E{R\sim \D_n}{\pregret[\pi \mid R_n, s, \gamma]}\\
    ={}&\lim_{n\to\infty} \pregret[\pi \mid R_n, s, \gamma]\\
    ={}& \sup_{R\in\R[true]} \pregret[\pi \mid R, s, \gamma].
\end{align}

On the other hand,
\begin{align}
     &\sup_{\D\in \Delta(\R[true])} \E{R\sim \D}{\pregret[\pi \mid R, s, \gamma]}\\
     \leq{}& \sup_{\D\in \Delta(\R[true])} \sup_{R\in\supp[\D]} \pregret[\pi \mid R, s, \gamma]\\
     \leq{}& \sup_{\D\in \Delta(\R[true])} \sup_{R\in\R[true]} \pregret[\pi \mid R, s, \gamma]\label{eq:supp-contain}\\
     ={}& \sup_{R\in\R[true]} \pregret[\pi \mid R, s, \gamma].
\end{align}
\Cref{eq:supp-contain} follows because $\supp[\D]\subseteq \R[true]$ by the definition of $\Delta(\R[true])$, the set of all probability distributions over $\R[true]$.

Therefore, the equality of \cref{eq:uncertain-equiv} holds.
\end{proof}

However, \cref{fig:robust-impossible} shows that in non-communicating environments, robustness against $\D$ is too restrictive, even if we assume perfect corrigibility. Therefore, we step away from worst-case regret minimization.

\begin{restatable}[Equal optimal average reward in communicating {\mdp}s]{lem}{eqGain}\label{lem:eq-gain}
If the environment is communicating, then $\forall R\in \rewardSpace, s,s' \in \St: \VfNorm{s,1}=\VfNorm{s',1}$.
\end{restatable}
\begin{proof}
Since the {\mdp} is communicating, $s$ can reach $s'$ with positive probability after at most $\abs{\St}$ timesteps under some policy $\pi$. If an agent following $\pi$ has not reached $s'$ within $\abs{\St}$ timesteps, all states can reach $s'$ with positive probability and so $\pi^\text{HD}$ once again attempts to navigate to $s'$. Since the {\mdp} is finite, there is a state with a minimal (but positive) probability $p$ of reaching $s'$ within $\abs{\St}$ time steps. Because $p$ is positive and minimal, $\pi^\text{HD}$ has probability at most $\prod_{t=1}^\infty (1-p)^t = 0$ of not reaching $s'$ eventually. Apply \cref{lem:prob-1-reach} to conclude that $\VfNorm{s,1}\geq \VfNorm{s',1}$.

The proof for $s'$ reaching $s$ is similar, and so $\VfNorm{s,1}=\VfNorm{s',1}$.
\end{proof}

\begin{restatable}[$\pregret$ is piecewise rational on $\gamma\in  {[0,1]}$]{conjSec}{pwRat}\label{prop:pw-rat-pregret}
\end{restatable}

\section{Varying the reward function distribution}
\begin{restatable}[$\pwrNoDist$, attainable utility distance, and optimality probability are convex over mixture distributions]{prop}{linearCombo}\label{thm:lin-mixture-dist}
Let $\D_1, \D_2$ be two bounded reward function distributions, let $\theta \in [0,1]$, and let $\D' \defeq \theta \D_1 + (1-\theta)\D_2$ be a mixture distribution of the two. Let $s$ be any state and $\gamma\in[0,1]$.
\begin{enumerate}
    \item $\pwrNoDist_{\D'}(s,\gamma) = \theta\pwrNoDist_{\D_1}(s,\gamma) + (1-\theta)\pwrNoDist_{\D_2}(s,\gamma)$.\label{item:pwr-mixture}
    \item $\forall \Delta_1, \Delta_2 \in \Delta(\St):$
    \begin{equation}
        \dauNorm[\D']{\Delta_1, \Delta_2}{\gamma} = \theta\dauNorm[\D_1]{\Delta_1, \Delta_2}{\gamma} + (1-\theta)\dauNorm[\D_2]{\Delta_1, \Delta_2}{\gamma}.
    \end{equation}\label{item:daunorm-mixture}
    \item $\forall F \subseteq \F(s): \optprob[\D']{F,\gamma} =
    \theta \optprob[\D_1]{F,\gamma}+(1-\theta)\optprob[\D_2]{F,\gamma}$.\label{item:opt-mixture}
\end{enumerate}
\end{restatable}
\begin{proof}
\Cref{item:pwr-mixture}: suppose $\gamma \in (0,1)$. Then
\begin{align}
&\pwrNoDist_{\D'}(s,\gamma) \\
    \defeq \,&\E{\rf\sim\D'}{\max_{\f\in \F(s)} \frac{1-\gamma}{\gamma}\prn{\f(\gamma)-\unitvec}^\top \rf}\\
    = \,&\theta \E{\rf\sim\D_1}{\max_{\f\in \F(s)} \frac{1-\gamma}{\gamma}\prn{\f(\gamma)-\unitvec}^\top \rf} + (1-\theta)\E{\rf\sim\D_2}{\max_{\f\in \F(s)} \frac{1-\gamma}{\gamma}\prn{\f(\gamma)-\unitvec}^\top \rf}\\
    = \,&\theta\pwrNoDist_{\D_1}(s,\gamma) + (1-\theta)\pwrNoDist_{\D_2}(s,\gamma).\label{eq:pwr-mixture-done}
\end{align}
Since \cref{eq:pwr-mixture-done} holds for arbitrary $\gamma \in (0,1)$, it must hold in the limits as $\gamma \to 0$ and $\gamma \to 1$; the limits of $\pwr$ exist by \cref{thm:cont-power}. Then \cref{item:pwr-mixture} follows.

Similar logic proves \cref{item:daunorm-mixture} via the linearity of expectation over reward functions.

Consider \cref{item:opt-mixture}. By the definition of optimality probability (\cref{def:prob-opt}), some $\f\in F$ is optimal at discount rate $\gamma$ with probability $\optprob[\D_1]{F,\gamma}$ when $R$ is drawn from $\D_1$ and with probability $\optprob[\D_2]{F,\gamma}$ when $R$ is drawn from $\D_2$. Then by the definition of the mixture distribution $\D'$, the total probability of this event is $\optprob[\D']{F,\gamma} = \theta \optprob[\D_1]{F,\gamma}+(1-\theta)\optprob[\D_2]{F,\gamma}$. This proves \cref{item:opt-mixture}.
\end{proof}

\begin{restatable}[Convexity in the space of probability distributions]{cor}{convexDists}
Let $\D_1, \D_2$ be two bounded reward function distributions, let $\theta \in [0,1]$, and let $\D' \defeq \theta \D_1 + (1-\theta)\D_2$ be a mixture distribution of the two. Let $s$ be any state, let $\gamma\in[0,1]$, and let $k \in \reals$.
\begin{enumerate}
    \item If $\pwrNoDist_{\D_1}(s,\gamma), \pwrNoDist_{\D_2}(s,\gamma)\geq k$, then $\pwrNoDist_{\D'}(s,\gamma) \geq k$.
    \item Let $\Delta_1, \Delta_2 \in \Delta(\St)$. If $\dauNorm[\D_1]{\Delta_1, \Delta_2}{\gamma},\dauNorm[\D_2]{\Delta_1, \Delta_2}{\gamma} \geq k$ then
    \begin{equation*}
        \dauNorm[\D']{\Delta_1, \Delta_2}{\gamma} \geq k.
    \end{equation*}
    \item Let $F \subseteq \F(s)$. If $\optprob[\D_1]{F,\gamma}$, $\optprob[\D_2]{F,\gamma} \geq k$, then $\optprob[\D']{F,\gamma} \geq k$.
\end{enumerate}
\end{restatable}
\begin{proof}
All items follow directly from \cref{thm:lin-mixture-dist}.
\end{proof}

\begin{restatable}[$\pwr$ difference bounded by total variation distance]{prop}{dtvPow}\label{thm:dtv-power}
Let $\D_1,\D_2$ be bounded reward function distributions on $\boundedRewards$.
\begin{equation}
    \abs{\pwrNoDist_{\D_1}(s,\gamma) - \pwrNoDist_{\D_2}(s,\gamma)} \leq \dtv{\D_1,\D_2}.\label{eq:thm-dtv-power}
\end{equation}
\end{restatable}
\begin{proof}
If $\dtv{\D_1,\D_2}=0$, then $\pwrNoDist_{\D_1}(s,\gamma)=\pwrNoDist_{\D_2}(s,\gamma)$ and the statement holds.

Suppose $\dtv{\D_1,\D_2}>0$. Letting $\D_1, \D_2$ respectively correspond to probability measures $F_1, F_2$, consider the finite signed probability measure $F_\text{diff} \defeq F_1 - F_2$. The positive sets of $F_\text{diff}$ are the sets to which $F_1$ assigns more probability; vice versa for the negative sets and $F_2$. By the Hahn decomposition theorem, there exist (non-negative) measures $F^+, F^-$ such that $F_\text{diff}= F^+ - F^-$.

By the fact that $F_1,F_2$ are probability measures with support contained in $\boundedRewards$,
\begin{align}
 F_1(\boundedRewards) -F_2(\boundedRewards)&= 0\\
    &= F_\text{diff}\prn{\boundedRewards}\\
    &= F^+\prn{\boundedRewards}-F^-\prn{\boundedRewards}.\label{eq:equal-measure-decomp}
\end{align}

Therefore, $F^+\prn{\boundedRewards}=F^-\prn{\boundedRewards}$. Furthermore, they both equal $\dtv{\D_1,\D_2}>0$ by the definition of total variation distance for probability measures. Let probability measures $F^+_{\text{renorm}}, F^-_{\text{renorm}}$ be the renormalized versions of the non-negative measures $F^+,F^-$; renormalization is possible because both measures assign finite positive probability to $\boundedRewards$.

Let $\gamma \in (0,1)$ and let $s$ be arbitrary. Let $f(R) \defeq \frac{1-\gamma}{\gamma}\prn{\OptVf{s,\gamma}-R(s)}$.
\begin{align}
    &\abs{\pwrNoDist_{\D_1}(s,\gamma) - \pwrNoDist_{\D_2}(s,\gamma)} \\
    =\,& \abs{\int_{\boundedRewards} f(R) \dif F_1(R) - \int_{\boundedRewards} f(R) \dif F_2(R)}\\
    =\,& \abs{\int_{\boundedRewards} f(R) \prn{\dif F_1(R) - \dif F_2(R)}}\\
    =\,& \abs{\int_{\boundedRewards} f(R) \dif F_\text{diff}(R)}\\
    =\,& \abs{\int_{\boundedRewards} f(R) \prn{\dif F^+(R) - \dif F^-(R)}}\\
    =\,& \abs{F^+\prn{\boundedRewards} \E{R \sim F^+_\text{renorm}}{f(R)} - F^-\prn{\boundedRewards}\E{R \sim F^-_\text{renorm}}{f(R)}}\label{eq:renorm-expect}\\
    =\,& \dtv{\D_1,\D_2} \abs{\E{R \sim F^+_\text{renorm}}{f(R)}-\E{R \sim F^-_\text{renorm}}{f(R)}}\label{eq:dtv-factor}\\
    \leq \, & \dtv{\D_1,\D_2}.\label{eq:dtv-pow-bounds}
\end{align}
\Cref{eq:renorm-expect} follows by the definitions of $F^+_{\text{renorm}}, F^-_{\text{renorm}}$. \Cref{eq:dtv-factor} follows from the fact that $F^+\prn{\boundedRewards}=F^-\prn{\boundedRewards}=\dtv{\D_1,\D_2}>0$. \Cref{eq:dtv-pow-bounds} follows from the fact that $f$ is bounded $[0,1]$, since its domain is $\boundedRewards$. Then \cref{eq:thm-dtv-power} holds.

Since \cref{eq:thm-dtv-power} holds for all $\gamma \in (0,1)$, it also holds in the limits $\gamma \to 0$ and $\gamma \to 1$.
\end{proof}

\begin{restatable}[Improved total variation bound]{conjSec}{sharpTV}
Let $F_{1,v}(v)$ be the optimal value \textsc{cdf} of $\D_1$ at state $s$ and discount rate $\gamma$; similarly define $F_{2,v}(v)$. The bound of \cref{eq:thm-dtv-power} can be improved to \[\dtv{\D_1,\D_2}\max\prn{\int_{F\inv_{1,v}(1-\dtv{\D_1,\D_2})}^1 v\dif F^+_\text{renorm}(v), 1-\int_0^{F\inv_{2,v}(\dtv{\D_1,\D_2})}v \dif F^-_\text{renorm}(v)}.\]
\end{restatable}

The bound of \cref{thm:dtv-power} is sharp; suppose $\D_1$ puts probability $1$ on the all-1 reward function, while $\D_2$ puts probability $1$ on the all-0 reward function. Then in any $\mdp$, at any $s$ and for any $\gamma \in [0,1]$,
\begin{equation}
    \abs{\pwrNoDist_{\D_1}(s,\gamma) - \pwrNoDist_{\D_2}(s,\gamma)}= \abs{1-0}=1=\dtv{\D_1,\D_2}.
\end{equation}

\begin{restatable}[$\pwr$ difference bounded by Wasserstein 1-distance]{thm}{wassPow}\label{thm:wass-power}
Let $\D_1,\D_2$ be any bounded reward function distributions.
\begin{equation}
    \abs{\pwrNoDist_{\D_1}(s,\gamma) - \pwrNoDist_{\D_2}(s,\gamma)} \leq \wasserstein{\D_1,\D_2}.\label{eq:thm-wass-power}
\end{equation}
\end{restatable}
\begin{proof}
Suppose $\gamma \in (0,1)$ and once again let $f(R)\defeq\frac{1-\gamma}{\gamma}\prn{\OptVf{s,\gamma}-R(s)}$. We first show that $f$ has Lipschitz constant $1$. Let $R_a, R_b \in \rewardVS$; without loss of generality, suppose $\OptVf[R_a]{s,\gamma} \geq \OptVf[R_b]{s,\gamma}$.
\begin{align}
    \abs{f(R_a)-f(R_b)}&=\frac{1-\gamma}{\gamma}\abs{\max_{\f_a\in\F(s)}\prn{\f_a(\gamma)-\unitvec}^\top\rf_a - \max_{\f_b\in\F(s)}\prn{\f_b(\gamma)-\unitvec}^\top\rf_b}\\
    &\leq \frac{1-\gamma}{\gamma}\abs{\max_{\f_a\in\F(s)}\prn{\f_a(\gamma)-\unitvec}^\top\rf_a - \prn{\f_a(\gamma)-\unitvec}^\top\rf_b}\label{eq:decrease-diff}\\
    &= \frac{1-\gamma}{\gamma}\abs{\max_{\f_a\in\F(s)}\prn{\f_a(\gamma)-\unitvec}^\top (\rf_a - \rf_b)}\\
    &\leq \frac{1-\gamma}{\gamma} \lone{\f_a(\gamma) - \unitvec} \lone{\rf_a-\rf_b}\label{eq:cs-wasserstein}\\
    &= \lone{\rf_a-\rf_b}.\label{eq:lip-1}
\end{align}
\Cref{eq:decrease-diff} follows because $\f_b$ was optimal for $\rf_b$, and so $(\f_a(\gamma)-\unitvec)^\top \rf_b \leq (\f_b(\gamma)-\unitvec)^\top \rf_b$. \Cref{eq:cs-wasserstein} is a simple application of the Cauchy-Schwarz inequality. \Cref{eq:lip-1} follows because \cref{prop:visit-dist-prop} shows that $\forall \f \in \F(s): \lone{\f(\gamma)}=\geom$, and $\unitvec$ is a unit vector. Therefore, $f$ has Lipschitz constant $1$.

We now show the desired inequality.
\begin{align}
    &\abs{\pwrNoDist_{\D_1}(s,\gamma) - \pwrNoDist_{\D_2}(s,\gamma)} \\
    =\,& \abs{\int_{\boundedRewards} f(R) \prn{\dif F_1(R) - \dif F_2(R)}}\\
    =\,& \abs{\int_{\boundedRewards} f(R) \dif \, (F_1 - F_2)(R)}\\
    \leq \, & \abs{\sup_{\substack{f_\text{lip}\,:\,\rewardVS \, \to \, \reals,\\f_\text{lip} \text{ has minimal Lipschitz constant $\leq 1$}}} \int_{\boundedRewards} f_\text{lip}(R) \dif \, (F_1 - F_2)(R)}\label{eq:lip-f-w1}\\
    =\,& W_1(\D_1,\D_2).\label{eq:dual-wass-formulation}
\end{align}
\Cref{eq:lip-f-w1} follows because $f$ has Lipschitz constant $1$ (and so its minimal constant is at most 1). \Cref{eq:dual-wass-formulation} follows by the dual formulation of Wasserstein 1-distance (which is applicable since both distributions have bounded support).
\end{proof}

\begin{restatable}[Optimality probability difference bounded by total variation distance]{thm}{dtvProb}\label{thm:dtv-opt-prob}
Let $\D_1,\D_2$ be any reward function distributions. Let $F \subseteq \F(s)$.
\begin{equation}
    \abs{\optprob[\D_1]{F,\gamma} - \optprob[\D_2]{F,\gamma}} \leq \dtv{\D_1,\D_2}.\label{eq:dtv-opt-prob}
\end{equation}
\end{restatable}
\begin{proof} Let $\D_1$ and $\D_2$ have probability measures $F_1$ and $F_2$.
\begin{align}
    \abs{\optprob[\D_1]{F,\gamma} - \optprob[\D_2]{F,\gamma}}&\defeq \abs{F_1\prn{\opt[F(\gamma)\geq \F(s,\gamma)]} - F_2\prn{\opt[F(\gamma)\geq \F(s,\gamma)]}} \\
    &\leq \sup_{B \in \mathscr{B}\prn{\rewardVS}} \abs{F_1\prn{B} - F_2\prn{B}}\label{eq:borel-opt-supp}\\
    &\eqdef \dtv{\D_1, \D_2}.
\end{align}
\Cref{eq:borel-opt-supp} follows because $\opt[F(\gamma)\geq \F(s,\gamma)]=\bigcup_{\f \in F} \opt$ is the finite union of closed sets (\cref{geom-prop}), and therefore $\opt[F(\gamma)\geq \F(s,\gamma)]$ is a Borel set.
\end{proof}

However,  \cref{fig:no-wasserstein-opt-prob} shows that optimality probability cannot be bounded by Wasserstein distance.

\begin{figure}[ht]
    \centering
     \begin{tikzpicture}[mygraph]
        %nodes
            \node[blue] (1) {$1$};
            \node[black] (2) [right of=1] {$2$};
        %arrows
            \path[->] (1) edge (2);
            \path[->] (1) edge [looseness=5, out=180, in=90] node[midway,left] {\texttt{stay}} (1);
            \path[->] (2) edge [looseness=5, out=0, in=90] (2);
    \end{tikzpicture}
    \caption[Optimality probability difference cannot be bounded by Wasserstein distance]{Let $\epsilon > 0$. If $\D_1$ assigns probability $1$ to $s \mapsto \frac{\epsilon}{2} \indic{s=\col{blue}{s_1}}$ and $\D_2$ assigns probability $1$ to $s \mapsto \frac{\epsilon}{2} \indic{s=s_2}$, then $W_1(\D_1,\D_2)=\epsilon$. However, $\optprob[\D_1]{\col{blue}{s_1},\texttt{stay},.5}=1$ while $\optprob[\D_2]{\col{blue}{s_1},\texttt{stay},.5}=0$.}
    \label{fig:no-wasserstein-opt-prob}
\end{figure}

Taken together, these results show that strict $\pwr$-seeking and optimality probability inequality holds within a $\dtv{}$ neighborhood in the space of reward function distributions, and that one action seeks strictly more $\pwr$ than another action in a 1-Wasserstein neighborhood of reward function distributions. In particular, the strict inequality conditions of \cref{graph-options} and \cref{rsdIC} hold within $\dtv{}$ neighborhoods of distributions which are finite mixtures of bounded continuous \textsc{iid} reward function distributions.

\begin{restatable}[Close $\D_1$, $\D_2$ induce similar $\dauNorm{}{}$ metrics]{conjSec}{conjEMDauDist}
Let $\D_1, \D_2$ be bounded reward function distributions. $\sup_{\Delta_a, \Delta_b \in \Delta(\St)} \abs{\dauNorm[\D_1]{\Delta_a,\Delta_b}{\gamma} - \dauNorm[\D_2]{\Delta_a,\Delta_b}{\gamma}} \leq W_1(\D_1, \D_2)$, where $W_1$ is the first Wasserstein distance on $\rewardVS$.
\end{restatable}

\subsection{Distributional transformations}

Positive affine transformation of $\D$ allows generalization of our results to other bounds, as optimal policy is invariant to positive affine transformation of the reward function. \Cref{item:dau-pos-affine} of \cref{prop:aff-transform} can be viewed as proving \citet{turner2020conservative}'s proposition 4, in the limit of infinitely many draws from the auxiliary reward function distribution $\D'$.

\begin{restatable}[How positive affine transformation affects optimality probability, $\pwrNoDist$, and normalized \textsc{au} distance]{prop}{posAffTransfProp}\label{prop:aff-transform}
Let $\gamma \in [0,1]$ and let $\D'$ be any bounded reward function distribution. Let $m>0,b\in\reals$ and let $\mathbf{b}\in\rewardVS$ be the ones vector times the scalar $b$. $m\D'+\mathbf{b}$ is the pushforward distribution over reward functions formed by applying the positive affine transformation $f(\rf)\defeq m\rf + \mathbf{b}$ to $\D'$.
\begin{enumerate}
    \item Let $F\subseteq \F(s)$. $\optprob[m\D'+\mathbf{b}]{F,\gamma}=\optprob[\D']{F,\gamma}$.\label{item:prob-pos-affine}
    \item $\pwr[s,\gamma][m\D'+\mathbf{b}]=m\cdot \pwr[s,\gamma][\D']+b$.\label{item:power-pos-affine}
    \item Let $\Delta_1,\Delta_2\in\Delta(\St)$. $\dauNorm[m\D'+\mathbf{b}]{\Delta_1,\Delta_2}{\gamma}=m\cdot \dauNorm[\D']{\Delta_1,\Delta_2}{\gamma}$.\label{item:dau-pos-affine}
\end{enumerate}
\end{restatable}
\begin{proof}
Let $\D'$ have probability measure $F'$ and let $m\D' + \mathbf{b}$ have probability measure $F'_\text{aff}$.

\Cref{item:prob-pos-affine}: let $\gamma\in(0,1)$. By \cref{def:prob-opt},
\begin{align}
\optprob[m\D'+\mathbf{b}]{F,\gamma}&\defeq \prob[R\sim m\D'+\mathbf{b}]{\exists \f^\pi \in F: \pi\in\optPi}\\
&=\prob[R\sim m\D'+\mathbf{b}]{\exists \f^\pi \in F: \pi\in\optPi[m\inv (R-b),\gamma]}\label{eq:invariant-pos-aff}\\
&= \prob[R'\sim \D']{\exists \f^\pi \in F: \pi\in\optPi[R',\gamma]}\\
&\eqdef \optprob[\D']{F,\gamma}.
\end{align}

\Cref{eq:invariant-pos-aff} holds because optimal policy sets are invariant to positive affine transformation $m\inv (R-b)$ of the reward function.

Since this equality holds for all $\gamma\in(0,1)$, it holds in the limits $\gamma \to 0$ and $\gamma \to 1$ as well.

\Cref{item:power-pos-affine}: let $\gamma \in (0,1)$.
\begin{align}
    &\phantom{=}\hspace{5pt}\vavg[s,\gamma][m\D'+\mathbf{b}]\\
    &=\int_{\rewardVS} \OptVf[R]{s,\gamma} \dF[R][F'_\text{aff}]\\
    &=\int_{\rewardVS} \OptVf[R]{s,\gamma} \dF[m\inv(R - b)][F']\label{eq:pushforward-vavg}\\
    &=\int_{\rewardVS} \OptVf[mR'+b]{s,\gamma} \dF[R'][F']\label{eq:vavg-cov}\\
    &=\int_{\rewardVS} m\OptVf[R']{s,\gamma}+\geom[b] \dF[R'][F']\label{eq:vavg-transform}\\
    &=m\cdot\vavg[s,\gamma][\D']+\geom[b].\label{eq:conclusion-vavg-transform}
\end{align}

\Cref{eq:pushforward-vavg} follows by the definition of a pushforward measure. \Cref{eq:vavg-cov} follows by substituting $R'\defeq m\inv(R - b)$. \Cref{eq:vavg-transform} follows because optimal policies are invariant to positive affine transformations.

\begin{align}
    \pwr[s,\gamma][m\D'+\mathbf{b}]&=\dfrac{1-\gamma}{\gamma}\prn{\vavg[s,\gamma][m\D'+\mathbf{b}]-\E{R\sim m\D'+\mathbf{b}}{R(s)}}\label{eq:pwr-id-transform}\\
    &= \dfrac{1-\gamma}{\gamma}\prn{m\vavg[s,\gamma][\D']+\geom[b]-m\E{R\sim \D'}{R(s)}-b}\label{eq:apply-conclusion}\\
    &=\dfrac{1-\gamma}{\gamma}\prn{m\vavg[s,\gamma][\D']-m\E{R\sim \D'}{R(s)}+\geom[b\gamma]}\\
    &=m\cdot \pwr[s,\gamma][\D']+b.\label{eq:pwr-id-transform-2}
\end{align}

\Cref{eq:pwr-id-transform} and \cref{eq:pwr-id-transform-2} follow by \cref{lem:power-id} since $\gamma \in (0,1)$. \Cref{eq:apply-conclusion} follows by \cref{eq:conclusion-vavg-transform} and the linearity of expectation.

Finally, \cref{eq:pwr-id-transform-2} holds in the limits as $\gamma\to 0$ and $\gamma\to 1$.

\Cref{item:dau-pos-affine}: let $\gamma \in [0,1)$.
\begin{align}
    &\phantom{=}\hspace{5pt}\dau[m\D'+\mathbf{b}]{\Delta, \Delta'}\\
    &=\int_{\rewardVS} \abs{\E{s \sim \Delta}{\OptVf{s,\gamma}}-\E{ s' \sim\Delta'}{\OptVf{s',\gamma}}} \dF[R][F'_\text{aff}]\\
    &=\int_{\rewardVS} \abs{\E{s \sim \Delta}{\OptVf{s,\gamma}}-\E{ s' \sim\Delta'}{\OptVf{s',\gamma}}} \dF[m^{-1}(R - b)][F']\label{eq:pushforward}\\
    &=\int_{\rewardVS} \abs{\E{s \sim \Delta}{\OptVf[mR'+b]{s,\gamma}}-\E{ s' \sim\Delta'}{\OptVf[mR'+b]{s',\gamma}}} \dF[R'][F']\label{eq:dau-cov}\\
    &=\int_{\rewardVS} \abs{\E{s \sim \Delta}{m\OptVf[R']{s,\gamma}+\geom[b]}-\E{ s' \sim\Delta'}{m\OptVf[R']{s',\gamma}+\geom[b]}} \dF[R'][F']\label{eq:invariant-transform}\\
    &=m\cdot\dau[\D']{\Delta, \Delta'}.\label{eq:dau-conclusion}
\end{align}

\Cref{eq:pushforward} follows by the definition of a pushforward measure. \Cref{eq:dau-cov} follows via substitution $R'\defeq m^{-1}(R-C)$. \Cref{eq:invariant-transform} follows because optimal policy is invariant to positive affine transformation of the reward function.

\begin{align}
    \dauNorm[m\D'+\mathbf{b}]{\Delta_1,\Delta_2}{\gamma}&\defeq \lim_{\gamma^*\to \gamma} (1-\gamma^*) \dau[m\D'+\mathbf{b}]{\Delta, \Delta'}\\
    &= m\cdot \lim_{\gamma^*\to \gamma} (1-\gamma^*)\dau[\D']{\Delta, \Delta'}\label{eq:apply-dau-conclusion}\\
    &=m\cdot\dauNorm[\D']{\Delta_1,\Delta_2}{\gamma}.\label{eq:dau-norm-conclusion}
\end{align}
\Cref{eq:apply-dau-conclusion} holds by \cref{eq:dau-conclusion}.

\Cref{eq:dau-norm-conclusion} shows the desired result for all $\gamma\in[0,1)$. \Cref{eq:dau-norm-conclusion} holds in the limit $\gamma\to 1$, and so the relationship holds for all $\gamma\in[0,1]$.
\end{proof}

\subsubsection{$\pwrNoDist$ inequalities under certain distributional transformations}

\begin{restatable}[How non-affine transformations affect $\pwrNoDist$]{prop}{nonAffTransf}\label{thm:non-aff-transf-pow}
Let $\gamma\in(0,1)$, let $s$ be a state, and let $\D'$ be any bounded reward function distribution. Suppose $g :\rewardVS \to \rewardVS$ is measurable and is such that $\forall \rf \in \supp[\D']:$
\begin{equation}
    \forall \f \in \Fnd(s): \f(\gamma)^\top \rf \geq \f(\gamma)^\top g(\rf).\label{eq:lower-reward}
\end{equation}

Then $\pwr[s,\gamma][\D']\geq \pwr[s,\gamma][g(\D')]$. This inequality is strict iff \cref{eq:lower-reward} is strict for a subset of $\supp[\D']$ with positive measure under $\D'$.

A similar statement holds when all of the above inequalities are flipped.
\end{restatable}
\begin{proof}
Let $\D'$ have probability measure $F'$. Let $g(\D')$ be the pushforward probability distribution induced by applying measurable function $g$ to $\D'$, and let $F'_g$ be its probability measure.
\begin{align}
    \vavg[s,\gamma][\D']&\defeq \int_{\rewardVS} \max_{\f \in \F(s)} \f(\gamma)^\top \rf \dF[\rf][F']\\
    &= \int_{\supp[\D']} \max_{\f \in \Fnd(s)} \f(\gamma)^\top \rf \dF[\rf][F']\label{eq:vavg-restrict-fnd}\\
    &\geq \int_{\supp[\D']} \max_{\f \in \Fnd(s)} \f(\gamma)^\top g(\rf) \dF[\rf][F']\label{eq:g-dec}\\
    &= \int_{\supp[g(\D')]} \max_{\f \in \Fnd(s)} \f(\gamma)^\top \rf' \dF[g\inv(\rf')][F']\label{eq:g-cov}\\
    &= \int_{\supp[g(\D')]} \max_{\f \in \Fnd(s)} \f(\gamma)^\top \rf' \dF[\rf'][F'_g]\label{eq:g-push}\\
    &= \vavg[s,\gamma][g(\D')].\label{eq:vavg-restrict-g-fnd}
\end{align}
\Cref{eq:vavg-restrict-fnd} and \cref{eq:vavg-restrict-g-fnd} follow by \cref{cor:opt-vf-restrict-fnd} and the definition of $\supp[\D']$. \Cref{eq:g-dec} follows by our assumptions on $g$. \Cref{eq:g-cov} holds by the substitution $\rf'\defeq g(\rf)$. \Cref{eq:g-push} follows by the definition of a pushforward measure $F'_g$.

Integration is invariant to strict optimal value decrease on a zero-measure subset of $\supp[\D']$, but not to strict optimal value decrease on a zero-measure subset of $\supp[\D']$. Therefore, \cref{eq:g-dec} is strict iff \cref{eq:lower-reward} is strict for a subset of $\supp[\D']$ with positive measure under $\D'$.

Since $\pwr[s,\gamma][\D']=\dfrac{1-\gamma}{\gamma}\prn{V^*_{\D'}(s,\gamma)-\E{R\sim \D'}{R(s)}}$ by \cref{lem:power-id},
\begin{equation*}
    \pwr[s,\gamma][\D']\geq \pwr[s,\gamma][g(\D')],
\end{equation*}
with strict inequality iff \cref{eq:lower-reward} is strict for a subset of $\supp[\D']$ with positive measure under $\D'$.

The proof for reward-increasing $g$ follows similarly.
\end{proof}

\begin{restatable}[How non-affine transformations affect optimality probability]{prop}{nonAffTransfOptProb}\label{thm:non-aff-transf-opt-prob}
Let $\gamma\in(0,1)$, let $s$ be a state, let $F\subseteq \F(s)$, and let $\D'$ be any bounded reward function distribution. Suppose $g :\rewardVS \to \rewardVS$ is measurable and is such that $\forall \rf \in \supp[\D']:$
\begin{align}
    \forall \f \in F&: \f(\gamma)^\top \rf \geq \f(\gamma)^\top g(\rf),\label{eq:lower-reward-opt}\\
    \forall \f' \in \Fnd(s)\setminus F&: \f'(\gamma)^\top \rf \leq \f'(\gamma)^\top g(\rf).\label{eq:greater-rest-reward-opt}
\end{align}

Then $\optprob[\D']{F,\gamma}\geq \optprob[g(\D')]{F,\gamma}$. Equality holds if \cref{eq:lower-reward-opt} and \cref{eq:greater-rest-reward-opt} are equalities for a subset of $\supp[\D']$ with measure $1$ under $\D'$.

A similar statement holds when all of the above inequalities are flipped.
\end{restatable}
\begin{proof}
Let $\D'$ have probability measure $F'$. Let $g(\D')$ be the pushforward probability distribution induced by applying measurable function $g$ to $\D'$, and let $F'_g$ be its probability measure.
\begin{align}
    \optprob[\D']{F,\gamma}&= \int_{\rewardVS} \indic{\max_{\f\in F} \f(\gamma)^\top \rf = \max_{\f'\in\Fnd(s)} \f'(\gamma)^\top \rf} \dF[\rf][F']\label{eq:opt-prob-id-non-aff}\\
    &= \int_{\supp[\D']} \indic{\max_{\f\in F} \f(\gamma)^\top \rf = \max_{\f'\in\Fnd(s)\setminus F} \f'(\gamma)^\top \rf} \dF[\rf][F']\label{eq:exclude-F}\\
    &\geq \int_{\supp[\D']} \indic{\max_{\f\in F} \f(\gamma)^\top g(\rf) = \max_{\f'\in\Fnd(s)\setminus F} \f'(\gamma)^\top g(\rf)} \dF[\rf][F']\label{eq:g-apply-opt}\\
    &= \int_{\supp[g(\D')]} \indic{\max_{\f\in F} \f(\gamma)^\top \rf' = \max_{\f'\in\Fnd(s)\setminus F} \f'(\gamma)^\top \rf'} \dF[g\inv(\rf')][F']\label{eq:g-subst-opt-prob}\\
    &= \int_{\supp[g(\D')]} \indic{\max_{\f\in F} \f(\gamma)^\top \rf' = \max_{\f'\in\Fnd(s)\setminus F} \f'(\gamma)^\top \rf'} \dF[\rf'][F'_g]\label{eq:g-push-opt-prob}\\
    &= \int_{\supp[g(\D')]} \indic{\max_{\f\in F} \f(\gamma)^\top \rf' = \max_{\f'\in\Fnd(s)} \f'(\gamma)^\top \rf'} \dF[\rf'][F'_g]\label{eq:g-reinclude-opt-prob}\\
    &= \optprob[g(\D')]{F,\gamma}.\label{eq:g-opt-prob-id}
\end{align}
\Cref{eq:opt-prob-id-non-aff} and \cref{eq:g-opt-prob-id} follow by \cref{lem:opt-prob-id}. \Cref{eq:exclude-F} and \cref{eq:g-reinclude-opt-prob} follow by \cref{lem:invariance-opt-prob} (\ref{item:help-exclusion}) and the definition of $\supp[\D']$. \Cref{eq:g-apply-opt} follows because our assumptions on $g$ imply that
\[\forall \rf \in \supp[\D']:\indic{\max_{\f\in F} \f(\gamma)^\top \rf = \max_{\f'\in\Fnd(s)\setminus F} \f'(\gamma)^\top \rf}\geq \indic{\max_{\f\in F} \f(\gamma)^\top g(\rf) = \max_{\f'\in\Fnd(s)\setminus F} \f'(\gamma)^\top g(\rf)}.\] \Cref{eq:g-subst-opt-prob} follows by the substitution $\rf'\defeq g(\rf)$. \Cref{eq:g-push-opt-prob} follows by the definition of the pushforward measure $F_g'$.

Integration is invariant to strict optimal value decrease on a zero-measure subset of $\supp[\D']$. Therefore, \cref{eq:g-dec} is an equality if \cref{eq:lower-reward-opt} and \cref{eq:greater-rest-reward-opt} are equalities for a subset of $\supp[\D']$ with measure $1$ under $\D'$.

A similar proof follows when the theorem statement's inequalities are flipped.
\end{proof}

\begin{remark}
Unlike \cref{thm:non-aff-transf-pow}, \cref{thm:non-aff-transf-opt-prob} does not have a strict inequality if-and-only-if. This is because even if $g$ strictly increases return for $\f\in F$, that increase may be insufficient to change the optimality status of $\f\in F$ for any $\rf\in\supp[\D']$.
\end{remark}

\Cref{thm:non-aff-transf-pow} and \cref{thm:non-aff-transf-opt-prob} significantly expand the initial $\pwr$\hyp{}seeking results to $g(\D)$ which distribute reward independently and non-identically across states.

Reconsider \cref{graph-options}. For \eg{} the max-entropy $\D$ over $[0,1]^{\abs{\St}}$, at $s'$, action $a$ is strictly $\pwr$-seeking and more probable under optimality compared to $a'$. \Cref{thm:non-aff-transf-pow} and \cref{thm:non-aff-transf-opt-prob} show that by \eg{} doubling reward at green states and zeroing the reward at two of the red states (call this distribution $g(\D)$), the same $\pwr[][g(\D)]$-seeking and optimality probability statements hold.

\subsection{$\Fnd$ symmetry}

\begin{restatable}[$\Fnd(s)$ symmetry group]{definition}{symGroupFnd}\label{def:sym-group-fnd}
Let $S_n$ be the permutation group on $n$ elements. For any state $s$,
\begin{equation}
    \symmFnd \defeq \set{\phi \in \mdpPermGroup \mid \phi\cdot\Fnd(s)=\Fnd(s)}.
\end{equation}
\end{restatable}

\begin{restatable}[$\symmFnd$ is a subgroup of $\mdpPermGroup$]{prop}{subgroupFnd}
\end{restatable}
\begin{proof} Let $s$ be any state. By \cref{def:sym-group-fnd}, $\symmFnd\subseteq \mdpPermGroup$. We show that $\symmFnd$ satisfies the group axioms under permutation composition.
\paragraph*{Identity.} Let $\phi_{id}\in \mdpPermGroup$ be the identity permutation. Then clearly $\phi_{id}\prn{\Fnd(s)}=\Fnd(s)$, so $\phi_{id} \in \symmFnd$.

\paragraph*{Composition.} Let $\phi_1, \phi_2 \in \mdpPermGroup$. Then
    \begin{align}
    (\phi_1\circ \phi_2)\prn{\Fnd(s)}&=\phi_1\prn{\phi_2\prn{\Fnd(s)}}\\
    &=\phi_1\prn{\phi_2\prn{\Fnd(s)}}\\
    &=\phi_1\prn{\Fnd(s)}\\
    &=\Fnd(s),
    \end{align}
   and so $(\phi_1\circ \phi_2)\in \symmFnd$.

\paragraph*{Inverse.} Let $\phi\in\symmFnd$. $\phi\cdot\Fnd(s)=\Fnd(s)$ implies that $\Fnd(s)=\phi\inv\prn{\Fnd(s)}$, and so $\phi\inv\in \symmFnd$.

Therefore, $\symmFnd$ is a group.
\end{proof}

\begin{restatable}[If $\phi\cdot\Fnd(s)=\Fnd(s')$, then $\phi(s)=s'$]{lem}{sendSPerm}\label{lem:send-s-perm}
\end{restatable}
\begin{proof}
$\phi\cdot\Fnd(s)=\Fnd(s')$ implies that
\begin{align}
    \set{\lim_{\gamma\to 0}\permute \f(\gamma) \mid \f \in \Fnd(s)}=\set{\lim_{\gamma\to 0} \f(\gamma) \mid \f \in \Fnd(s')}.
\end{align}

But the right-hand side equals $\set{\unitvec[s']}$ by the definition of a state visit distribution function $\f\in\F(s')$ (\cref{def:visit}). The left-hand side also equals $\set{\unitvec[s']}$, implying that $\permute \unitvec=\unitvec[s']$ and so $\phi(s)=s'$.
\end{proof}

\begin{restatable}[$\pwr$ across certain distributional symmetries]{prop}{pwrInvariant}\label{thm:pwr-invariant-symmetry}
Let $s,s'$ be states. If $\exists \phi \in \mdpPermGroup$ such that $\phi\cdot\Fnd(s)=\Fnd(s')$, then $\pwr[s,\gamma][\Dbd]=\pwr[s',\gamma][\phi\cdot \Dbd]$.
\end{restatable}
\begin{proof}
Let $\gamma^*\in(0,1),F\defeq \set{\f(\gamma^*)-\unitvec\mid \f\in \Fnd(s)},F'\defeq \set{\f(\gamma^*)-\unitvec[s']\mid \f\in \Fnd(s')}$.
\begin{align}
    &\pwr[s,\gamma][\Dbd] \\
    &\defeq \lim_{\gamma^*\to\gamma} \frac{1-\gamma^*}{\gamma^*}\E{\rf \sim \Dbd}{\max_{\f\in\F(s)} \prn{\f(\gamma^*)-\unitvec}^\top\rf}\\
    &=\lim_{\gamma^*\to\gamma} \frac{1-\gamma^*}{\gamma^*} \E{\rf \sim \Dbd}{\max_{\f\in\Fnd(s)} \prn{\f(\gamma^*)-\unitvec}^\top\rf}\label{eq:restrict-fnd-symm}\\
    &= \lim_{\gamma^*\to\gamma} \frac{1-\gamma^*}{\gamma^*}\E{\rf \sim \phi\cdot \Dbd}{\max_{\f'\in\phi\cdot F} \f'^\top \rf}\label{eq:sym-fix-state}\\
    &= \lim_{\gamma^*\to\gamma} \frac{1-\gamma^*}{\gamma^*}\E{\rf \sim \phi\cdot \Dbd}{ \max_{\f'\in F'} \f'^{\top}\rf}\label{eq:similar-visit-sym}\\
    &\eqdef \pwr[s',\gamma][\phi\cdot \Dbd].
\end{align}

\Cref{eq:restrict-fnd-symm} follows because \cref{cor:opt-vf-restrict-fnd} shows that optimal value is invariant to restriction to $\Fnd$. Let $g:\reals \to \reals$ be the identity function, and let $f(B\mid \Dbd)\defeq \E{\rf \sim \Dbd}{g\prn{\max_{\f \in B} \f^\top \rf}}$. \Cref{eq:sym-fix-state} follows by applying \cref{lem:helper-perm} to conclude that $f(F\mid \Dbd)=f(\phi\cdot F\mid \phi\cdot \Dbd)$.

$\phi\cdot F=\set{\permute\f(\gamma^*) - \permute\unitvec\mid \f \in \Fnd(s)}=F'$ by the assumptions on $\phi$ and the fact that $\phi(s)=s'$ by \cref{lem:send-s-perm}. Therefore, \cref{eq:similar-visit-sym} follows.

Since this holds for all $\gamma\in(0,1)$, it holds in the limits $\gamma\to 0$ and $\gamma\to 1$ as well.
\end{proof}

\begin{figure}[ht]
    \centering
    \captionsetup[subfigure]{labelformat=empty}
    \subfloat[]{
    \begin{tikzpicture}[mygraph]
        %nodes
            \node[blue] (1) {$1$};
            \node[black] (2) [right of=1] {};
            \node[black] (3) [below of=1] {};
        %arrows
            \path[->] (1) edge (2);
            \path[->] (1) edge (3);
            \path[->] (2) edge [looseness=5, out=0, in=90] (2);
            \path[->] (3) edge [looseness=5, out=-90, in=0] (3);
    \end{tikzpicture}
     }\qquad\quad\subfloat[]{
     \begin{tikzpicture}[mygraph,shorten <=1pt]
        %nodes
            \node[blue] (1) {$2$};
            \node[black] (2) [right of=1] {};
            \node[black] (3) [below of=1] {};
        %arrows
            \path[->] (1) edge (2);
            \path[->] (1) edge (3);
            \path[<->] (2) edge [color=lightgray] (3);
            \path[->] (2) edge [looseness=5, out=0, in=90] (2);
            \path[->] (3) edge [looseness=5, out=-90, in=0] (3);
    \end{tikzpicture}
    }\qquad\quad\subfloat[]{
     \begin{tikzpicture}[mygraph,shorten <=1pt]
        %nodes
            \node[blue] (1) {$3$};
            \node[black] (2) [right of=1] {$4$};
            \node[black] (3) [below of=1] {$5$};
        %arrows
            \path[->] (1) edge (2);
            \path[->] (1) edge (3);
            \path[<->] (2) edge [color=red]  node[midway,right] {$\phi$}(3);
            \path[->] (2) edge [looseness=5, out=0, in=90] (2);
            \path[->] (3) edge [looseness=5, out=-90, in=0] (3);
    \end{tikzpicture}
    }
    \caption[Similarity of non-dominated visit distributions allows concluding $\pwrNoDist$ invariance under those permutations]{By \cref{thm:pwr-invariant-symmetry}, $\forall\gamma\in[0,1]:\pwr[\col{blue}{s_1},\gamma][\Diid]=\pwr[\col{blue}{s_2},\gamma][\Diid]$. In general, $\pwr[\col{blue}{s_3},\gamma]=\pwr[\col{blue}{s_3},\gamma][\col{red}{\phi}\cdot \Dbd]$.
    \label{fig:sim-pwr-nd}}
\end{figure}

\subsection{State similarity}
\begin{restatable}[State similarity]{definition}{similarState}\label{def:similar-state}
State $s$ is \emph{similar} to $s'$ if there exists a stochastic model isomorphism $\phi$ such that $\phi(s)=s'$. If all states are similar, the model is \emph{vertex transitive}.
\end{restatable}

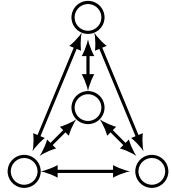
\begin{figure}
\centering
 \begin{tikzpicture}[mygraph,shorten >=2pt,node distance=1.2cm]
        %nodes
            \node[black] (1) {};
            \node[black] (2) [above of=1] {};
            \node[black] (3) [below left of=1] {};
            \node[black] (4) [below right of=1] {};
        %arrows
            \path[<->] (1) edge (2);
            \path[<->] (1) edge (3);
            \path[<->] (1) edge (4);
            \path[<->] (2) edge (3);
            \path[<->] (2) edge (4);
            \path[<->] (3) edge (4);
    \end{tikzpicture}
    \caption[If the {\mdp} model has vertex transitivity, $\pwrNoDist_{\Diid}$ is constant]{The tetrahedral graph is vertex transitive.}\label{fig:tetra-v-trans}
\end{figure}

When the dynamics are deterministic, \cref{def:similar-state} reduces to the standard graph\hyp{}theoretic vertex similarity, as shown in \cref{fig:tetra-v-trans}.

\begin{restatable}[State similarity criterion]{cor}{permSimilar}\label{perm-similar}
$s$ and $s'$ are similar via permutation $\phi$ iff $\F(s')=\phi\cdot\F(s)$.
\end{restatable}
\begin{proof}
Apply \cref{model-iso}.
\end{proof}

\begin{restatable}[Similar states have similar non-dominated visit distribution functions]{lem}{similarFND}\label{F-similar-Fnd-similar}
If $s$ and $s'$ are similar via permutation $\phi$, then $\Fnd(s')=\phi\cdot\Fnd(s)$.
\end{restatable}
\begin{proof}
State similarity implies visit distribution function similarity (\cref{perm-similar}), so $\F(s')=\set{\permute \f \mid \f \in \F(s)}$. Without loss of generality, suppose that $\permute\f$ is non-dominated at $s'$; this implies $\permute\f$ is strictly optimal for reward function $R$ at discount rate $\gamma$. Then
\begin{align}
    (\permute\f(\gamma))^\top \rf &> \max_{\f' \in \F(s')\setminus \set{\permute\f}} \f'(\gamma)^\top\rf.\\
    \f(\gamma)^\top (\permute[\phi^{-1}]\rf) &> \max_{\f' \in \F(s)\setminus \set{\f}} \f'(\gamma)^\top(\permute[\phi^{-1}]\rf).
\end{align}
Then $\f$ is strictly optimal at state $s$ for reward function $\permute[\phi^{-1}]\rf$ at discount rate $\gamma$, and thus $\f\in\Fnd(s)$.
\end{proof}

\begin{restatable}[Similar states have equal $\protect\pwr$]{prop}{powSimilar}\label{pow-similar}
If $s$ and $s'$ are similar, $\forall\gamma\in[0,1]:\pwr[s,\gamma][\Diid]=\pwr[s',\gamma][\Diid]$.
\end{restatable}
\begin{proof}
Suppose $\phi\cdot \F(s)=\F(s')$. Apply \cref{F-similar-Fnd-similar} to conclude that $\phi\cdot \Fnd(s)=\Fnd(s')$. Then apply \cref{thm:pwr-invariant-symmetry} to conclude the desired equality (with identical distribution ensuring that $\phi\cdot \Diid=\Diid$).
\end{proof}

\begin{restatable}[Vertex transitivity implies $\pwrNoDist_{\Diid}$ is equal everywhere]{cor}{vTrans}\label{vtrans}
If the model is vertex transitive, then all states have equal $\pwr[][\Diid]$.
\end{restatable}

\Cref{fig:sim-pwr-nd} shows that non-dominated similarity (\cref{thm:pwr-invariant-symmetry}) allows us to conclude $\pwr$ equality in a greater range of situations than does \cref{pow-similar}.

\subsection{Strong visitation distribution set similarity}
\begin{restatable}[Strong visitation distribution set similarity]{definition}{similarVisitFn}\label{def:strong-similar-visit-fn}
Let $F,F'\in\F(s)$. We say that $F$ and $F'$ are \textit{strongly similar} if they are similar via a permutation $\phi\in\symmFnd$.
\end{restatable}

Strong visitation distribution similarity depends on the totality of $\Fnd(s)$. We aren't just interested in whether $F'=\phi\cdot F$ — we want to know whether $F$ and $F'$ ``interact with $\Fnd(s)$ in the same way.''

\begin{restatable}[Similar visit distribution functions have the same optimality probability]{prop}{visitSim}\label{possSim}
If $F,F'\subseteq\F(s)$ are strongly similar via permutation $\phi$, then $\optprob[\Dany]{F,\gamma}=\optprob[\phi\cdot \Dany]{F',\gamma}$.
\end{restatable}
\begin{proof} Let $\gamma\in(0,1)$.
\begin{align}
    \optprob{F,\gamma}&= \E{\rf \sim \Dany}{\mathbbm{1}_{\max_{\f \in F}\f(\gamma)^\top \rf \geq \max_{\f^i\in\Fnd(s)} \f^i(\gamma)^\top\rf}}\label{eq:restrict-opt-prob}\\
    &= \E{\rf \sim \phi\cdot \Dany}{\mathbbm{1}_{\max_{\f'\in F'}\f'(\gamma)^{\top} \rf \geq \max_{\f^i\in\Fnd(s)} \f^i(\gamma)^\top\rf}}\label{eq:similar-visit-opt-prob}\\
    &= \optprob[\phi\cdot \Dany]{F',\gamma}.
\end{align}

\Cref{eq:restrict-opt-prob} follows by \cref{lem:opt-prob-id}. Let $g(b_1,b_2)\defeq \indic{b_1\geq b_2}$ and $f(B_1,B_2\mid \Dany)\defeq \E{\rf\sim \Dany}{g\prn{\max_{\mathbf{b}_1 \in B_1}\mathbf{b}_1^\top \rf,\max_{\mathbf{b}_2 \in B_2}\mathbf{b}_2^\top \rf}}$. Then by \cref{lem:helper-perm} and the definition of strong similarity,
\begin{align*}
    f\prn{F(\gamma),\Fnd(s,\gamma)\mid \Dany}
        & = f\prn{\phi\cdot F(\gamma),\phi\cdot \Fnd(s,\gamma)\mid \phi\cdot \Dany}\\
        & = f\prn{F'(\gamma), \Fnd(s,\gamma) \mid \phi\cdot \Dany}.
\end{align*}
This implies \cref{eq:similar-visit-opt-prob}.

The $\gamma=0$ and $\gamma=1$ cases hold via the appropriate limits.
\end{proof}

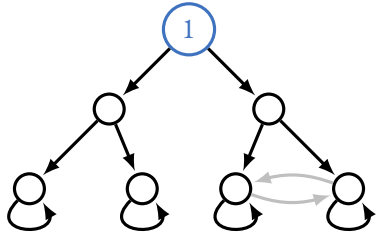
\begin{figure}[h]
    \centering
     \begin{tikzpicture}[mygraph]
        %nodes
            \node[blue] (1) {$1$};
            \node[black] (2) [below right of=1] {};
            \node[black] (3) [below left of=1] {};
            \node[black] (4) [below right of=2] {};
            \node[black] (5) [left of=4] {};
            \node[black] (7) [below left of=3] {};
            \node[black] (6) [right of=7] {};
        %arrows
            \path[->] (1) edge (2);
            \path[->] (1) edge (3);
            \path[->] (2) edge (4);
            \path[->] (2) edge (5);
            \path[->] (3) edge (6);
            \path[->] (3) edge (7);
            \path[->] (4) edge [looseness=6, out=-135, in=-45] (4);
            \path[->] (5) edge [looseness=6, out=-135, in=-45] (5);
            \path[->] (6) edge [looseness=6, out=-135, in=-45] (6);
            \path[->] (7) edge [looseness=6, out=-135, in=-45] (7);
        % Dominated paths
            \path[->] (4) edge [color=lightgray, out=160, in=20] (5);
            \path[->] (5) edge [color=lightgray, out=-20, in=-160] (4);
    \end{tikzpicture}
    \caption[Optimality probability should ignore dominated visit distributions]{Gray actions are only taken by the policies of dominated visit distribution functions. Considered as singleton sets, all four non-dominated visitation distribution functions are strongly similar. \protect\Cref{simNoIC} allows us to conclude they all have equal optimality probability, even though this is not obvious just from looking at the full model. Indeed, the left-most and right-most trajectories would not be classified as ``similar'' under a definition where $\phi$ has to be a stochastic model isomorphism.}
    \label{fig:sim_nd}
\end{figure}

The existence of such a stochastic model isomorphism $\phi$ on the full model is sufficient (but not necessary) for strong similarity (see \cref{fig:sim_nd}).

\begin{figure}[h]
    \centering
    \begin{tikzpicture}[mygraph,shorten >=2pt]
        %nodes
            \node[black] (botleft) {};
            \node[black] (botright) [right of=botleft] {};
        %arrows
            \path[->] (botleft) edge [looseness=6, out=-135, in=135] (botleft);
            \path[<->] (botleft) edge [shorten <= 2pt] (botright);
            \path[->] (botright) edge [looseness=6, out=-45, in=45] (botright);
    \end{tikzpicture}
    \caption[Vertex transitivity doesn't imply strong similarity of all of each state's visit distributions]{The complete directed graph on two vertices is vertex transitive, but each self-loop is not strongly similar to each state's other visit distribution functions.}
    \label{fig:v-trans-not-imply-f-sim}
\end{figure}
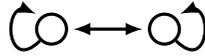

\Cref{fig:v-trans-not-imply-f-sim} shows that vertex transitivity does not imply that all visitation distributions \emph{at a state} are strongly similar to each other. By \cref{model-iso}, vertex transitivity shows that all states have similar visitation distribution function sets, but not that all visit distribution functions within each set are strongly similar to each other.

\begin{restatable}[Strongly similar non-dominated visit distributions and determinism imply no instrumental convergence]{conjSec}{similarnoIC}\label{simNoIC}
If the environment is deterministic and all non-dominated visitation distributions of a state are strongly similar, then no instrumental convergence exists downstream of that state.
\end{restatable}

\section{Orbits}
\begin{restatable}[Trivial satisfaction of $\geq^n_{\text{most}}$]{lem}{trivialRetarget}
Let $f:\retarget \to \reals$, and suppose $\retarget$ is a subset of a set acted on by $\genSym$. $\forall n\geq 0: f(\rtparam)\geqMost[n][\retarget] f(\rtparam)$.
\end{restatable}
\begin{proof}
Because $f(\rtparam)>f(\rtparam)$ is impossible, both cardinalities in \cref{def:restate-orbit} are zero, and so the claim holds trivially.
\end{proof}

\begin{restatable}[Lower-bound on joint $\geqMost$ agreement strength]{conjSec}{lbJointAgreement}\strut\\
Suppose $\prn{f_i \geqMost[C_i] f_i'}_{i\in I}$ for countable index set $I$. Then for all $\Dbd$, the proportion of  the $\Dbd^\phi \in\prn{\orbi}$ which satisfy $\bigwedge_{i\in I} f_i\prn{\Dbd^{\phi}}\geq f_i'\prn{\Dbd^{\phi}}$ is greater than $1-\sum_{i\in I} (1-\frac{C_i}{C_i+1})$.
\end{restatable}

\begin{restatable}[Orbit tendencies lower-bound measure under $\Diid$]{prop}{lbMeasOrbit}
Suppose $f, g:\genVS \to \reals$ are measurable, let $n$ be a positive integer, and suppose $f(\genVS) \geqMost[n] g(\genVS)$. For $\uf \in \genVS$, let $O_{f<g}(\uf)\defeq \set{\uf' \in S_\dimGen \cdot \uf \mid f(\uf')<g(\uf')}$. Suppose there exist $\phi_1,\ldots,\phi_n \in \genSym$ such that for all $\uf \in \genVS: \phi_i \cdot O_{f<g}(\uf) \subseteq \prn{S_\dimGen\cdot \uf}\setminus O_{f<g}(\uf)$, and when $i\neq j$, $\phi_i\cdot O_{f<g}(\uf)$ and $\phi_j\cdot O_{f<g}(\uf)$ are disjoint.

Then for any {\iid} distribution $\D\in\Delta(\genVS)$, $\optprob[\uf \sim \D]{f(\uf)\geq g(\uf)}\geq \frac{n}{n+1}$.
\end{restatable}
\begin{proof}
$f$ and $g$ are measurable, and thus $f-g$ is measurable, and the set $(-\infty,0)$ is measurable in $\reals$, and so $(f-g)\inv\prn{(-\infty,0)}$ is measurable. But this preimage equals $X\defeq \set{\uf\in\genVS \mid f(\uf) < g(\uf)}$, and so $X$ is measurable.

Let $\D$ have probability measure $F$.
\begin{align}
     \optprob[\uf \sim \D]{f(\uf)\geq g(\uf)}&\defeq F\prn{\set{\uf \in \genVS \mid f(\uf)\geq g(\uf)}}\label{eq:exists-measure}\\
    &\geq F\prn{\cup_{i=1}^n \phi_i \cdot X}\label{eq:cup-inside}\\
    &=\sum_{i=1}^n F(\phi_i\cdot X)\label{eq:disjoint-inside}\\
    &=\sum_{i=1}^n F(X)\label{eq:iid-sum}\\
        &=nF(X).
\end{align}
\Cref{eq:exists-measure} is well-defined because the right-hand set is measurable for the same reason that $X$ is measurable. \Cref{eq:cup-inside} holds because assumption on $\phi_i$ ensures that $\phi_i\cdot X \subseteq \set{\uf \in \genVS \mid f(\uf)\geq g(\uf)}$.

If $\phi_i\neq \phi_j$, then suppose $\uf,\uf'\in\genVS$. If they are not in the same orbit under $S_\dimGen$, then $\phi_i\cdot \uf \neq \phi_j\cdot \uf'$ by the definition of an orbit. Otherwise, the inequality holds by the assumed disjointness. Therefore the elements of $\set{\phi_i\cdot X\mid i=1,\ldots,n}$ are pairwise disjoint, and so \cref{eq:disjoint-inside} holds by the finite additivity of probability measures on disjoint sets. \Cref{eq:iid-sum} follows because probability measure $F$ is {\iid} across states and so is invariant to permutation of $X$, and because permutations are measurable transformations with unitary Jacobian determinant.

Since $X\cup \set{\uf \in \genVS \mid f(\uf)\geq g(\uf)}=\genVS$ and the two sets are disjoint, their probability sums to $1$ and so $\optprob[\uf \sim \D]{f(\uf)\geq g(\uf)}\geq \frac{n}{n+1}$.
\end{proof}

\begin{restatable}[Generalized measure lower-bound for orbit tendencies]{conjSec}{genOrbitLB}
Suppose $f, g:\genVS \to \reals$ are measurable, let $C\geq 0$, and suppose $f(\genVS) \geqMost[C] g(\genVS)$. Then for all $\Diid$, $\optprob[\uf \sim \Diid]{f(\uf)\geq g(\uf)}\geq \frac{C}{C+1}$.
\end{restatable}

\subsection{Blackwell versus average optimality}\label{sec:bw-avg}
\citet{turner_optimal_2020}'s \cref{rsdIC} applies to \emph{average-optimal} policies ($\gamma=1$), but not to Blackwell-optimal policies. Therefore, their results are inapplicable to discount rates $\gamma\approx 1$. \Cref{prop:almost-all-bw-avg} shows \cref{res:orbitAlmostEq}, which shows that \cref{rsdIC} holds when $\gamma\approx1$ for almost all reward function orbits.

\begin{restatable}[Visit distribution functions which induce {\rsd}s]{definition}{shortDefinitionRSD}\label{def:induce-rsds}Let $D\subseteq \RSD$. The set of visit distribution functions which induce $\dbf\in D$ is:
    \begin{equation}
    \FRestrictRSD{D}\defeq \set{\f \in \F(s) \mid \NormF{\f,1}\in D}.
    \end{equation}
    $\optprob[\Dany]{D,1}\defeq \optprob[\Dany]{\FRestrictRSD{D},1}$.
\end{restatable}

\begin{restatable}[Average optimality probability is greater than Blackwell optimality probability]{lem}{Avggeq}\label{lem:avg-geq-bw}
Let $D\subseteq \RSD$. $\optprob[\Dany]{\FRestrictRSD{D},1}\leq \avgprob[\Dany]{D}$.
\end{restatable}
\begin{proof}
If some $\f\in\FRestrictRSD{D}$ is Blackwell optimal for reward function $R\in\rewardSpace$, then $\NormF{\f,1}$ is average optimal for $R$ -- \ie{} Blackwell optimality at a state implies average optimality at a state \citep{puterman_markov_2014} . Therefore, the set of reward functions for which some $\f\in\FRestrictRSD{D}$ is Blackwell optimal is a subset of the set of reward functions for which some $\dbf\in D$ is average optimal. Then the desired inequality follows by the monoticity of probability.
\end{proof}

\begin{restatable}[Average optimality probability equals Blackwell optimality probability for almost all reward functions]{prop}{almostAverageEq}\label{prop:almost-all-bw-avg}
Let $D\subseteq \RSD$. For almost all reward functions $\rf\in\rewardVS$, $\optprob[\D_\rf]{D,1}=\avgprob[\D_\rf]{D}$ (where $\D_\rf$ is the degenerate probability distribution which places probability $1$ on $\rf$).
\end{restatable}
\begin{proof}
\Cref{lem:avg-geq-bw} shows that $\optprob[\D_\rf]{D,1}\leq \avgprob[\D_\rf]{D}$. So inequality only holds if $\optprob[\D_\rf]{D,1}< \avgprob[\D_\rf]{D}$. Since $\D_\rf$ is a degenerate distribution, this implies that $\optprob[\D_\rf]{D,1}=0<1= \avgprob[\D_\rf]{D}$. $D$ is non-empty because $1= \avgprob[\D_\rf]{D}$; in particular, there must exist $\dbf_A\in D$ which is average optimal for $\rf$ but which is not induced by a Blackwell optimal policy for $\rf$.

Let $D_B\defeq \set{\dbf^{\pi,s} \mid \pi\in\optPi[\rf,1]}$ be the set of {\rsd}s induced by Blackwell optimal policies for $\rf$. Since every reward function has a Blackwell optimal policy \citep{blackwell_discrete_1962}, $D_B$ must be non-empty. Let $\dbf_B \in D_B$ be one of its elements.

We know that $\dbf_A \not \in D_B$, and so $\dbf_A\neq \dbf_B$. Since Blackwell optimal policies must be average optimal \citep{puterman_markov_2014}, $\dbf_A$ and $\dbf_B$ are both average optimal. By \cref{cor:distinct-maximized},
\begin{equation*}
    \set{\rf \in \rewardVS \mid \abs{\argmax_{\dbf\in\RSD} \dbf^\top \rf}>1}
\end{equation*}
has measure zero under all absolutely continuous measures on $\rewardVS$. In particular, this set has zero Lebesgue measure.
\end{proof}

\begin{restatable}[Average optimality probability equals Blackwell optimality probability for $\Dcont$]{cor}{dContAgreeAvg}\label{cor:cont-agree-avg}
Let $D\subseteq \RSD$. $\optprob[\Dcont]{D,1}=\avgprob[\Dcont]{D}$.
\end{restatable}
\begin{proof}
\begin{align}
    \optprob[\Dcont]{D,1}&\defeq \prob[R\sim\Dcont]{\exists \f^\pi \in \FRestrictRSD{D}: \pi\in\optPi[R,1]}\\
    &=\E{\rf\sim\Dcont}{\optprob[\D_\rf]{D,1}}\\
    &=\E{\rf\sim\Dcont}{\avgprob[\D_\rf]{D,1}}\label{eq:avg-eq-almost}\\
    &=\avgprob[\Dcont]{D}.
\end{align}
By \cref{prop:almost-all-bw-avg}, almost all $\rf\in\rewardVS$ agree that $\optprob[\D_\rf]{D,1}=\avgprob[\D_\rf]{D}$. Since $\Dcont$ is absolutely continuous with respect to the Lebesgue measure, it also assigns zero probability measure to disagreeing reward functions. Then \cref{eq:avg-eq-almost} follows.
\end{proof}

\begin{restatable}[Average optimality probability equals Blackwell optimality probability for all orbit elements in almost all orbits]{cor}{orbitAlmostEq}\label{res:orbitAlmostEq}
Let $D\subseteq \RSD$. For almost all $\rf\in\rewardVS$, all $\D_{\rf'}\in\orbi[\D_{\rf}]$ satisfy $\optprob[\D_{\rf'}]{D,1}=\avgprob[\D_{\rf'}]{D}$.
\end{restatable}
\begin{proof}
By \cref{prop:almost-all-bw-avg}, the set $X\defeq \set{\rf \in \rewardVS\mid \optprob[\D_\rf]{D,1}\neq \avgprob[\D_\rf]{D}}$ has Lebesgue measure zero. Let $O\defeq \set{\rf' \in \rewardVS \mid \exists \rf\in \orbi[\rf']: \rf\in X}$ be the set of reward functions whose orbits contain an element of $X$. Alternatively, $O=\set{\cup_{\phi\in\mdpPermGroup}\permute \rf \mid \rf \in X}$.

Let $\mu$ be the Lebesgue measure; we want to show that $\mu(O)=0$.
\begin{align}
    \mu\prn{O}&\defeq \mu\prn{\cup_{\phi\in\mdpPermGroup}\set{\permute \rf \mid \rf \in X}}\\
    &= \mu\prn{\cup_{\phi\in\mdpPermGroup}\phi\cdot X}\\
    &\leq \sum_{\phi\in\mdpPermGroup} \mu\prn{\phi\cdot X} \label{eq:union-bd}\\
    &= \sum_{\phi\in\mdpPermGroup} \mu\prn{X}\label{eq:measure-preserve}\\
    &= \abs{\mathcal{S}}!\cdot 0\label{eq:0-ub}\\
    &=0.
\end{align}
\Cref{eq:union-bd} follows by the union bound. \Cref{eq:measure-preserve} follows because permutations have Jacobian determinant $1$ and therefore are measure-preserving operators. \Cref{eq:0-ub} follows because $\abs{\mdpPermGroup}=\abs{\mathcal{S}}!$ and $\mu(X)=0$.

Therefore, $\mu(O)=0$, and so for almost all $\rf\in\rewardVS$, all $\D_{\rf'}\in\orbi[\D_{\rf}]$ satisfy $\optprob[\D_{\rf'}]{D,1}=\avgprob[\D_{\rf'}]{D}$.
\end{proof}

\begin{figure}[ht]
    \centering
    \begin{tikzpicture}[mygraph]
        % Upper subgraph
            \node[blue] (1) {$1$};
            \node[black] (2) [above right of=1] {$2$};
            \path[->] (1) edge node[midway,left] {\texttt{up}} (2) ;
            \node[black] (a) [right of=2] {};
            \path[->] (2) edge (a);
            \node[black] (b) [above right of=a] {};
            \path[->] (a) edge (b);
            \path[->] (b) edge (c);
            \node[black] (c) [right of=a] {};
            \path[->] (a) edge (c);
            \path[->] (c) edge [looseness=6, out=-45, in=45] (c);
            \node[black] (d) [below right of=a] {};
            \path[->] (a) edge (d);
            \path[->] (d) edge (c);
        % Lower subgraph
            \node[black] (3) [below right of=1] {$3$};
            \path[->] (1) edge node[midway,left] {\texttt{down}} (3);
            \node[black] (4) [above right of=3] {};
            \path[->] (3) edge (4);
            \path[->] (4) edge [looseness=6, out=45, in=135] (4);
            \node[black] (5) [right of=3] {};
            \path[->] (3) edge (5);
            \path[->] (5) edge [looseness=6, out=0, in=90] (5);
     \end{tikzpicture}

    \caption[On a zero measure set of reward functions, Blackwell-optimal orbit tendencies differ from the average-optimal tendencies]{More states are reachable by going \texttt{up} compared to \texttt{down}. Thus, most orbit elements of $\orbi[\unitvec]$ make it strictly Blackwell-optimal to go \texttt{up}, while also most elements make it (at least weakly) average-optimal to go \texttt{down}. Therefore, the orbit $\orbi[\unitvec]$ does not satisfy \cref{res:orbitAlmostEq}.}\label{fig:indicator-orbit}
\end{figure}

\begin{restatable}[\Cref{sec:bw-avg}'s results hold for child-state distributions $\childDists$ instead of $\RSD$]{conjSec}{conjAnalogy}\label{conj:analogy}
\end{restatable}

\Cref{conj:analogy} will probably be aided by \cref{def:induce-children}, which takes the place of \cref{def:induce-rsds}.

\begin{restatable}[Visit dist.\ functions which induce child visit distributions]{definition}{shortDefinitionChildren}\label{def:induce-children}Let $N\subseteq \childDists$.
    \begin{equation}
        \FRestrictChild{N}\defeq \set{\f \in \F(s) \mid \lim_{\gamma\to 0} \gamma\inv\prn{\f(\gamma)-\unitvec}\in N}.
    \end{equation}

    $\optprob{N,0}\defeq \optprob{\FRestrictChild{N},0}$.\label{item:short-limit-greedy}
\end{restatable}

\subsection{\texorpdfstring{$\pwrNoDist$}{POWER}}

\begin{restatable}[Orbit incentives characterization for $\pwrNoDist$]{conjSec}{mostEqIff}
Let $\gamma\in(0,1)$. The following statements are equivalent:
\begin{enumerate}
    \item $\pwr \geqMost[][\DSetBd] \pwr[s',\gamma]$.
    \item For all continuous bounded {\iid} $\Diid$, $\pwr[s,\gamma][\Diid]\geq\pwr[s',\gamma][\Diid]$.
    \item $\Fnd(s')$ is similar to a subset of $\F(s)$.
\end{enumerate}
\end{restatable}

\begin{restatable}[$\geq_\text{most: $\DSetBd$}$ is not a complete ordering for the $\pwrNoDist$ of states]{conjSec}{completeOrder}
\end{restatable}

Lastly, a remark motivating the requirement of \emph{involution} in the copy containment definition (\cref{def:dist-sim}).

\begin{remark}
\Cref{lem:expect-superior} and \cref{lem:opt-prob-superior} require similarity via involution (not just via permutation). Consider $A\defeq \set{\colvec{3}{0}{5}{10}}, B\defeq \set{\colvec{3}{10}{0}{5}},\phi\defeq (1\, 2\, 3)$. $A$ is similar to $B$ via permutation $\phi$. However, there is no involution which enforces the similarity.

Furthermore, consider the orbit $S_3 \cdot \D_{\unitvec[3]}=\set{\D_{\unitvec[1]},\D_{\unitvec[2]},\D_{\unitvec[3]}}$ and consider the conditions of \cref{lem:expect-superior} with $g$ the identity function. \[\E{\rf \sim \D_{\unitvec[1]}}{\max_{\bv\in B}\bv^\top \rf}=10>0=\E{\rf \sim \D_{\unitvec[1]}}{\max_{\av\in A}\bv^\top \rf},\] but the opposite is true for the other two orbit elements. Therefore,
\begin{equation*}
    \E{\rf \sim \Dbd}{\max_{\av\in A}\bv^\top \rf}\not\leqMost[][\DSetBd]\E{\rf \sim \Dbd}{\max_{\bv\in B}\bv^\top \rf}.
\end{equation*}
\end{remark}

\begin{restatable}[Nontrivial copy containment guarantee]{prop}{copyLem}
If $B$ contains $n>1$ copies of $A$ via $\phi_1,...,\phi_n$, then for any $i=1,...,n$, $A\neq B_i$.
\end{restatable}
\begin{proof}
Suppose $A=B_i\defeq \phi_i\cdot A$ for some $\phi_i$. Then consider any $j\neq i$ (which exists since $n>1$). $\phi_j\cdot B_i = \phi_j\cdot A = B_j$. But by the set copy definition (\cref{def:copies}), $\phi_j\cdot B_i = B_i$ since $j\neq i$. Then $B_i=B_j$, which contradicts the set copy assumption.
\end{proof}

\section{Featurized utility functions}\label{app:feat}
Consider the case where utility functions over outcomes are \emph{featurized}: $u(o)=\feat(o)^\top \alpha$, with $\feat(o): \set{1,\ldots, \dimGen} \to \featVS$ linearly mapping deterministic outcomes to $\nfeat$-dimensional feature vectors. Let $\featMat\defeq \prn{\feat(o_1) \quad \cdots \quad \feat(o_\dimGen)}$ be the feature matrix. For outcome lottery set $X\subseteq \genVS$, $\featMat \cdot X \defeq \set{\featMat \x \mid \x \in X}$ is the left-coset.

For feature weighting $\alpha \in \featVS$, the expected utility of an outcome lottery $\x \in \genVS$ is then the expected value of the $\alpha$-weighted combination of each outcome's features: $\x^\top \uf = \x^\top (\featMat^\top \alpha)=(\featMat \x)^\top \alpha$. For example, if an agent maximizes expected utility, it essentially makes decisions according to the available feature vectors of outcomes, and so instead of looking for copy containment in the outcome lottery sets $A,B\subseteq \genVS$, we can look for copy containment in the feature vector sets $\featMat \cdot A, \featMat \cdot B \subseteq \featVS$.

\begin{restatable}[Feature-level tendencies guaranteed by featurizations which commute with outcome symmetries]{prop}{featGuarantee} \label{res:feat-guarantee-commute}
Let $A,B\subsetneq \genVS$, and suppose that $B$ contains a copy of $A$ via $\phi$ with permutation matrix $\permute \in \reals^{\dimGen\times \dimGen}$. Consider featurization $\featMat \in \reals^{\nfeat\times \dimGen}$ which maps each of $\dimGen$ outcomes to their $\nfeat$-feature vectors.

If there exists a feature involution $\phi_f \in S_{\nfeat}$ such that $\permute[\phi_f] \featMat = \featMat \permute$, then $\featMat \cdot B$ contains a copy of $\featMat \cdot A$ via $\phi_f$.
\end{restatable}
\begin{proof}
\begin{align}
    \phi_f \cdot \prn{\featMat \cdot A} &\defeq \set{\permute[\phi_f] \featMat \av \mid \av \in A}\\
    &= \set{\featMat \permute \av \mid \av \in A}\label{eq:feat-commute}\\
    &\subseteq \set{\featMat \bv \mid \bv \in B}\label{eq:subset-feat}\\
    &\eqdef \featMat \cdot B.
\end{align}
\Cref{eq:feat-commute} follows by the assumption that $\permute[\phi_f] \featMat = \featMat \permute$. \Cref{eq:subset-feat} follows because we assumed that $\phi\cdot A \subseteq B$.

Since $\phi_f \cdot \prn{\featMat \cdot A}\subseteq \featMat \cdot B$ and $\phi_f$ is an involution by assumption, then we conclude that $\featMat \cdot B$ contains a copy of $\featMat \cdot A$ via $\phi_f$.
\end{proof}

\begin{restatable}[Multiple feature copy containment ensured by feature commutation]{conjSec}{multiContain}
\Cref{res:feat-guarantee-commute}  generalizes to the case with multiple feature involutions $\phi_{f_1},\ldots,\phi_{f_n}$.
\end{restatable}

\section{\texorpdfstring{$\epsilon$}{Epsilon}-Optimal policies}
\citet{turner_optimal_2020}'s results assume perfectly optimal agents. We now relax that assumption slightly to $\epsilon$-optimality, although $\epsilon$ may be extremely small (and so these results do not seem very practically interesting). For a more practical extension, see \cref{chp:rationalities}'s total abandonment of the optimality requirement.

Traditionally, $\epsilon$-optimal policies are defined as $\pi$ such that for all $s\in\St$:
\begin{equation}
    \OptVf{s,\gamma} - \Vf[\pi][R]{s,\gamma} \leq \epsilon\label{eq:traditional-eps}
\end{equation}
for fixed $\gamma \in (0,1)$. However, in the state-based reward setting, this is equivalent to
\begin{align}
    \prn{R(s)+\gamma\max_{\pi'}\E{s' \sim T(s,\pi'(s))}{\Vf[\pi'][R]{s',\gamma}}}&-\prn{R(s)+\gamma\E{s' \sim T(s,\pi(s))}{\Vf[\pi][R]{s',\gamma}}} \leq \epsilon\\
    \max_{\pi'}\E{s' \sim T(s,\pi'(s))}{\Vf[\pi'][R]{s',\gamma}}&-\E{s' \sim T(s,\pi(s))}{\Vf[\pi][R]{s',\gamma}} \leq \gamma\inv\epsilon.
\end{align}

To avoid superfluous division by $\gamma$, we modify the traditional criterion to
\begin{equation}
    \max_{\pi'}\E{s' \sim T(s,\pi'(s))}{\Vf[\pi'][R]{s',\gamma}}-\E{s' \sim T(s,\pi(s))}{\Vf[\pi][R]{s',\gamma}} \leq \epsilon.
\end{equation}

However, since the magnitude of on-policy value often diverges as $\gamma \to 1$, we instead consider a \emph{discount-averaged per-time step suboptimality} of $\epsilon$:
\begin{equation}
    \max_{\pi'}\E{s' \sim T(s,\pi'(s))}{\Vf[\pi'][R]{s',\gamma}}-\E{s' \sim T(s,\pi(s))}{\Vf[\pi][R]{s',\gamma}} \leq \geom[\epsilon].
\end{equation}

Multiplying both sides by $1-\gamma$, we arrive at our definition.

\begin{restatable}[$\epsilon$-optimal policy]{definition}{defEpsOpt}\label{def:eps-opt}
Let $\epsilon\geq 0$. Policy $\pi$ is \emph{$\epsilon$-optimal for reward function $R$ at discount rate $\gamma\in [0,1]$} when for all $s\in\St$,
\begin{equation}
    \max_{\pi'}\E{s' \sim T(s,\pi'(s))}{\VfNorm[\pi'][R]{s',\gamma}}-\E{s' \sim T(s,\pi(s))}{\VfNorm[\pi][R]{s',\gamma}} \leq \epsilon.
\end{equation}
\end{restatable}

For any fixed $\gamma\in(0,1)$, the traditional definition (\cref{eq:traditional-eps}) is equivalent to \cref{def:eps-opt}, in that traditional $\epsilon$-optimality is equivalent to \cref{def:eps-opt}'s $\frac{\epsilon}{\gamma(1-\gamma)}$-optimality.

\begin{restatable}[$\epsilon$-optimal policy set]{definition}{defEpsOptSet}\label{def:eps-opt-piset}
Let $\epsilon\geq 0$. The \emph{$\epsilon$-optimal policy set} for reward function $R$ at discount rate $\gamma\in [0,1]$ is
\begin{equation}
    \piSet[\epsilon] \defeq \set{\pi \in \Pi \mid \pi \text{ is $\epsilon$-optimal for $R$ at $\gamma$}}.
\end{equation}
\end{restatable}

\begin{remark}
When $\gamma\in(0,1)$, many reward functions may have multiple $\epsilon$-optimal visit distributions, whereas \cref{unique-opt-value} shows that continuous distributions place zero probability on reward functions with multiple optimal visit distributions.
\end{remark}

\subsection{\texorpdfstring{$\epsilon$}{Epsilon}-optimal \texorpdfstring{$\pwrNoDist$}{POWER}}
The situation for $\epsilon$-optimal $\pwr$ is quite simple.

\begin{restatable}[$\epsilon$-optimal policy-generating function]{definition}{epsOptPolGen}
$\pol[]: \rewardSpace \times [0,1]\to \Pi$ is an \emph{$\epsilon$-optimal policy-generating function} when $\forall R\in \rewardSpace,\gamma\in[0,1]: \pol\in\piSet[\epsilon]$.
\end{restatable}

\begin{restatable}[$\epsilon$-optimal $\pwr$ bound]{prop}{epsOptPower}\label{thm:eps-opt-power}
Let $\epsilon\geq 0$ and $\gamma \in [0,1]$. Let $\pol[]$ be an $\epsilon$-optimal policy-generating function. Then for all states $s$,
\begin{equation}
    \pwr[s,\gamma] - \pwrPol{s,\gamma} \leq \epsilon.
\end{equation}
\end{restatable}
\begin{proof}
\begin{align}
    &\pwr[s,\gamma] - \pwrPol{s,\gamma} \\
    =\,& \E{R\sim \Dany}{\max_{\pi'}\E{s' \sim T(s,\pi'(s))}{\VfNorm[\pi'][R]{s',\gamma}}-\E{s' \sim T(s,\pi(s))}{\VfNorm[\pi][R]{s',\gamma}}}\label{eq:pwr-id-expect}\\
    \leq\,& \E{R\sim \Dany}{\epsilon}\label{eq:pol-eps-opt}\\
    =\,& \epsilon.
\end{align}

\Cref{eq:pwr-id-expect} follows from \cref{lem:pwr-avg} and \cref{def:pow-pol} of $\pwrPol{}$. \Cref{eq:pol-eps-opt} follows from its assumed $\epsilon$-optimality for all reward functions.
\end{proof}

\begin{restatable}[Optimal $\pwrNoDist$-seeking implies $\epsilon$-optimal $\pwrNoDist$-seeking]{thm}{epsPowerSeek}\label{thm:eps-pwr-seek}
Let $k\geq 1$, $s\in\St,a$, $a'\in\A,\gamma\in[0,1]$. If
\begin{equation*}
    \E{s' \sim T(s,a)} {\pwr[s',\gamma]} > k\E{s' \sim T(s,a')} {\pwr[s',\gamma]},
\end{equation*}
then there exists $\epsilon > 0$ such that
\begin{equation*}
    \E{s' \sim T(s,a)} {\pwrPol{s',\gamma}} > k\E{s' \sim T(s,a')} {\pwrPol{s',\gamma}},
\end{equation*}
where $\pol[]$ is any $\epsilon$-optimal policy-generating function.
\end{restatable}
\begin{proof}
Let $\delta \defeq \E{s' \sim T(s,a)} {\pwr[s',\gamma]} - k\E{s' \sim T(s,a')} {\pwr[s',\gamma]}$.
\begin{align}
    &\E{\substack{s_a \sim T(s,a),\\s_{a'} \sim T(s,a')}}{\pwrPol{s_a,\gamma} - k\pwrPol{s_{a'},\gamma}}\\
    ={}&\mathbb{E}_{\substack{s_a \sim T(s,a),\\s_{a'} \sim T(s,a')}}\bigg[\prn{\pwrPol{s_a,\gamma} - \pwr[s_a,\gamma]} \nonumber\\
    &\phantom{\mathbb{E}_{\substack{s_a \sim T(s,a),\\s_{a'} \sim T(s,a')}}}+ \prn{\pwr[s_a,\gamma] - k\pwr[s_{a'},\gamma]} \nonumber \\
    &\phantom{\mathbb{E}_{\substack{s_a \sim T(s,a),\\s_{a'} \sim T(s,a')}}}+\prn{\pwr[s_{a'},\gamma] -\pwrPol{s_{a'},\gamma}}\bigg]\\
    ={}&\mathbb{E}_{\substack{s_a \sim T(s,a),\\s_{a'} \sim T(s,a')}}\bigg[\prn{\pwr[s_a,\gamma] - k\pwr[s_{a'},\gamma]}\nonumber\\
    &\phantom{\mathbb{E}_{\substack{s_a \sim T(s,a),\\s_{a'} \sim T(s,a')}}}- \prn{\pwr[s_a,\gamma]-\pwrPol{s_a,\gamma}} \nonumber \\
    &\phantom{\mathbb{E}_{\substack{s_a \sim T(s,a),\\s_{a'} \sim T(s,a')}}}+\prn{\pwr[s_{a'},\gamma] -\pwrPol{s_{a'},\gamma}}\bigg]\\
    \geq{}& \E{\substack{s_a \sim T(s,a),\\s_{a'} \sim T(s,a')}}{\prn{\pwr[s_a,\gamma] - k\pwr[s_{a'},\gamma]}-\epsilon+0}\\
    ={}& \delta -\epsilon\label{eq:pwr-eps-worst-case}\\
    >{}& 0.\label{eq:pwr-ineq-eps}
\end{align}

\Cref{eq:pwr-eps-worst-case} follows because $\pwr[s_{a'},\gamma]\geq\pwrPol{s_{a'},\gamma}$ and because \cref{thm:eps-opt-power} bounds $\pwr[s_a,\gamma]-\pwrPol{s_a,\gamma}\leq \epsilon$ because $\pol[]$ is an $\epsilon$-optimal policy generating function. \Cref{eq:pwr-ineq-eps} follows for any $\epsilon<\delta$; since $\delta$ is positive, we can ensure that $\epsilon$ is as well.
\end{proof}

\subsection{\texorpdfstring{$\epsilon$}{Epsilon}-optimality probability}
The situation for $\epsilon$-optimality probability is less simple.

\begin{restatable}[$\epsilon$-optimality probability]{definition}{epsOptProb}\label{def:eps-opt-prob}
For state $s$, let $F\subseteq \F(s)$, $\gamma \in [0,1]$, and $\epsilon \geq 0$.
\begin{equation}
    \optprob[\Dany][\epsilon]{F,\gamma}\defeq \prob[R\sim\Dany]{\exists \f^\pi \in F: \pi\in\piSet[\epsilon]}.
\end{equation}
\end{restatable}

\begin{restatable}[Average-optimal policies]{definition}{avgOpt}\label{def:avg-opt}
$\average$ is the set of average reward-optimal policies for reward function $R$.
\end{restatable}

\begin{restatable}[Characterizing $0$-optimal policy sets]{prop}{charEpsOptSet}\label{prop:char-0-opt}
Let $\gamma \in (0,1)$.
\begin{enumerate}
    \item $\optPi=\piSet[0][R,\gamma]$.
    \item $\optPi[R,0]\subseteq \piSet[0][R,0]=\greedy$. \item $\optPi[R,1]\subseteq \piSet[0][R,1]=\average$.
\end{enumerate}
\end{restatable}
\begin{proof}
If $\gamma \in (0,1)$, $\piSet[0][R,\gamma]$ must be optimal at every $s$ by \cref{def:eps-opt}. $\piSet[0][R,0]=\greedy$ by \cref{def:greedy-state} and $\piSet[0][R,1]=\average$ by \cref{def:avg-opt}. $\optPi[R,0]\subseteq\greedy$ by \cref{lem:greedy-reward} and $\optPi[R,1]\subseteq\average$ by the fact that Blackwell optimal policies must be average optimal (\cite{puterman_markov_2014}).
\end{proof}

\begin{figure}[ht]
    \centering
    \begin{tikzpicture}[mygraph,shorten > = -3pt,node distance = 1.1cm]
        %nodes
            \node (1) {$\col{blue}{s_0}$};
            \node [left of=1, node distance=.4cm] {\col{green}{$0$}};
            \node (2) [above right of=1] {$s_1$};
            \node [below of=2, node distance=.4cm] {\col{green}{$1$}};
            \node (3) [below right of=1] {$s_1'$};
            \node [below of=3, node distance=.4cm] {\col{green}{$1$}};
            \node (4) [right of=2] {$s_2$};
            \node [below of=4, node distance=.4cm] {\col{green}{$1$}};
            \node (5) [right of=3] {$s_2'$};
            \node [below of=5, node distance=.4cm] {\col{green}{$0$}};
            \node (6) [right of=1, node distance=2.7cm] {$s_\text{sink}$};
            \node [below of=6, node distance=.4cm] {\col{green}{$0$}};
        %arrows
            \path[->] (1) edge node[midway,left] {\texttt{up}} (2);
            \path[->] (1) edge node[midway,left] {\texttt{down}} (3);
            \path[->] (2) edge (4);
            \path[->] (3) edge  (5);
            \path[->] (4) edge (6);
            \path[->] (5) edge  (6);
            \path[->] (6) edge [looseness=4, out=-45, in=45] (6);
    \end{tikzpicture}
    \caption[Average optimality and Blackwell optimality are distinct concepts]{All policies are average-optimal, since there is only one {\rsd}. However, only going \texttt{up} is Blackwell-optimal. Similarly, all policies are greedily optimal, since $R(s_1)=R(s_1')=1$, but only going \texttt{up} is asymptotically greedily optimal. In this {\mdp}, $\piSet[0][R,0]\neq \optPi[R,0]$ and $\piSet[0][R,1]\neq \optPi[R,1]$ by \cref{prop:char-0-opt}.}
    \label{fig:gain-subopt-diff}
\end{figure}
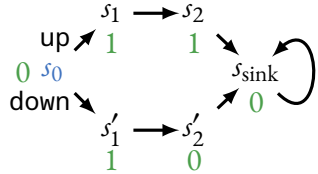

\begin{restatable}[When $\gamma \in (0,1)$, $0$-optimality probability coincides with optimality probability]{cor}{zeroOptProb}\label{cor:zero-opt-prob}
Let $F\subseteq \F(s)$. If $\gamma\in(0,1)$, then $\optprob[\Dany]{F,\gamma} = \optprob[\Dany][0]{F,\gamma}$.
\end{restatable}
\begin{proof}
When $\gamma \in (0,1)$, $\optPi=\piSet[0][R,\gamma]$ by \cref{prop:char-0-opt}. $\epsilon$-optimality probability (\cref{def:eps-opt-prob}) then reduces to optimality probability (\cref{def:prob-opt}).
\end{proof}

\begin{restatable}[$\epsilon$-optimal policy set monotonicity]{lem}{epsMonotone}\label{lem:eps-monotone}
Let $0 \leq \epsilon_1 \leq \epsilon_2$. Then $\piSet[\epsilon_1]\subseteq \piSet[\epsilon_2]$.
\end{restatable}
\begin{proof}
Suppose $\pi \in \piSet[\epsilon_1]$. Then by \cref{def:eps-opt-piset},
\begin{align}
    \max_{\pi'}\E{s' \sim T(s,\pi'(s))}{\VfNorm[\pi'][R]{s',\gamma}}-\E{s' \sim T(s,\pi(s))}{\VfNorm[\pi][R]{s',\gamma}} &\leq \epsilon_1\\
    &\leq \epsilon_2.
\end{align}
Therefore, $\pi \in \piSet[\epsilon_2]$.
\end{proof}

\begin{restatable}[$\epsilon$-optimal policy set containment]{cor}{epsContained}\label{cor:eps-contained}
Let $\epsilon \geq 0$. For all $R\in\rewardSpace,\gamma\in[0,1]$, $\optPi \subseteq \piSet[0]\subseteq \piSet[\epsilon]$.
\end{restatable}
\begin{proof}
The first containment holds by \cref{prop:char-0-opt}. The second containment holds by \cref{lem:eps-monotone}.
\end{proof}

\begin{restatable}[$\epsilon$-optimality probability is monotonically increasing in $\epsilon$]{prop}{epsProbIneq}\label{prop:eps-prob-ineq}
Let $s$ be a state and consider $F \subseteq \F(s)$, $\gamma \in [0,1]$. Let $0 \leq \epsilon_1\leq \epsilon_2$. $\optprob{F,\gamma} \leq \optprob[\Dany][\epsilon_1]{F,\gamma}\leq \optprob[\Dany][\epsilon_2]{F,\gamma}$.
\end{restatable}
\begin{proof}
By \cref{lem:eps-monotone} and \cref{cor:eps-contained}, $\optPi \subseteq \piSet[\epsilon_1]\subseteq \piSet[\epsilon_2]$. Then $\optprob{F,\gamma} \leq \optprob[\Dany][\epsilon_1]{F,\gamma}\leq \optprob[\Dany][\epsilon_2]{F,\gamma}$ by the monotonicity of probability.
\end{proof}

As $\epsilon$ increases, more policies are $\epsilon$-optimal for any given reward function (\cref{lem:eps-monotone}). The following result shows that for bounded reward function distributions, ``anything goes'' for sufficiently large $\epsilon$.
\begin{restatable}[Under $\Dbd$, every policy can be $\epsilon$-optimal (for the right $\epsilon$)]{prop}{maxEpsProb}
Let $\Dbd$ be a reward function distribution which is bounded $[b,c]$. Then for all $R \in \supp[\Dbd]$ and $\gamma \in [0,1]$, $\piSet[(c-b)]=\Pi$ and for all $s$, non-empty $F\subseteq \F(s)$, $\optprob[\Dbd][(c-b)]{F,\gamma}=1$.
\end{restatable}
\begin{proof}
Let $\gamma \in [0,1]$. \begin{equation}
\forall R \in \supp[\Dbd],s\in \St:\OptVfNorm{s,\gamma}\leq \lim_{\gamma^*\to\gamma}\frac{c}{1-\gamma^*}=c.
\end{equation}

Its minimal normalized value is likewise at least  $b$. Therefore, for all $\pi \in \Pi$, we have
\begin{align}
    &\max_{\pi'}\E{s' \sim T(s,\pi'(s))}{\VfNorm[\pi'][R]{s',\gamma}}-\E{s' \sim T(s,\pi(s))}{\VfNorm[\pi][R]{s',\gamma}}\\
    \leq\,& \max_{\pi'}\E{s' \sim T(s,\pi'(s))}{c}-\E{s' \sim T(s,\pi(s))}{b}\\
    =\,&c-b.
\end{align}

This holds at all states $s$. So every $\pi$ is $(c-b)$-optimal, and $\piSet[(c-b)]=\Pi$.

Since $F$ is not empty, it must contain some $\fpi{} \in F$. Since $\piSet[(c-b)]=\Pi$, $\pi \in \piSet[(c-b)]$ for all $R$, and so $\optprob[\Dbd][(c-b)]{F,\gamma}=1$ by \cref{def:eps-opt-prob}.
\end{proof}

Suppose that we know that action $a$ is both strictly $\pwrNoDist$-seeking and strictly more probable under optimality compared to another action $a'$. \Cref{thm:eps-opt-power} implies that there exists an $\epsilon>0$ for which $a$ is $\epsilon$-optimal $\pwrNoDist$-seeking compared to $a'$. The following result implies a similar result with respect to $\epsilon$-optimality probability.
\begin{restatable}[$\epsilon$-optimality probability approaches $0$-optimality probability in a continuous fashion]{prop}{epsZeroCont}\label{thm:eps-opt-prob-cont}
Let $F\subseteq \F(s)$. $\lim_{\epsilon \to 0}\optprob[\Dany][\epsilon]{F,\gamma}=\optprob[\Dany][0]{F,\gamma}$.
\end{restatable}
\begin{proof}
Let $f(\epsilon)\defeq \optprob[\Dany][\epsilon]{F,\gamma}$. $f$ is monotonically increasing in $\epsilon$ and has range $[\optprob[\Dany][0]{F,\gamma},1]$ by \cref{prop:eps-prob-ineq}. Suppose the sequence $\mathbf{s}\defeq \prn{\epsilon_i}_{i\geq 1}$ converges to $0$.  Choose a monotonically decreasing subsequence $\mathbf{s}'\defeq \prn{\epsilon_j'}_{j \in I}$.  $\mathbf{s}'$ must also converge to $0$. Then $\mathbf{s}'_f\defeq \prn{f(\epsilon_j')}_{j \in I}$ monotonically decreases and is bounded below by $\optprob[\Dany][0]{F,\gamma}$. Therefore, by the monotone convergence theorem, $\mathbf{s}'_f$ has non-negative limit $L$.

Suppose $L >\optprob[\Dany][0]{F,\gamma}$. This implies that all elements of $\mathbf{s}'$ are positive. Then there exist reward functions for which some $\fpi{} \in F$ is $\epsilon$-optimal for arbitrarily small $\epsilon>0$, but for which $\f$ is not $0$-optimal. This is impossible, and so $L =\optprob[\Dany][0]{F,\gamma}$.

Furthermore, since $\mathbf{s}\to 0$, $\prn{f(\epsilon_i)}_{i\geq 1} \to \optprob[\Dany][0]{F,\gamma}$ as well. Since this limit applies for any sequence $\mathbf{s}\to 0$, the result follows.
\end{proof}

\begin{restatable}[For small $\epsilon$, $\epsilon$-optimality probability approximates optimality probability]{cor}{epsCloseProb}\label{cor:eps-close-prob}
Let $F\subseteq \F(s)$. When $\gamma \in (0,1)$, for any $\delta > 0$, there exists $\epsilon > 0$ such that $\optprob[\Dany][\epsilon]{F,\gamma}-\optprob[\Dany]{F,\gamma}<\delta$.
\end{restatable}
\begin{proof}
By \cref{cor:zero-opt-prob}, when $\gamma \in (0,1)$, $\optprob[\Dany][0]{F,\gamma}=\optprob[\Dany]{F,\gamma}$. Therefore, when $\gamma \in (0,1)$, \cref{thm:eps-opt-prob-cont} shows that $\lim_{\epsilon \to 0}\optprob[\Dany][\epsilon]{F,\gamma}=\optprob[\Dany]{F,\gamma}$. \Cref{thm:eps-opt-prob-cont} showed that $f(\epsilon)\defeq \optprob[\Dany][\epsilon]{F,\gamma}$ is sequentially continuous at $\epsilon=0$; since $f:\reals \to \reals$ has metric spaces for both its domain and its range, sequential continuity implies (topological) continuity at $\epsilon=0$. Then the claim follows from the definition of continuity on metric spaces.
\end{proof}

\begin{restatable}[Strict optimality probability inequalities are preserved for small enough $\epsilon$]{lem}{contDifferenceEps}\label{cont-diff-eps-opt}
Let $k\geq 1,s,s'\in \St,F\subseteq \F(s),F'\subseteq \F(s')$, and $\gamma \in (0,1)$. If $\optprob[\Dany]{F,\gamma}>k\optprob[\Dany]{F',\gamma}$, then there exists $\epsilon>0$ such that $\optprob[\Dany][\epsilon]{F,\gamma}>k\optprob[\Dany][\epsilon]{F',\gamma}$.
\end{restatable}
\begin{proof}
Let $\delta \defeq \optprob[\Dany]{F,\gamma}-k\optprob[\Dany]{F',\gamma}$; note that $\delta>0$ by assumption.
\begin{align}
    &\optprob[\Dany][\epsilon]{F,\gamma}-k\optprob[\Dany][\epsilon]{F',\gamma}\\
    ={} &         \prn{\optprob[\Dany][\epsilon]{F,\gamma}-\optprob[\Dany]{F,\gamma}}\nonumber\\
        & \qquad+ \prn{\optprob[\Dany]{F,\gamma}-k\optprob[\Dany]{F',\gamma}}\nonumber\\
        & \qquad+k\prn{\optprob[\Dany]{F',\gamma}-\optprob[\Dany][\epsilon]{F',\gamma}}\label{eq:eps-prob-ineq}\\
    ={}&\delta+\prn{\optprob[\Dany][\epsilon]{F,\gamma}-\optprob[\Dany]{F,\gamma}}-k\prn{\optprob[\Dany][\epsilon]{F',\gamma}-\optprob[\Dany]{F',\gamma}}\\
    \geq{}&\delta+0-k\prn{\optprob[\Dany][\epsilon]{F',\gamma}-\optprob[\Dany]{F',\gamma}}\label{eq:geq-0-eps}\\
    >{}&0.\label{eq:ineq-close-eps}
\end{align}
\Cref{eq:geq-0-eps} follows by \cref{prop:eps-prob-ineq}. By \cref{cor:eps-close-prob}, $\gamma \in (0,1)$ and $\delta>0$ implies that we can choose $\epsilon$ so that $\optprob[\Dany][\epsilon]{F,\gamma}-\optprob[\Dany]{F,\gamma}<\frac{\delta}{k}$; by doing so, \cref{eq:ineq-close-eps} follows.
\end{proof}

\begin{restatable}[$\epsilon$-optimality results hold for {\rsd} optimality probability]{conjSec}{rsdEpsOpt}
\end{restatable}

For shorthand, we define action $\epsilon$-optimality probability, mirroring \cref{def:action-optimality}'s definition of action optimality probability.
\begin{restatable}[Action $\epsilon$-optimality probability]{definition}{actEpsOpt}\label{def:action-eps-optimality}
Let $\epsilon\geq 0$. At discount rate $\gamma$ and at state $s$, the \emph{$\epsilon$-optimality probability of action $a$} is
\begin{equation*}
    \optprob[\Dany][\epsilon]{s,a,\gamma}\defeq \prob[R \sim \Dany]{\exists \pi \in \piSet[\epsilon]: \pi(s)=a}.
\end{equation*}
\end{restatable}

\begin{restatable}[Action $\epsilon$-opt.\ probability is a special case of visit distribution $\epsilon$-opt.\ prob.]{prop}{agreeEpsOptProb}\label{prop:agree-eps-opt-prob} For any $\gamma \in [0,1]$ and $\epsilon\geq 0$,  $\optprob[\Dany][\epsilon]{s,a,\gamma}=\optprob[\Dany][\epsilon]{\FRestrictAction{s}{a},\gamma}$.
\end{restatable}
\begin{proof}
Let $F_a\defeq \FRestrictAction{s}{a}$. For $\gamma \in [0,1]$ and any state $s$,
\begin{align}
    \optprob[\Dany][\epsilon]{s,a,\gamma}&\defeq\prob[R \sim \Dany]{\exists \pi \in \piSet[\epsilon]: \pi(s)=a}\\
    &= \prob[\rf \sim \Dany]{\exists \fpi{} \in F_a: \pi\in \piSet[\epsilon]}\label{eq:f-action-equiv-eps}\\
    &\eqdef\optprob[\Dany][\epsilon]{F_a,\gamma}.
\end{align}
\Cref{eq:f-action-equiv-eps} follows because $\f\in F_a$ iff $\exists \pi\in\Pi:\fpi{}=\f,\pi(s)=a$ by the definition of $\FRestrictAction{s}{a}$ (\cref{def:restrict-single}).
\end{proof}

\begin{restatable}[Optimal $\pwrNoDist$-seeking incentives imply $\epsilon$-optimal $\pwrNoDist$-seeking incentives]{thm}{espSeekImply}\label{thm:eps-tend-seek}
Let $s\in\St,a,a'\in\A,\gamma\in(0,1)$. If
\begin{equation*}
    \E{s' \sim T(s,a)} {\pwr[s',\gamma]} > \E{s' \sim T(s,a')} {\pwr[s',\gamma]}
\end{equation*}
and
\begin{equation*}
    \optprob[\Dany]{s,a,\gamma}>\optprob[\Dany]{s,a',\gamma},
\end{equation*}
then there exists $\epsilon > 0$ such that
\begin{equation*}
    \E{s' \sim T(s,a)}{\pwrPol{s',\gamma}} > \E{s' \sim T(s,a')} {\pwrPol{s',\gamma}}
\end{equation*}
and
\begin{equation*}
    \optprob[\Dany][\epsilon]{s,a,\gamma}>\optprob[\Dany][\epsilon]{s,a',\gamma},
\end{equation*}
where $\pol[]$ is any $\epsilon$-optimal policy-generating function.
\end{restatable}
\begin{proof}
\Cref{thm:eps-pwr-seek} guarantees the existence of $\epsilon_{\pwrNoDist}>0$ such that
\begin{equation*}
    \E{s' \sim T(s,a)} {\pwrPol{s',\gamma}} > \E{s' \sim T(s,a')} {\pwrPol{s',\gamma}}.
\end{equation*}
\Cref{cont-diff-eps-opt} guarantees the existence of $\epsilon_{\Prb}>0$ such that
\begin{align}
    \optprob[\Dany][\epsilon_{\Prb}]{s,a,\gamma}&=\optprob[\Dany][\epsilon_{\Prb}]{\FRestrictAction{s}{a},\gamma}\label{eq:opt-act-1}\\
    &>\optprob[\Dany][\epsilon_{\Prb}]{\FRestrictAction{s}{a'},\gamma}\\
    &=\optprob[\Dany][\epsilon_{\Prb}]{s,a',\gamma}.\label{eq:opt-act-2}
\end{align}
\Cref{eq:opt-act-1} and \cref{eq:opt-act-2} follow from \cref{prop:agree-eps-opt-prob}.

Choosing $\epsilon<\min\prn{\epsilon_{\pwrNoDist},\epsilon_{\Prb}}$ ensures that both strict inequalities hold. Since both $\epsilon_{\pwrNoDist}$ and $\epsilon_{\Prb}$ are positive, we can ensure that $\epsilon>0$.
\end{proof}

The $\pwr$-seeking theorems are only concerned with the probability of the \emph{existence} of $\epsilon$-optimal policies which seek $\pwr$. Supposing that an agent is only constrained to follow some $\epsilon$-optimal policy, the agent may end up seeking $\pwr$ with lower probability due to the influence of \eg{} some tie-breaking rule for policy selection.

\begin{restatable}[Continuous distributions have continuous $\epsilon$-optimality probability functions]{conjSec}{contEpsProb}
Fixing any $s\in\St,F\subseteq\F(s),\gamma\in[0,1]$, $\optprob[\Dcont][\epsilon]{F,\gamma}$ is continuous on $\epsilon \in [0,\infty)$.
\end{restatable}
%%%%%%%%%%%%%%%%%%%%%%%%%%%%%%%%%%%
\end{appendices}

% backmatter
\clearpage
\phantomsection
\printendnotes
\clearpage\phantomsection
\begingroup
\RaggedRight
\clearpage\phantomsection
\addcontentsline{toc}{chapter}{Lists of Figures \& Results}

\renewcommand{\listfigurename}{List of Figures}
\listoffigures
\listoftables

\def\thmtformatoptarg#1{~#1}
\makeatletter
\let\Contentsline\thmt@contentsline
\renewcommand\thmt@contentsline[4]{\Contentsline{}{\parbox[t]{1.45em}{\strut}\parbox[t]{\linewidth-2.45em}{\RaggedRight\hyperlink{#4}{\em#2\nobreak\hspace{\pagespace}#3}}\\[2pt]}{}{}}
\makeatother
\def\thmtformatoptarg#1{\hspace{0.5em}\em#1}

\ignoretheorems{conjSec,question,questionSec,definition}
\boolfalse{definitions}
\booltrue{theorems}
\boolfalse{conjectures}
\listoftheorems[swapnumber]

\booltrue{definitions}
\boolfalse{theorems}
\boolfalse{conjectures}
\renewcommand\listtheoremname{List of Definitions}
\listoftheorems[ignoreall,show={definition},swapnumber]

\boolfalse{definitions}
\boolfalse{theorems}
\booltrue{conjectures}
\renewcommand\listtheoremname{List of Conjectures}
\listoftheorems[ignoreall,show={conjSec,questionSec,question},swapnumber]

\endgroup
\clearpage\phantomsection
\addcontentsline{toc}{chapter}{References}
\bibliographystyle{apalike2}
\bibliography{AlexanderMattTurnerDissertation}
\end{document}